\documentclass[letterpaper]{article} 
\usepackage{aaai25}  
\usepackage{times}  
\usepackage{helvet}  
\usepackage{courier}  
\usepackage[hyphens]{url}  
\usepackage{graphicx} 
\usepackage{natbib}  
\usepackage{caption} 
\usepackage{algorithm}
\usepackage{algorithmic}

\usepackage{newfloat}
\usepackage{listings}
\DeclareCaptionStyle{ruled}{labelfont=normalfont,labelsep=colon,strut=off} 
\floatstyle{ruled}
\newfloat{listing}{tb}{lst}{}
\floatname{listing}{Listing}
\title{Knowledge in Superposition: Unveiling the Failures of Lifelong Knowledge Editing for Large Language Models}
\author{
    Chenhui Hu\textsuperscript{\rm 1,\rm 2},
    Pengfei Cao\textsuperscript{\rm 1,\rm 2},
    Yubo Chen\textsuperscript{\rm 1,\rm 2},
    Kang Liu\textsuperscript{\rm 1,\rm 2},
    Jun Zhao\textsuperscript{\rm 1,\rm 2}
}
\affiliations{
    \textsuperscript{\rm 1} The Laboratory of Cognition and Decision Intelligence for Complex Systems, \\
    Institute of Automation, Chinese Academy of Sciences, Beijing, China\\
    \textsuperscript{\rm 2} School of Artificial Intelligence, University of Chinese Academy of Sciences, Beijing, China\\

    huchenhui2024@ia.ac.cn, \{pengfei.cao, yubo.chen, kliu, jzhao\}@nlpr.ia.ac.cn
}

\usepackage{bibentry}

\usepackage{amsmath}
\usepackage{amssymb}
\usepackage{graphicx}
\usepackage{subfigure}
\usepackage{subcaption}
\usepackage{caption}
\usepackage{booktabs}

\usepackage{alphalph}

\usepackage{morefloats}
\newtheorem{definition}{Definition}

\begin{document}

\maketitle

\begin{abstract}
Knowledge editing aims to update outdated or incorrect knowledge in large language models (LLMs). However, current knowledge editing methods have limited scalability for lifelong editing\footnote{Lifelong editing means lifelong knowledge editing.}. This study explores the fundamental reason why knowledge editing fails in lifelong editing. We begin with the closed-form solution derived from linear associative memory, which underpins state-of-the-art knowledge editing methods. We extend the solution from single editing to lifelong editing, and through rigorous mathematical derivation, identify an interference term in the final solution, suggesting that editing knowledge may impact irrelevant knowledge. Further analysis of the interference term reveals a close relationship with superposition between knowledge representations. When knowledge superposition does not exist in language models, the interference term vanishes, allowing for lossless knowledge editing. Experiments across numerous language models reveal that knowledge superposition is universal, exhibiting high kurtosis, zero mean, and heavy-tailed distributions with clear scaling laws. \textbf{Ultimately, by combining theory and experiments, we demonstrate that knowledge superposition is the fundamental reason for the failure of lifelong editing.} Moreover, this is the first study to investigate knowledge editing from the perspective of superposition and provides a comprehensive observation of superposition across numerous real-world language models. Code available at https://github.com/ChenhuiHu/knowledge\_in\_superposition.
\end{abstract}

%

\section{Introduction}\label{section:Introduction}

In large language models (LLMs), outdated or incorrect knowledge may persist (\citealp{radford2019language}; \citealp{wang2022gpt}; \citealp{biderman2023pythia}; \citealp{touvron2023llama}). However, retraining these models to update knowledge incurs prohibitively high costs. To address this problem, knowledge editing (\citealp{de2021editing}; \citealp{mitchell2021fast}) is introduced to edit specific knowledge by directly updating the internal parameters of language models.

\begin{figure}
    \centering
    \includegraphics[width=0.48\textwidth]{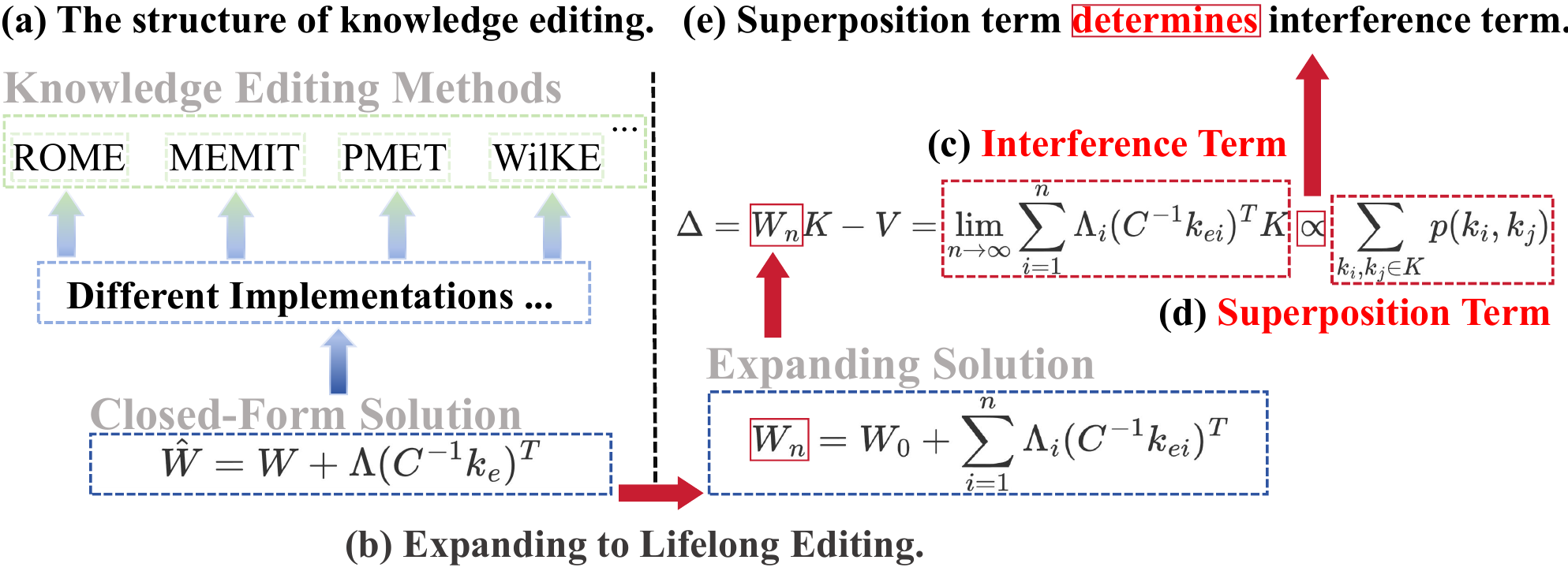}
    \caption{Illustration of our work. (a) Current knowledge editing methods use a unified closed-form solution, which means adding \(\Lambda(C^{-1}k_e)^T\) to parameters matrix \(W\) to achieve knowledge updating. (b) Extend the closed-form solution to lifelong editing, where \(W_n\) represents parameters matrix after the \(n\)-th edit. (c) Interference term accumulating sufficiently will cause language models to forget knowledge. (d) Superposition term, where \(p(\cdot,\cdot)\) denotes the degree of superposition between two knowledge representations. (e) Superposition term actually determines interference term.}
  \label{fig:closed-form solution}
\end{figure}

Despite achieving significant progress in single editing, where knowledge is edited only once, knowledge editing should be continuous in fact. However, current methods struggle with scalability for lifelong editing \cite{huang2023transformer}, where continuous editing and performance monitoring are required throughout the language models' lifecycle. For example, using the representative editing methods ROME \cite{meng2022locating} or MEMIT \cite{meng2022mass} for lifelong editing results in severe performance degradation after dozens or hundreds of editing steps, rendering the edited models unusable \cite{hu2024wilke}. However, the underlying reason has been rarely explored.

In this paper, we explore the fundamental reason why knowledge editing fails in lifelong editing. We start from the closed-form solution for knowledge editing derived by \cite{meng2022locating} through linear associative memory, which is the foundation of the state-of-the-art knowledge editing methods (Figure~\ref{fig:closed-form solution}a), including ROME \cite{meng2022locating}, MEMIT \cite{meng2022mass}, PMET \cite{li2023pmet}, WilKE \cite{hu2024wilke} and so on. Specifically, we extend the closed-form solution from single editing to lifelong editing (Figure~\ref{fig:closed-form solution}b). Our rigorous mathematical derivation reveals that the extended solution introduces an \textbf{interference term} (Figure~\ref{fig:closed-form solution}c), which, when accumulated sufficiently, leads language models to forgetting the knowledge, including original knowledge and previously edited knowledge.

\begin{figure}
    \centering
    \includegraphics[width=0.48\textwidth]{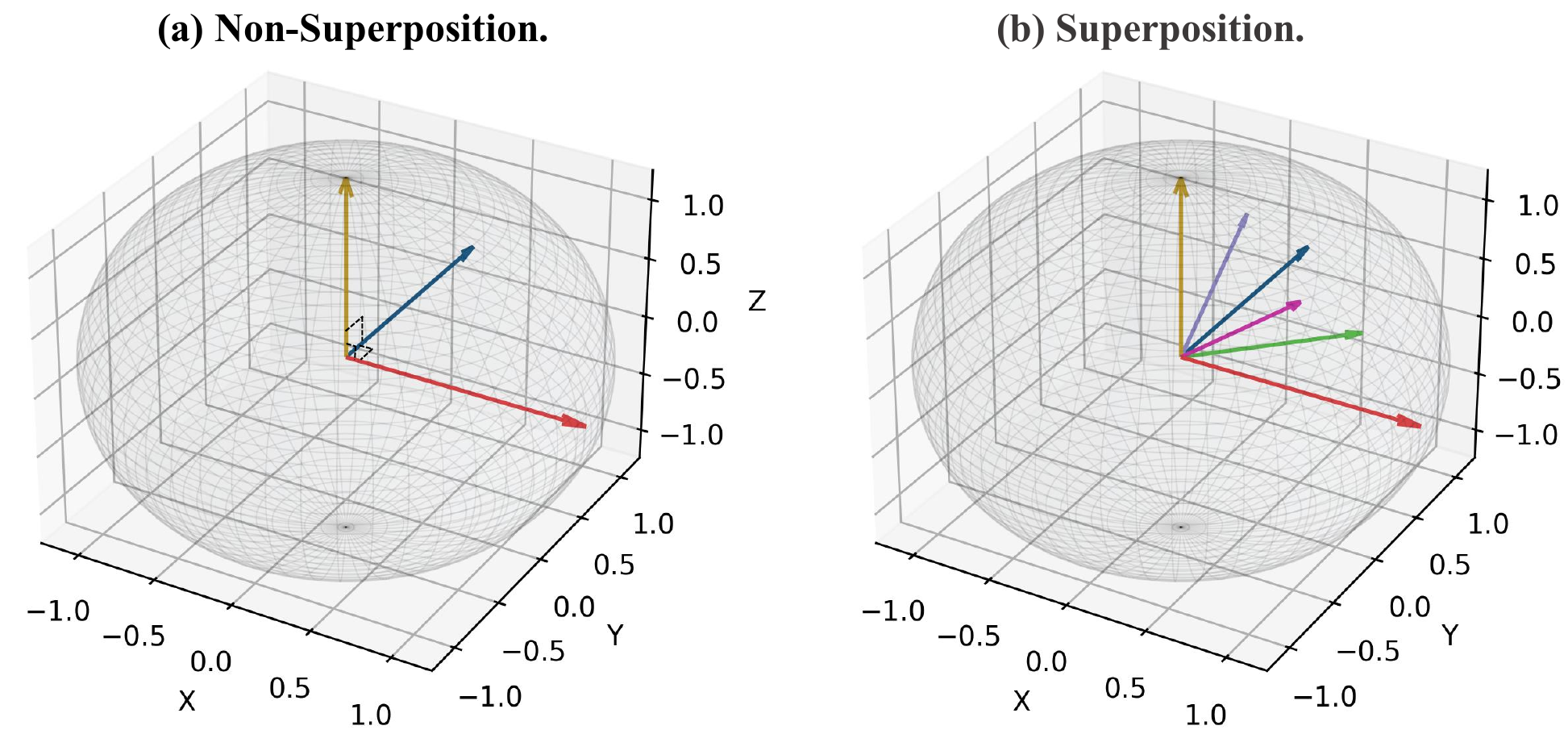}
    \caption{A neural network with only three neurons, corresponding to three dimensions, (a) can directly represent three features orthogonally, but (b) to represent six features (or more features), it will using superposition to noisily encode them nearly orthogonally.}
  \label{fig:superposition_in_ball}
\end{figure}

Further analysis of the interference term reveals a close relationship with superposition: if superposition does not exist in language models, the superposition term (Figure~\ref{fig:closed-form solution}d) is zero, causing the interference term to vanish (Figure~\ref{fig:closed-form solution}e), allowing for lossless lifelong editing. Superposition \cite{elhage2022toy} refers to the situation where neural networks attempt to represent more features than the available dimensions. For example, if a simple neural network with three neurons (corresponding to a three-dimensional space) attempts to represent three features, each neuron can be assigned to one feature, forming an orthogonal basis in three-dimensional space (Figure~\ref{fig:superposition_in_ball}a). But if the same network tries to represent six features (or more features), it will use superposition strategy, noisily encoding these features, where the directions corresponding to each feature will not be fully orthogonal (Figure~\ref{fig:superposition_in_ball}b). Specifically, we find that the magnitude of interference term depends on the orthogonality of knowledge representations (Figure \ref{fig:closed-form solution}e). If these knowledge representations are perfectly orthogonal (\textbf{non-superposition}), the interference term is zero, enabling perfectly lossless knowledge editing, where only the target knowledge is updated without affecting unrelated knowledge. However, if they are not orthogonal (\textbf{superposition}), the interference will accumulate linearly and eventually tend toward infinity, leading to language models' failure (Appendix H). \textbf{In other words, whether superposition exists in language models is equivalent to whether lossless knowledge editing can be achieved, which will determine whether lifelong editing can be achieved}.

However, to what extent does superposition hold true in real-world language models? \textbf{Our experiments reveal that knowledge superposition\footnote{Knowledge superposition refers to the superposition of knowledge representations in language models.} is universal across all language model families, characterized by high kurtosis, zero mean, heavy-tailed distributions, with a clear scaling law}. 

Specifically, we conduct experiments on language model families including GPT-2 \cite{radford2019language}, Llama-2 \cite{touvron2023llama}, Pythia \cite{biderman2023pythia}, GPT-J \cite{wang2022gpt}, Llama-3 \cite{meta2024introducing}, and Llama-3.1 \cite{dubey2024llama3herdmodels}, to capture the real situation of knowledge superposition. Firstly, we observe that knowledge superposition exists across all layers of all language models we examined (Figure~\ref{fig:visualized_superposition}), indicating that superposition is widespread in real-world language models. Secondly, the knowledge superposition exhibits a high kurtosis, zero-mean heavy-tailed distribution (Figure~\ref{fig:kde_superposition}). This means that the distribution consistently has a very high peak near zero, which corresponds to the positions where all knowledge representations are perfectly orthogonal, suggesting models attempt to store different knowledge orthogonally but resort to stores them nearly orthogonally due to capacity constraints, i.e., through superposition. Thirdly, we observe a clear scaling law (Figure~\ref{fig:scaling_law_superposition}): as the size of language models increases, they attempt to represent knowledge in a more orthogonal manner, thereby reducing noise between knowledge representations, which explains why larger language models can more effectively handle knowledge-intensive question-answering scenarios. Finally, to provide further direct evidence of superposition, we calculate the angular distribution between these knowledge representations in high-dimensional space, which also shows a high kurtosis, heavy-tailed distribution, but centered around 90 degrees (Figure~\ref{fig:whitening_vs_activation}).

Overall, our contributions are as follows:

\begin{itemize}
    \item We extend the closed-form solution for knowledge editing from single editing to lifelong editing. Our rigorous derivation reveals an interference term in the final solution, indicating that editing can affect both original and previously edited knowledge. 
    \item We conduct further analysis of the interference term, which actually reflects knowledge superposition. The magnitude of this interference is determined by the superposition of knowledge representations, with non-superposition enabling lossless knowledge editing.
    \item We investigate the phenomenon of knowledge superposition across multiple language model families and find it universal, exhibiting high kurtosis, zero mean, heavy-tailed distributions, with a clear scaling law. To our knowledge, this is the first observation of widespread knowledge superposition. 
\end{itemize}

\section{Related Work}

\subsection{Knowledge Editing}

In general, knowledge editing aims to modify the knowledge of a language model so that its outputs reflect the revised state when confronted with relevant inputs \cite{de2021editing}. \citet{yao2023editing} survey knowledge editing methods and classify them into two main categories: preserving model parameters and modifying model parameters. Preserving model parameters includes memory-based methods (\citealp{mitchell2022memory}; \citealp{zhong2023mquake}; \citealp{hartvigsen2024aging}), which stores all edit examples explicitly in memory, and additional parameter methods (\citealp{dong2022calibrating}; \citealp{huang2023transformer}), which use extra trainable parameters within language models. However, these methods usually have poor generalization, and the required additional content also increases significantly with the number of editing. In contrast, modifying model parameters methods directly update the model parameters for editing, thereby avoiding these issues. Such methods are also divided into two categories: meta-learning and locate-then-edit methods. Meta-learning methods (\citealp{de2021editing}; \citealp{mitchell2021fast}; \citealp{tan2023massive}), which use a hyper network to learn and apply gradients for fine-tuning, typically require a lengthy hyper network training process before each edit, making them less effective for lifelong editing. Locate-then-edit methods first identify and then update the parameters associated with specific knowledge. For instance, KnowledgeNeuron \cite{dai2021knowledge} uses knowledge attribution to locate neurons and updates parameters to achieve knowledge editing; ROME \cite{meng2022locating} employs causal mediation analysis to identify the center of causal effects, namely MLP, and performs rank-one editing on MLP; MEMIT \cite{meng2022mass} extends ROME by allocating residuals to multiple layers and enabling batch editing; PMET \cite{li2023pmet} further refines residual allocation based on MEMIT; and WilKE \cite{hu2024wilke} dynamically selects editing layers based on ROME to avoid toxicity flash.

This paper focuses on methods that modify model parameters, among which the state-of-the-art methods are ROME and its derivatives, MEMIT, PMET, and WilKE. Therefore, we concentrate on the same closed-form solution (Figure~\ref{fig:closed-form solution}a) of them and extend it from single to lifelong editing. From a theoretical perspective, we identify the fundamental reason for the failure of knowledge editing methods in lifelong editing, namely superposition.

\subsection{Superposition}

Superposition refers to a strategy used by neural networks to express features that far exceed their dimensions, typically assigning approximately orthogonal directions in the representation space to these features (Figure~\ref{fig:superposition_in_ball}b). However, this superposition phenomenon has only been observed in toy models so far \cite{elhage2022toy}. Some studies hypothesize that this superposition phenomenon also exists in language models and propose sparse autoencoder (SAE) \cite{cunningham2023sparse} and its various variants (\citealp{rajamanoharan2024improving}; \citealp{gao2024scaling}) to attempt to disentangle superposition from the activation space of language models (\citealp{bricken2023monosemanticity}; \citealp{templeton2024scaling}). Additionally, \citet{gurnee2023finding} claim to observe superposition in the wild, their study was limited to a few neurons in a specific layer of a single model, focusing on neuron superposition. This differs from the superposition concept in \citet{elhage2022toy} and our study.

In this paper, we identify a widespread phenomenon of knowledge superposition across multiple language model families and explore its characteristics. Furthermore, this is the first study of knowledge superposition in knowledge editing, explaining the reason for failures in lifelong editing from the perspective of superposition. Notably, unlike previous studies, we find that superposition occurs in the whitening space rather than the activation space (Figure~\ref{fig:whitening_vs_activation}).

\section{Preliminary}\label{section:Preliminary}

\citet{geva2020transformer} discover that MLPs are the key components for memory storage in Transformers, and \citet{meng2022locating} identify through causal tracing that MLPs are crucial for storing factual knowledge. The MLPs in the Feed-Forward Network (FFN) of a Transformer consist of two layers, represented as follows (bias terms are omitted)
\begin{equation}
    FFN(\pmb{x})=W_{proj}\ \sigma(W_{fc}\ \pmb{x}),
\end{equation}
where \( W_{fc} \in \mathbb{R}^{d_m \times d} \) and \( W_{proj} \in \mathbb{R}^{d \times d_m} \) are the parameter matrices of the FFN, \( d_m \) is the dimensionality of the hidden layer in the FFN, \( \sigma \) is the activation function, and \( \pmb{x} \in \mathbb{R}^d \) is the input of the FFN.

\citet{meng2022locating} model MLPs as linear associative memory (\citealp{kohonen1972correlation}; \citealp{anderson1972simple}), viewing \( W_{proj} \) as a linear associative store. This perspective observes that any linear operation \( W \) can be represented as a key-value store with a set of vector keys \( K = [k_1 | k_2 | \cdots ] \) and corresponding vector values \( V = [v_1 | v_2 | \cdots ] \). To optimally insert a new key-value pair \((k_e, v_e)\) into the store for knowledge updates, one can solve a constrained least-squares problem, leading to a closed-form solution
\begin{equation}\label{equ:closed-form_solution}
    \begin{gathered}
    \mathop{minimize}\|\hat{W}K-V\|\text{ such that } \hat{W}k_e=v_e\\ 
    \text{ by setting }\hat{W}=W+\Lambda(C^{-1}k_e)^T,
    \end{gathered}
\end{equation}
where \( C = KK^T \) is a constant that actually represents the covariance matrix, and \( \Lambda = (v_e - Wk_e)/(C^{-1}k_e)^T k_e \). The detailed proof can be found in the original paper \cite{meng2022locating}, and a more thorough derivation is provided in Appendix~\ref{appendix:A. Proof in Preliminary}A.

\section{Expanding to Lifelong Editing}\label{section:Expanding to Lifelong Editing}

For convenience, let the initial \( W_{proj} \) be \( W_0 \). Following the first update using \((k_{e_1}, v_{e_1})\), we obtain \( W_1 \). Subsequently, after the \( n \)-th edit with \((k_{e_n}, v_{e_n})\), we derive \( W_n \). Extending the closed-form solution from Equation~\ref{equ:closed-form_solution} to the lifelong editing scenario, we obtain
\vspace{-1ex}
\begin{equation}
    \begin{gathered}
        W_n=\begin{cases}
        W_0,&n=0\\
        W_{n-1}+\Lambda_{n}(C^{-1}k_{e_n})^T,&n\ge1
        \end{cases}
    \end{gathered}
\end{equation}
where \( C = KK^T \), and
\begin{equation}
    \begin{gathered}
        \Lambda_n=\frac{v_{e_n}-W_{n-1}k_{e_n}}{(C^{-1}k_{e_n})^Tk_{e_n}},n\ge1.
    \end{gathered}
\end{equation}
Expanding \( W_n \), we obtain
\vspace{-1ex}
\begin{equation}
    \begin{gathered}
        W_n = W_0+ \sum_{i=1}^n\Lambda_i(C^{-1}k_{e_i})^T.
    \end{gathered}
\end{equation}

For convenience, we let
\vspace{-1ex}
\begin{equation}
    K^*=
    \left[
    \begin{array}{ccc|ccc}
    \\
    & K_o^* & &  & K_e^* &\\
    \\
    \end{array}
    \right],
\end{equation}
where \( K_o^*=[k_1 | k_2 | \cdots] \) represents the keys corresponding to the model's original knowledge, and \( K_e^*= [k_{e_1} | k_{e_2} | \cdots | k_{e_n}]\) represents the keys for the edited knowledge.

We then calculate \( V^* \) after \( W_0 \) has been updated to \( W_n \), which is given by

\begin{align}
    V^* &= W_nK^*\notag\\
    &= (W_0+ \sum_{i=1}^n\Lambda_i(C^{-1}k_{e_i})^T)K^*\notag\\
    &= W_0K^*+\left(\sum_{i=1}^n\Lambda_i(C^{-1}k_{e_i})^T\right)K^*\notag\\
    &= W_0K^*+\sum_{i=1}^n\Lambda_i(C^{-1} k_{e_i})^TK^*.
\end{align}

For convenience, we can represent \( V^* \) using a block matrix notation
\begin{equation}
    V^*=
    \left[
    \begin{array}{ccc|ccc}
    \\
    & V_o^* & &  & V_e^* &\\
    \\
    \end{array}
    \right],
\end{equation}
where \( V_o^* \) and \( V_e^* \) represent the updated value matrices corresponding to the original knowledge and the edited knowledge, respectively. We then analyze these two components separately.

\subsection{Interference Term of Original Knowledge \(\Delta_o\)}

Ideally, the value vectors corresponding to the original knowledge after editing, namely \(V_o^*\), are same as \( V_o = [v_1 | v_2 | \cdots] \), because we want to ensure that editing does not affect unrelated original knowledge. Thus, we define \( \Delta_o \) as
\begin{align}
    \Delta_o&=V^*_o-V_o\notag\\
    &=\left[\sum_{i=1}^n\Lambda_i(C^{-1}k_{e_i})^Tk_1|\sum_{i=1}^n\Lambda_i(C^{-1}k_{e_i})^Tk_2|\cdots\right]\!.
\end{align}

For convenience, we study \( \Delta_o[:, j] \), which represents the \( j \)-th column vector of \( \Delta_o \). This can be viewed as the interference term introduced to the \( j \)-th piece of original knowledge after \( n \) edits,
\begin{align}
    \Delta_{o}[:,j]&= \sum_{i=1}^n\Lambda_i(C^{-1}k_{e_i})^Tk_j\notag\\
    &= \sum_{i=1}^n\frac{v_{e_i}-W_{i-1}k_{e_i}}{(C^{-1}k_{e_i})^Tk_{e_i}}(C^{-1}k_{e_i})^Tk_j\notag\\
    &= \sum_{i=1}^n\frac{(C^{-1}k_{e_i})^Tk_j}{(C^{-1}k_{e_i})^Tk_{e_i}}(v_{e_i}-W_{i-1}k_{e_i}).
\end{align}

Next, let \(\delta(v_{e_i}) = v_{e_i} - W_{i-1} k_{e_i} \in \mathbb{R}^d\) represent the difference vector between the optimized value and the current value when editing the \(i\)-th piece of knowledge, and define the coefficient \( p(k_{e_i}, k_j) = \frac{(C^{-1}k_{e_i})^T k_j}{(C^{-1}k_{e_i})^T k_{e_i}} \in \mathbb{R} \). Then, we obtain
\begin{equation}
    \Delta_o[:,j]=\sum_{i=1}^np(k_{e_i},k_{j})\delta(v_{e_i}).
\end{equation}

Ideally, we want \(\Delta_o[:, j] = \pmb{0}\) to achieve lossless knowledge editing, ensuring that the original knowledge of the model remains unaffected. However, since \(\delta(v_{e_i}) = v_{e_i} - W_{i-1}k_{e_i} \neq \pmb{0}\) (as this is the premise; if \(\delta(v_{e_i}) = \pmb{0}\), then there is no need for editing), we can only hope that \(p(k_{e_i}, k_j) = 0\). This would eliminate the interference on the \(j\)-th original knowledge.

\subsection{Interference Term of Edited Knowledge \(\Delta_e\)}

Ideally, the value vectors corresponding to the edited knowledge after editing, \(V_e^*\), are same as \( V_e = [v_{e_1} | v_{e_2} | \cdots | v_{e_n}] \), because we want to avoid affecting the already edited knowledge during further editing. Thus, we define \( \Delta_e \) as
\begin{align}
    \Delta_e &= V_e^*-V_e\notag\\
    &\!\!= \left[\sum_{i=2}^n\Lambda_i(C^{-1}k_{e_i})^Tk_{e_1}|\cdots|\Lambda_n(C^{-1}k_{e_n})^Tk_{e_{n-1}}|\pmb 0\right]\!.
\end{align}

Unlike the case with original knowledge, as editing progresses, earlier edits experience more interference, while later edits are less affected. This is consistent with \citet{jang2021towards}. For convenience, we study \( \Delta_e[:, j] \), which represents the \( j \)-th column vector of \( \Delta_e \). This can be seen as the impact introduced to the \( j \)-th edited knowledge after \( n \) edits,
\begin{equation}
    \begin{gathered}
        \Delta_e[:,j]=
        \begin{cases}
        \sum_{i=j+1}^n\Lambda_i(C^{-1}k_{e_i})^Tk_{e_j},&j< n\\
        \pmb 0,&j=n
        \end{cases}
    \end{gathered}
\end{equation}
In a similar manner, we can express it as
\begin{equation}
    \begin{gathered}
        \Delta_e[:,j]=
        \begin{cases}
        \sum_{i=j+1}^n\!\frac{(C^{-1}k_{e_i})^Tk_{e_j}}{(C^{-1}k_{e_i})^Tk_{e_i}}(v_{e_i}-W_{i-1}k_{e_i}),\!\!\!\!\!&j< n\\
        \pmb0,\!\!\!\!\!&j=n
        \end{cases}
    \end{gathered}
\end{equation}

Next, let \(\delta(v_{e_i}) = v_{e_i} - W_{i-1} k_{e_i} \in \mathbb{R}^d\) represent the difference vector between the optimized value and the current value when editing the \(i\)-th piece of knowledge. Define the coefficient \( p(k_{e_i}, k_{e_j}) = \frac{(C^{-1}k_{e_i})^T k_{e_j}}{(C^{-1}k_{e_i})^T k_{e_i}} \in \mathbb{R} \). Then, we obtain

\begin{equation}
    \begin{gathered}
        \Delta_e[:,j]=
        \begin{cases}
        \sum_{i=j+1}^n p(k_{e_i},k_{e_j})\delta(v_{e_i}),&j< n\\
        \pmb0,& j=n
        \end{cases}
    \end{gathered}
\end{equation}

Ideally, we want \(\Delta_e[:, j] = \pmb{0}\) to achieve lossless knowledge editing, ensuring that previously edited knowledge is not affected. However, since \(\delta(v_{e_i}) = v_{e_i} - W_{i-1}k_{e_i} \neq \pmb{0}\), we can only hope that \(p(k_{e_i}, k_{e_j}) = 0\). This would eliminate the interference on the \(j\)-th edited knowledge.

\subsection{How to Understand \(p(\cdot,\cdot)\)}

In summary, we find that for both the original knowledge and the edited knowledge, there is a similar coefficient \( p(\cdot, \cdot) \) in interference term, which directly determines whether lossless knowledge editing can be achieved.

To study this coefficient more generally, we do not limit to specific keys (which correspond to specific knowledge activations) but consider any keys (or any knowledge activations) \(k_i\) and \(k_j\). We further investigate 
\begin{equation}
    \begin{gathered}
        p(k_i, k_j) = \frac{(C^{-1}k_i)^T k_j}{(C^{-1}k_i)^T k_i} \in \mathbb{R}.
    \end{gathered}
\end{equation}
Since \(C = KK^T\) is a symmetric matrix, \(C^{-1}\) is also symmetric. Therefore, we can rewrite it as
\begin{align}
    p(k_i,k_j) &= \frac{k_{i}^TC^{-1}k_j}{k_{i}^TC^{-1}k_{i}}= \frac{(C^{-\frac12}k_{i})^T(C^{-\frac12}k_{j})}{(C^{-\frac12}k_{i})^T(C^{-\frac12}k_{i})}.
\end{align}

Here, \(C = KK^T\) is the covariance matrix, and both \(k_i\) and \(k_j\) are column vectors of \(K\). Therefore, \(C^{-\frac{1}{2}}k_i\) and \(C^{-\frac{1}{2}}k_j\) are the representations of \(k_i\) and \(k_j\) in the whitening space (\citealp{koivunen1999feasibility}; \citealp{kawahara2007face}) after the whitening transformation \(C^{-\frac{1}{2}}\) (proof in Appendix~\ref{appendix:B. Proof of Whitening Matrix}B). The coefficient \(p(\cdot, \cdot)\) can be understood as the dot product of \(k_i\) and \(k_j\) in the whitening space, normalized. When \(p(\cdot, \cdot) = 0\), it indicates that the knowledge activations \(k_i\) and \(k_j\) are orthogonal in the whitening space.

\begin{definition}[Matrix Whitening]
For a matrix \(X\), its covariance matrix \( \text{Cov}(X) = XX^T \) is not necessarily the identity matrix. Matrix whitening involves finding a transformation matrix \(P\) such that the covariance matrix of \(Y = PX\), denoted as \(\text{Cov}(Y)\), becomes the identity matrix.
\end{definition}

Returning to the problem, if we aim to achieve perfectly lossless knowledge editing, this is equivalent to requiring both \(\Delta_o\) and \(\Delta_e\) to be zero matrices. This in turn means we expect \(p(\cdot, \cdot)\) to be zero, which is equivalent to expecting that knowledge activations are orthogonal in the whitening space. Consequently, this implies that we expect the language model to store different pieces of knowledge in orthogonal directions in the whitening space, and thus that knowledge superposition does not exist in the whitening space. \textbf{Conversely, if such superposition exists, then the representations are not orthogonal and we cannot achieve perfectly lossless knowledge editing}.

\section{Knowledge in Superposition}

In this section, \textbf{we focus on the phenomenon of knowledge superposition in real-world language models, combining previous theoretical derivations to demonstrate that knowledge superposition is the fundamental reason for the failure in lifelong editing}.

Specifically, we calculate the degree of superposition between pieces of knowledge by computing the matrix \(P\) for \(m\) pieces of knowledge, where
\begin{equation}
    \begin{gathered}
        P[i,j]=p(k_i,k_j),1\le i,j\le m.
    \end{gathered}
\end{equation}
As \( p(\cdot,\cdot) \) approaches 0, knowledge representations become more orthogonal in the whitening space, and the degree of superposition becomes weaker. Conversely, as \( p(\cdot,\cdot) \) approaches 1, knowledge representations become more similar in the whitening space, and the degree of superposition becomes stronger.

In practice, we choose \( m = 128 \) to compute the superposition between \( 128 \times 128 \) pairs of knowledge and resulting in a \( 128 \times 128 \) matrix \(P\). Empirical evidence shows that \( m = 128 \) is sufficient because, at this point, the kurtosis of the superposition distribution has converged (details in Appendix~\ref{appendix:C. Number of Knowledge Items}C). This indicates that the data size is adequate for describing the distribution of superposition.

The specific experimental setup is described as follows:

\noindent\textbf{Models} In this study, we employ a variety of models including GPT2 family \cite{radford2019language} —GPT2-Small, GPT2-Medium, and GPT2-Large, Pythia family \cite{biderman2023pythia} —Pythia-1B, Pythia-2.8B, and Pythia-6.9B, Llama2 family \cite{touvron2023llama} -Llama2-7B and Llama2-13B. Additionally, we use the classic GPT-J \cite{wang2022gpt}, the latest Llama3-8B \cite{meta2024introducing} and Llama3.1-8B \cite{dubey2024llama3herdmodels}.

\noindent\textbf{Datasets} For the extraction of knowledge representations, we utilized the CounterFact dataset \cite{meng2022locating}. It is important to note that while CounterFact is commonly used for counterfactual knowledge editing, our experimental setup does not involve this aspect. Instead, we focus on subject-related knowledge rather than counterfactual objects (details in Appendix~\ref{appendix:G. Experimental Detail}G).

\begin{figure}
    \centering
    \subfigure[GPT2-Small]{\includegraphics[width=0.22\textwidth]{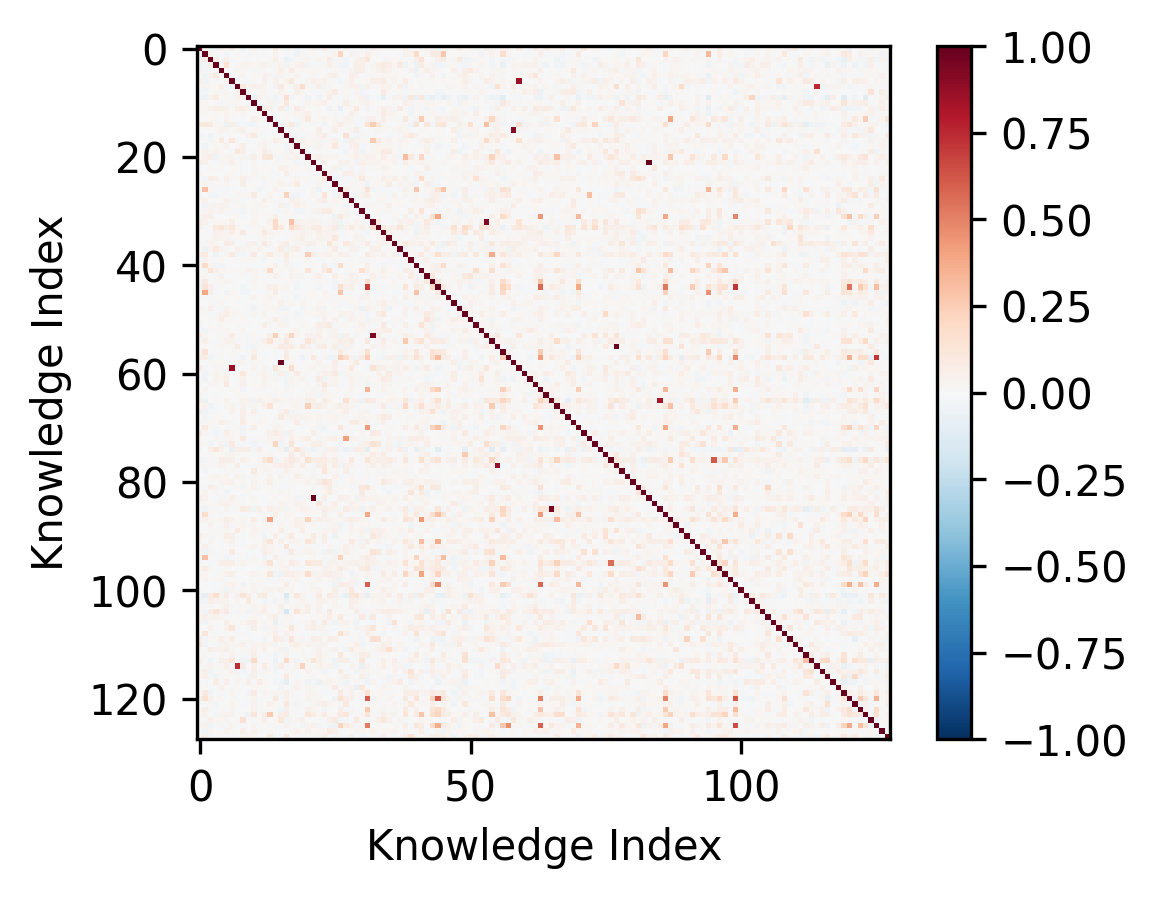}
    }
    \hfill
    \subfigure[GPT2-Medium]{\includegraphics[width=0.22\textwidth]{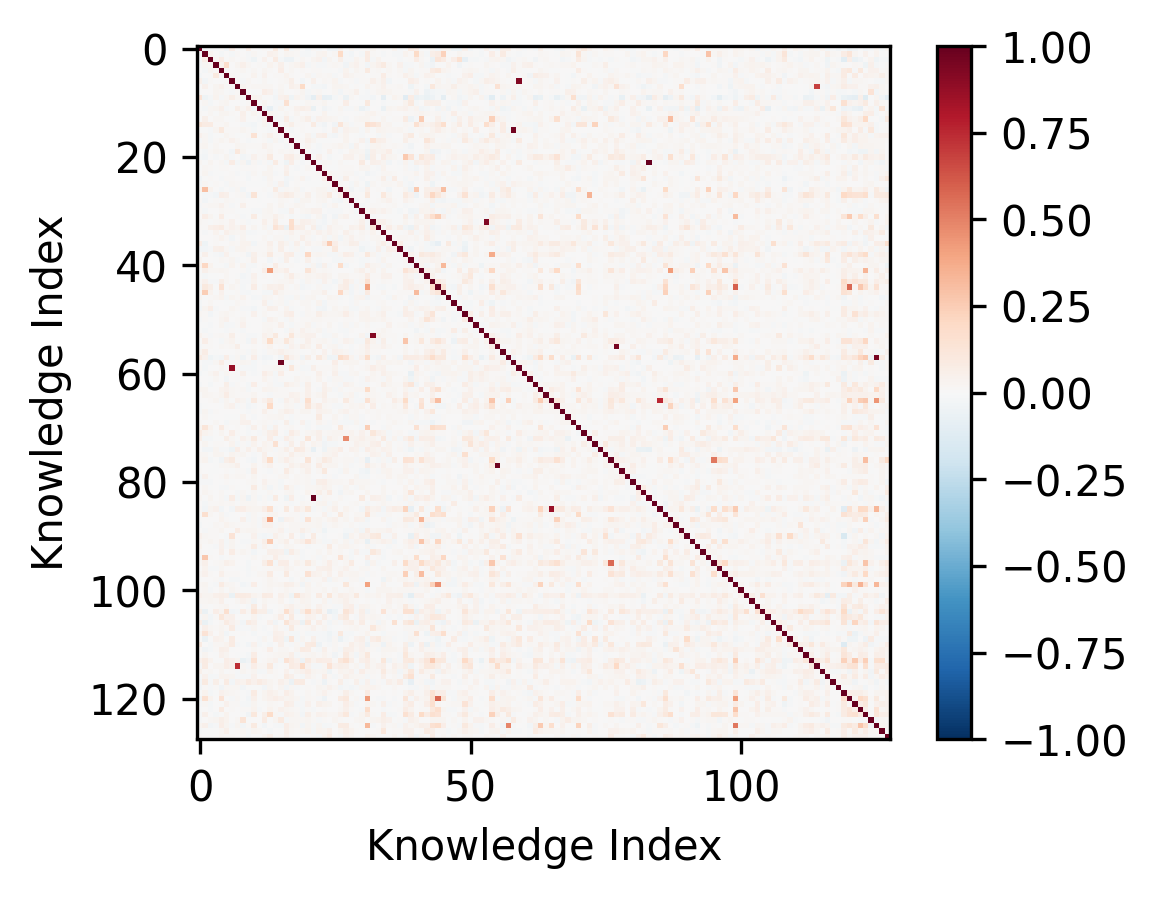}
    }
    \hfill
    \subfigure[GPT2-Large]{\includegraphics[width=0.22\textwidth]{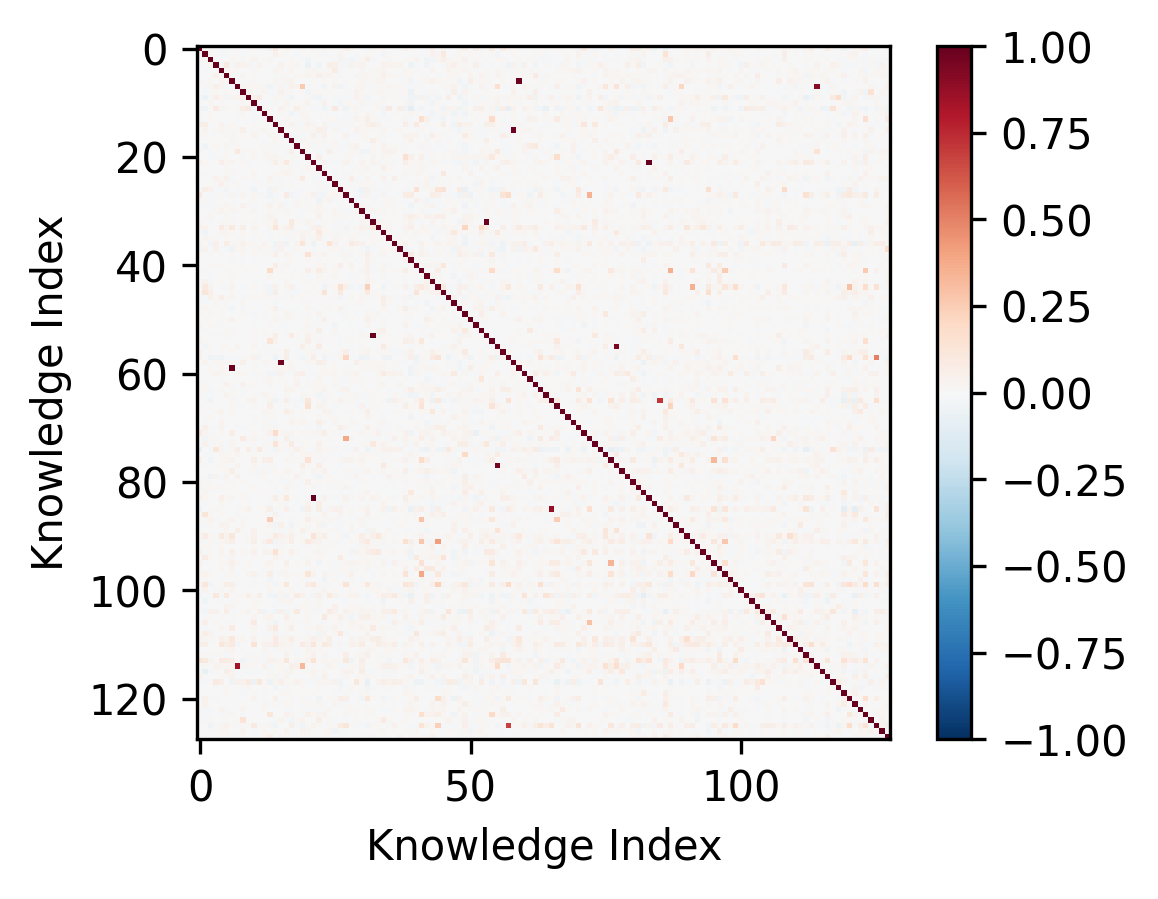}
    }
    \hfill
    \subfigure[GPT-J-6B]{\includegraphics[width=0.22\textwidth]{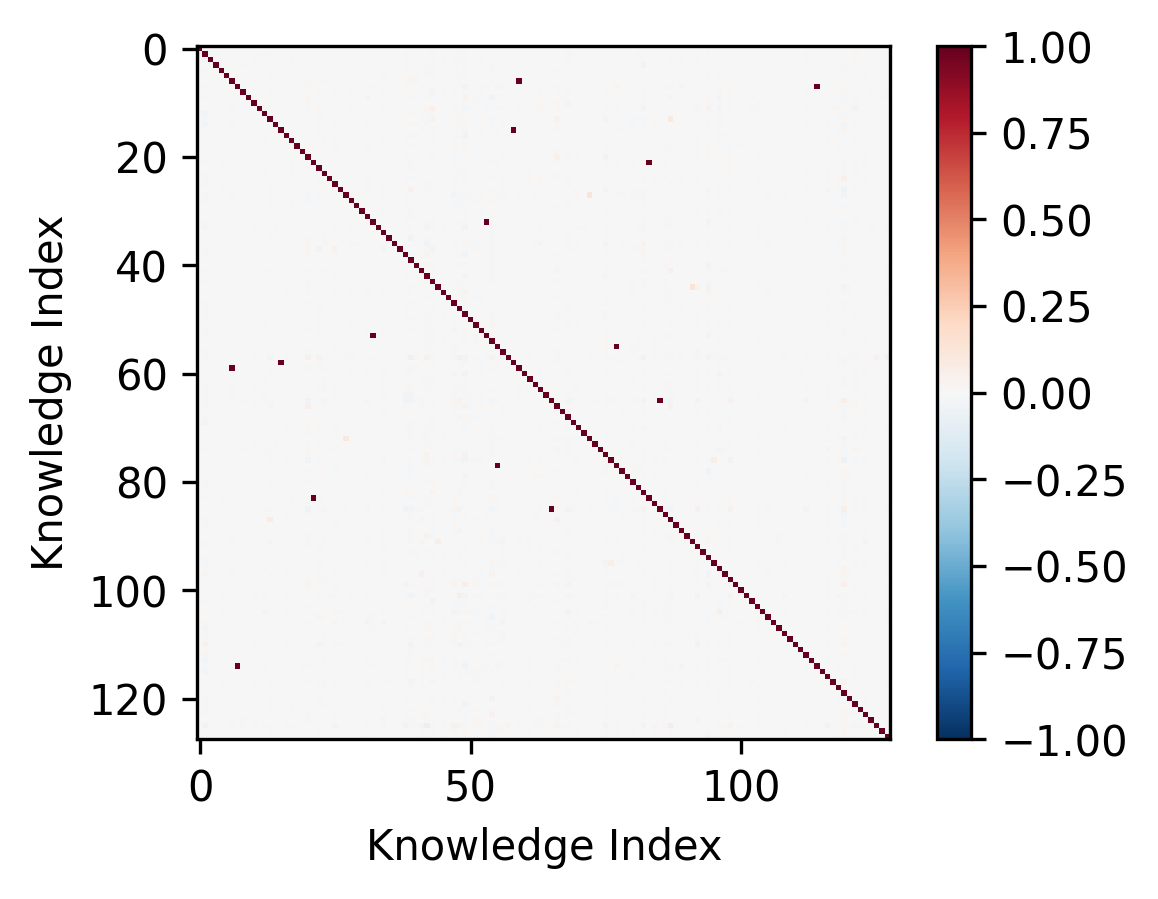}
    }
  \caption{Superposition at layer 0 across different language models visualized using P matrices, ordered by model size. Each point in these 128x128 \( P \) matrices corresponds to the \(p(\cdot,\cdot)\) value between two pieces of knowledge.}
  \label{fig:visualized_superposition}
\end{figure}

\subsection{Universal in Language Models}

Ideally, the matrix \(P\) we obtain should be such that all positions except the diagonal are zero, indicating that no superposition exists and allowing for lossless knowledge editing. However, in all layers of all the language models we studied, the \(P\) matrices we obtained consistently show noisy non-zero entries at positions other than the diagonal, indicating the presence of superposition at these points. As shown in Figure~\ref{fig:visualized_superposition}, we present the heatmaps of \(P\) matrices for GPT2-Small, GPT2-Medium, GPT2-Large, and the GPT-J at layer 0, ordered by model size. Additional heatmaps of \(P\) matrices for all layers of all models are provided in Appendix~\ref{appendix:D. Visualization of Matrix for All Layers}D.

It is evident that as the model size increases, the \(P\) matrices become progressively "cleaner." This indicates that as language models gain more storage capacity to store knowledge, they tend to store this knowledge with weaker superposition, thereby reducing the interference between pieces of knowledge caused by superposition.

Additionally, we observe that even as model size increases, fixed interference points remain at positions off the diagonal, even in models with different architectures and trained on different corpora. We hypothesize that the knowledge pairs corresponding to these points may actually be closely related, despite having different expressions, leading to this phenomenon. This point has been validated through detailed case studies. We find that the knowledge pairs corresponding to these points are indeed closely related. For instance, in the first layer (layer 0) of these four language models, the \(p(\cdot,\cdot)\) values for "Vladimir Mayakovsky" and "Vladimir Bukovsky" (both of whom are native Russian speakers) are above 0.95, and even 1.00 in GPT-J, indicating that the operations performed by the MLP in the first layer for these two pieces of knowledge are similar or even identical. This also implies that if we attempt to edit the knowledge of subject "Vladimir Mayakovsky" in this layer, it will have a consistent impact on "Vladimir Bukovsky," suggesting that they are bound together. For example, if we attempt to edit "Vladimir Mayakovsky" to have French as his native language, the edited model will also output French as "Vladimir Bukovsky's" native language. Similar examples include "Windows 8.1" and "Mac OS X 10.1," which generally have very high \(p(\cdot,\cdot)\) values (1.00 in the first layer of GPT-J), even though they are produced by entirely different manufacturers, which is fascinating! The case study is detailed in Appendix~\ref{appendix:F. Case Study}F.

\begin{figure*}
    \centering
    \subfigure[GPT2 Family]{\includegraphics[width=0.32\textwidth]{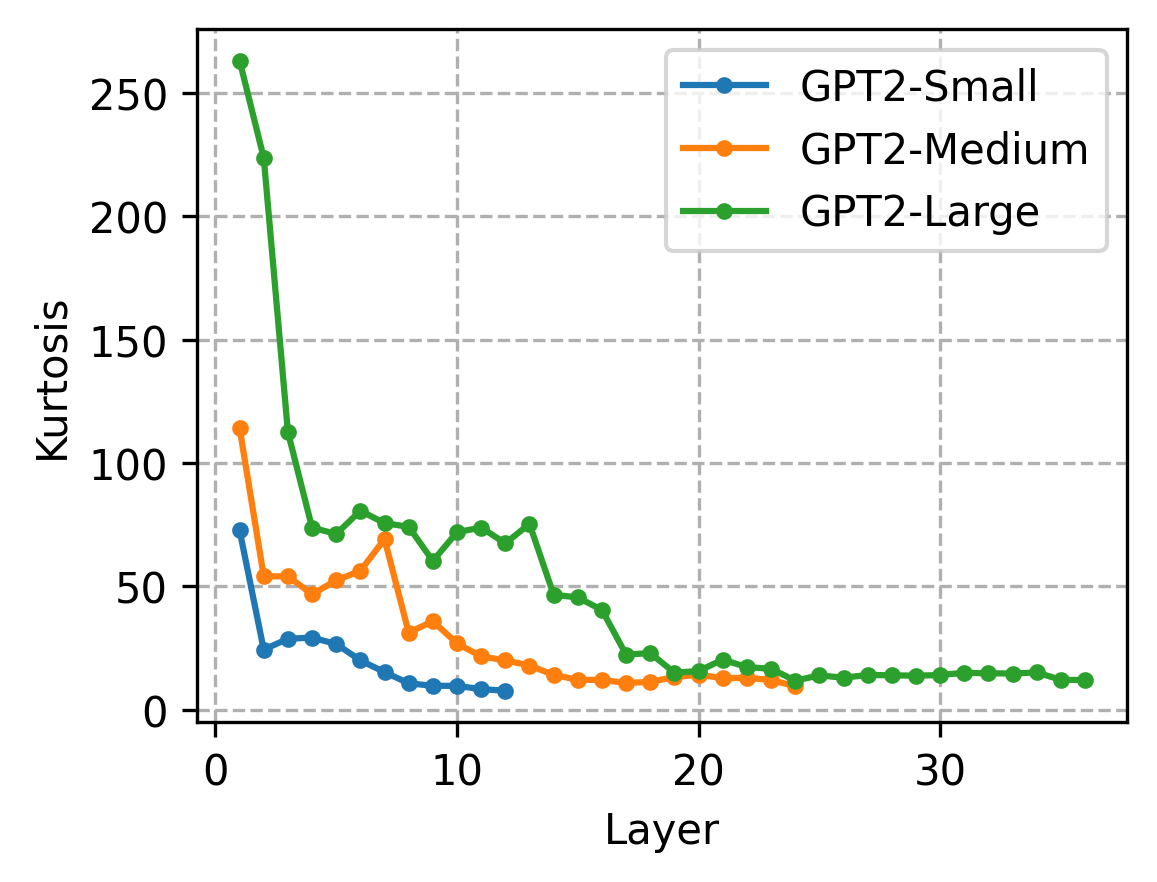}
    }
    \hfill
    \subfigure[Pythia Family]{\includegraphics[width=0.32\textwidth]{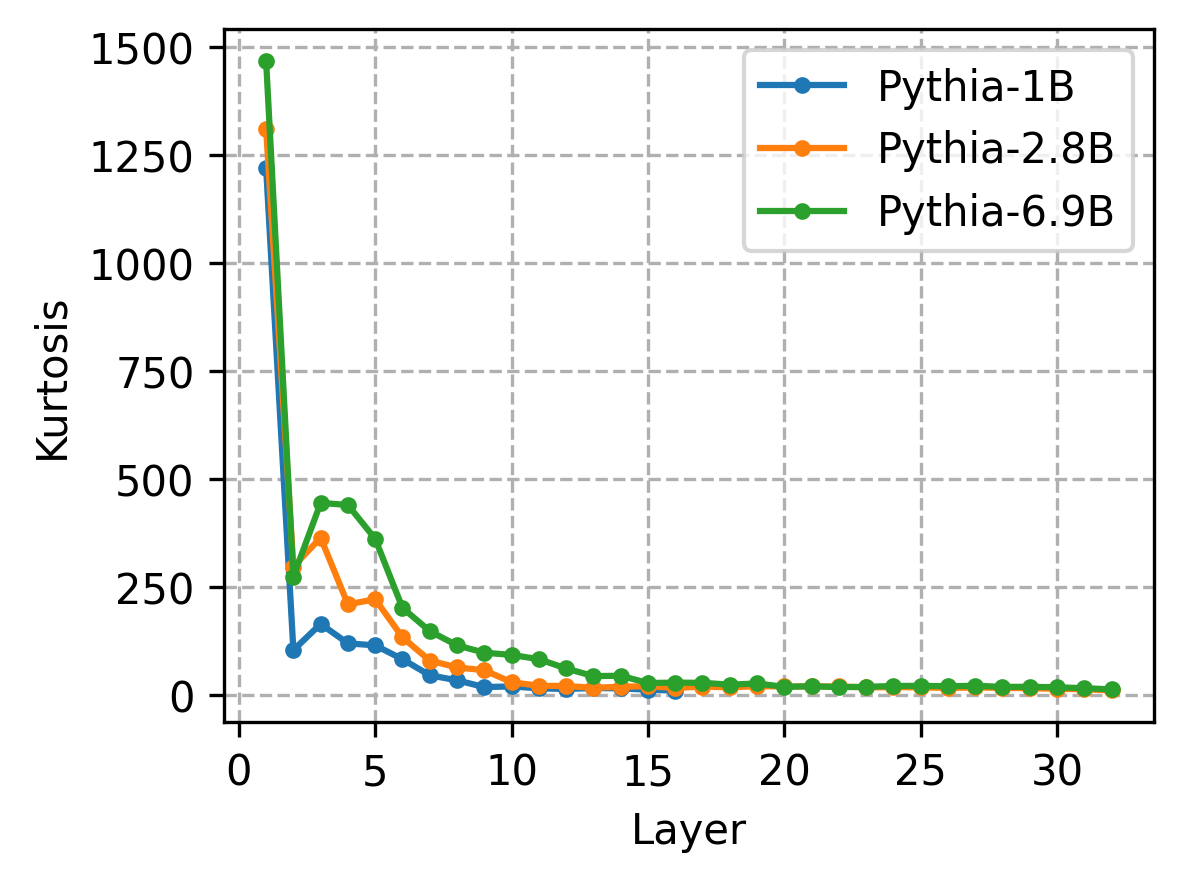}
    }
    \hfill
    \subfigure[Llama2 Family]{\includegraphics[width=0.32\textwidth]{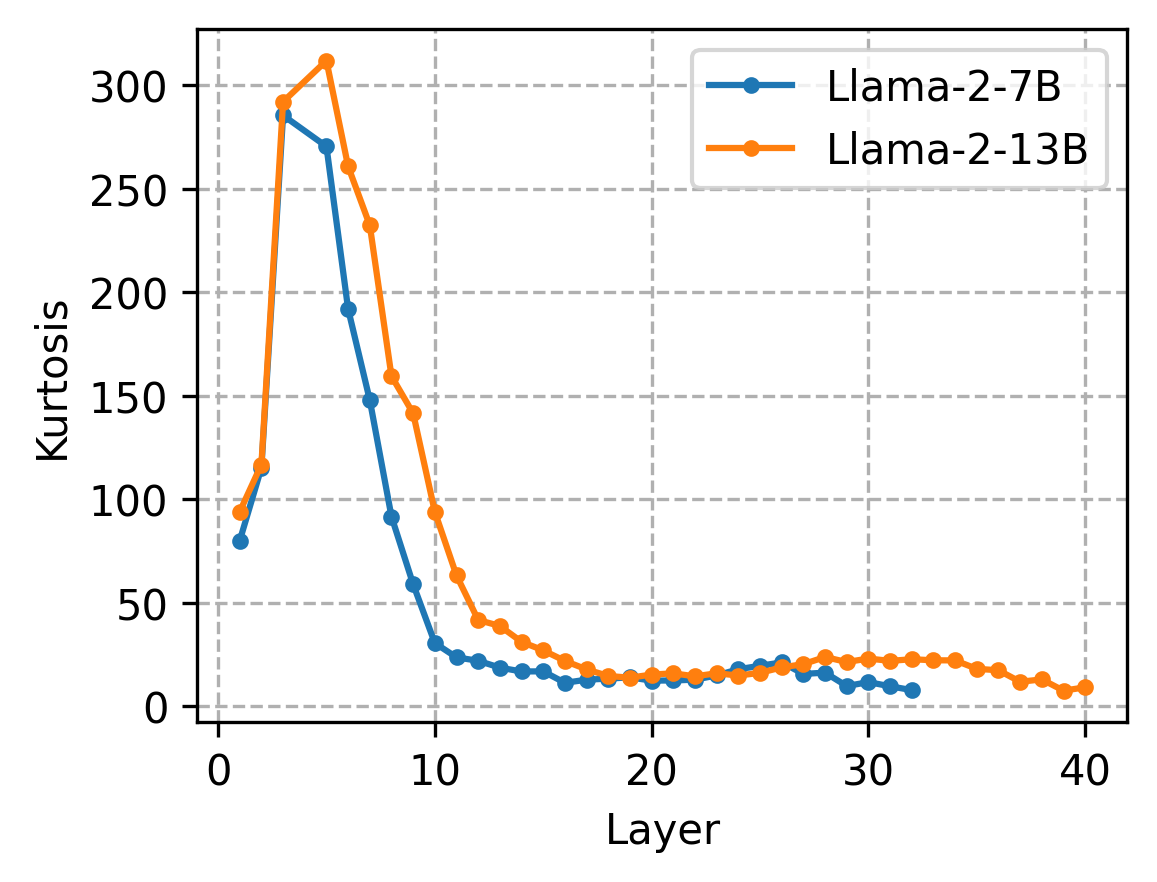}
    }
  \caption{The scaling law of knowledge superposition. Higher kurtosis means less superposition.}
  \label{fig:scaling_law_superposition}
\end{figure*}

Furthermore, choosing different editing layers may lead to certain changes, as described by \citet{hu2024wilke}. However, other layers also contain other knowledge in superposition, as shown in Figure~\ref{fig:scaling_law_superposition}.

\subsection{Heavy-Tailed Distribution in Language Models}

\begin{figure}
    \centering
    \subfigure[GPT2-Small]{\includegraphics[width=0.22\textwidth]{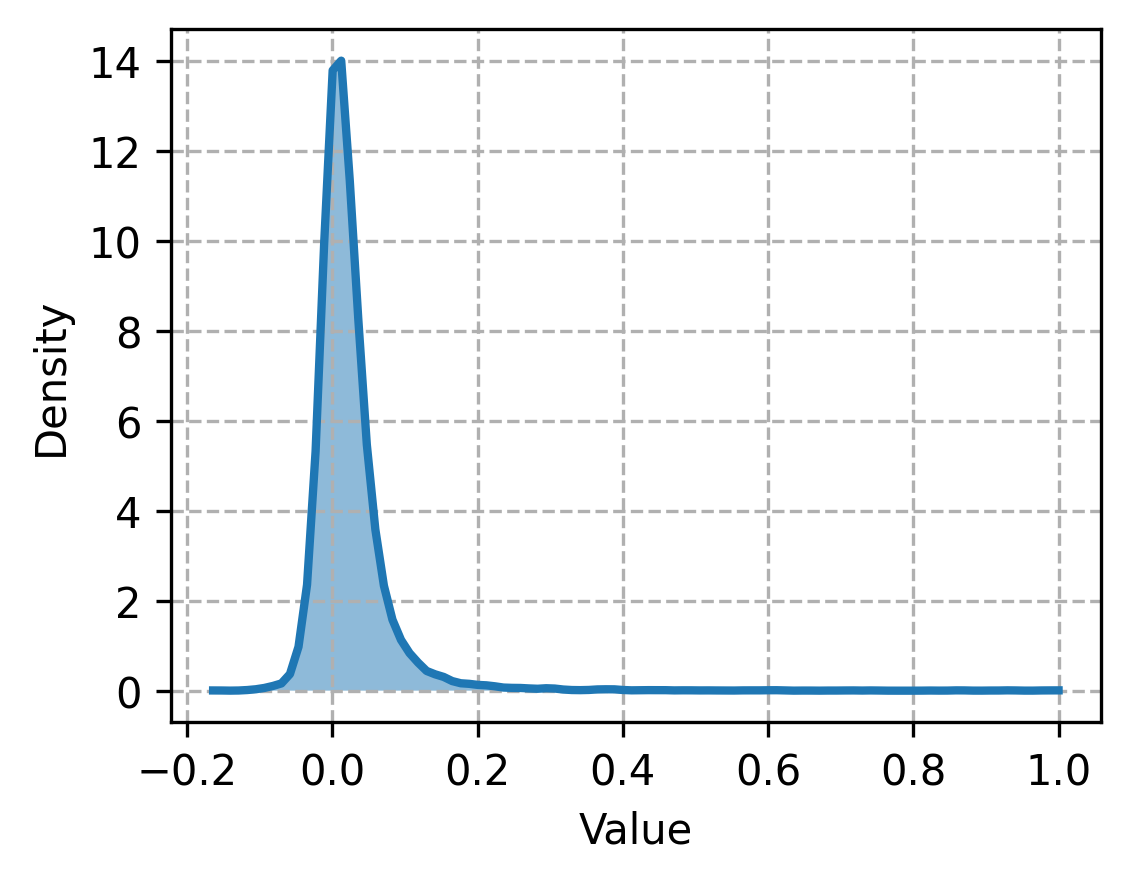}
    }
    \hfill
    \subfigure[GPT2-Medium]{\includegraphics[width=0.22\textwidth]{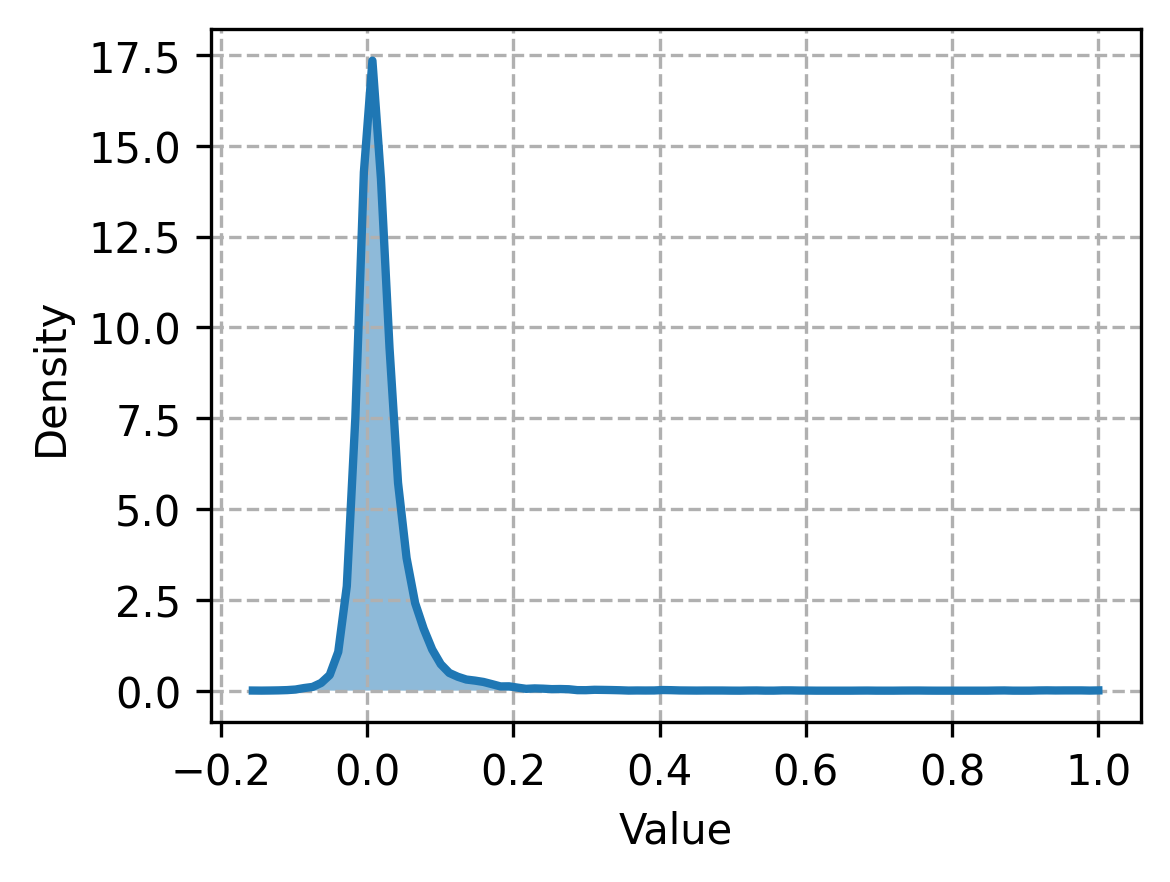}
    }
    \hfill
    \subfigure[GPT2-Large]{\includegraphics[width=0.22\textwidth]{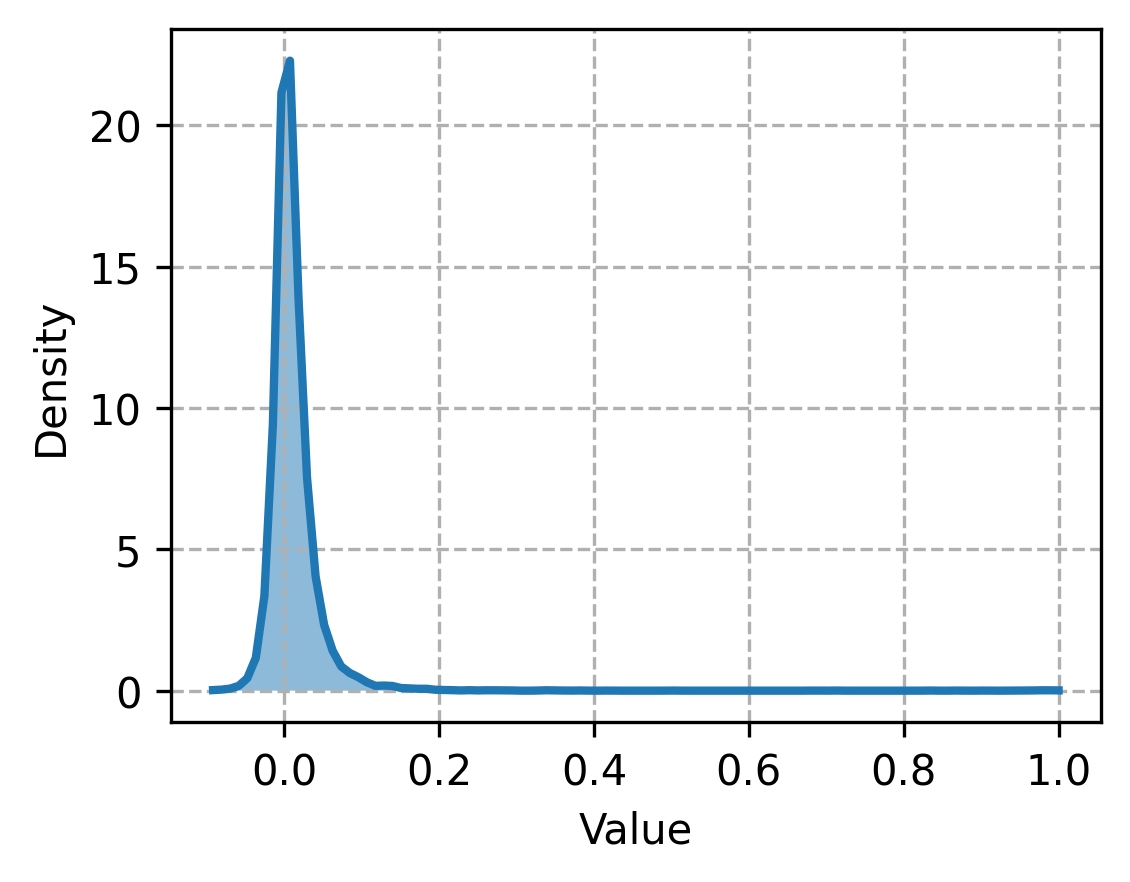}
    }
    \hfill
    \subfigure[GPT-J-6B]{\includegraphics[width=0.22\textwidth]{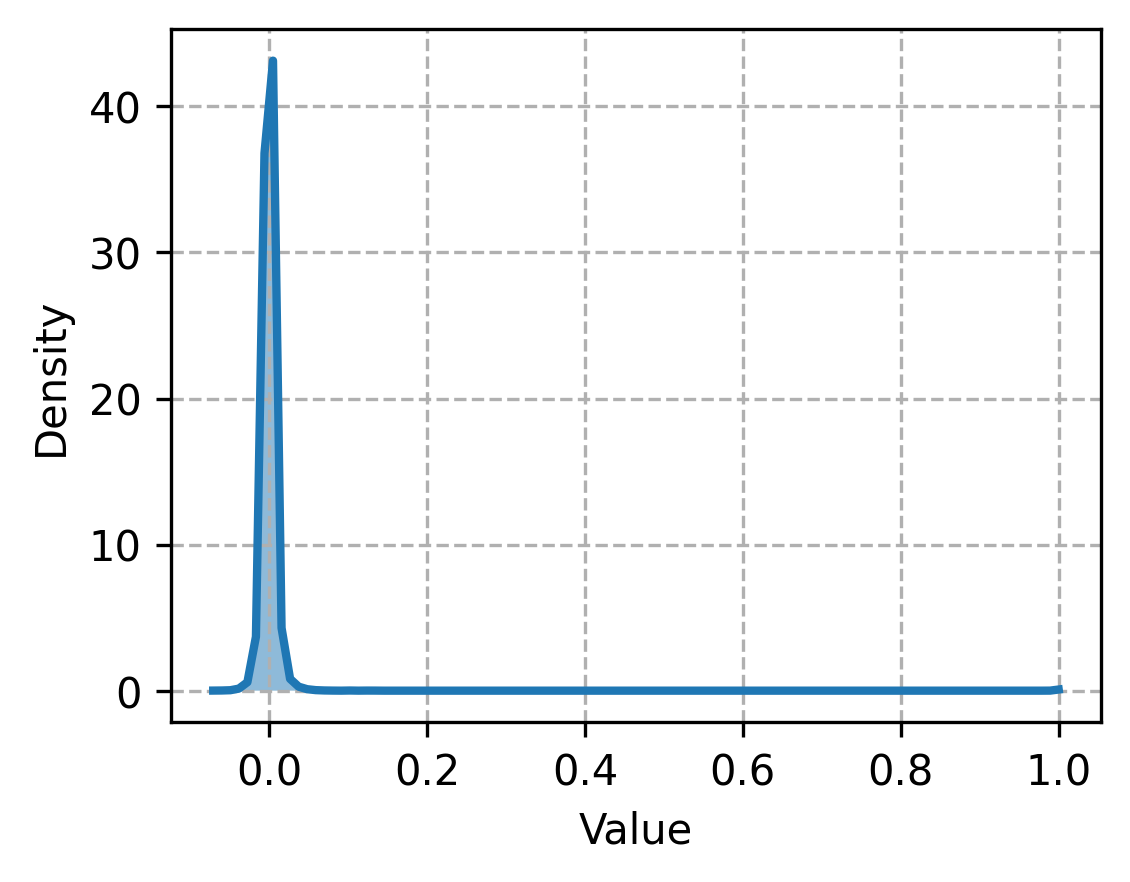}
    }
  \caption{KDE of \(p(\cdot,\cdot)\) values in P matrics at layer 0 across different language models, ordered by model size. In \( P \) matrices, \(p(\cdot,\cdot)\) values are concentrated around 0, with high kurtosis, which increases as model size grows.}
  \label{fig:kde_superposition}
\end{figure}

To gain a more intuitive understanding of the distribution characteristics of knowledge superposition, we remove the diagonal elements from the \(P\) matrices across all layers of all the language models we studied and then plot the kernel density estimation (KDE) of the remaining elements. In Figure~\ref{fig:kde_superposition}, we present the kernel density estimation of the superposition distribution at layer 0 for GPT2-Small, GPT2-Medium, GPT2-Large, and the GPT-J, ordered by model size. Additional kernel density estimations for all layers of all models are provided in Appendix~\ref{appendix:E. KDE of Matrix for All Layers}E.

It can be observed that the superposition distribution exhibits characteristics of a heavy-tailed distribution with high kurtosis and zero mean. As the model size increases, the kurtosis of distribution becomes larger and the distribution becomes more concentrated around 0. This indicates that smaller models, constrained by capacity, exhibit more superposition, attempting to store knowledge representations in a relatively orthogonal manner. In contrast, larger models have greater capacity to store knowledge, allowing them to store it in a more orthogonal manner compared to smaller models, resulting in a kernel density estimation more concentrated around 0.

\subsection{Scaling Law for Superposition}

We have already seen that the degree of knowledge superposition varies with model size. In this section, we formally study the scaling law of superposition by experimenting with different-sized models from the same language model family. These models have consistent architecture and training corpora, differing only in size. As shown in Figure~\ref{fig:scaling_law_superposition}, we examine the kurtosis across all layers of a total of 8 language models from the GPT2, Pythia, and Llama-2 families. Higher kurtosis reflects  weaker superposition.

We can observe a clear scaling law: as model size increases, kurtosis tends to rise, indicating that the degree of superposition between pieces of knowledge decreases with larger models. This also explains why larger language models exhibit a higher degree of intelligence. A plausible explanation is that as model size grows, the knowledge from the corpus can be stored with weaker superposition, allowing for effective handling of more complex or knowledge-dense scenarios. since the sparser the features, the stronger the superposition \cite{elhage2022toy}, which also means they can only effectively handle scenarios where knowledge features are sparser.

Additionally, we find that the degree of superposition varies across different layers in the same language model. Earlier layers exhibit higher kurtosis and thus lower degree of superposition, while later layers exhibit lower kurtosis and higher degree of superposition. This is reasonable, as early layers in a language model focus on shallow syntactic features, which are dense and result in weaker superposition. In contrast, later layers focus on deep semantic features, which are sparse and result in stronger superposition. This aligns with \citet{elhage2022toy}.

Furthermore, we find that different architectures have varying impacts on knowledge superposition. As shown in Figure~\ref{fig:scaling_law_superposition}, by observing the vertical axis, it is evident that the Pythia and GPT2 architectures seem to encourage more orthogonal representations of knowledge, achieving higher kurtosis with a smaller number of parameters, indicating weaker superposition. In contrast, the Llama2 architecture seems to encourage greater knowledge superposition, with even larger parameter sizes corresponding to relatively lower kurtosis, indicating stronger superposition. This is consistent with \citet{elhage2022solu}, which suggests that some architectures may encourage sparsity.

\subsection{Superposition in Whitening Space}

\begin{figure}
    \centering
    \subfigure[Llama3-8B (whitening)]{\includegraphics[width=0.22\textwidth]{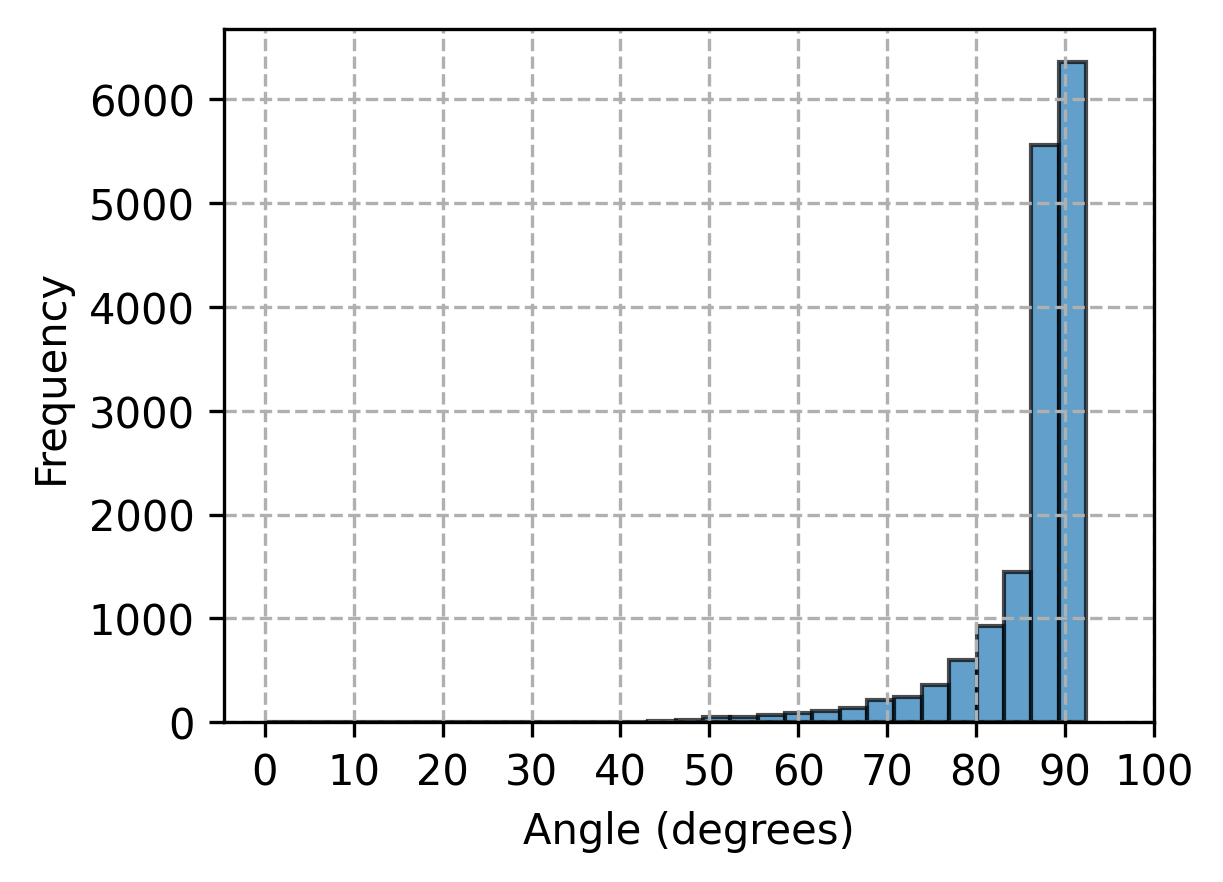}
    }
    \hfill
    \subfigure[Llama3.1-8B (whitening)] {\includegraphics[width=0.22\textwidth]{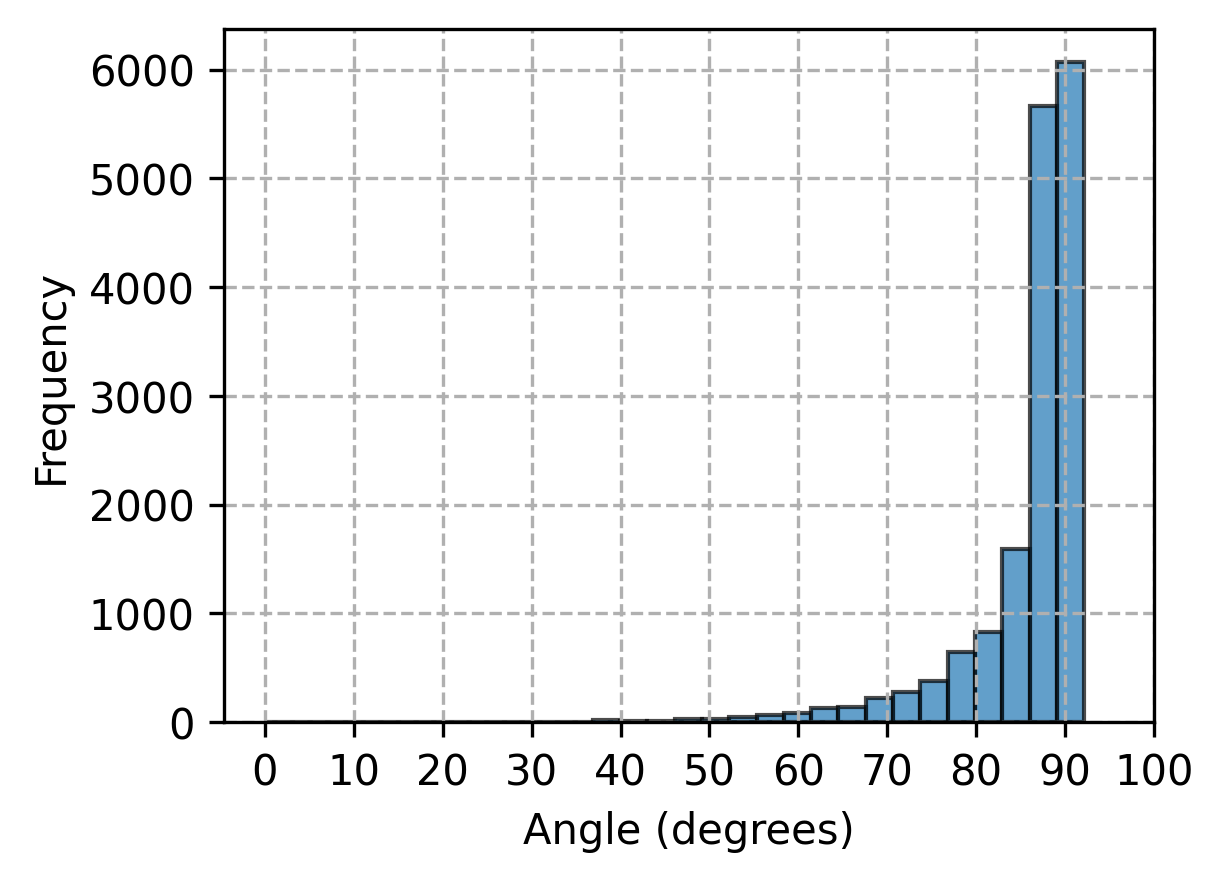}
    }
    \hfill
    \subfigure[Llama3-8B (activation)]{\includegraphics[width=0.22\textwidth]{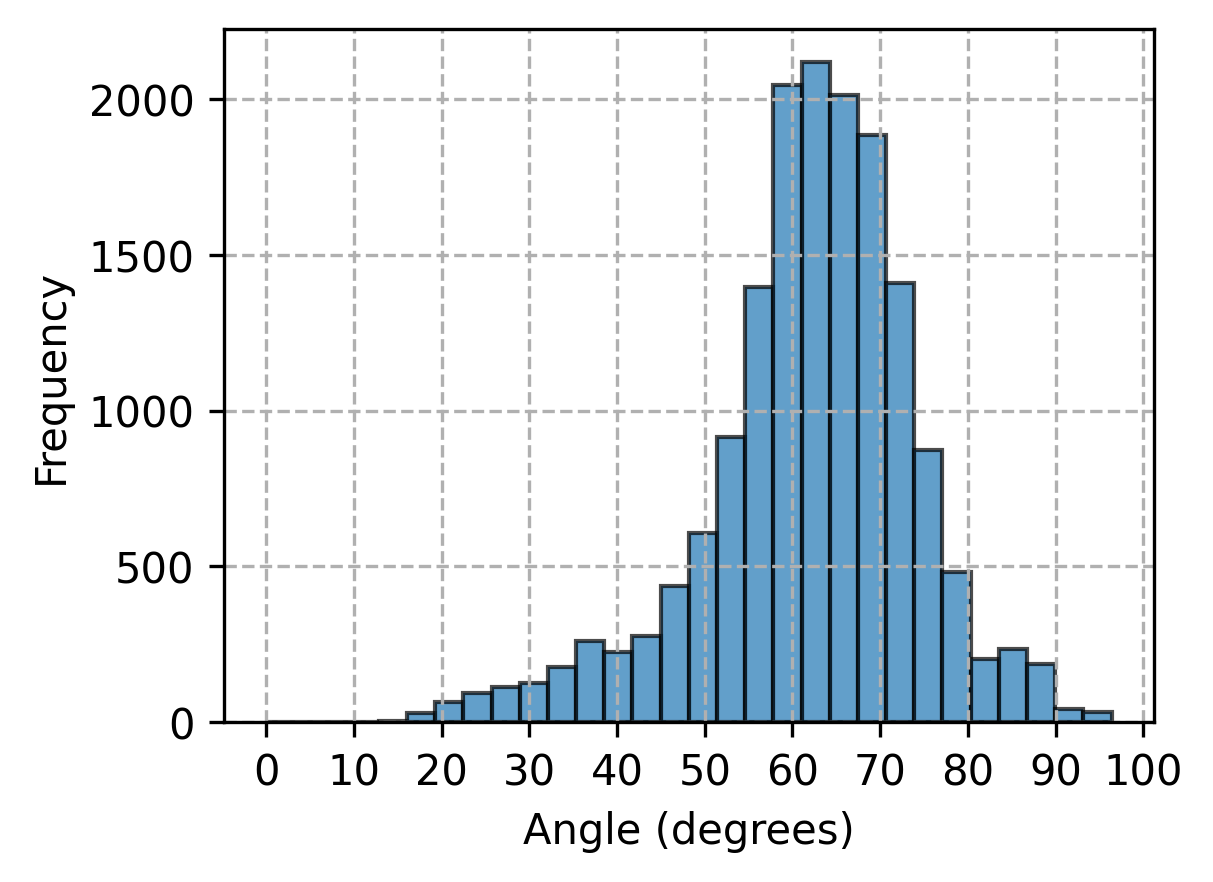}
    }
    \hfill
    \subfigure[Llama3.1-8B  (activation)]{\includegraphics[width=0.22\textwidth]{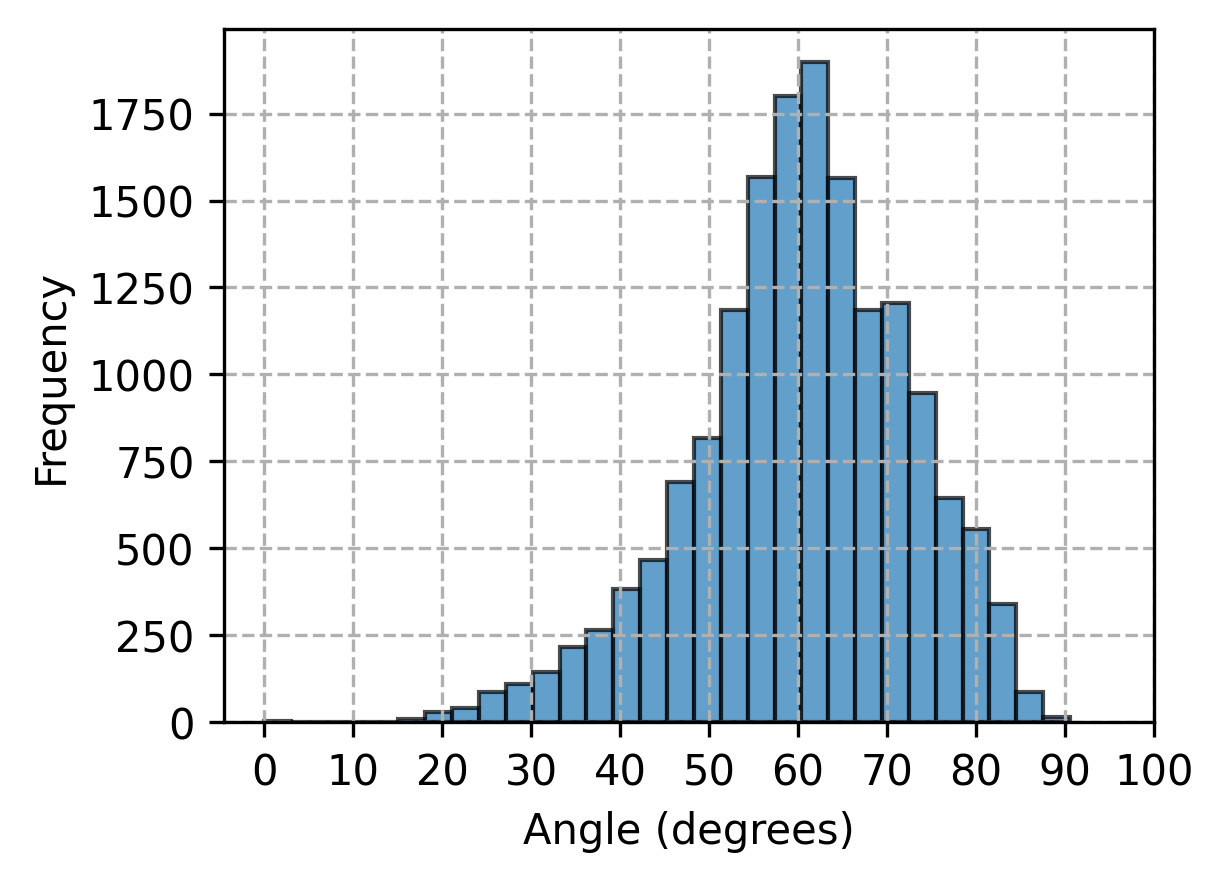}
    }
  \caption{Angular distribution of knowledge representations in whitening space and activation space in last layer.}
  \label{fig:whitening_vs_activation}
\end{figure}

In this section, we provide further direct evidence of superposition by calculating the angular distribution of knowledge representations, demonstrating its presence in the whitening space. Specifically, we calculate the angles between 128x128 pairs of knowledge representations in the whitening space on Llama3-8B and Llama-3.1-8B. For comparison, we also calculate the angles in the activation space (i.e., directly from MLP's hidden activations). 

In the whitening space (Figure~\ref{fig:whitening_vs_activation}ab), the angles are primarily concentrated around 90 degrees, indicating that language models attempt to represent knowledge orthogonally, though limited model capacity results in a long tail. However, in the activation space (Figure~\ref{fig:whitening_vs_activation}cd), the relationships between these knowledge representations are indiscernible, showing no orthogonal tendencies.

Importantly, superposition in the whitening space is derived through rigorous mathematical reasoning, whereas superposition in the activation space is based on a heuristic perspective \cite{elhage2022toy}. However, this heuristic perspective has been used as a fundamental assumption in many works that address superposition (\citealp{cunningham2023sparse}; \citealp{bricken2023monosemanticity}; \cite{templeton2024scaling}).

\section{Conclusion}

In this research, we focus on the failure of lifelong knowledge editing and explore its fundamental reason. Our rigorous mathematical derivation indicates that continuous knowledge editing can interfere with both original and previously edited knowledge in language models. Further analysis shows that the degree of interference is determined by the extent of knowledge superposition. Subsequently, we explore the presence and properties of superposition in real-world language models. Our experiments reveal that knowledge superposition is universal across language models, characterized by a high-kurtosis, zero-mean, heavy-tailed distribution, with a clear scaling law. In conclusion, our findings indicate that knowledge superposition is the fundamental reason for the failure in lifelong editing. Nevertheless, the scaling law of superposition still suggests future directions for knowledge editing, including: on the model side, attempting knowledge editing in larger language models with sparser architectures; and on the algorithm side, attempting to decompose knowledge within language models and performing edits based on this decomposition.

\section{Acknowledgements}

This work is supported by the National Natural Science Foundation of China (No.U24A20335). This work is also supported by Beijing Natural Science Foundation (L243006) and the National Natural Science Foundation of China (No.62176257, No.62406321). This work is also supported by the Youth Innovation Promotion Association CAS and the China Postdoctoral Science Foundation under Grant Number 2024M753500.

\bibliography{aaai25}

\clearpage

\appendix

\section{A. Proof in Preliminary}\label{appendix:A. Proof in Preliminary}

We present a detailed derivation of Equation~\ref{equ:closed-form_solution} from Section Preliminary.

We assume that \( W \) is the optimal least-squares solution for the mapping from a set of previous keys \( K \) to values \( V \), described as follows,
\begin{equation}
    W=\mathop{argmin}_{W}\|WK-V\|_F^2,
\end{equation}
where the Frobenius norm of \( WK - V \) is used to represent the total squared error, as the variable being optimized is the matrix \( W \).

We then define the function \( L(W) \) with respect to \( W \),
\begin{align}
    L(W) &= \|WK-V\|_F^2\\
    &=\sum_{i,j}|(WK-V)_{ij}|^2\\
    &=\sum_i\sum_j(WK-V)_{ij}(WK-V)_{ij}\\
    &=\sum_i\sum_j(WK-V)_{ij}(WK-V)_{ji}^T\\
    &=\sum_{i}\left( (WK-V)(WK-V)^T \right)_{ii}\\
    &=\text{Tr}\left( (WK-V)(WK-V)^T \right)\\
    &=\text{Tr}(WKK^TW^T\!\!-WKV^T\!\!-VK^TW^T\!\!+VV^T)
\end{align}
Using the linear property of the trace, \( \text{Tr}(A + B) = \text{Tr}(A) + \text{Tr}(B) \), and the fact that \( \text{Tr}(A) = \text{Tr}(A^T) \), we obtain
\begin{align}
    L(W)&=\text{Tr}(WKK^TW^T-2WKV^T+VV^T)\\
    &=\text{Tr}(WKK^TW^T)-2\text{Tr}(WKV^T)+\text{Tr}(VV^T)
\end{align}

Compute the gradient of \( L(W) \) with respect to \( W \),

\begin{align}
    \nabla_WL(W)&=\nabla_W(\text{Tr}(WKK^TW^T) \notag\\
    &-2\text{Tr}(WKV^T)+\text{Tr}(VV^T)).
\end{align}

\begin{itemize}
    \item To compute the gradient of \( \text{Tr}(WKK^TW^T) \) with respect to \( W \): Because for any matrix \( A \), we have \(\frac{\partial}{\partial W}\text{Tr}(WAW^T)=2WA\), then we get
    \begin{equation}
        \frac{\partial}{\partial W}\text{Tr}(WKK^TW^T)=2WKK^T.
    \end{equation}

    \item To compute the gradient of \( -2\text{Tr}(WKV^T) \) with respect to \( W \): The property of the trace, \(\text{Tr}(AB) = \text{Tr}(BA)\), is utilized, and for any matrices $A$ and $B$, we have $\frac{\partial}{\partial W}(AWB) = A^TB^T$, then we get
    \begin{equation}
        \frac{\partial}{\partial W}(-2\text{Tr}(V^TWK))=-2VK^T.
    \end{equation}
\end{itemize}

Combine the gradients we obtain
\begin{equation}
    \nabla_W L(W)=2(WKK^T-VK^T).
\end{equation}
Setting the gradient to zero, we obtain
\begin{align}
    WKK^T-VK^T=0,\\
    WKK^T=VK^T.
\end{align}
Thus, we obtain the optimal least squares solution for the mapping from the previous set of keys $K$ to values $V$.

To insert the new key-value pair \((k_e, v_e)\), we actually aim to find a new matrix \(\hat W\) that solves the same least squares problem while satisfying an additional equality constraint,
\begin{equation}
    \hat{W}=\mathop{argmin}_{\hat{W}}\|\hat{W}K-V\| \text{ subject to } \hat{W}k_e=v_e.
\end{equation}

This is a well-studied least squares problem with linear equality constraints. The direct solution can be derived by defining and minimizing the Lagrangian function, where \(\Lambda \in \mathbb{R}^d\) minimizes the following

\begin{align}
    \text{define }L(\hat W,\Lambda)&=\frac12\|\hat{W}K-V\|_F^2-\Lambda^T(\hat{W}k_e-v_e)\\
    &\!\!\!\!\!\!\!\!\!\!\!\!\!\!\!\!\!\!\!\!\!\!\!\!\!\!\!\!\!\!\!\!\!\!\!\!\!\!\!\!=\frac12(\hat{W}K)(\hat WK)^T\!\!-\!V(\hat{W}K)+\!\frac12VV^T\!\!-\!\Lambda^T(\hat {W}k_e-v_e)\\
    \text{setting }0=\frac{\partial L}{\partial \hat W}&=\hat{W}(KK^T)-VK^T-\Lambda k_e^T\\
    \hat WKK^T&=VK^T+\Lambda k_e^T
\end{align}

Subtracting the equations \(\hat WKK^T = VK^T + \Lambda k_e^T\) and \(WKK^T = VK^T\), we obtain
\begin{align}
    (\hat{W}-W)KK^T&=\Lambda k_e^T\\
    \hat{W}&=W+\Lambda k_e^T(KK^T )^{-1}  \\
    \hat{W}&=W+\Lambda(C^{-1}k_e)^T
\end{align}
And we have \(\hat{W}k_e = v_e\). Finally, we obtain
\begin{align}
    \hat{W}k_e&=(W+\Lambda (C^{-1}k_e)^T)k_e\\
    &=Wk_e+\Lambda((C^{-1}k_e)^Tk_e)\\
    &=v_e,
\end{align}
then we get
\begin{equation}
    \Lambda=\frac{v_e-Wk_e}{(C^{-1}k_e)^Tk_e}.
\end{equation}
Q.E.D.

\section{B. Proof of Whitening Matrix}\label{appendix:B. Proof of Whitening Matrix}

To prove that \(C^{-\frac{1}{2}}\) is a whitening matrix, it is natural to show that \(C^{-\frac{1}{2}}K\) results in an identity matrix. Since \(C = KK^T\) is a symmetric positive definite matrix and matrix transpose and inversion operations are commutative, we can prove as follows,
\begin{align}
    Cov(C^{-\frac12}K)&=(C^{-\frac12}K)(C^{-\frac12}K)^T\notag\\
    &=C^{-\frac12}KK^T(C^{-\frac12})^T\notag\\
    &=C^{-\frac12}C(C^{T})^{-\frac12}\notag\\
    &=C^{-\frac12}CC^{-\frac12}  \notag\\
    &=I.\notag
\end{align}

\section{C. Number of Knowledge Items \(m\)}\label{appendix:C. Number of Knowledge Items}

\begin{figure*}
    \centering
    \subfigure[\(m=16\)]{\includegraphics[width=0.22\textwidth]{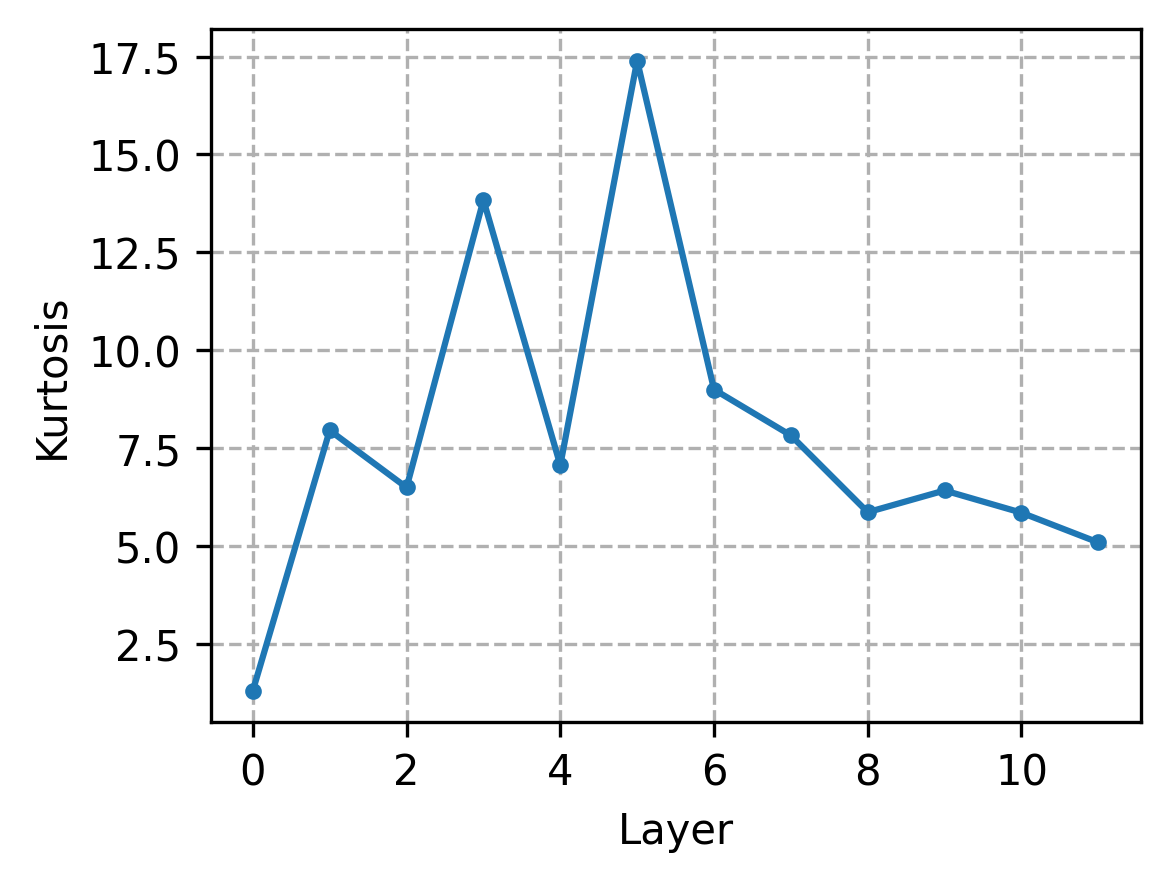}
    }
    \hfill
    \subfigure[\(m=32\)]{\includegraphics[width=0.22\textwidth]{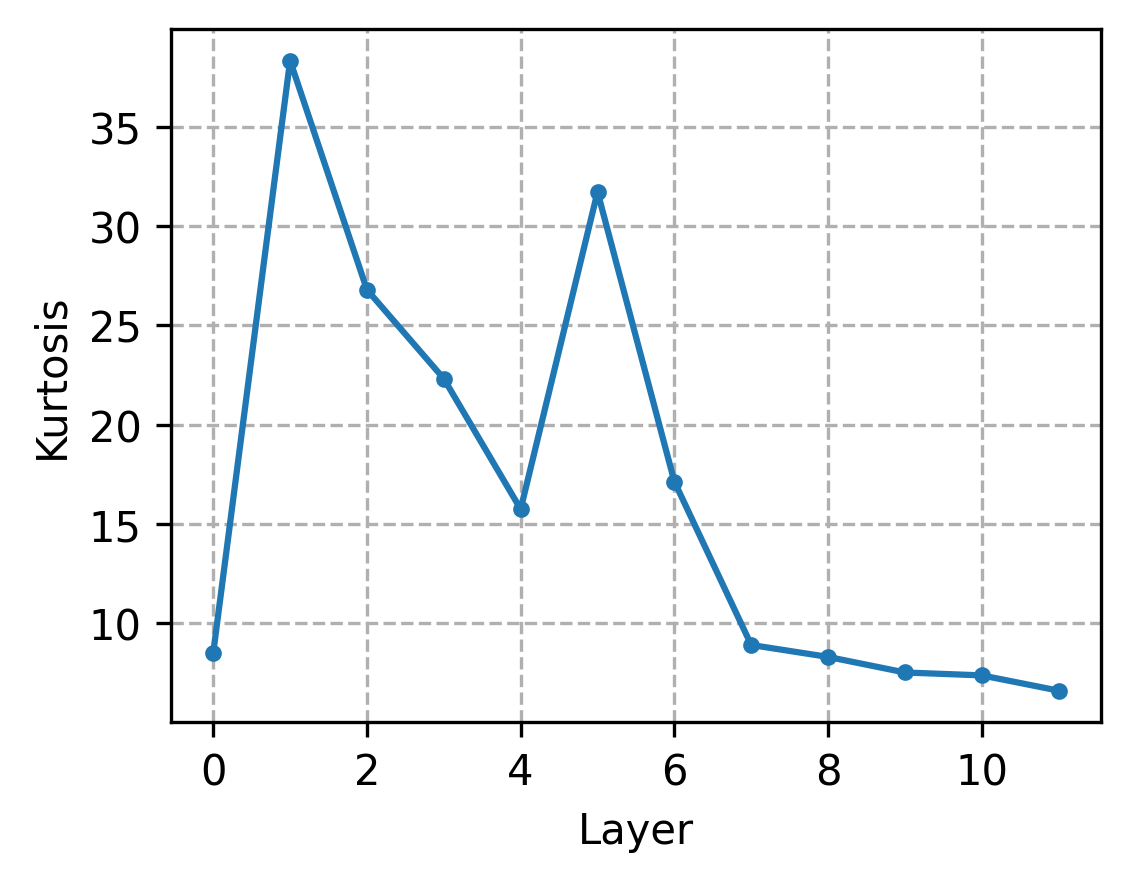}
    }
    \hfill
    \subfigure[\(m=48\)]{\includegraphics[width=0.22\textwidth]{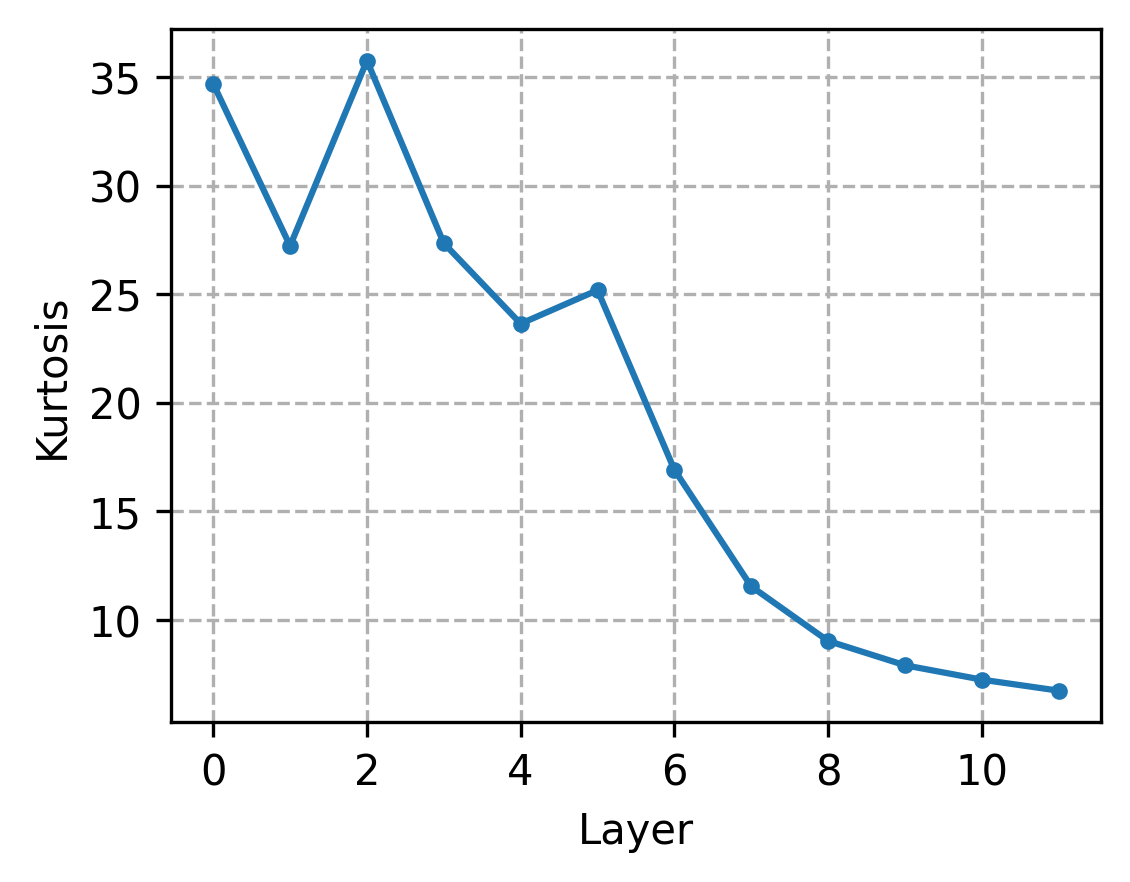}
    }
    \hfill
    \subfigure[\(m=64\)]{\includegraphics[width=0.22\textwidth]{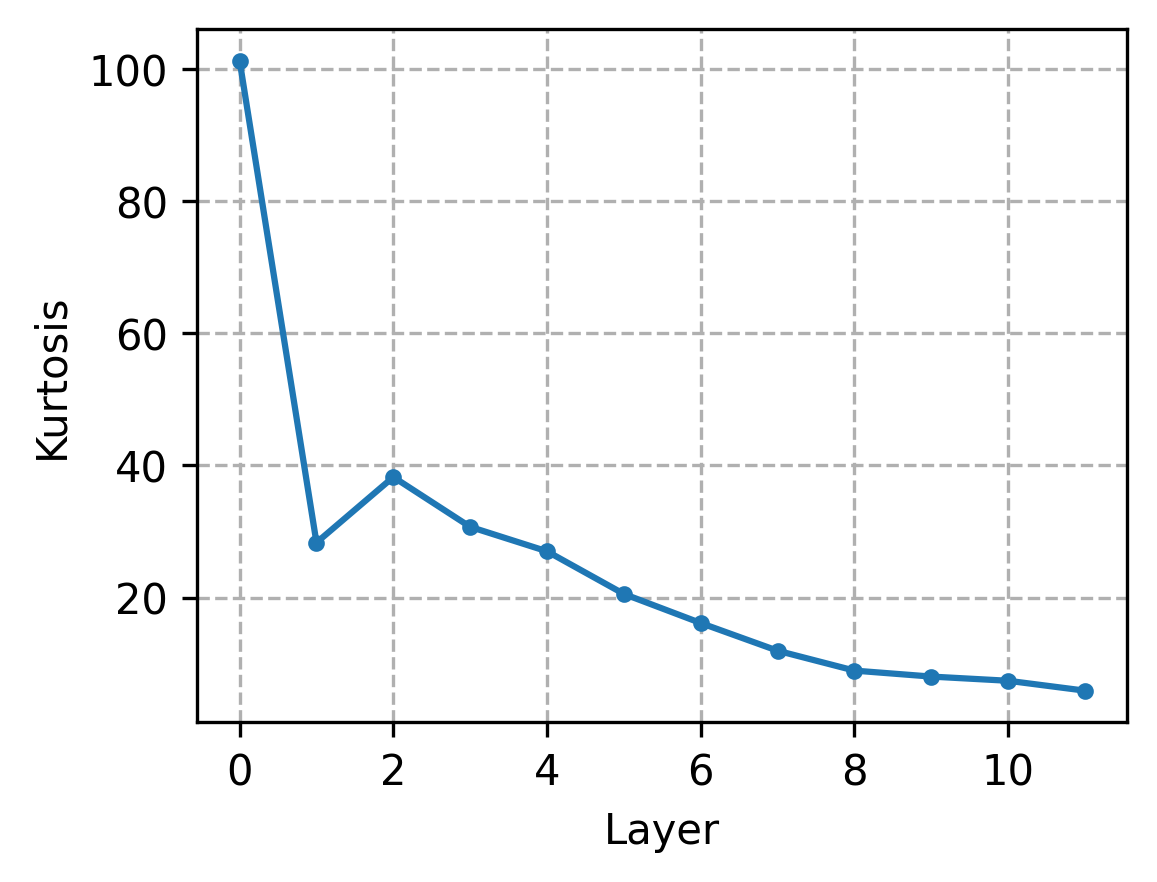}
    }
    \hfill
    \subfigure[\(m=80\)]{\includegraphics[width=0.22\textwidth]{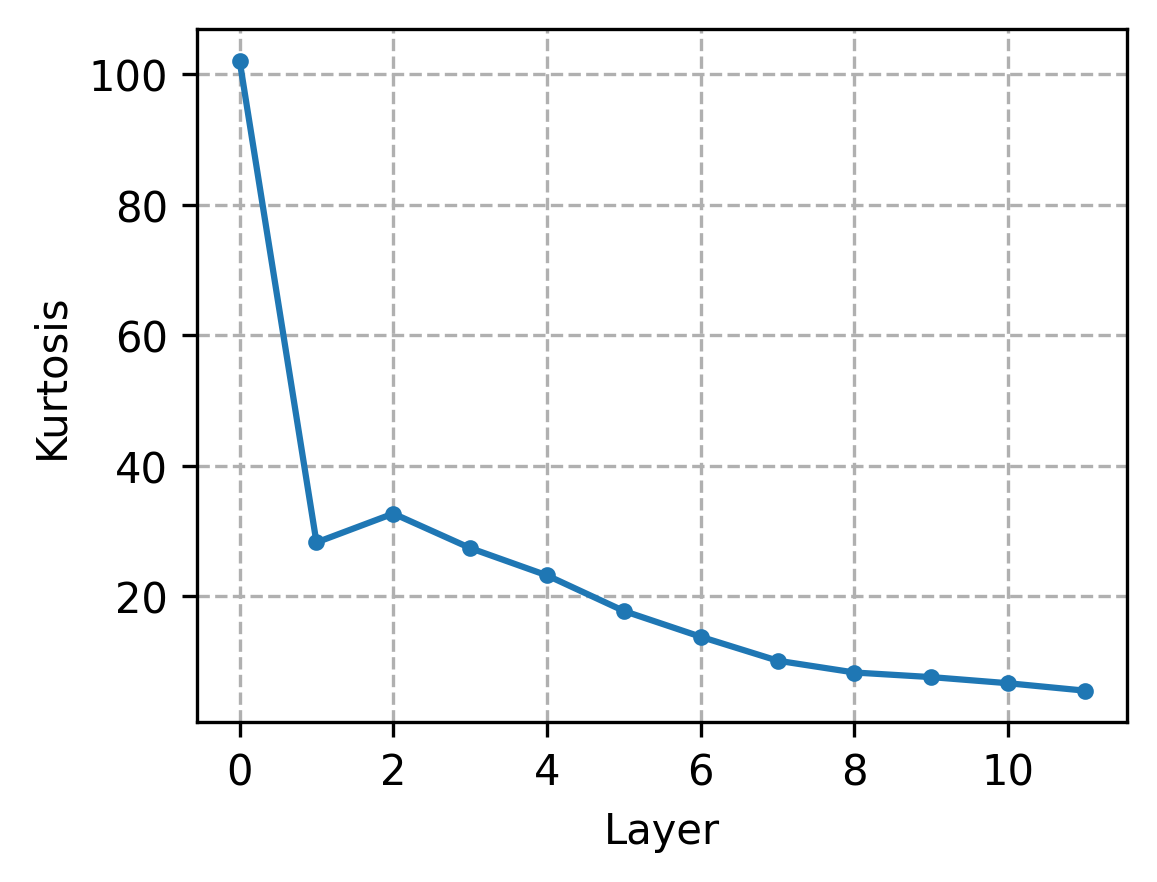}
    }
    \hfill
    \subfigure[\(m=96\)]{\includegraphics[width=0.22\textwidth]{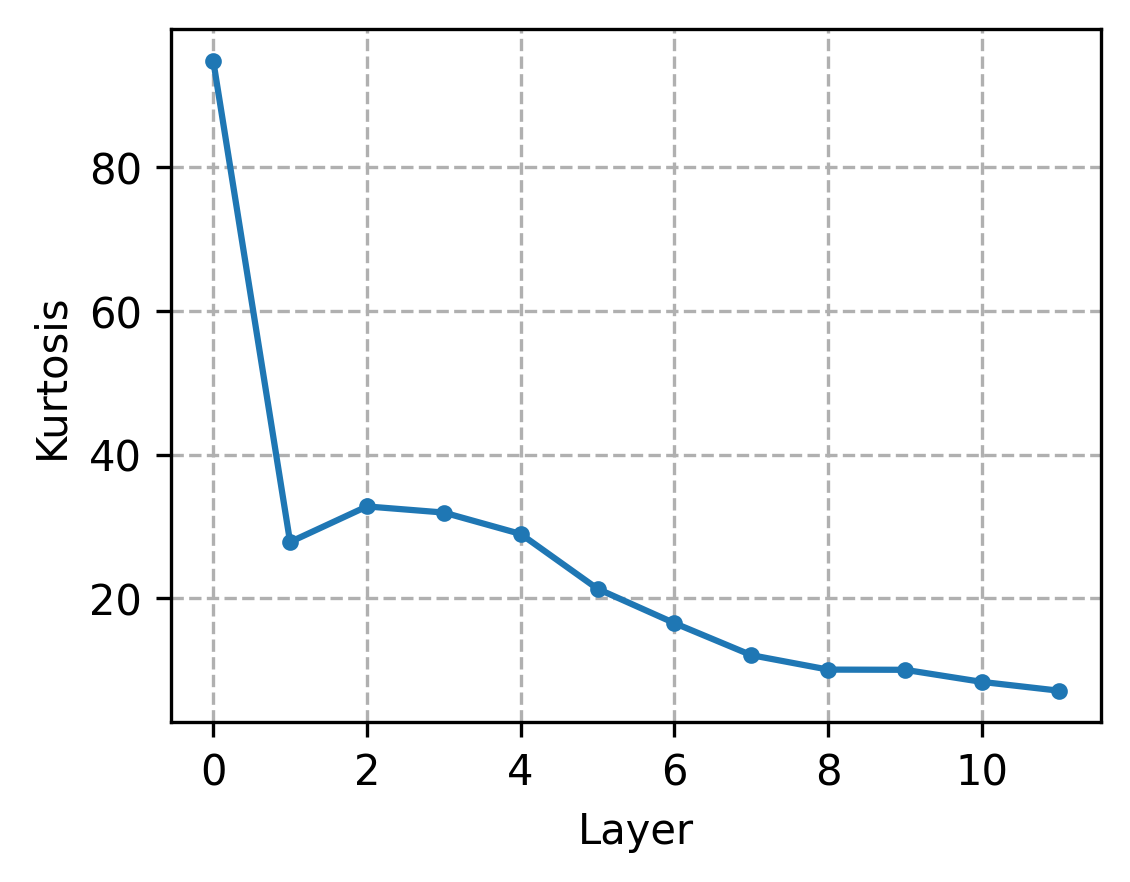}
    }
    \hfill
    \subfigure[\(m=112\)]{\includegraphics[width=0.22\textwidth]{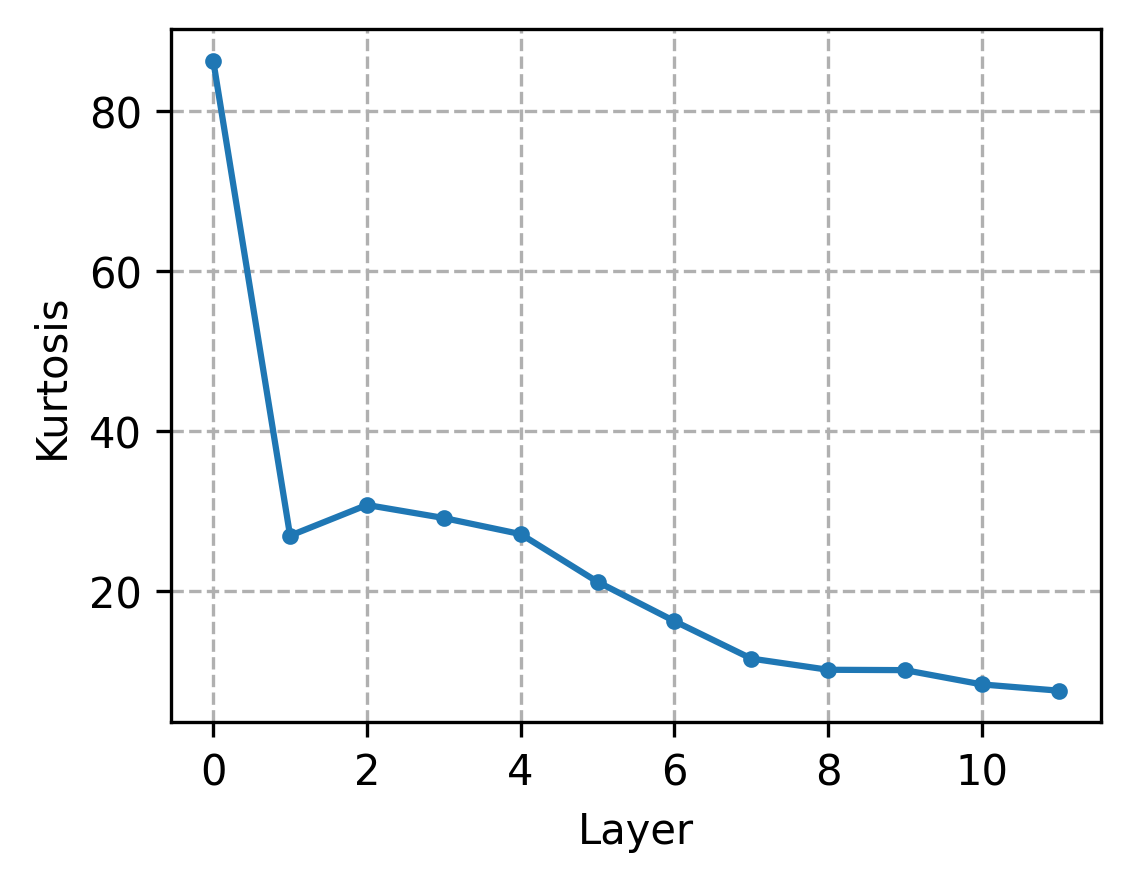}
    }
    \hfill
    \subfigure[\(m=128\)]{\includegraphics[width=0.22\textwidth]{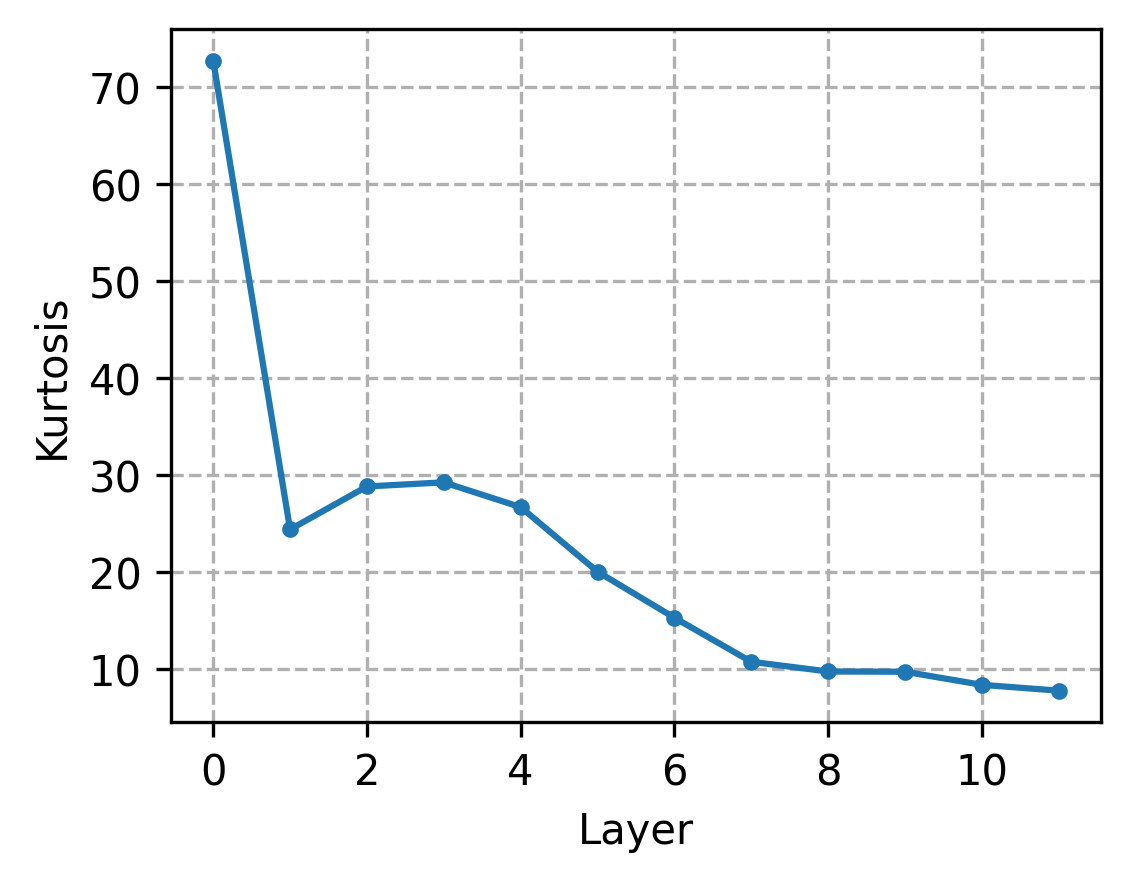}
    }
  \caption{In GPT2-Small, the superposition distribution converges before \( m = 128 \).}
  \label{fig:superposition_converges_gpt2-small}
\end{figure*}

\begin{figure*}
    \centering
    \subfigure[\(m=16\)]{\includegraphics[width=0.22\textwidth]{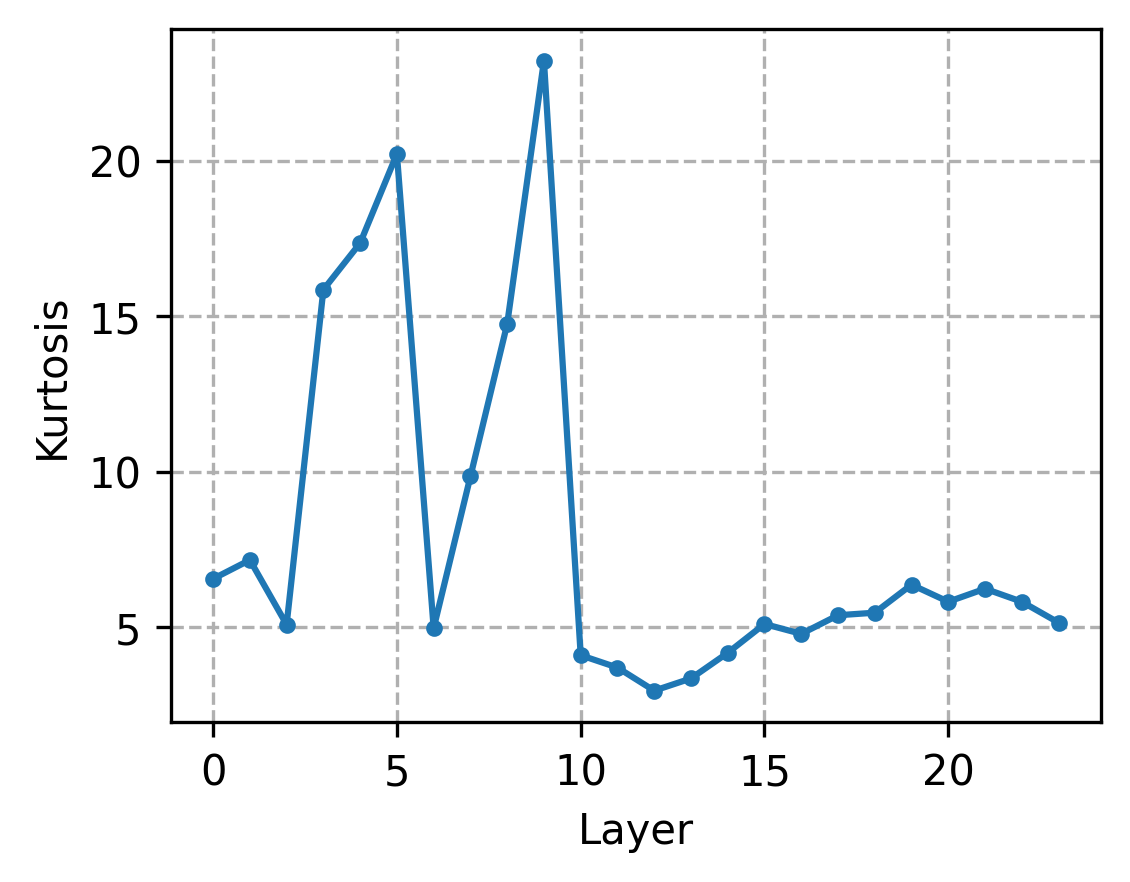}
    }
    \hfill
    \subfigure[\(m=32\)]{\includegraphics[width=0.22\textwidth]{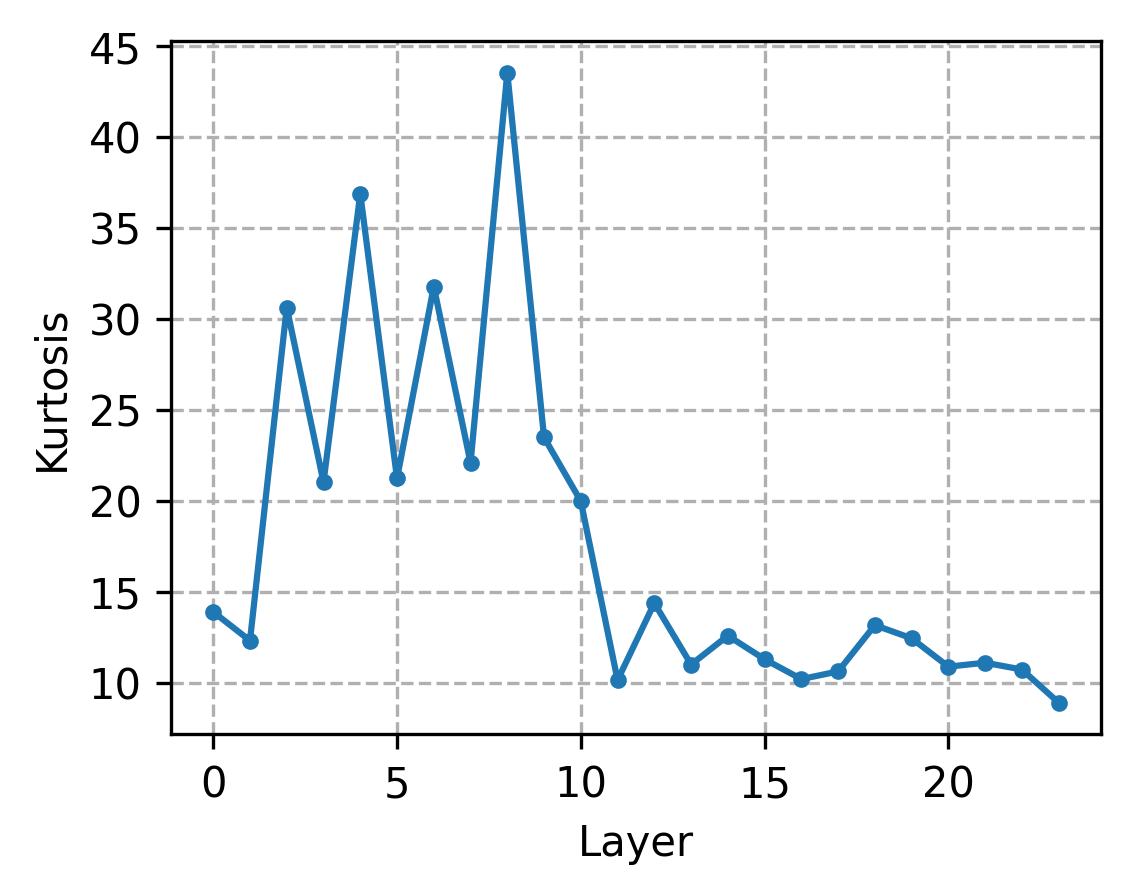}
    }
    \hfill
    \subfigure[\(m=48\)]{\includegraphics[width=0.22\textwidth]{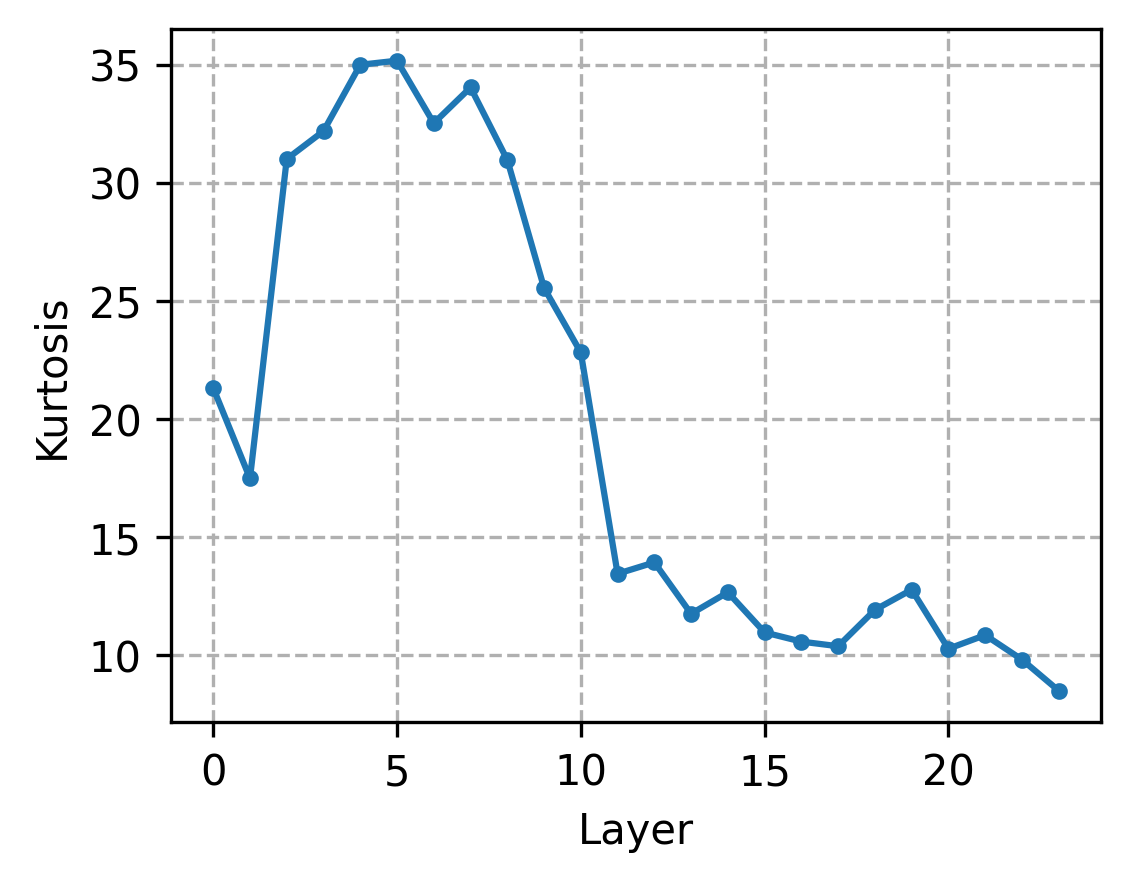}
    }
    \hfill
    \subfigure[\(m=64\)]{\includegraphics[width=0.22\textwidth]{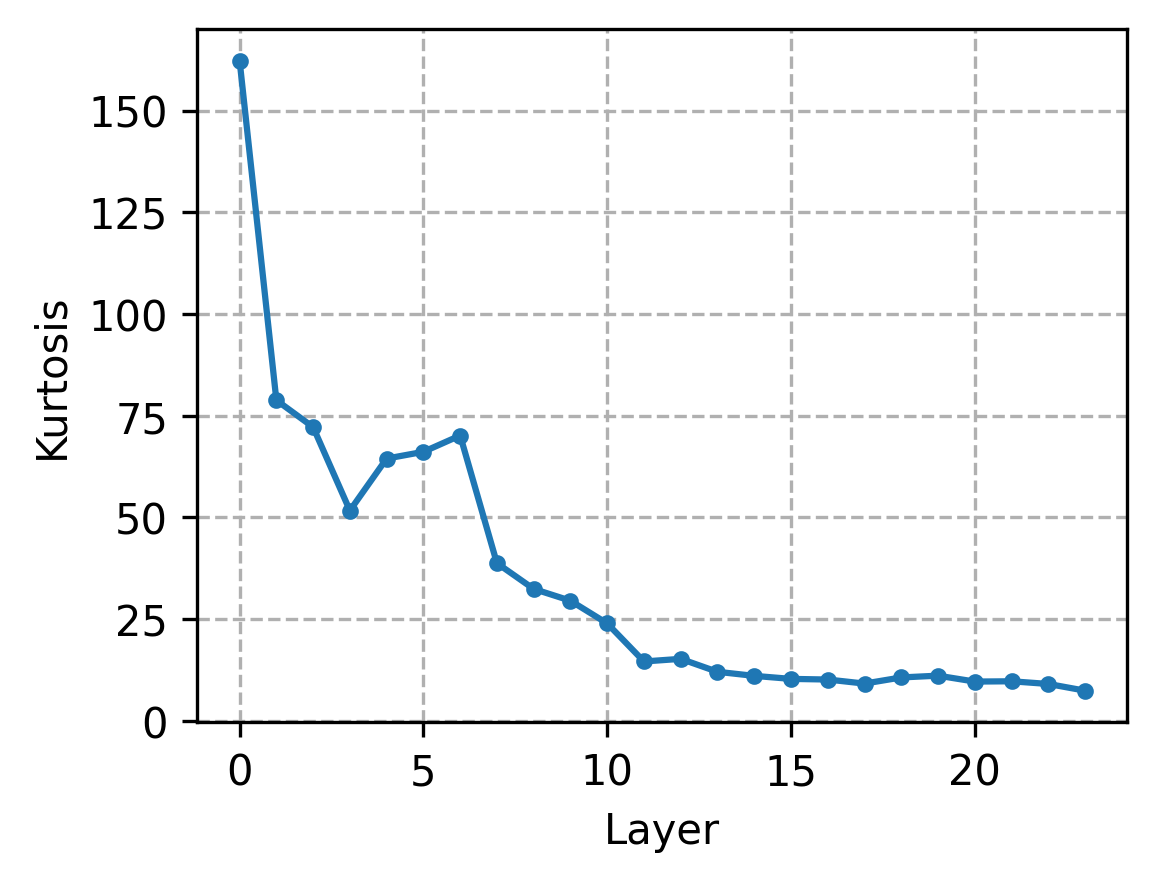}
    }
    \hfill
    \subfigure[\(m=80\)]{\includegraphics[width=0.22\textwidth]{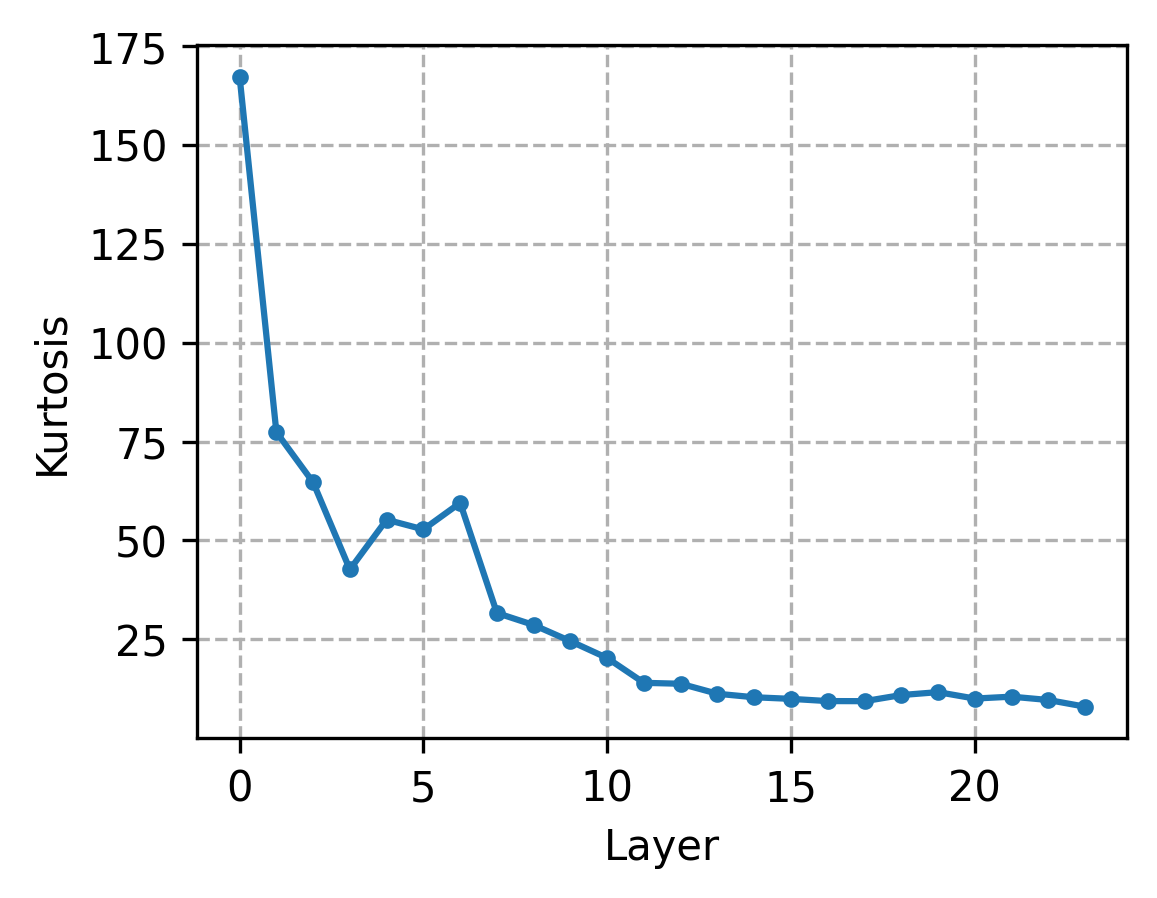}
    }
    \hfill
    \subfigure[\(m=96\)]{\includegraphics[width=0.22\textwidth]{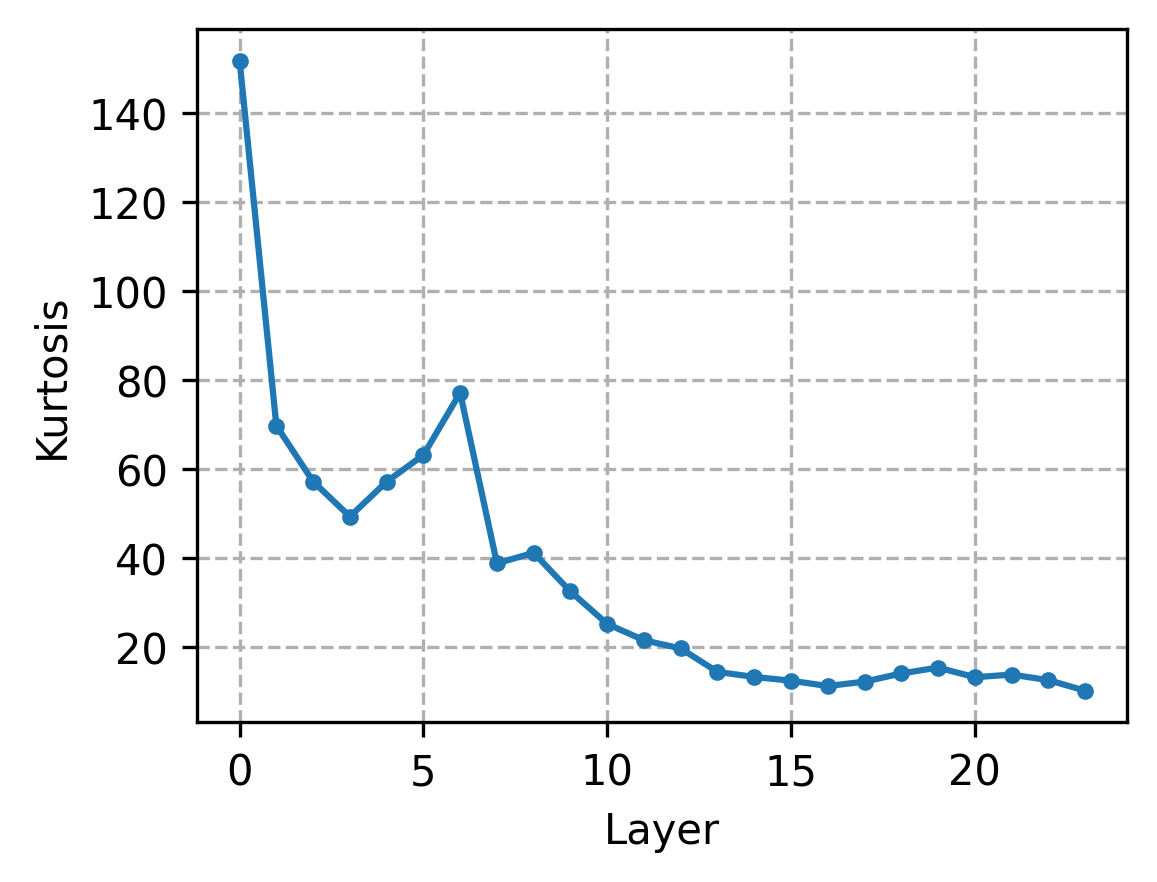}
    }
    \hfill
    \subfigure[\(m=112\)]{\includegraphics[width=0.22\textwidth]{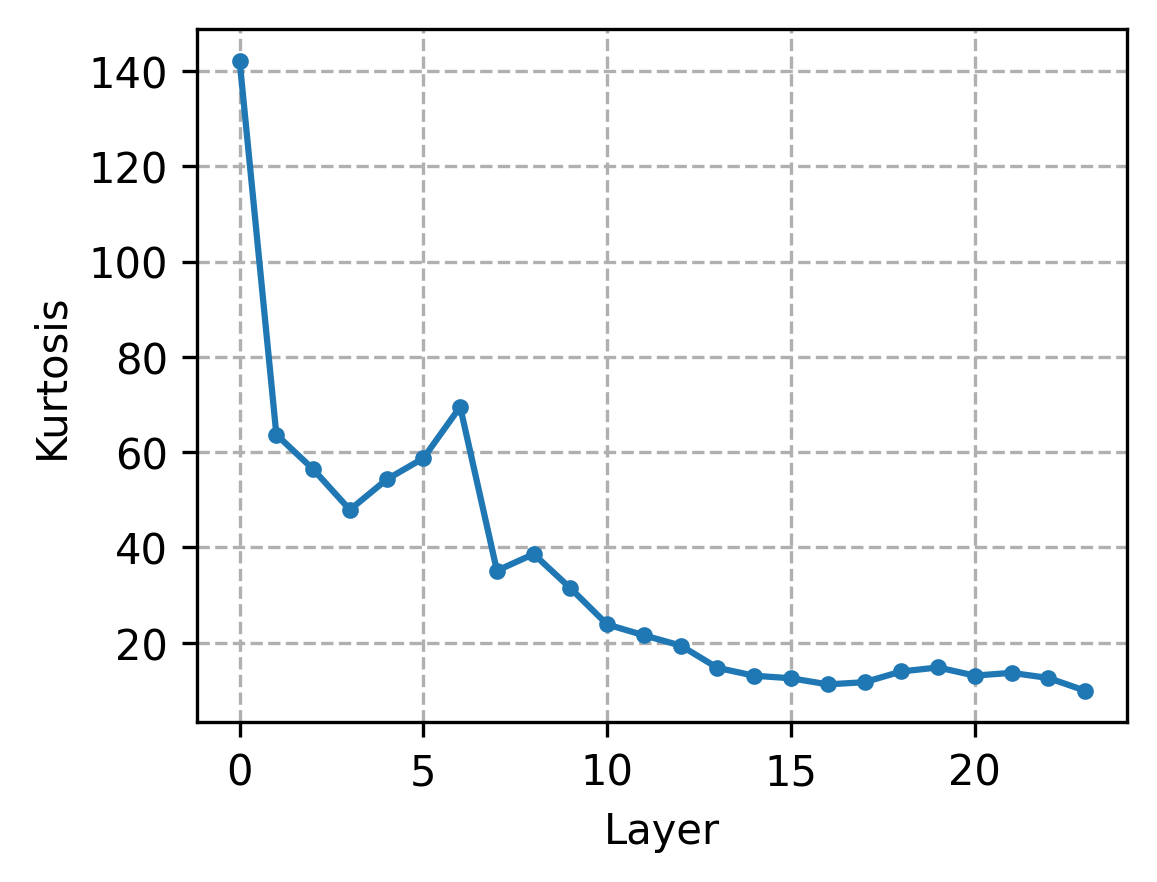}
    }
    \hfill
    \subfigure[\(m=128\)]{\includegraphics[width=0.22\textwidth]{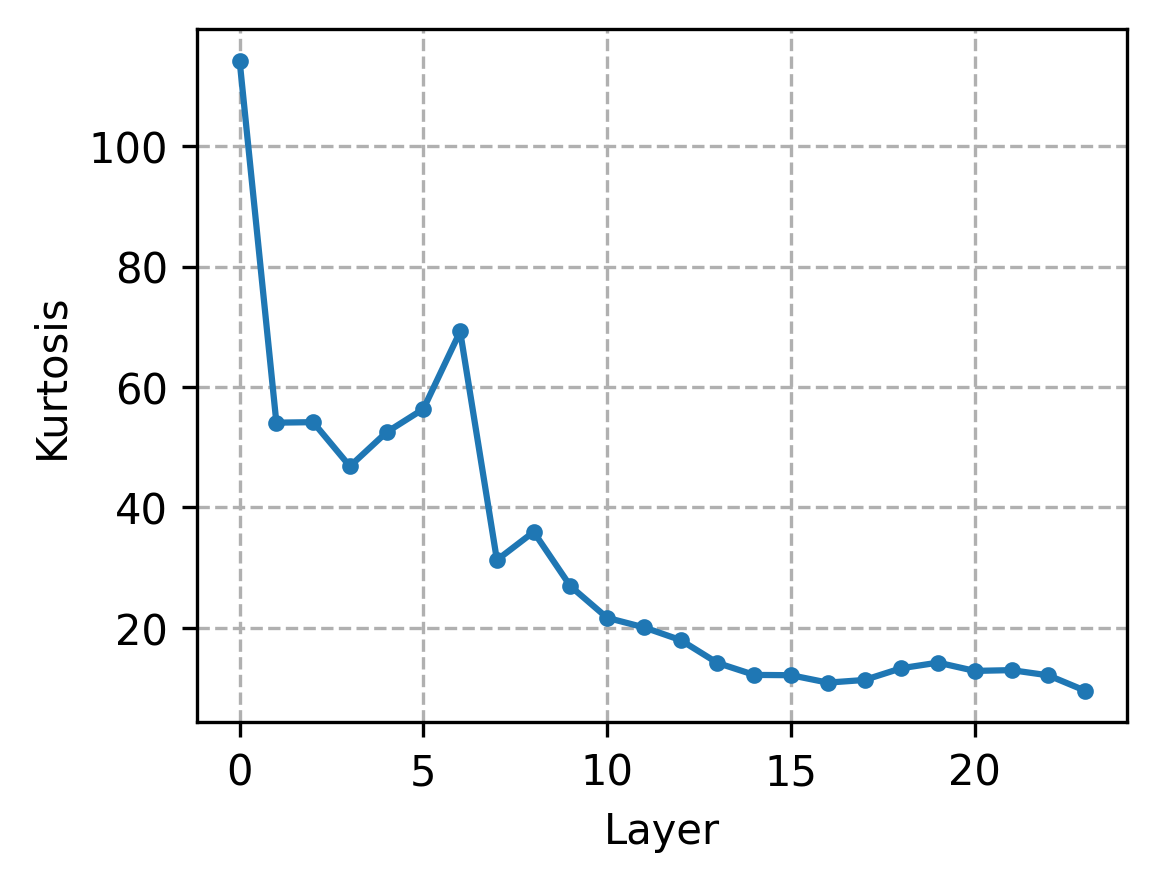}
    }
  \caption{In GPT2-Medium, the superposition distribution converges before \( m = 128 \).}
  \label{fig:superposition_converges_gpt2-medium}
\end{figure*}

\begin{figure*}
    \centering
    \subfigure[\(m=16\)]{\includegraphics[width=0.22\textwidth]{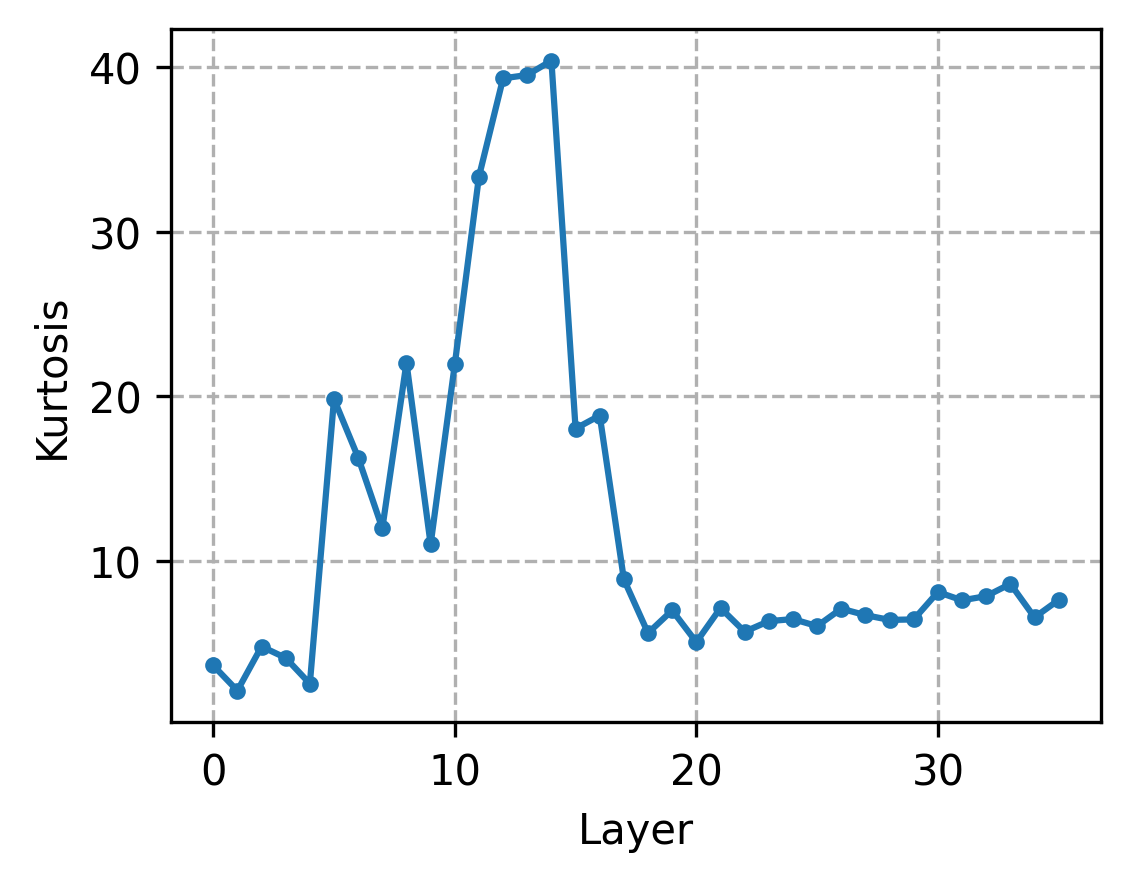}
    }
    \hfill
    \subfigure[\(m=32\)]{\includegraphics[width=0.22\textwidth]{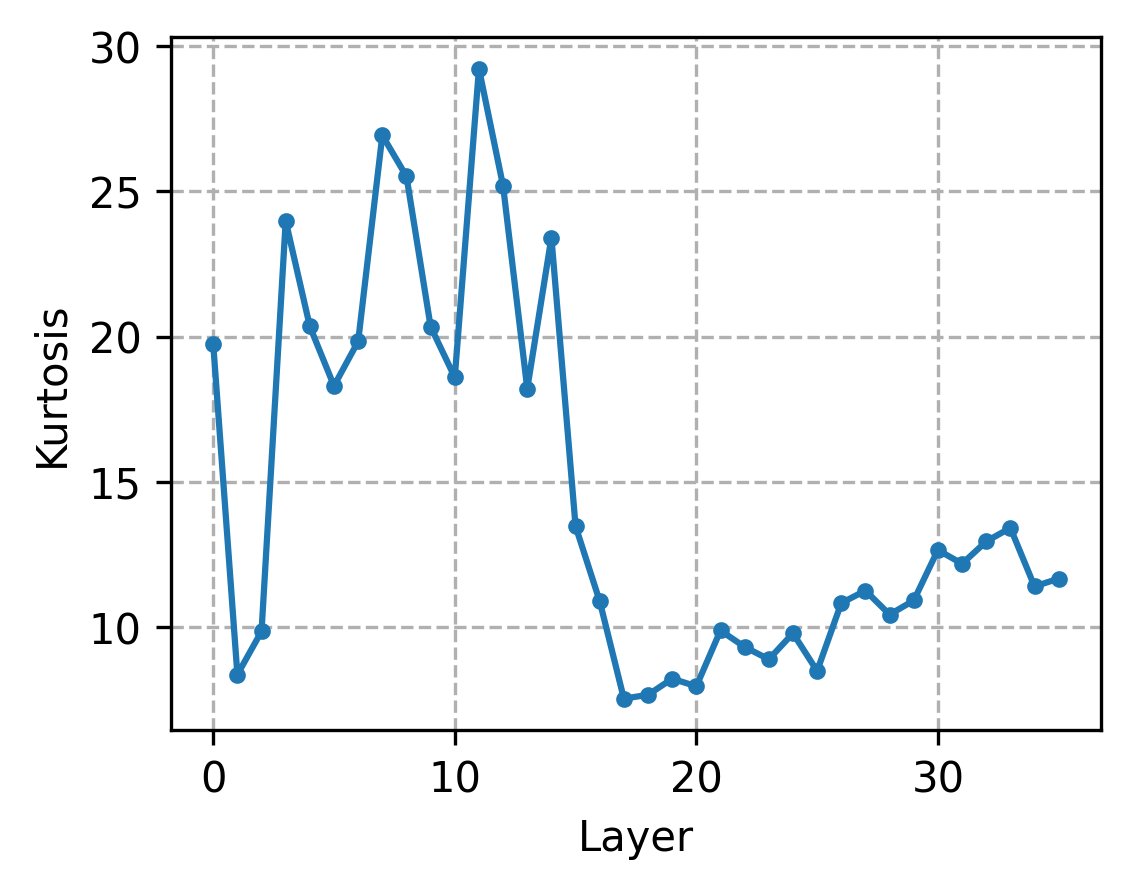}
    }
    \hfill
    \subfigure[\(m=48\)]{\includegraphics[width=0.22\textwidth]{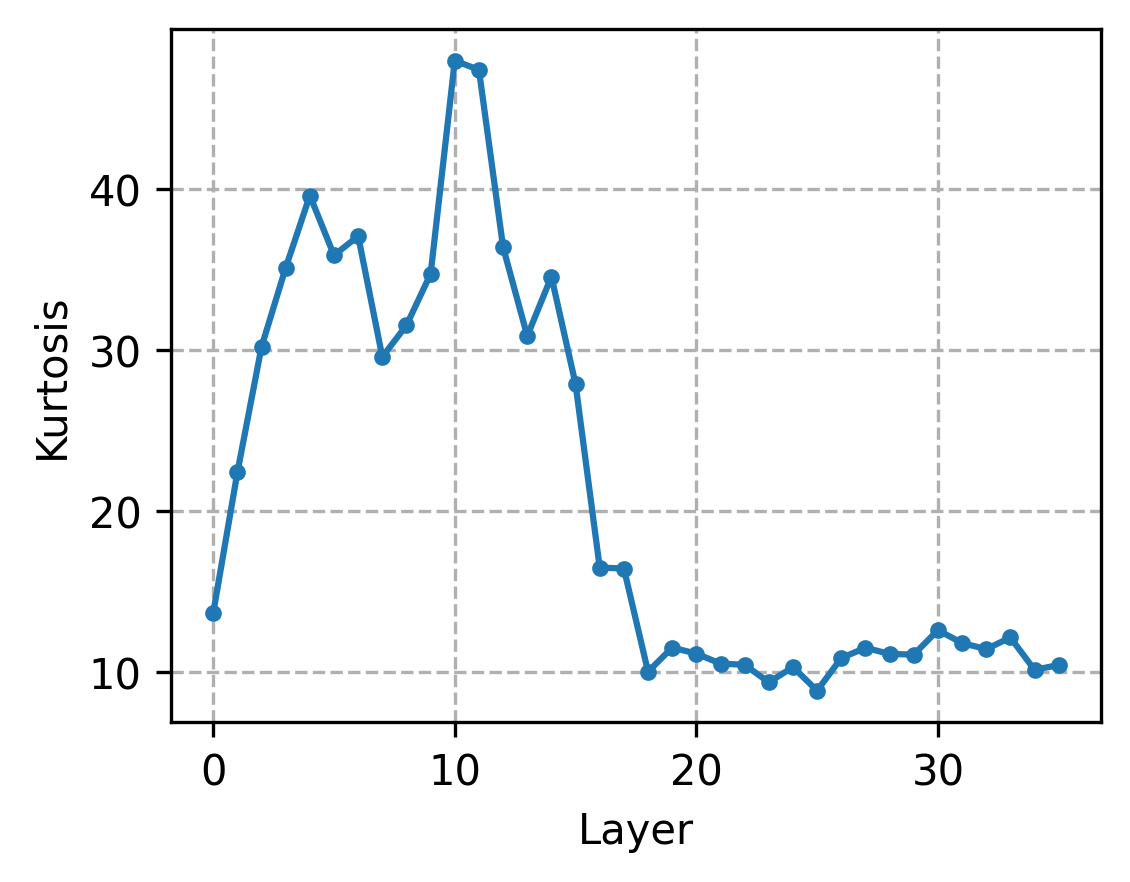}
    }
    \hfill
    \subfigure[\(m=64\)]{\includegraphics[width=0.22\textwidth]{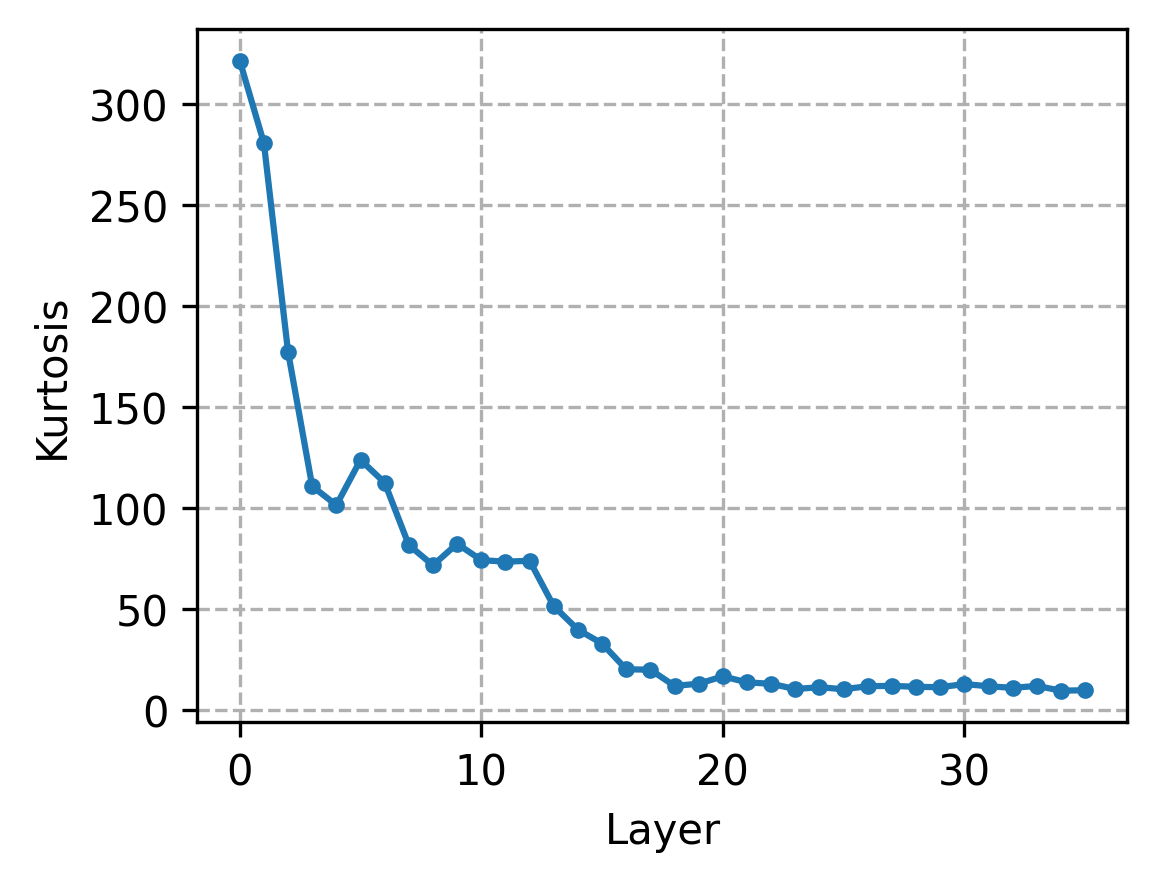}
    }
    \hfill
    \subfigure[\(m=80\)]{\includegraphics[width=0.22\textwidth]{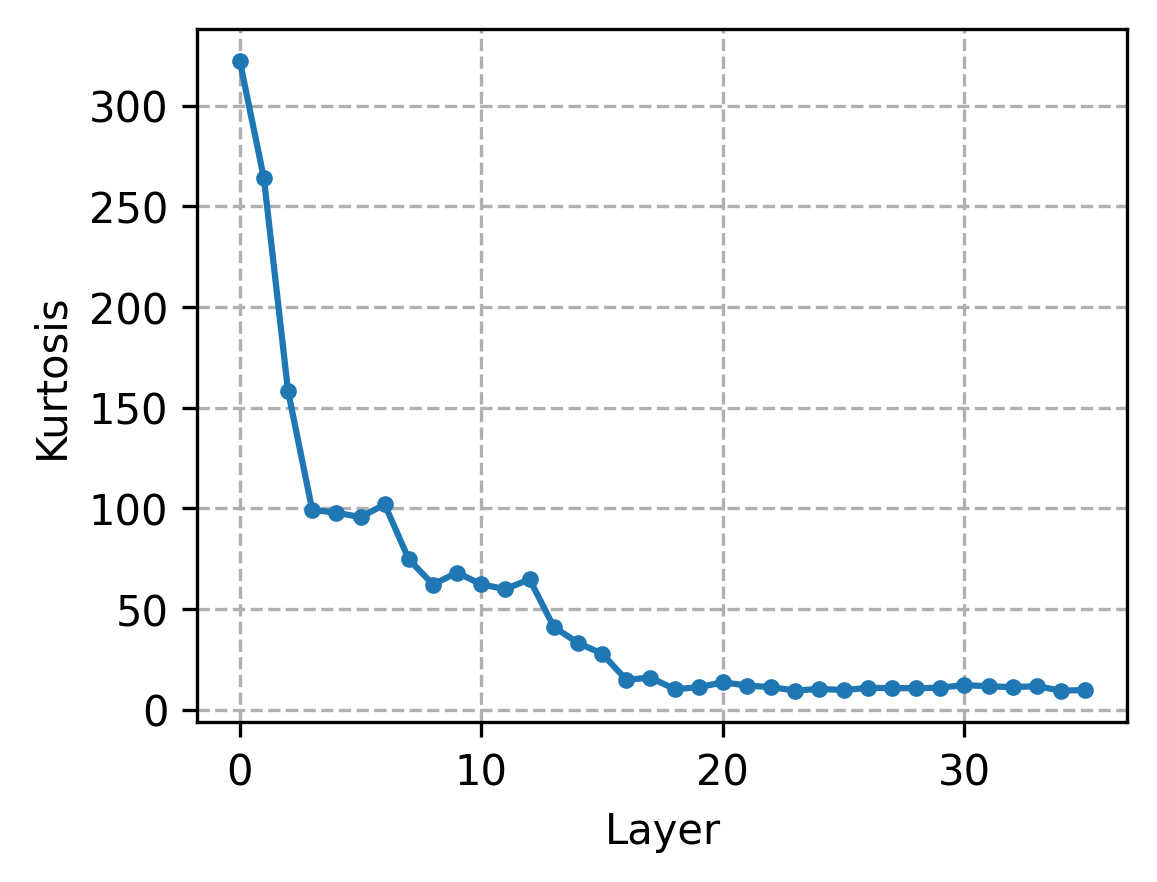}
    }
    \hfill
    \subfigure[\(m=96\)]{\includegraphics[width=0.22\textwidth]{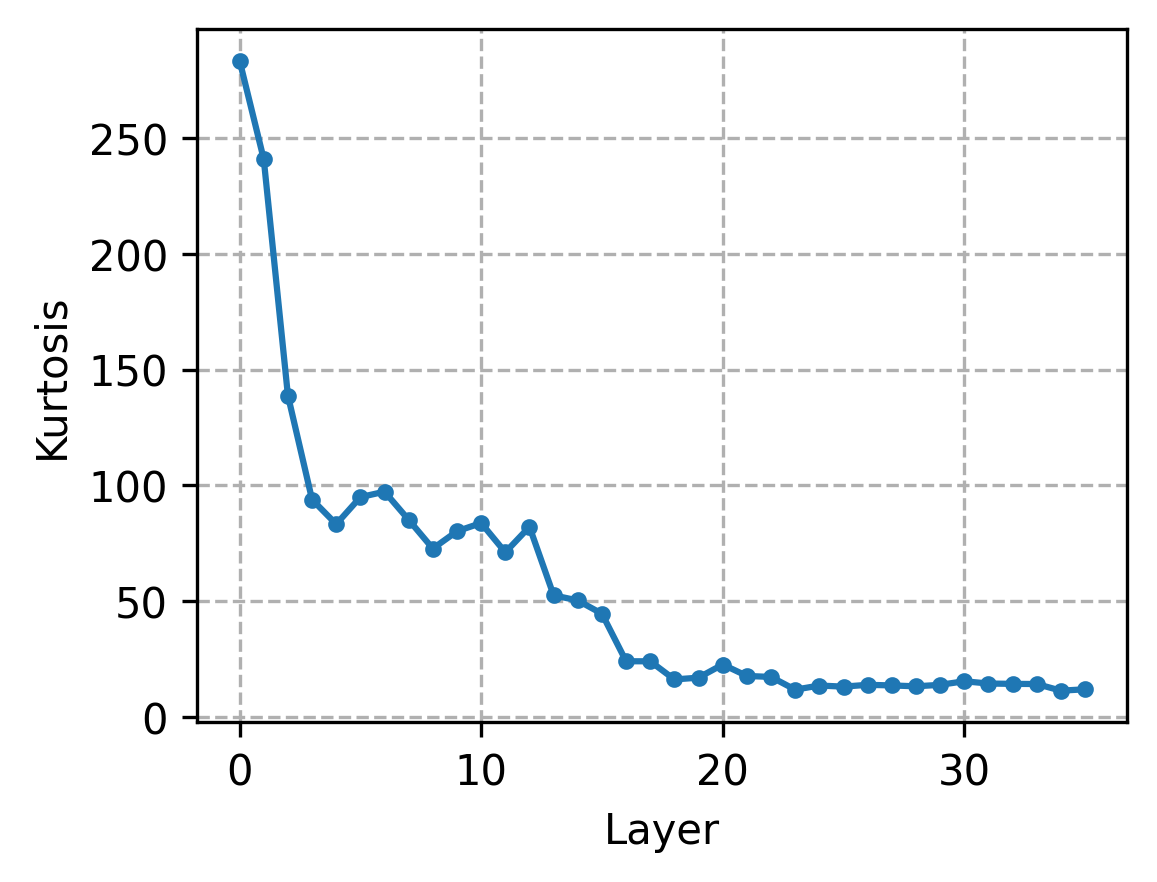}
    }
    \hfill
    \subfigure[\(m=112\)]{\includegraphics[width=0.22\textwidth]{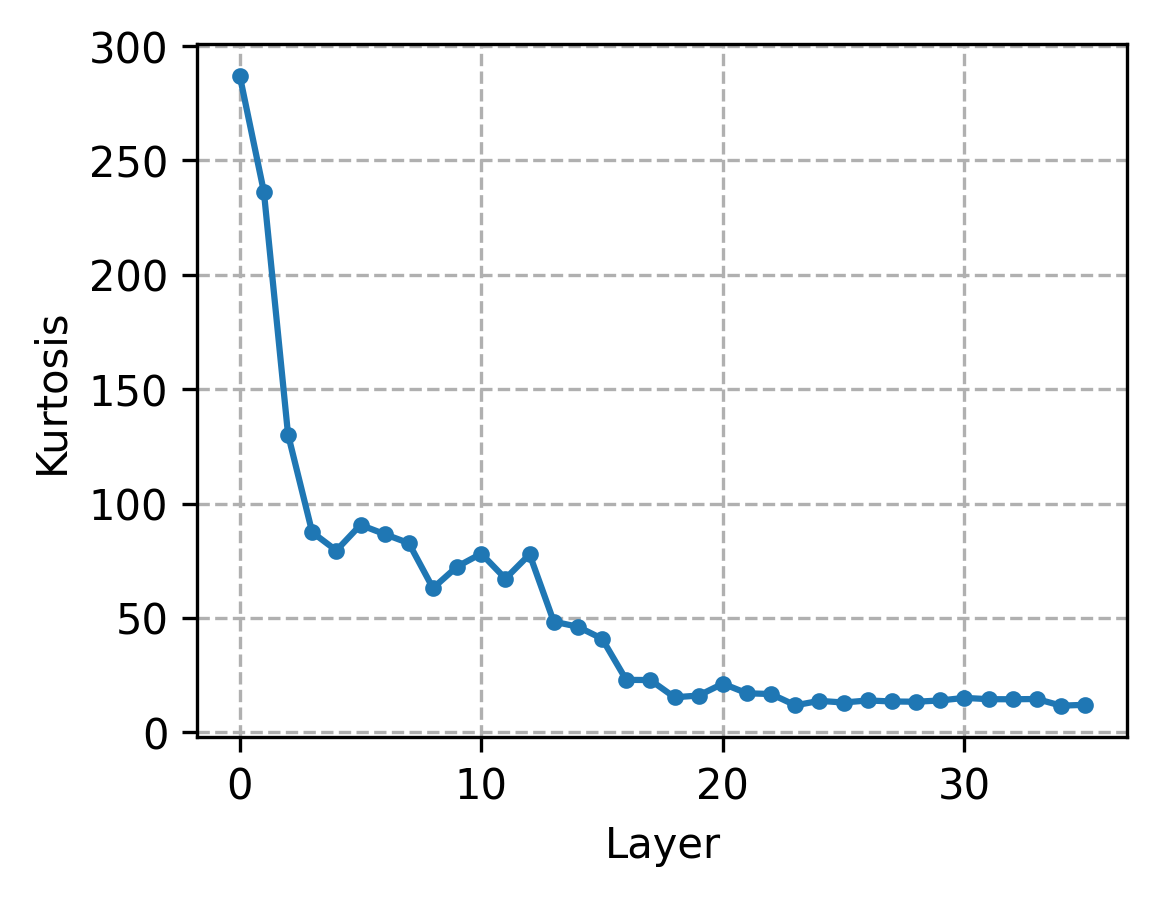}
    }
    \hfill
    \subfigure[\(m=128\)]{\includegraphics[width=0.22\textwidth]{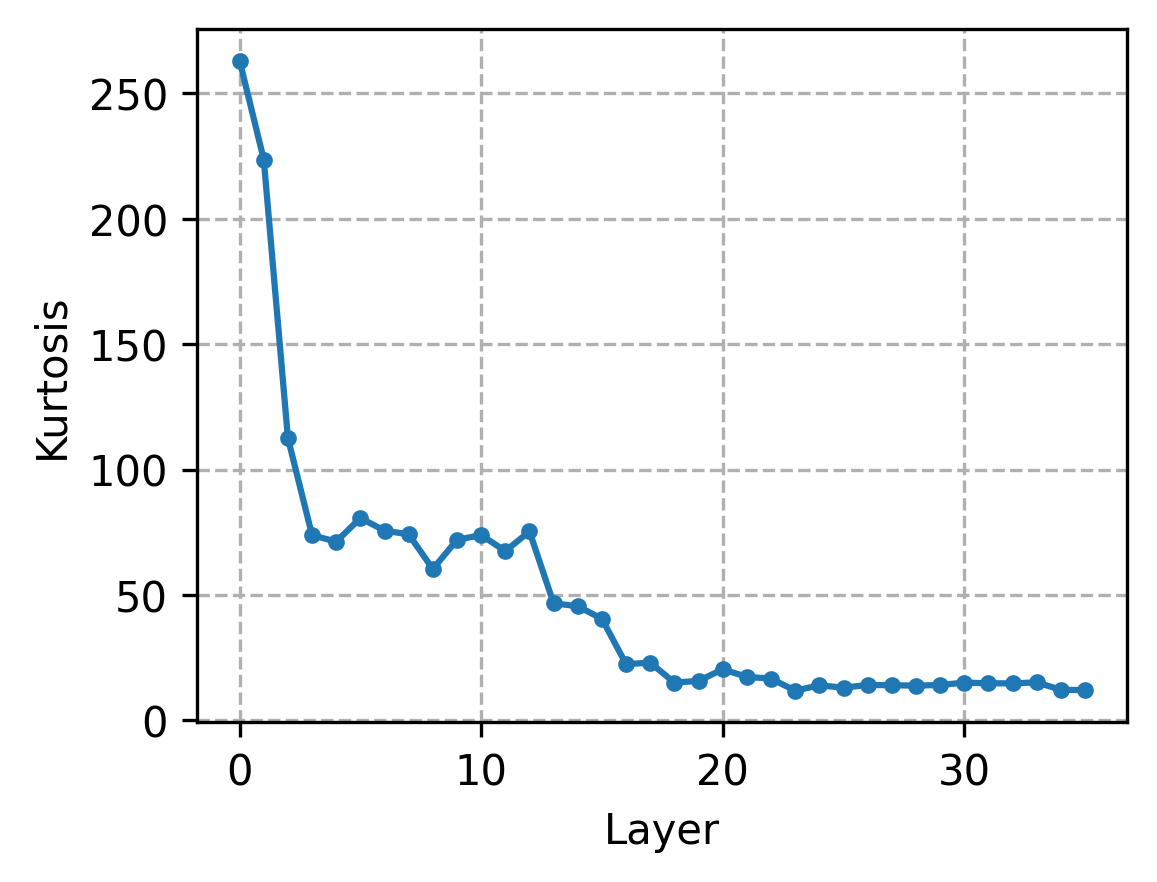}
    }
  \caption{In GPT2-Large, the superposition distribution converges before \( m = 128 \).}
  \label{fig:superposition_converges_gpt2-large}
\end{figure*}

\begin{figure*}
    \centering
    \subfigure[\(m=16\)]{\includegraphics[width=0.22\textwidth]{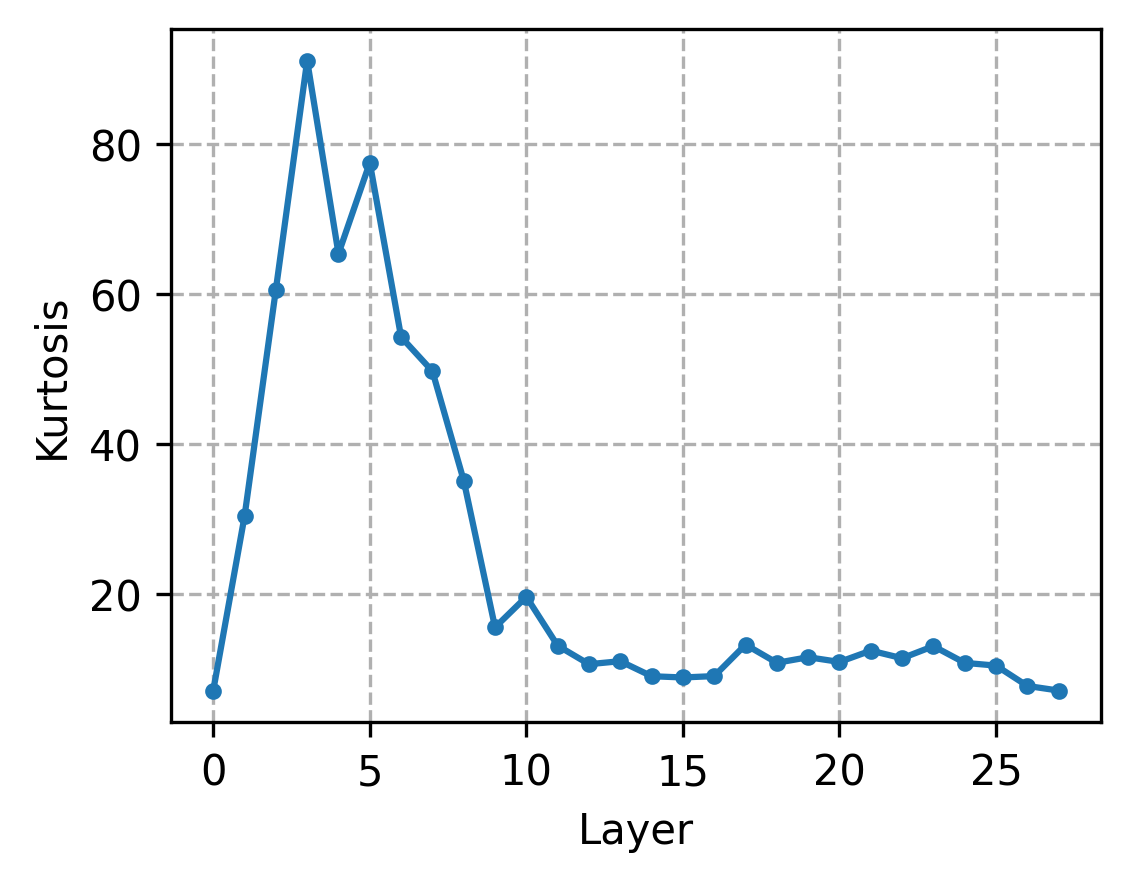}
    }
    \hfill
    \subfigure[\(m=32\)]{\includegraphics[width=0.22\textwidth]{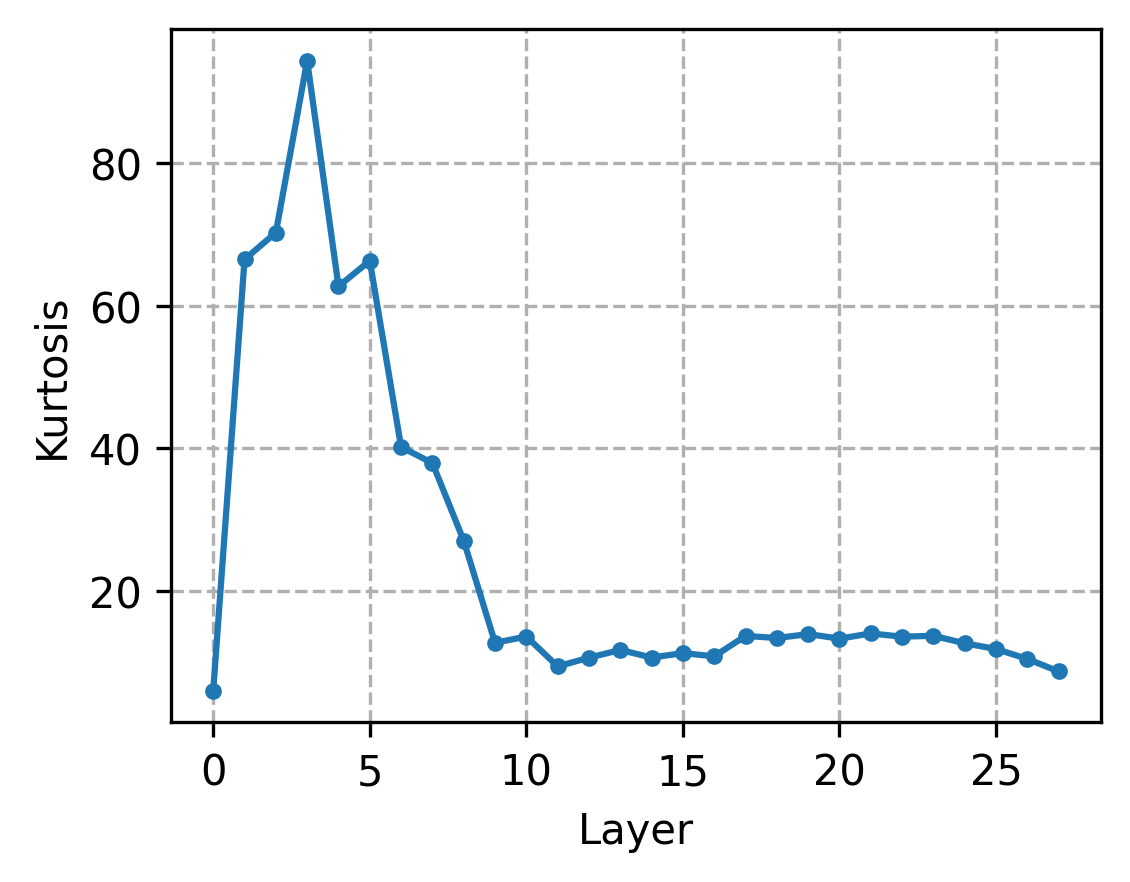}
    }
    \hfill
    \subfigure[\(m=48\)]{\includegraphics[width=0.22\textwidth]{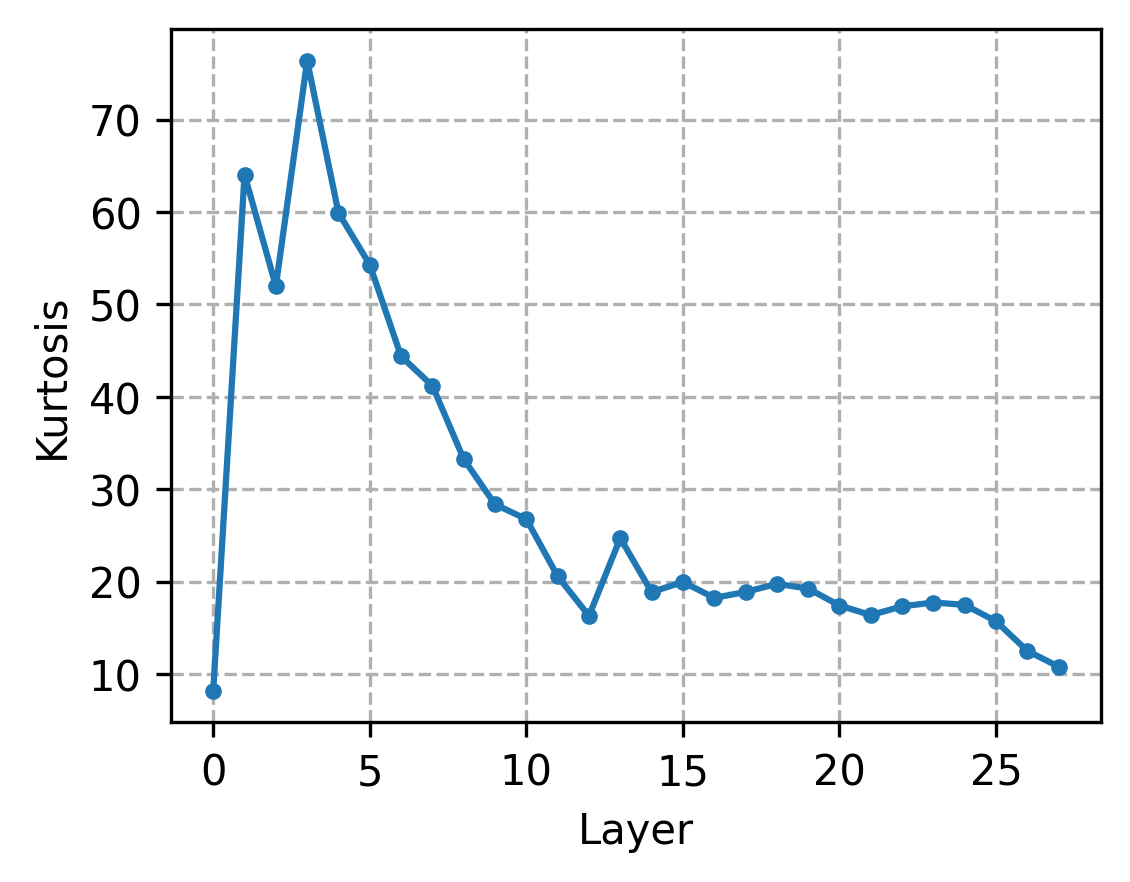}
    }
    \hfill
    \subfigure[\(m=64\)]{\includegraphics[width=0.22\textwidth]{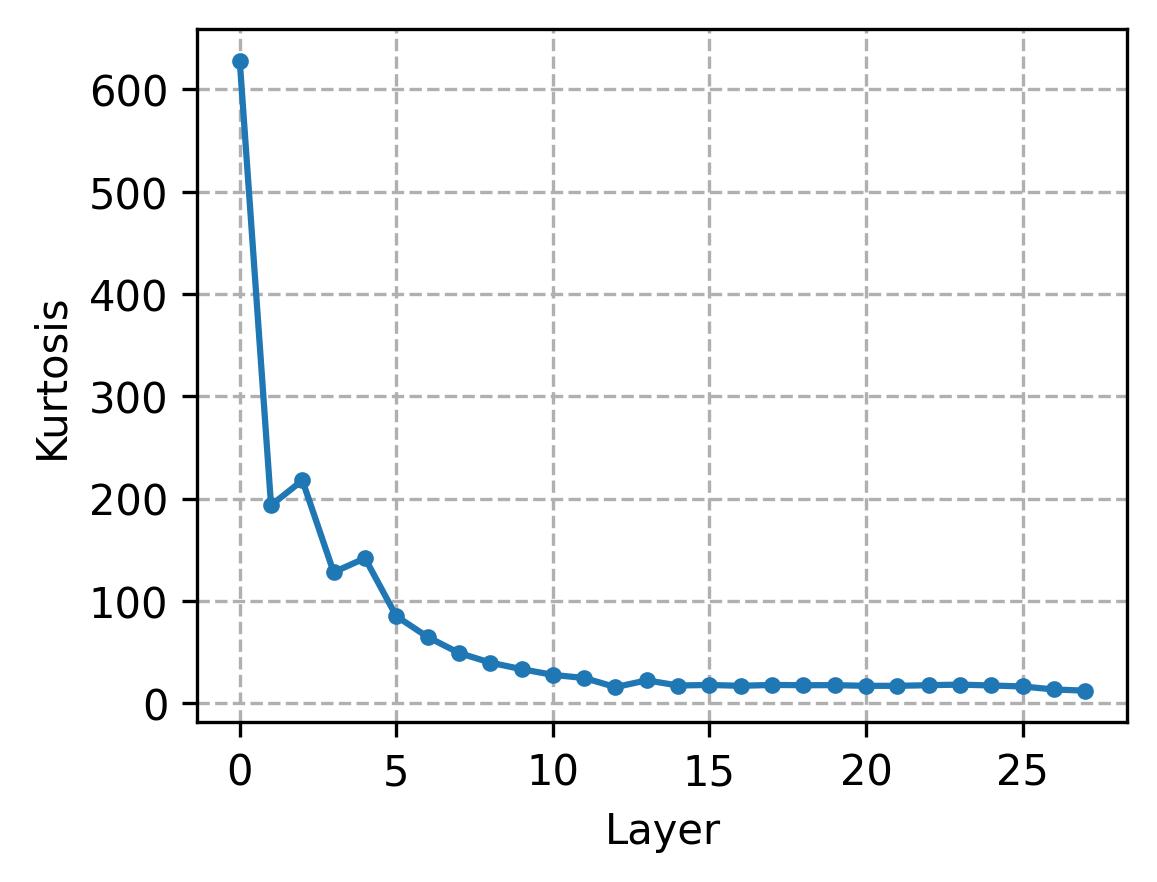}
    }
    \hfill
    \subfigure[\(m=80\)]{\includegraphics[width=0.22\textwidth]{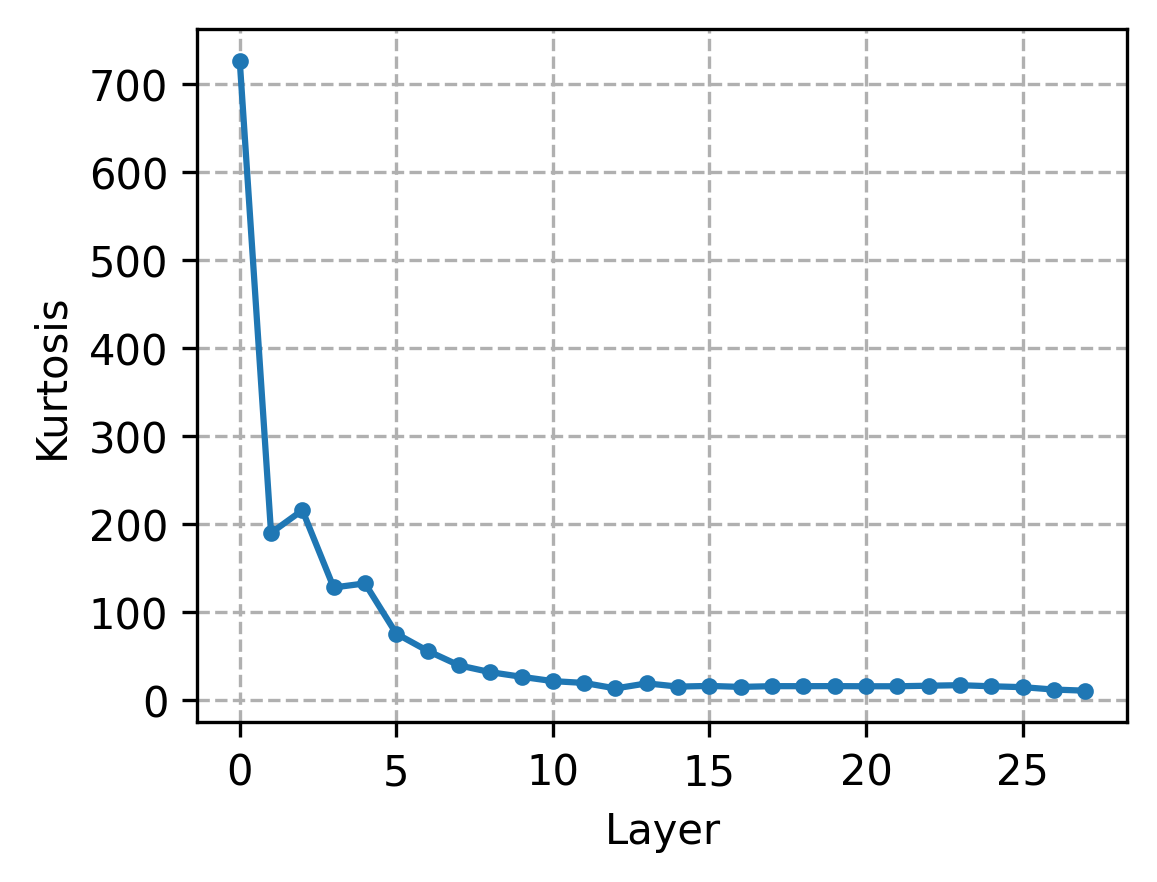}
    }
    \hfill
    \subfigure[\(m=96\)]{\includegraphics[width=0.22\textwidth]{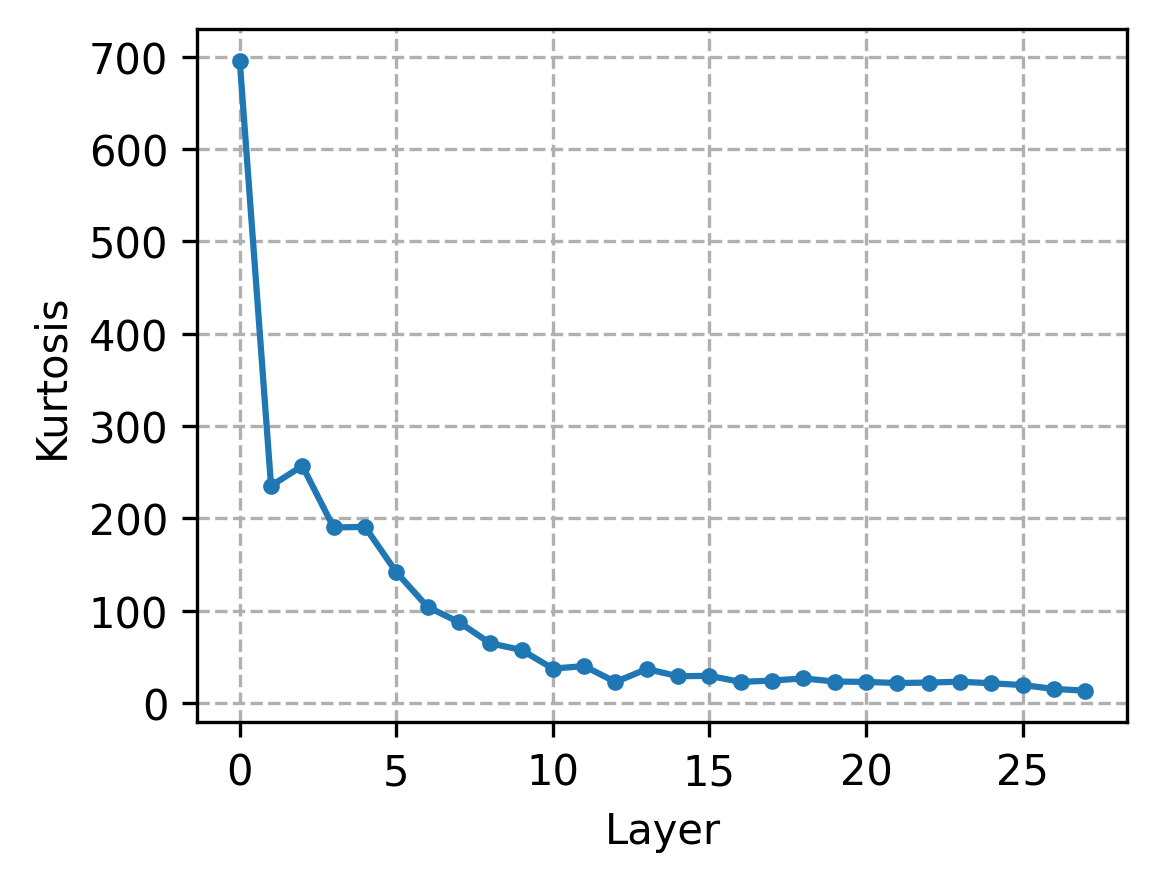}
    }
    \hfill
    \subfigure[\(m=112\)]{\includegraphics[width=0.22\textwidth]{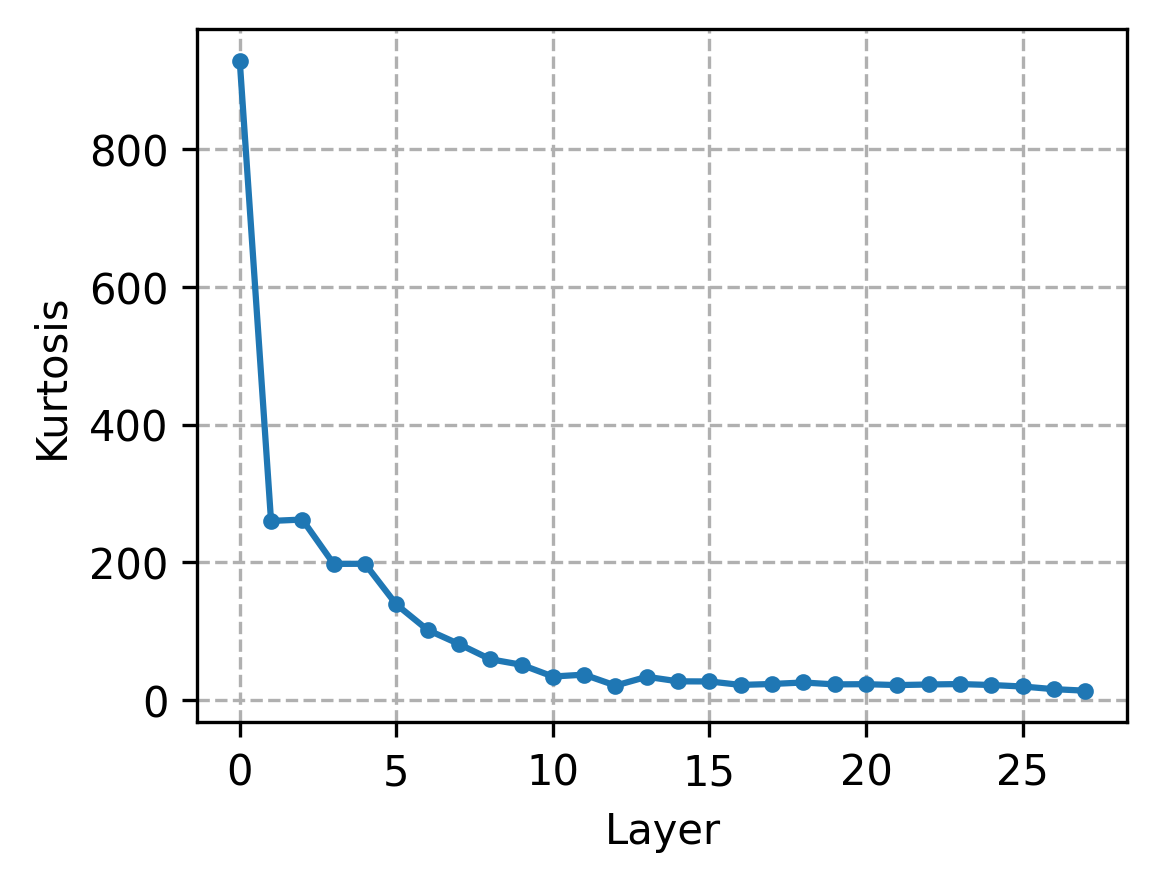}
    }
    \hfill
    \subfigure[\(m=128\)]{\includegraphics[width=0.22\textwidth]{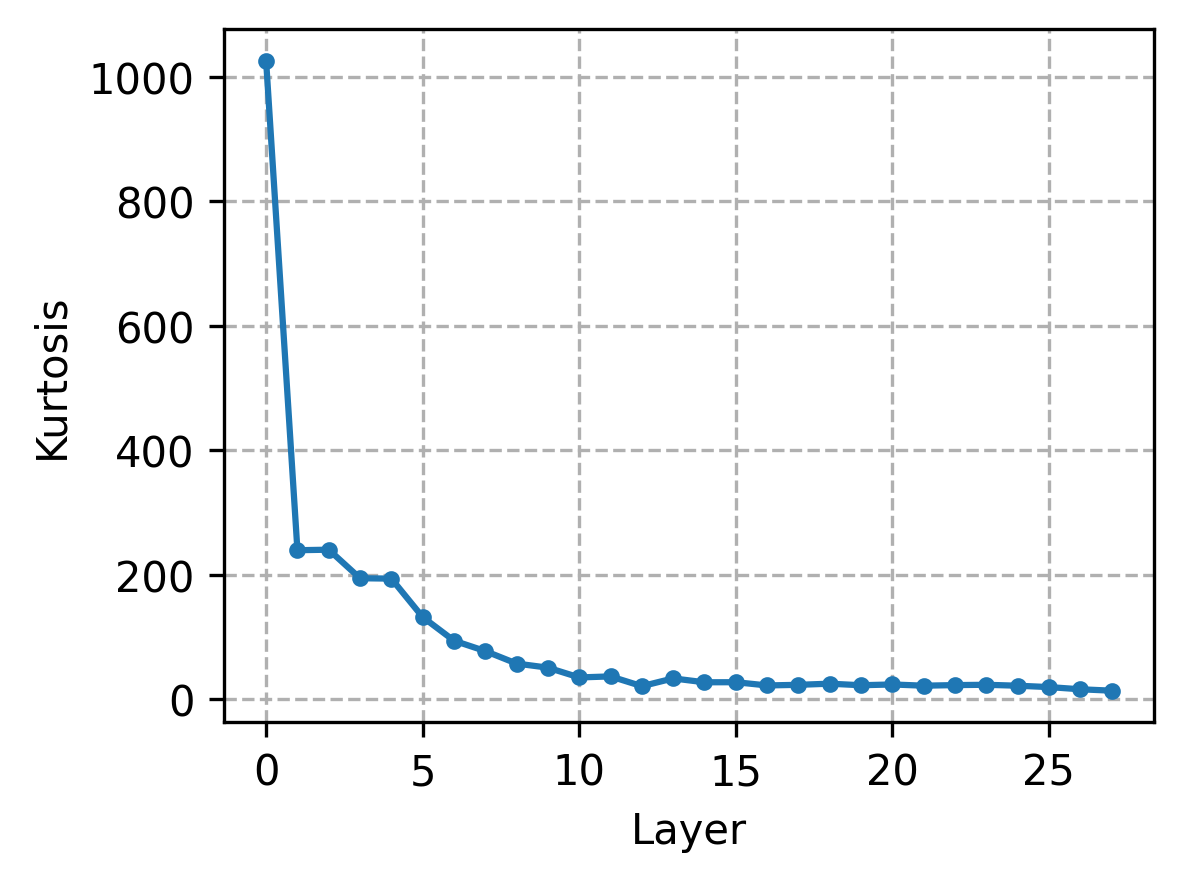}
    }
  \caption{In GPT-J-6B, the superposition distribution converges before \( m = 128 \).}
  \label{fig:superposition_converges_gpt-j-6b}
\end{figure*}

\begin{figure*}
    \centering
    \subfigure[\(m=16\)]{\includegraphics[width=0.22\textwidth]{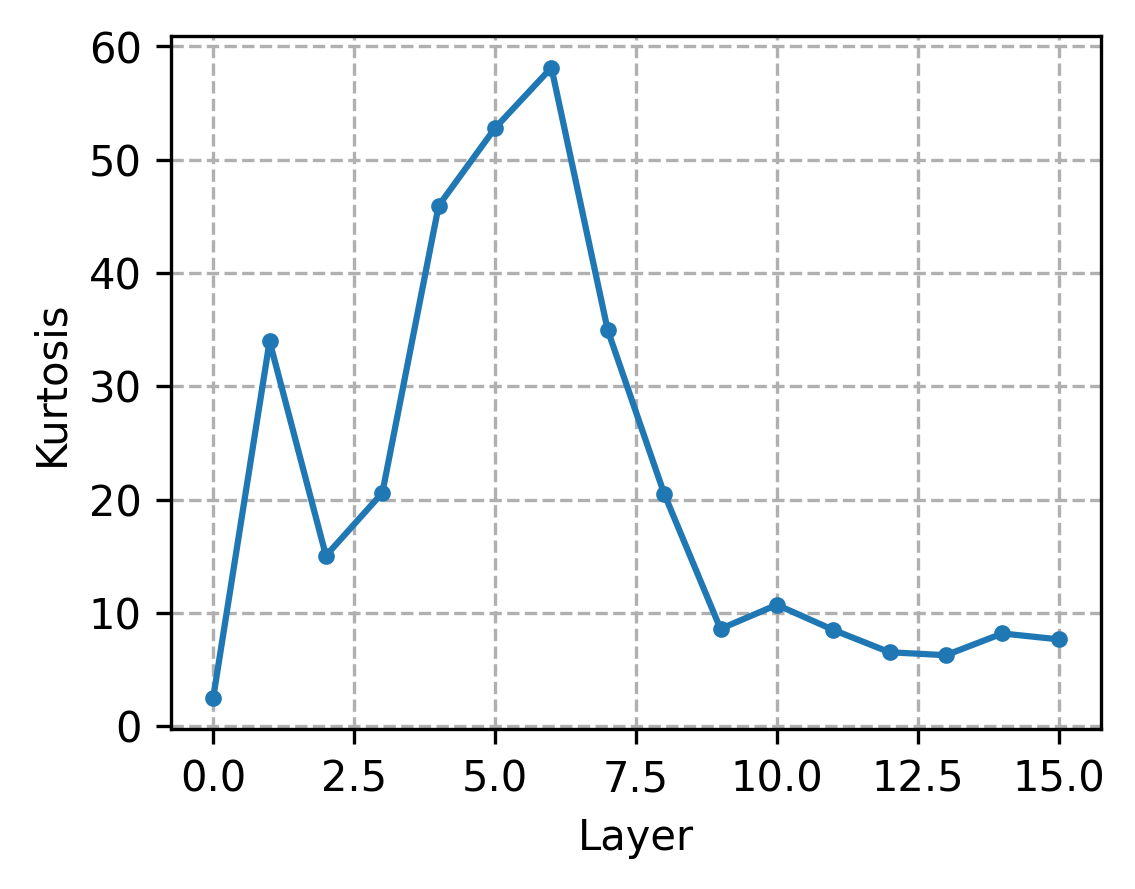}
    }
    \hfill
    \subfigure[\(m=32\)]{\includegraphics[width=0.22\textwidth]{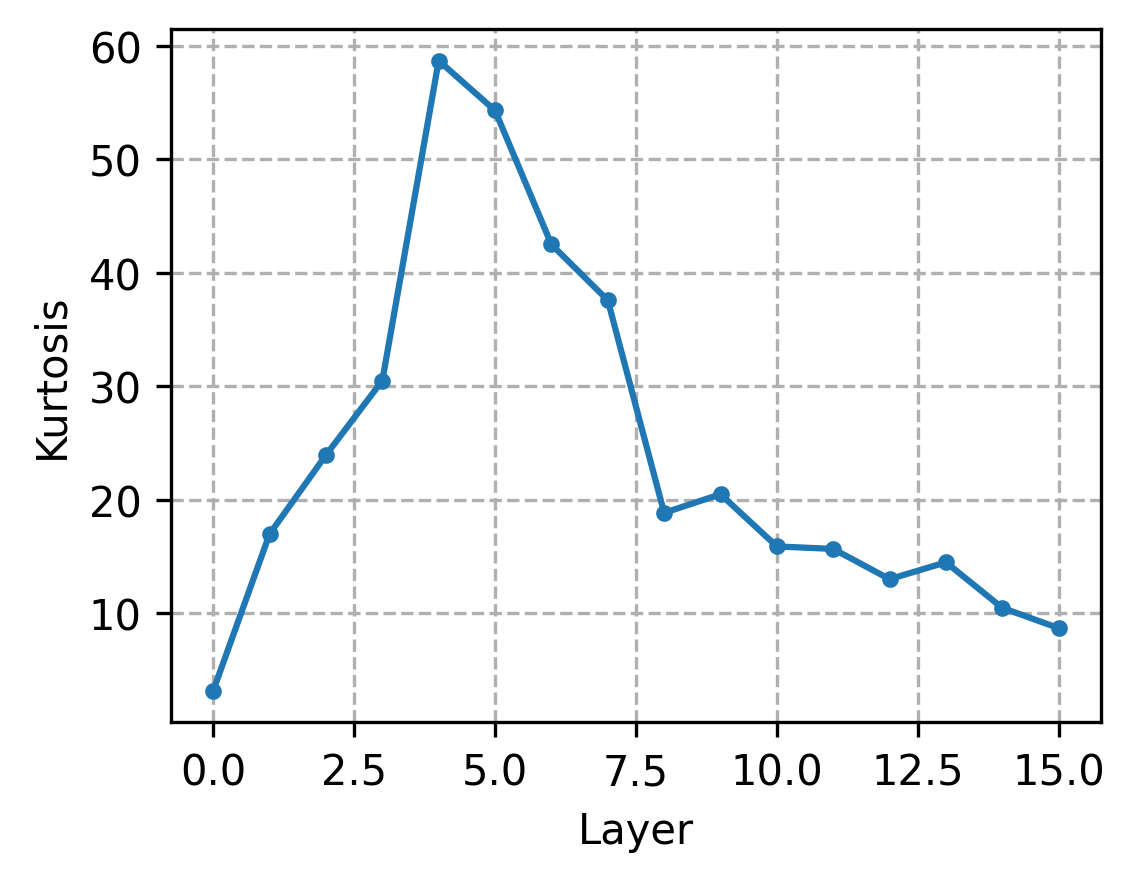}
    }
    \hfill
    \subfigure[\(m=48\)]{\includegraphics[width=0.22\textwidth]{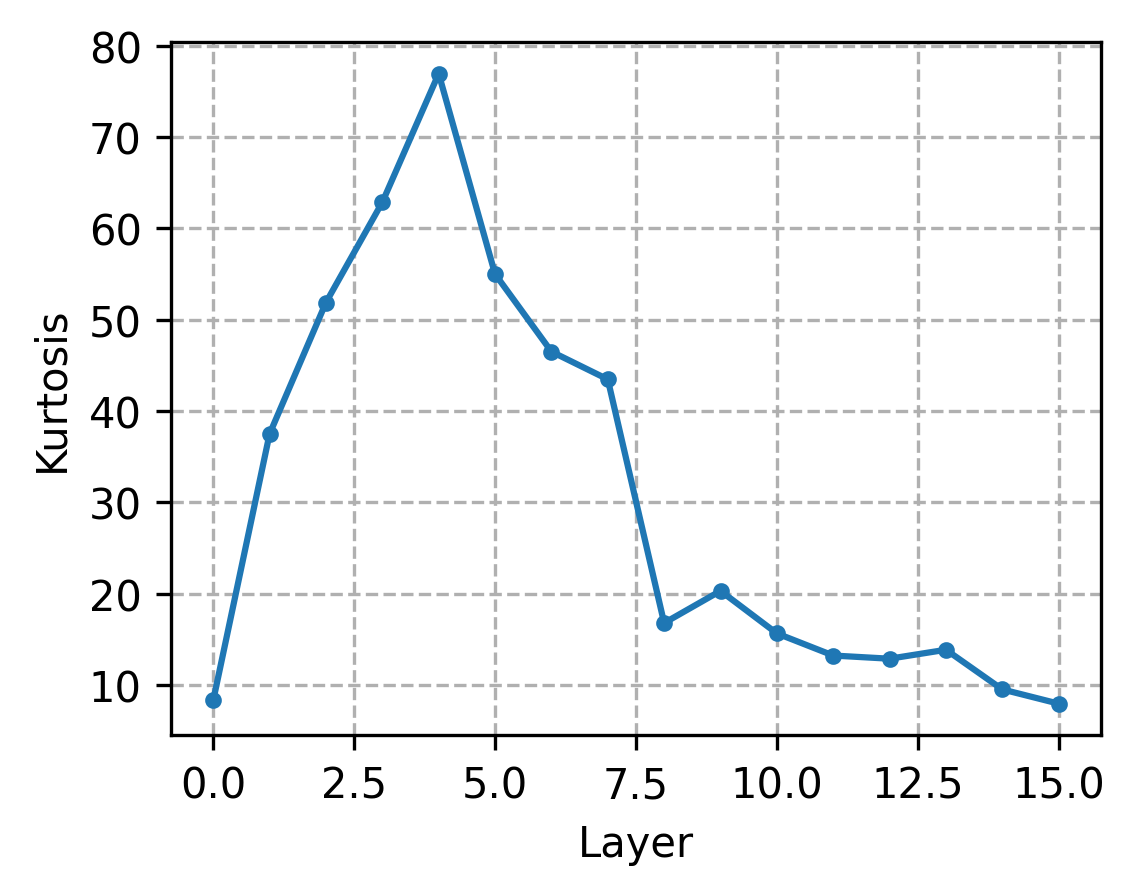}
    }
    \hfill
    \subfigure[\(m=64\)]{\includegraphics[width=0.22\textwidth]{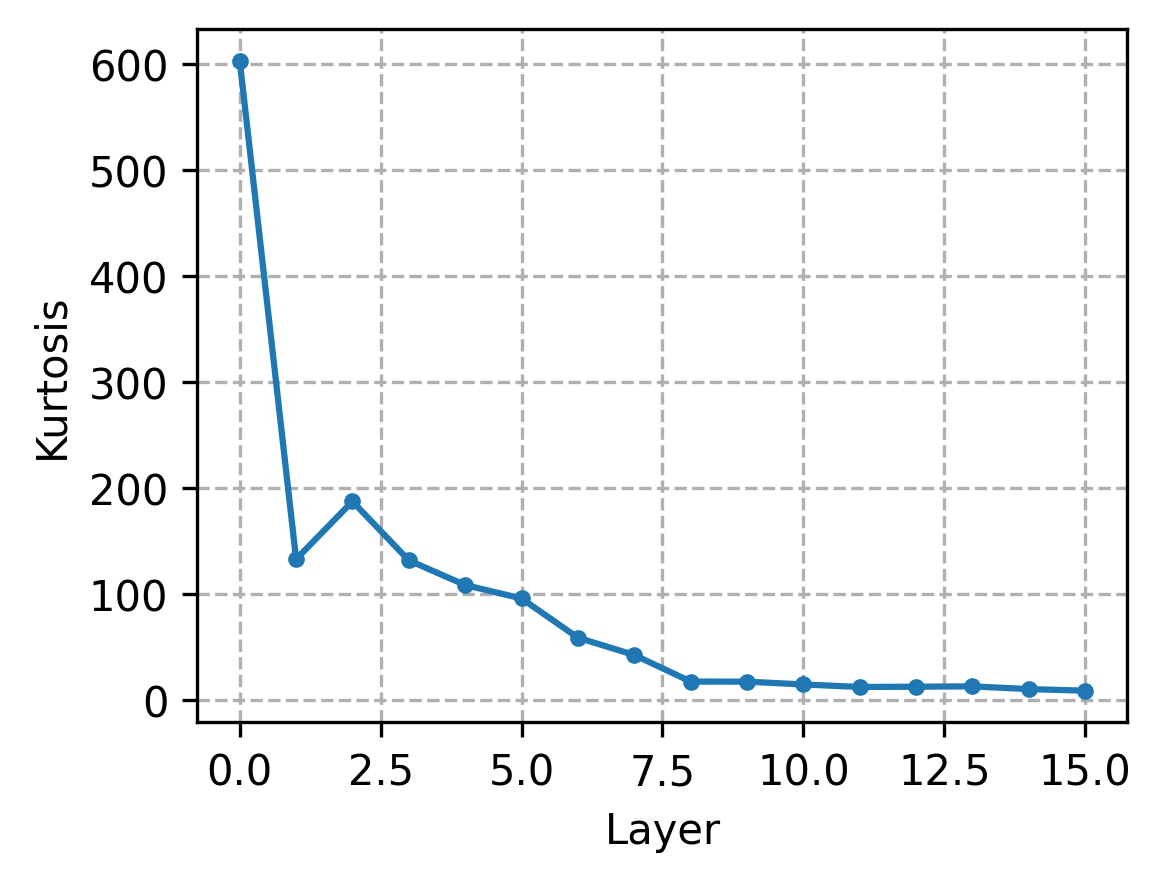}
    }
    \hfill
    \subfigure[\(m=80\)]{\includegraphics[width=0.22\textwidth]{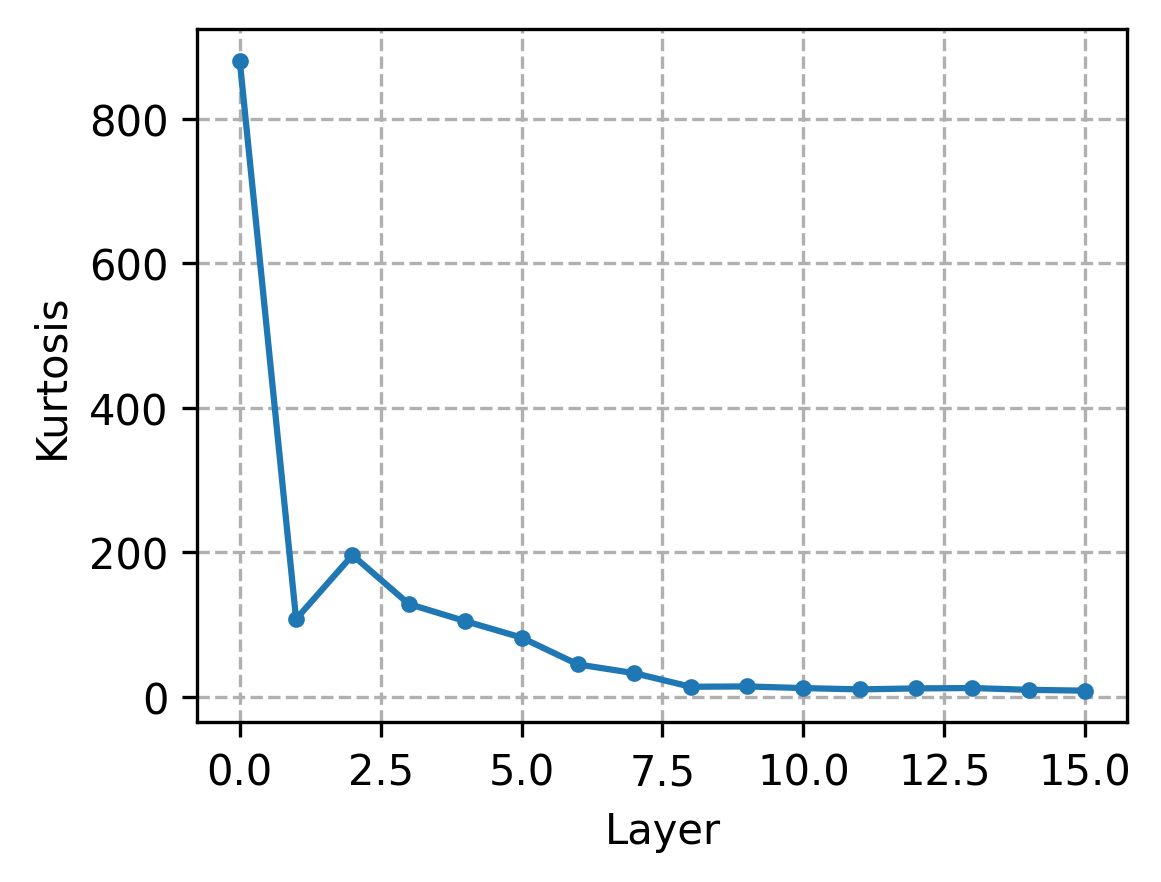}
    }
    \hfill
    \subfigure[\(m=96\)]{\includegraphics[width=0.22\textwidth]{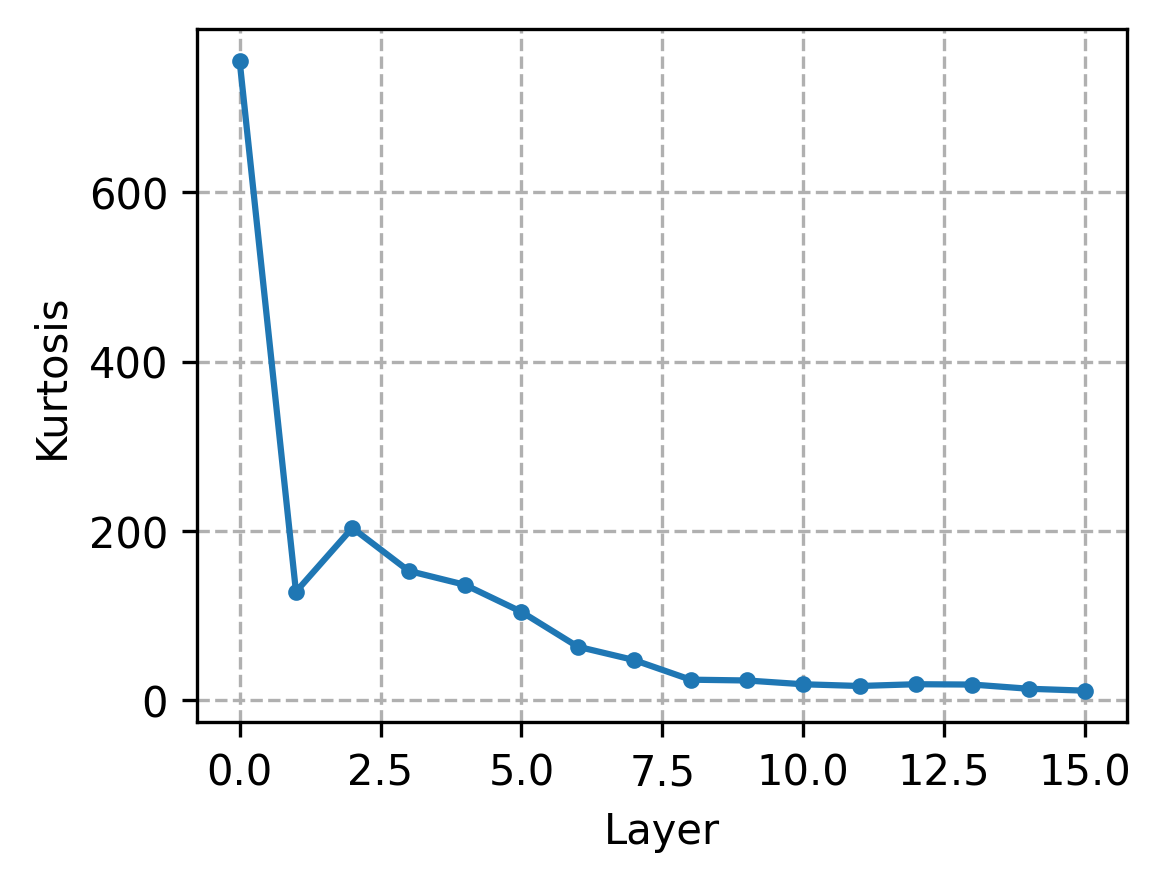}
    }
    \hfill
    \subfigure[\(m=112\)]{\includegraphics[width=0.22\textwidth]{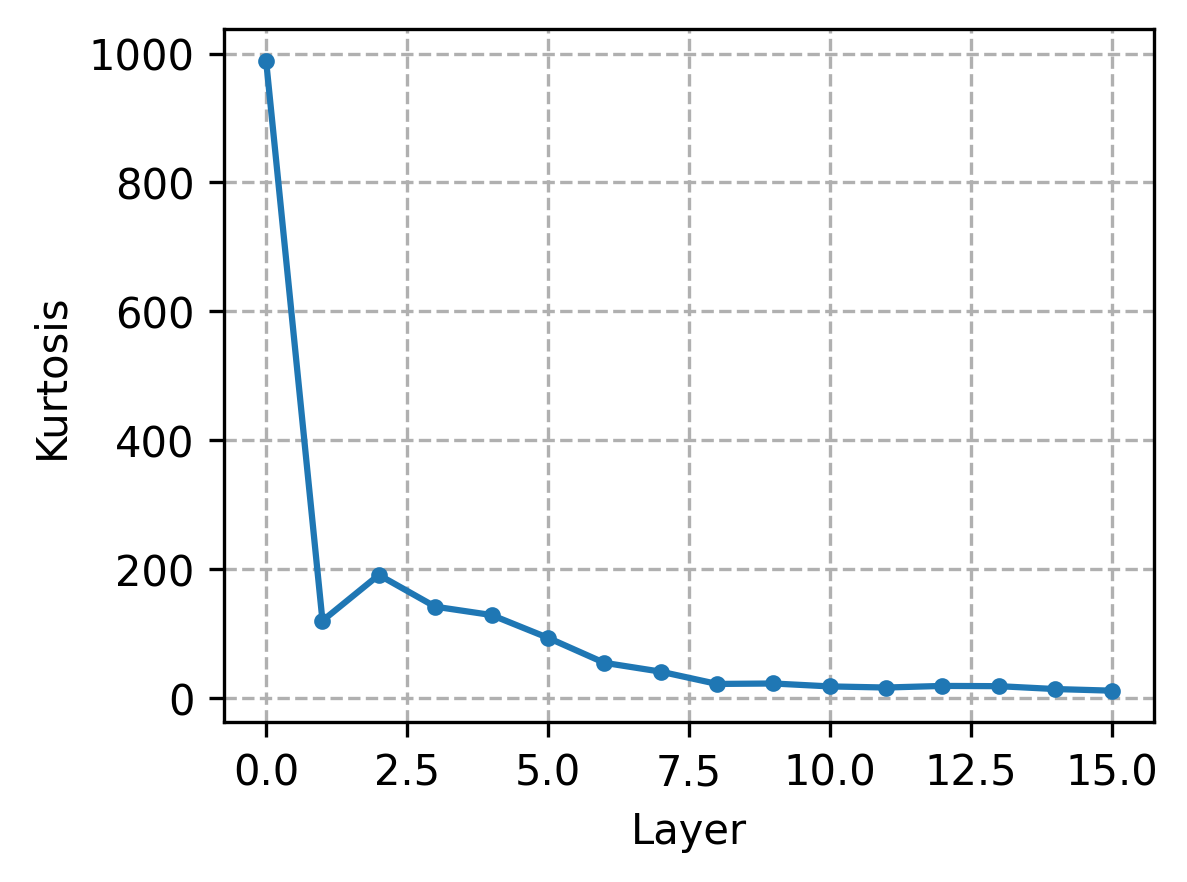}
    }
    \hfill
    \subfigure[\(m=128\)]{\includegraphics[width=0.22\textwidth]{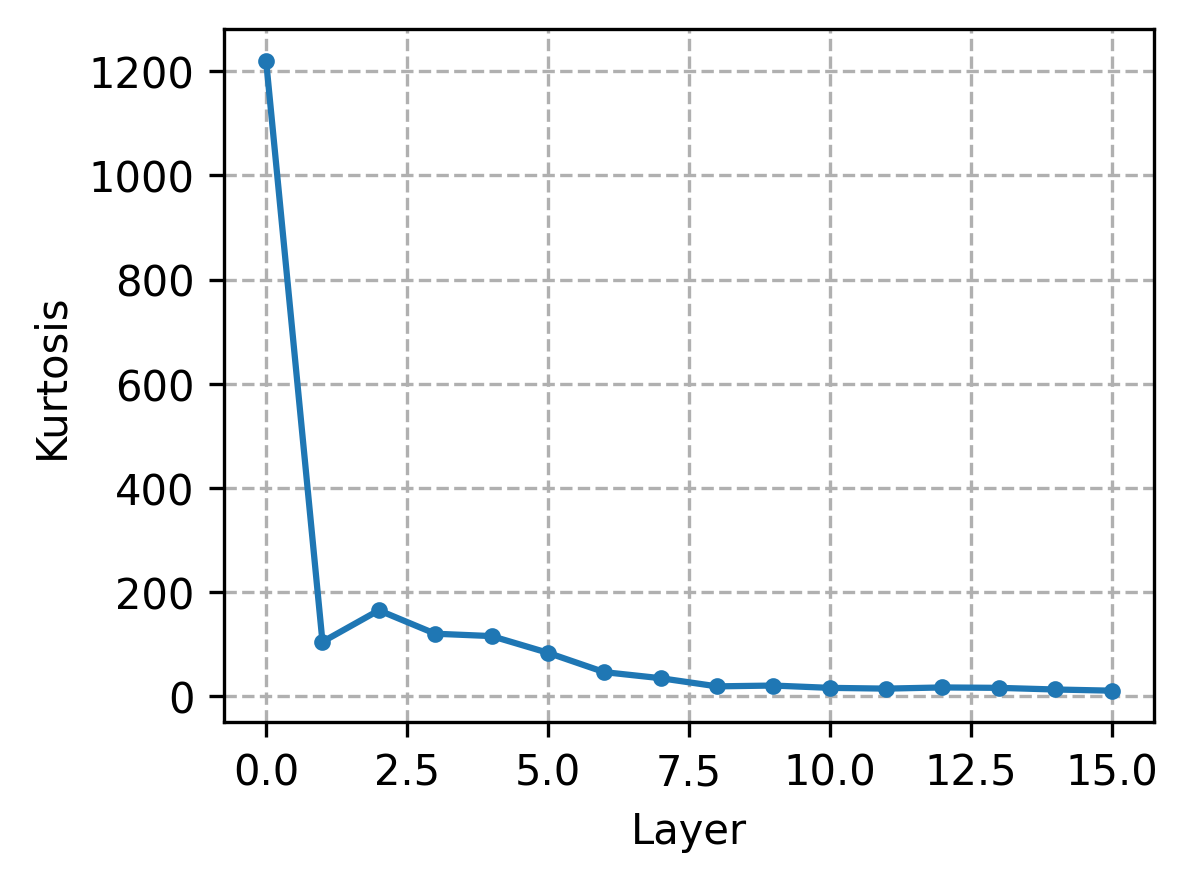}
    }
  \caption{In Pythia-1B, the superposition distribution converges before \( m = 128 \).}
  \label{fig:superposition_converges_pythia-1b}
\end{figure*}

\begin{figure*}
    \centering
    \subfigure[\(m=16\)]{\includegraphics[width=0.22\textwidth]{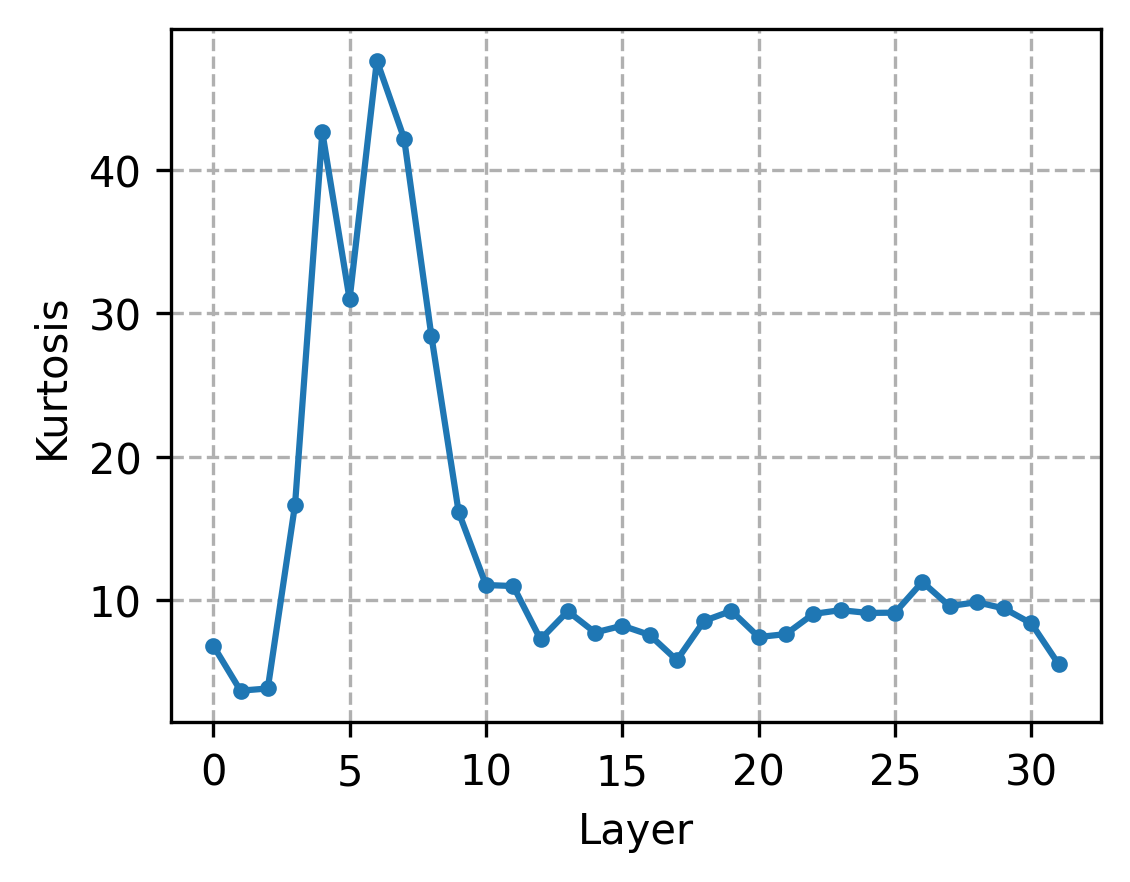}
    }
    \hfill
    \subfigure[\(m=32\)]{\includegraphics[width=0.22\textwidth]{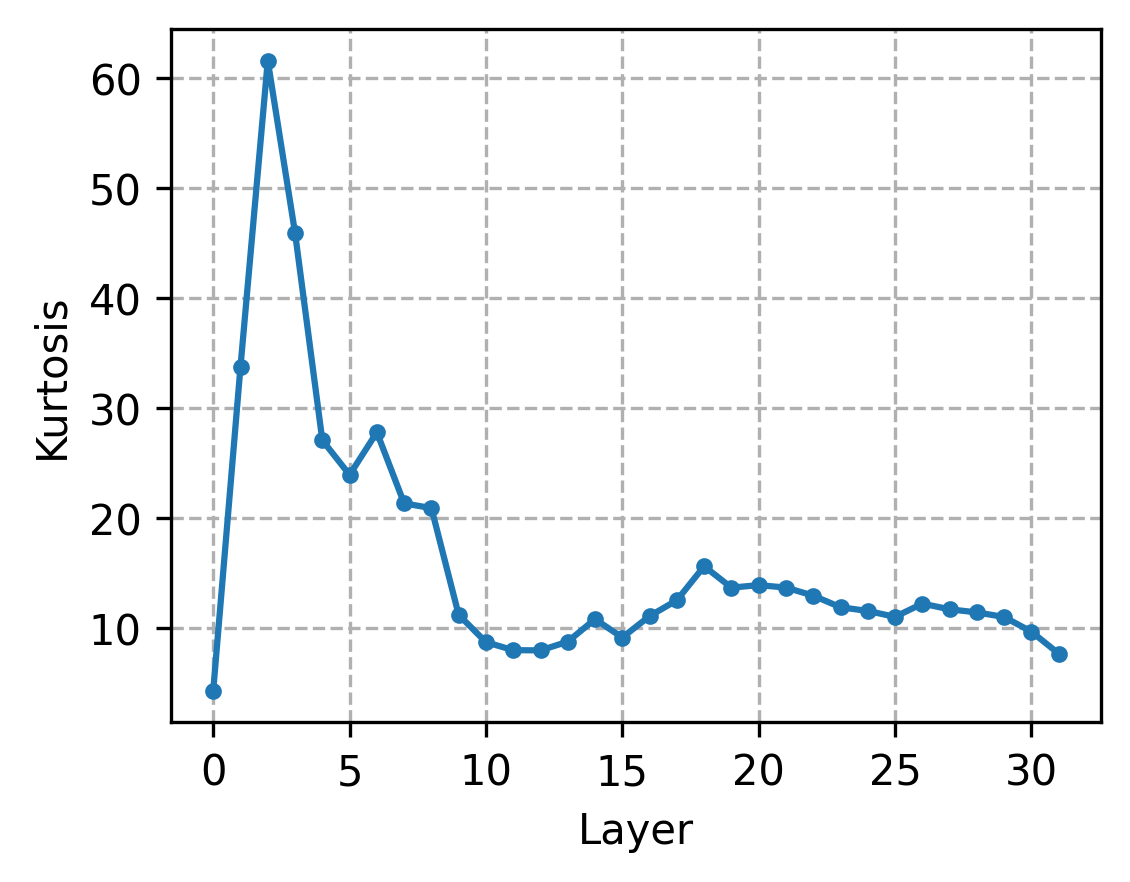}
    }
    \hfill
    \subfigure[\(m=48\)]{\includegraphics[width=0.22\textwidth]{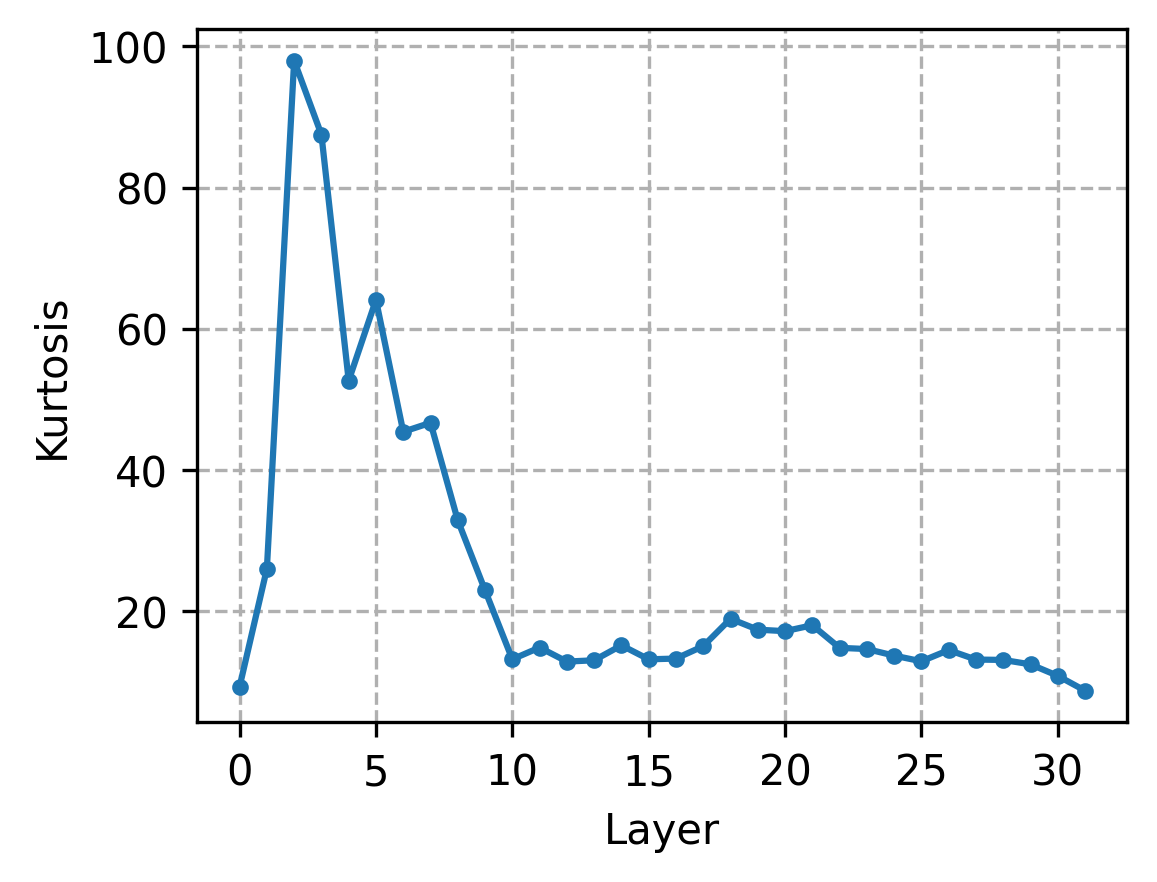}
    }
    \hfill
    \subfigure[\(m=64\)]{\includegraphics[width=0.22\textwidth]{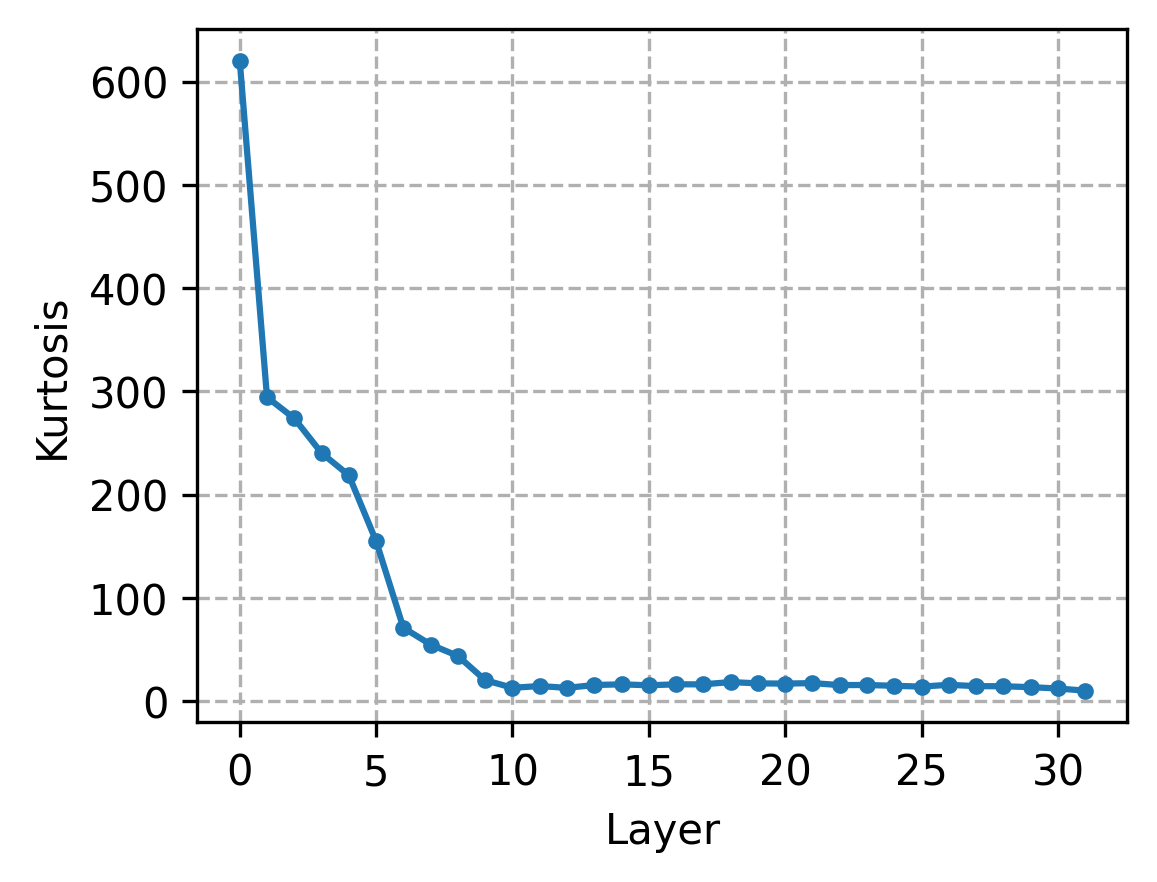}
    }
    \hfill
    \subfigure[\(m=80\)]{\includegraphics[width=0.22\textwidth]{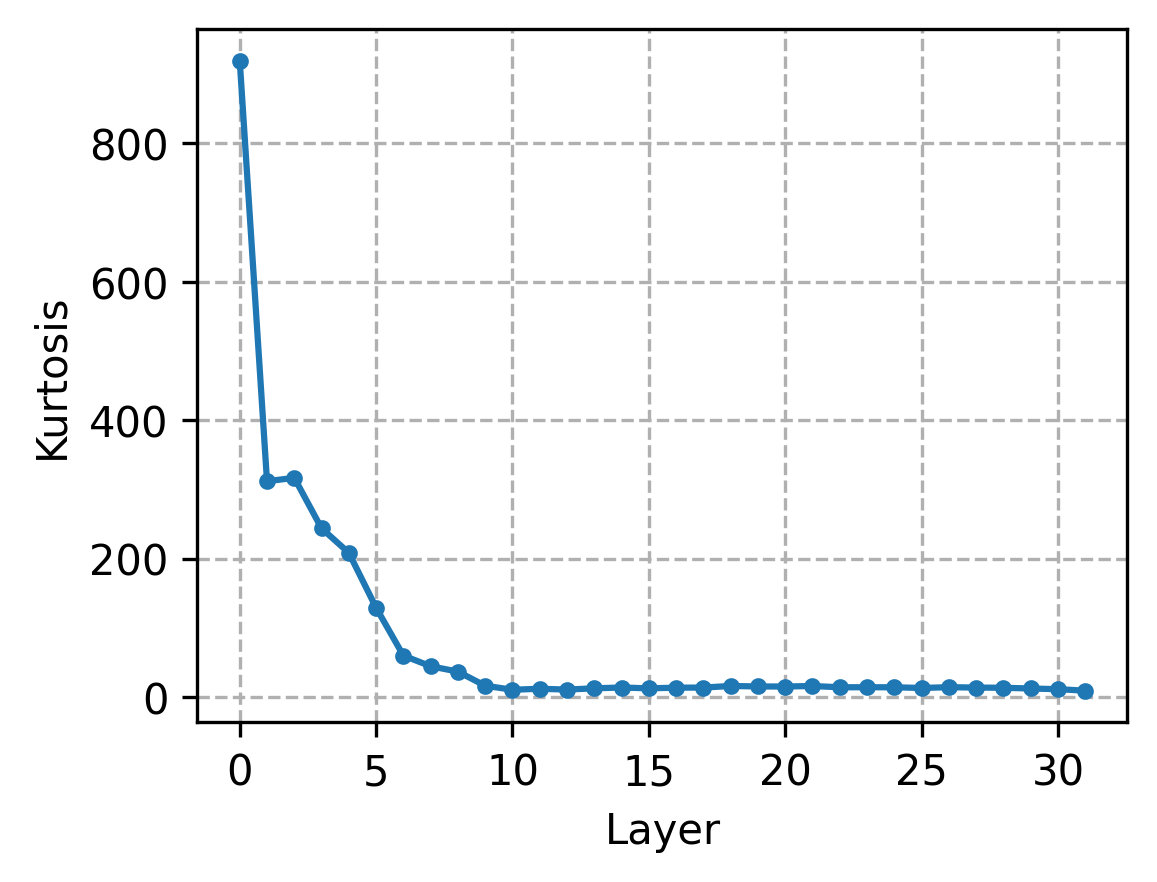}
    }
    \hfill
    \subfigure[\(m=96\)]{\includegraphics[width=0.22\textwidth]{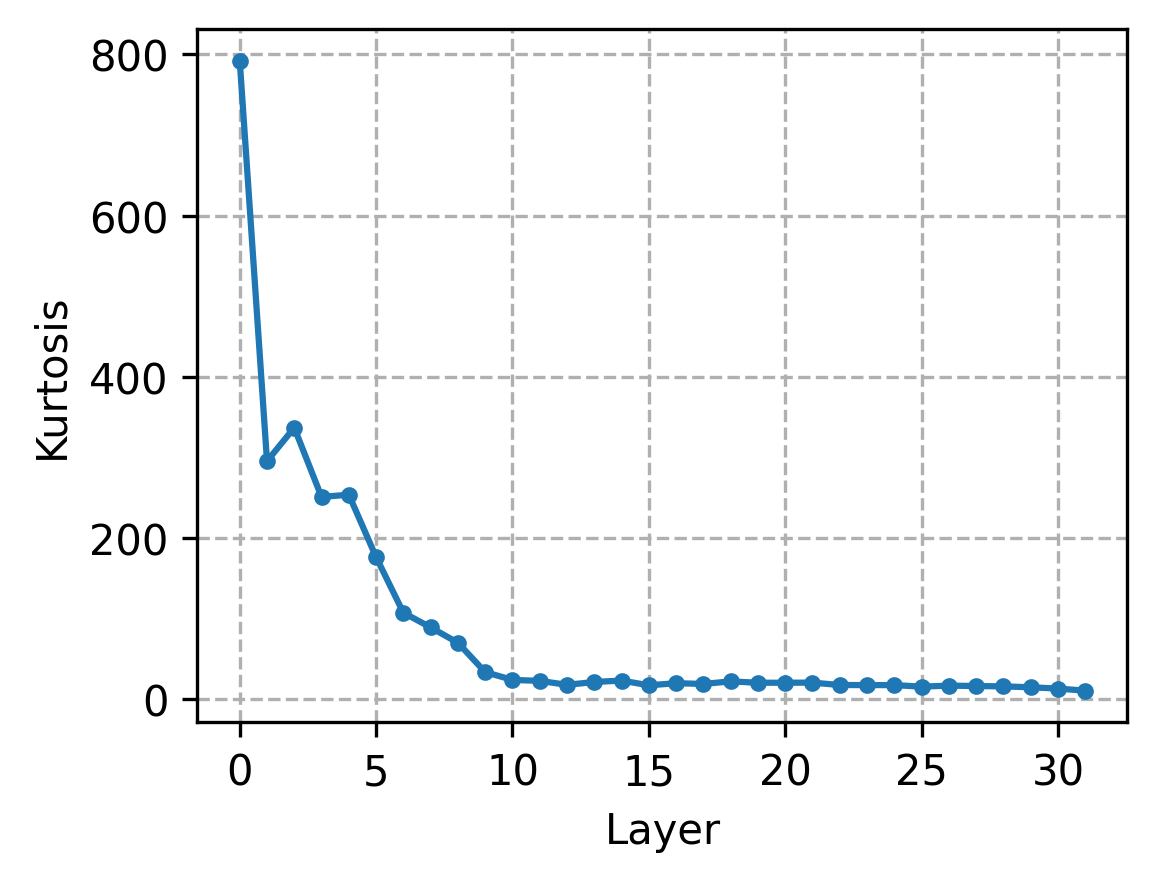}
    }
    \hfill
    \subfigure[\(m=112\)]{\includegraphics[width=0.22\textwidth]{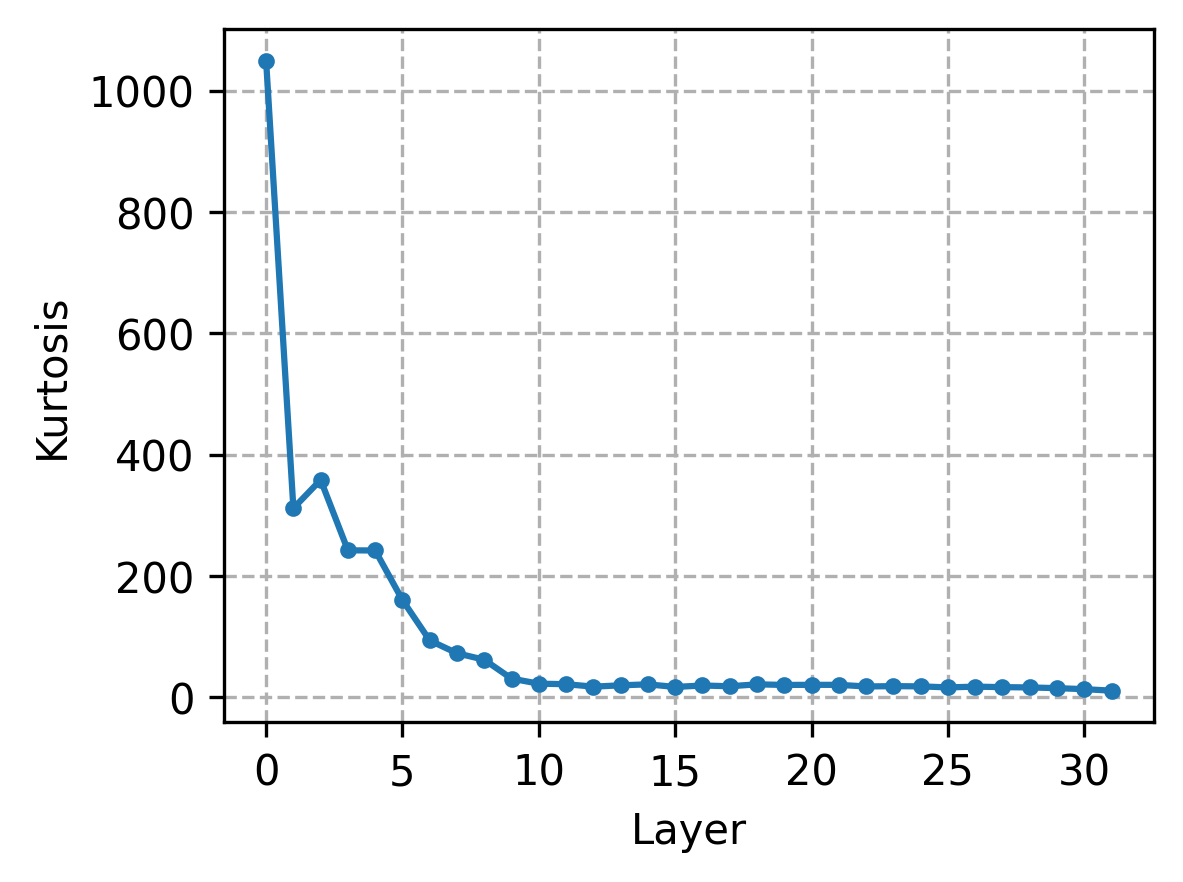}
    }
    \hfill
    \subfigure[\(m=128\)]{\includegraphics[width=0.22\textwidth]{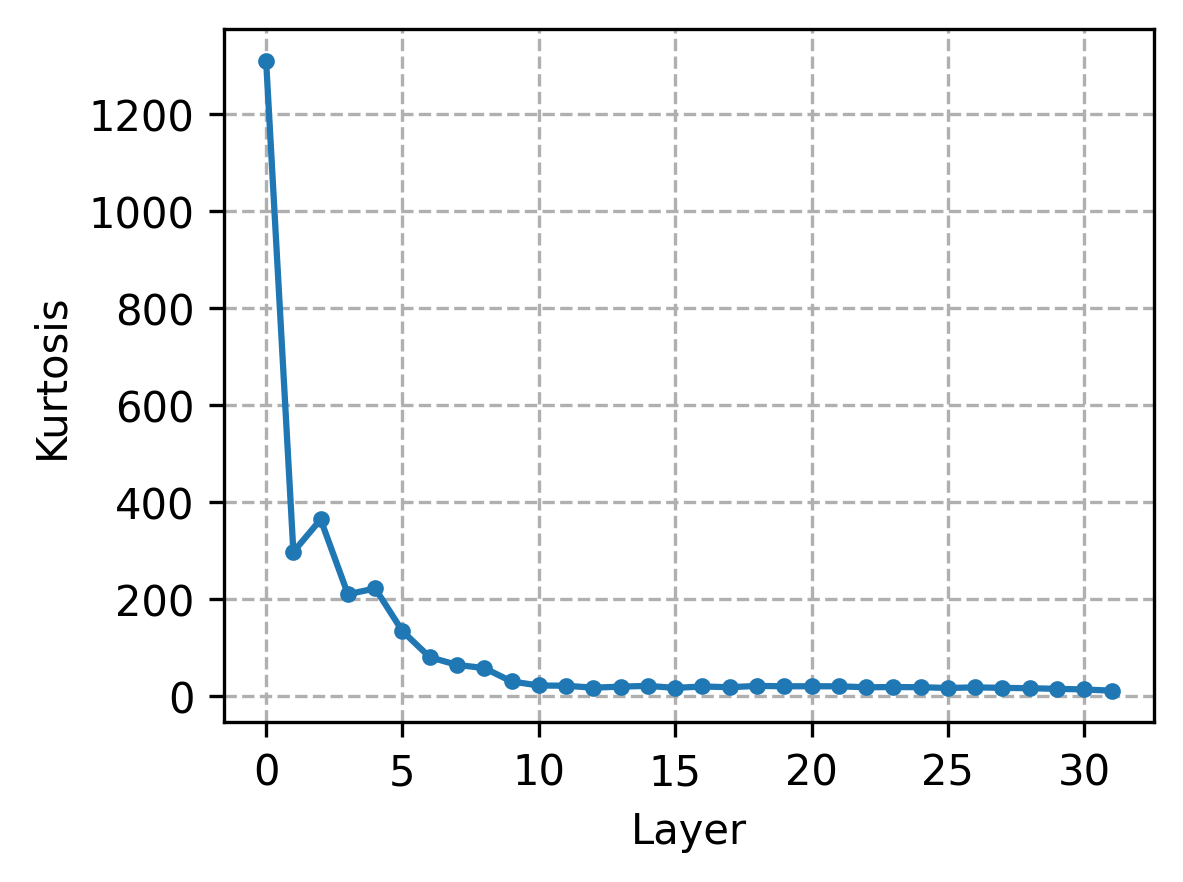}
    }
  \caption{In Pythia-2.8B, the superposition distribution converges before \( m = 128 \).}
  \label{fig:superposition_converges_pythia-2.8b}
\end{figure*}

\begin{figure*}
    \centering
    \subfigure[\(m=16\)]{\includegraphics[width=0.22\textwidth]{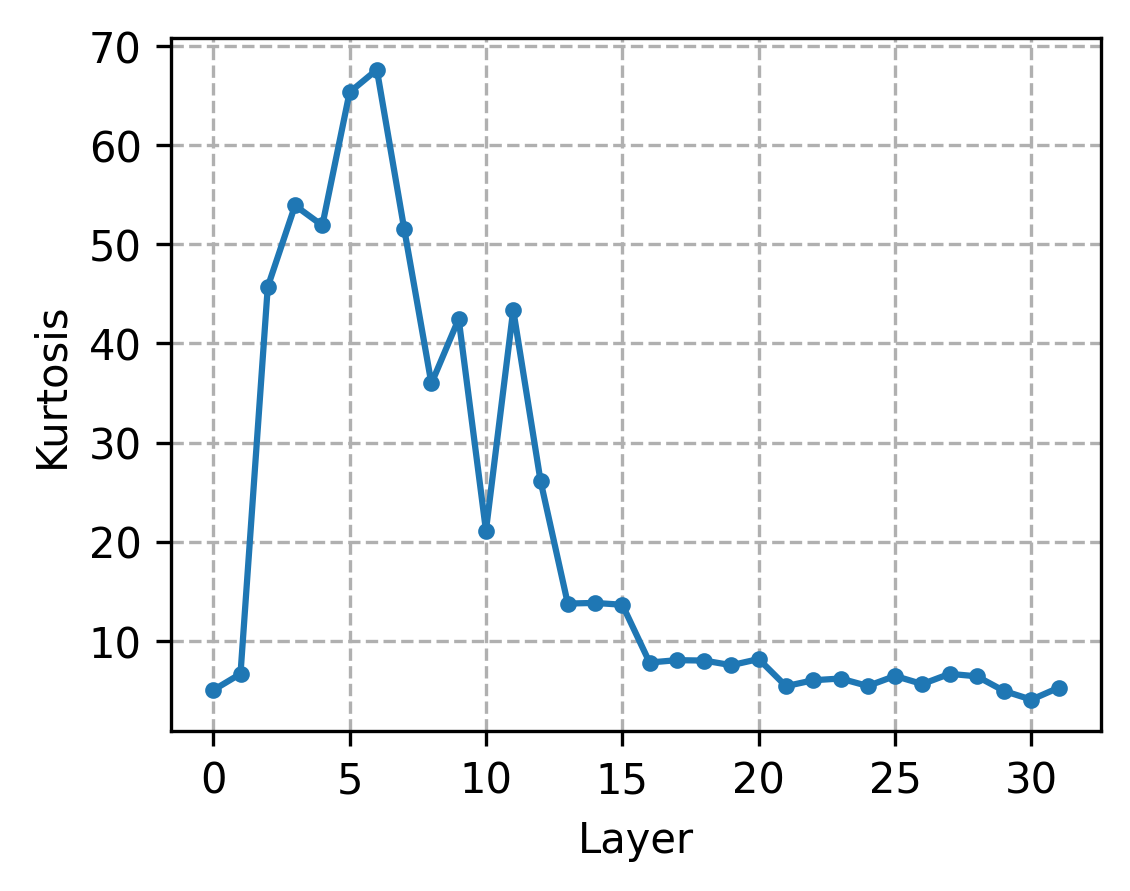}
    }
    \hfill
    \subfigure[\(m=32\)]{\includegraphics[width=0.22\textwidth]{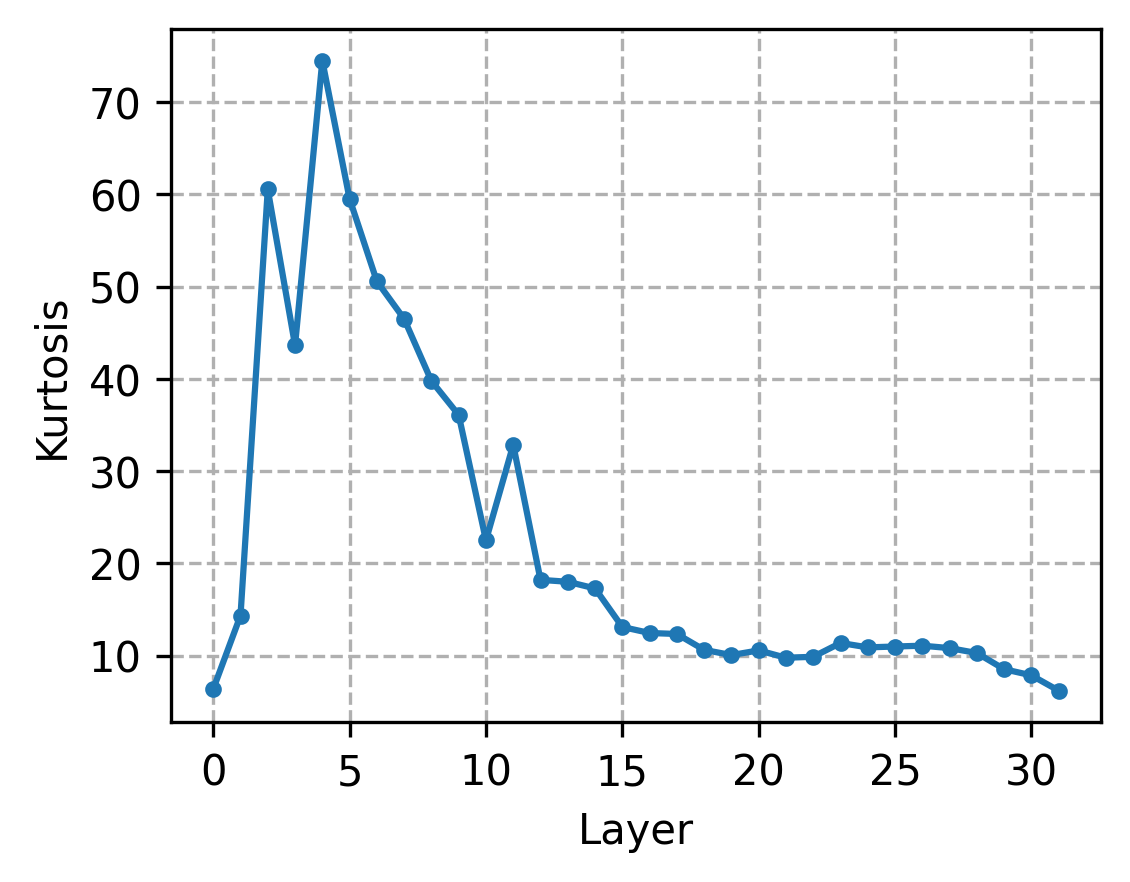}
    }
    \hfill
    \subfigure[\(m=48\)]{\includegraphics[width=0.22\textwidth]{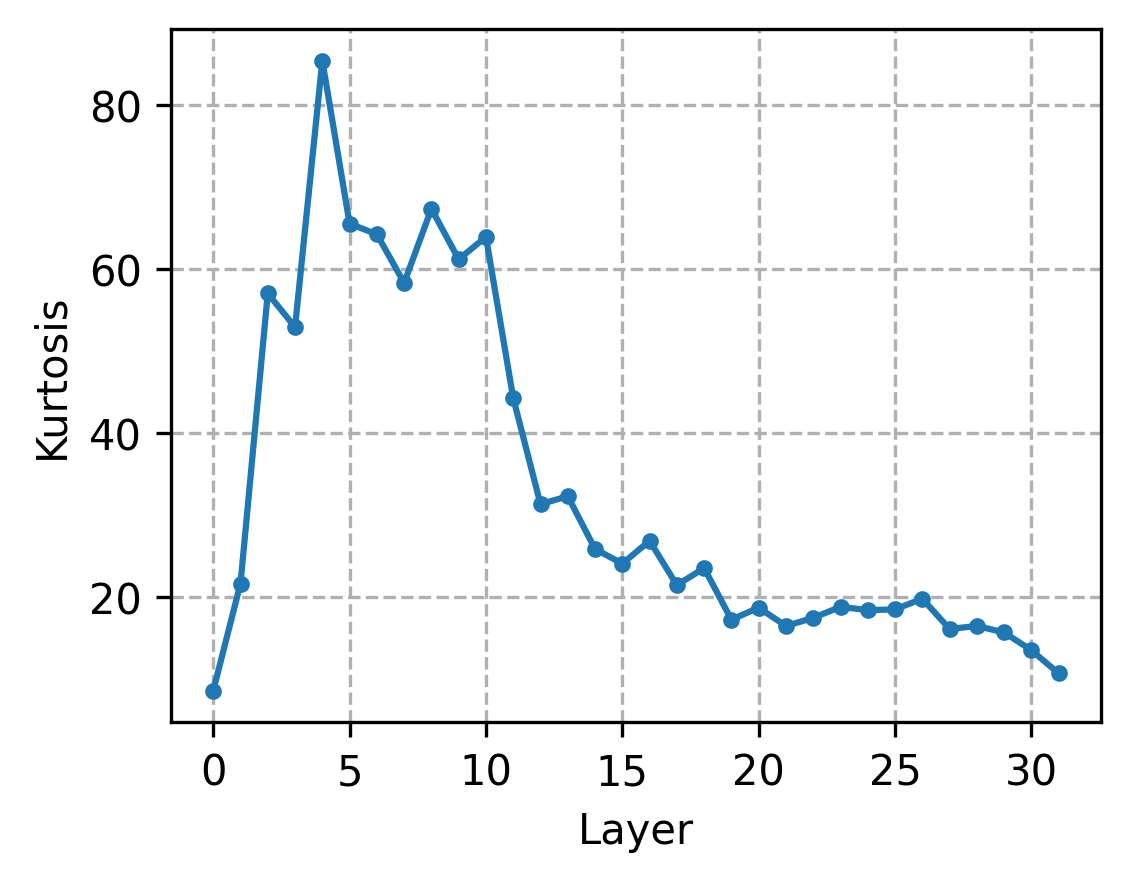}
    }
    \hfill
    \subfigure[\(m=64\)]{\includegraphics[width=0.22\textwidth]{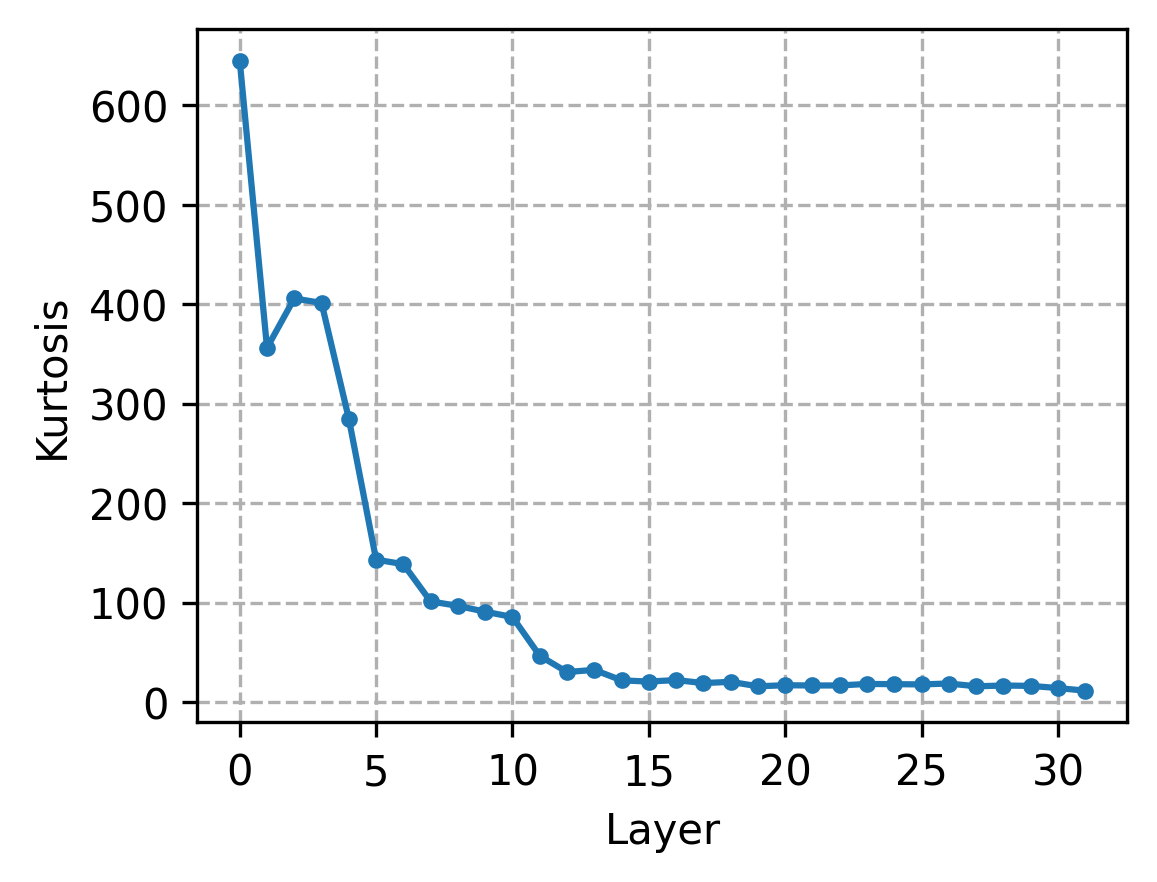}
    }
    \hfill
    \subfigure[\(m=80\)]{\includegraphics[width=0.22\textwidth]{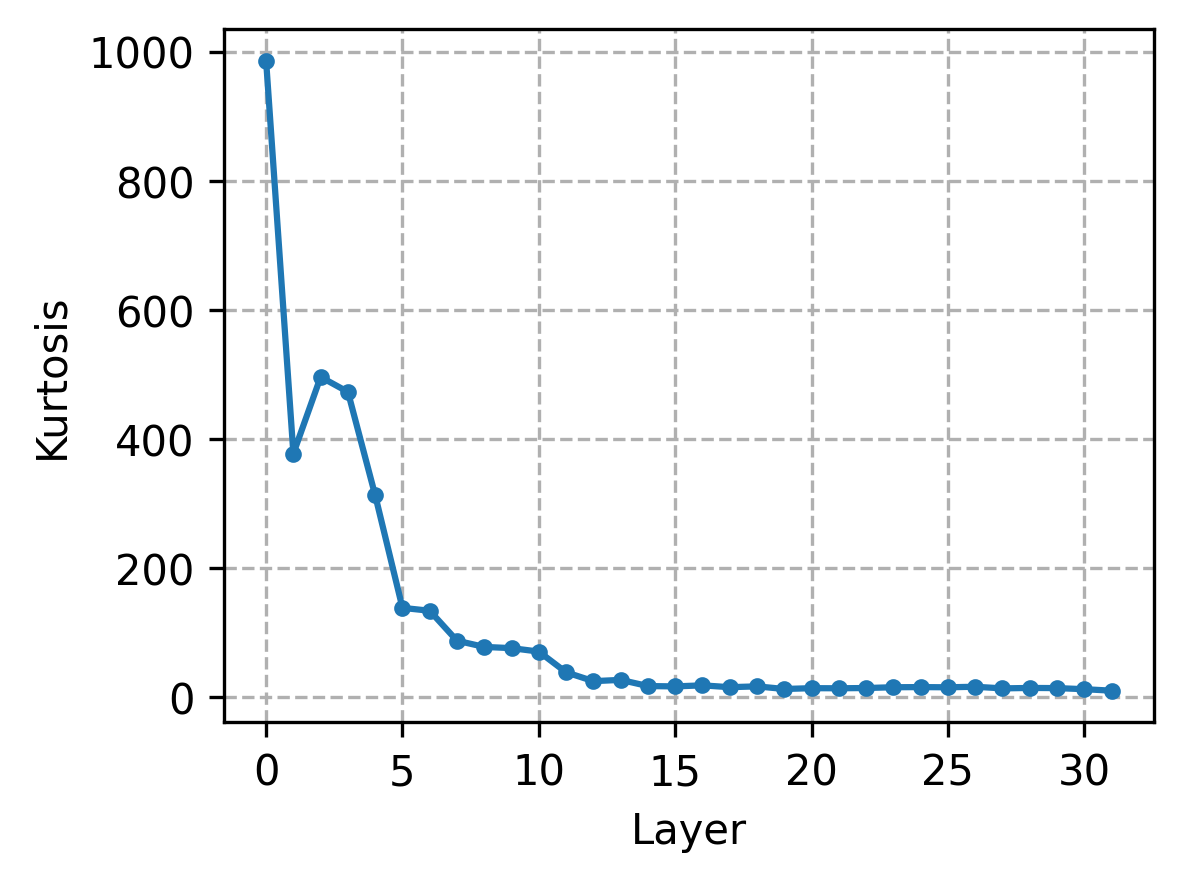}
    }
    \hfill
    \subfigure[\(m=96\)]{\includegraphics[width=0.22\textwidth]{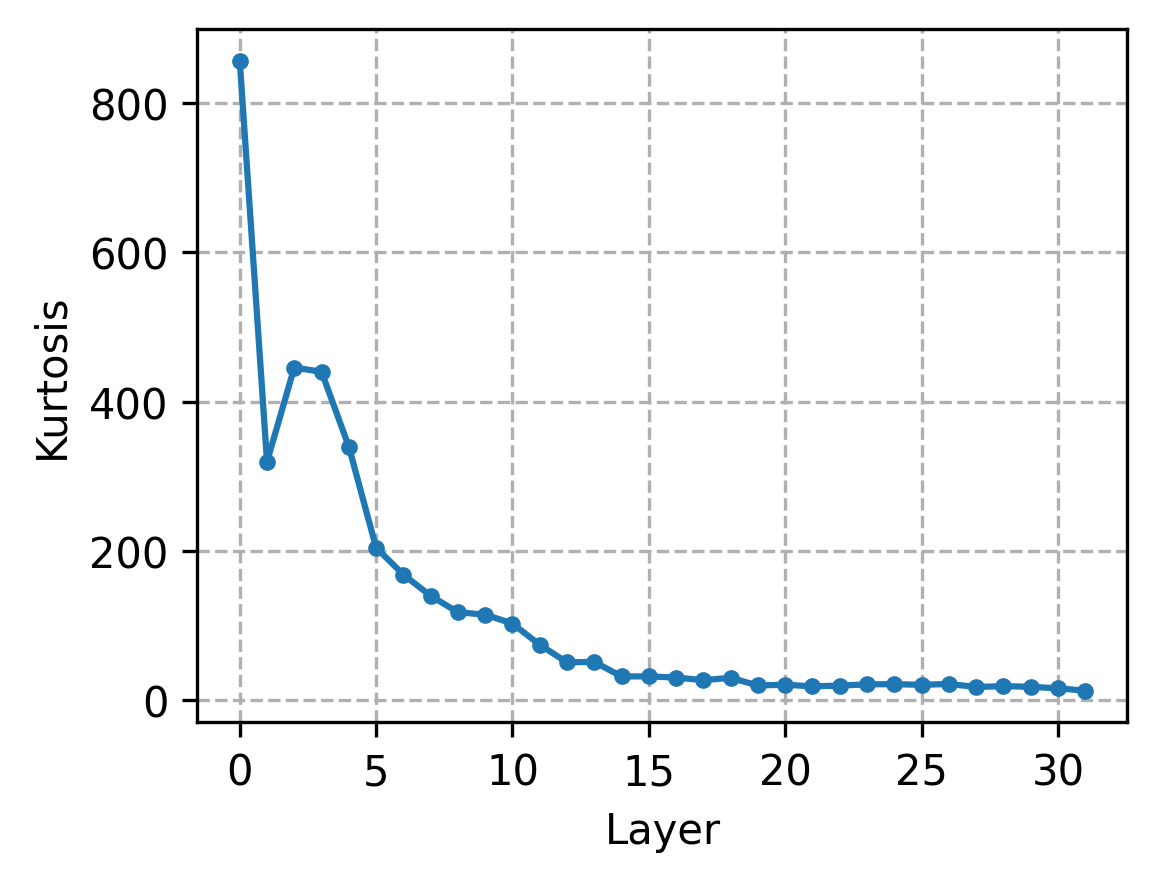}
    }
    \hfill
    \subfigure[\(m=112\)]{\includegraphics[width=0.22\textwidth]{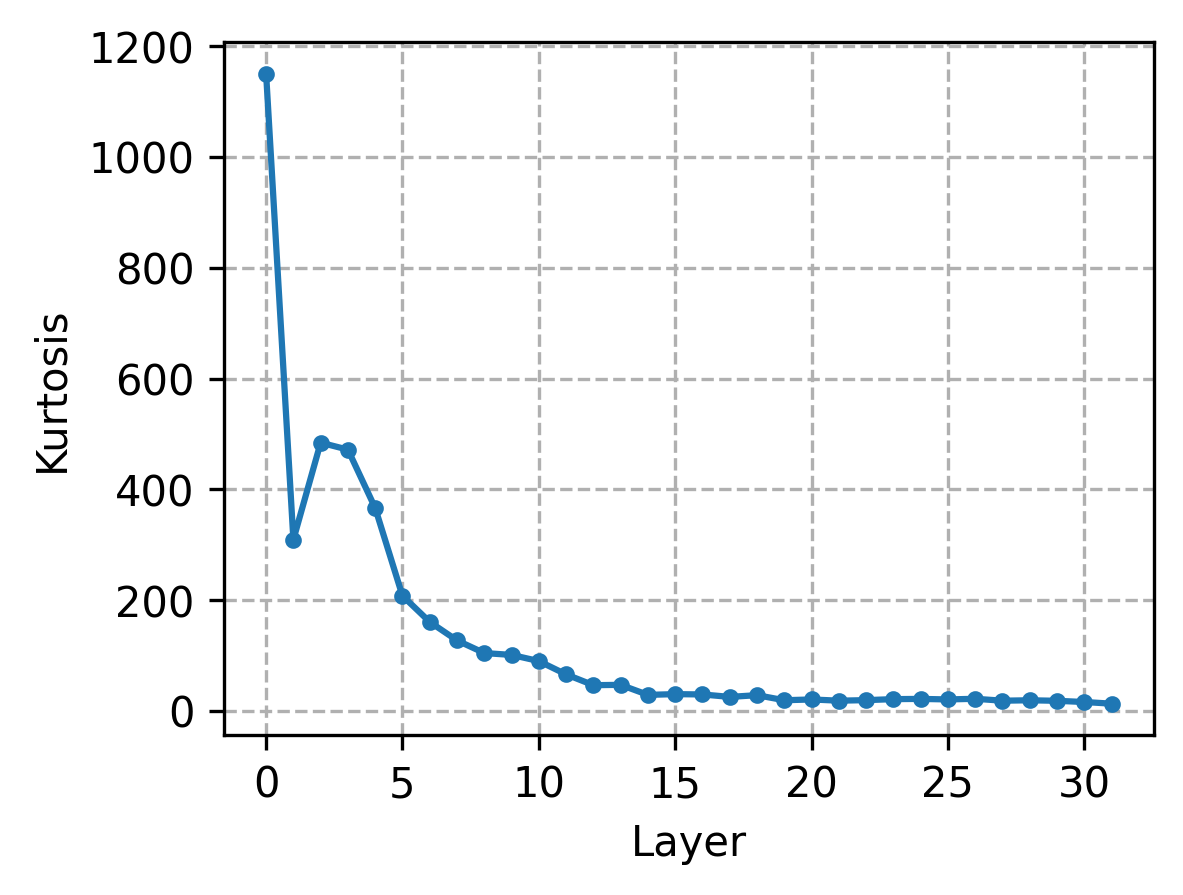}
    }
    \hfill
    \subfigure[\(m=128\)]{\includegraphics[width=0.22\textwidth]{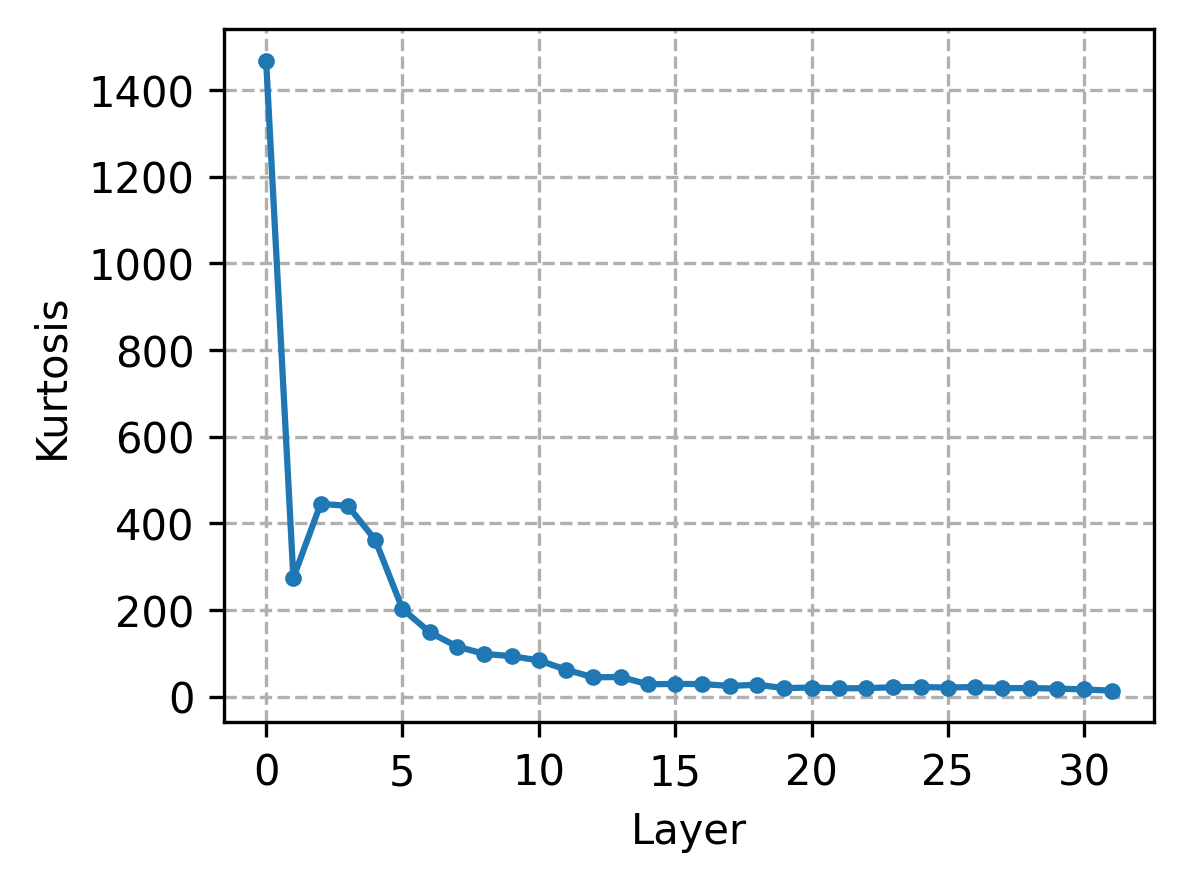}
    }
  \caption{In Pythia-6.9B, the superposition distribution converges before \( m = 128 \).}
  \label{fig:superposition_converges_pythia-6.9b}
\end{figure*}

\begin{figure*}
    \centering
    \subfigure[\(m=16\)]{\includegraphics[width=0.22\textwidth]{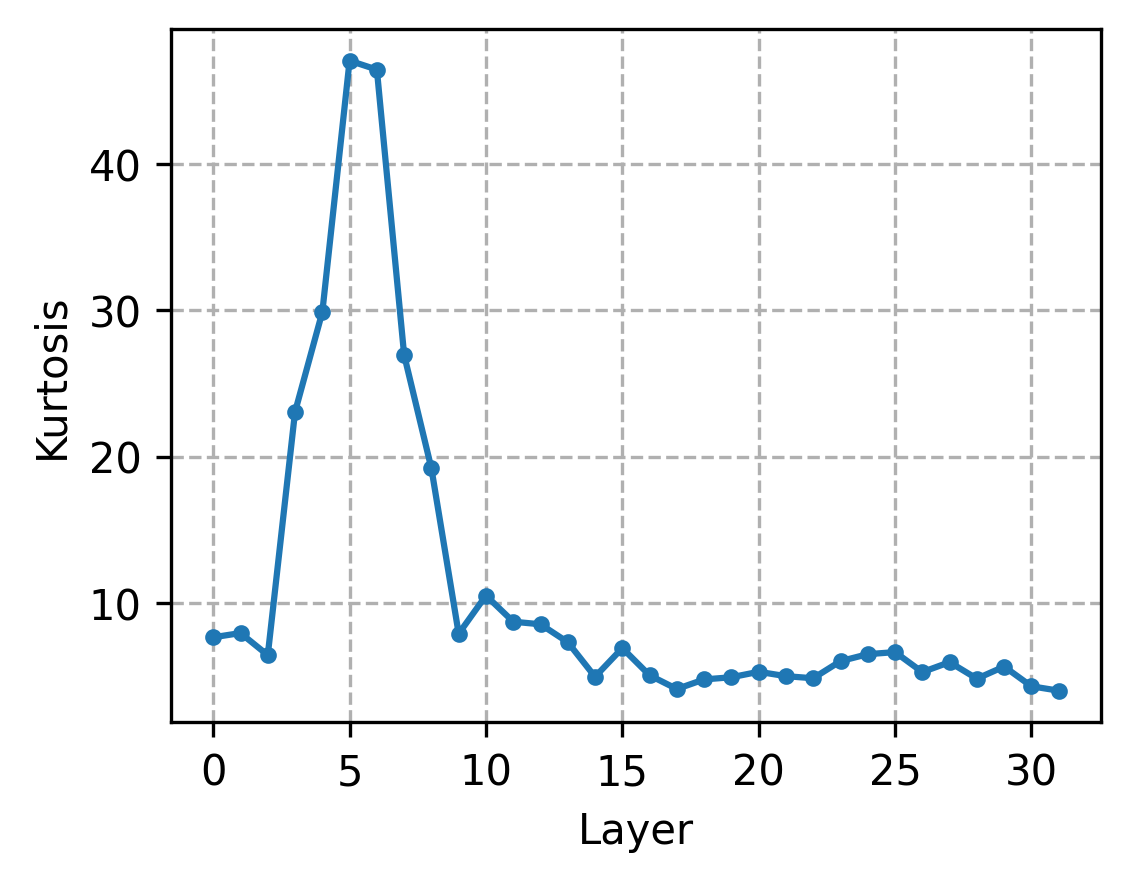}
    }
    \hfill
    \subfigure[\(m=32\)]{\includegraphics[width=0.22\textwidth]{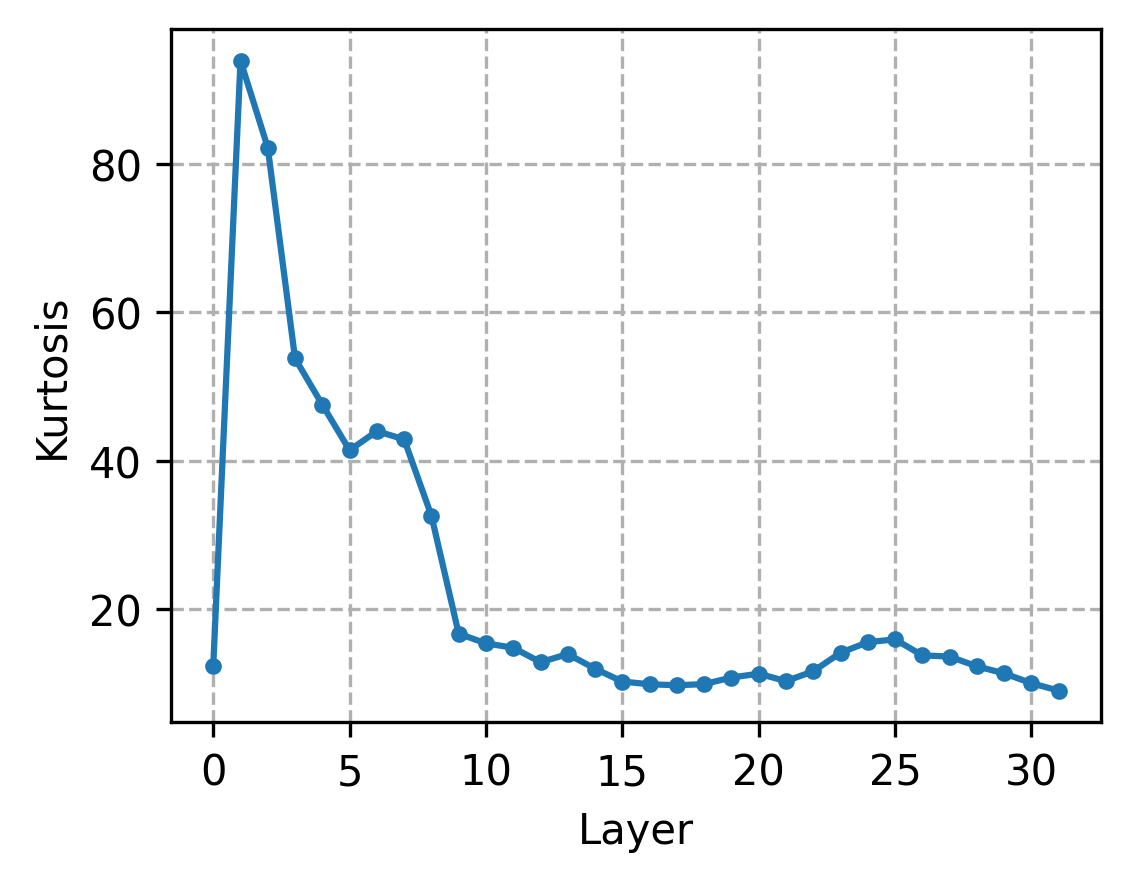}
    }
    \hfill
    \subfigure[\(m=48\)]{\includegraphics[width=0.22\textwidth]{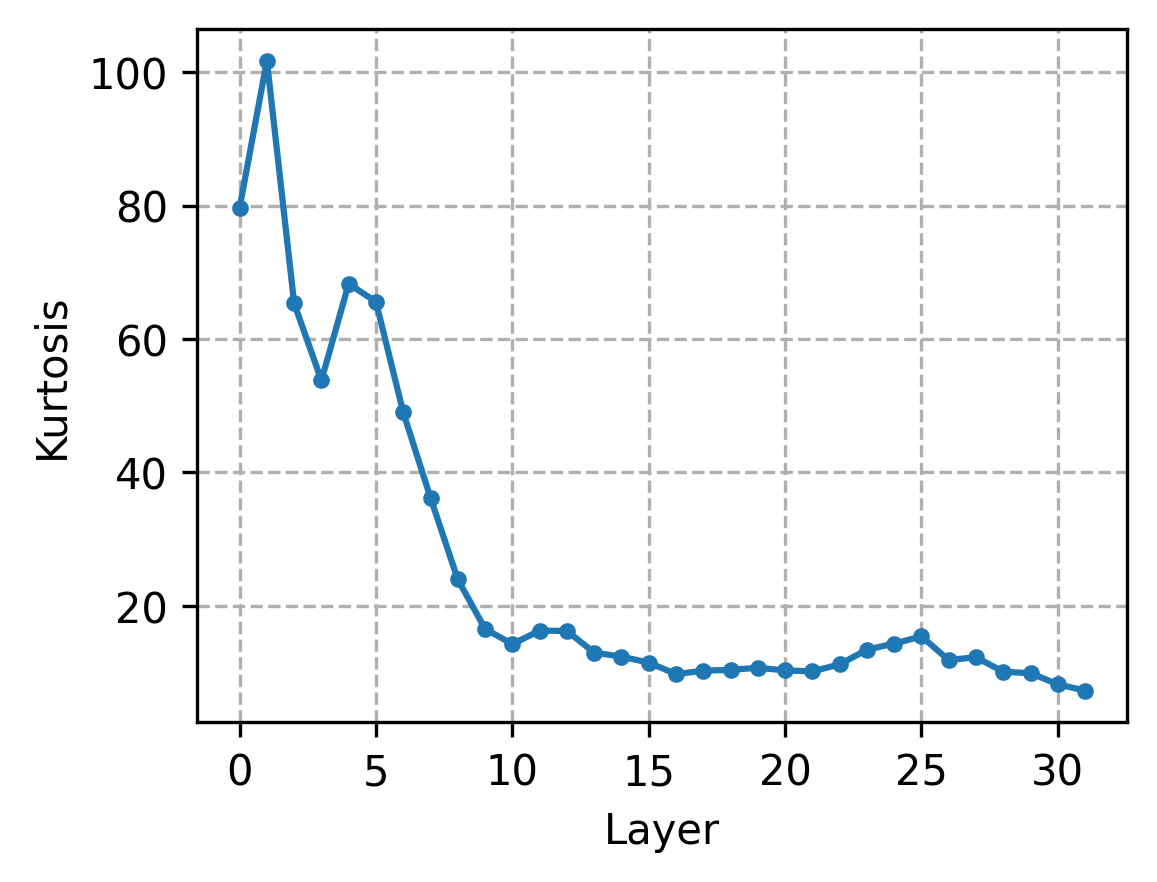}
    }
    \hfill
    \subfigure[\(m=64\)]{\includegraphics[width=0.22\textwidth]{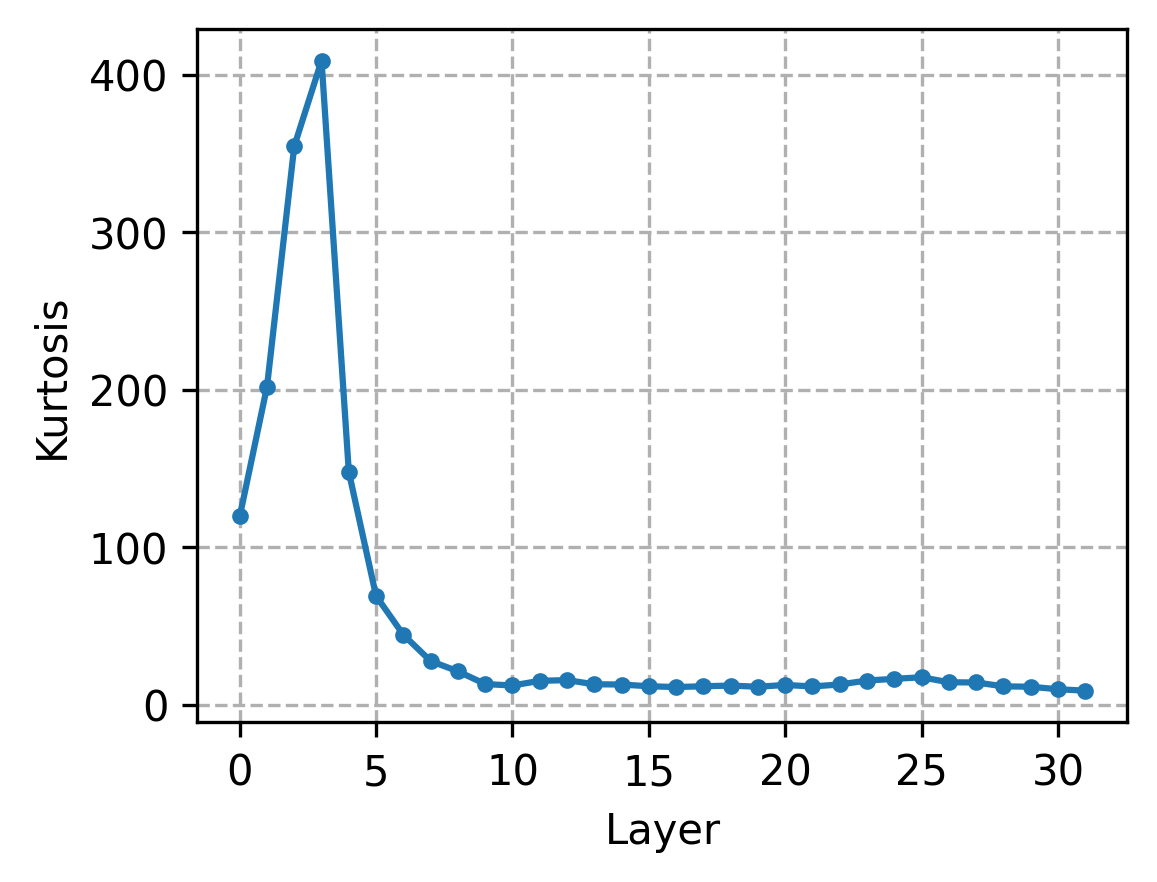}
    }
    \hfill
    \subfigure[\(m=80\)]{\includegraphics[width=0.22\textwidth]{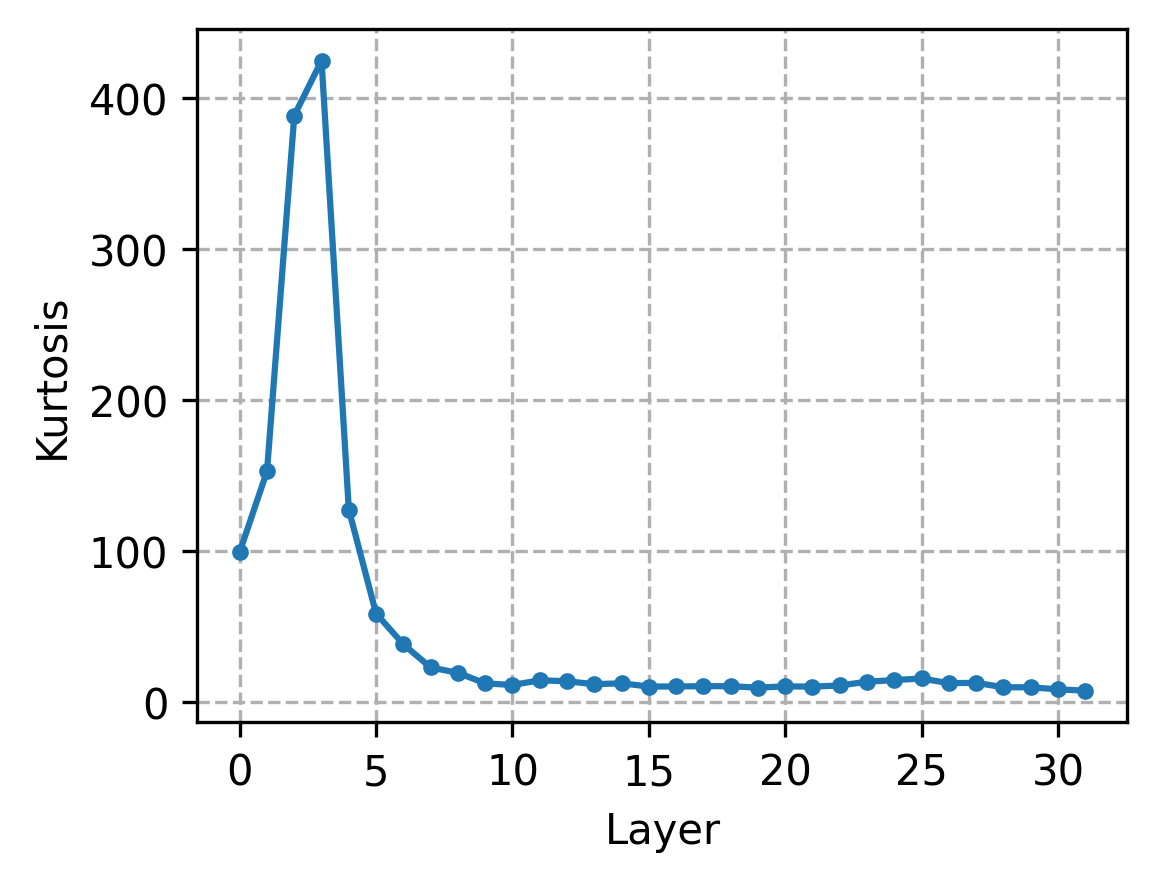}
    }
    \hfill
    \subfigure[\(m=96\)]{\includegraphics[width=0.22\textwidth]{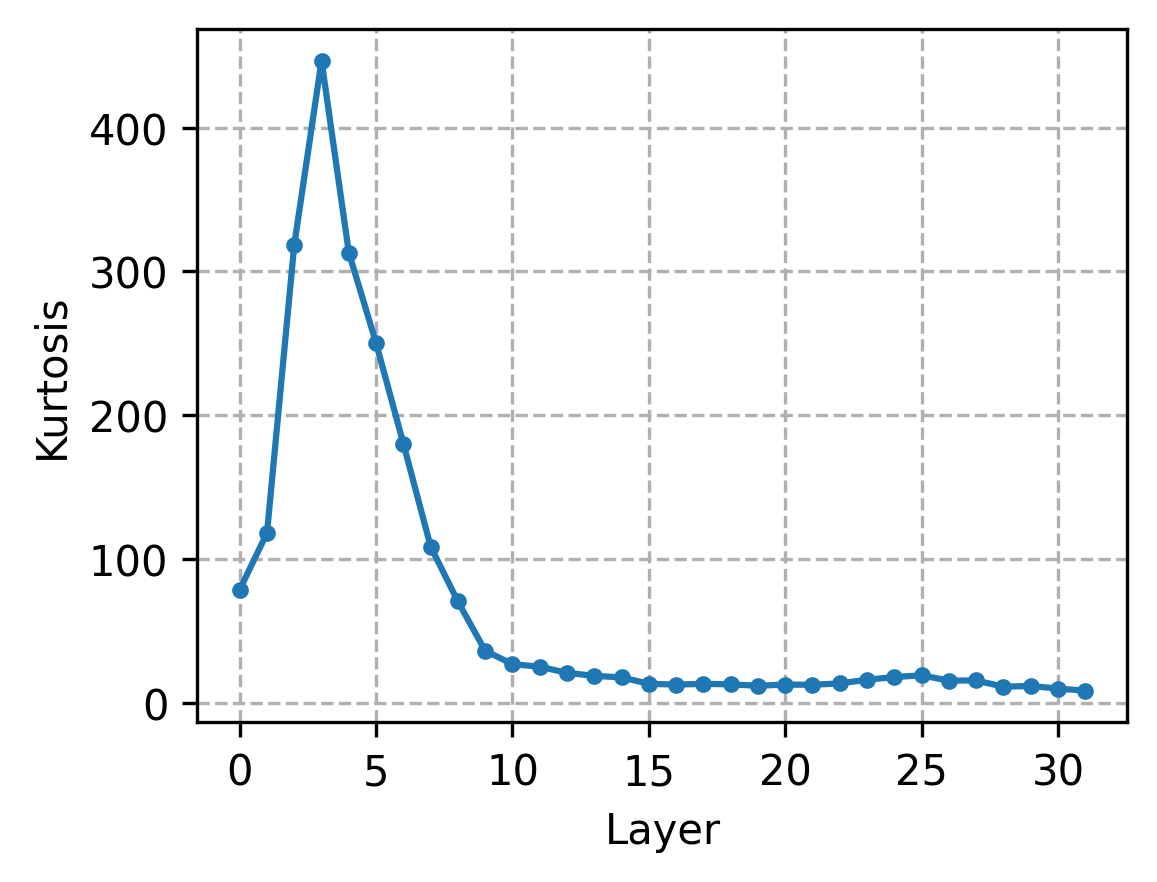}
    }
    \hfill
    \subfigure[\(m=112\)]{\includegraphics[width=0.22\textwidth]{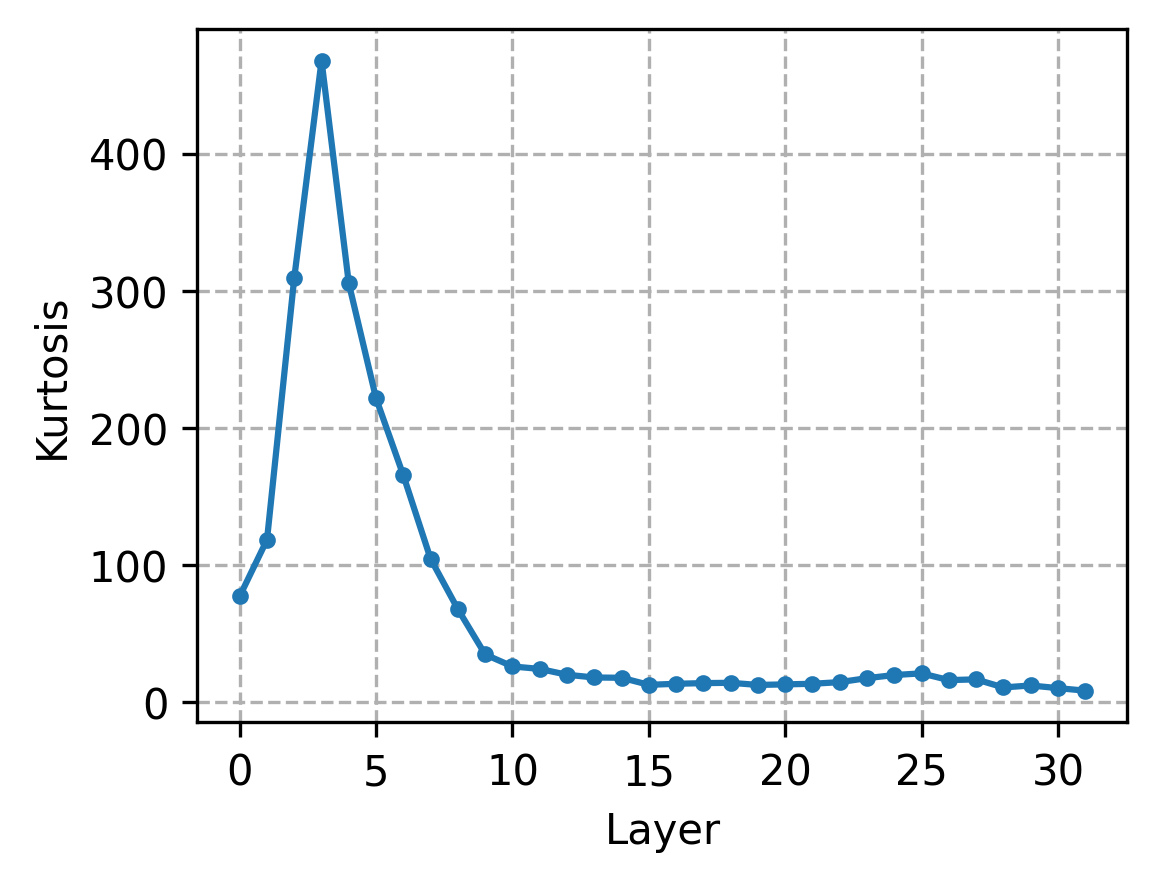}
    }
    \hfill
    \subfigure[\(m=128\)]{\includegraphics[width=0.22\textwidth]{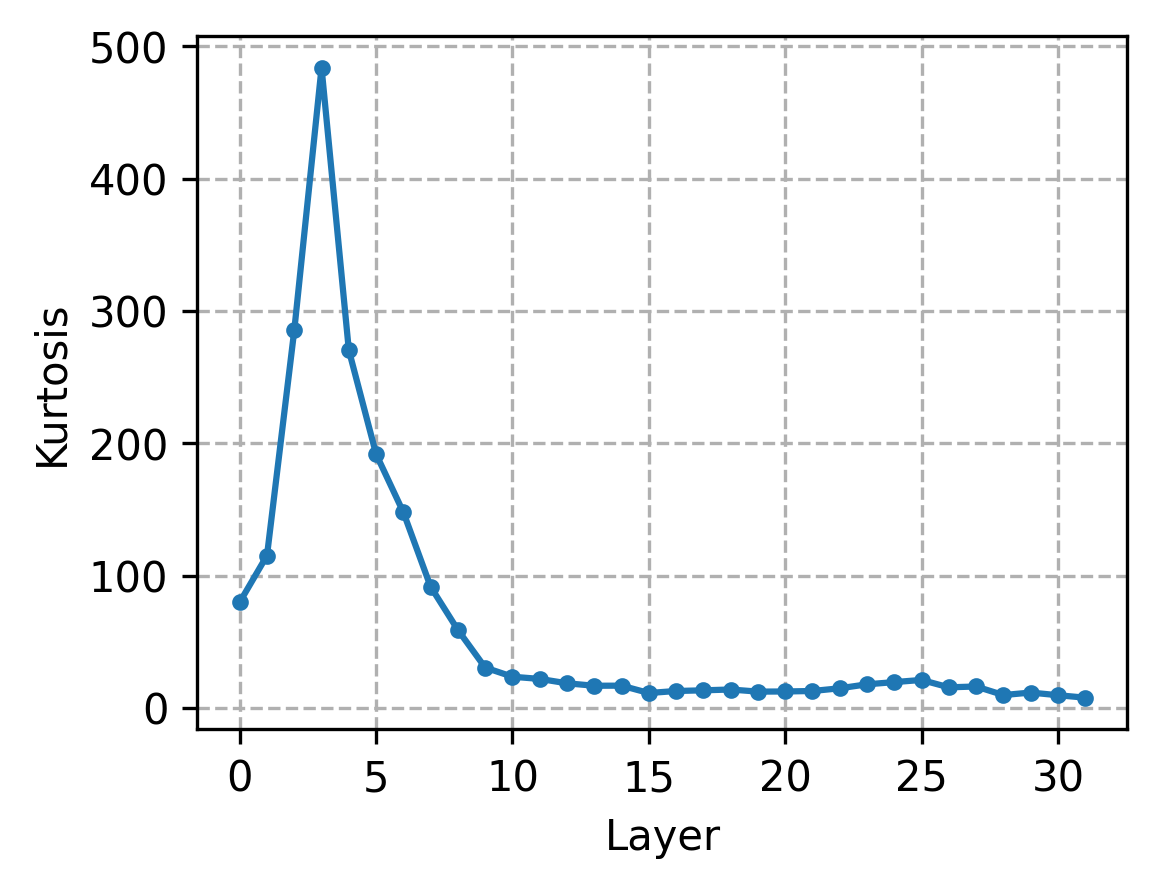}
    }
  \caption{In Llama2-7B, the superposition distribution converges before \( m = 128 \).}
  \label{fig:superposition_converges_llama2-7b}
\end{figure*}

\begin{figure*}
    \centering
    \subfigure[\(m=16\)]{\includegraphics[width=0.22\textwidth]{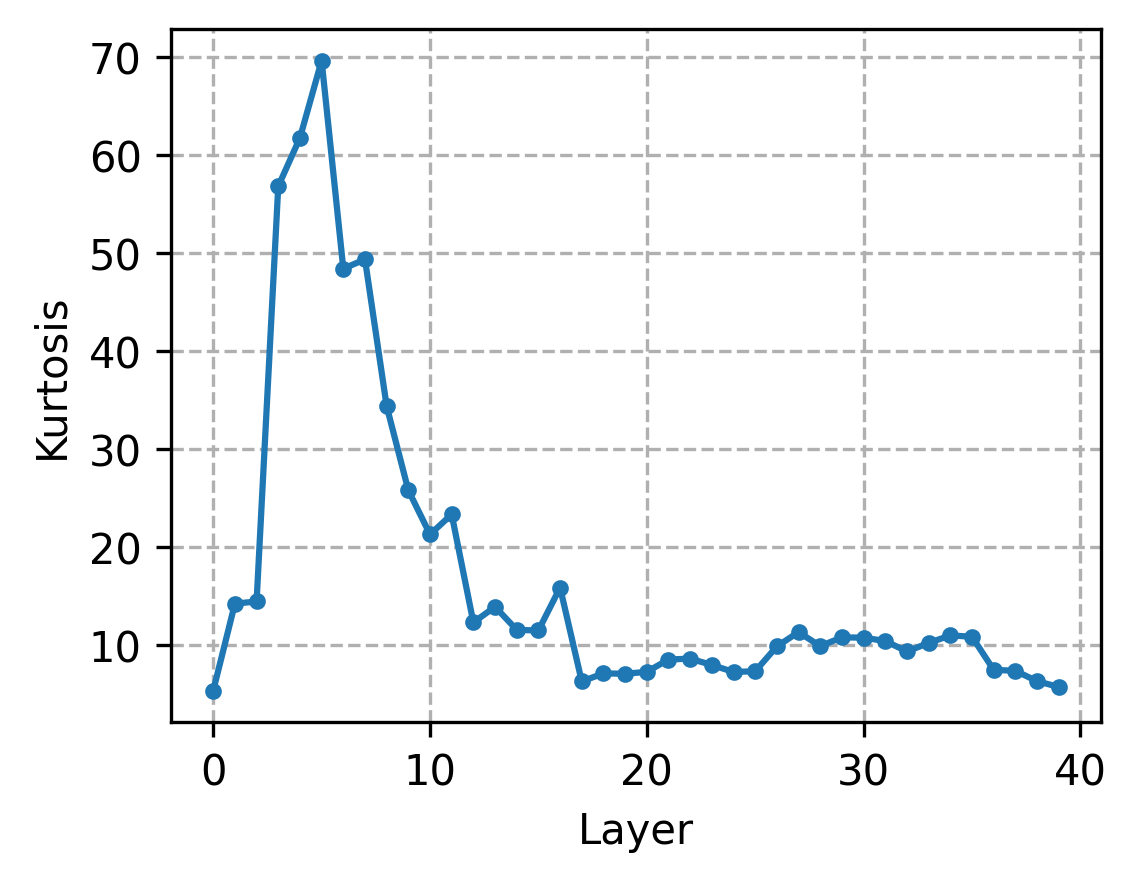}
    }
    \hfill
    \subfigure[\(m=32\)]{\includegraphics[width=0.22\textwidth]{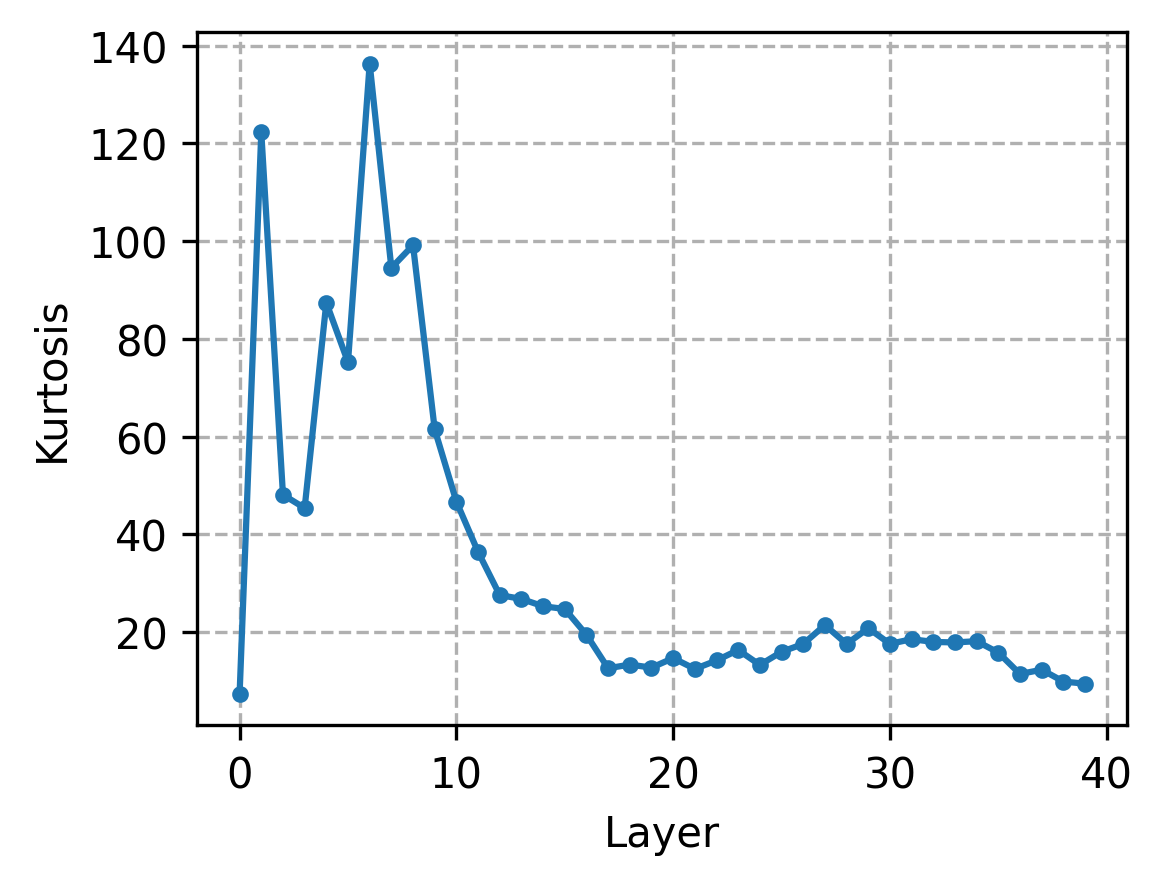}
    }
    \hfill
    \subfigure[\(m=48\)]{\includegraphics[width=0.22\textwidth]{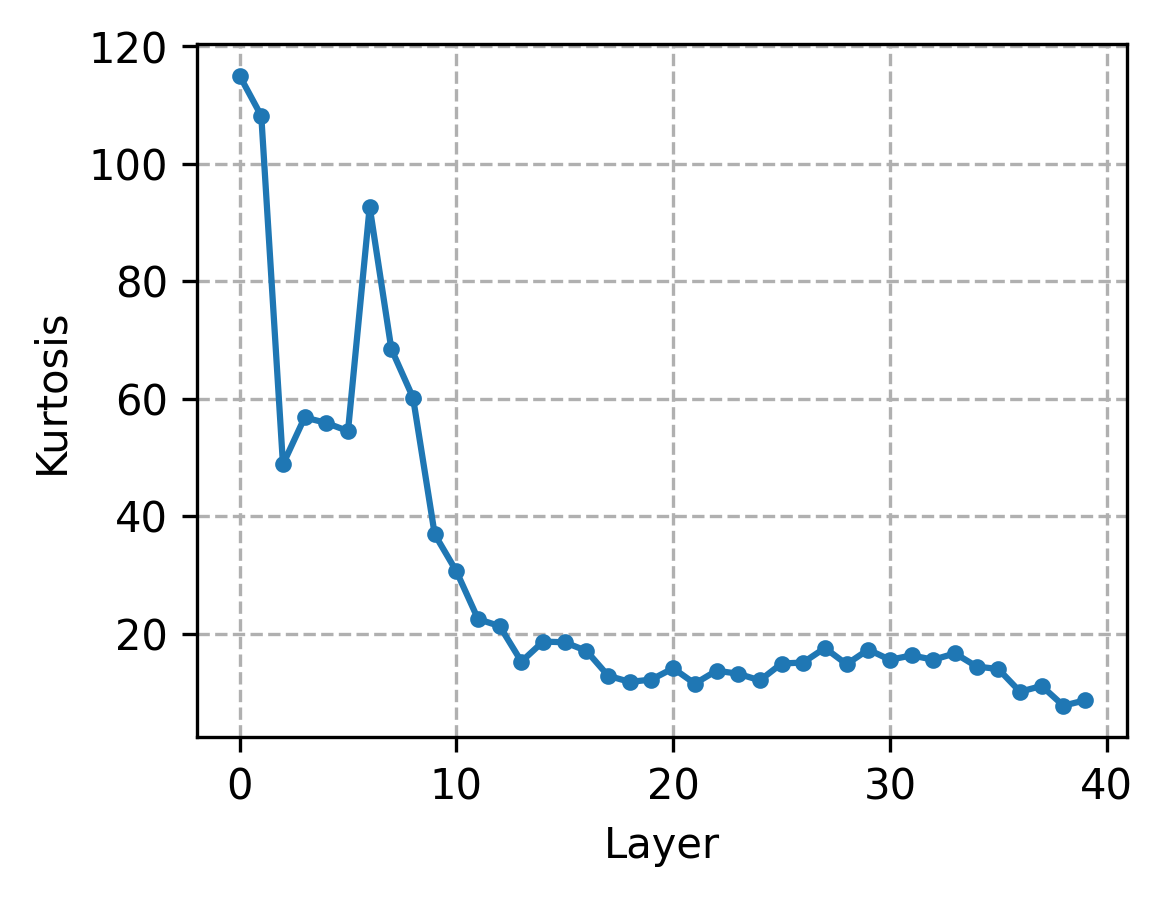}
    }
    \hfill
    \subfigure[\(m=64\)]{\includegraphics[width=0.22\textwidth]{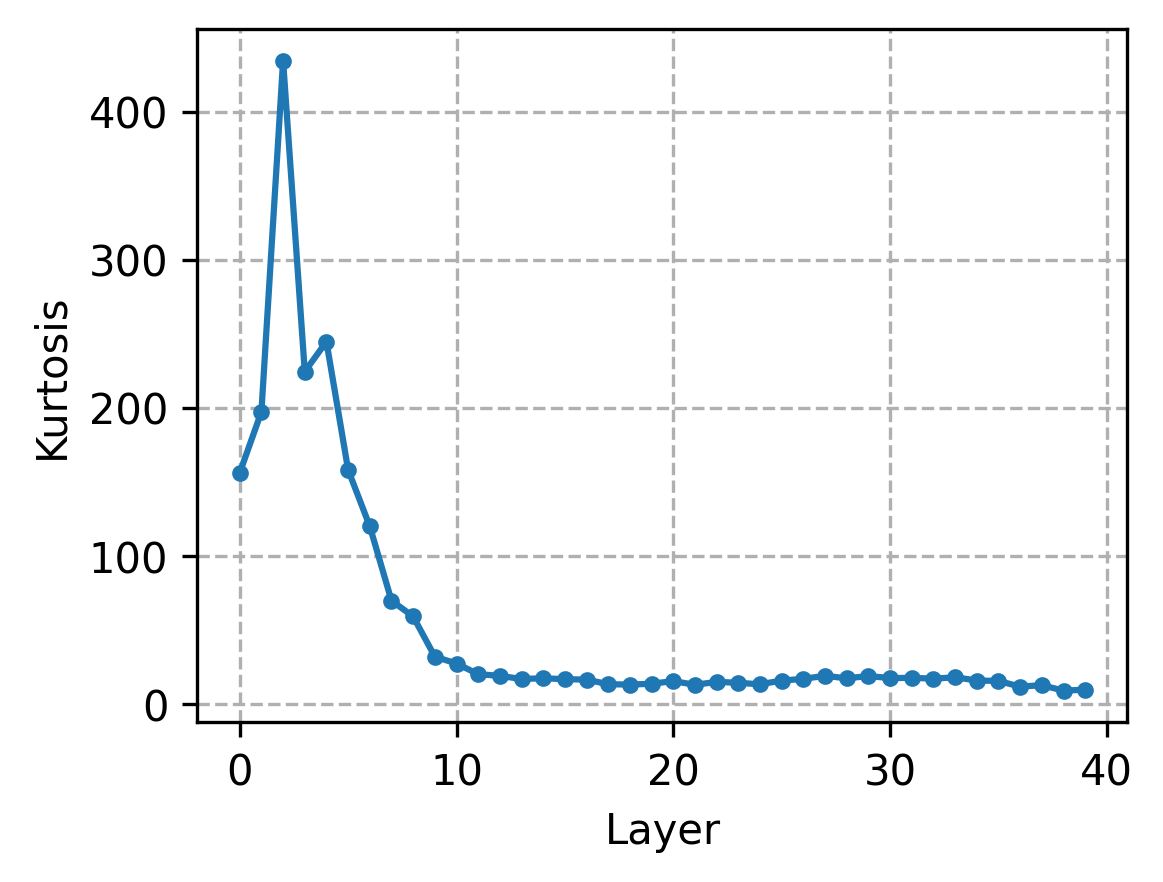}
    }
    \hfill
    \subfigure[\(m=80\)]{\includegraphics[width=0.22\textwidth]{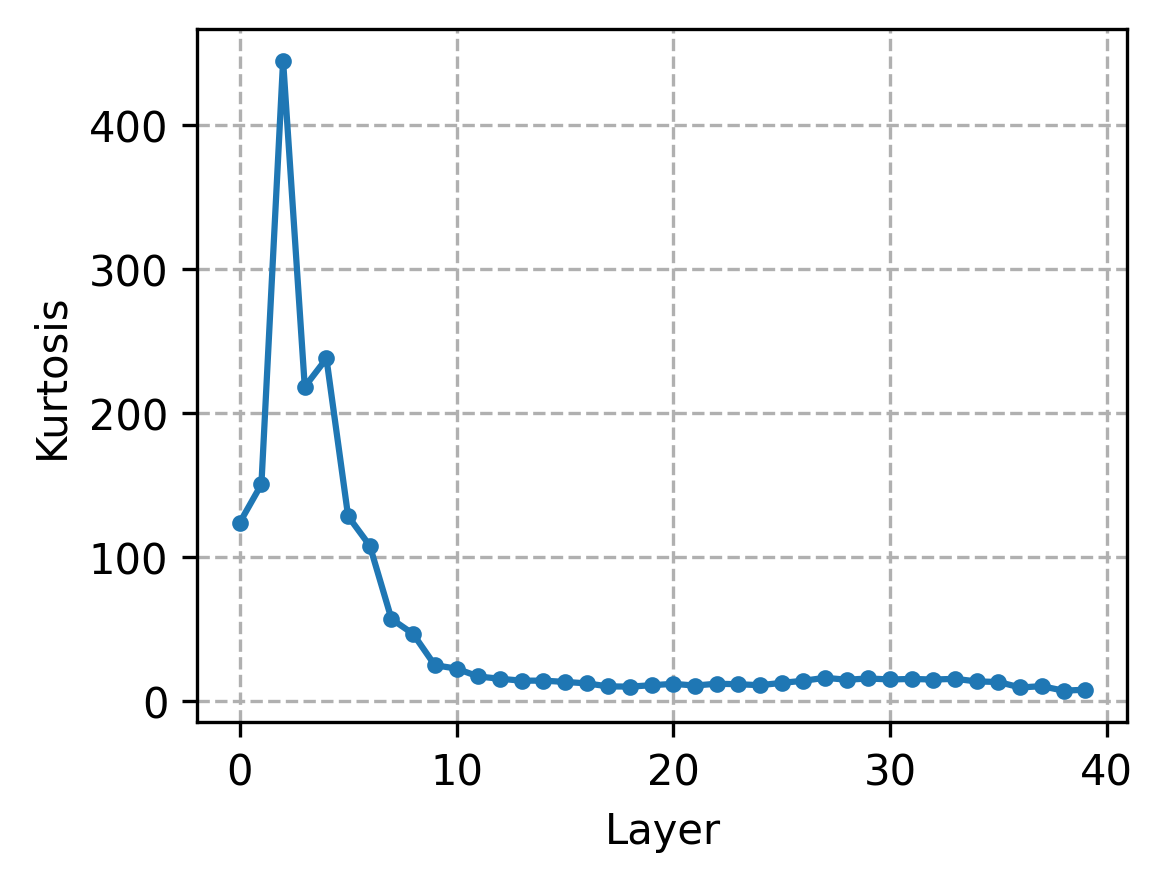}
    }
    \hfill
    \subfigure[\(m=96\)]{\includegraphics[width=0.22\textwidth]{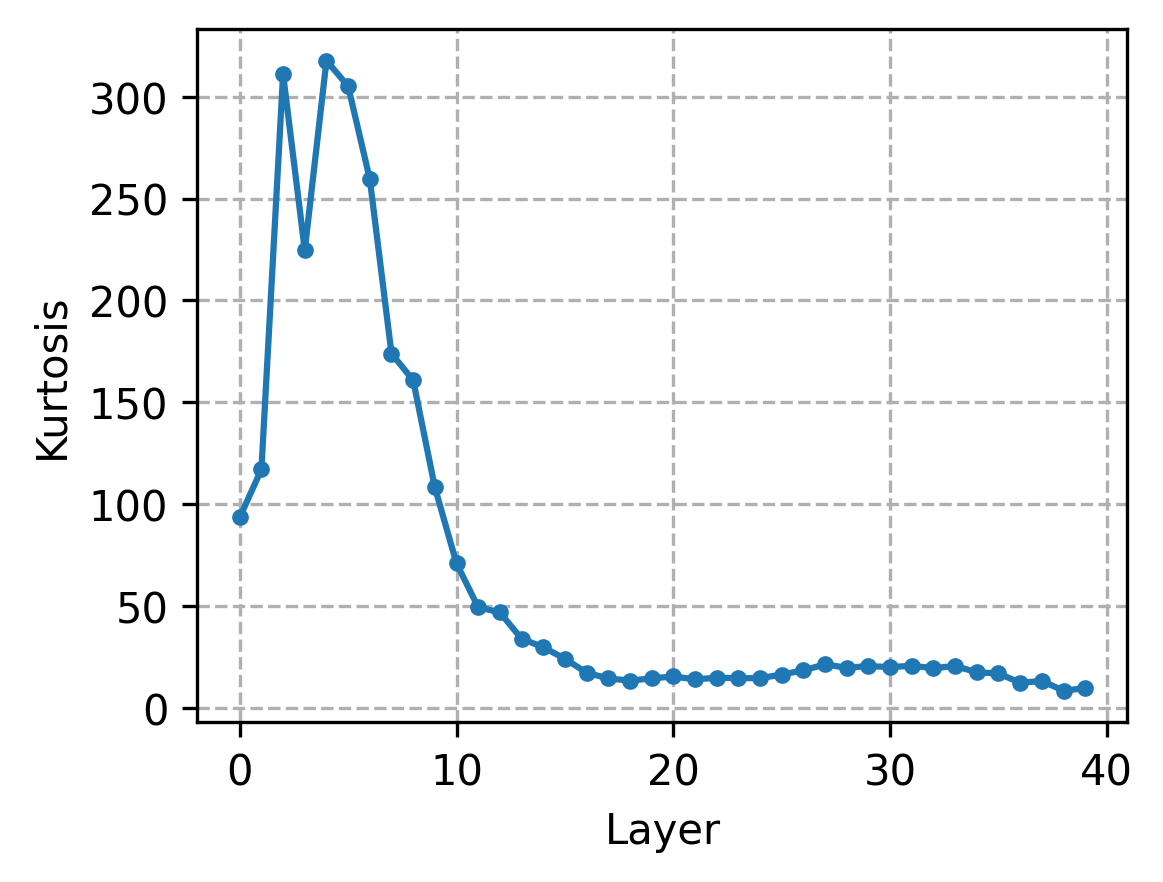}
    }
    \hfill
    \subfigure[\(m=112\)]{\includegraphics[width=0.22\textwidth]{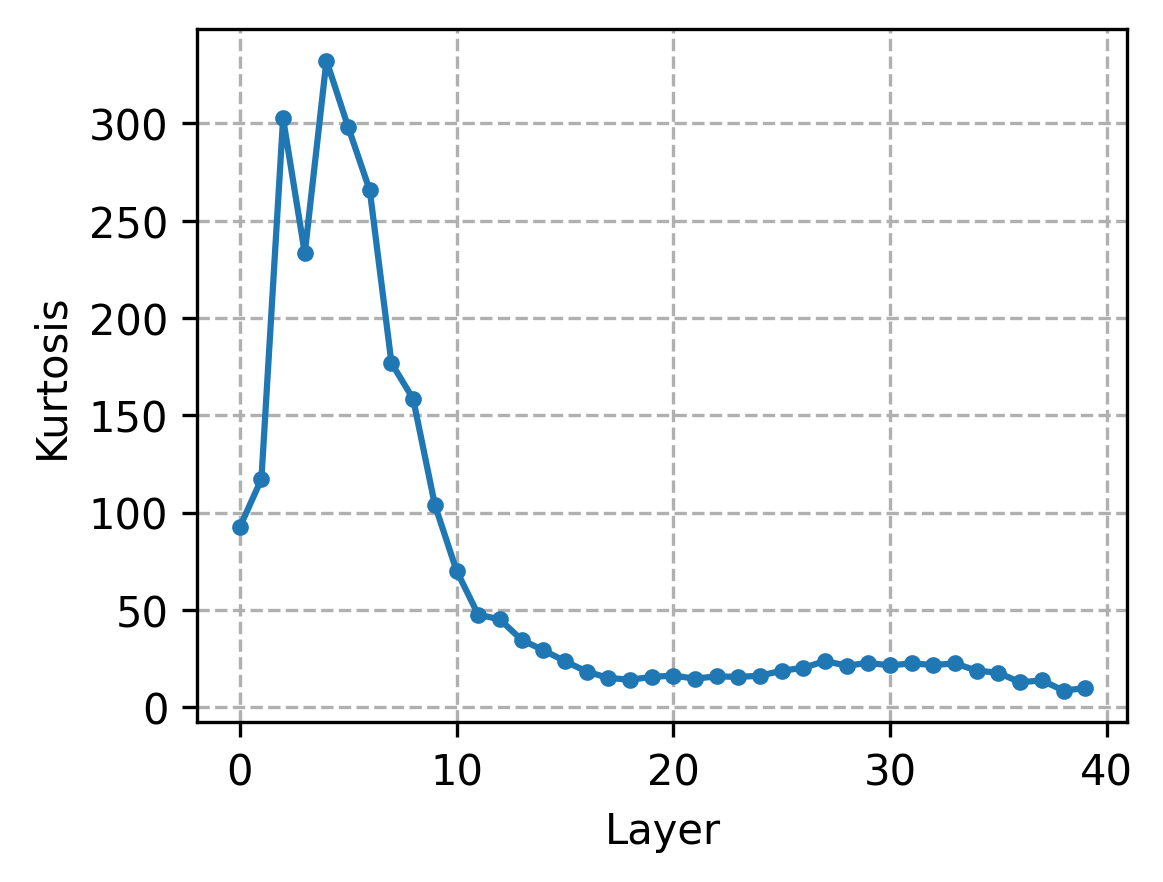}
    }
    \hfill
    \subfigure[\(m=128\)]{\includegraphics[width=0.22\textwidth]{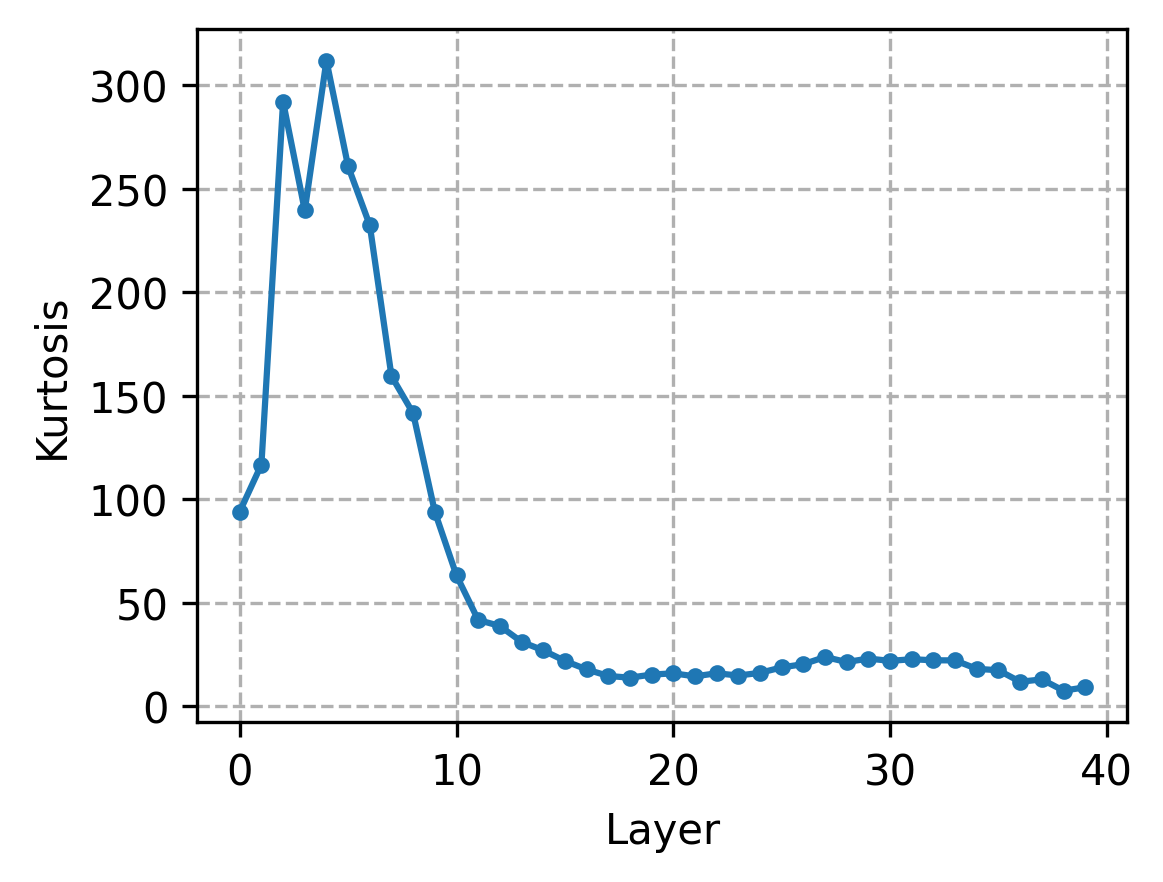}
    }
  \caption{In Llama2-13B, the superposition distribution converges before \( m = 128 \).}
  \label{fig:superposition_converges_llama2-13b}
\end{figure*}

\begin{figure*}
    \centering
    \subfigure[\(m=16\)]{\includegraphics[width=0.22\textwidth]{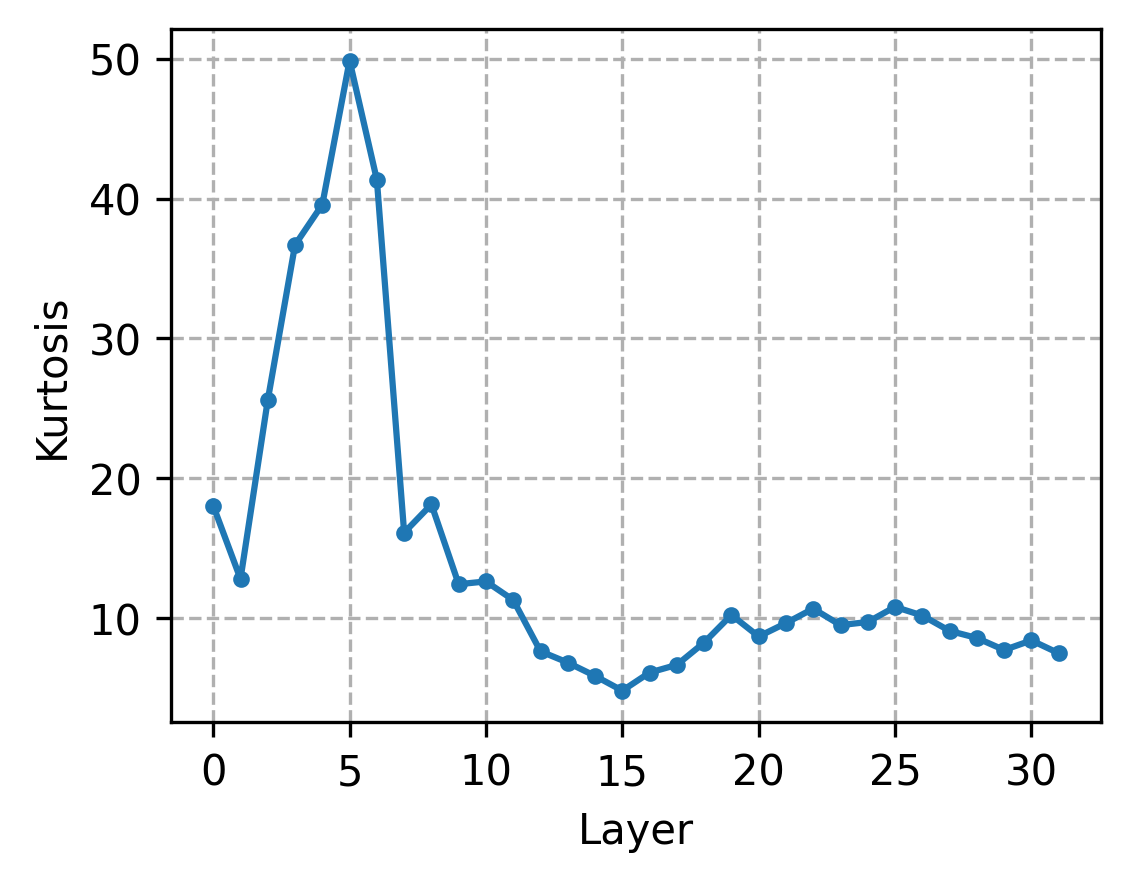}
    }
    \hfill
    \subfigure[\(m=32\)]{\includegraphics[width=0.22\textwidth]{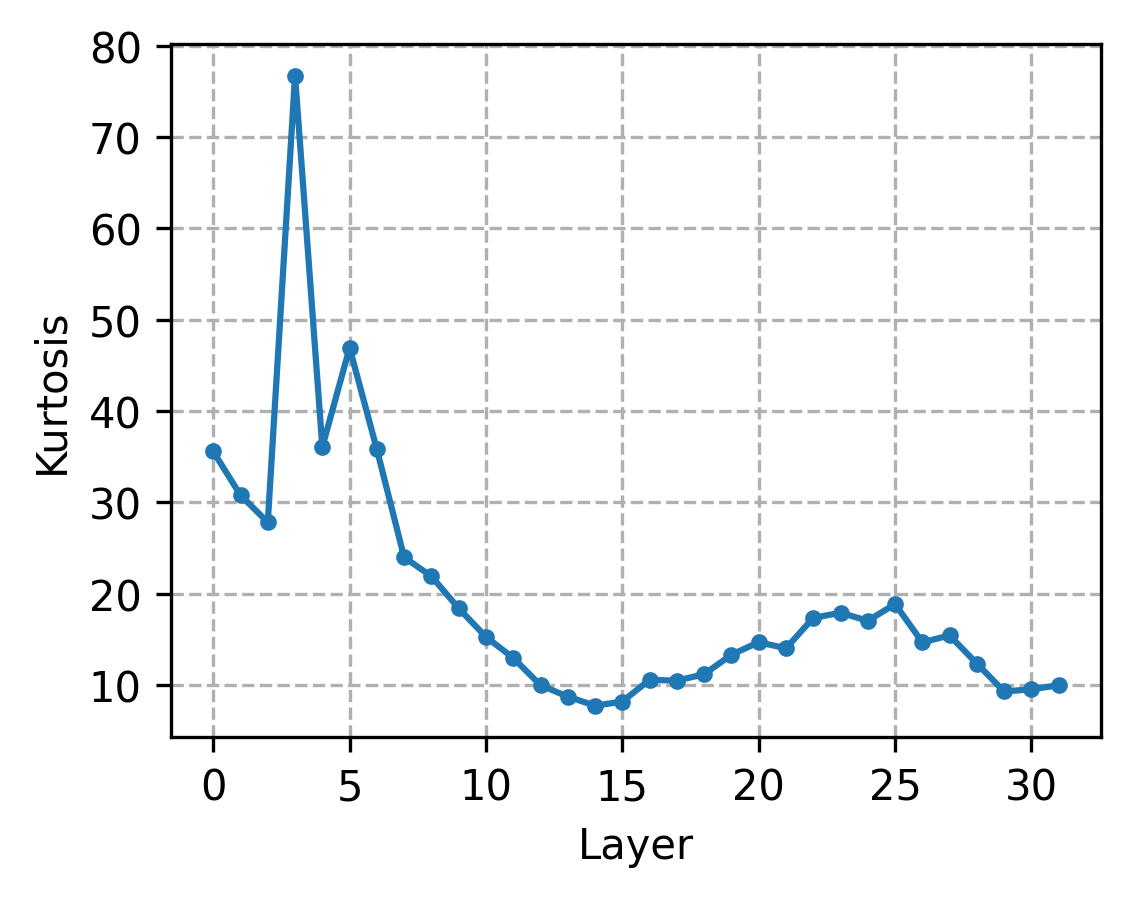}
    }
    \hfill
    \subfigure[\(m=48\)]{\includegraphics[width=0.22\textwidth]{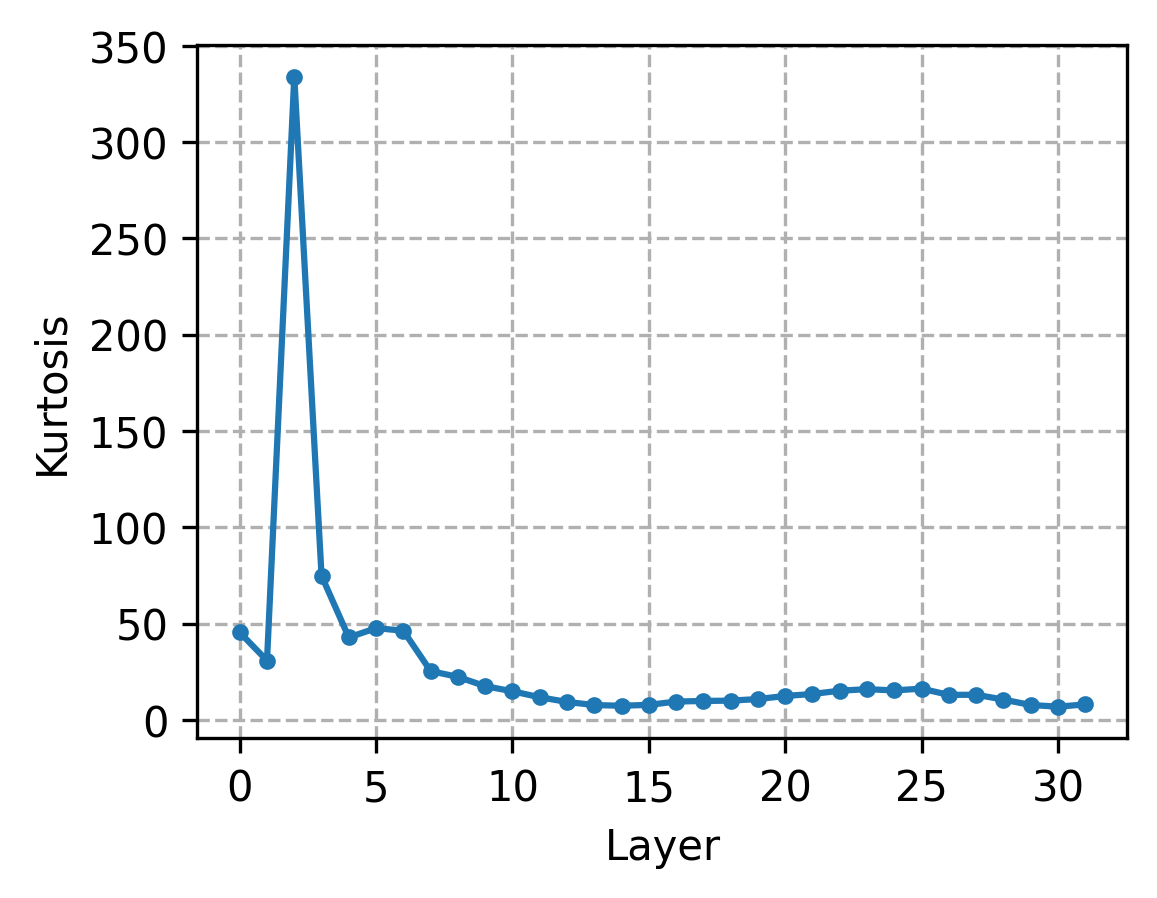}
    }
    \hfill
    \subfigure[\(m=64\)]{\includegraphics[width=0.22\textwidth]{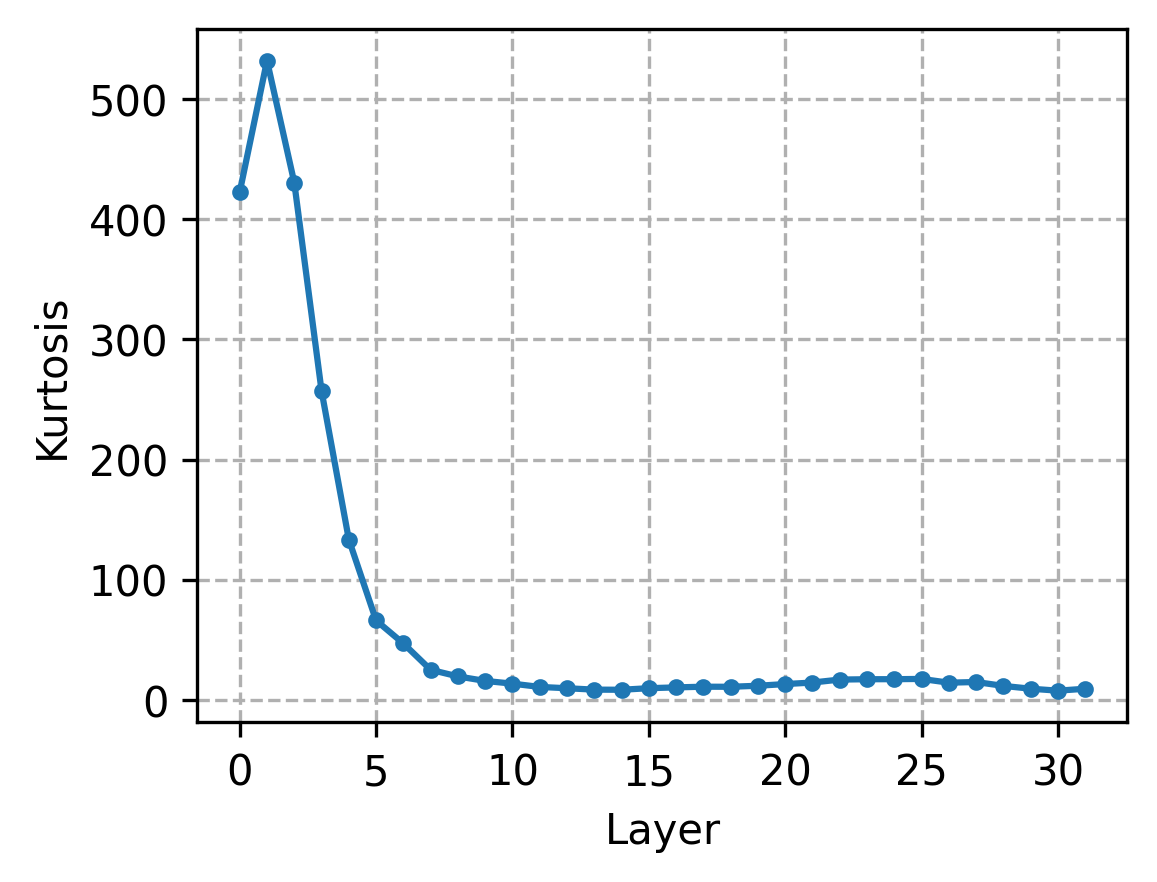}
    }
    \hfill
    \subfigure[\(m=80\)]{\includegraphics[width=0.22\textwidth]{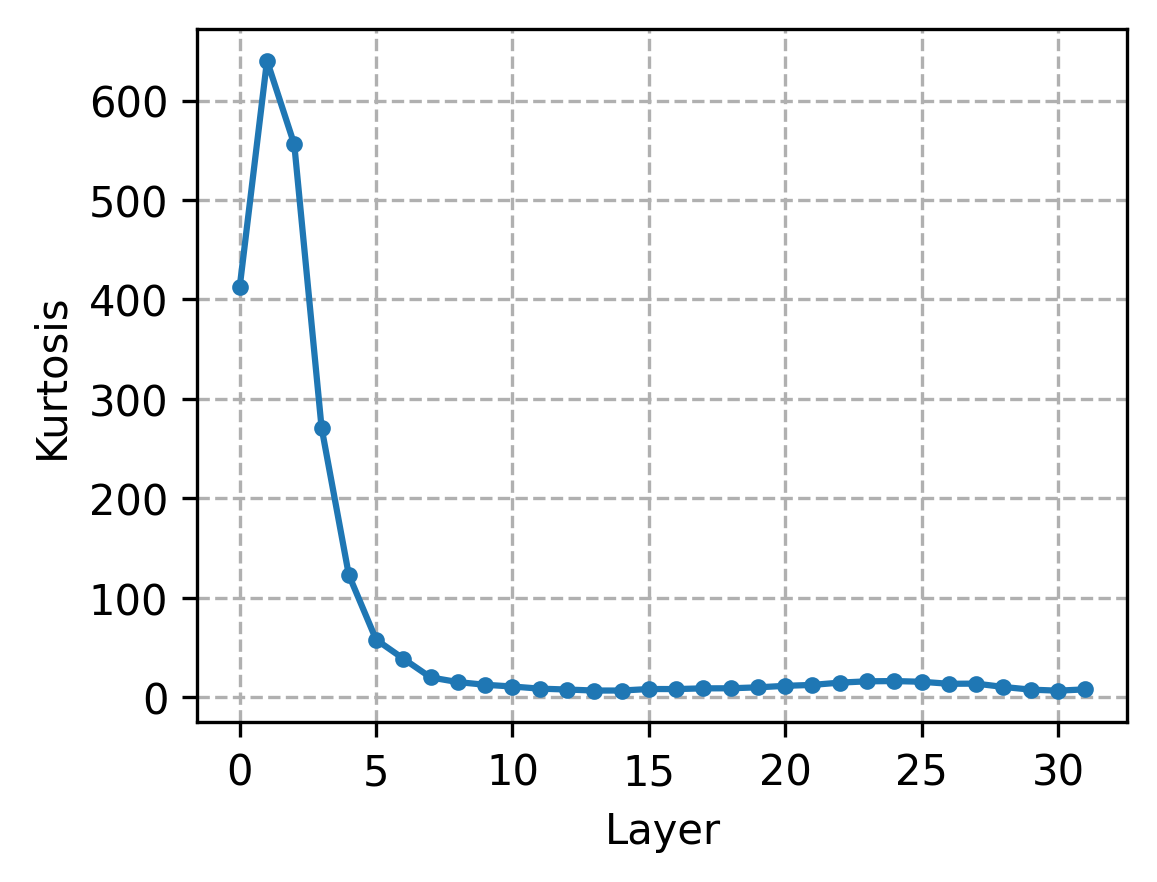}
    }
    \hfill
    \subfigure[\(m=96\)]{\includegraphics[width=0.22\textwidth]{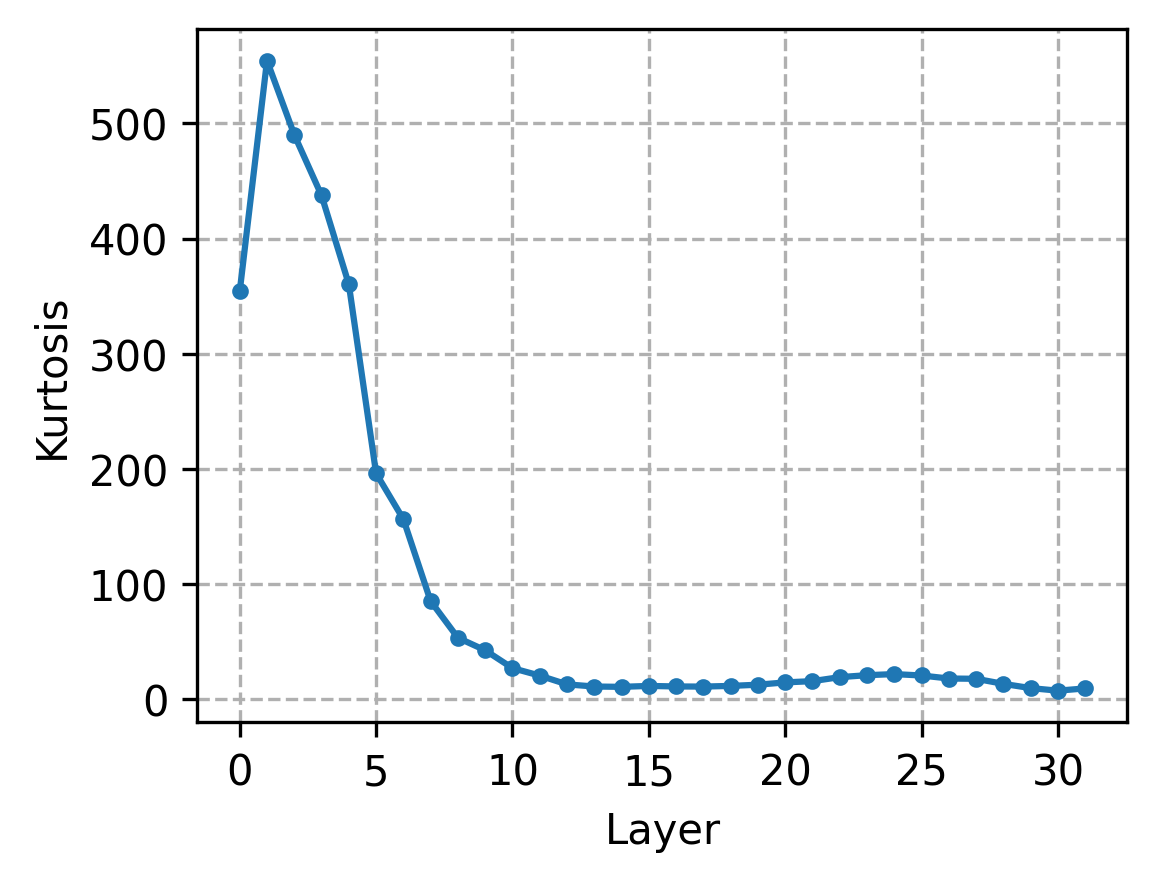}
    }
    \hfill
    \subfigure[\(m=112\)]{\includegraphics[width=0.22\textwidth]{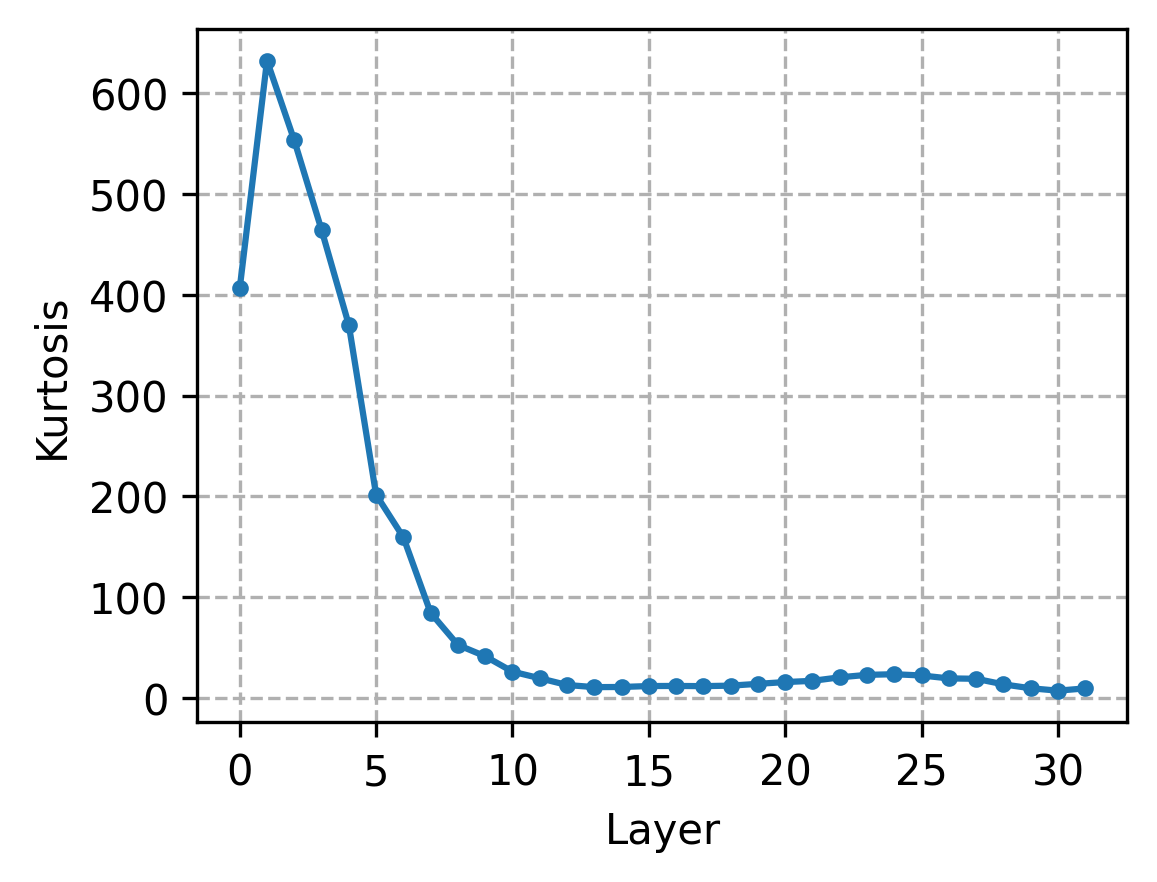}
    }
    \hfill
    \subfigure[\(m=128\)]{\includegraphics[width=0.22\textwidth]{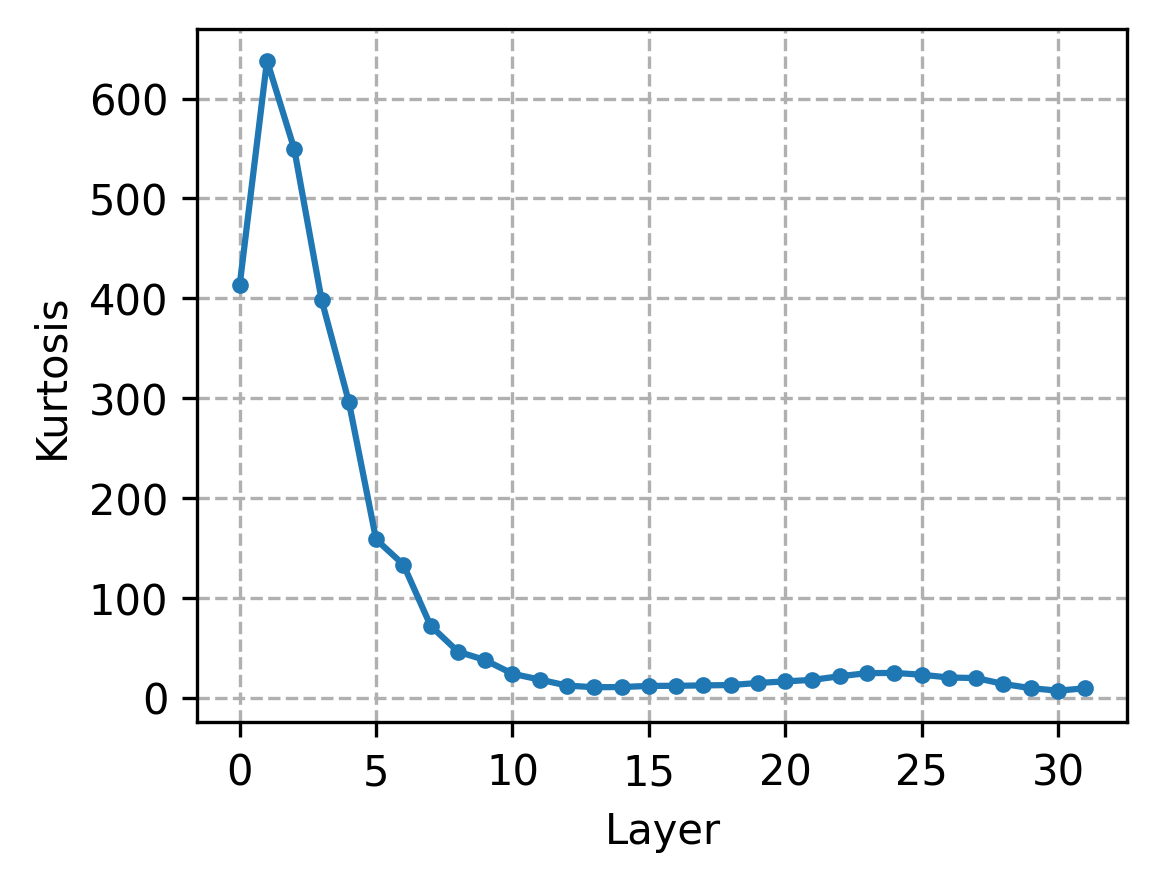}
    }
  \caption{In Llama3-8B, the superposition distribution converges before \( m = 128 \).}
  \label{fig:superposition_converges_llama3-8b}
\end{figure*}

\begin{figure*}
    \centering
    \subfigure[\(m=16\)]{\includegraphics[width=0.22\textwidth]{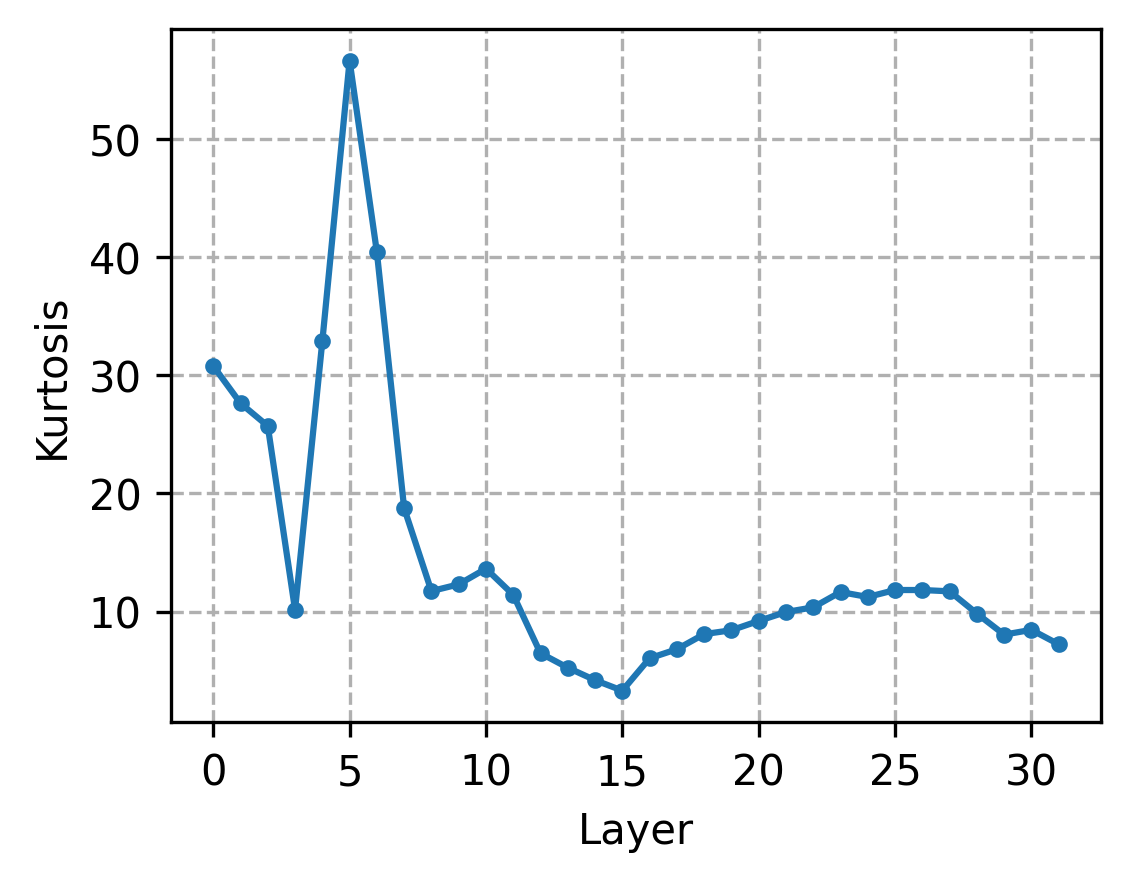}
    }
    \hfill
    \subfigure[\(m=32\)]{\includegraphics[width=0.22\textwidth]{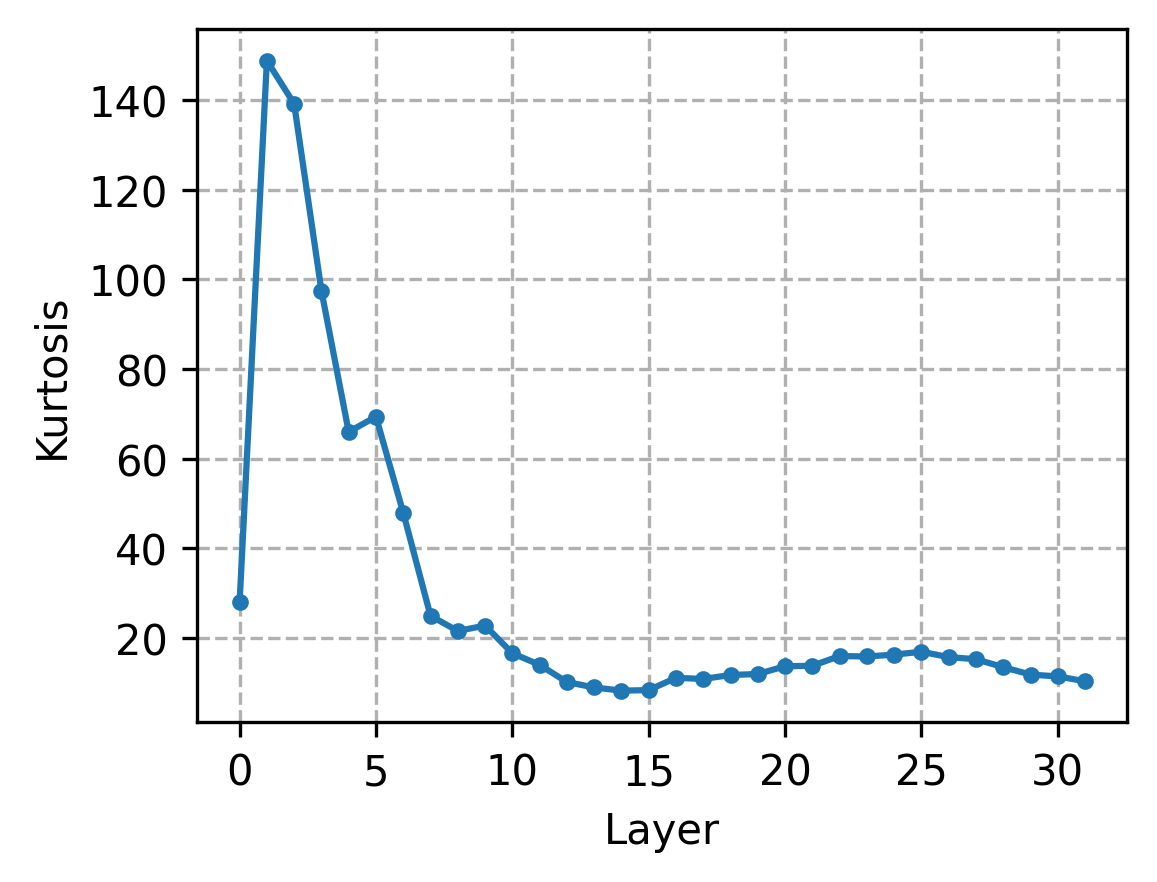}
    }
    \hfill
    \subfigure[\(m=48\)]{\includegraphics[width=0.22\textwidth]{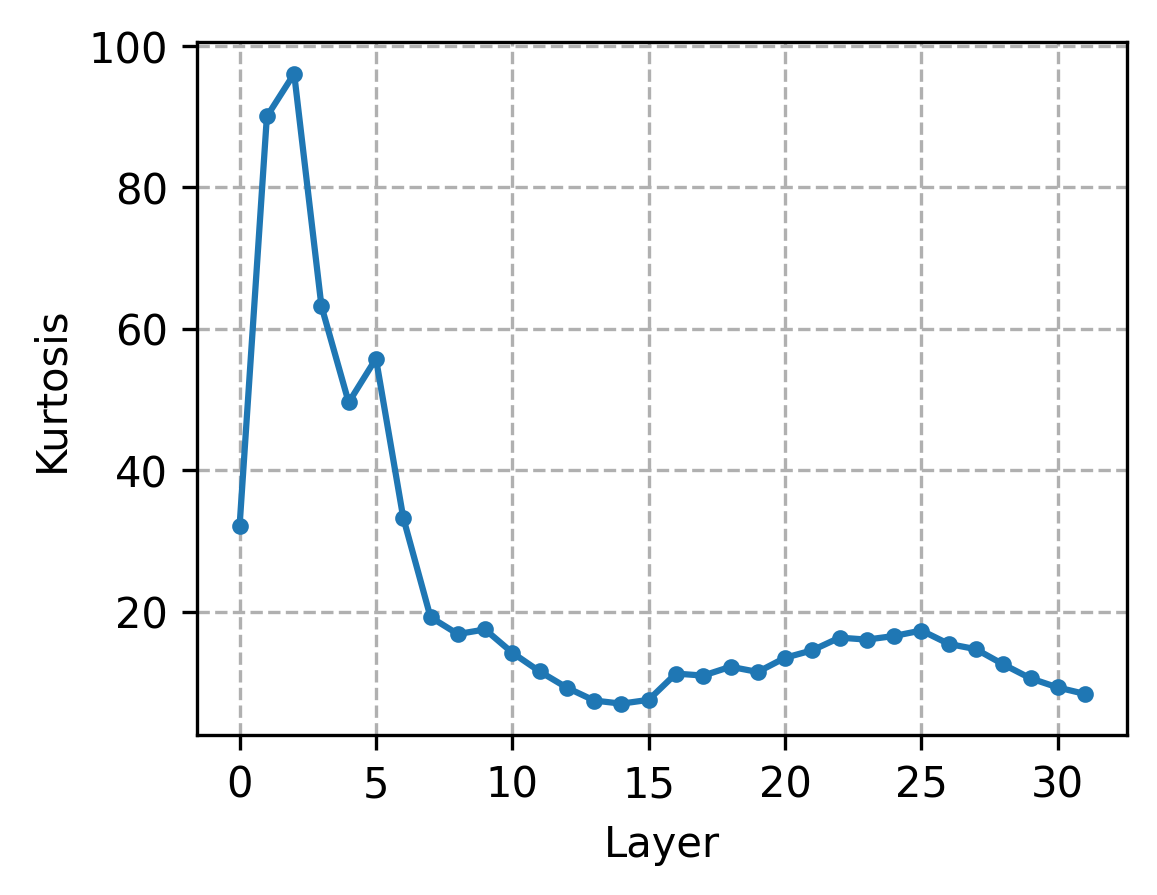}
    }
    \hfill
    \subfigure[\(m=64\)]{\includegraphics[width=0.22\textwidth]{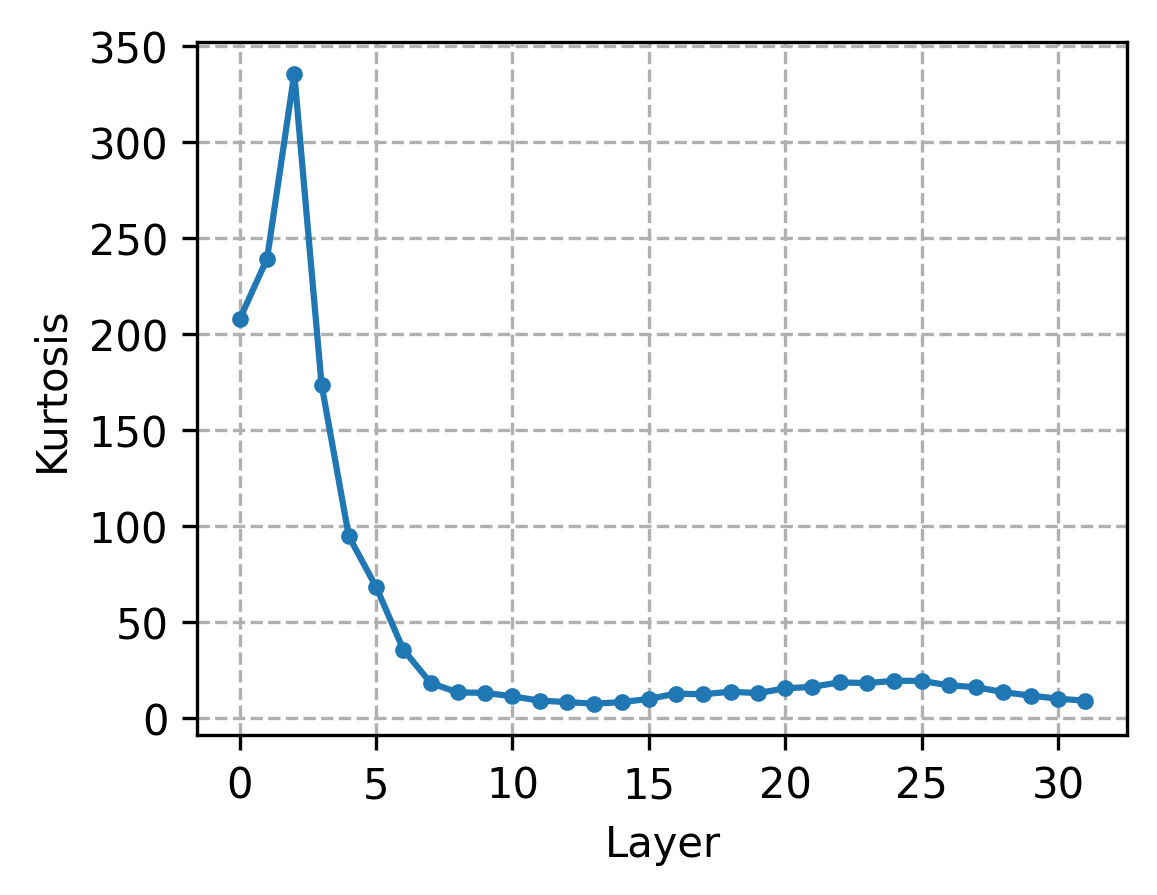}
    }
    \hfill
    \subfigure[\(m=80\)]{\includegraphics[width=0.22\textwidth]{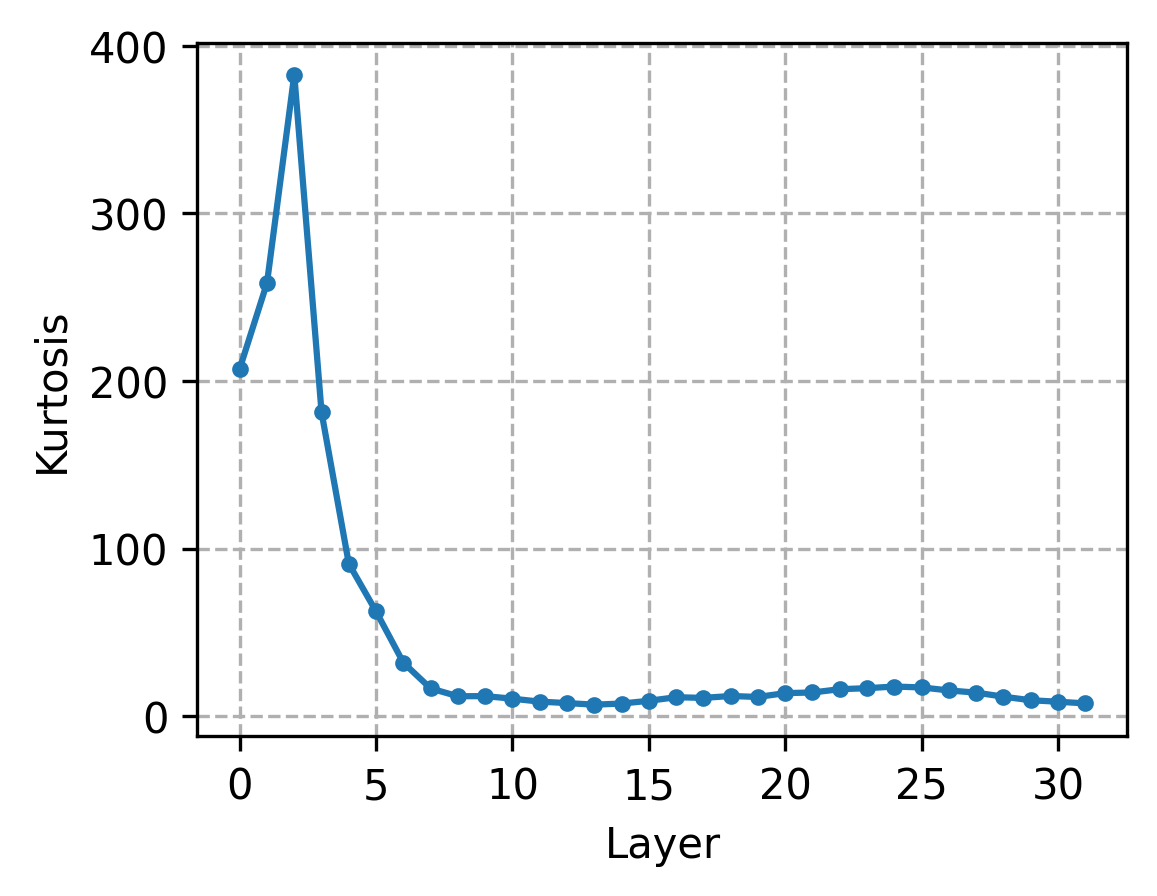}
    }
    \hfill
    \subfigure[\(m=96\)]{\includegraphics[width=0.22\textwidth]{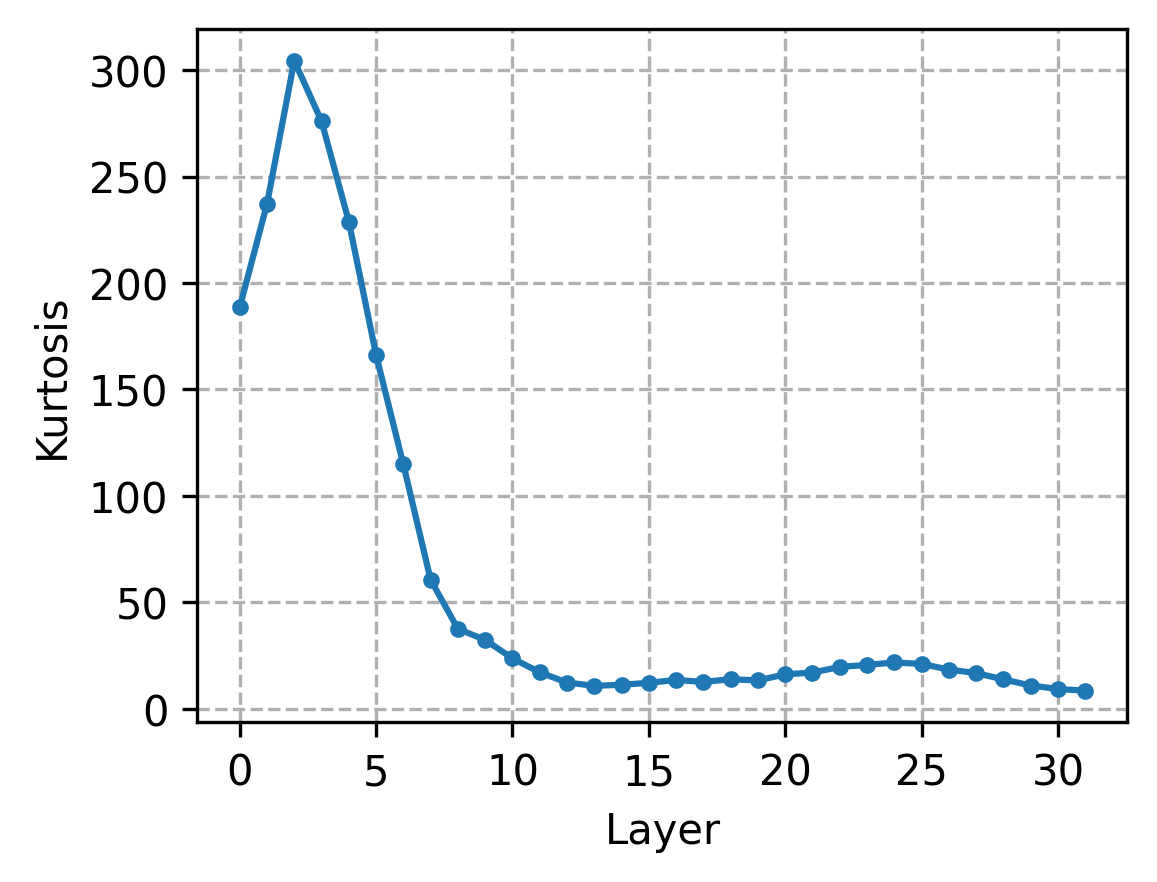}
    }
    \hfill
    \subfigure[\(m=112\)]{\includegraphics[width=0.22\textwidth]{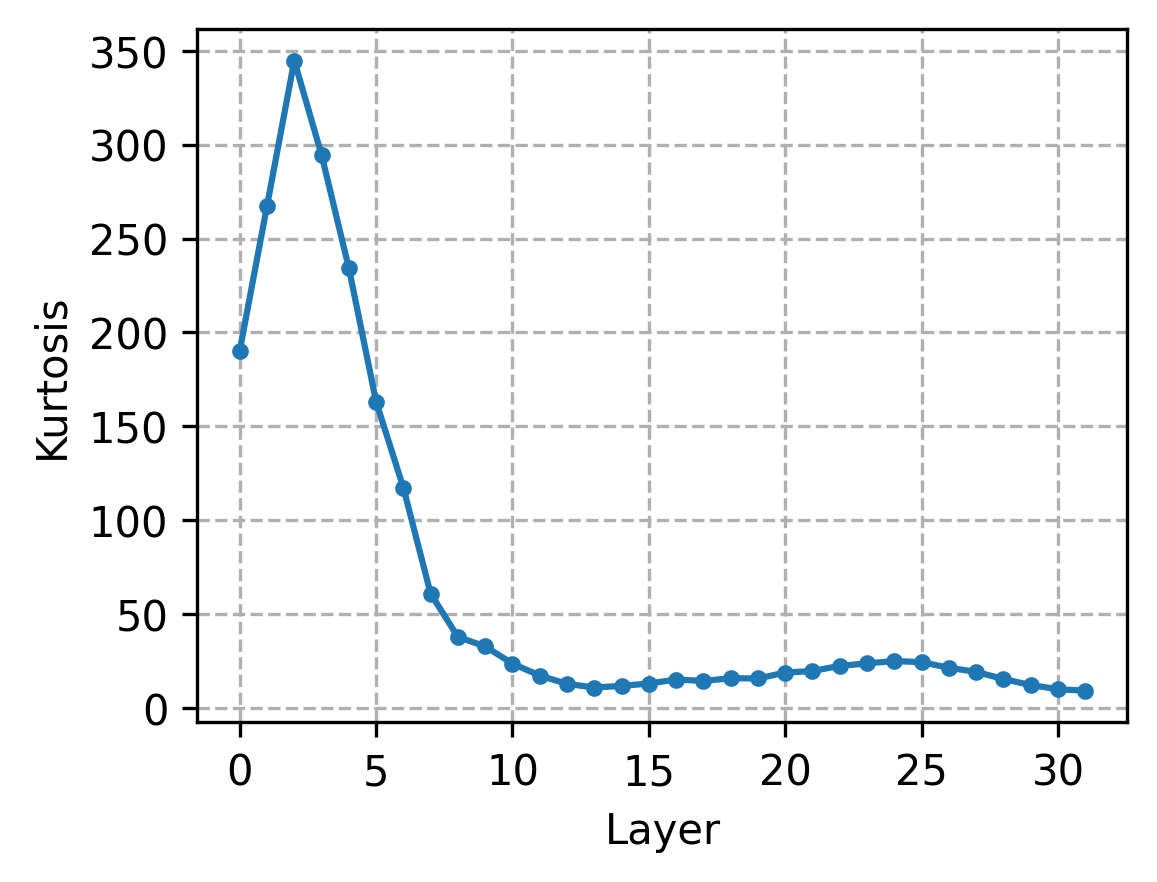}
    }
    \hfill
    \subfigure[\(m=128\)]{\includegraphics[width=0.22\textwidth]{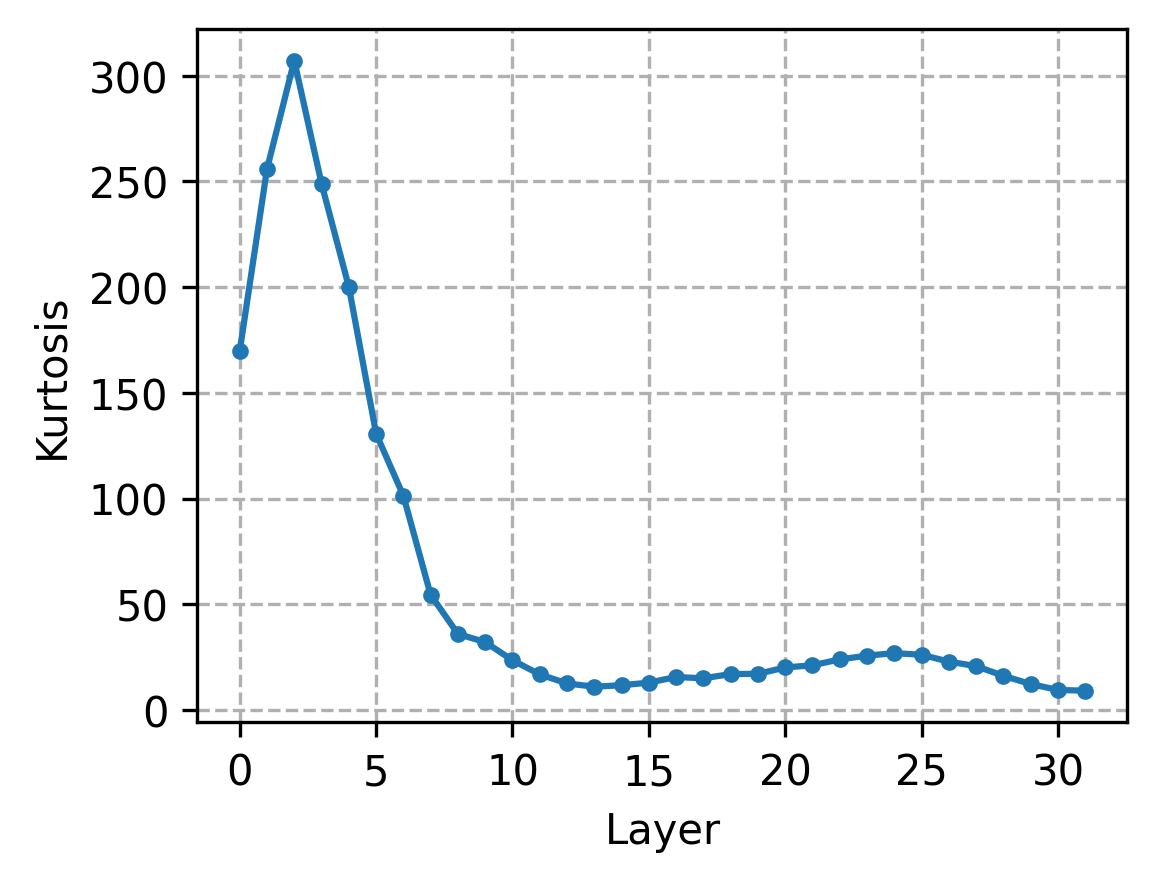}
    }
  \caption{In Llama3.1-8B, the superposition distribution converges before \( m = 128 \).}
  \label{fig:superposition_converges_llama3.1-8b}
\end{figure*}

The choice of \( m \) in Section Knowledge in Superposition, is crucial because if \( m \) is not large enough, it will be insufficient to represent a relatively complete superposition distribution with \( m \times m \) data points. Therefore, we will demonstrate that \( m = 128 \), which means that \(128\times128\) data points are completely adequate to depict a relatively complete superposition distribution. We will visualize the superposition distribution for values of \( m \) ranging from 16 to 128 using a step size of 16. As shown in Figure~\ref{fig:superposition_converges_gpt2-small}-~\ref{fig:superposition_converges_llama3.1-8b}, it can be observed that the distribution shape has converged before reaching \( m = 128 \). Thus, \( m = 128 \), or 128\*128 data points, is fully sufficient to illustrate a relatively complete superposition distribution in our study.

\begin{table*}
    \centering
    \caption{Top 20 knowledge pairs in GPT2-Small.} 
    \begin{tabular}{lp{0.55\textwidth}} 
    \toprule
          \textbf{\(p(i,j)\)}& \multicolumn{1}{c}{\textbf{Item \(i\) \& \(j\)}}\\ 
         \midrule
         1.00& [University of Washington] is based in [Seattle].\\& [University of Washington] is located in [Seattle].\\ 
         \midrule
         1.00& [University of Washington] is located in [Seattle].\\& [University of Washington] is based in [Seattle].\\ 
         \midrule
         0.98& [Vladimir Mayakovsky] is a native speaker of [Russian].\\& [Vladimir Bukovsky] is a native speaker of [Russian].\\ 
         \midrule
         0.97& The native language of [Patrick Rambaud] is [French].\\& The native language of [Pierre Trabaud] is [French].\\ 
         \midrule
         0.94& [Mac OS X 10.1], a product developed by [Apple].\\& [Windows 8.1], a product developed by [Microsoft].\\ 
         \midrule
         0.93& In [Independent State of Croatia], the language spoken is [Croatian].\\& The official language of [Croatia] is [Croatian].\\ 
         \midrule
         0.92& The official language of [Croatia] is [Croatian].\\& In [Independent State of Croatia], the language spoken is [Croatian].\\ 
         \midrule
         0.90& The native language of [Pierre Trabaud] is [French].\\& The native language of [Patrick Rambaud] is [French].\\ 
         \midrule
         0.87& [Vladimir Bukovsky] is a native speaker of [Russian].\\& [Vladimir Mayakovsky] is a native speaker of [Russian].\\ 
         \midrule
         0.86& [Kingdom of Afghanistan]'s capital city, [Kabul].\\& The official religion of [Afghanistan] is [Islam].\\ 
         \midrule
         0.85& The official religion of [Afghanistan] is [Islam].\\& [Kingdom of Afghanistan]'s capital city, [Kabul].\\ 
         \midrule
         0.82& [Windows 8.1], a product developed by [Microsoft].\\& [Mac OS X 10.1], a product developed by [Apple].\\ 
         \midrule
         0.74& The native language of [Ilya Nikolaevich Ulyanov] is [Russian].\\& [Nikolay Nekrasov] is a native speaker of [Russian].\\ 
         \midrule
         0.73& [Nikolay Nekrasov] is a native speaker of [Russian].\\& The native language of [Ilya Nikolaevich Ulyanov] is [Russian].\\ 
         \midrule
         0.71& [Windows 2000] was developed by [Microsoft].\\& [Windows Server 2012] was developed by [Microsoft].\\ 
         \midrule
         0.71& [Internet Explorer 5] was developed by [Microsoft].\\& [Microsoft Internet Explorer 6], a product created by [Microsoft].\\ 
         \midrule
         0.68& [Windows 2000] was developed by [Microsoft].\\& [Windows Vista] is a product of [Microsoft].\\ 
         \midrule
         0.65& [Microsoft Internet Explorer 6], a product created by [Microsoft].\\& [Windows Server 2012] was developed by [Microsoft].\\ 
         \midrule
         0.61& [Windows 95], a product of [Microsoft].\\& [Windows 2000] was developed by [Microsoft].\\ 
         \midrule
         0.61& [Nokia N70], produced by [Nokia].\\& [Nokia N80] is produced by [Nokia].\\ 
    \bottomrule
    \end{tabular}
    \label{tab:case_study_gpt2-small}
\end{table*}

\begin{table*}
    \centering
    \caption{Top 20 knowledge pairs in GPT2-Medium.} 
    \begin{tabular}{lp{0.55\textwidth}} 
    \toprule
          \textbf{\(p(i,j)\)}& \multicolumn{1}{c}{\textbf{Item \(i\) \& \(j\)}}\\ 
         \midrule
         1.00& [University of Washington] is based in [Seattle].\\& [University of Washington] is located in [Seattle].\\ 
         \midrule
         1.00& [University of Washington] is located in [Seattle].\\& [University of Washington] is based in [Seattle].\\ 
         \midrule
         0.97& [Vladimir Bukovsky] is a native speaker of [Russian].\\& [Vladimir Mayakovsky] is a native speaker of [Russian].\\ 
         \midrule
         0.97& The native language of [Patrick Rambaud] is [French].\\& The native language of [Pierre Trabaud] is [French].\\ 
         \midrule
         0.96& The native language of [Pierre Trabaud] is [French].\\& The native language of [Patrick Rambaud] is [French].\\ 
         \midrule
         0.95& [Vladimir Mayakovsky] is a native speaker of [Russian].\\& [Vladimir Bukovsky] is a native speaker of [Russian].\\ 
         \midrule
         0.94& [Internet Explorer 5] was developed by [Microsoft].\\& [Microsoft Internet Explorer 6], a product created by [Microsoft].\\ 
         \midrule
         0.92& [Kingdom of Afghanistan]'s capital city, [Kabul].\\& The official religion of [Afghanistan] is [Islam].\\ 
         \midrule
         0.92& The official language of [Croatia] is [Croatian].\\& In [Independent State of Croatia], the language spoken is [Croatian].\\ 
         \midrule
         0.92& In [Independent State of Croatia], the language spoken is [Croatian].\\& The official language of [Croatia] is [Croatian].\\ 
         \midrule
         0.86& The official religion of [Afghanistan] is [Islam].\\& [Kingdom of Afghanistan]'s capital city, [Kabul].\\ 
         \midrule
         0.86& [Mac OS X 10.1], a product developed by [Apple].\\& [Windows 8.1], a product developed by [Microsoft].\\ 
         \midrule
         0.75& [Windows 8.1], a product developed by [Microsoft].\\& [Mac OS X 10.1], a product developed by [Apple].\\ 
         \midrule
         0.73& [Nikolay Nekrasov] is a native speaker of [Russian].\\& The native language of [Ilya Nikolaevich Ulyanov] is [Russian].\\ 
         \midrule
         0.68& The native language of [Ilya Nikolaevich Ulyanov] is [Russian].\\& [Nikolay Nekrasov] is a native speaker of [Russian].\\ 
         \midrule
         0.59& [Windows 2000] was developed by [Microsoft].\\& [Windows Server 2012] was developed by [Microsoft].\\ 
         \midrule
         0.57& [Windows 95], a product of [Microsoft].\\& [Windows 2000] was developed by [Microsoft].\\ 
         \midrule
         0.57& [Windows 2000] was developed by [Microsoft].\\& [Windows 95], a product of [Microsoft].\\ 
         \midrule
         0.56& [Nokia N80] is produced by [Nokia].\\& [Nokia N70], produced by [Nokia].\\ 
         \midrule
         0.53& [Microsoft Internet Explorer 6], a product created by [Microsoft].\\& [Windows Server 2012] was developed by [Microsoft].\\ 
    \bottomrule
    \end{tabular}
    \label{tab:case_study_gpt2-medium}
\end{table*}

\begin{table*}
    \centering
    \caption{Top 20 knowledge pairs in GPT2-Large.} 
    \begin{tabular}{lp{0.55\textwidth}} 
    \toprule
          \textbf{\(p(i,j)\)}& \multicolumn{1}{c}{\textbf{Item \(i\) \& \(j\)}}\\ 
         \midrule
         1.00& [University of Washington] is based in [Seattle].\\& [University of Washington] is located in [Seattle].\\ 
         \midrule
         1.00& [University of Washington] is located in [Seattle].\\& [University of Washington] is based in [Seattle].\\ 
         \midrule
         0.98& The official language of [Croatia] is [Croatian].\\& In [Independent State of Croatia], the language spoken is [Croatian].\\ 
         \midrule
         0.98& In [Independent State of Croatia], the language spoken is [Croatian].\\& The official language of [Croatia] is [Croatian].\\ 
         \midrule
         0.98& The official religion of [Afghanistan] is [Islam].\\& [Kingdom of Afghanistan]'s capital city, [Kabul].\\ 
         \midrule
         0.97& [Kingdom of Afghanistan]'s capital city, [Kabul].\\& The official religion of [Afghanistan] is [Islam].\\ 
         \midrule
         0.97& The native language of [Pierre Trabaud] is [French].\\& The native language of [Patrick Rambaud] is [French].\\ 
         \midrule
         0.96& The native language of [Patrick Rambaud] is [French].\\& The native language of [Pierre Trabaud] is [French].\\ 
         \midrule
         0.95& [Vladimir Bukovsky] is a native speaker of [Russian].\\& [Vladimir Mayakovsky] is a native speaker of [Russian].\\ 
         \midrule
         0.94& [Vladimir Mayakovsky] is a native speaker of [Russian].\\& [Vladimir Bukovsky] is a native speaker of [Russian].\\ 
         \midrule
         0.90& The native language of [Ilya Nikolaevich Ulyanov] is [Russian].\\& [Nikolay Nekrasov] is a native speaker of [Russian].\\ 
         \midrule
         0.87& [Mac OS X 10.1], a product developed by [Apple].\\& [Windows 8.1], a product developed by [Microsoft].\\ 
         \midrule
         0.83& [Nikolay Nekrasov] is a native speaker of [Russian].\\& The native language of [Ilya Nikolaevich Ulyanov] is [Russian].\\ 
         \midrule
         0.71& [Windows 8.1], a product developed by [Microsoft].\\& [Mac OS X 10.1], a product developed by [Apple].\\ 
         \midrule
         0.68& [Microsoft Internet Explorer 6], a product created by [Microsoft].\\& [Internet Explorer 5] was developed by [Microsoft].\\ 
         \midrule
         0.50& [Internet Explorer 5] was developed by [Microsoft].\\& [Microsoft Internet Explorer 6], a product created by [Microsoft].\\ 
         \midrule
         0.41& [Honda S2000] is developed by [Honda].\\& [Windows 2000] was developed by [Microsoft].\\ 
         \midrule
         0.38& [Nissan NV200] is developed by [Nissan].\\& [Honda S600], produced by [Honda].\\ 
         \midrule
         0.37& [Honda CBX] is developed by [Honda].\\& [Final Fantasy X], a product created by [Square].\\ 
         \midrule
         0.35& [Honda S600], produced by [Honda].\\& [Honda CB550] is produced by [Honda].\\ 
    \bottomrule
    \end{tabular}
    \label{tab:case_study_gpt2-large}
\end{table*}

\begin{table*}
    \centering
    \caption{Top 20 knowledge pairs in GPT-J-6B.} 
    \begin{tabular}{lp{0.55\textwidth}} 
    \toprule
          \textbf{\(p(i,j)\)}& \multicolumn{1}{c}{\textbf{Item \(i\) \& \(j\)}}\\ 
         \midrule
         1.00& [Kingdom of Afghanistan]'s capital city, [Kabul].\\& The official religion of [Afghanistan] is [Islam].\\ 
         \midrule
         1.00& The native language of [Ilya Nikolaevich Ulyanov] is [Russian].\\& [Nikolay Nekrasov] is a native speaker of [Russian].\\ 
         \midrule
         1.00& The native language of [Pierre Trabaud] is [French].\\& The native language of [Patrick Rambaud] is [French].\\ 
         \midrule
         1.00& [University of Washington] is based in [Seattle].\\& [University of Washington] is located in [Seattle].\\ 
         \midrule
         1.00& The official language of [Croatia] is [Croatian].\\& In [Independent State of Croatia], the language spoken is [Croatian].\\ 
         \midrule
         1.00& In [Independent State of Croatia], the language spoken is [Croatian].\\& The official language of [Croatia] is [Croatian].\\ 
         \midrule
         1.00& [Vladimir Mayakovsky] is a native speaker of [Russian].\\& [Vladimir Bukovsky] is a native speaker of [Russian].\\ 
         \midrule
         1.00& The native language of [Patrick Rambaud] is [French].\\& The native language of [Pierre Trabaud] is [French].\\ 
         \midrule
         1.00& The official religion of [Afghanistan] is [Islam].\\& [Kingdom of Afghanistan]'s capital city, [Kabul].\\ 
         \midrule
         1.00& [Windows 8.1], a product developed by [Microsoft].\\& [Mac OS X 10.1], a product developed by [Apple].\\ 
         \midrule
         1.00& [Vladimir Bukovsky] is a native speaker of [Russian].\\& [Vladimir Mayakovsky] is a native speaker of [Russian].\\ 
         \midrule
         1.00& [University of Washington] is located in [Seattle].\\& [University of Washington] is based in [Seattle].\\ 
         \midrule
         1.00& [Mac OS X 10.1], a product developed by [Apple].\\& [Windows 8.1], a product developed by [Microsoft].\\ 
         \midrule
         1.00& [Nikolay Nekrasov] is a native speaker of [Russian].\\& The native language of [Ilya Nikolaevich Ulyanov] is [Russian].\\ 
         \midrule
         0.13& [Windows 2000] was developed by [Microsoft].\\& [Honda S2000] is developed by [Honda].\\ 
         \midrule
         0.12& [Final Fantasy X], a product created by [Square].\\& [Honda CBX] is developed by [Honda].\\ 
         \midrule
         0.10& [Honda CBX] is developed by [Honda].\\& [Final Fantasy X], a product created by [Square].\\ 
         \midrule
         0.10& [Suzuki GSX-R750], produced by [Suzuki].\\& [Honda CB550] is produced by [Honda].\\ 
         \midrule
         0.07& [Honda CB550] is produced by [Honda].\\& [Suzuki GSX-R750], produced by [Suzuki].\\ 
         \midrule
         0.07& [Microsoft Internet Explorer 6], a product created by [Microsoft].\\& [Honda CB550] is produced by [Honda].\\ 
    \bottomrule
    \end{tabular}
    \label{tab:case_study_gpt-j}
\end{table*}

\section{D. Visualization of \(P\) Matrices for All Layers}\label{appendix:D. Visualization of Matrix for All Layers}

\begin{figure*}
    \centering
    \subfigure[Layer 0]{\includegraphics[width=0.22\textwidth]{fig/gpt2-small/p_matrix/known/heatmap/superposition_for_layer_0.png}
    }
    \hfill
    \subfigure[Layer 1]{\includegraphics[width=0.22\textwidth]{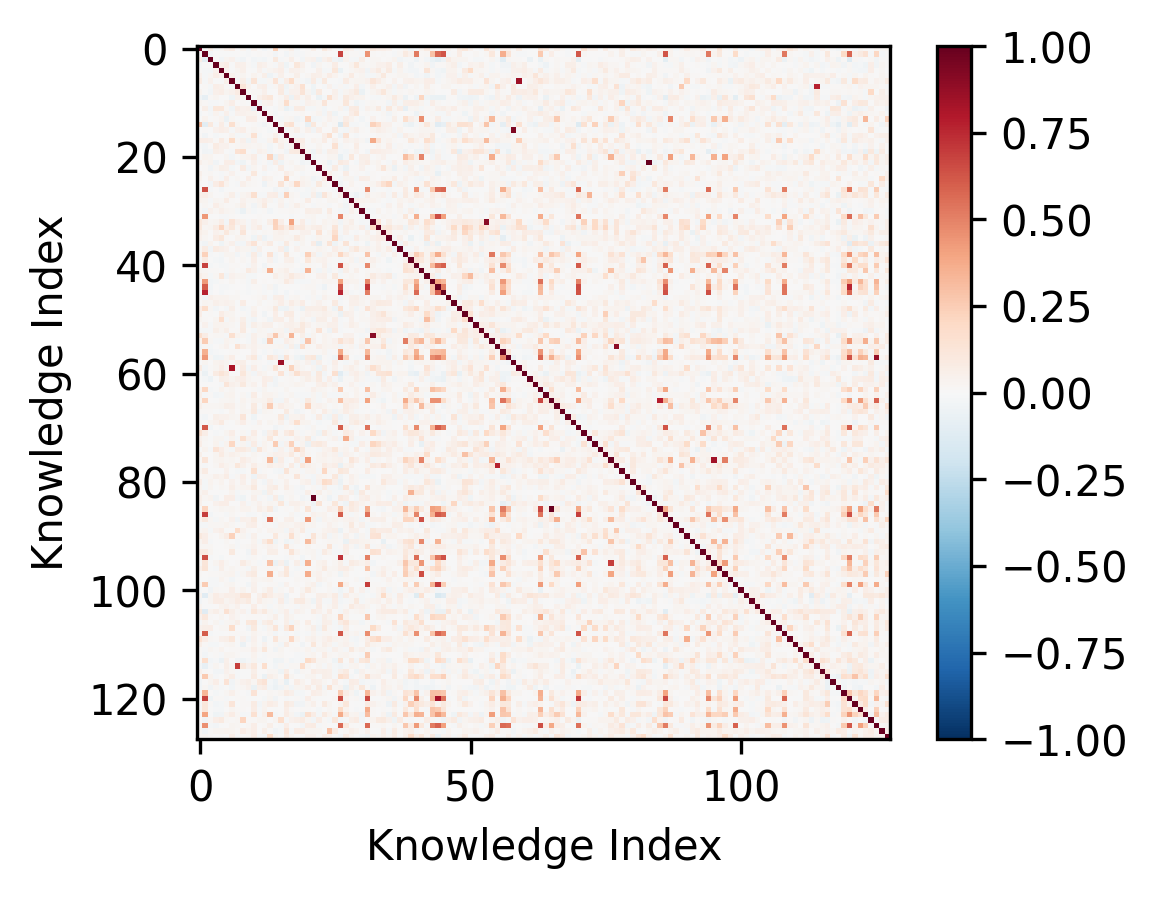}
    }
    \hfill
    \subfigure[Layer 2]{\includegraphics[width=0.22\textwidth]{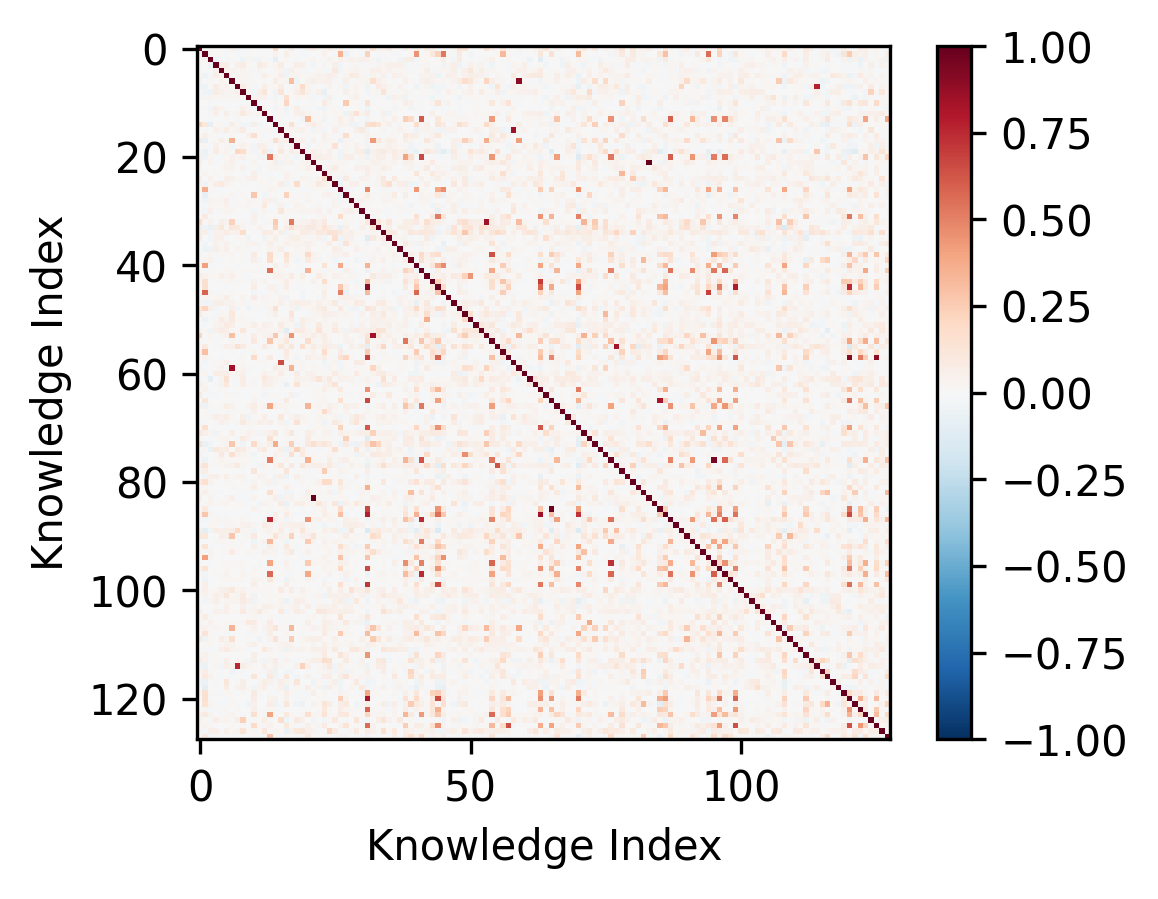}
    }
    \hfill
    \subfigure[Layer 3]{\includegraphics[width=0.22\textwidth]{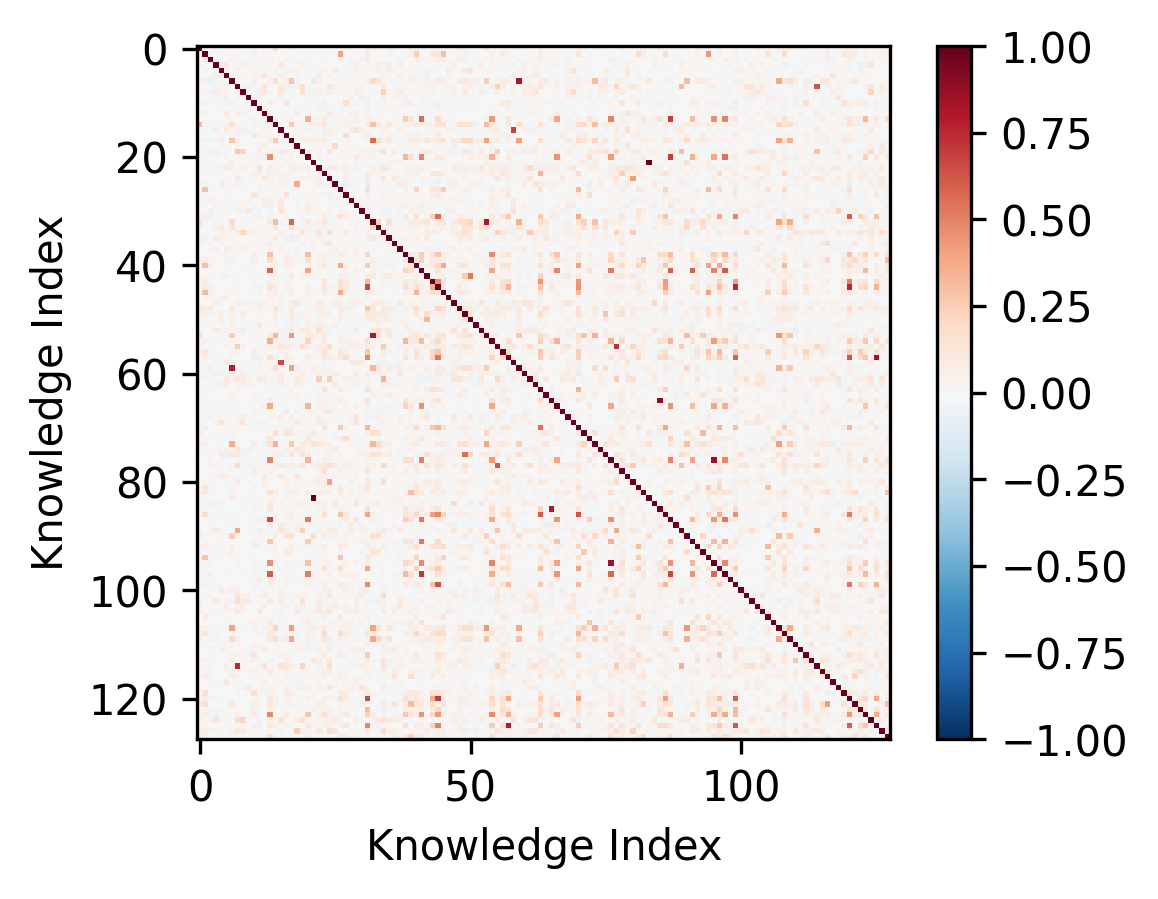}
    }
    \hfill
    \subfigure[Layer 4]{\includegraphics[width=0.22\textwidth]{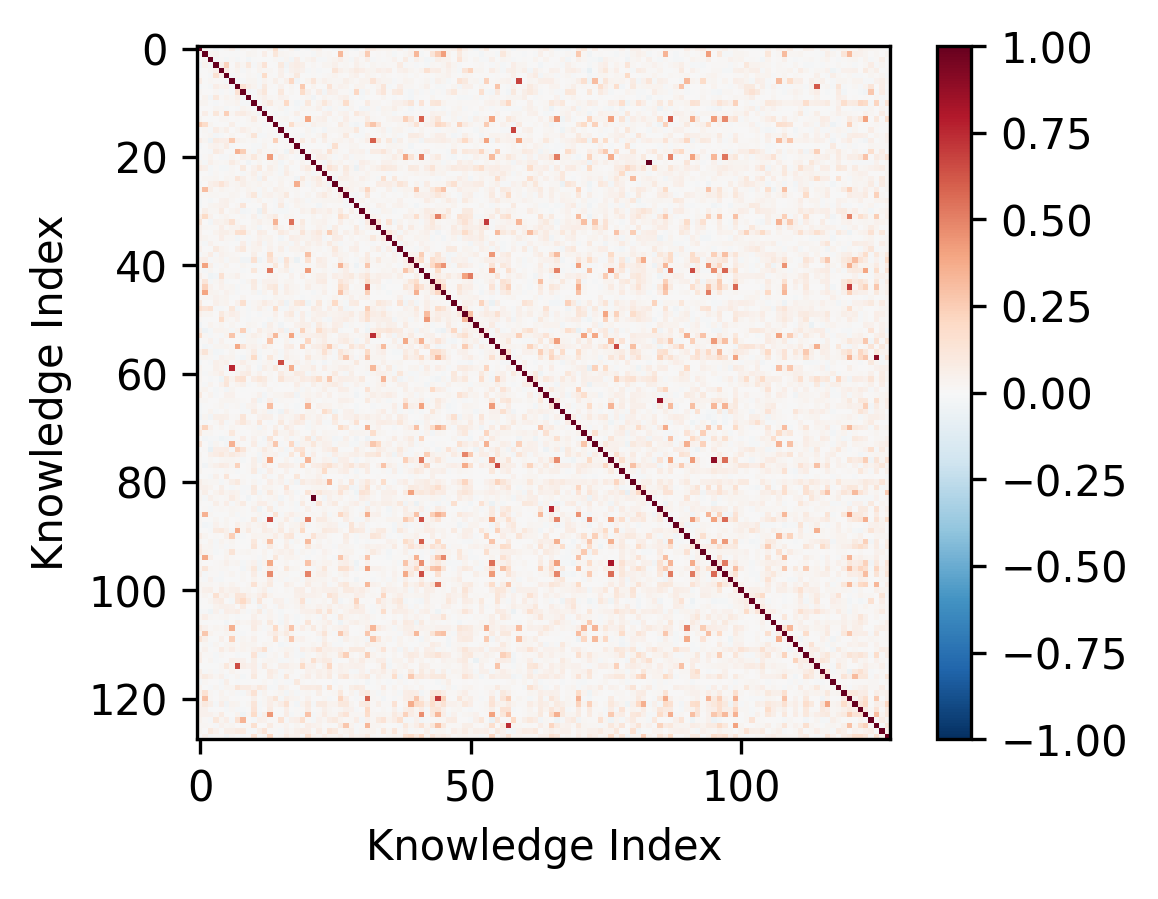}
    }
    \hfill
    \subfigure[Layer 5]{\includegraphics[width=0.22\textwidth]{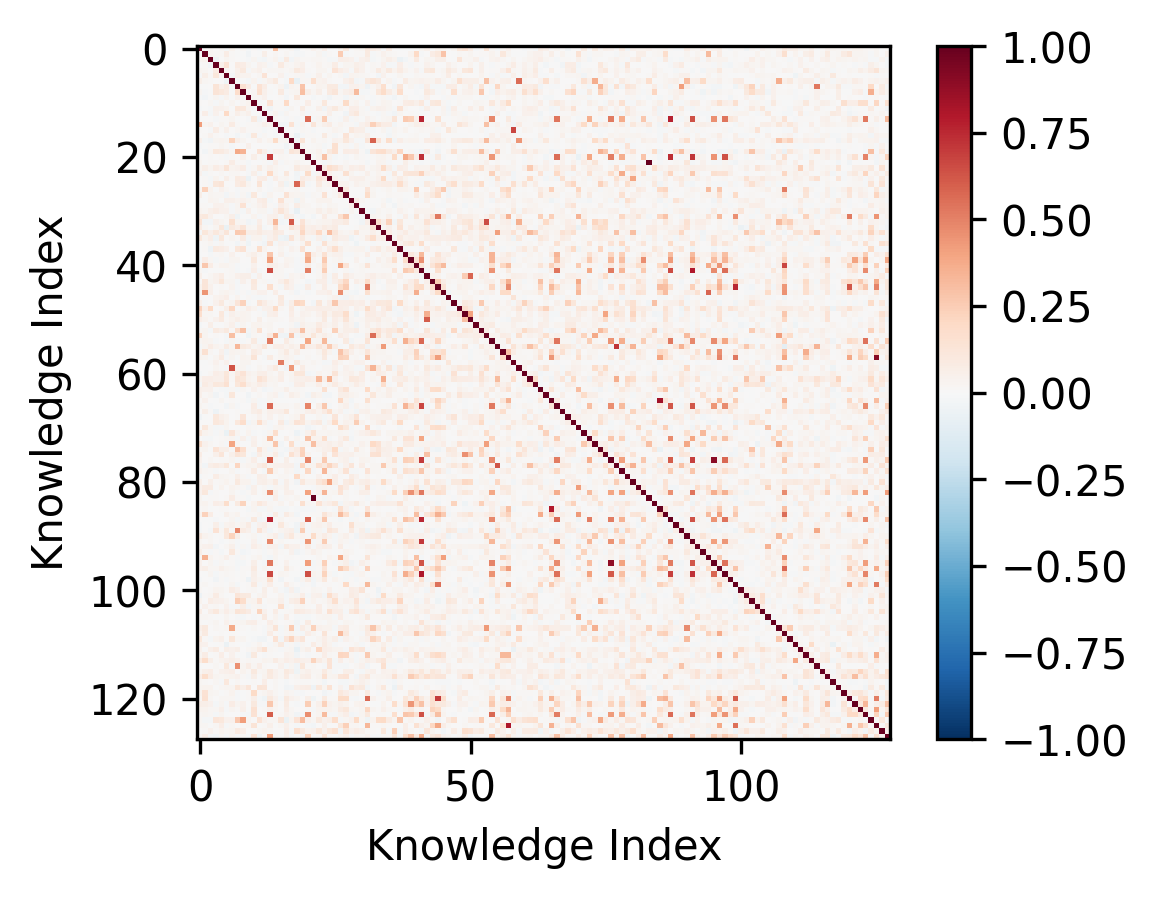}
    }
    \hfill
    \subfigure[Layer 6]{\includegraphics[width=0.22\textwidth]{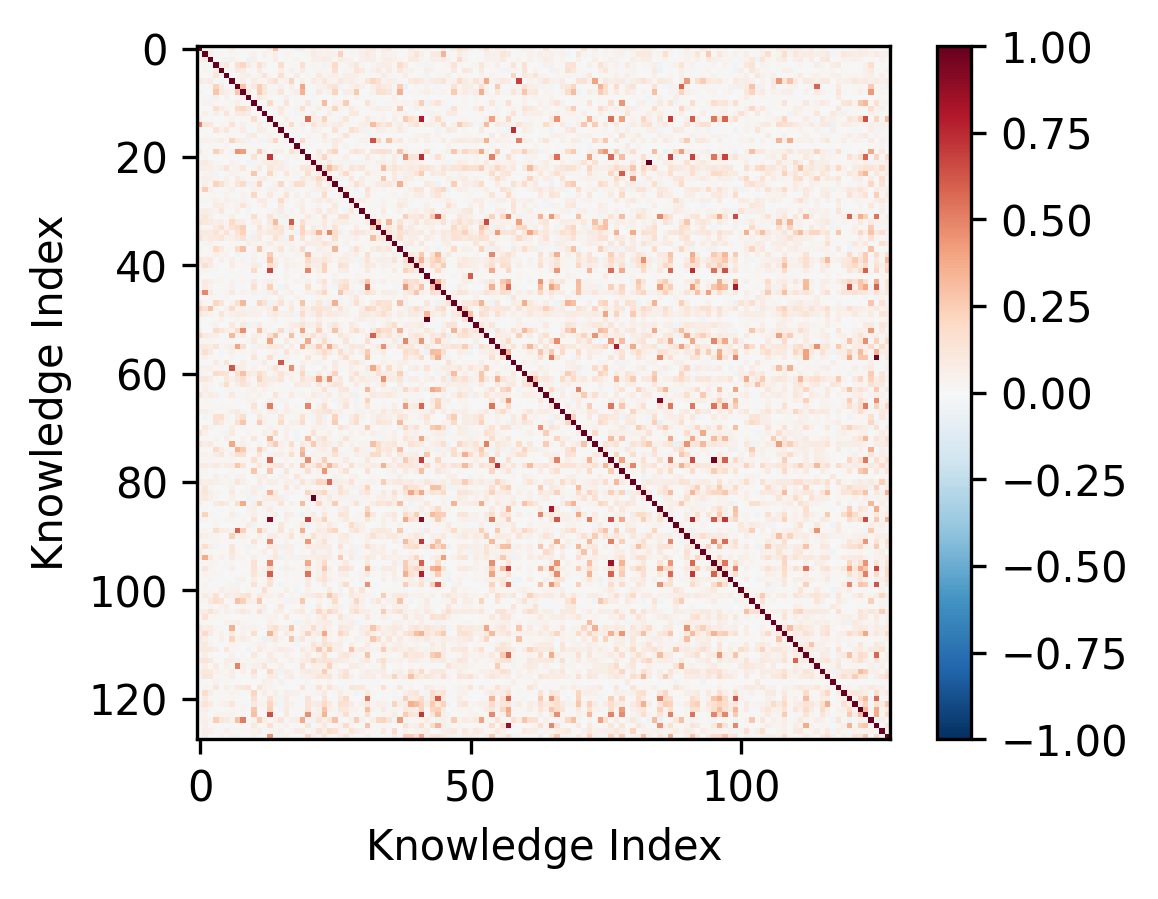}
    }
    \hfill
    \subfigure[Layer 7]{\includegraphics[width=0.22\textwidth]{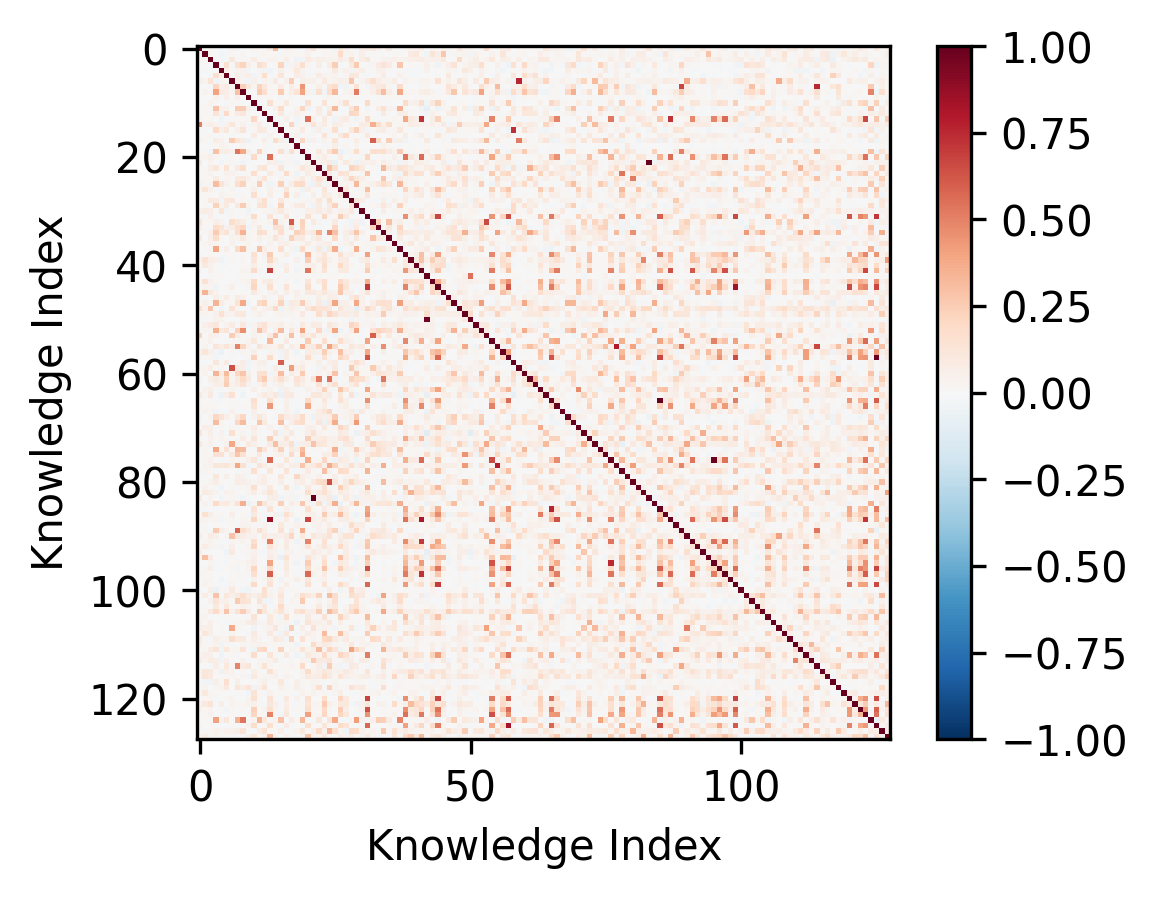}
    }
    \hfill
    \subfigure[Layer 8]{\includegraphics[width=0.22\textwidth]{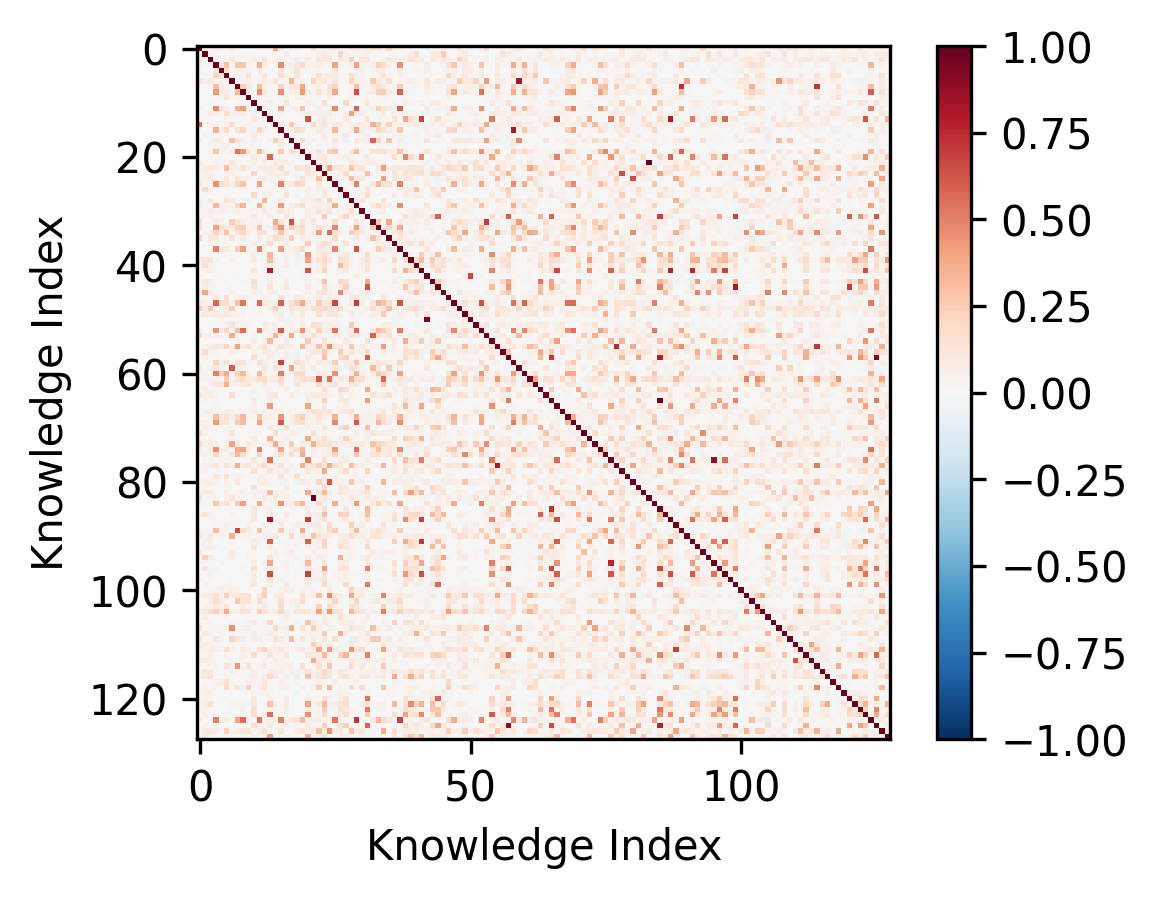}
    }
    \hfill
    \subfigure[Layer 9]{\includegraphics[width=0.22\textwidth]{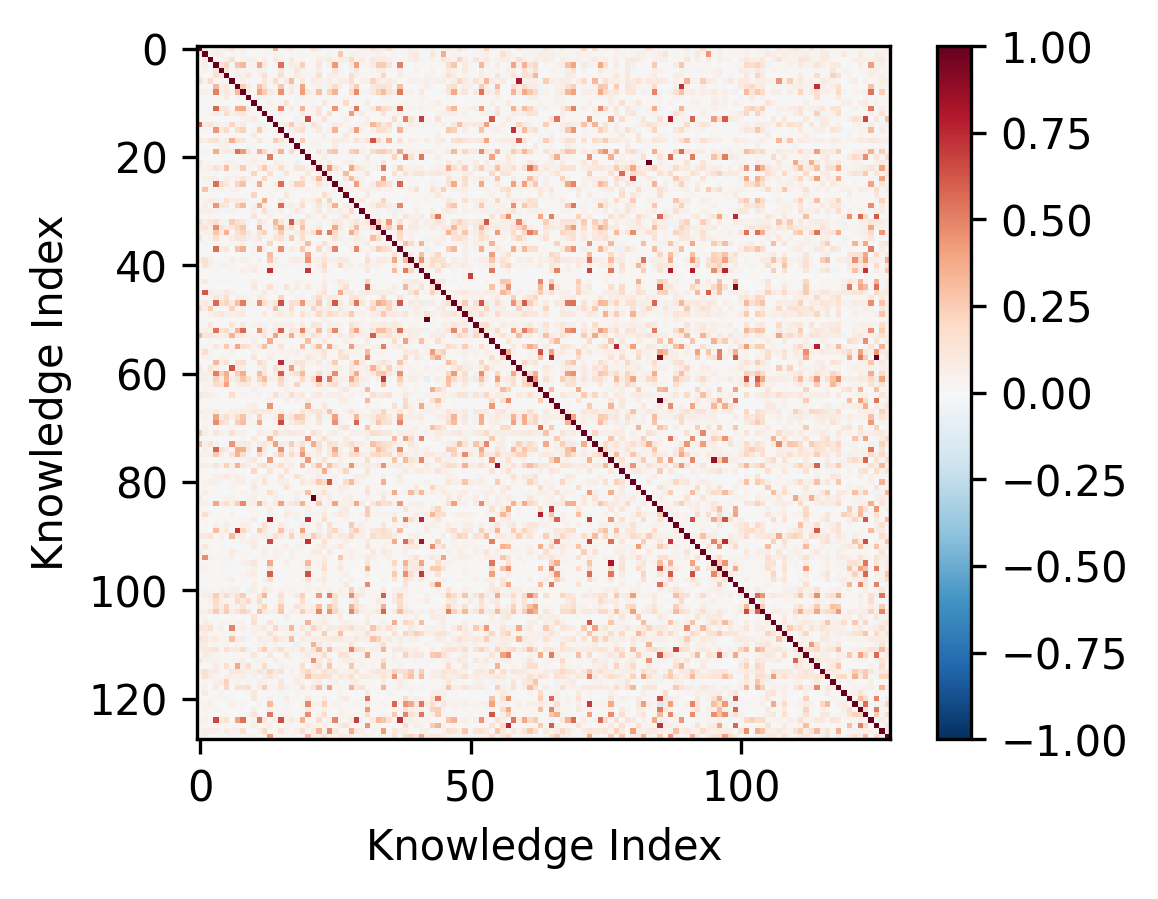}
    }
    \hfill
    \subfigure[Layer 10]{\includegraphics[width=0.22\textwidth]{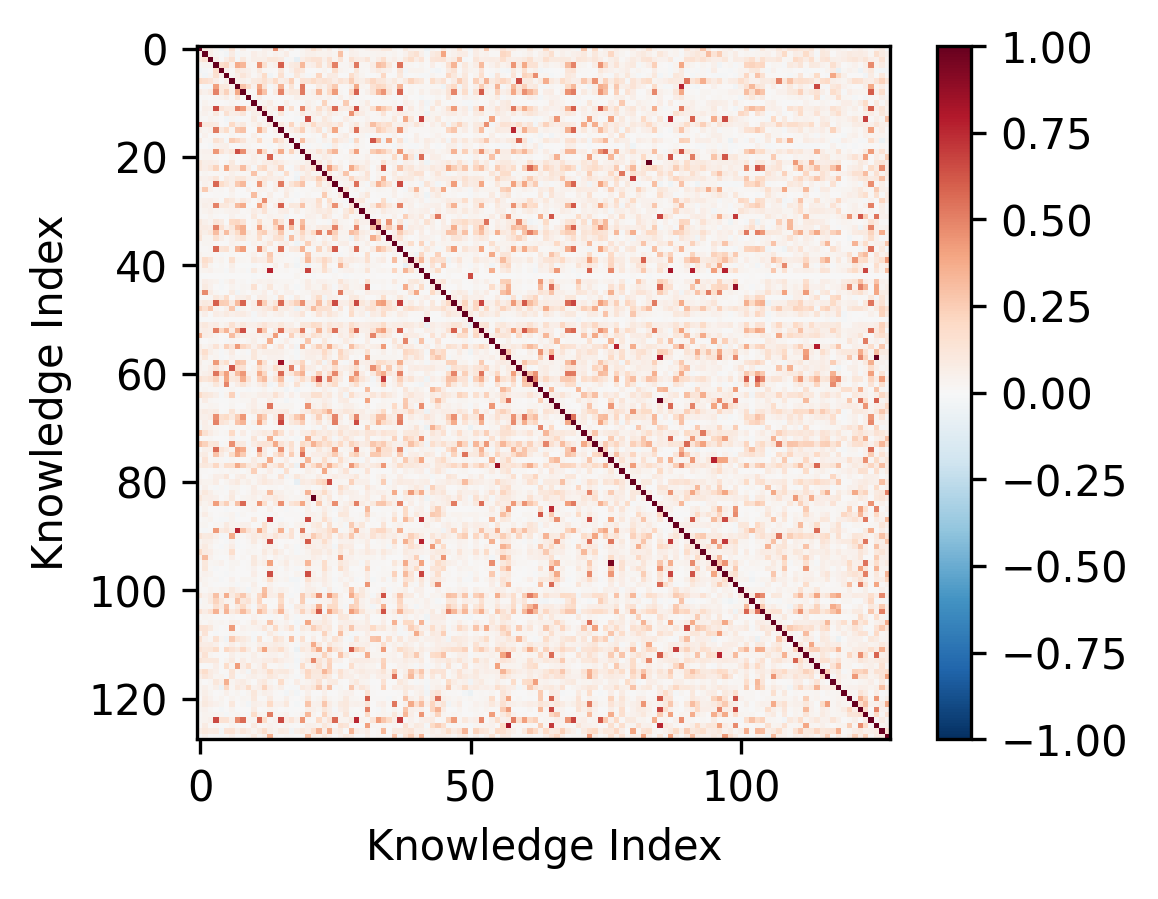}
    }
    \hfill
    \subfigure[Layer 11]{\includegraphics[width=0.22\textwidth]{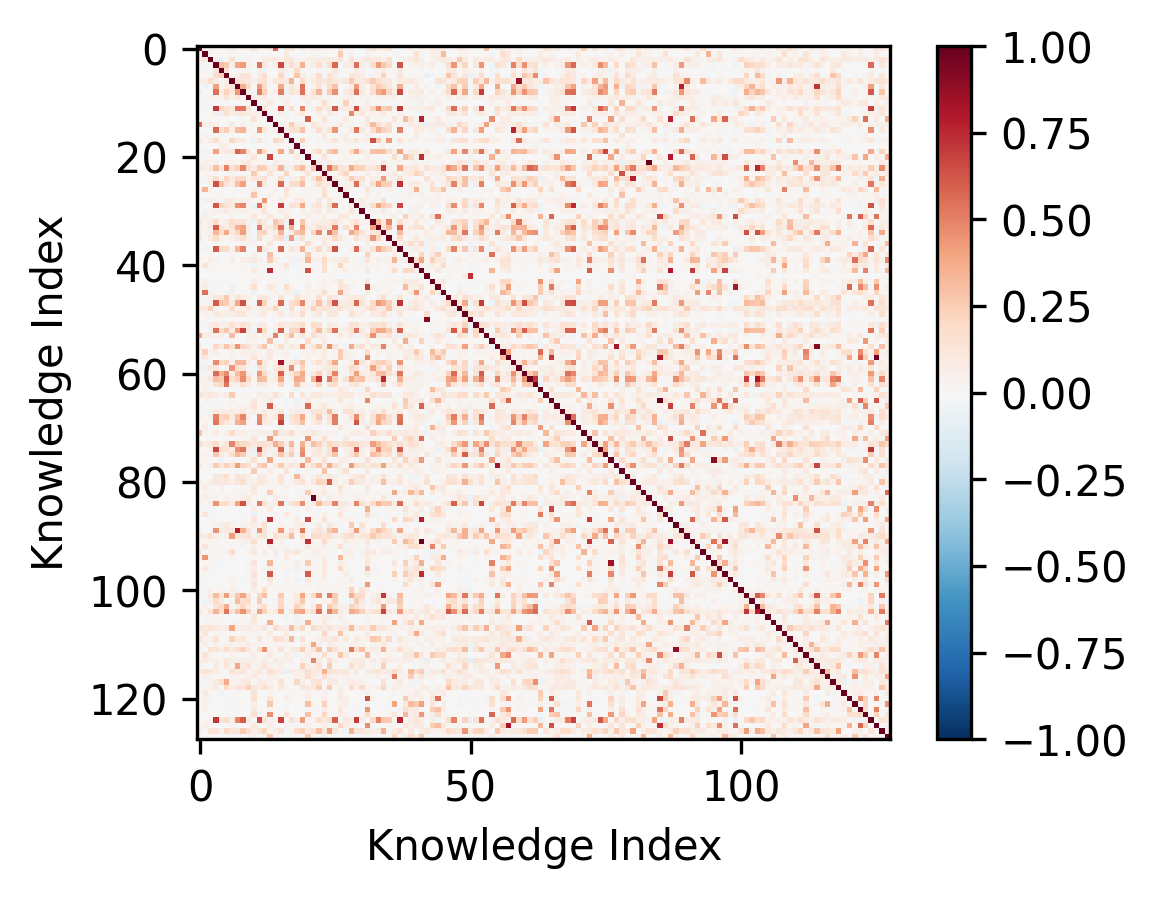}
    }
  \caption{In GPT2-Small, the visualization of \(P\) matrices across layers.}
  \label{fig:superposition_visualization_gpt2-small}
\end{figure*}

\begin{figure*}
    \centering
    \subfigure[Layer 0]{\includegraphics[width=0.22\textwidth]{fig/gpt2-medium/p_matrix/known/heatmap/superposition_for_layer_0.png}
    }
    \hfill
    \subfigure[Layer 1]{\includegraphics[width=0.22\textwidth]{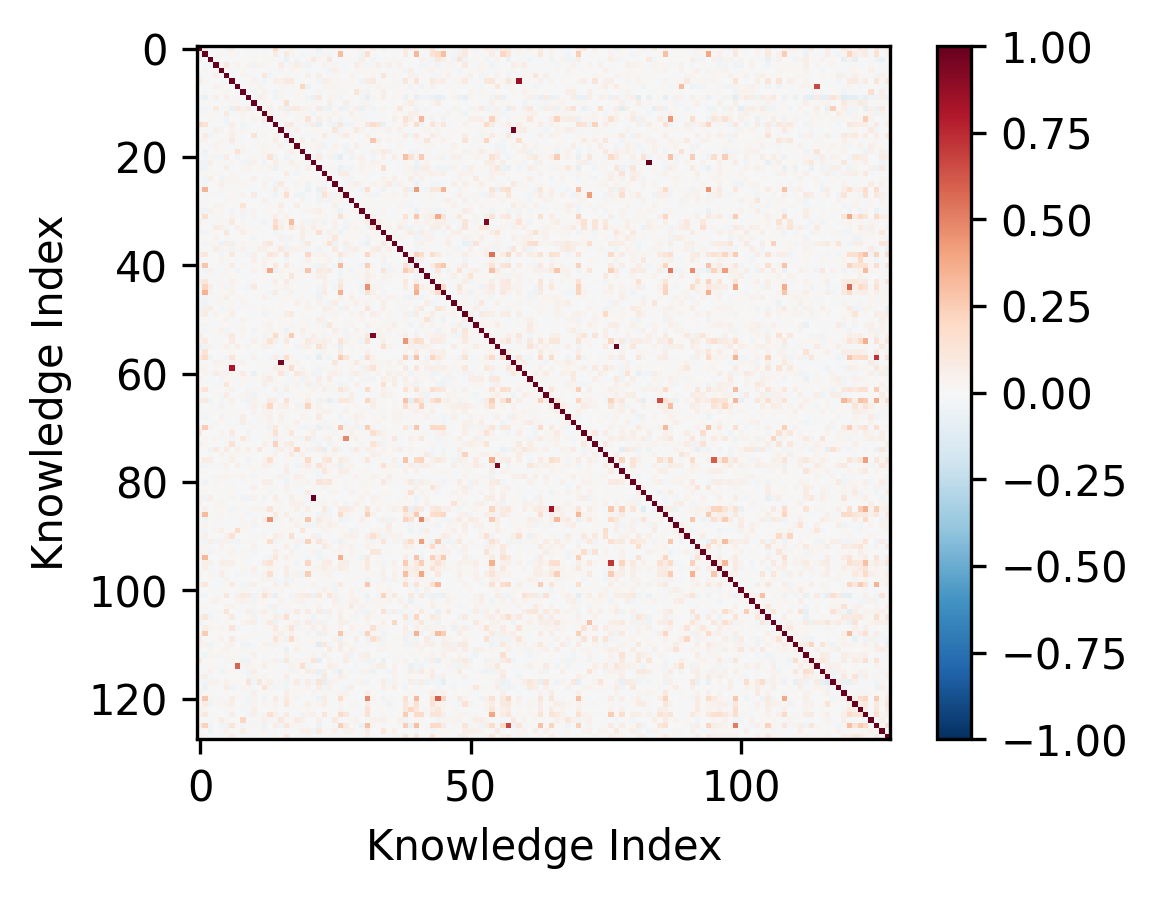}
    }
    \hfill
    \subfigure[Layer 2]{\includegraphics[width=0.22\textwidth]{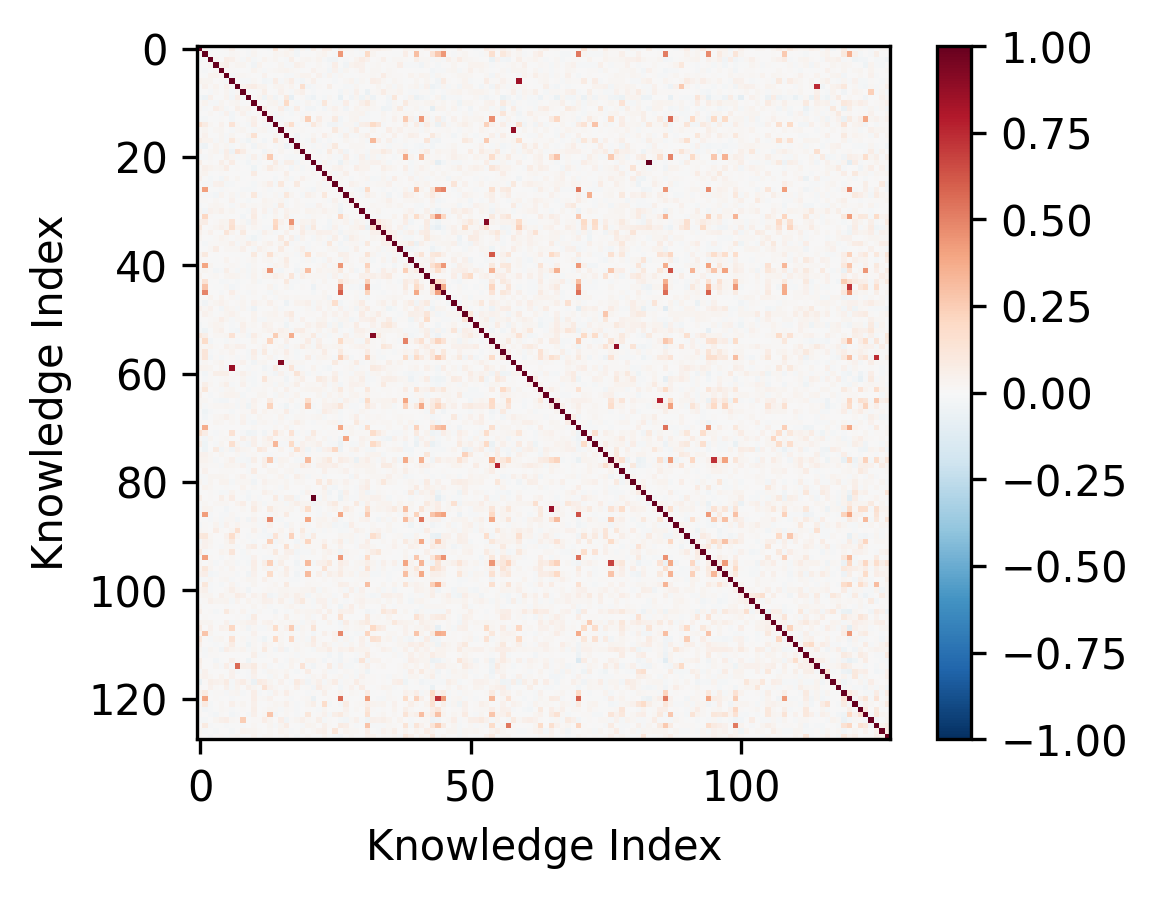}
    }
    \hfill
    \subfigure[Layer 3]{\includegraphics[width=0.22\textwidth]{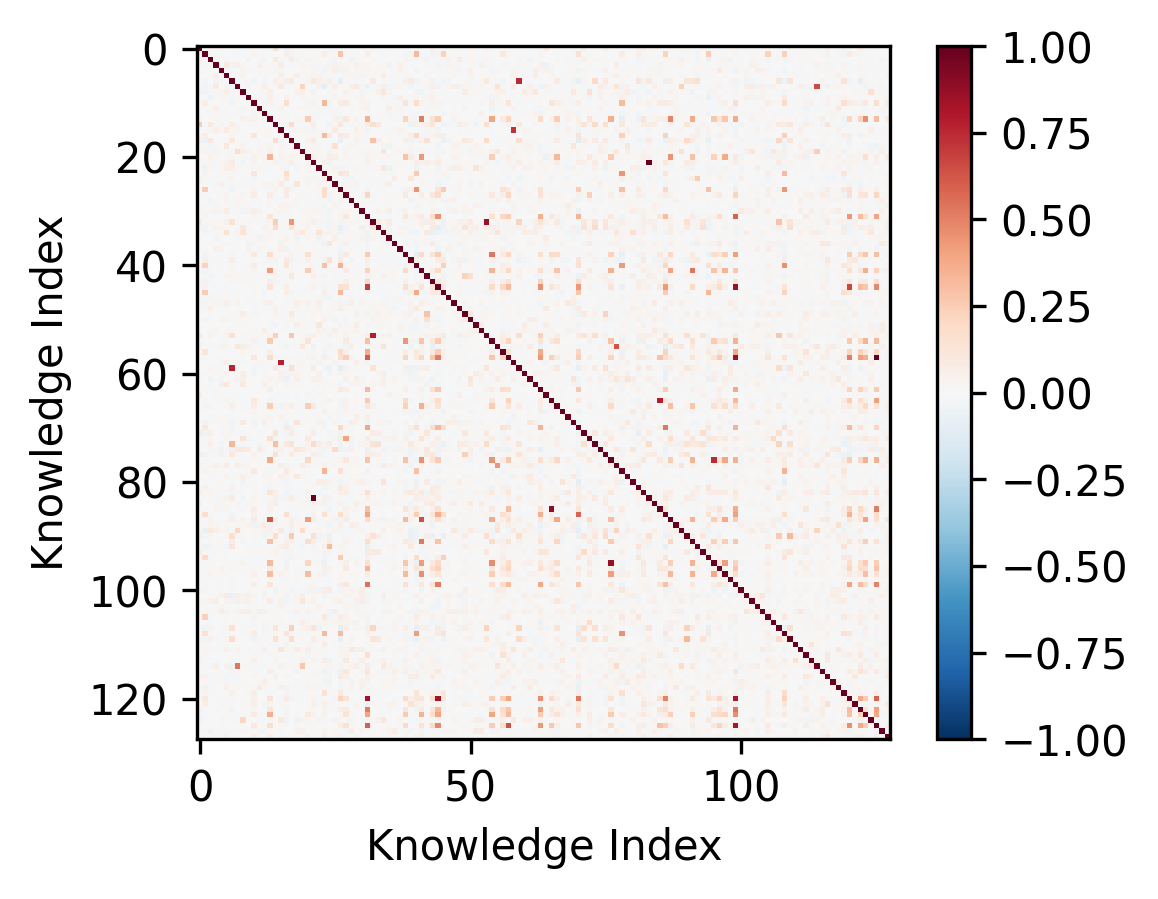}
    }
    \hfill
    \subfigure[Layer 4]{\includegraphics[width=0.22\textwidth]{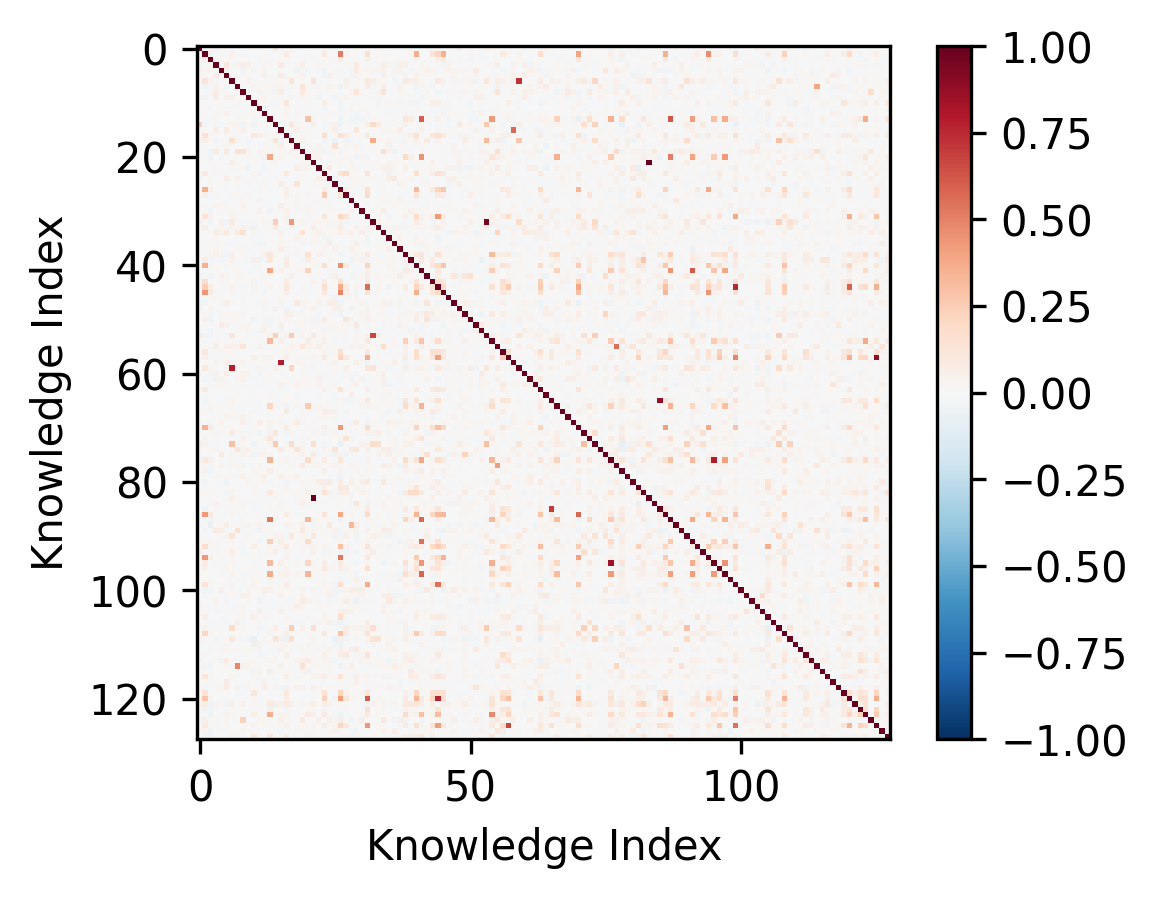}
    }
    \hfill
    \subfigure[Layer 5]{\includegraphics[width=0.22\textwidth]{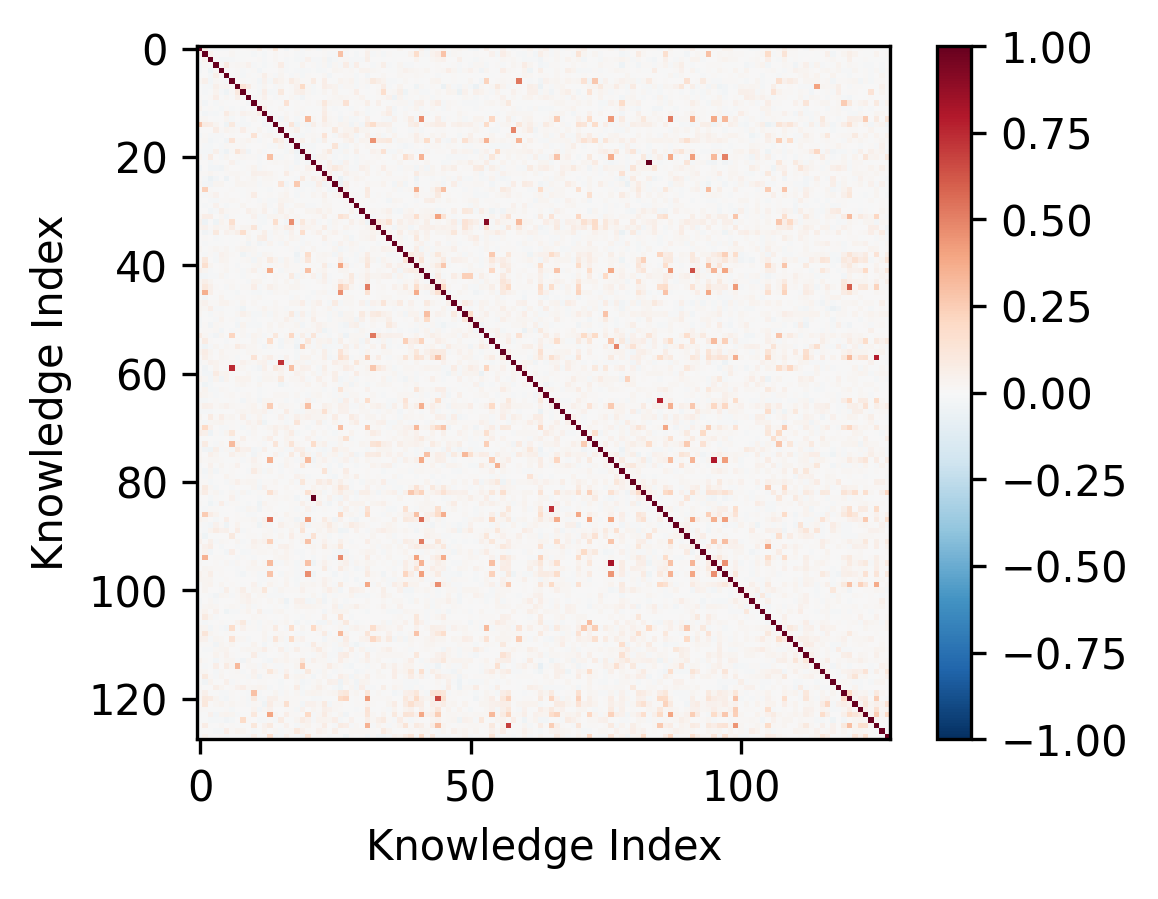}
    }
    \hfill
    \subfigure[Layer 6]{\includegraphics[width=0.22\textwidth]{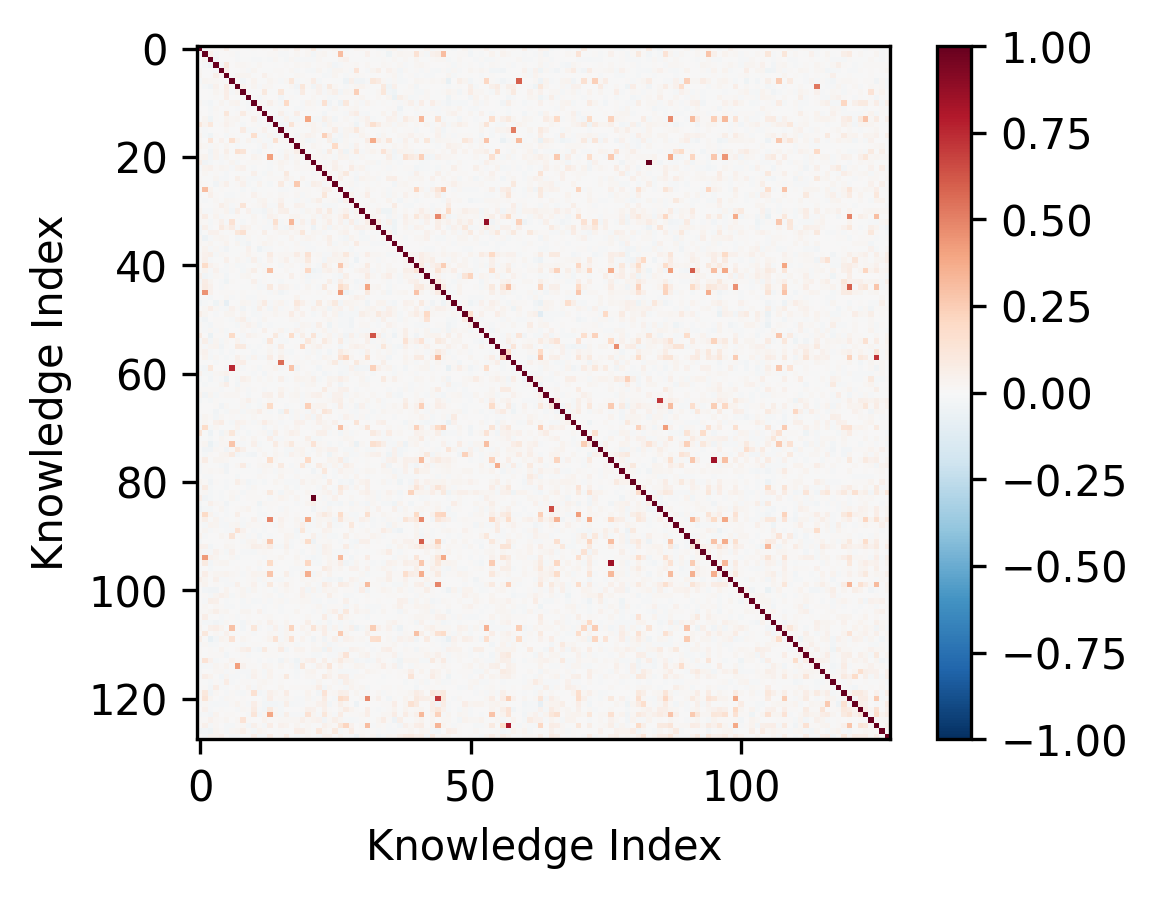}
    }
    \hfill
    \subfigure[Layer 7]{\includegraphics[width=0.22\textwidth]{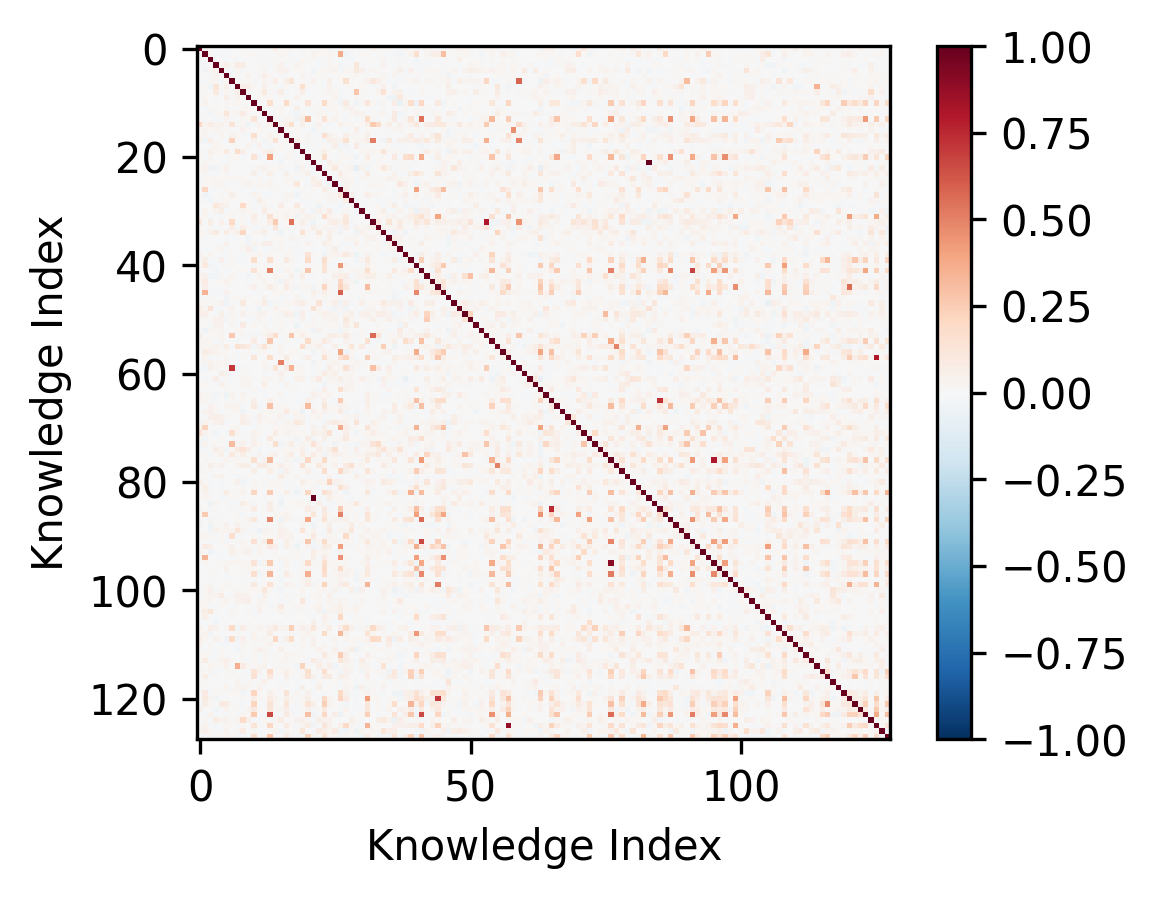}
    }
    \hfill
    \subfigure[Layer 8]{\includegraphics[width=0.22\textwidth]{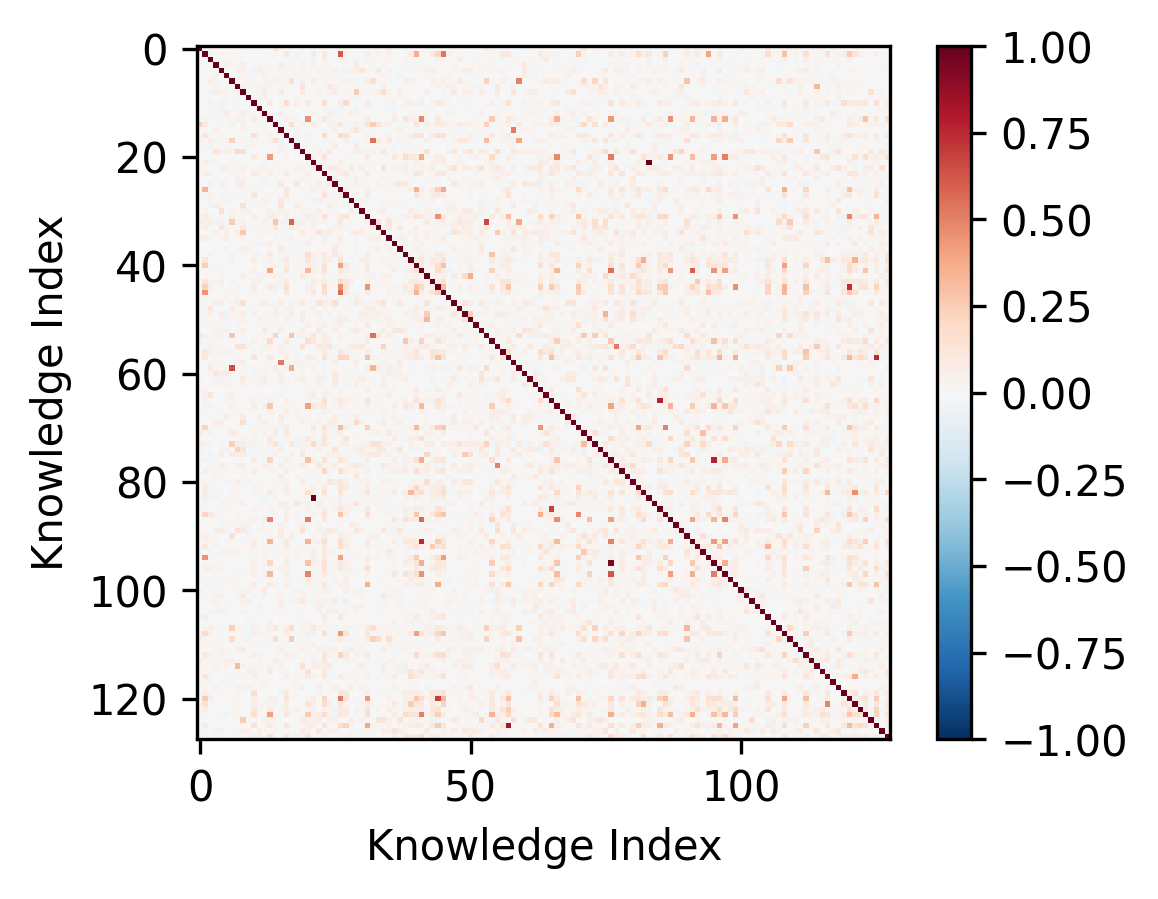}
    }
    \hfill
    \subfigure[Layer 9]{\includegraphics[width=0.22\textwidth]{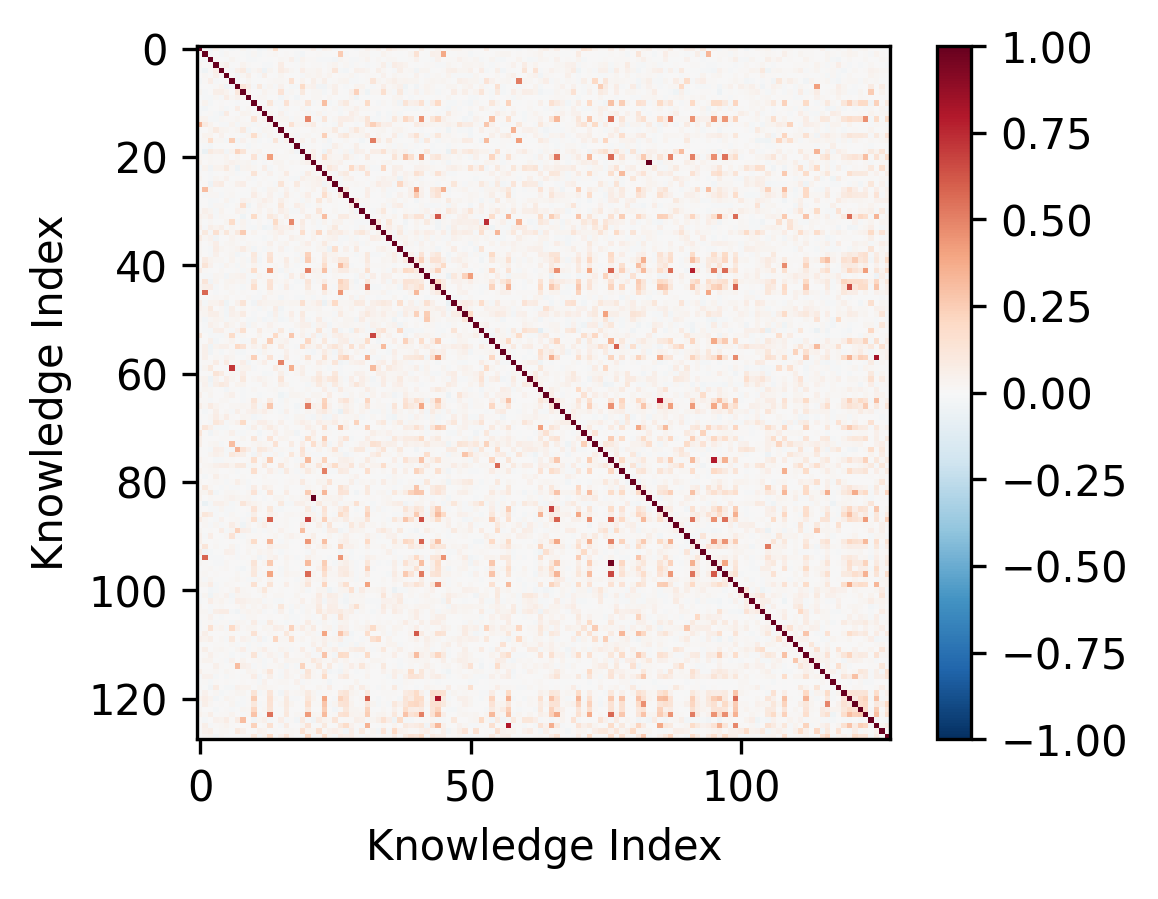}
    }
    \hfill
    \subfigure[Layer 10]{\includegraphics[width=0.22\textwidth]{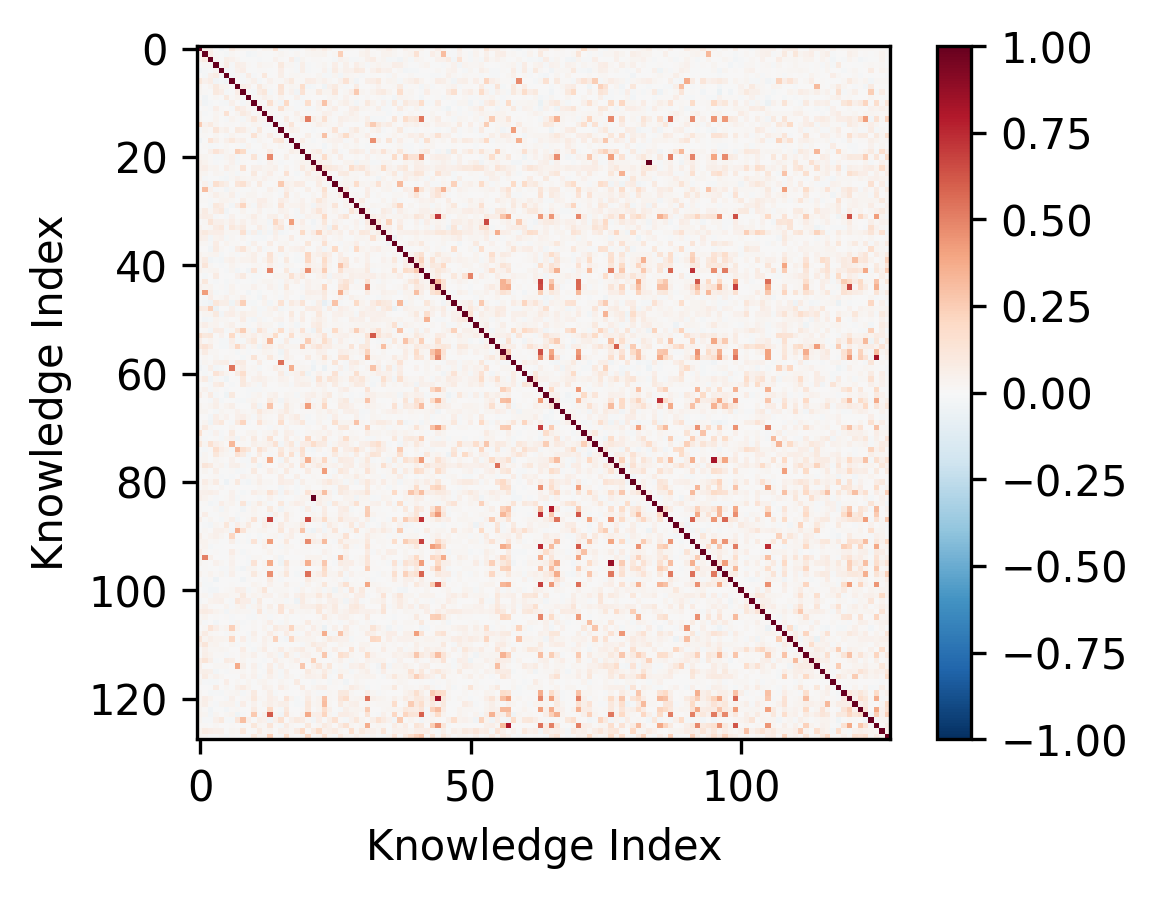}
    }
    \hfill
    \subfigure[Layer 11]{\includegraphics[width=0.22\textwidth]{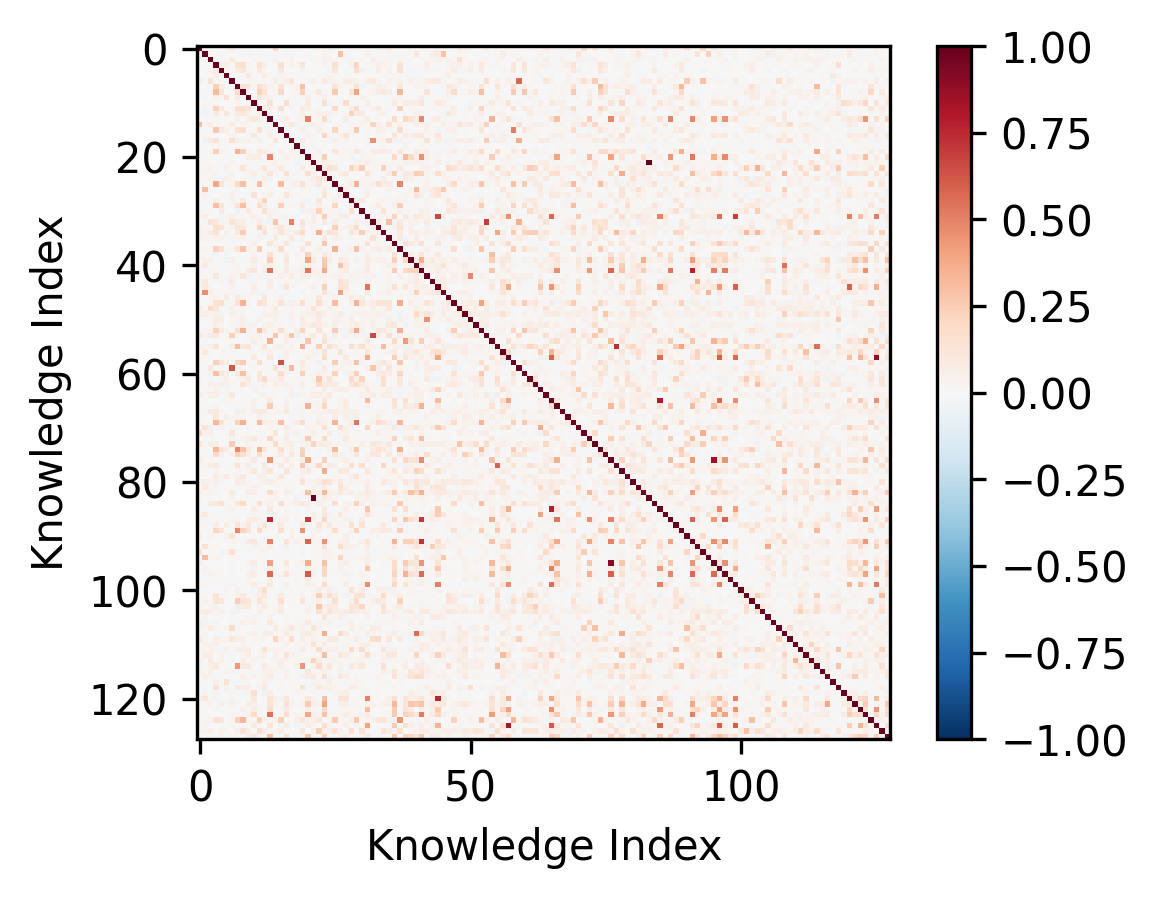}
    }
    \hfill
    \subfigure[Layer 12]{\includegraphics[width=0.22\textwidth]{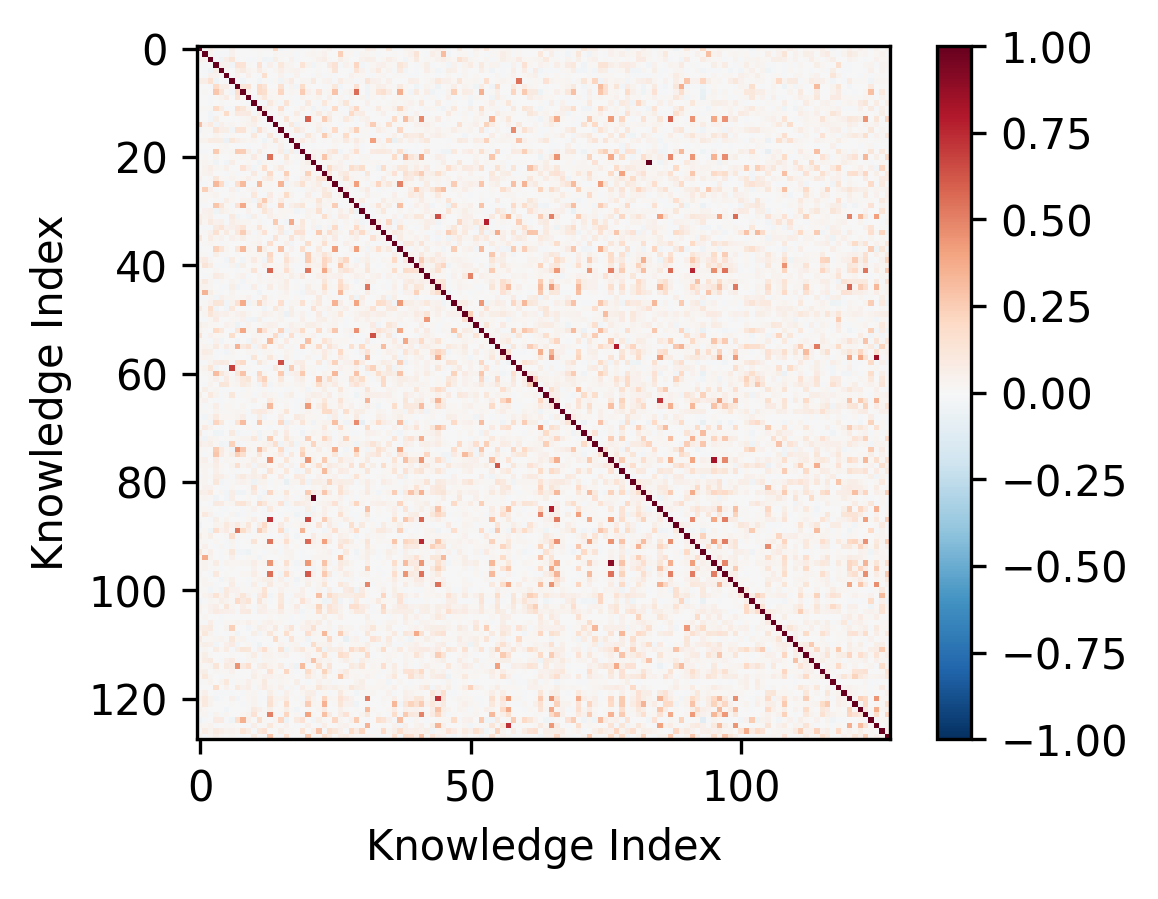}
    }
    \hfill
    \subfigure[Layer 13]{\includegraphics[width=0.22\textwidth]{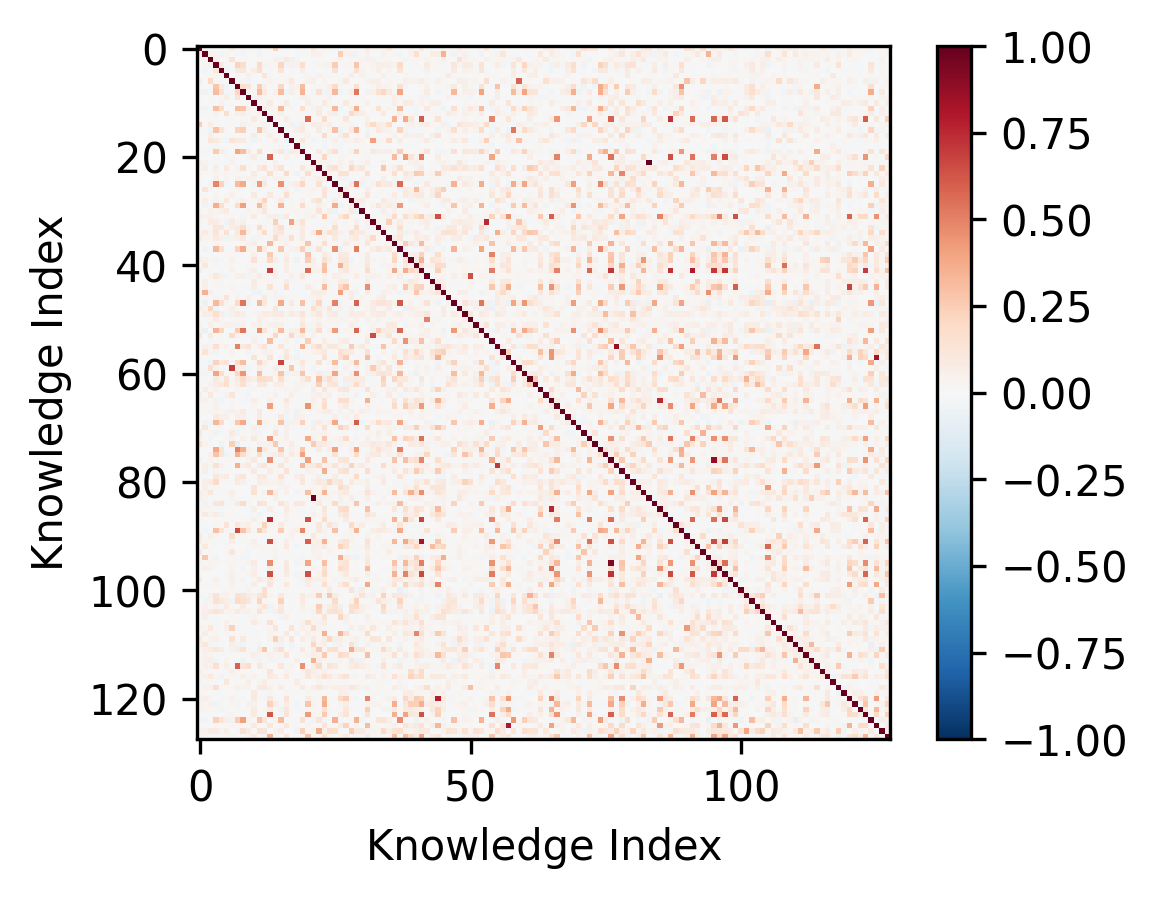}
    }
    \hfill
    \subfigure[Layer 14]{\includegraphics[width=0.22\textwidth]{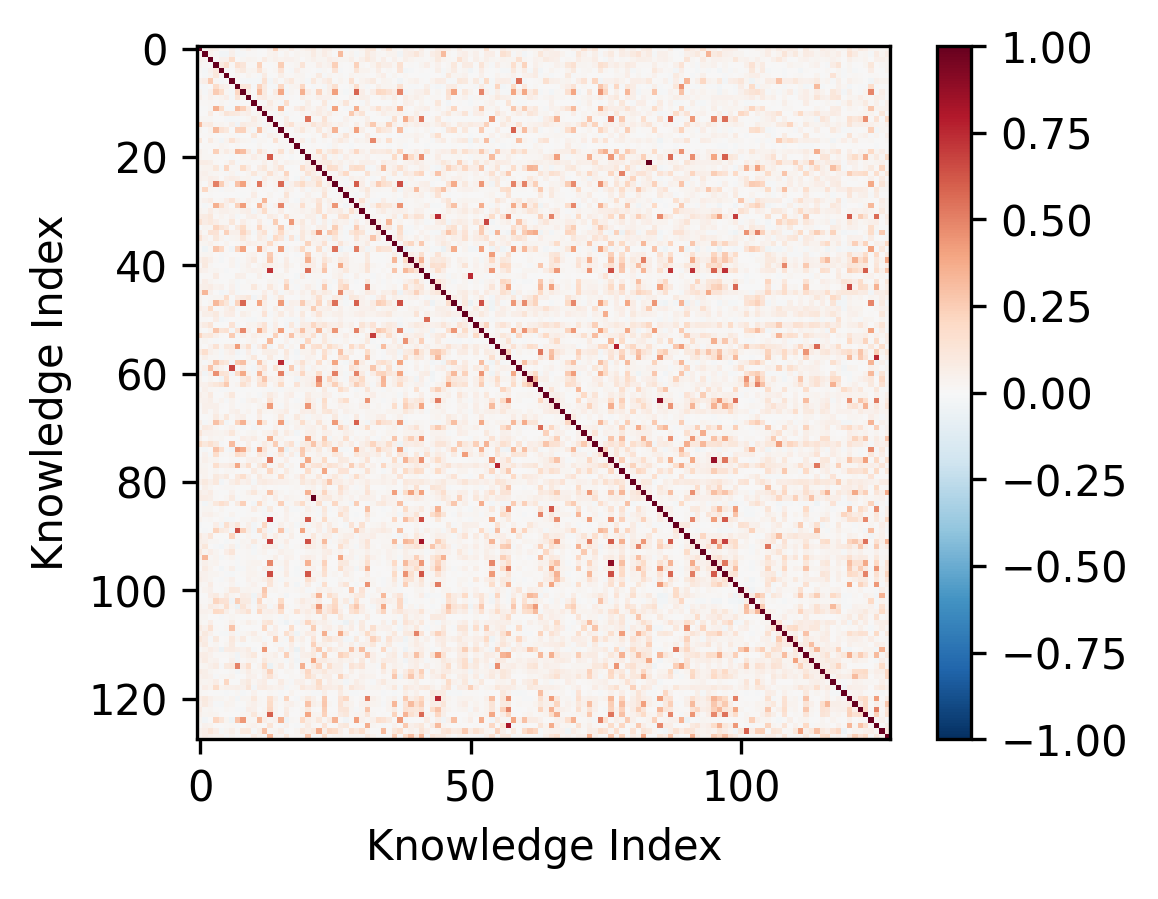}
    }
    \hfill
    \subfigure[Layer 15]{\includegraphics[width=0.22\textwidth]{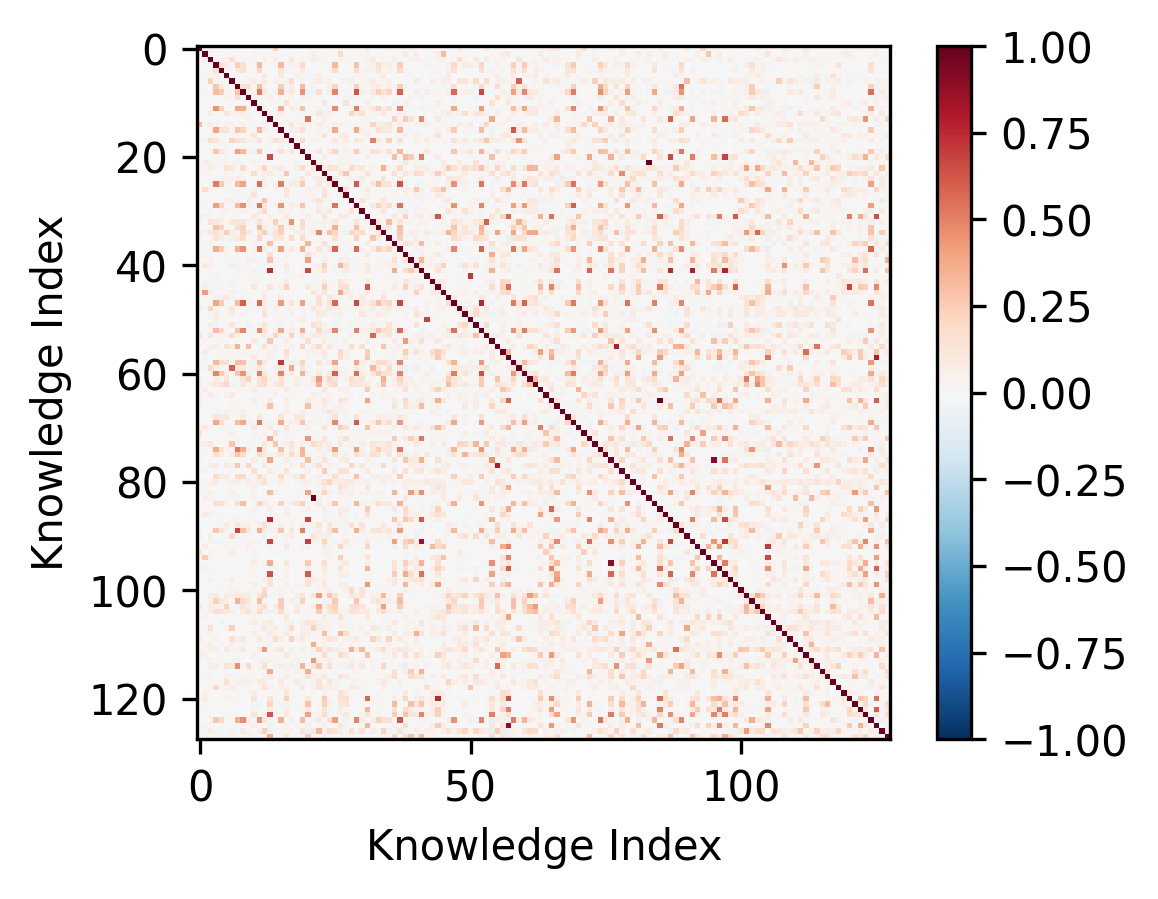}
    }
    \hfill
    \subfigure[Layer 16]{\includegraphics[width=0.22\textwidth]{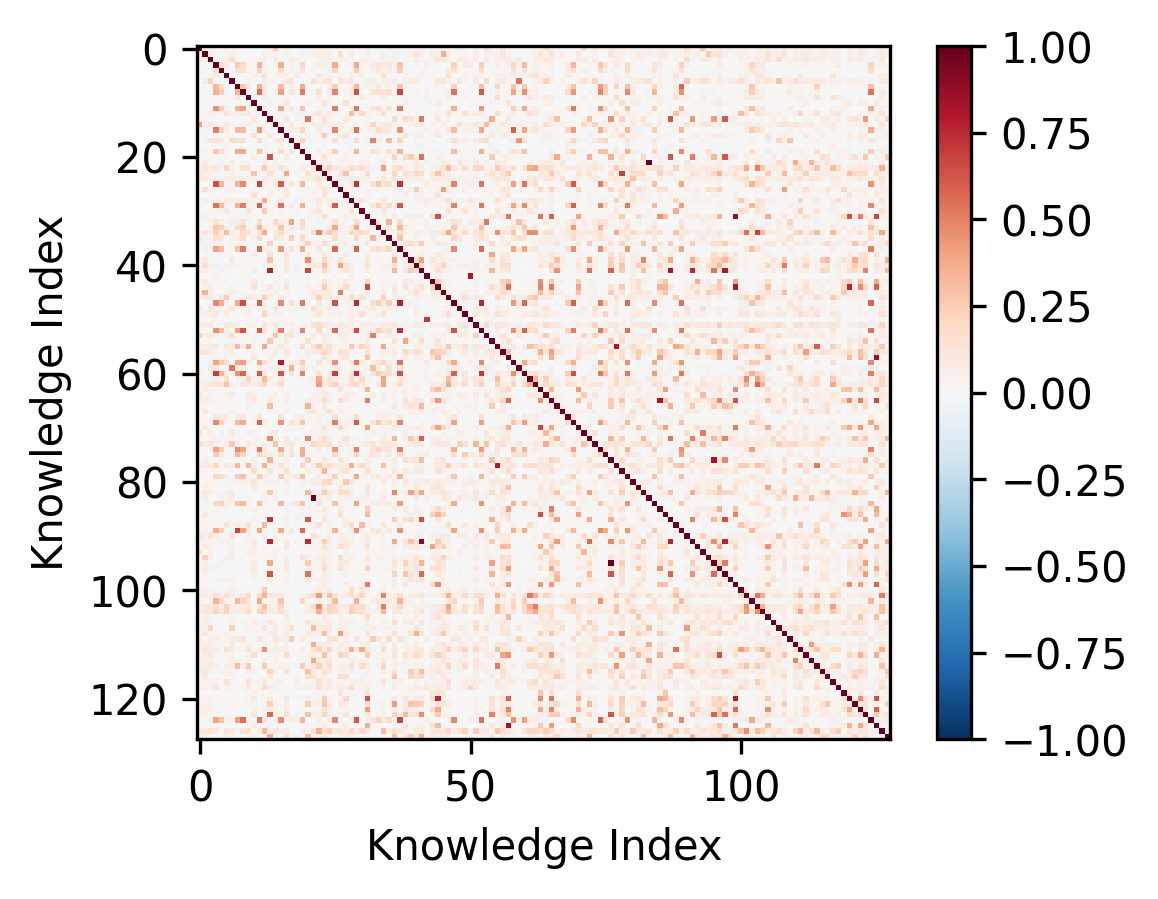}
    }
    \hfill
    \subfigure[Layer 17]{\includegraphics[width=0.22\textwidth]{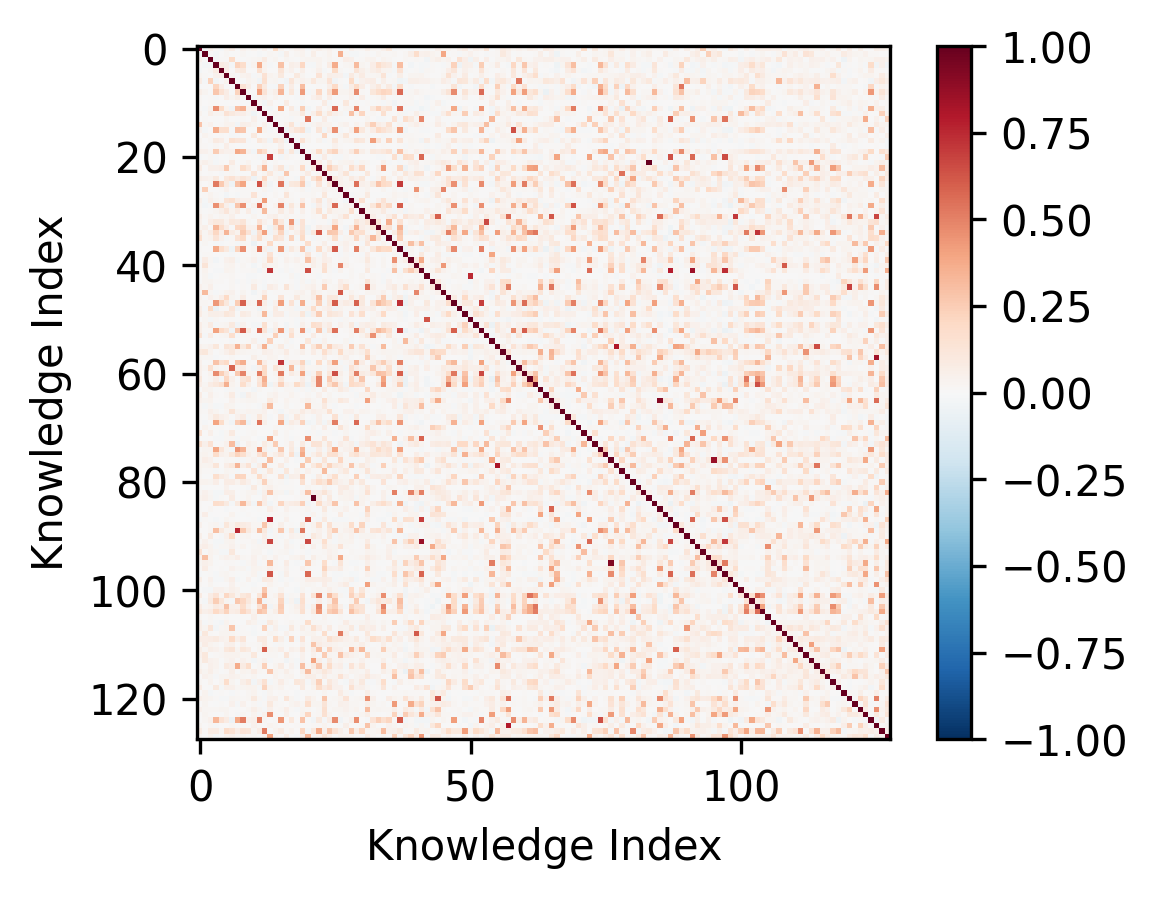}
    }
    \hfill
    \subfigure[Layer 18]{\includegraphics[width=0.22\textwidth]{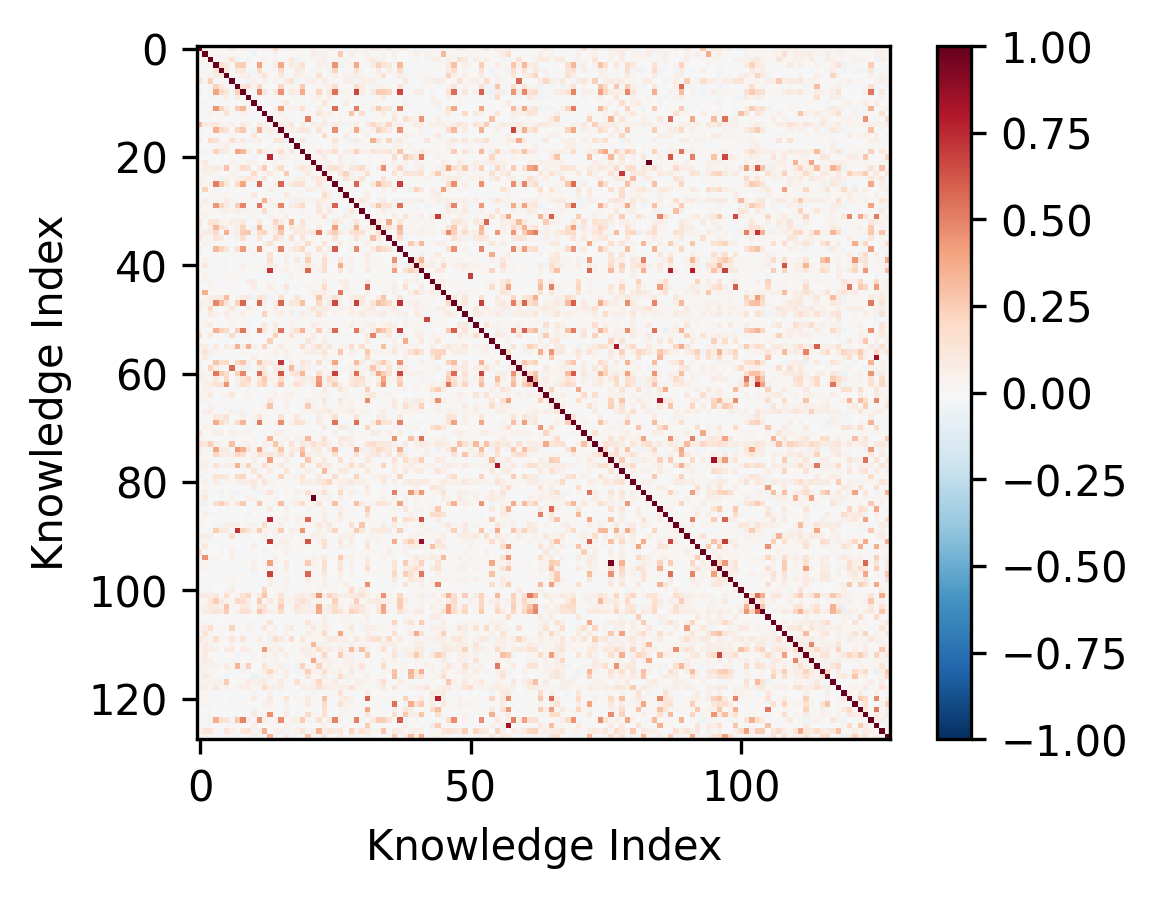}
    }
    \hfill
    \subfigure[Layer 19]{\includegraphics[width=0.22\textwidth]{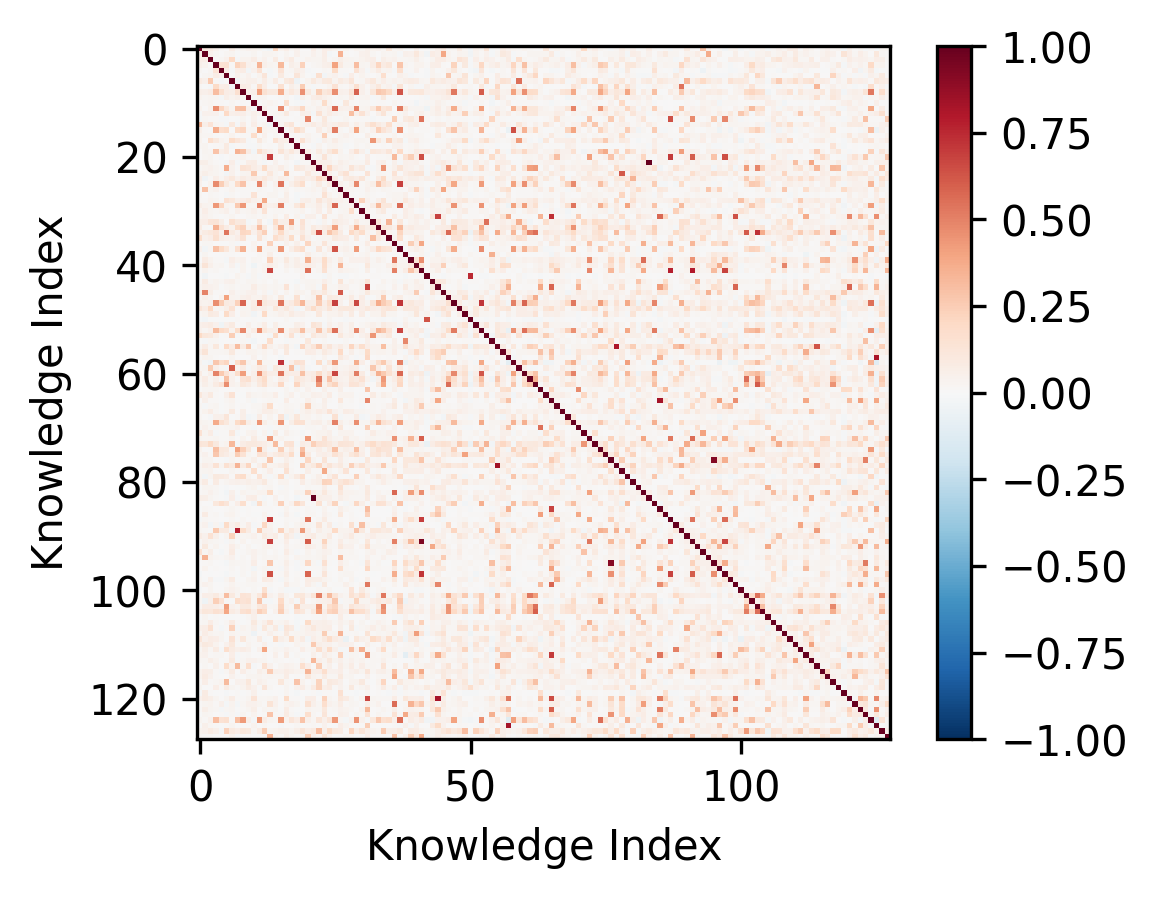}
    }
  \caption{In GPT2-Medium, the visualization of \(P\) matrices across layers (0-19).}
  \label{fig:superposition_visualization_gpt2-medium-part1}
\end{figure*}

\begin{figure*}
    \centering
    \subfigure[Layer 20]{\includegraphics[width=0.22\textwidth]{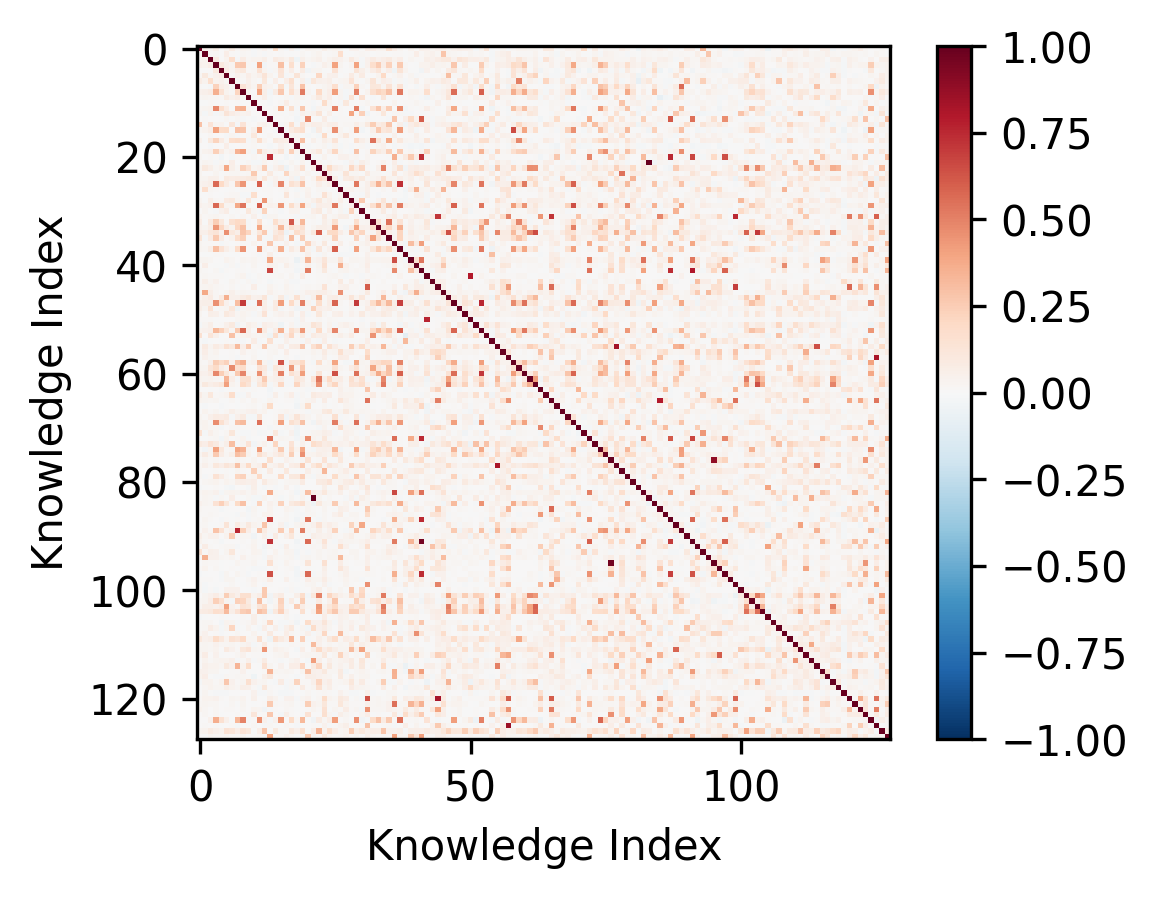}
    }
    \hfill
    \subfigure[Layer 21]{\includegraphics[width=0.22\textwidth]{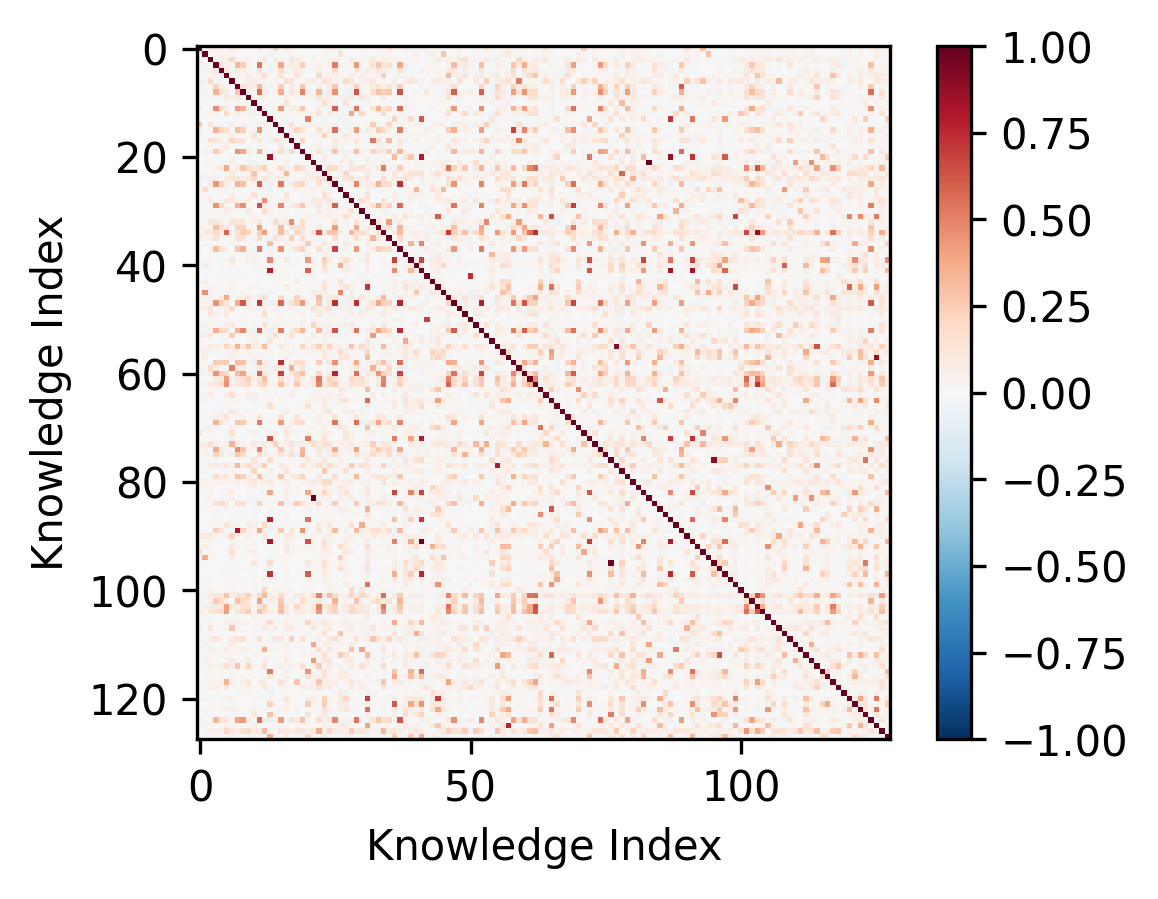}
    }
    \hfill
    \subfigure[Layer 22]{\includegraphics[width=0.22\textwidth]{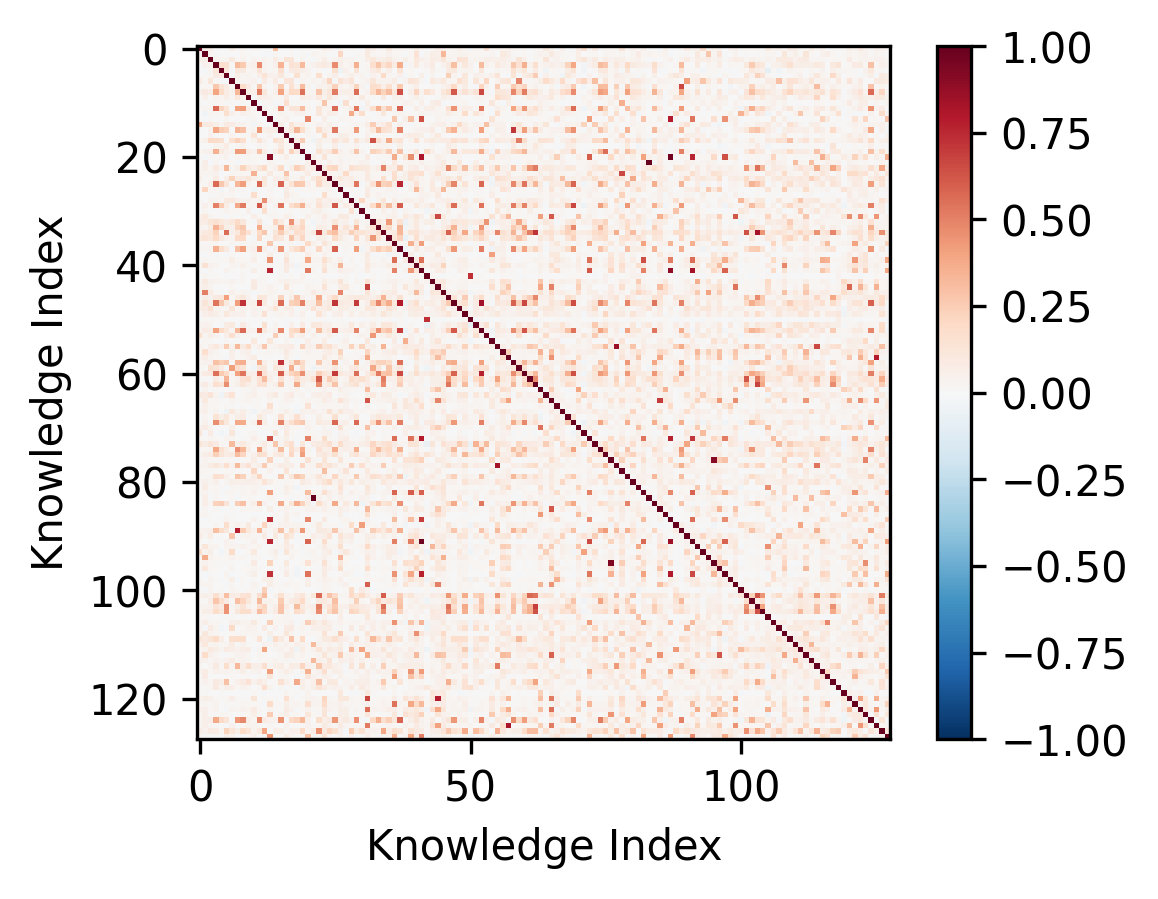}
    }
    \hfill
    \subfigure[Layer 23]{\includegraphics[width=0.22\textwidth]{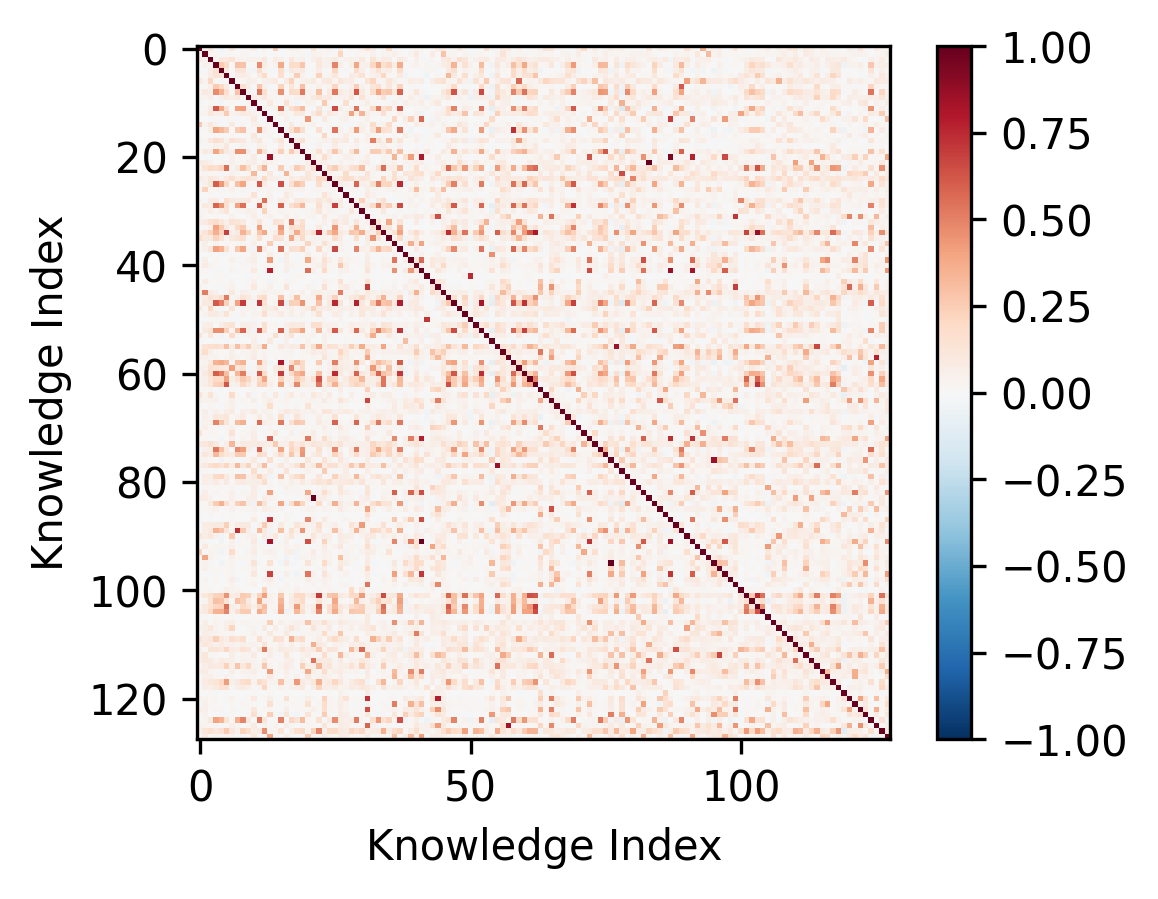}
    }
  \caption{In GPT2-Medium, the visualization of \(P\) matrices across layers (20-23).}
  \label{fig:superposition_visualization_gpt2-medium-part2}
\end{figure*}

\begin{figure*}
    \centering
    \subfigure[Layer 0]{\includegraphics[width=0.22\textwidth]{fig/gpt2-large/p_matrix/known/heatmap/superposition_for_layer_0.png}
    }
    \hfill
    \subfigure[Layer 1]{\includegraphics[width=0.22\textwidth]{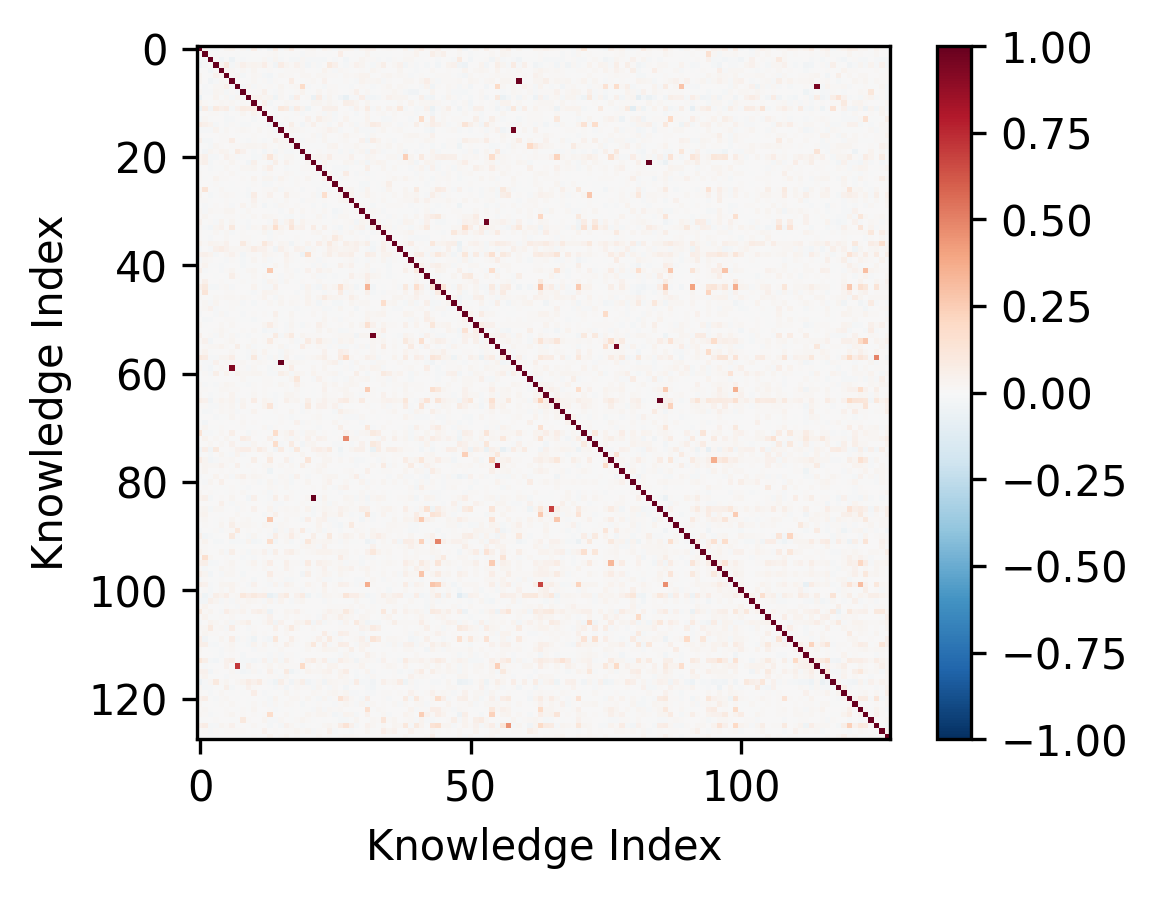}
    }
    \hfill
    \subfigure[Layer 2]{\includegraphics[width=0.22\textwidth]{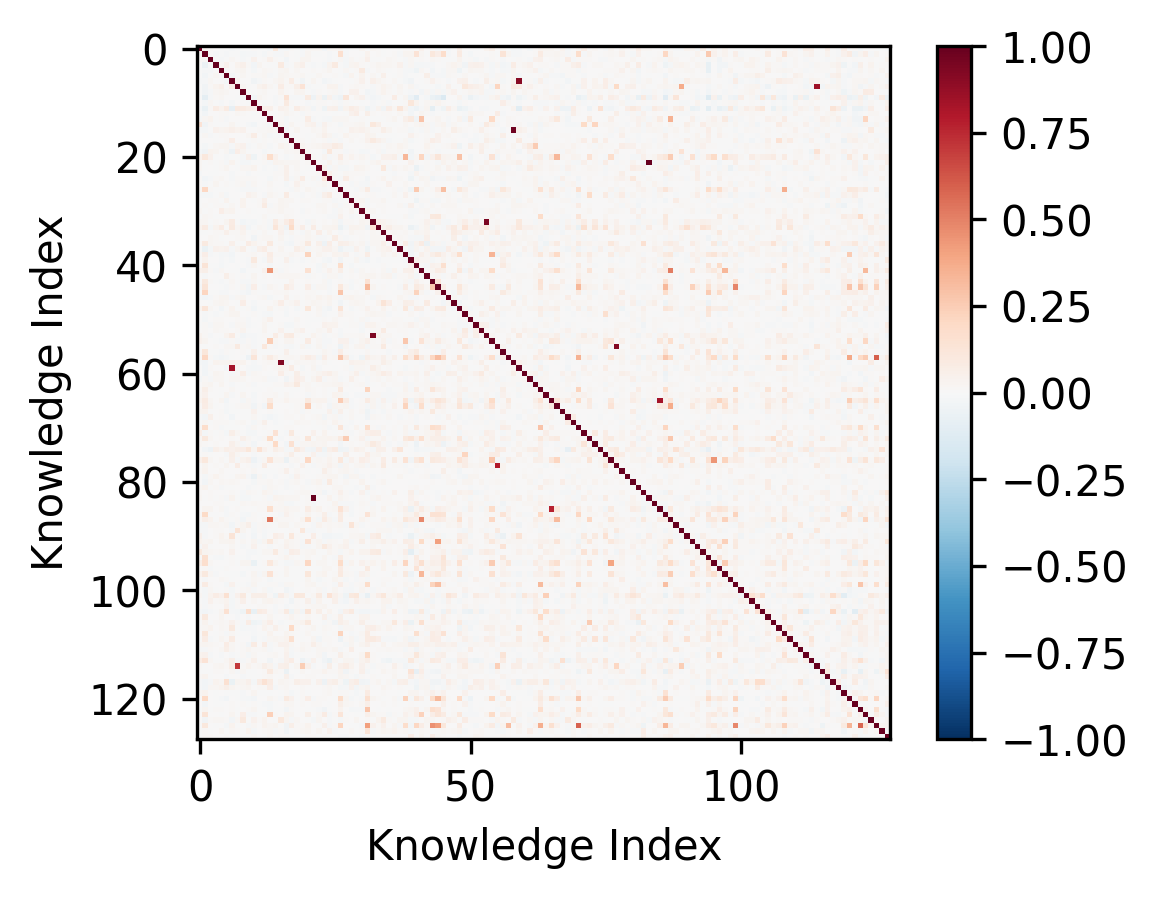}
    }
    \hfill
    \subfigure[Layer 3]{\includegraphics[width=0.22\textwidth]{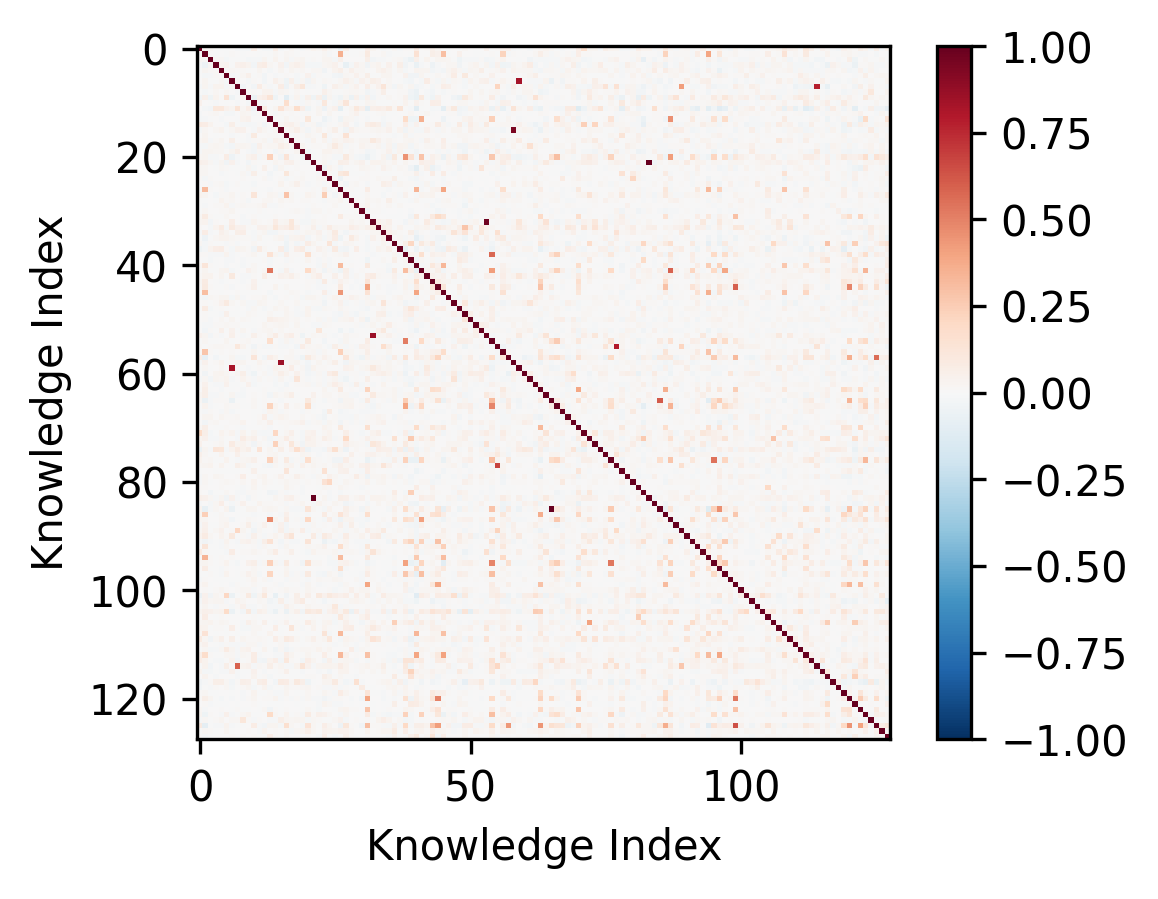}
    }
    \hfill
    \subfigure[Layer 4]{\includegraphics[width=0.22\textwidth]{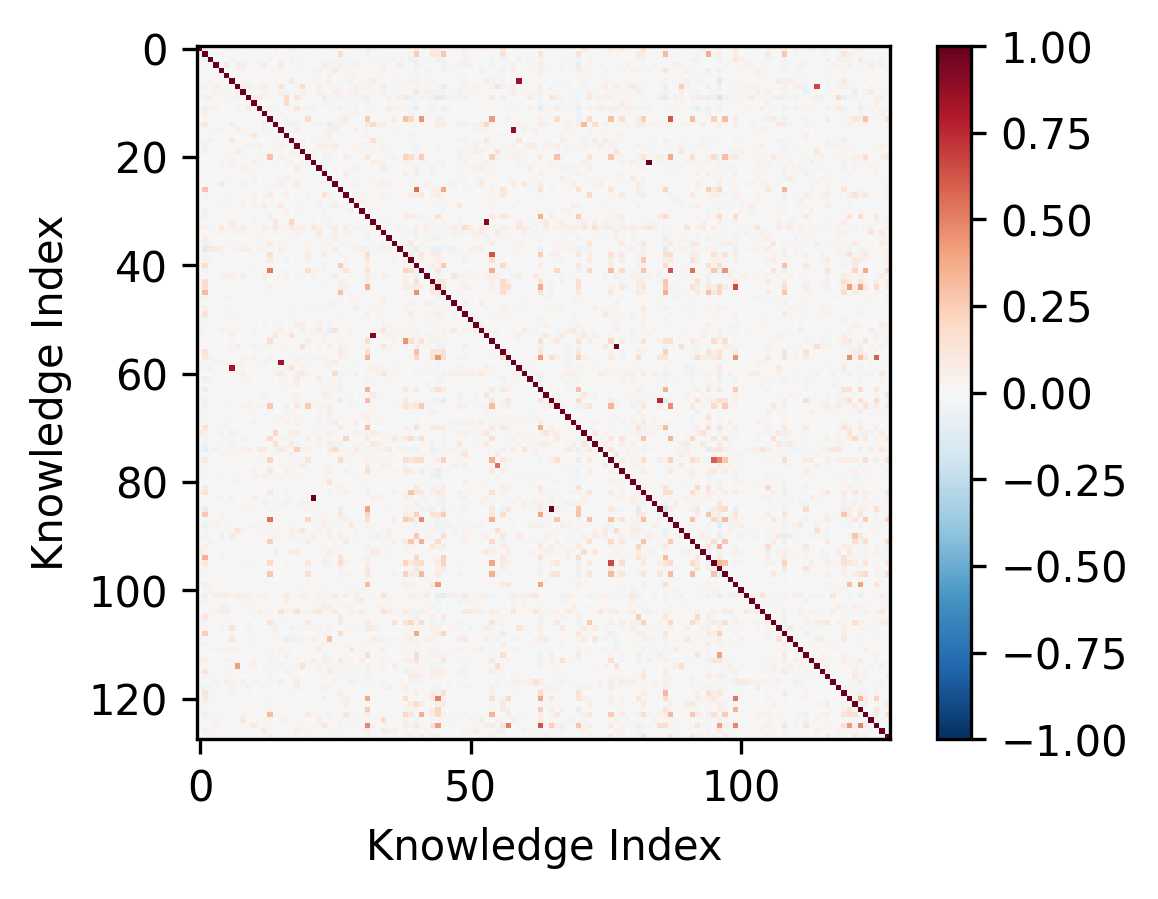}
    }
    \hfill
    \subfigure[Layer 5]{\includegraphics[width=0.22\textwidth]{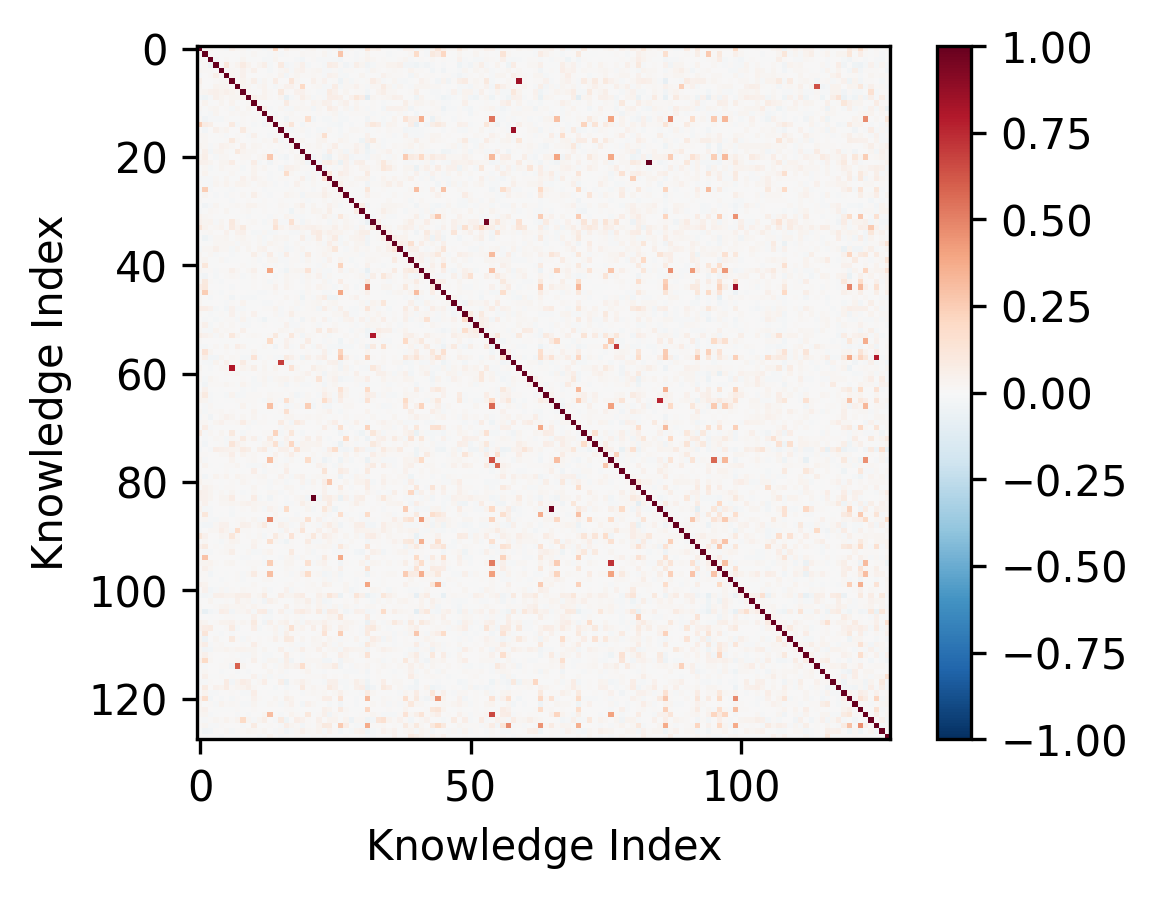}
    }
    \hfill
    \subfigure[Layer 6]{\includegraphics[width=0.22\textwidth]{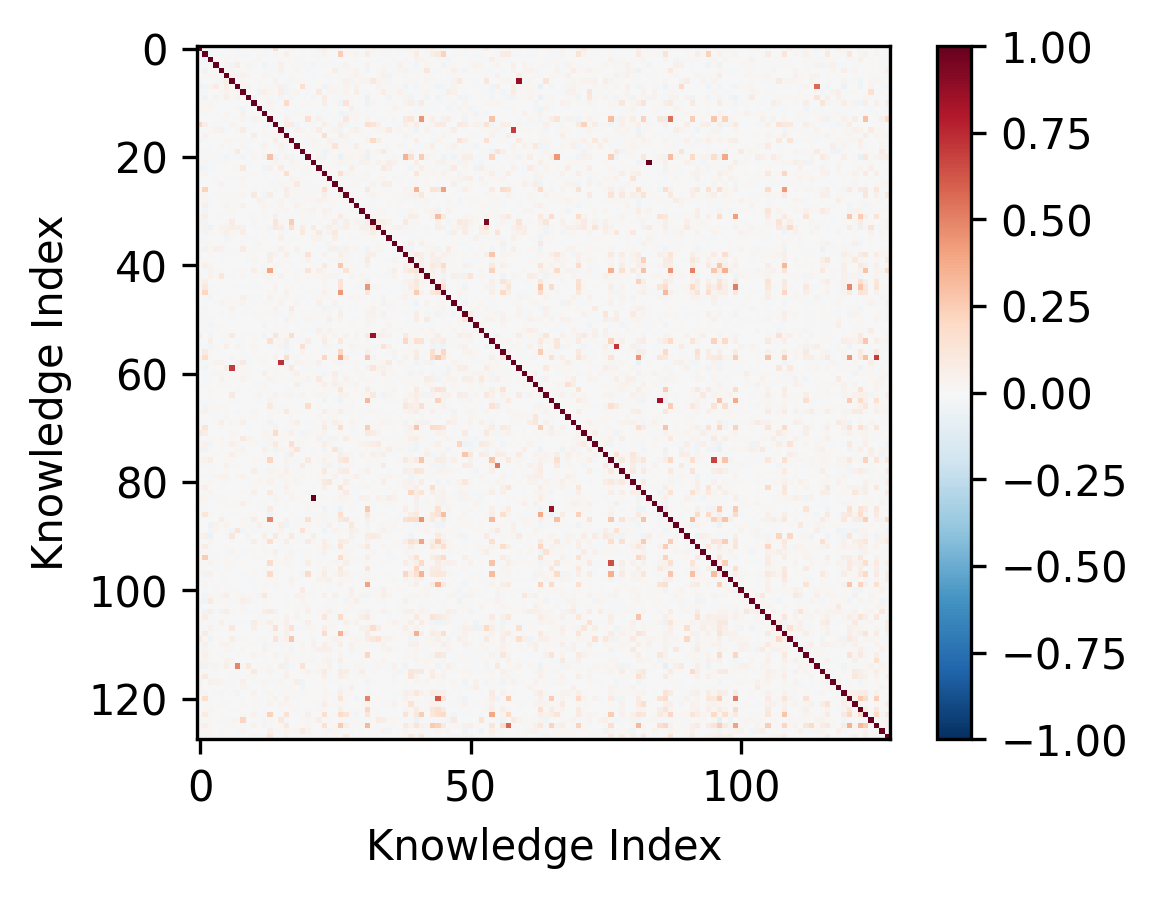}
    }
    \hfill
    \subfigure[Layer 7]{\includegraphics[width=0.22\textwidth]{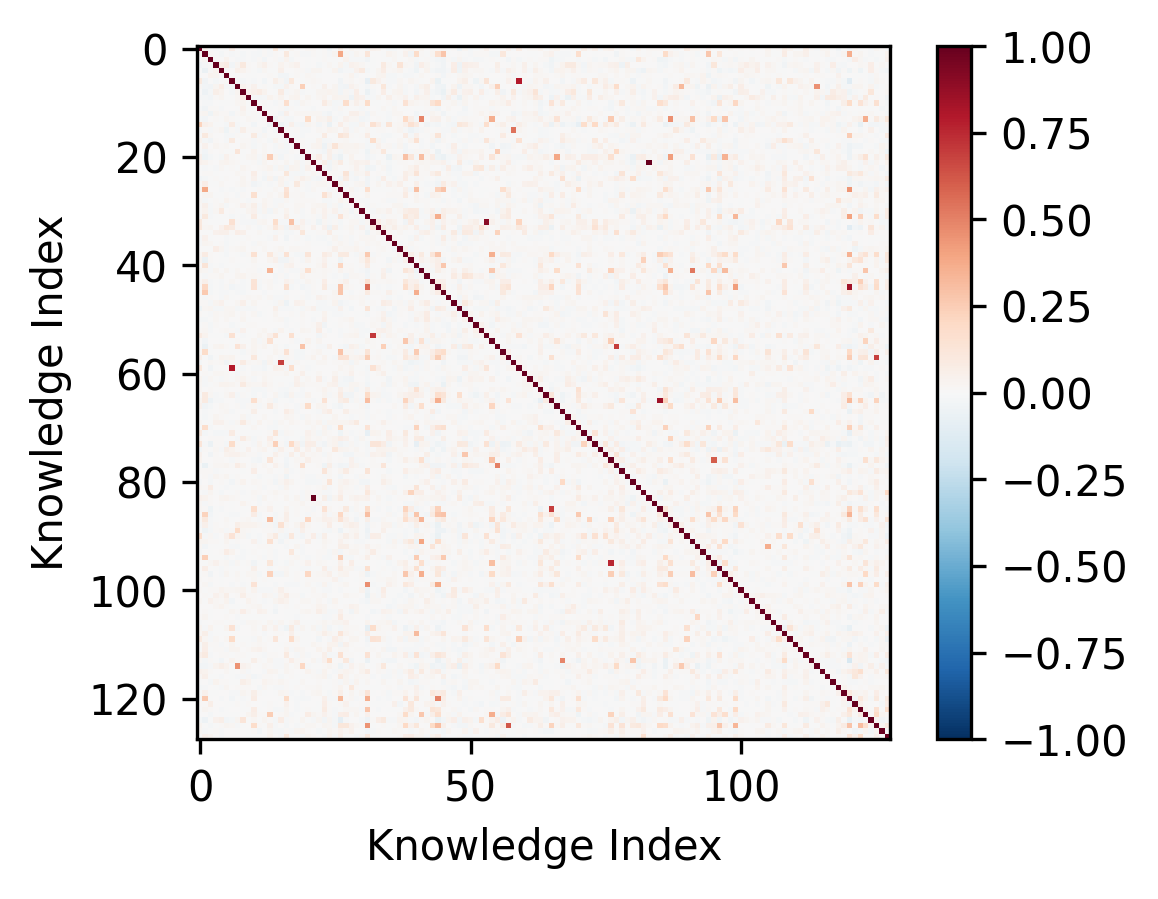}
    }
    \hfill
    \subfigure[Layer 8]{\includegraphics[width=0.22\textwidth]{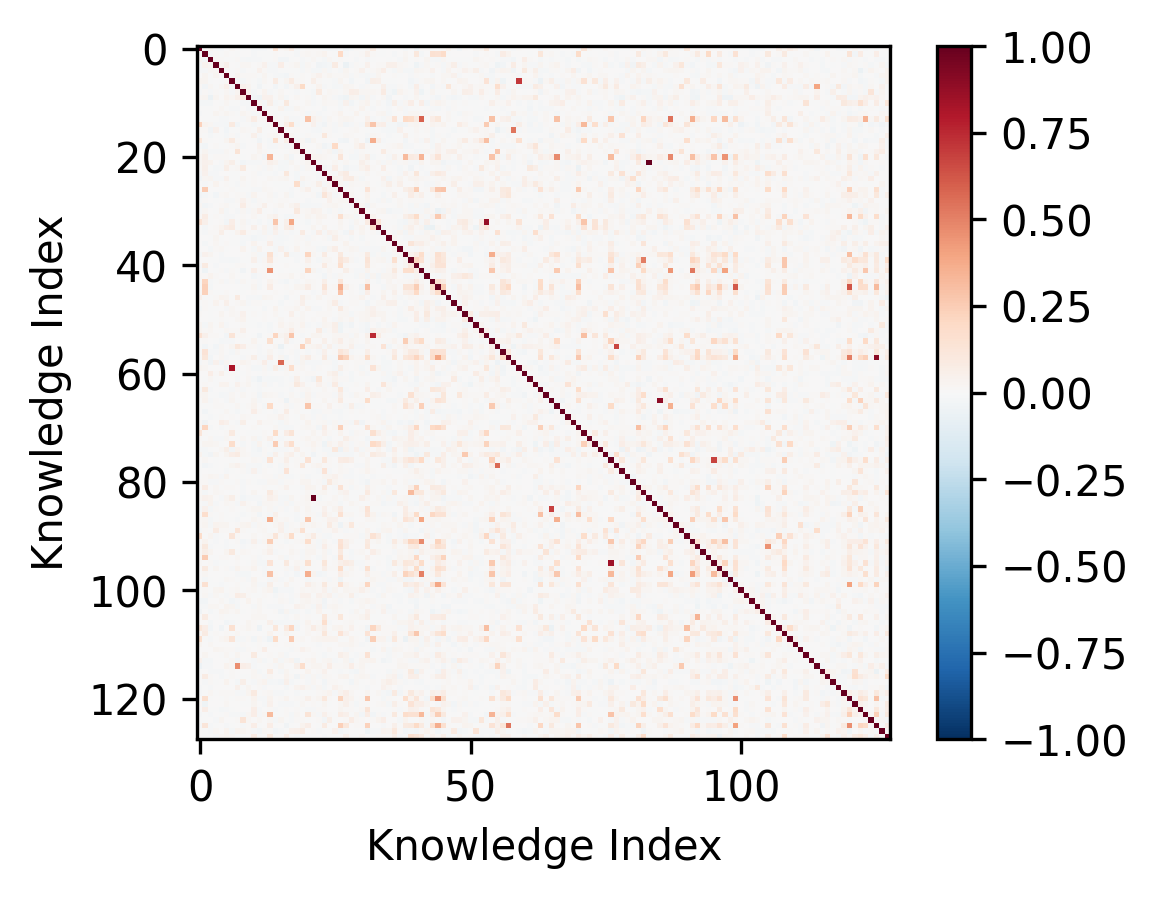}
    }
    \hfill
    \subfigure[Layer 9]{\includegraphics[width=0.22\textwidth]{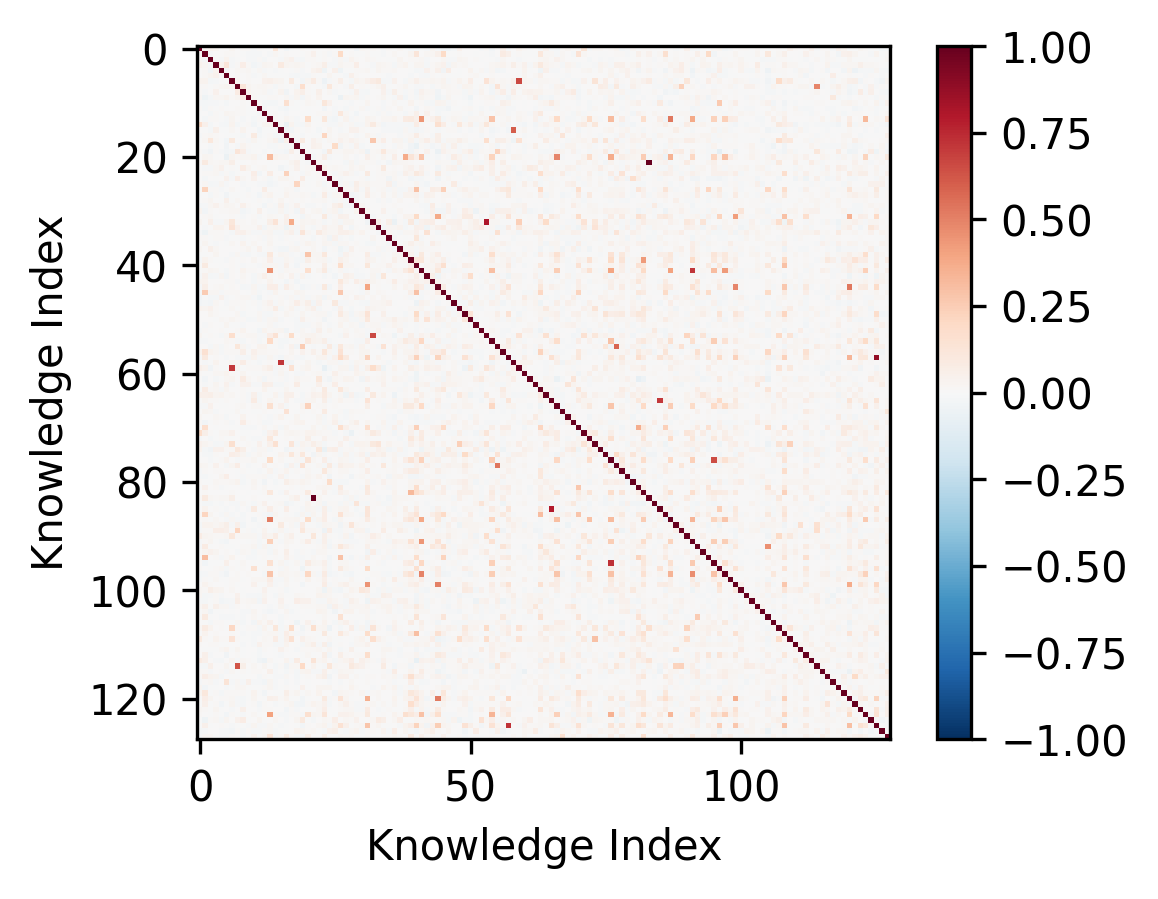}
    }
    \hfill
    \subfigure[Layer 10]{\includegraphics[width=0.22\textwidth]{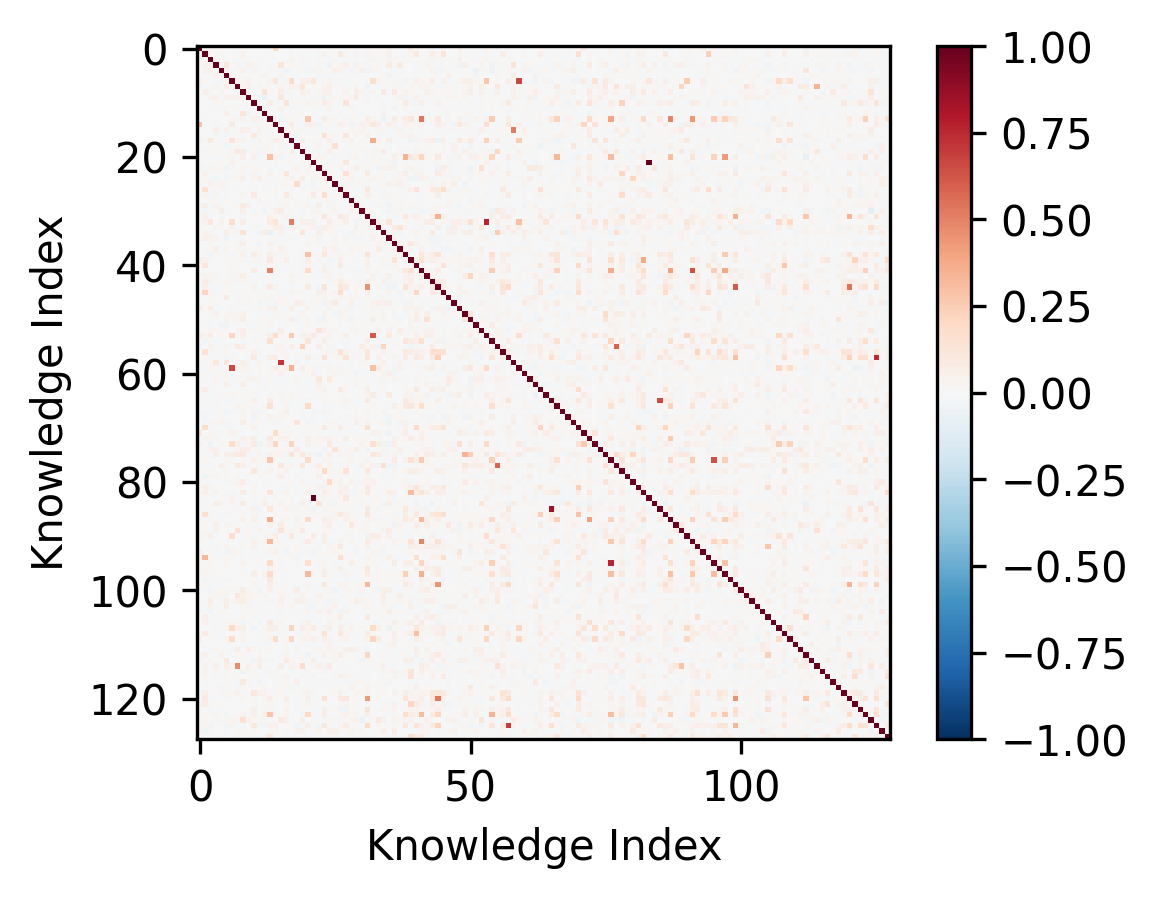}
    }
    \hfill
    \subfigure[Layer 11]{\includegraphics[width=0.22\textwidth]{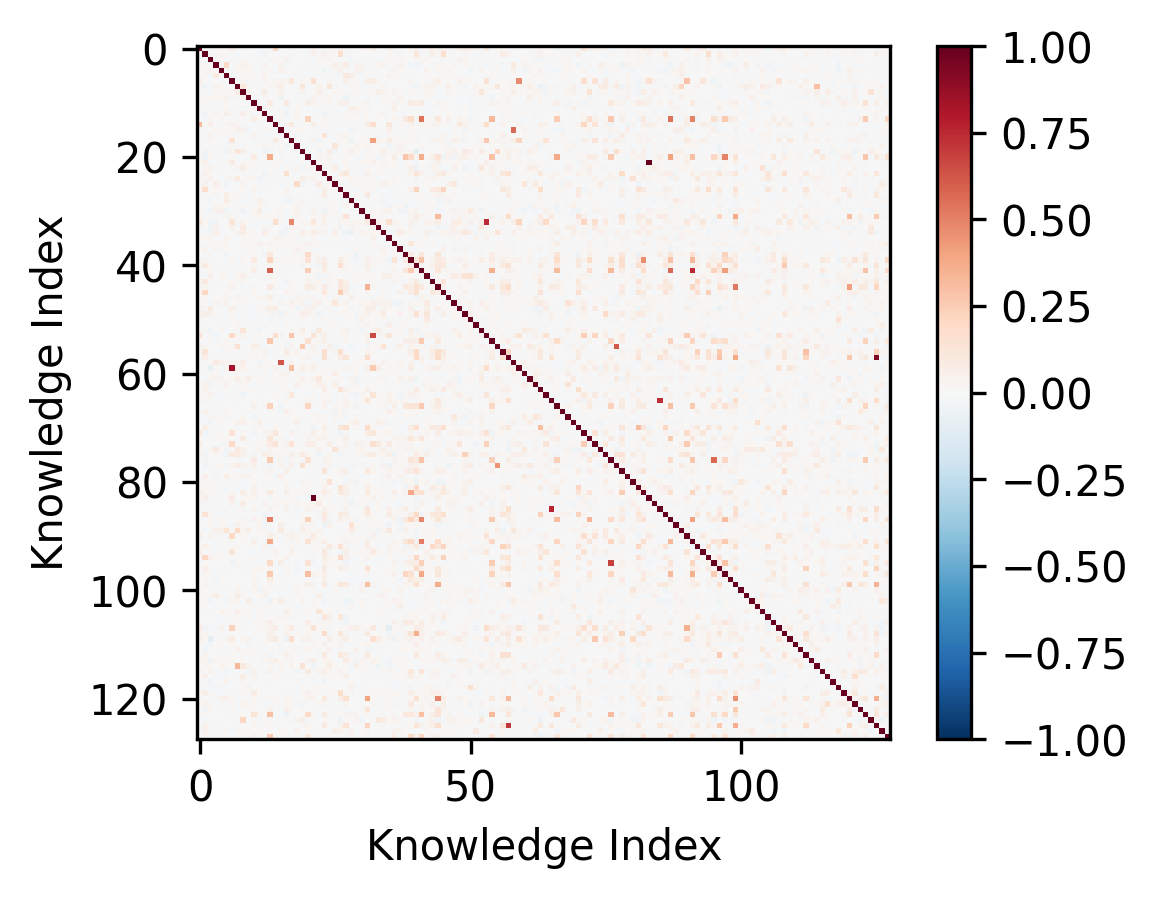}
    }
    \hfill
    \subfigure[Layer 12]{\includegraphics[width=0.22\textwidth]{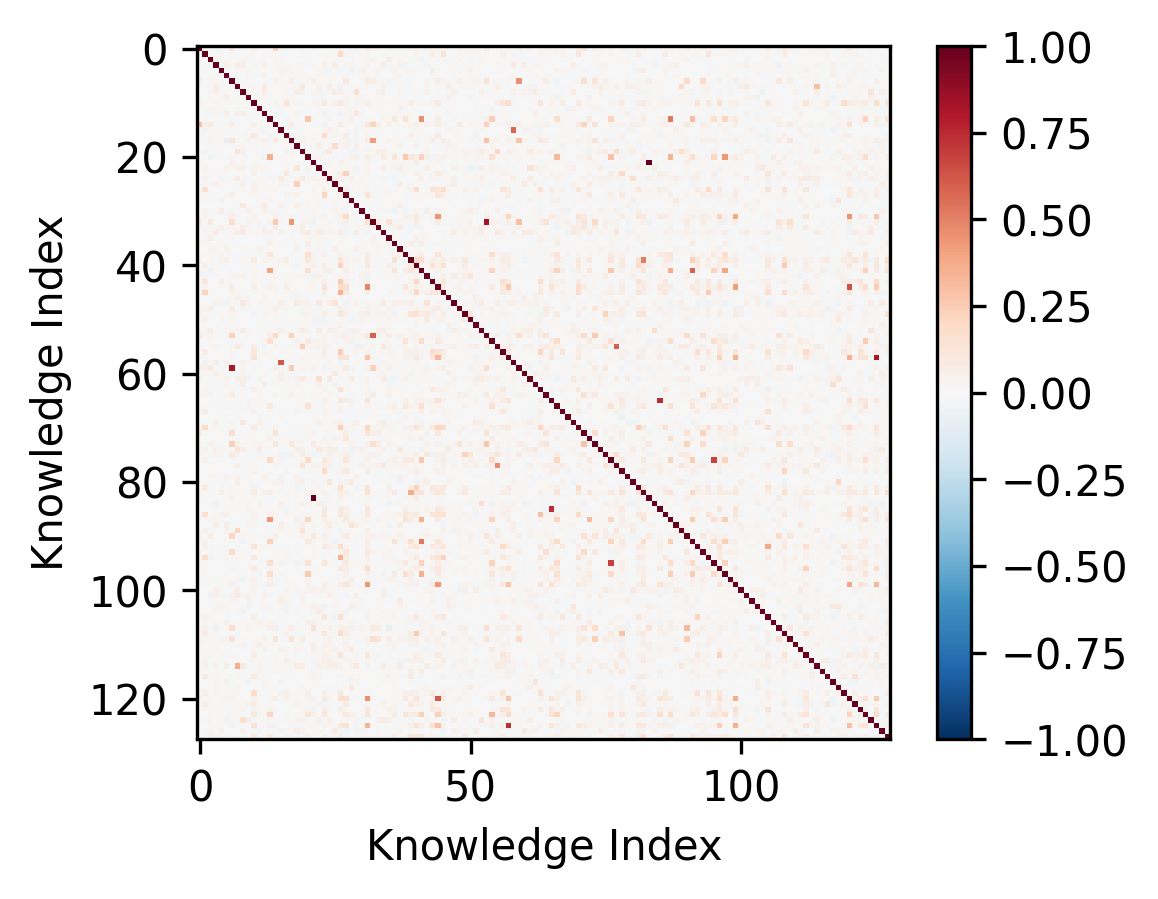}
    }
    \hfill
    \subfigure[Layer 13]{\includegraphics[width=0.22\textwidth]{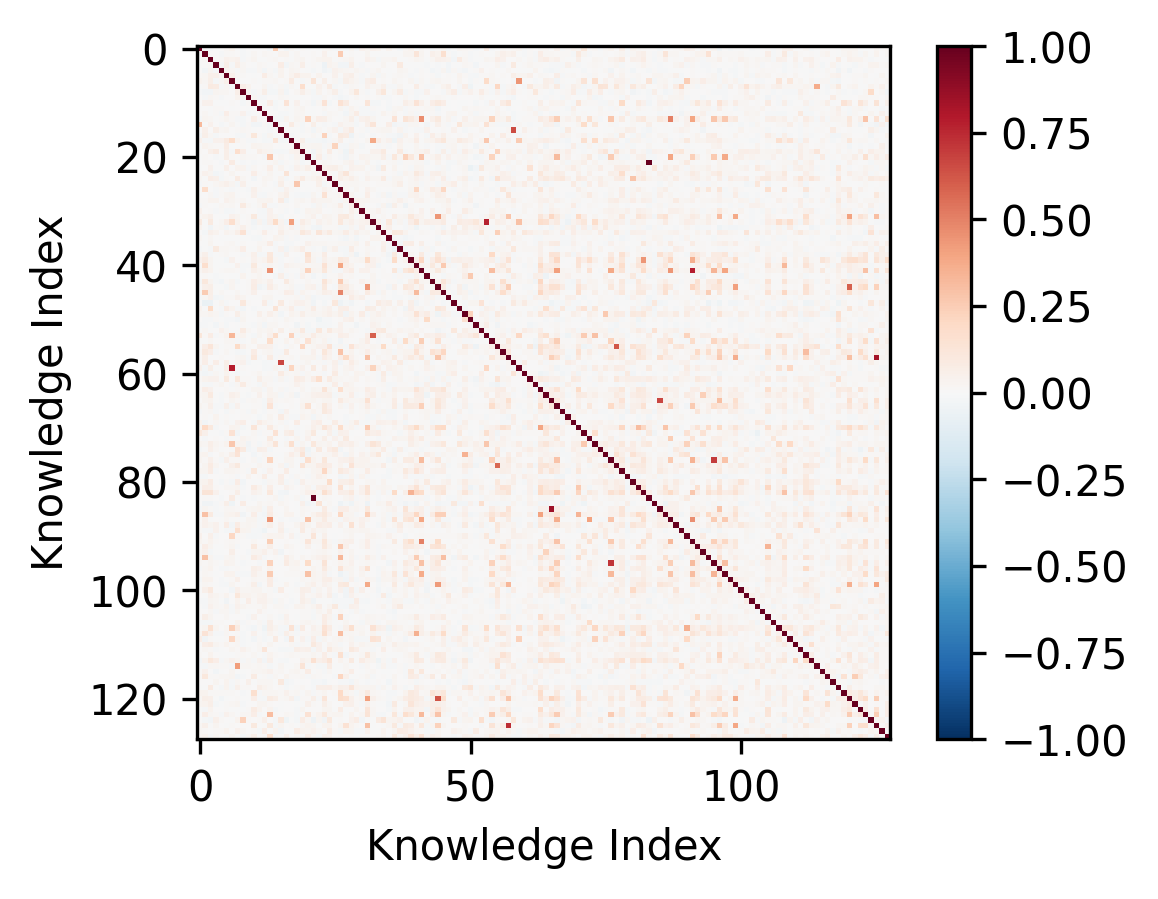}
    }
    \hfill
    \subfigure[Layer 14]{\includegraphics[width=0.22\textwidth]{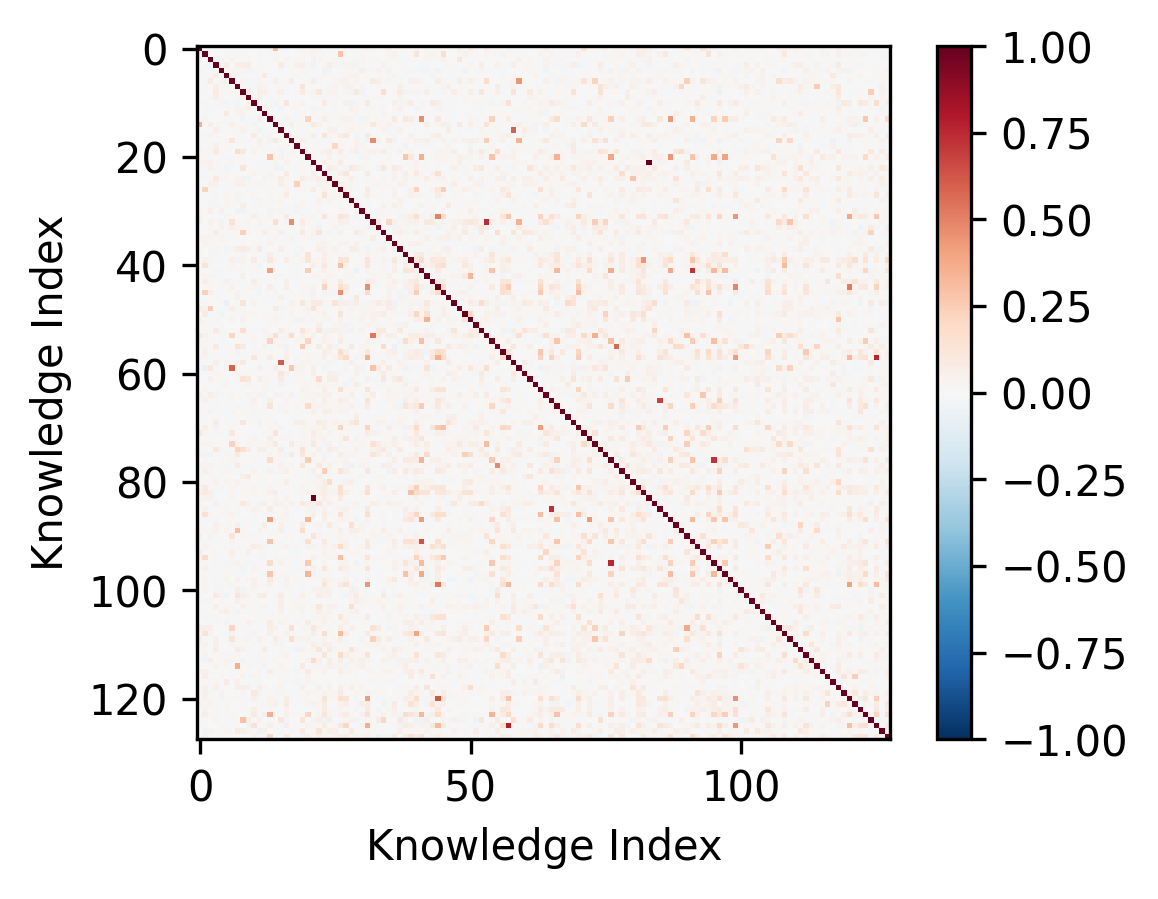}
    }
    \hfill
    \subfigure[Layer 15]{\includegraphics[width=0.22\textwidth]{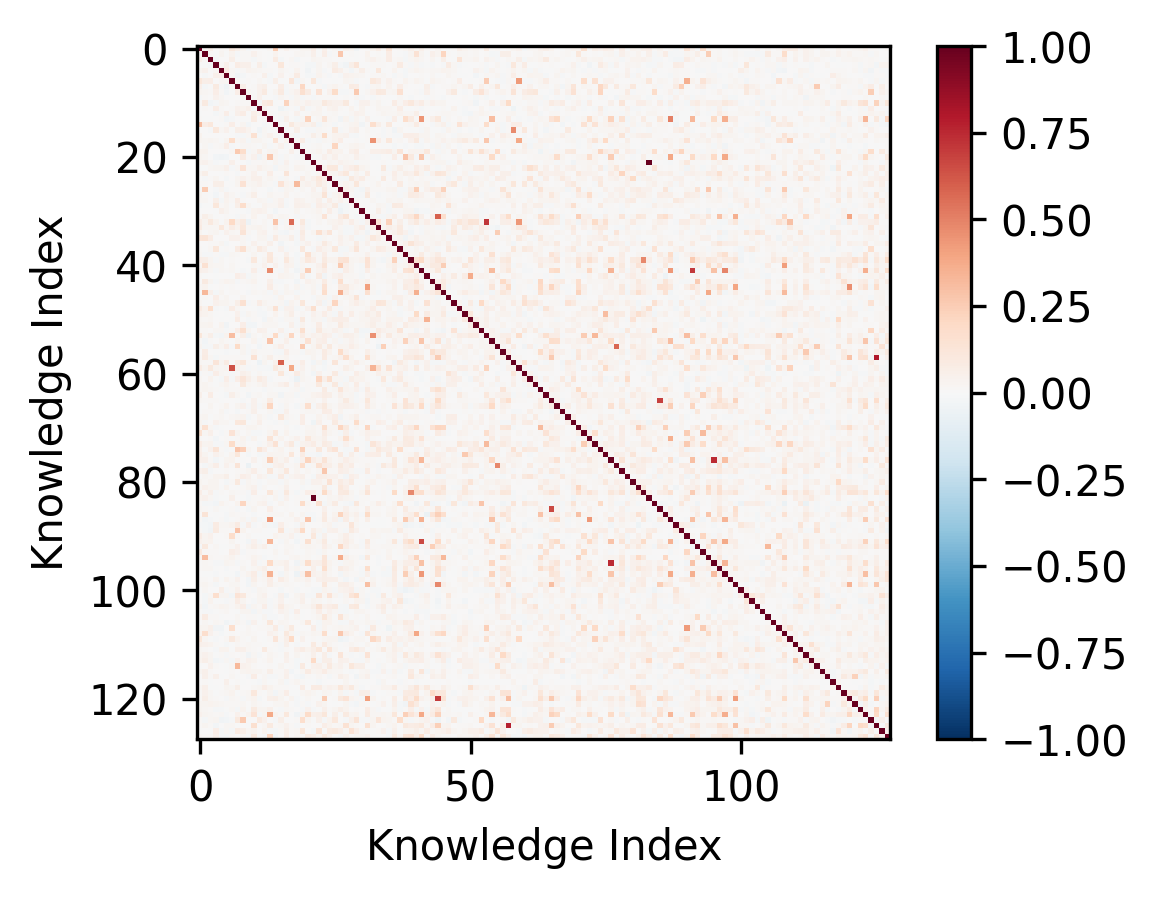}
    }
    \hfill
    \subfigure[Layer 16]{\includegraphics[width=0.22\textwidth]{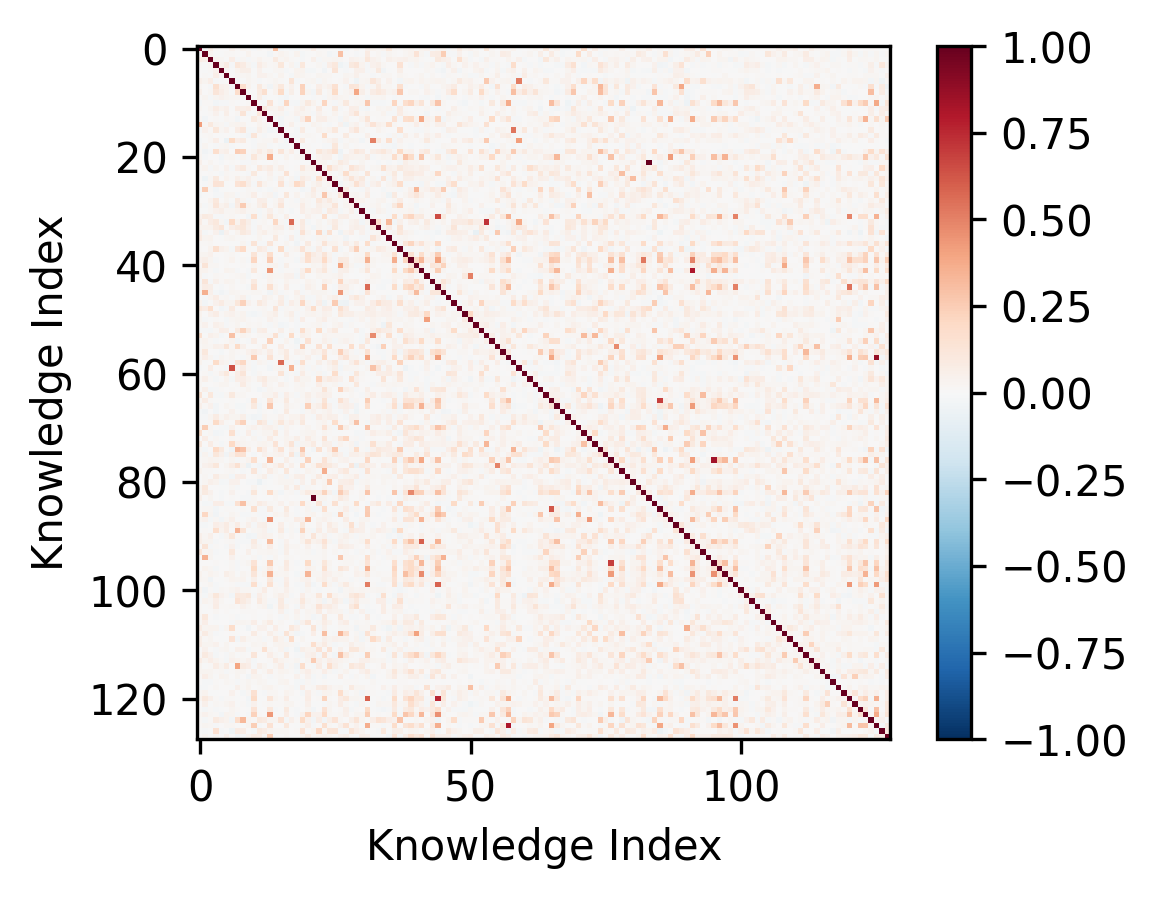}
    }
    \hfill
    \subfigure[Layer 17]{\includegraphics[width=0.22\textwidth]{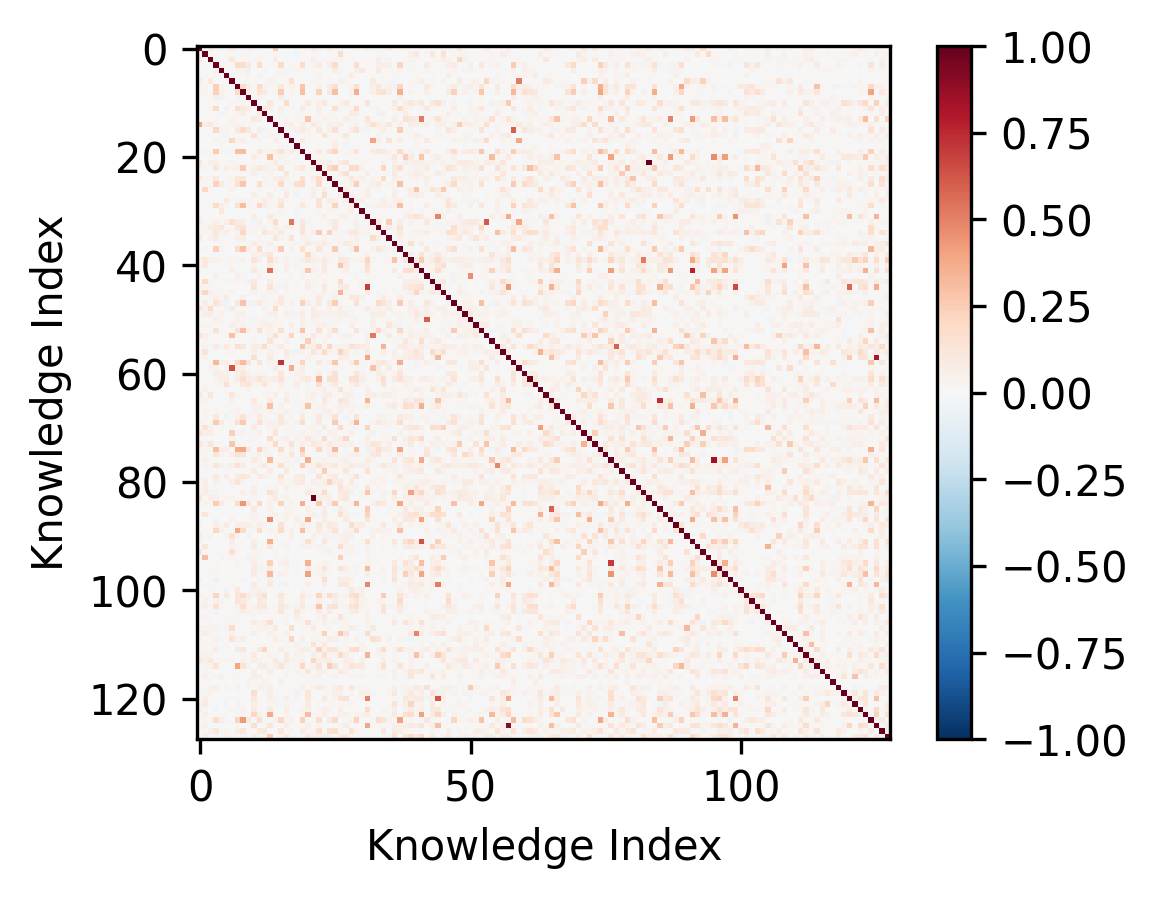}
    }
    \hfill
    \subfigure[Layer 18]{\includegraphics[width=0.22\textwidth]{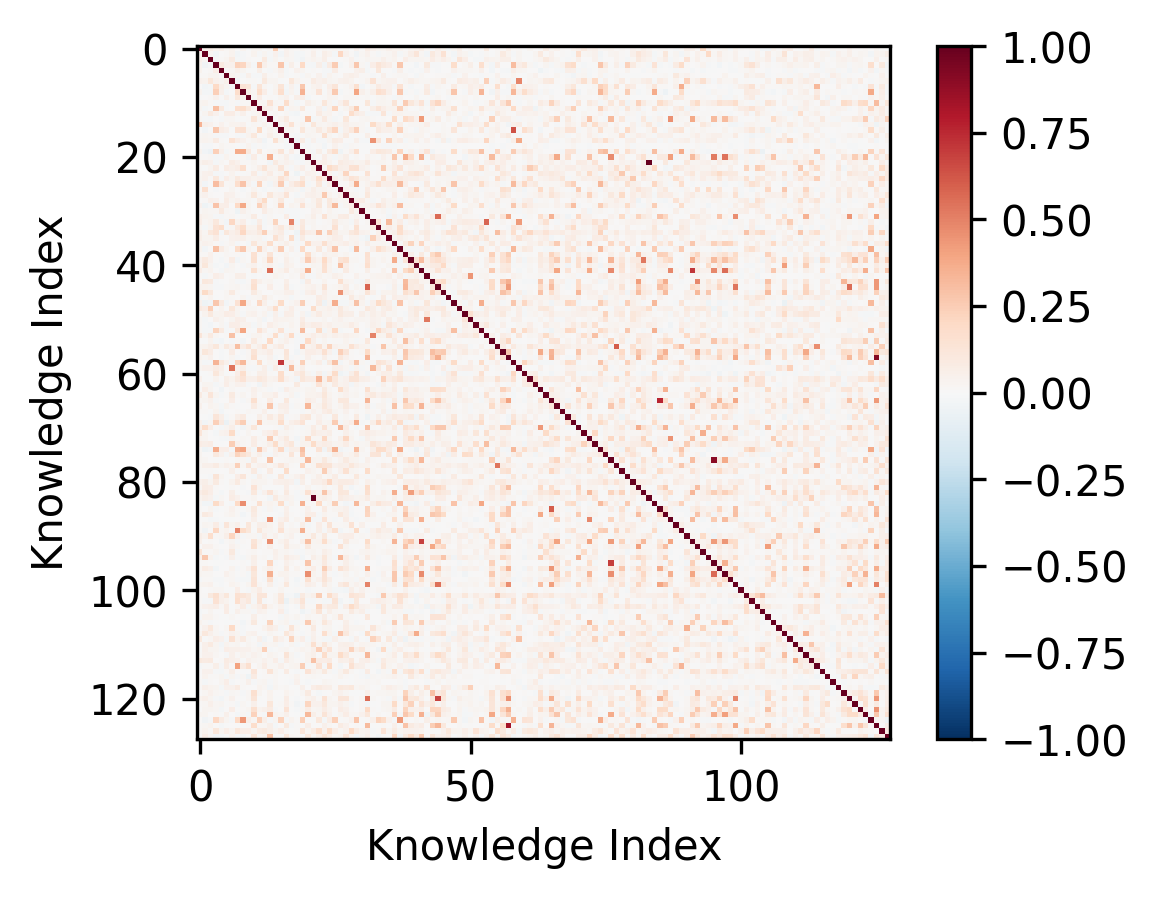}
    }
    \hfill
    \subfigure[Layer 19]{\includegraphics[width=0.22\textwidth]{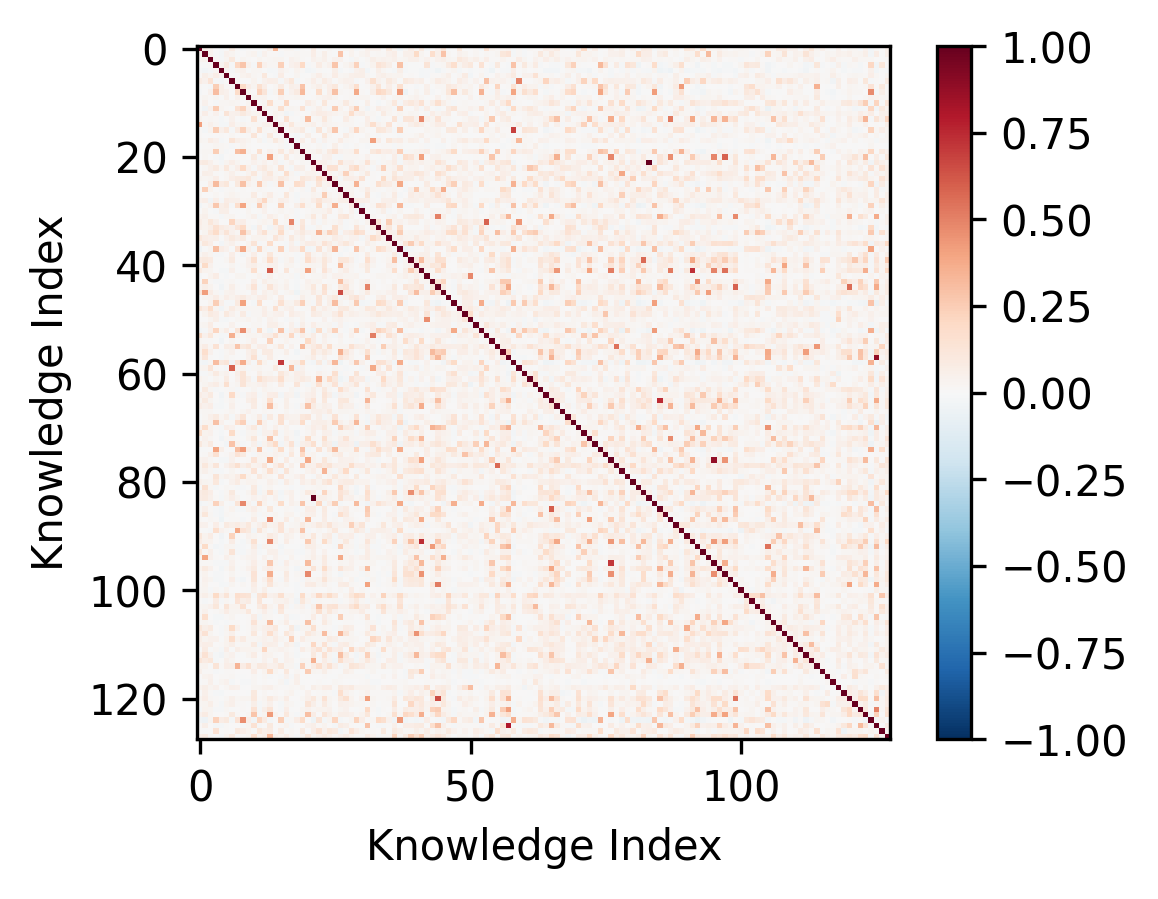}
    }
  \caption{In GPT2-Large, the visualization of \(P\) matrices across layers (0-19).}
  \label{fig:superposition_visualization_gpt2-large-part1}
\end{figure*}

\begin{figure*}
    \centering
    \subfigure[Layer 20]{\includegraphics[width=0.22\textwidth]{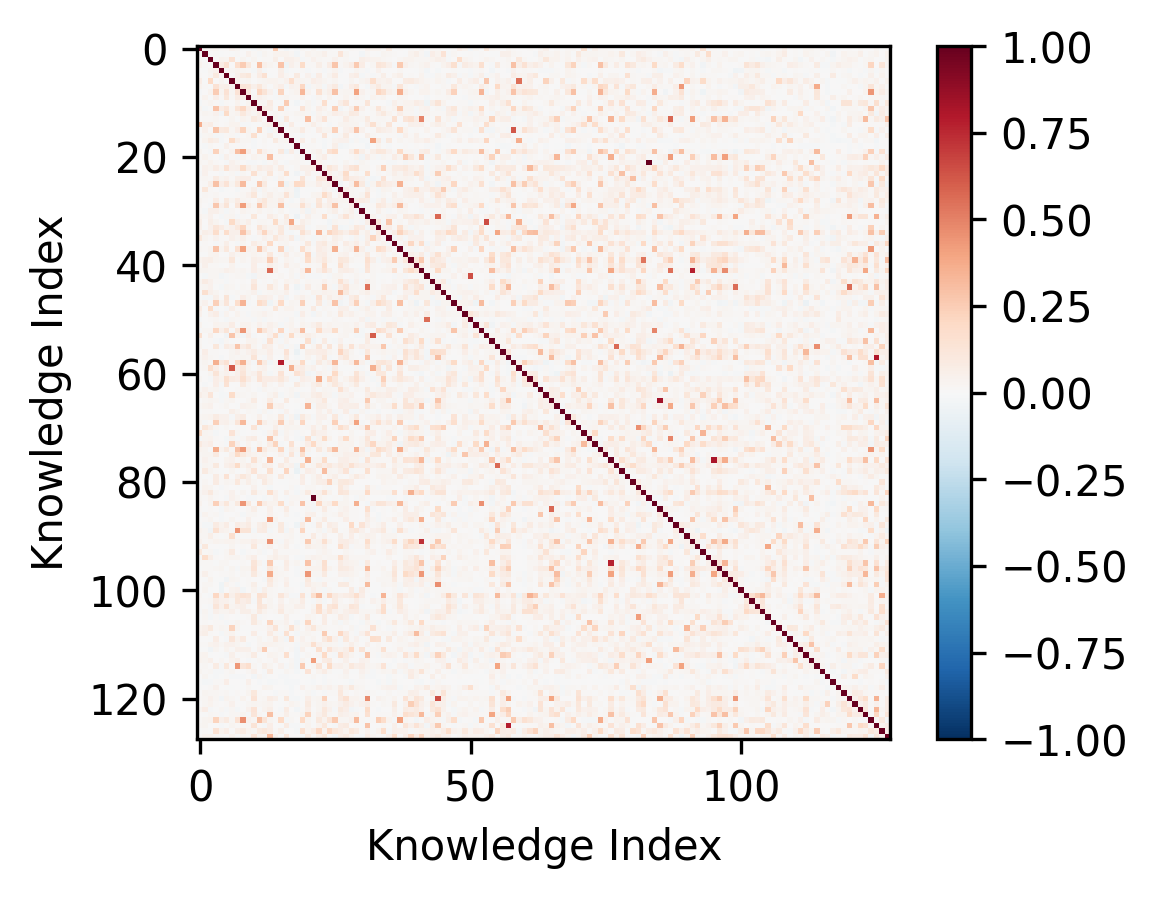}
    }
    \hfill
    \subfigure[Layer 21]{\includegraphics[width=0.22\textwidth]{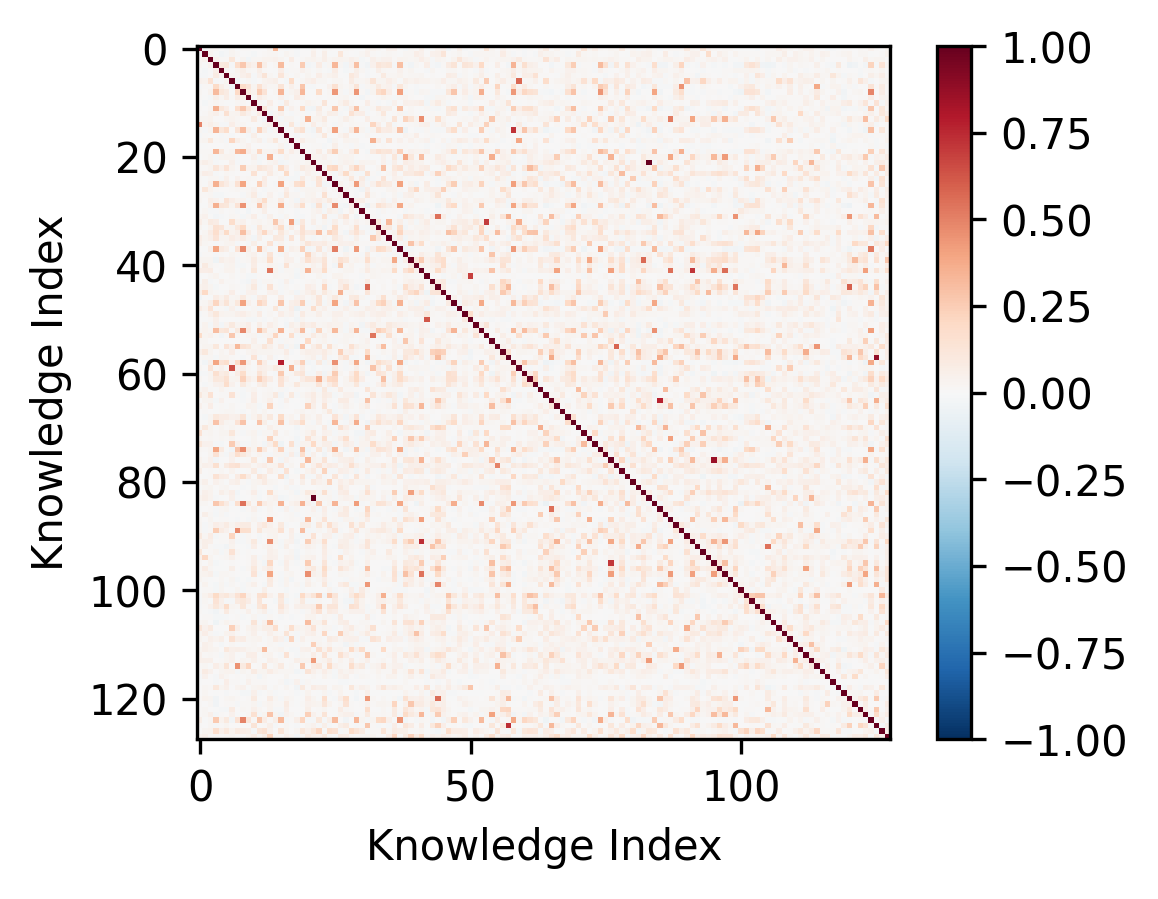}
    }
    \hfill
    \subfigure[Layer 22]{\includegraphics[width=0.22\textwidth]{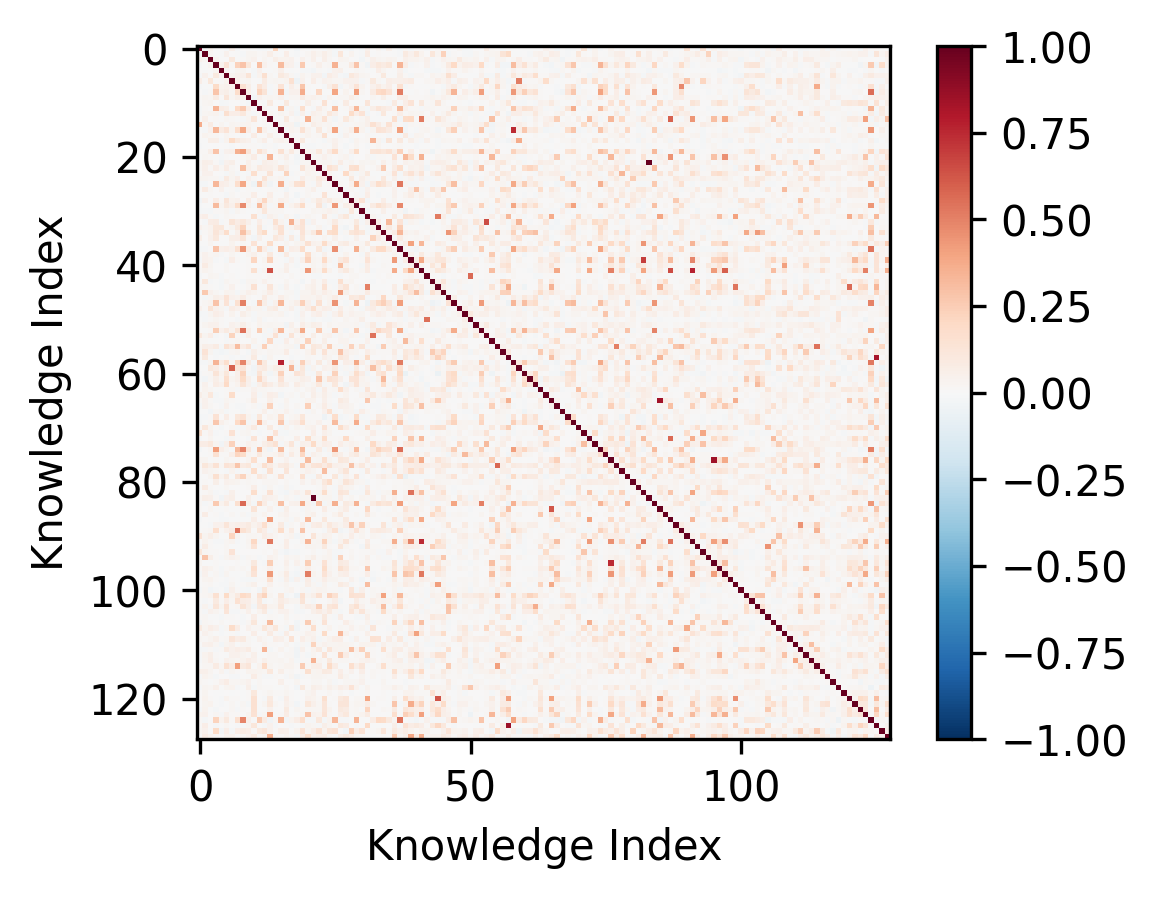}
    }
    \hfill
    \subfigure[Layer 23]{\includegraphics[width=0.22\textwidth]{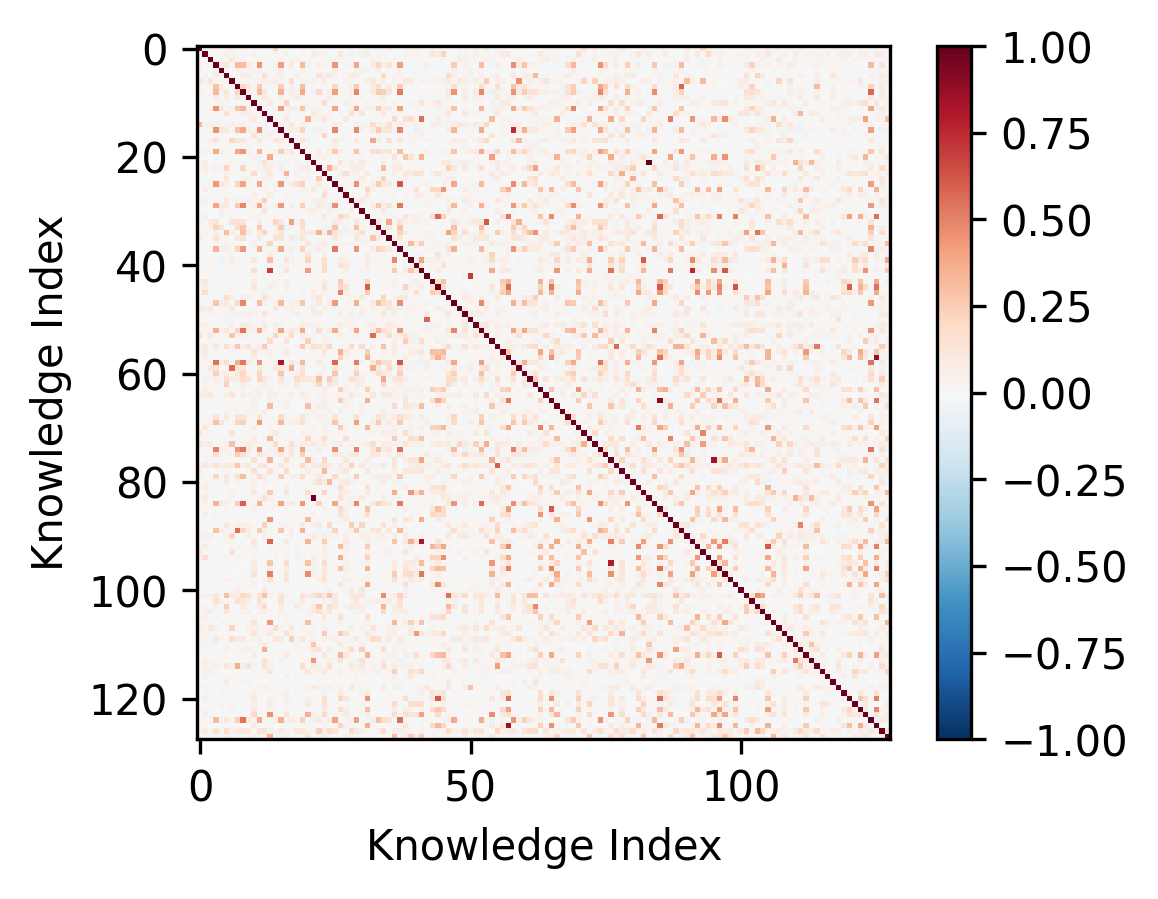}
    }
    \hfill
    \subfigure[Layer 24]{\includegraphics[width=0.22\textwidth]{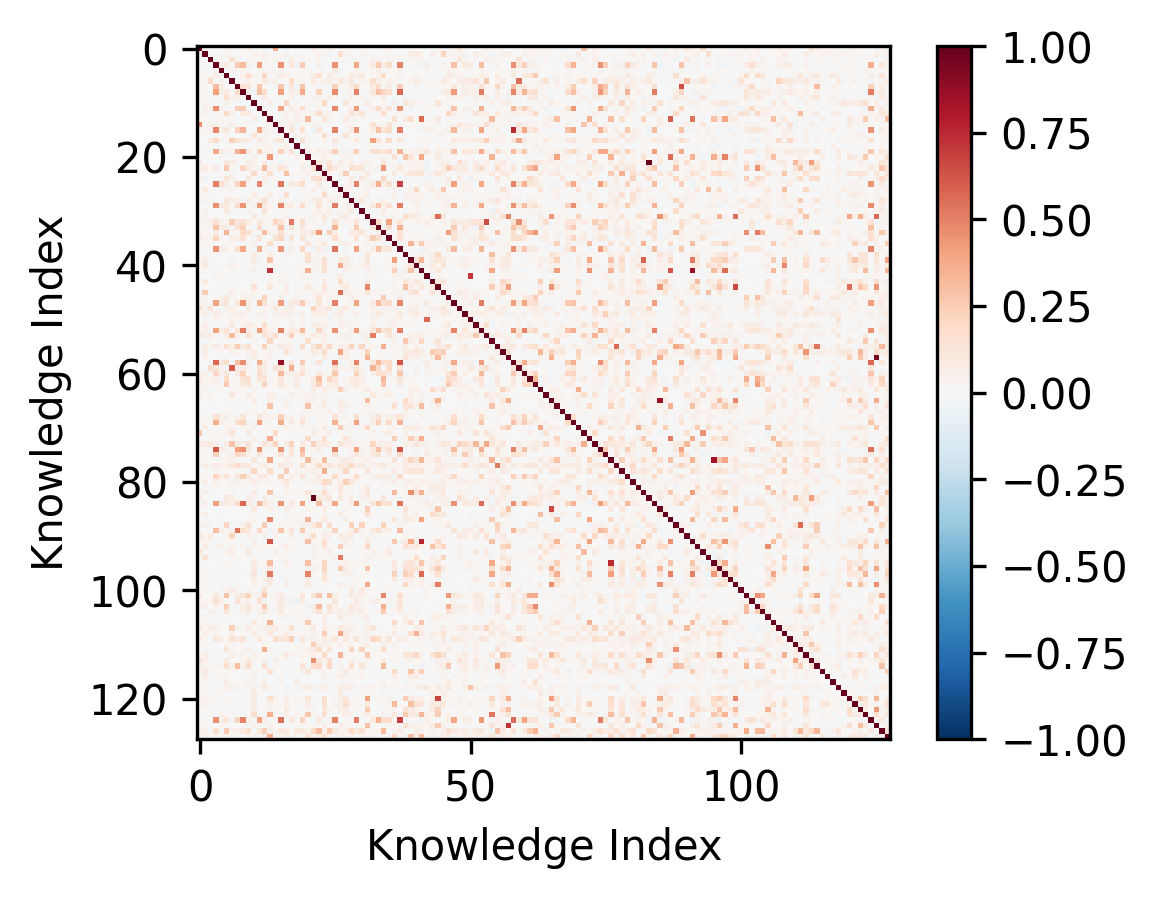}
    }
    \hfill
    \subfigure[Layer 25]{\includegraphics[width=0.22\textwidth]{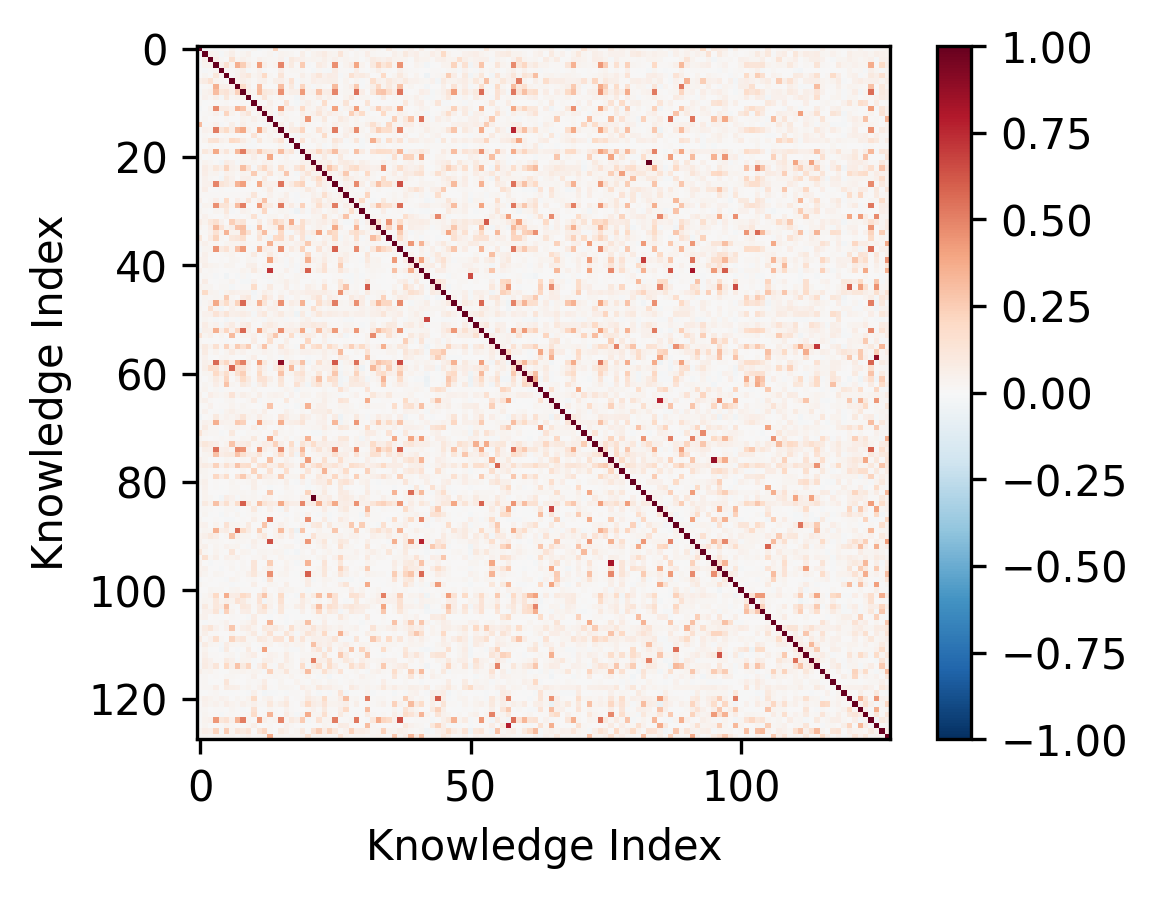}
    }
    \hfill
    \subfigure[Layer 26]{\includegraphics[width=0.22\textwidth]{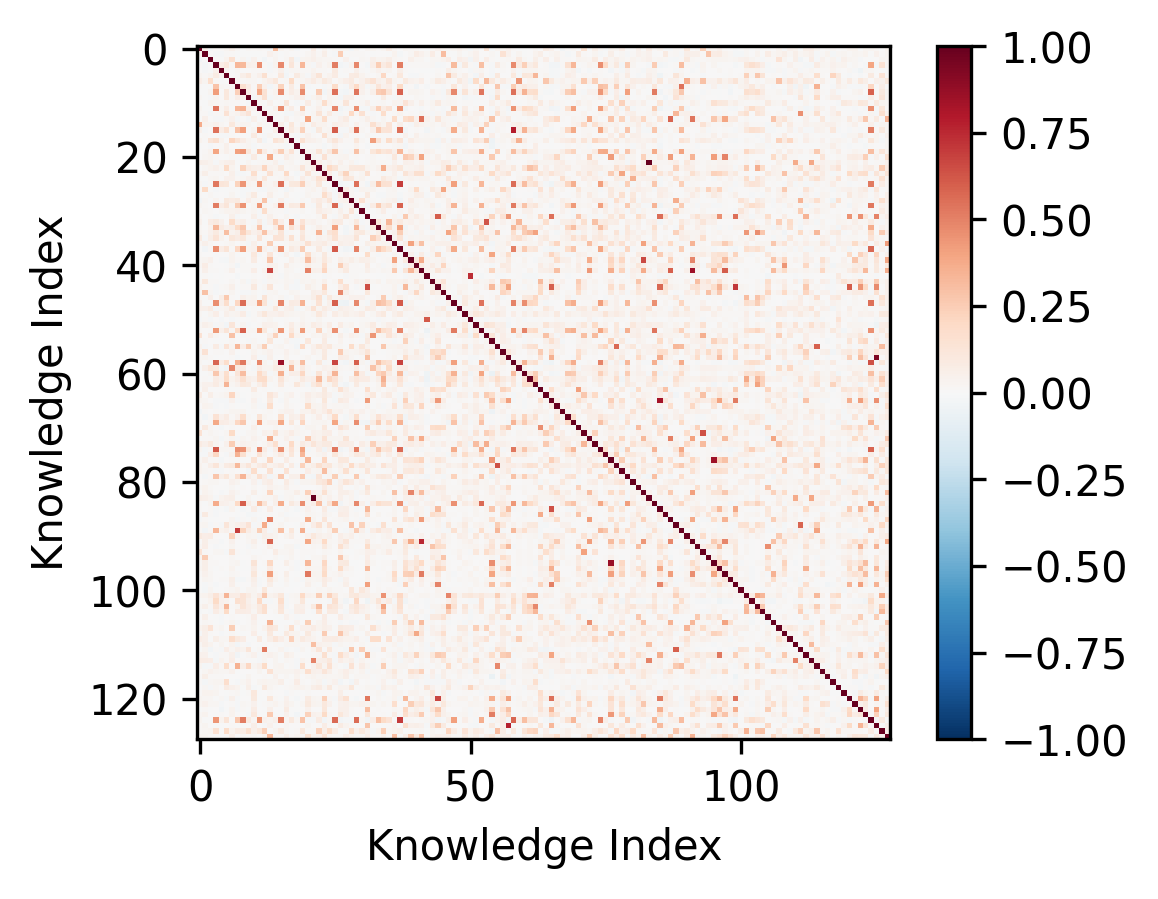}
    }
    \hfill
    \subfigure[Layer 27]{\includegraphics[width=0.22\textwidth]{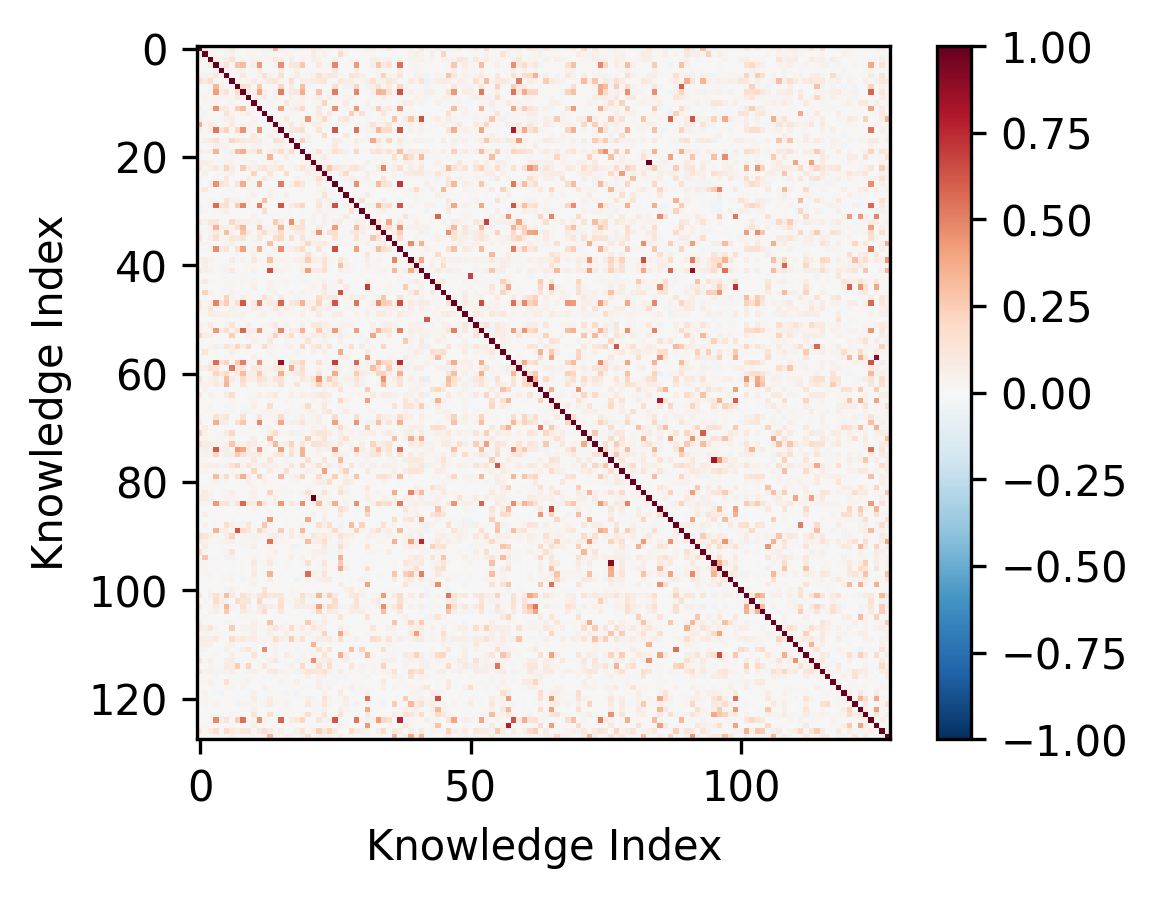}
    }
    \hfill
    \subfigure[Layer 28]{\includegraphics[width=0.22\textwidth]{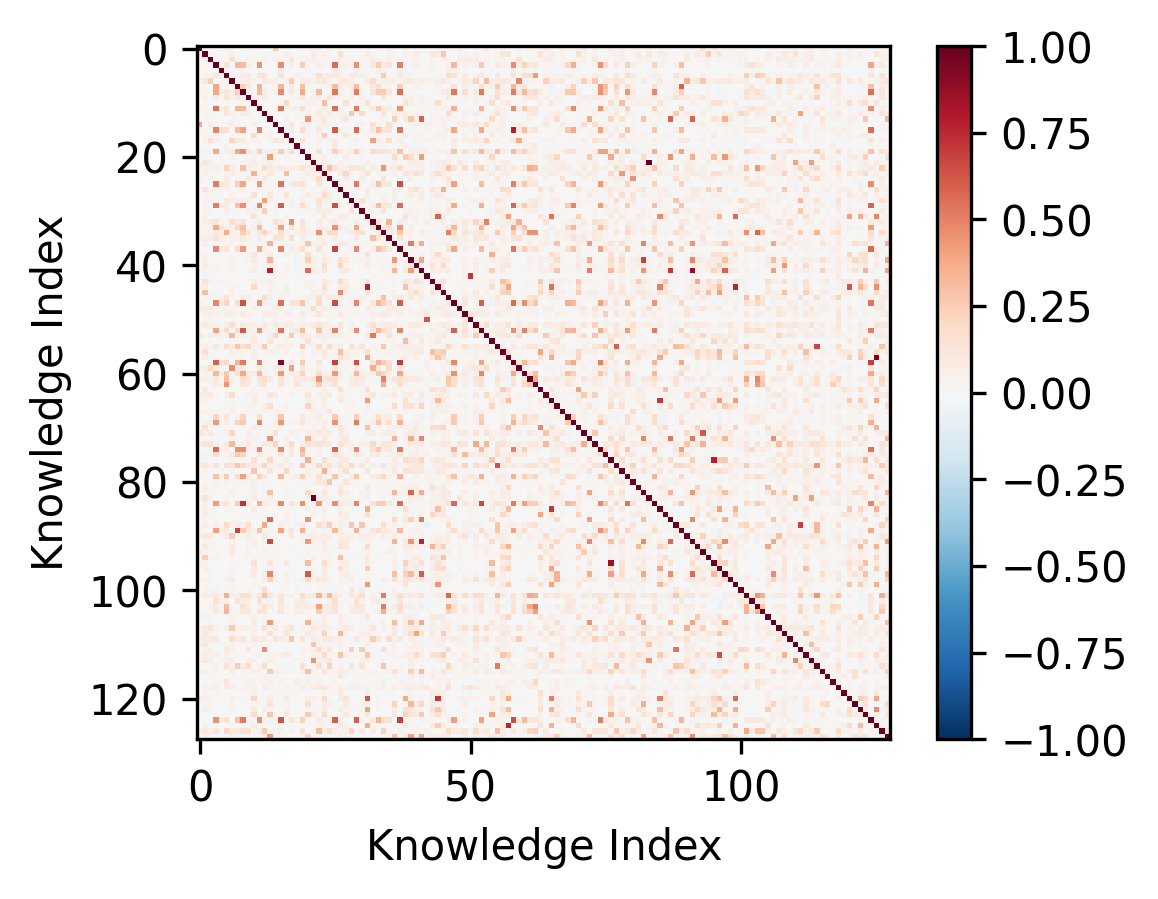}
    }
    \hfill
    \subfigure[Layer 29]{\includegraphics[width=0.22\textwidth]{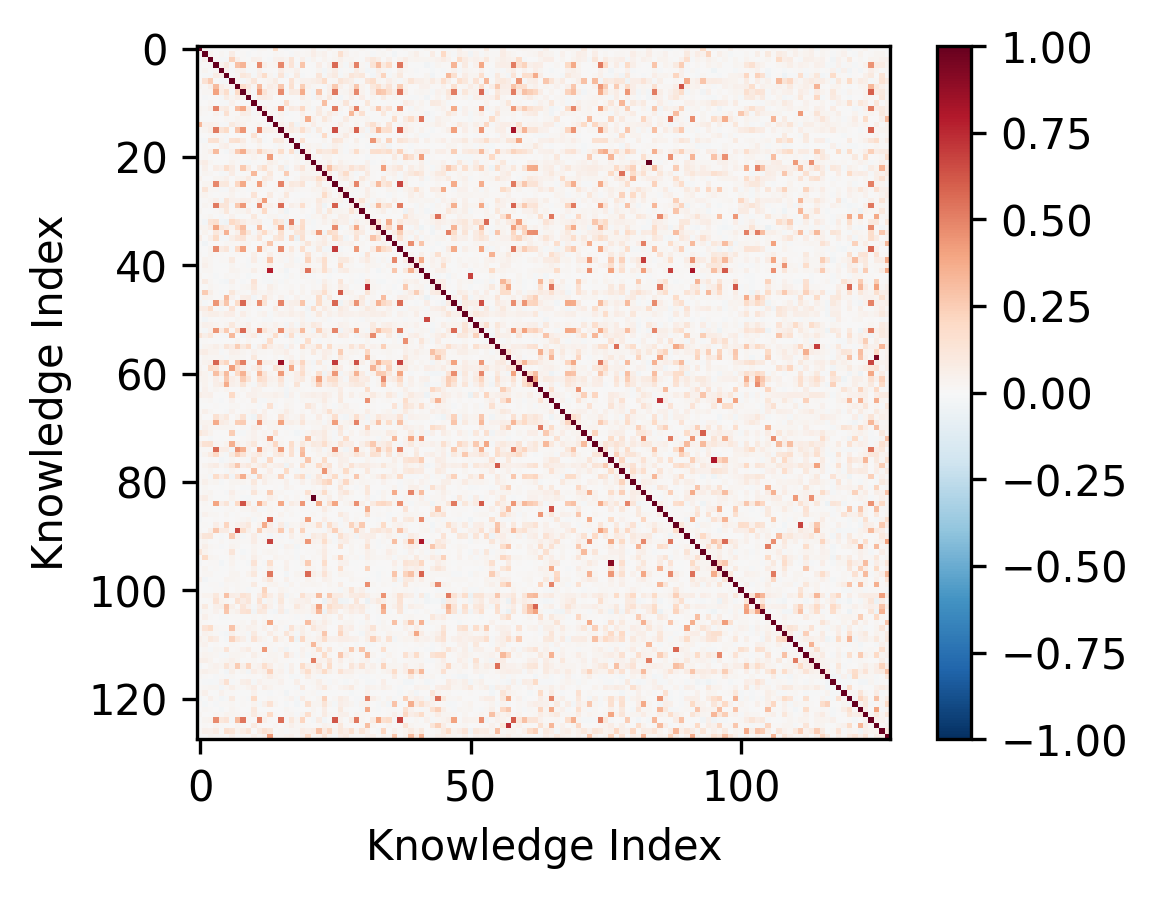}
    }
    \hfill
    \subfigure[Layer 30]{\includegraphics[width=0.22\textwidth]{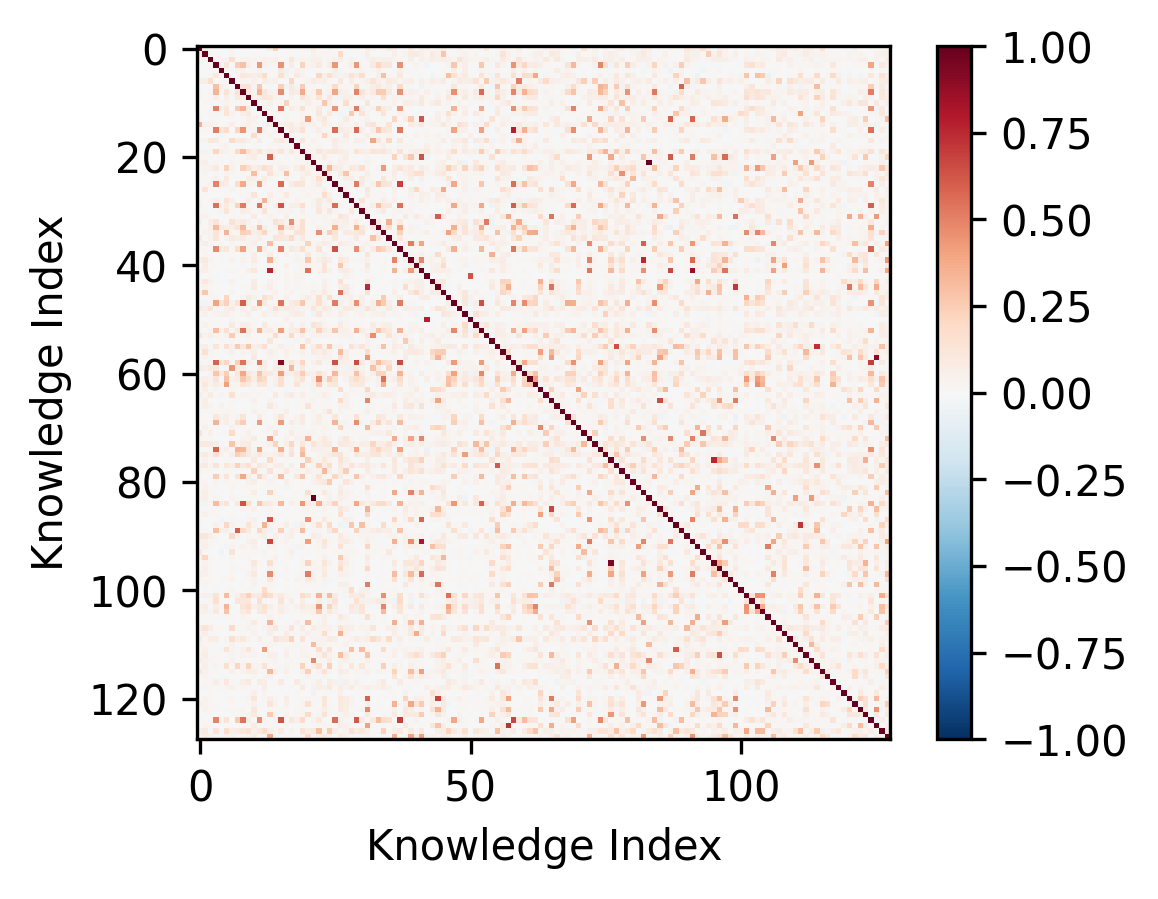}
    }
    \hfill
    \subfigure[Layer 31]{\includegraphics[width=0.22\textwidth]{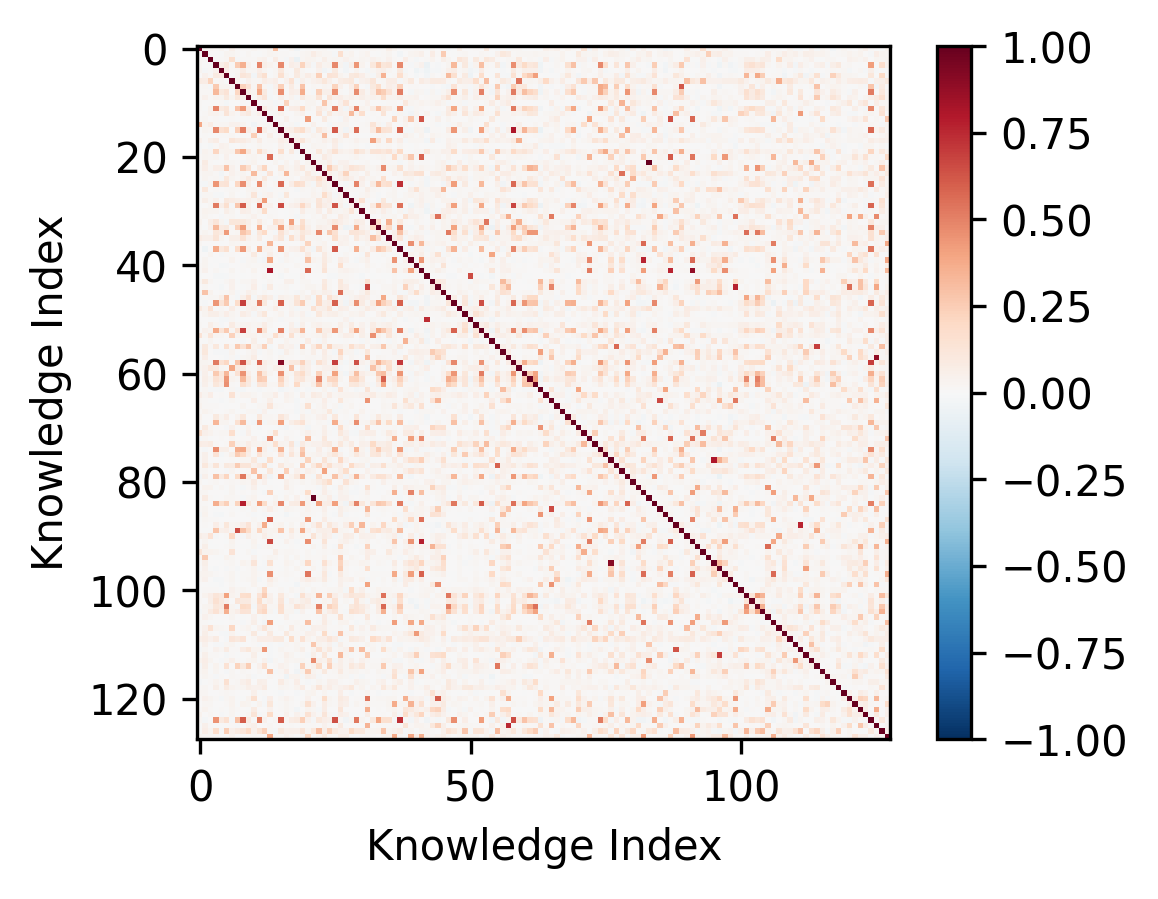}
    }
    \hfill
    \subfigure[Layer 32]{\includegraphics[width=0.22\textwidth]{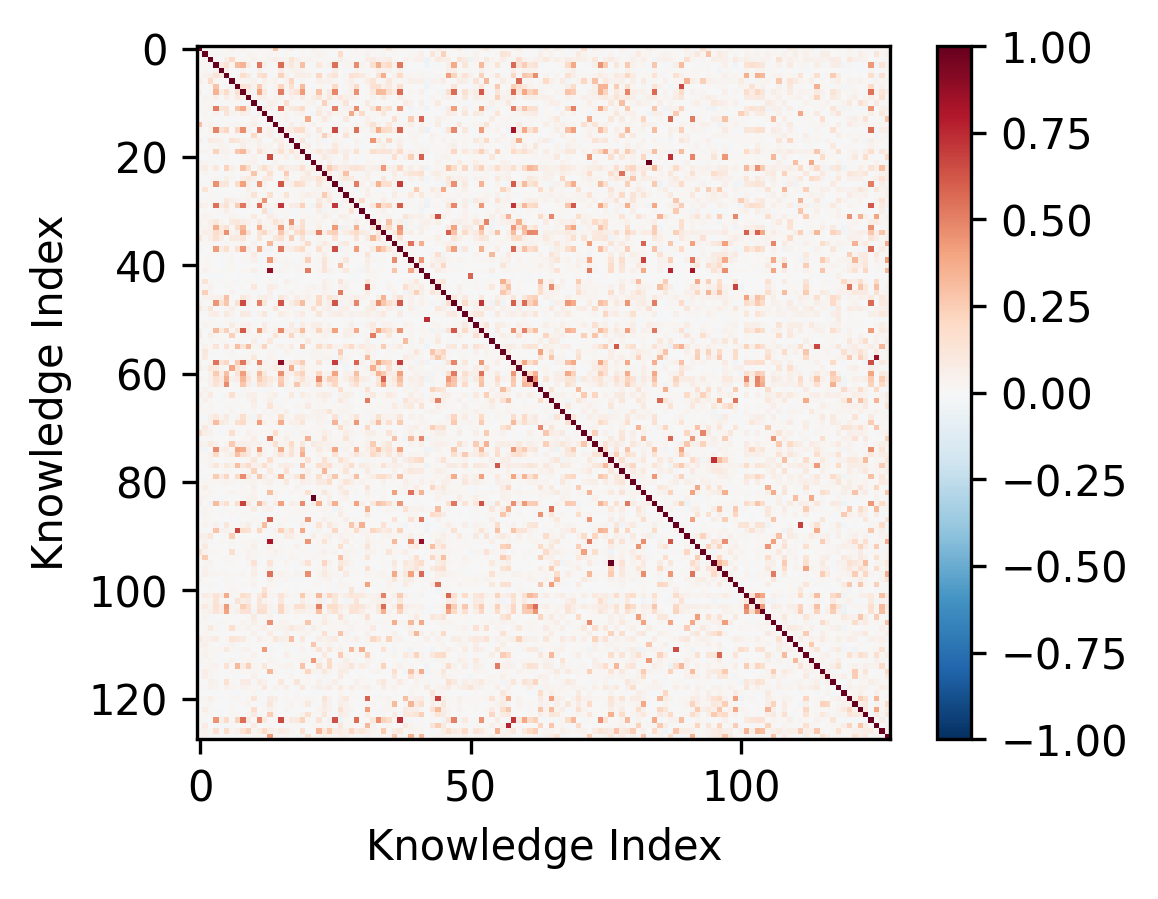}
    }
    \hfill
    \subfigure[Layer 33]{\includegraphics[width=0.22\textwidth]{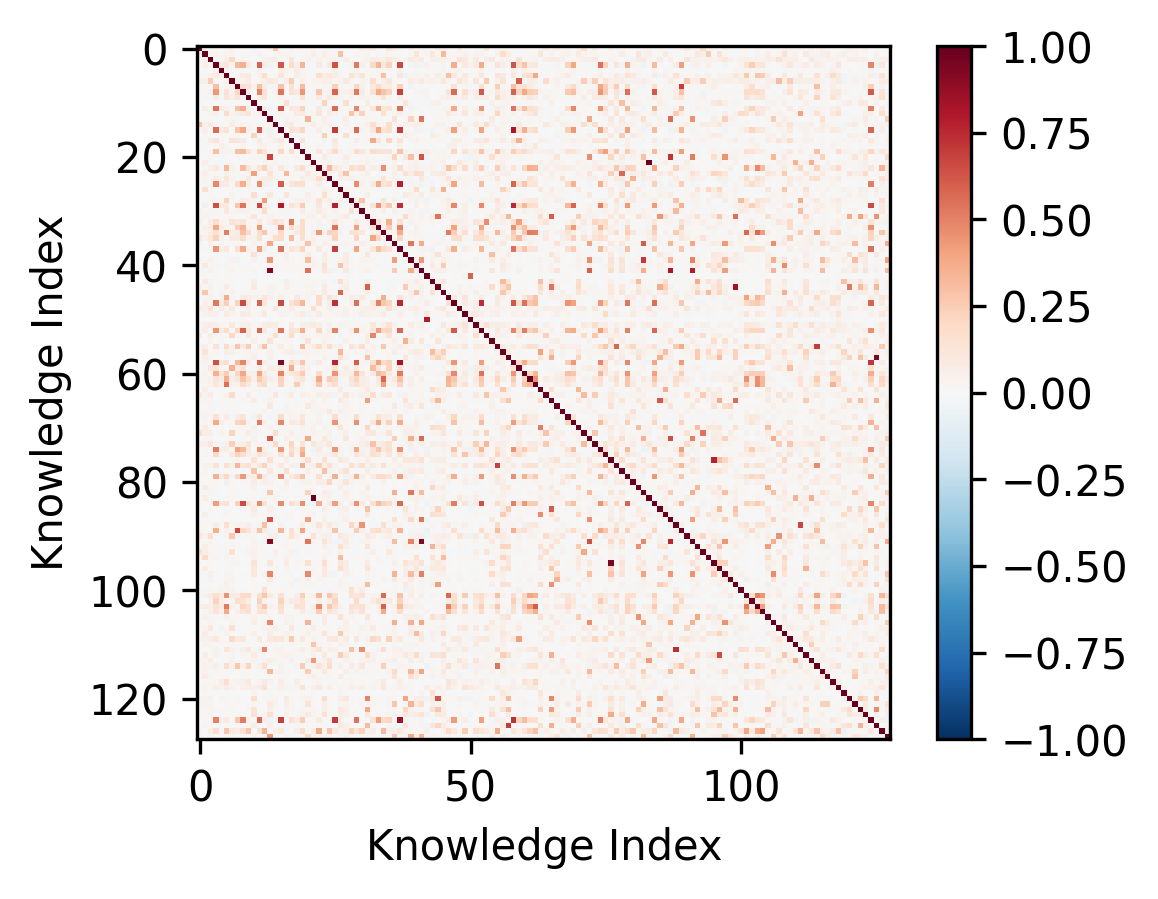}
    }
    \hfill
    \subfigure[Layer 34]{\includegraphics[width=0.22\textwidth]{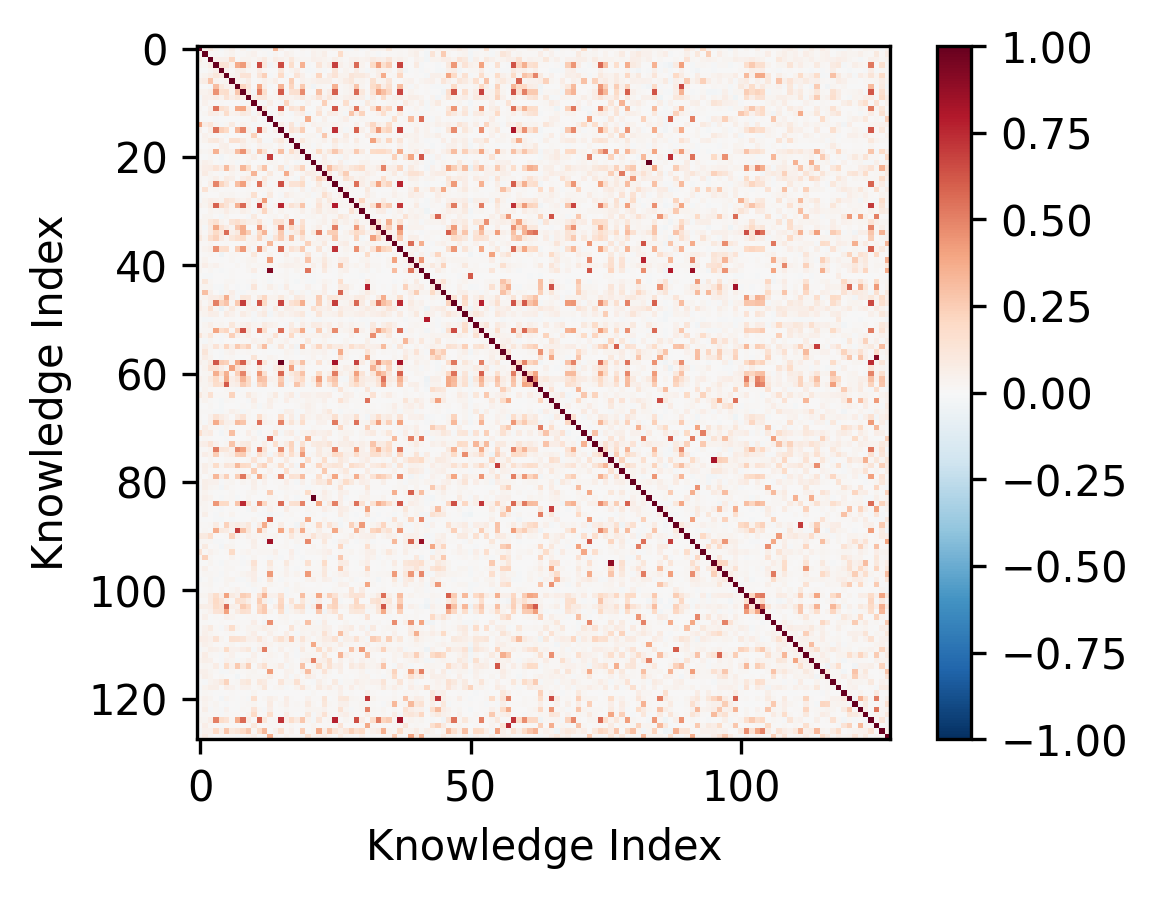}
    }
    \hfill
    \subfigure[Layer 35]{\includegraphics[width=0.22\textwidth]{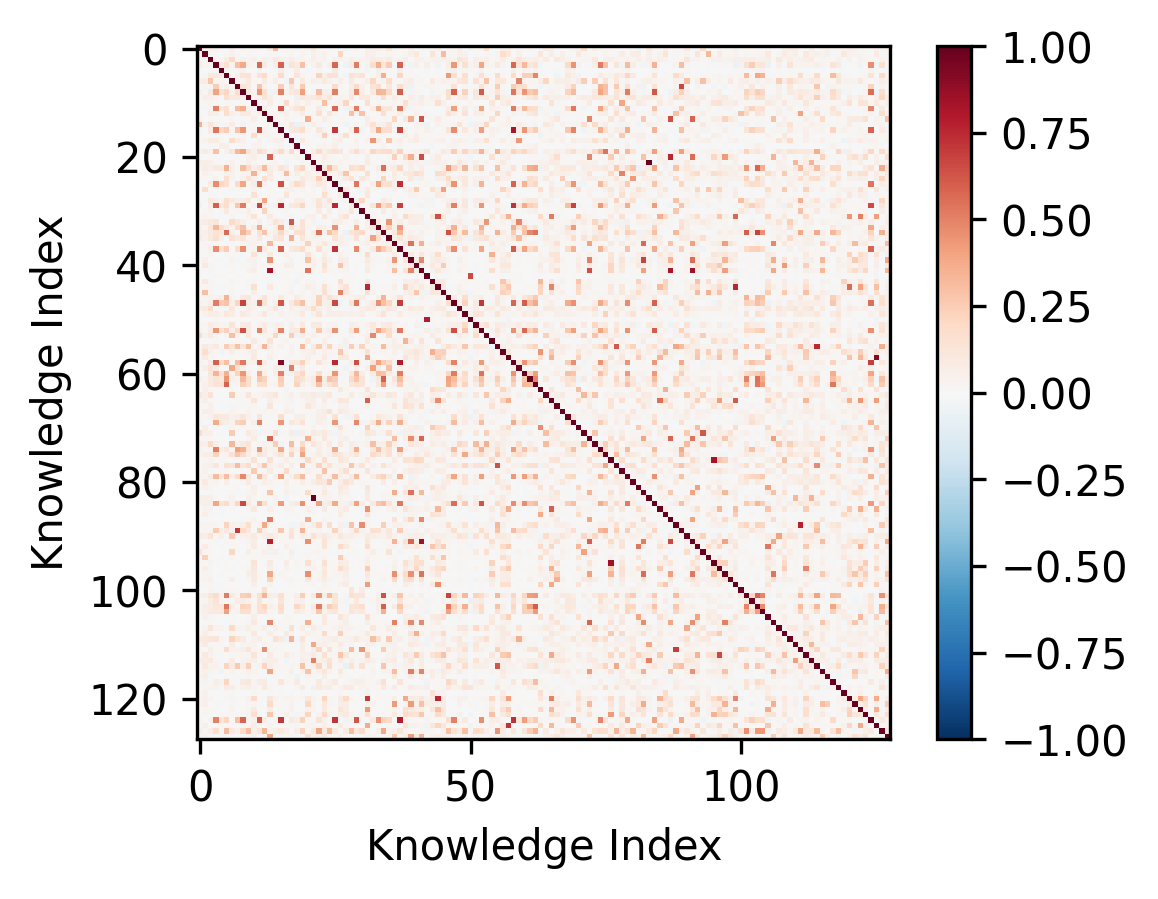}
    }
  \caption{In GPT2-Large, the visualization of \(P\) matrices across layers (20-35).}
  \label{fig:superposition_visualization_gpt2-large-part2}
\end{figure*}

\begin{figure*}
    \centering
    \subfigure[Layer 0]{\includegraphics[width=0.22\textwidth]{fig/EleutherAI/gpt-j-6B/p_matrix/known/heatmap/superposition_for_layer_0.png}
    }
    \hfill
    \subfigure[Layer 1]{\includegraphics[width=0.22\textwidth]{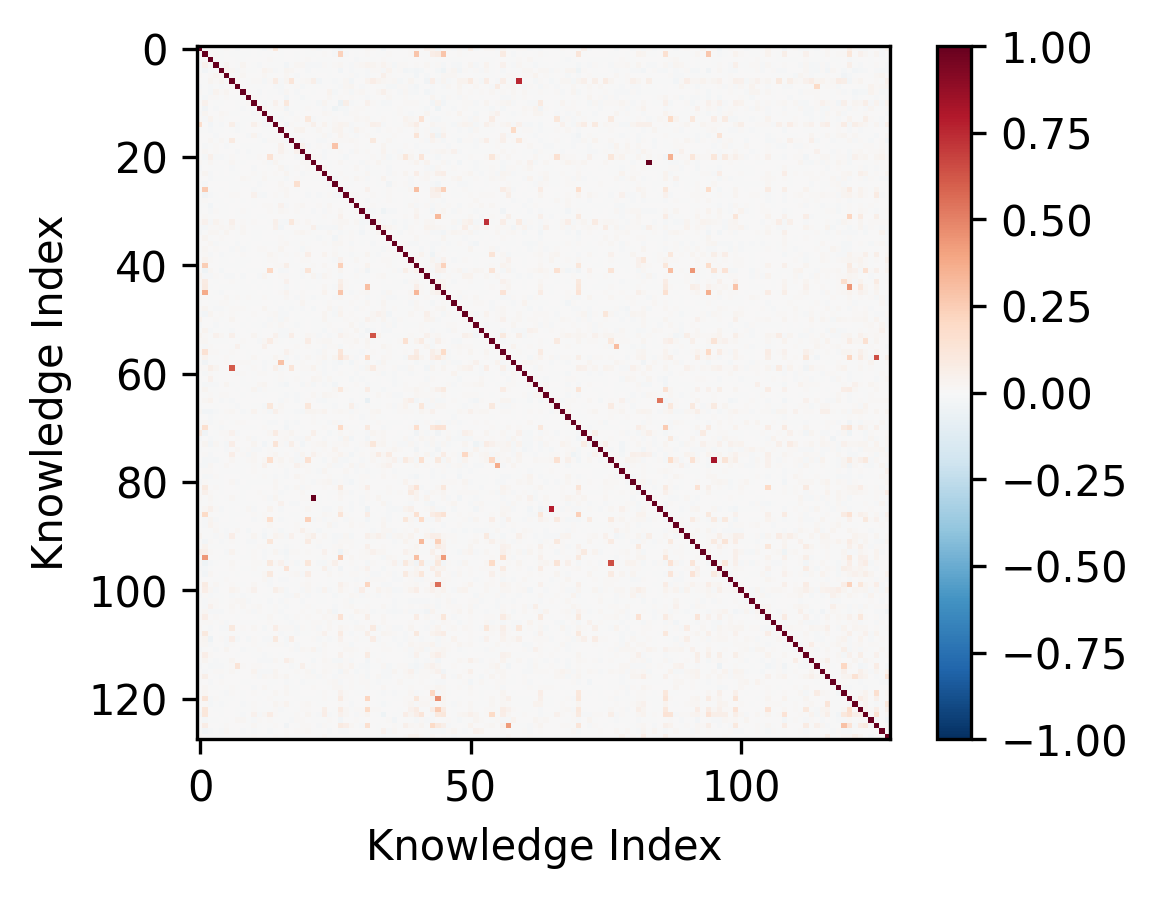}
    }
    \hfill
    \subfigure[Layer 2]{\includegraphics[width=0.22\textwidth]{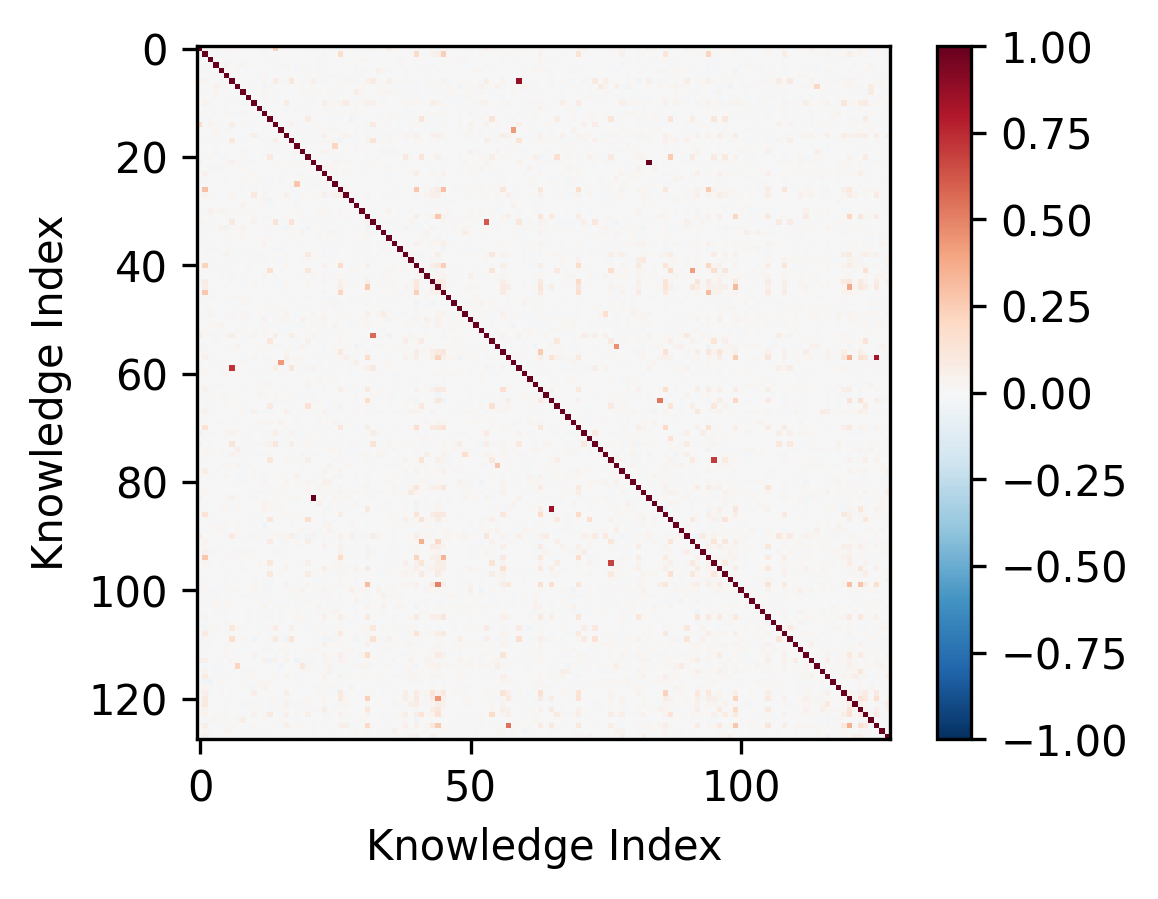}
    }
    \hfill
    \subfigure[Layer 3]{\includegraphics[width=0.22\textwidth]{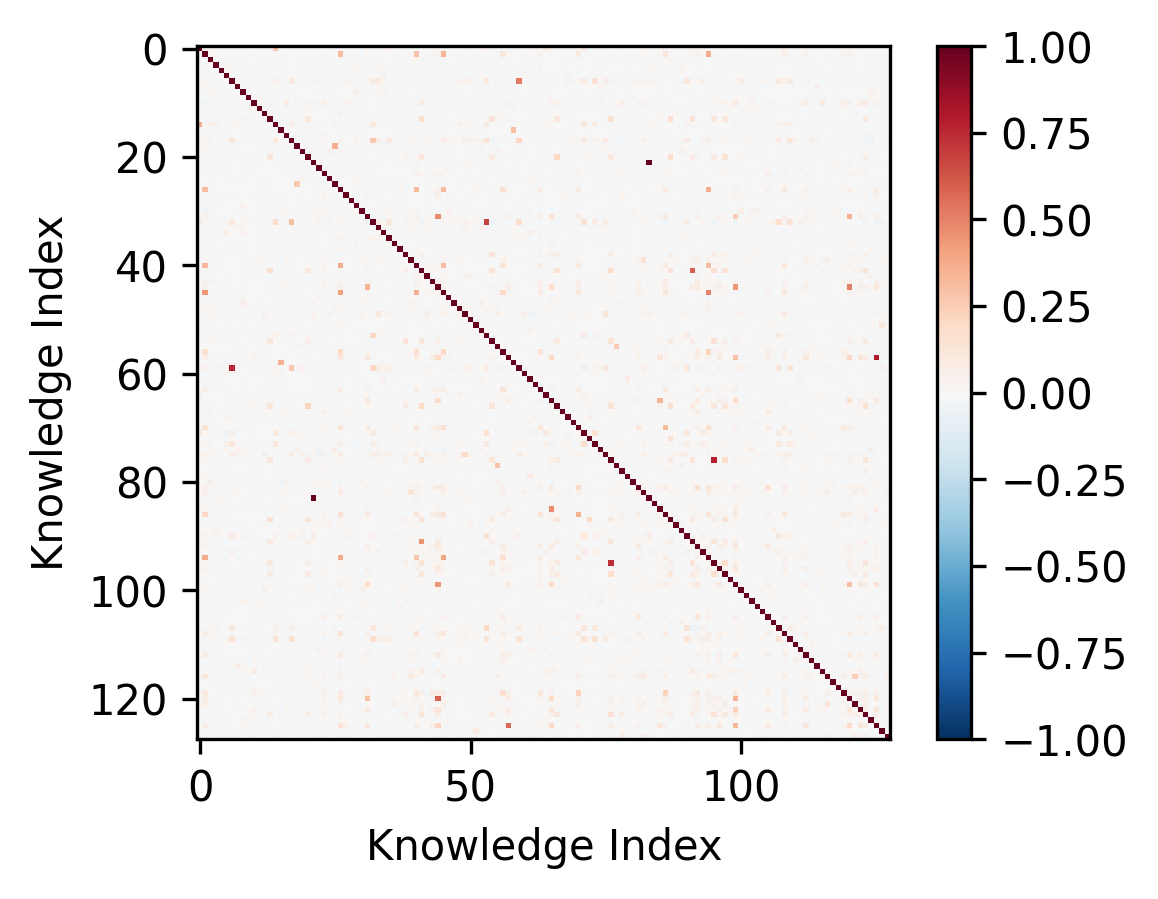}
    }
    \hfill
    \subfigure[Layer 4]{\includegraphics[width=0.22\textwidth]{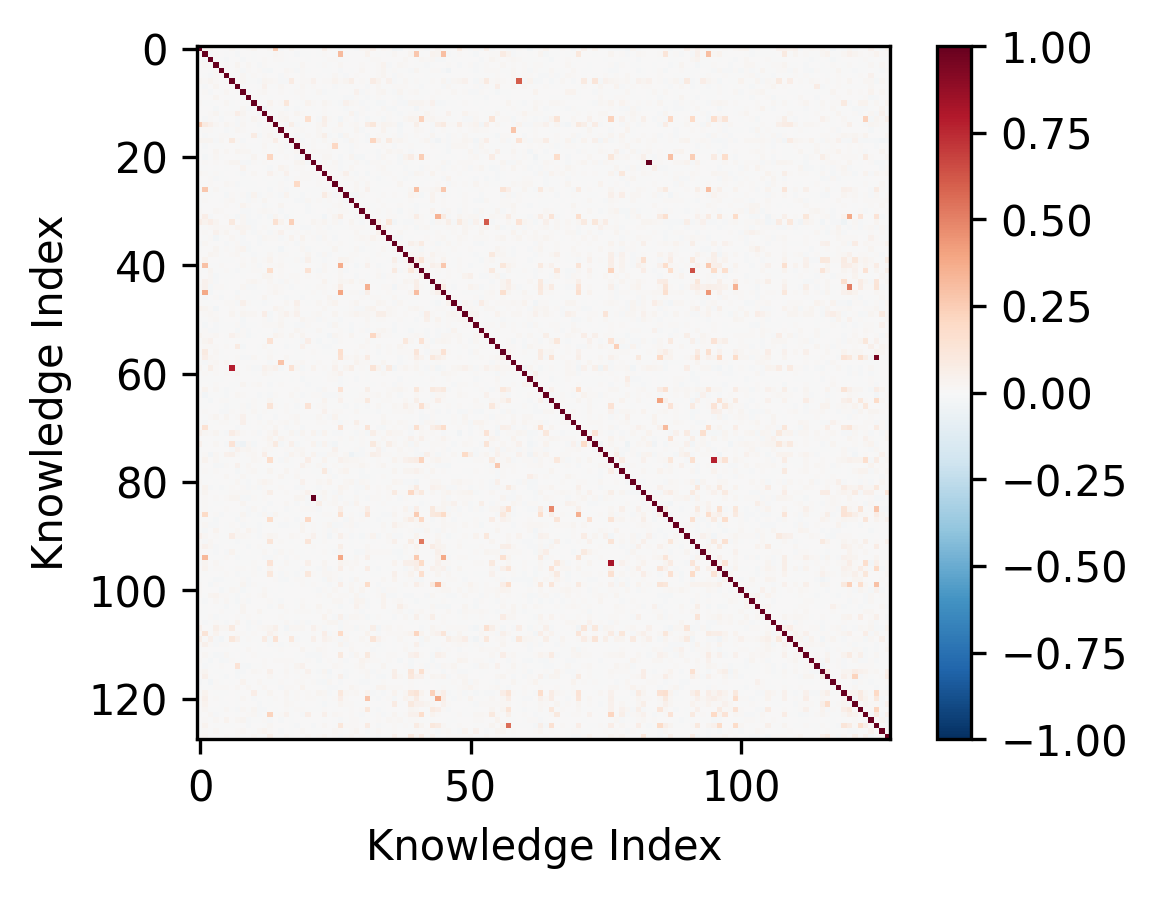}
    }
    \hfill
    \subfigure[Layer 5]{\includegraphics[width=0.22\textwidth]{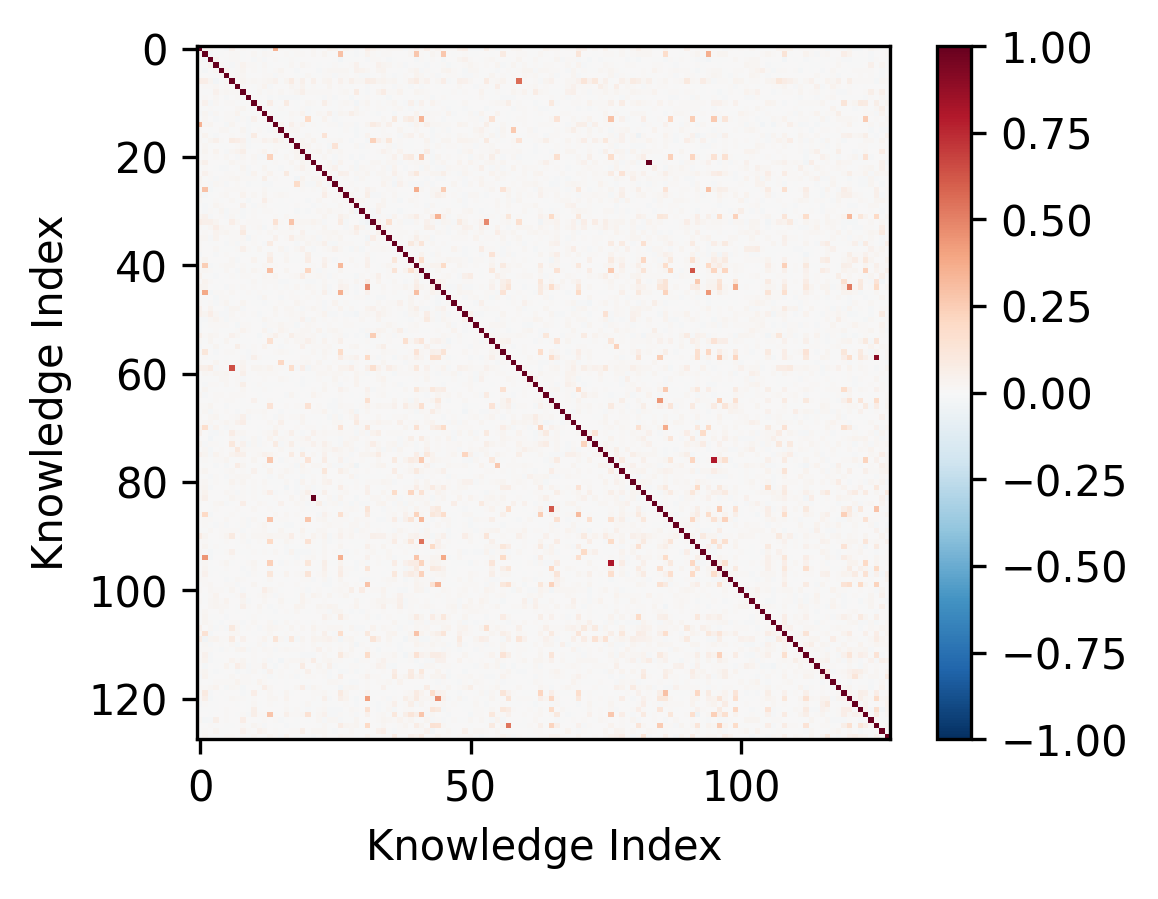}
    }
    \hfill
    \subfigure[Layer 6]{\includegraphics[width=0.22\textwidth]{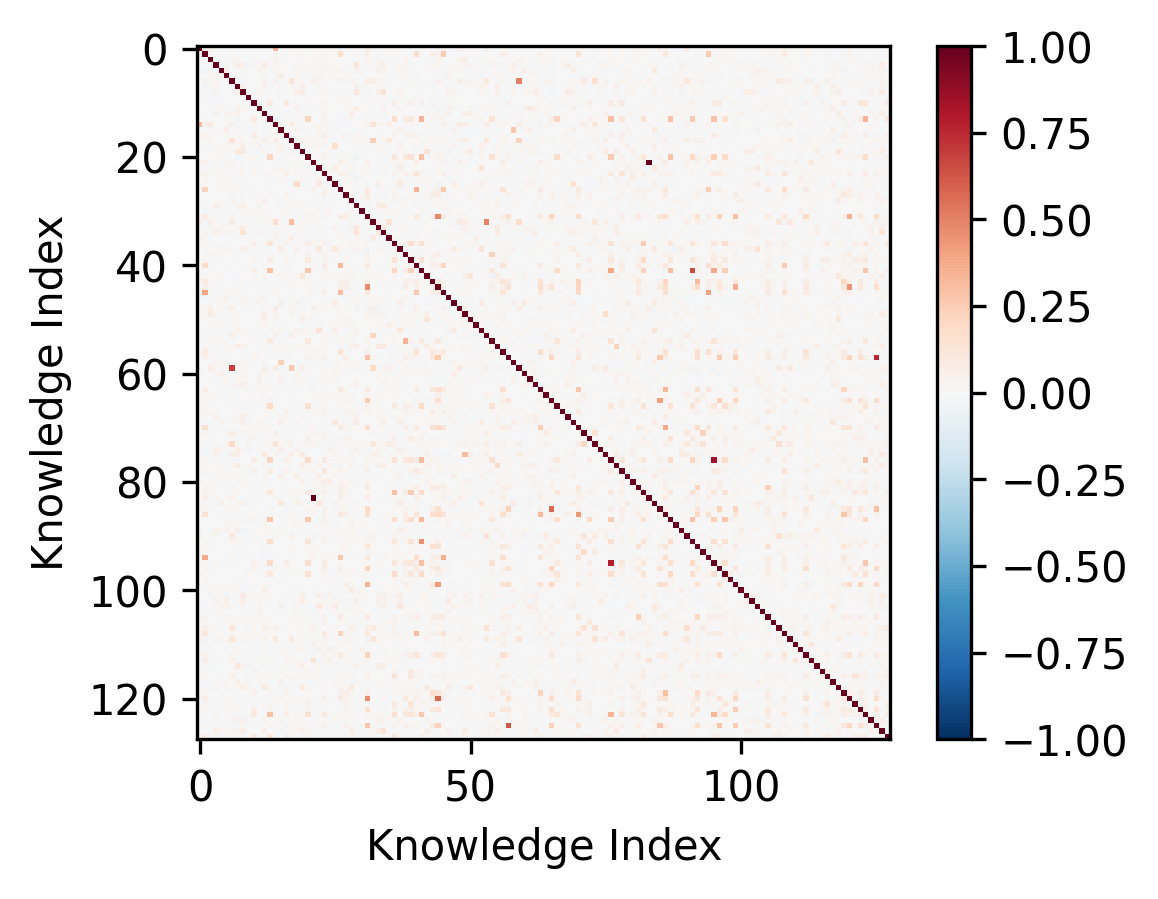}
    }
    \hfill
    \subfigure[Layer 7]{\includegraphics[width=0.22\textwidth]{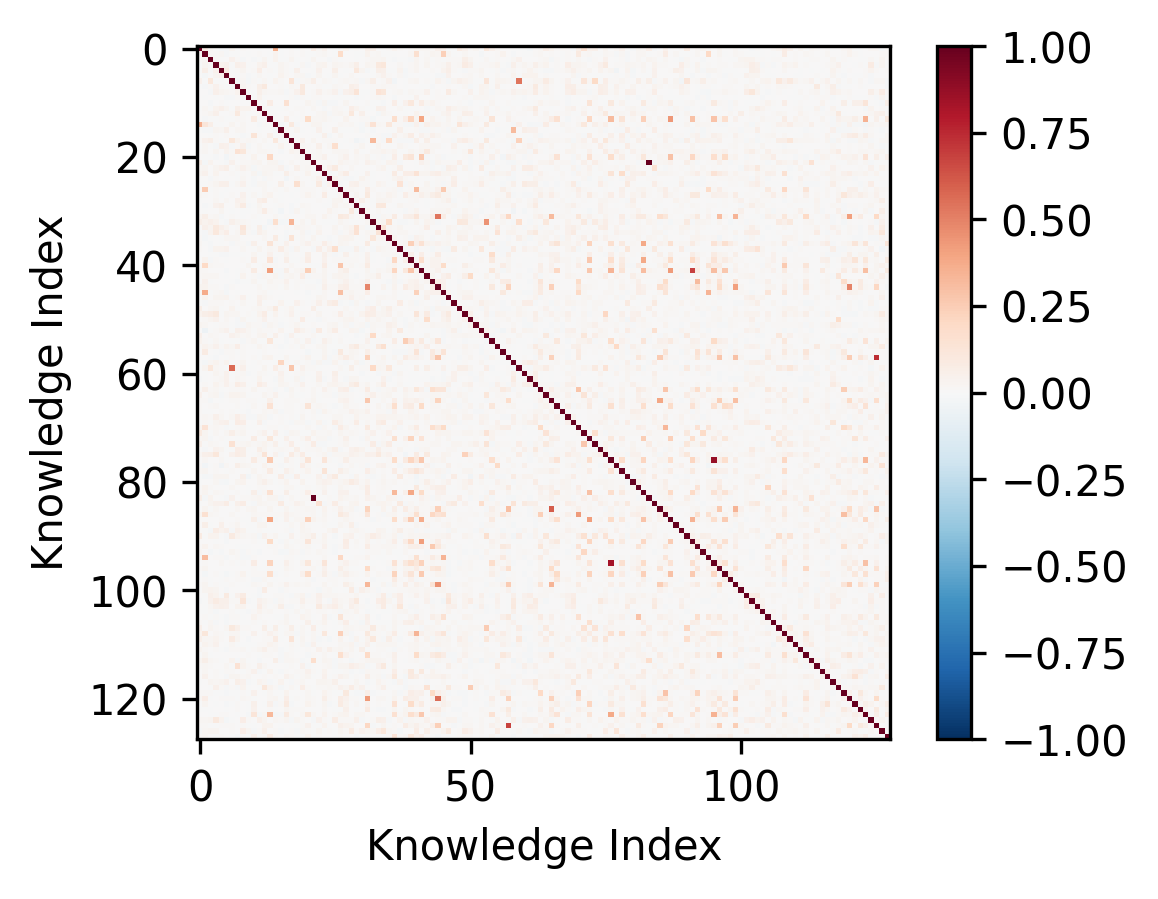}
    }
    \hfill
    \subfigure[Layer 8]{\includegraphics[width=0.22\textwidth]{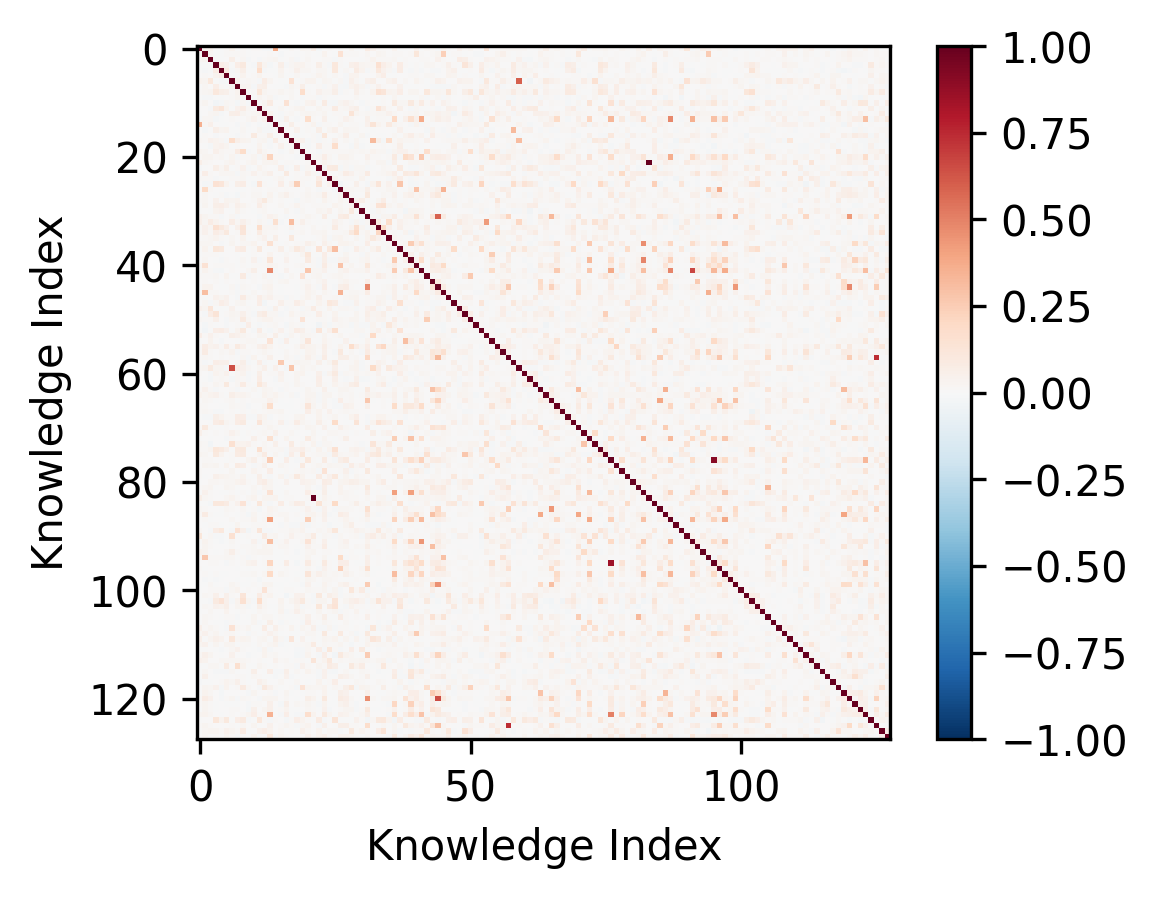}
    }
    \hfill
    \subfigure[Layer 9]{\includegraphics[width=0.22\textwidth]{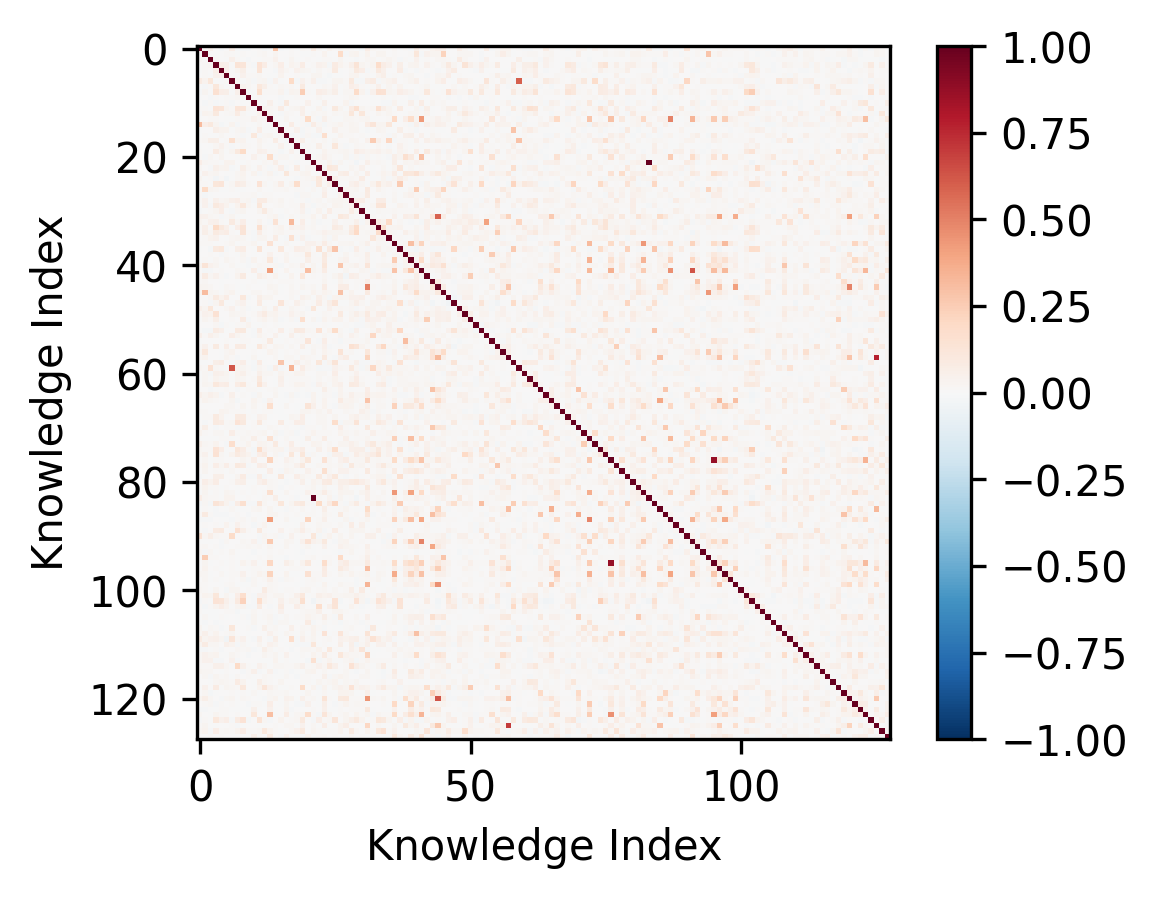}
    }
    \hfill
    \subfigure[Layer 10]{\includegraphics[width=0.22\textwidth]{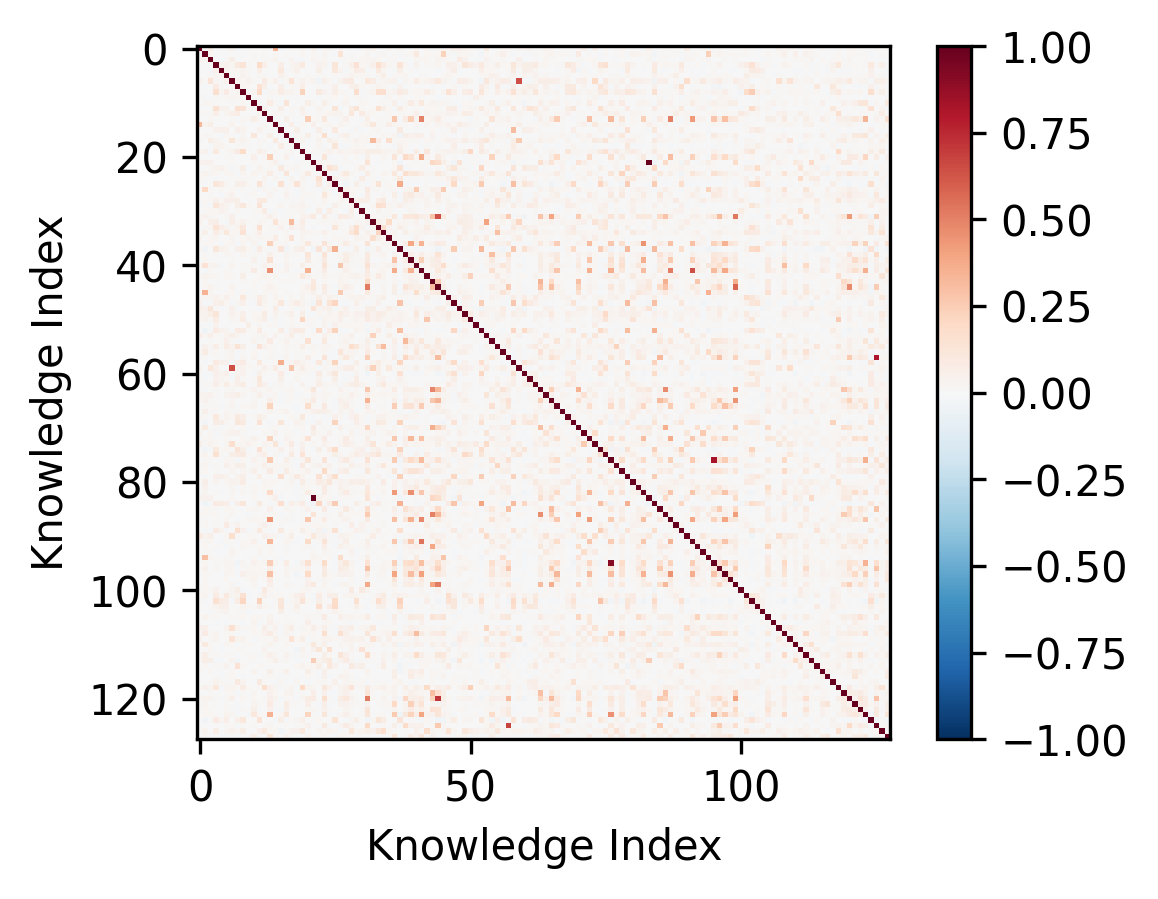}
    }
    \hfill
    \subfigure[Layer 11]{\includegraphics[width=0.22\textwidth]{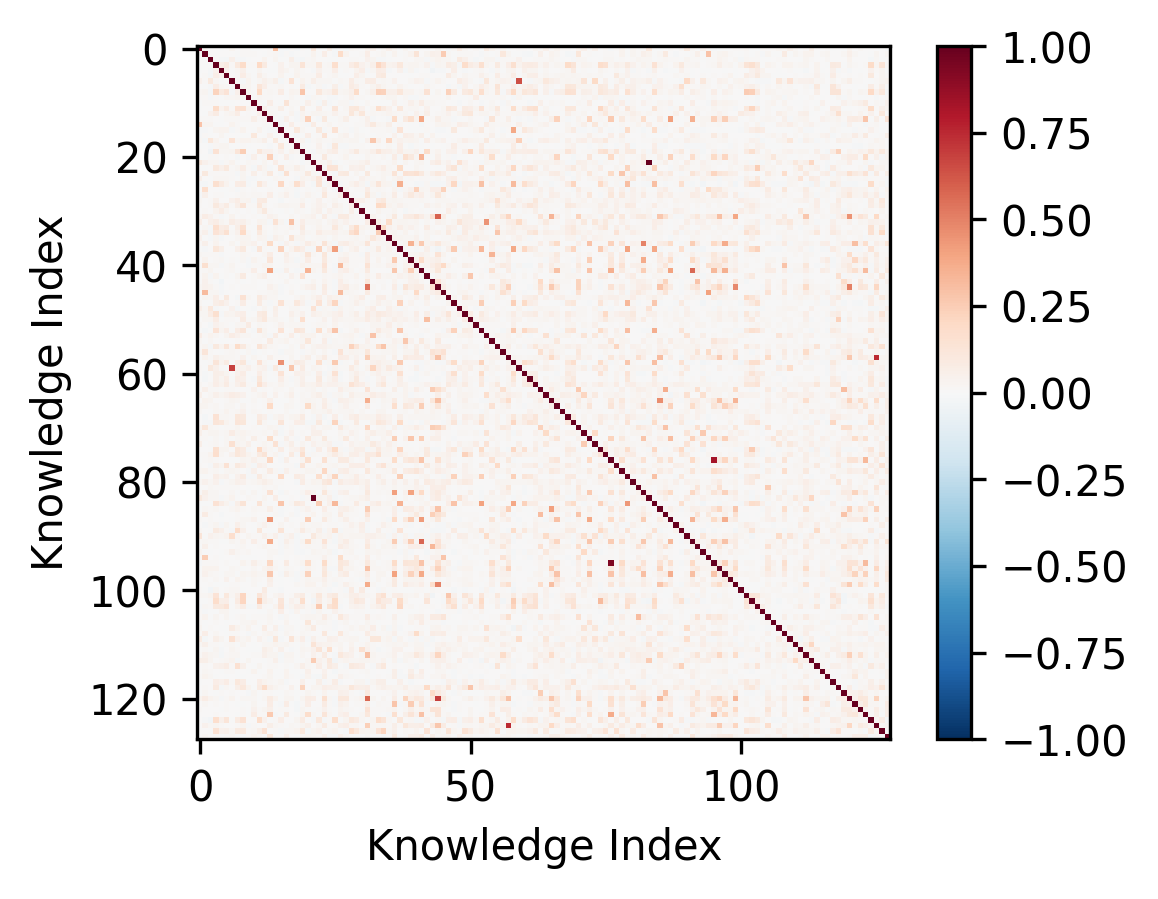}
    }
    \hfill
    \subfigure[Layer 12]{\includegraphics[width=0.22\textwidth]{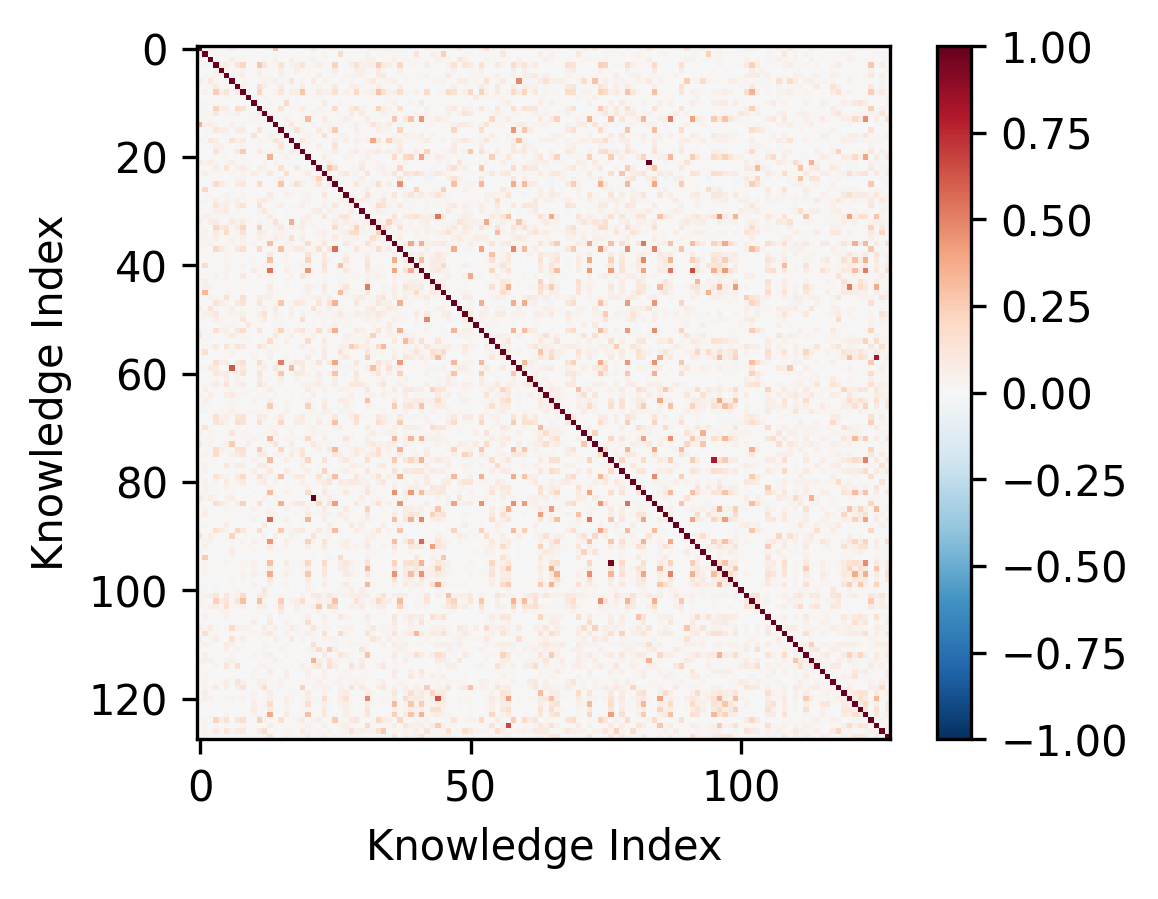}
    }
    \hfill
    \subfigure[Layer 13]{\includegraphics[width=0.22\textwidth]{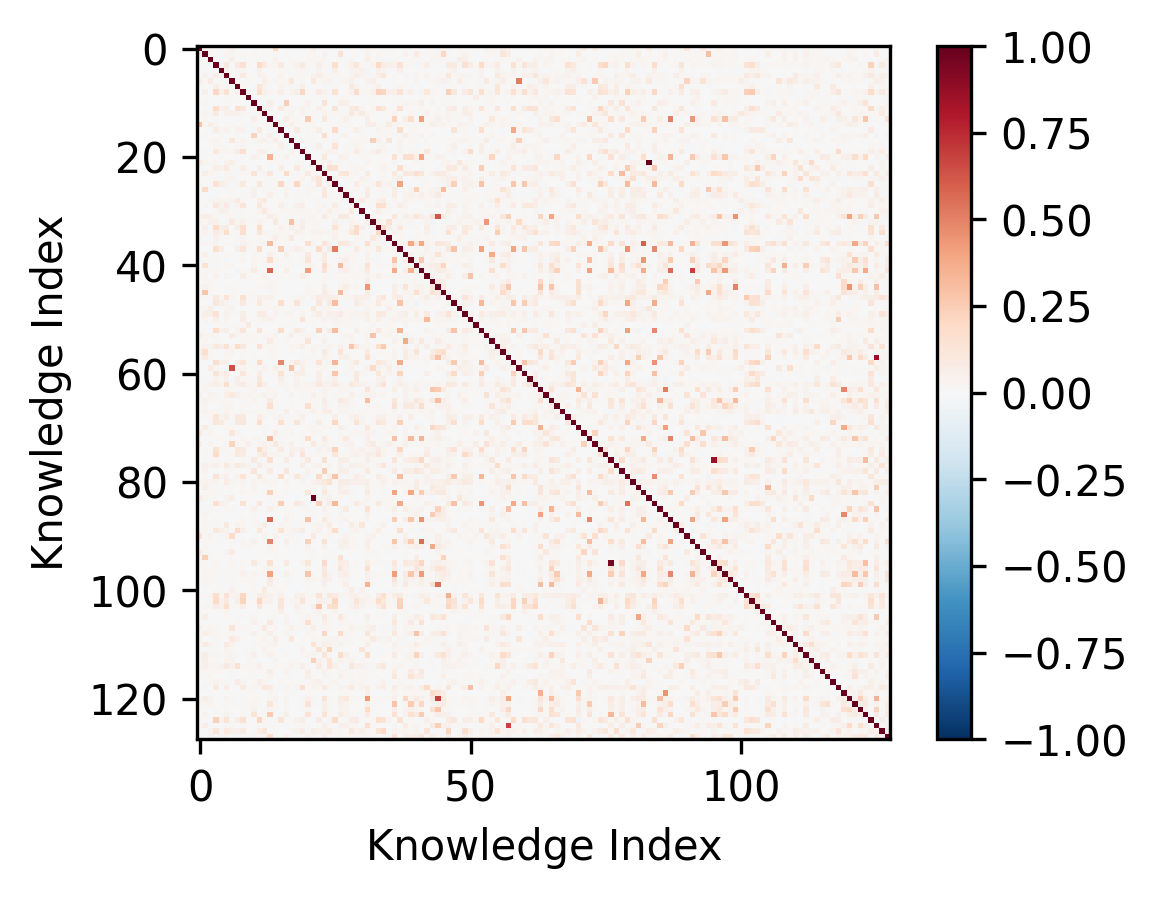}
    }
    \hfill
    \subfigure[Layer 14]{\includegraphics[width=0.22\textwidth]{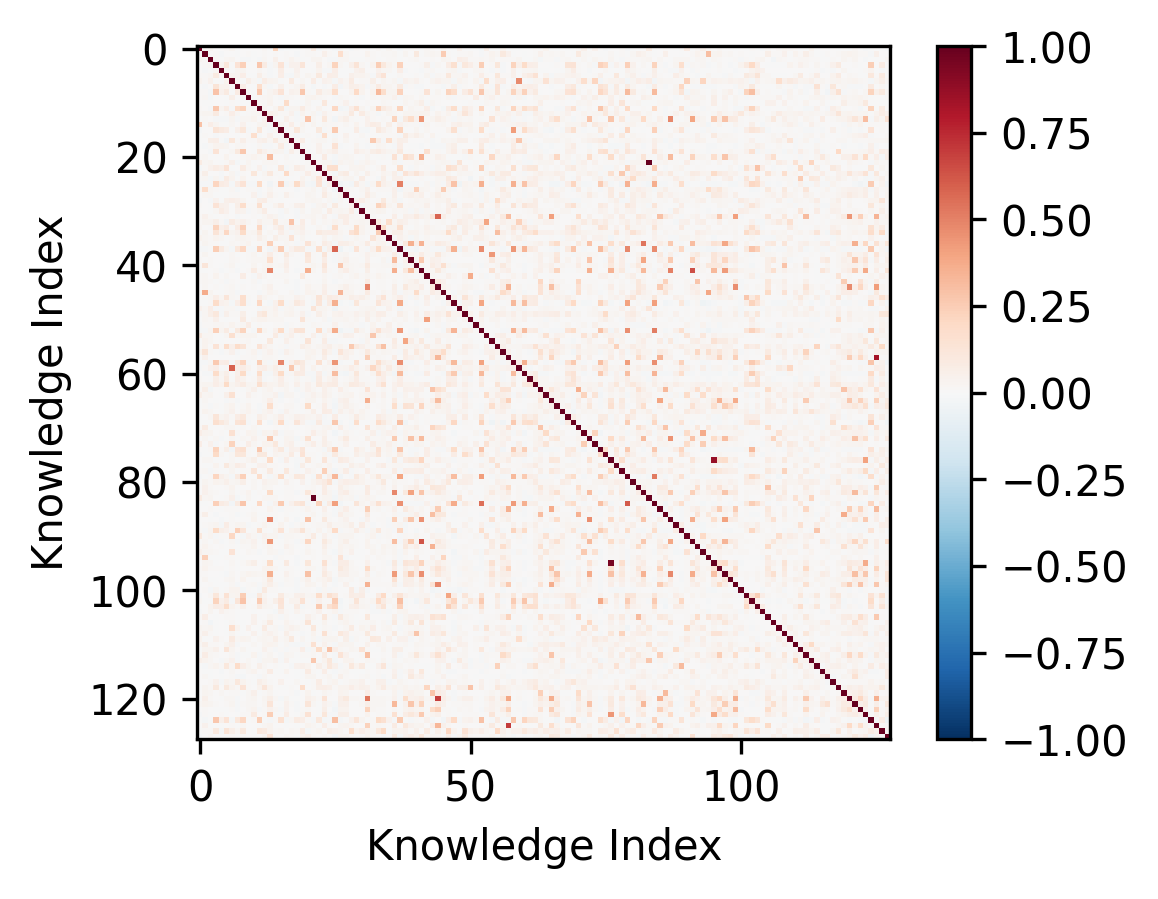}
    }
    \hfill
    \subfigure[Layer 15]{\includegraphics[width=0.22\textwidth]{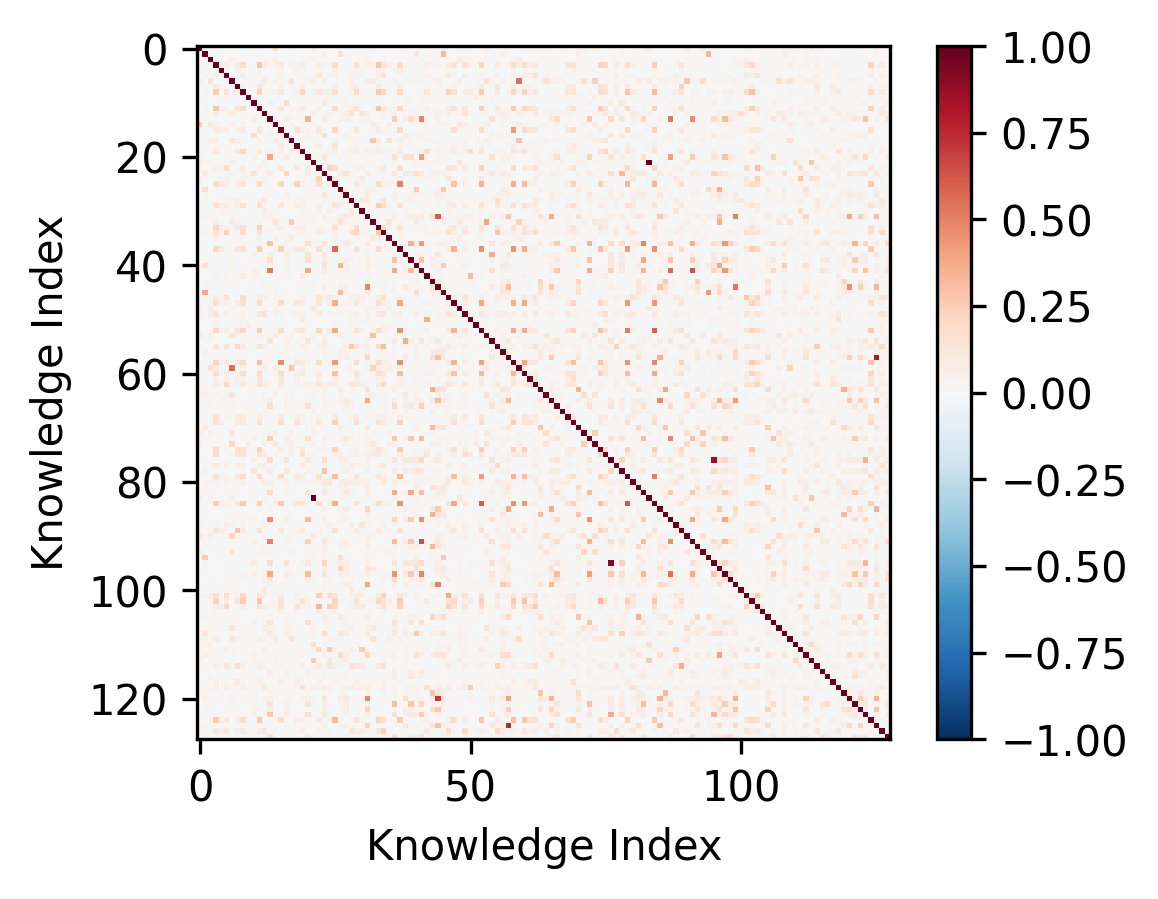}
    }
    \hfill
    \subfigure[Layer 16]{\includegraphics[width=0.22\textwidth]{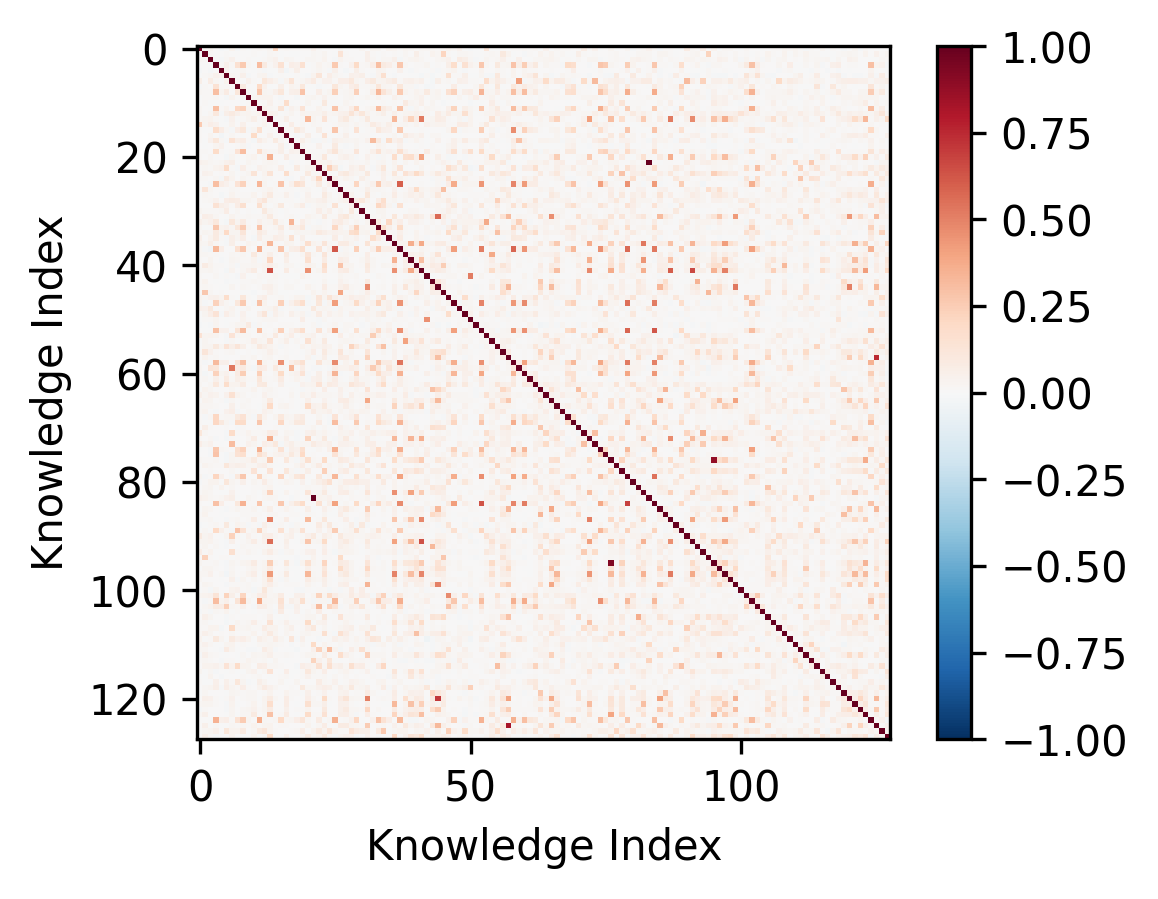}
    }
    \hfill
    \subfigure[Layer 17]{\includegraphics[width=0.22\textwidth]{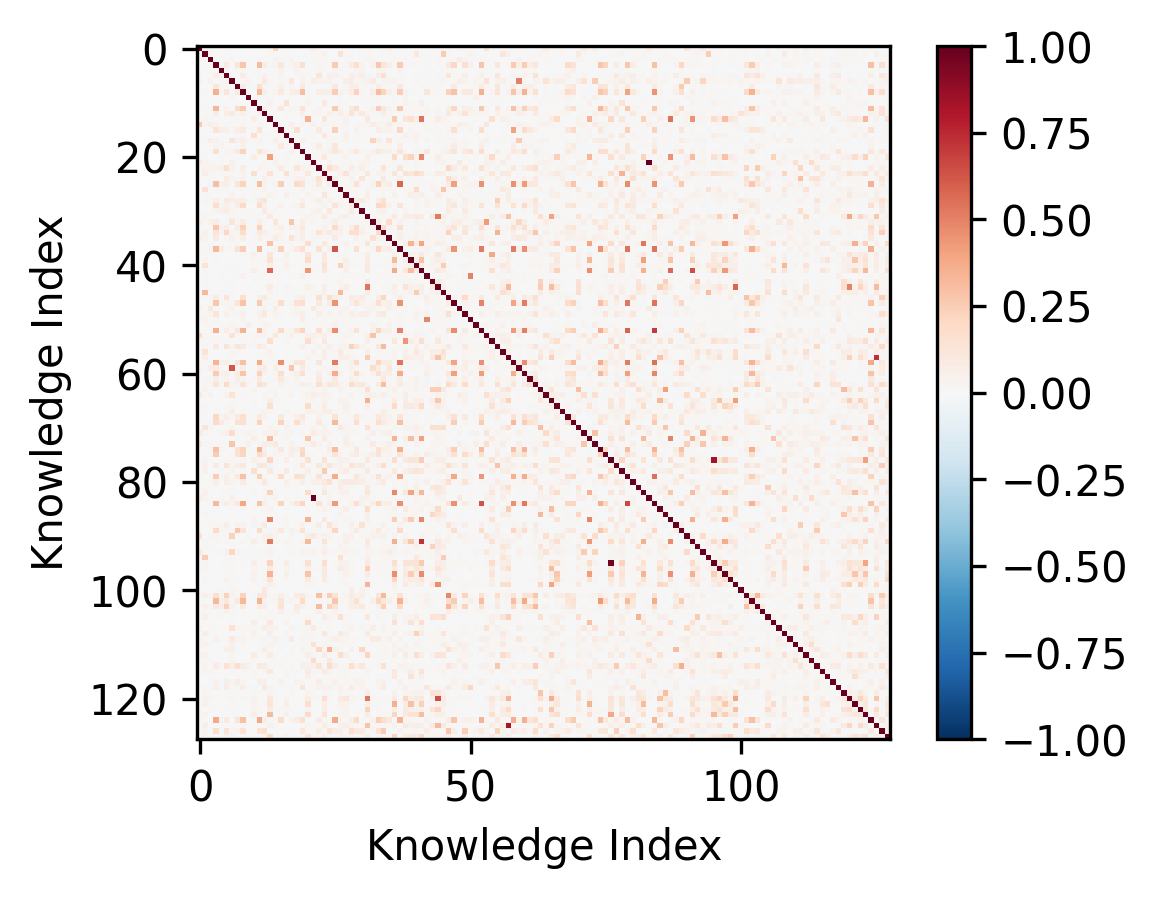}
    }
    \hfill
    \subfigure[Layer 18]{\includegraphics[width=0.22\textwidth]{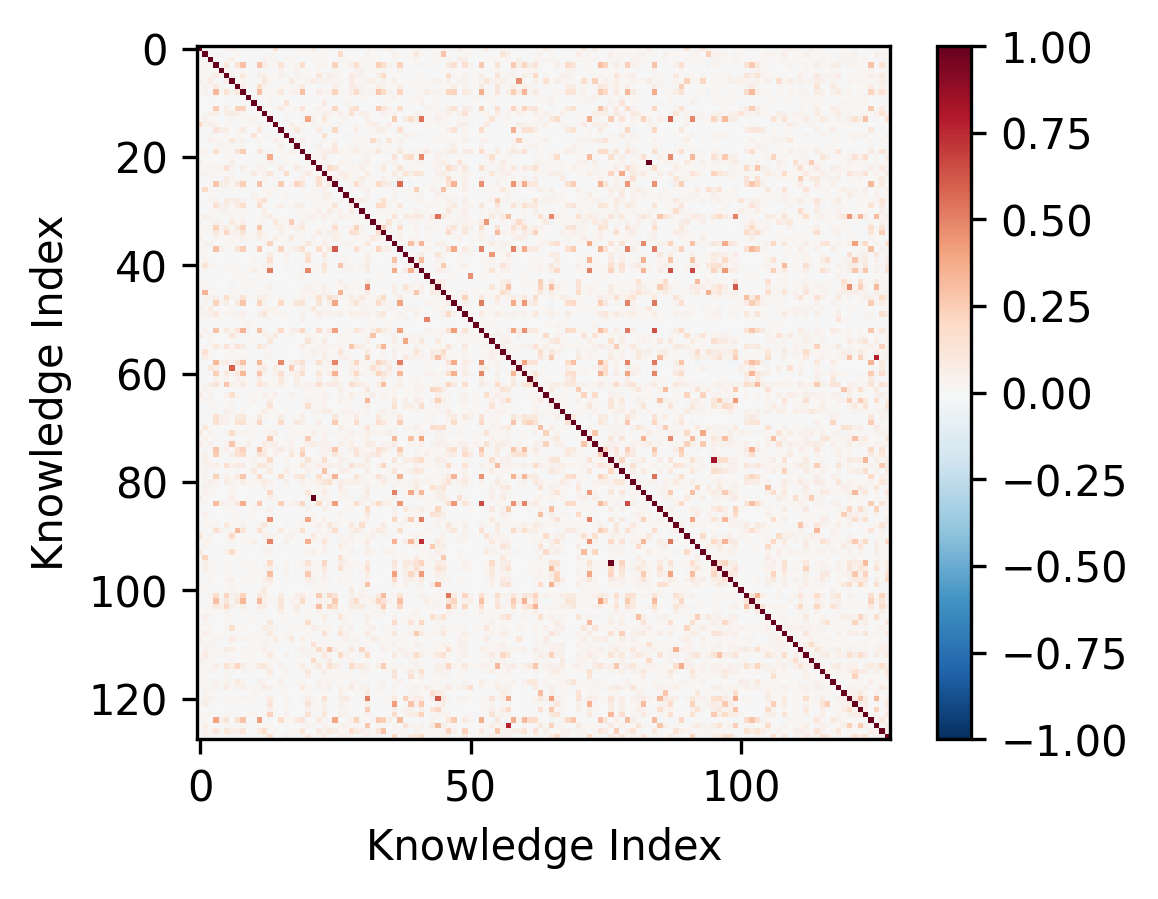}
    }
    \hfill
    \subfigure[Layer 19]{\includegraphics[width=0.22\textwidth]{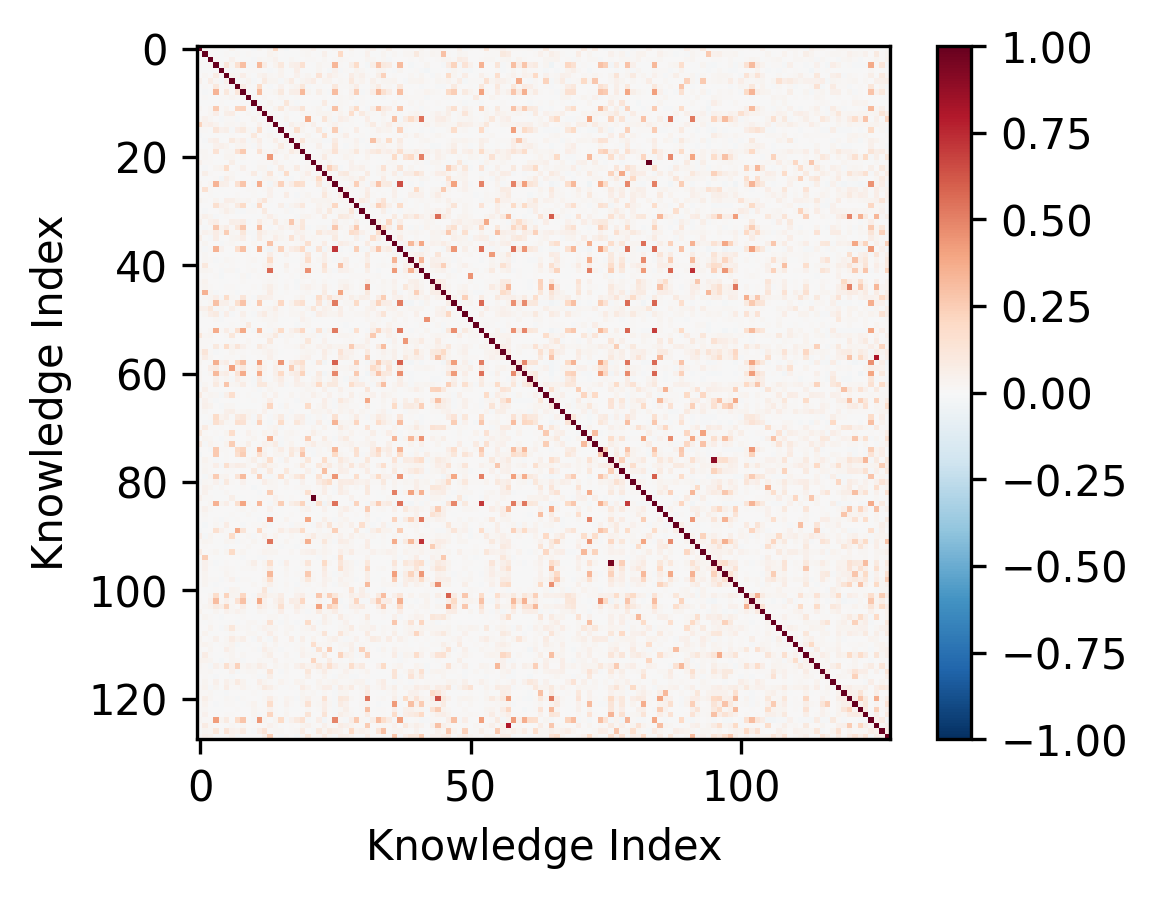}
    }
  \caption{In GPT-J-6B, the visualization of \(P\) matrices across layers (0-19).}
  \label{fig:superposition_visualization_gpt-j-6b-part1}
\end{figure*}

\begin{figure*}
    \centering
    \subfigure[Layer 20]{\includegraphics[width=0.22\textwidth]{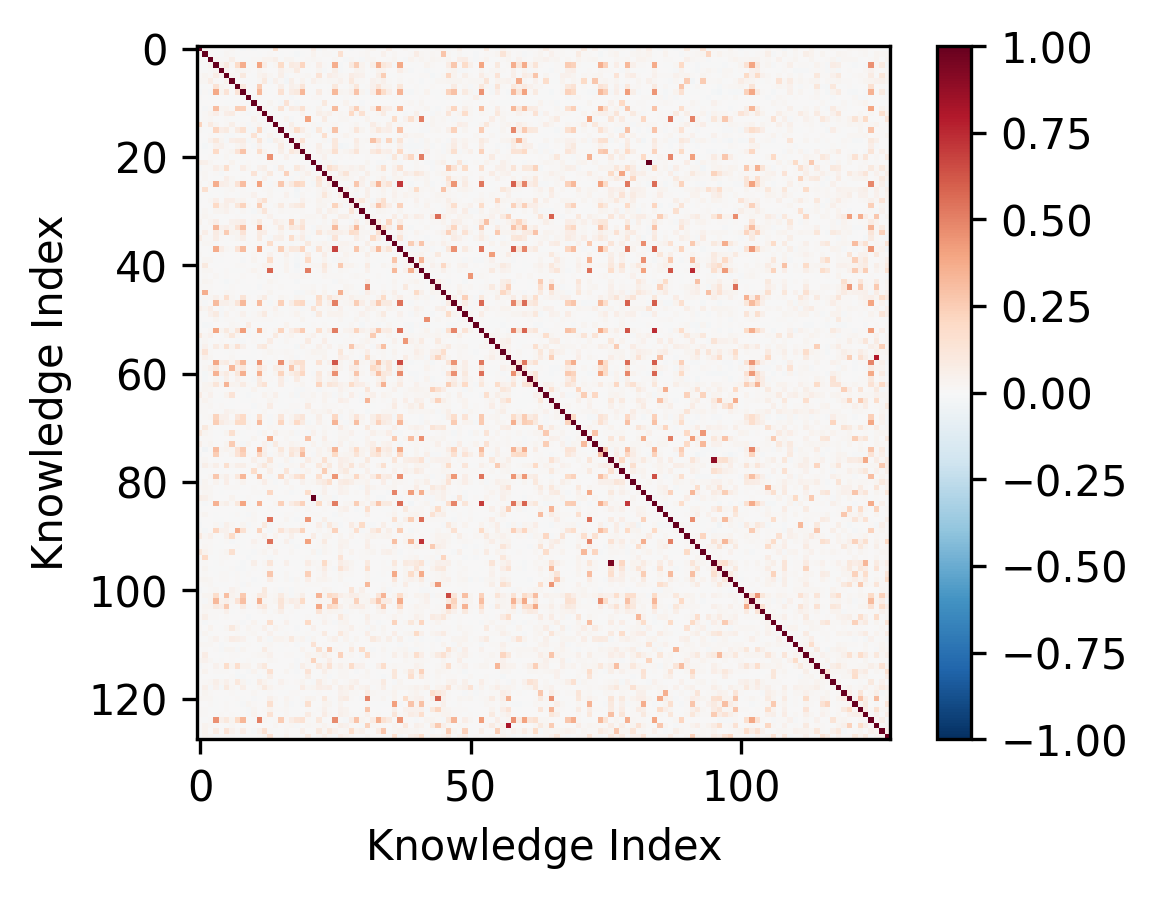}
    }
    \hfill
    \subfigure[Layer 21]{\includegraphics[width=0.22\textwidth]{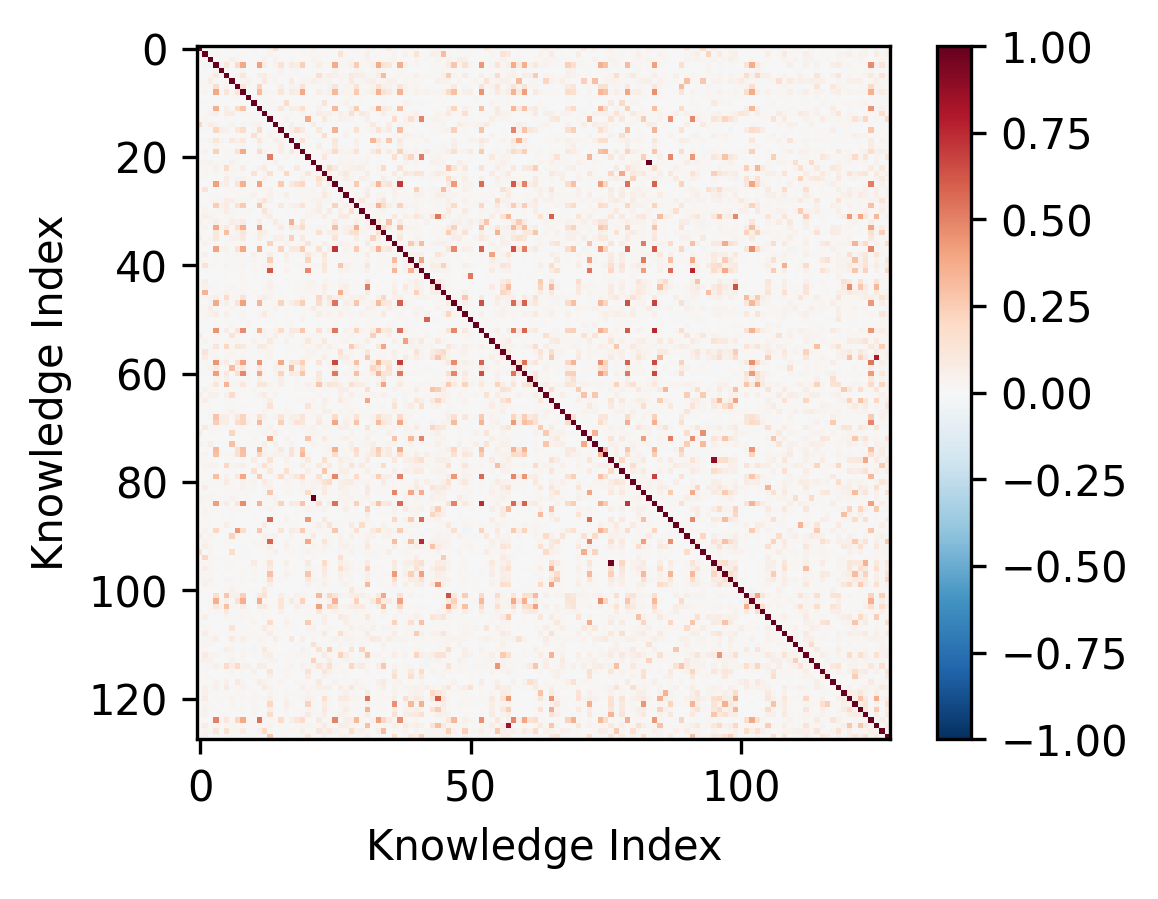}
    }
    \hfill
    \subfigure[Layer 22]{\includegraphics[width=0.22\textwidth]{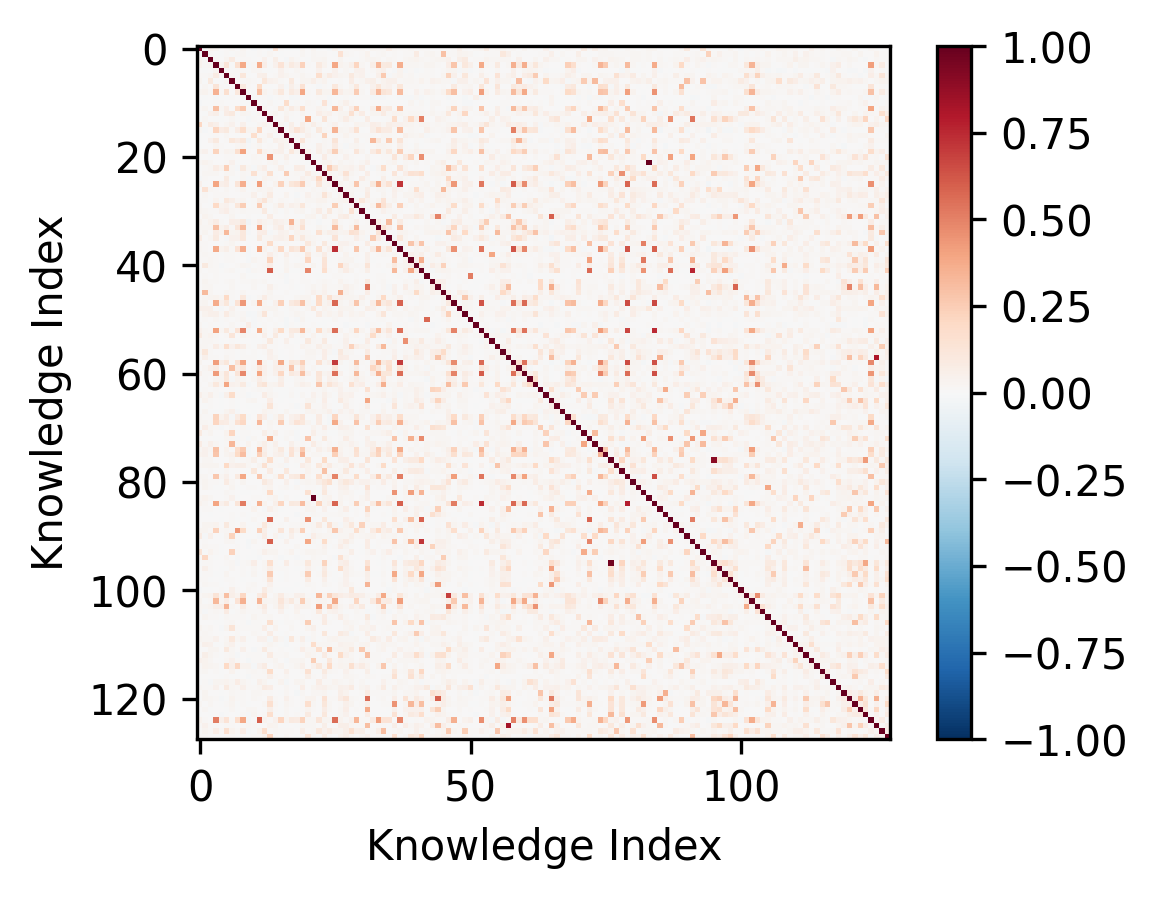}
    }
    \hfill
    \subfigure[Layer 23]{\includegraphics[width=0.22\textwidth]{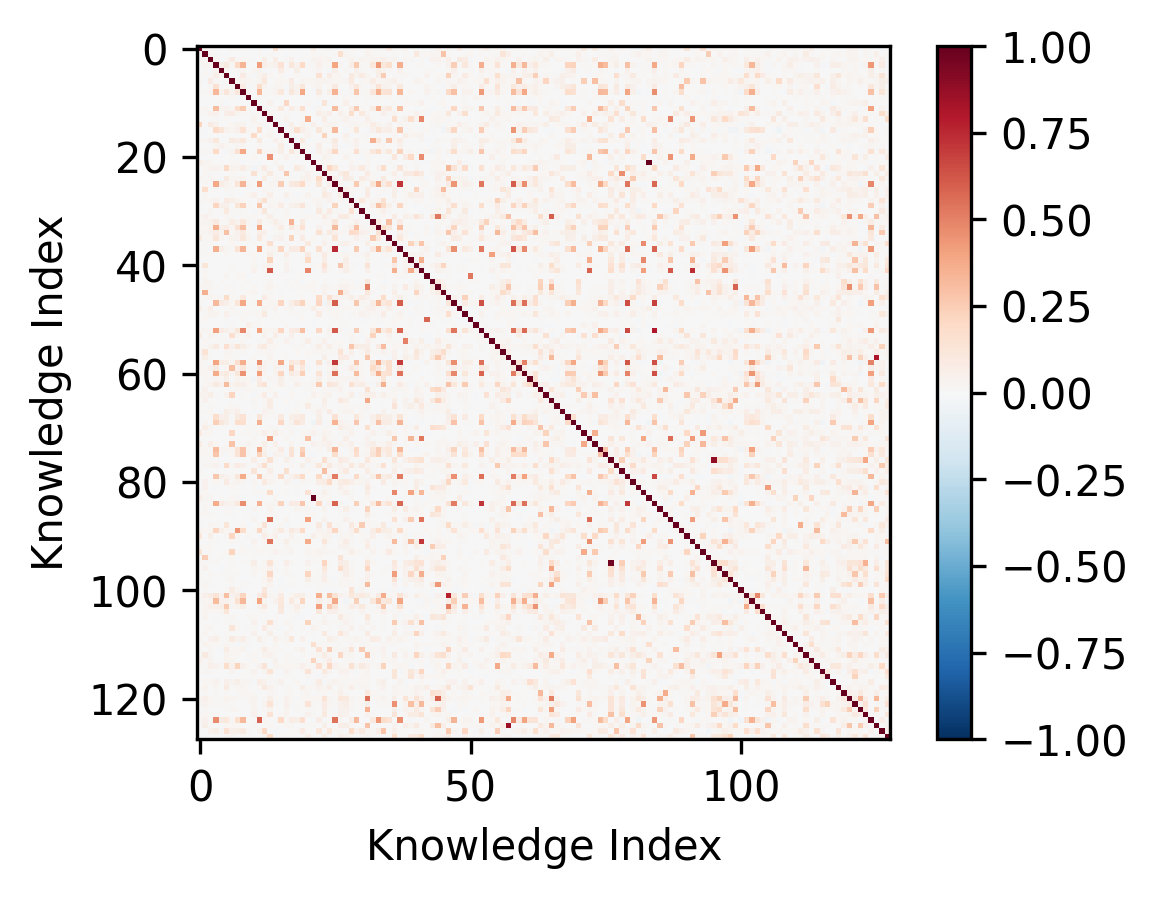}
    }
    \hfill
    \subfigure[Layer 24]{\includegraphics[width=0.22\textwidth]{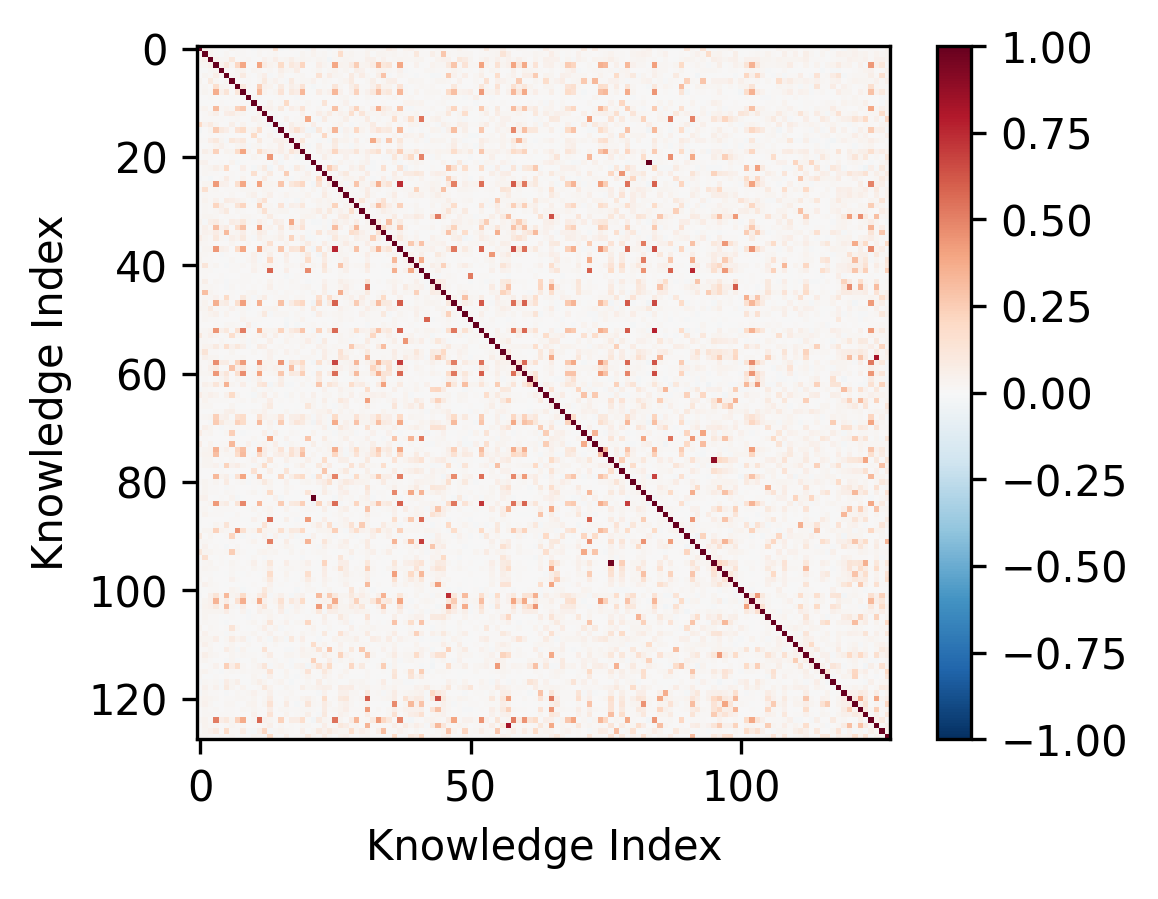}
    }
    \hfill
    \subfigure[Layer 25]{\includegraphics[width=0.22\textwidth]{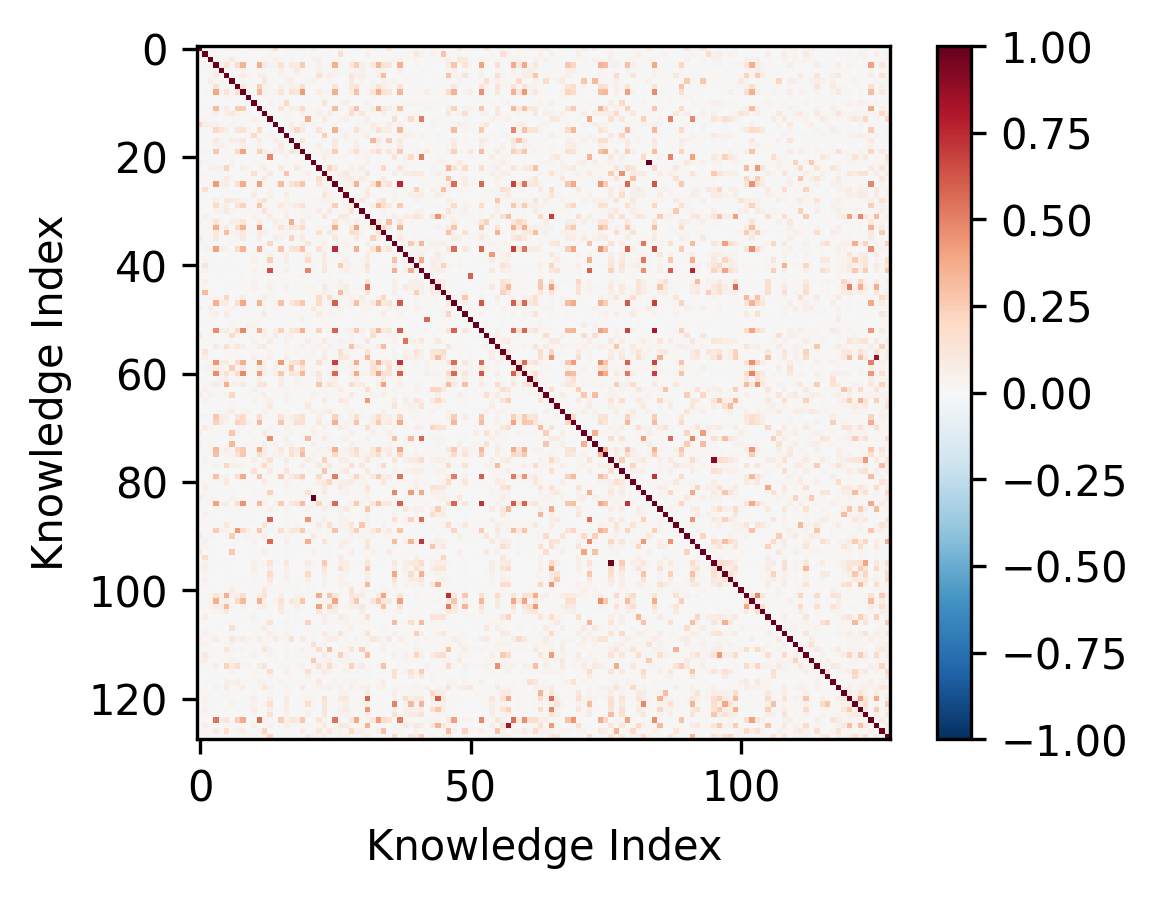}
    }
    \hfill
    \subfigure[Layer 26]{\includegraphics[width=0.22\textwidth]{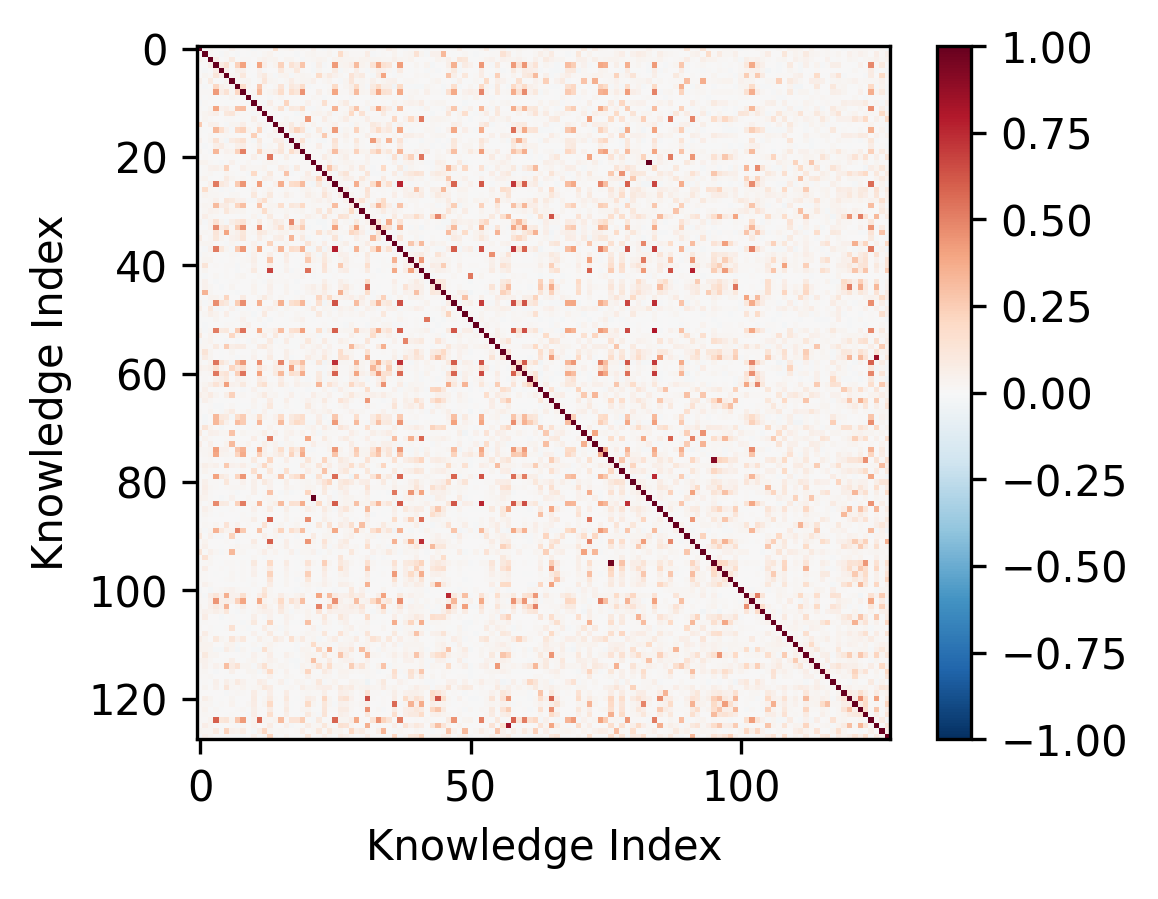}
    }
    \hfill
    \subfigure[Layer 27]{\includegraphics[width=0.22\textwidth]{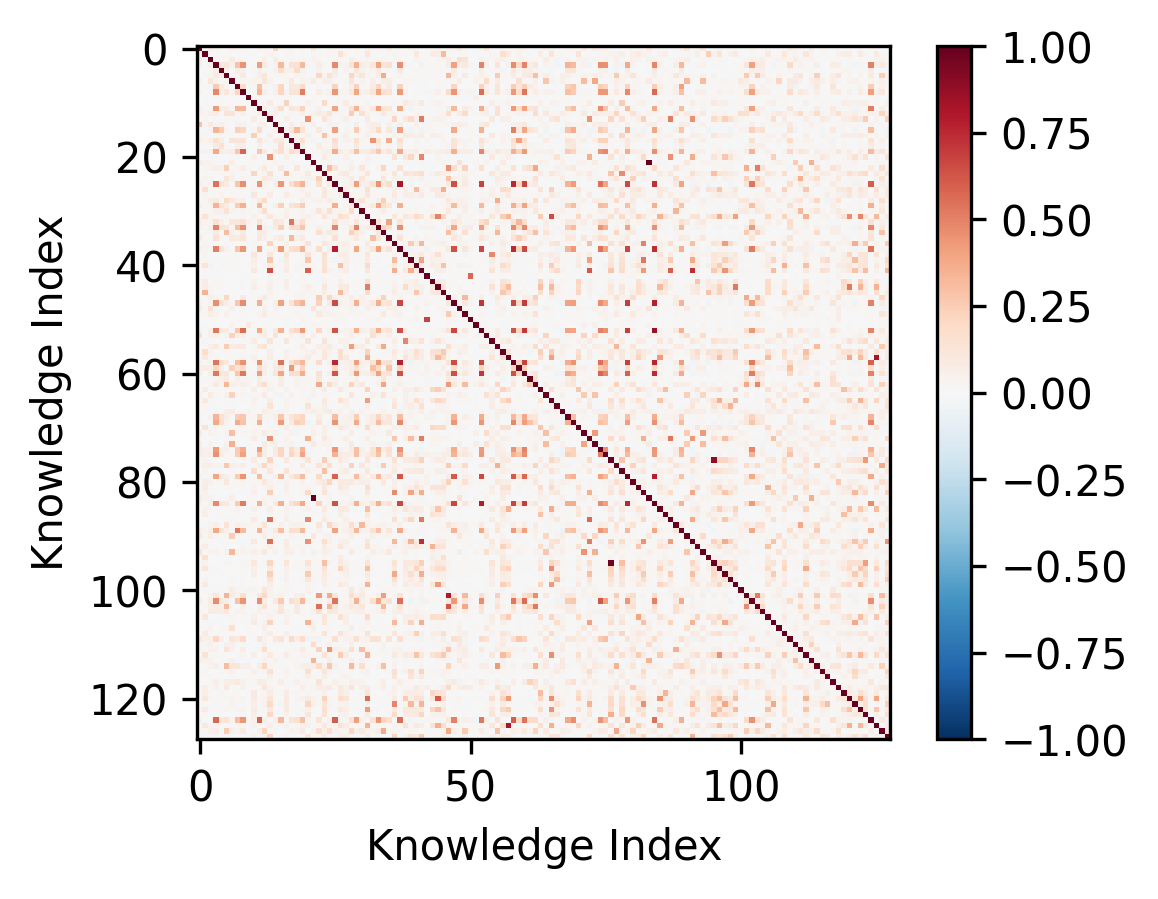}
    }
  \caption{In GPT-J-6B, the visualization of \(P\) matrices across layers (20-27).}
  \label{fig:superposition_visualization_gpt-j-6b-part2}
\end{figure*}

In this section, we will present visualizations of the \( P \) matrices across layers for GPT2-Small, GPT2-Medium, GPT2-Large, GPT-J-6B, Pythia-1B, Pythia-2.8B, Pythia-6.9B, Llama2-7B, Llama2-13B, Llama3-8B and Llama3.1-8B. These will be displayed as heatmaps to provide readers with a more direct understanding of knowledge superposition, as illustrated in Figures~\ref{fig:superposition_visualization_gpt2-small}-~\ref{fig:superposition_visualization_gpt-j-6b-part2}. Some experiments have been deleted due to Arxiv upload restrictions. If you really need these experimental results, please contact us.

\section{E. KDE of \(P\) Matrices for All Layers}\label{appendix:E. KDE of Matrix for All Layers}

\begin{figure*}
    \centering
    \subfigure[Layer 0]{\includegraphics[width=0.22\textwidth]{fig/gpt2-small/p_matrix/known/KDE/superposition_for_layer_0.png}
    }
    \hfill
    \subfigure[Layer 1]{\includegraphics[width=0.22\textwidth]{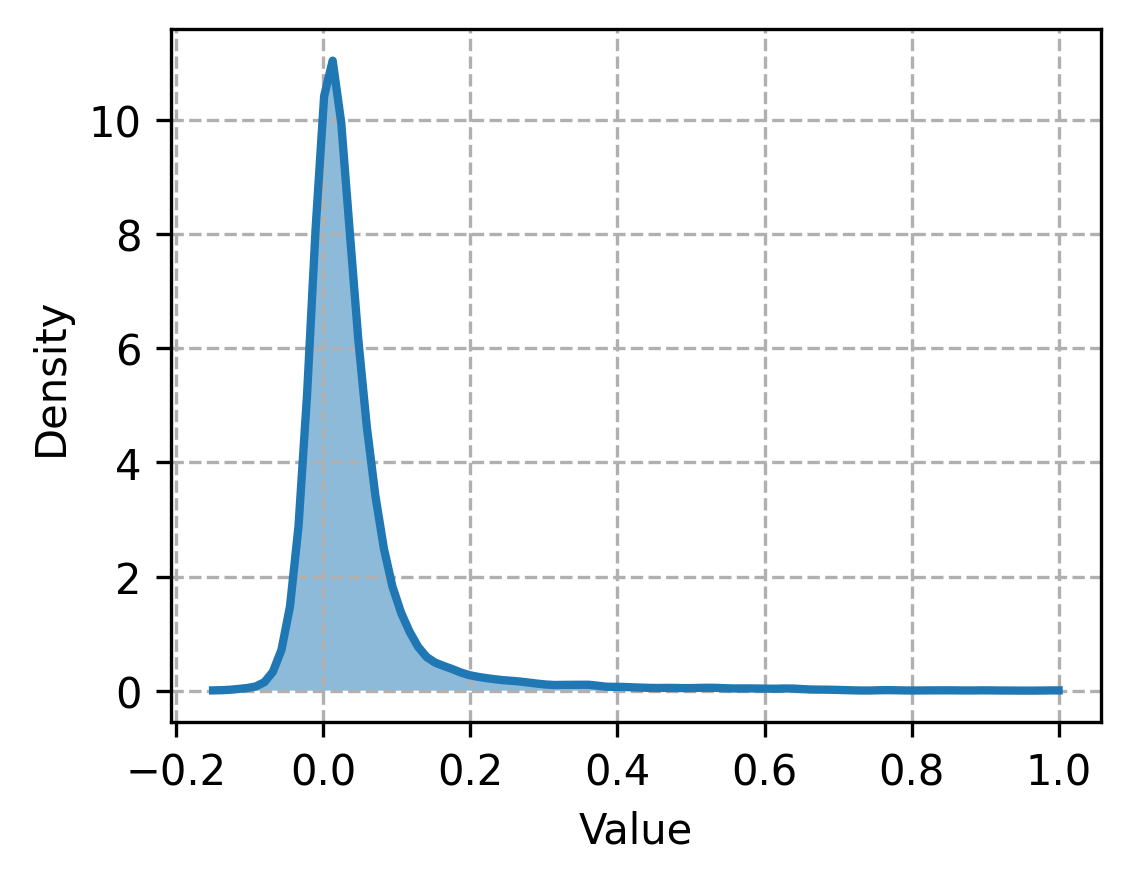}
    }
    \hfill
    \subfigure[Layer 2]{\includegraphics[width=0.22\textwidth]{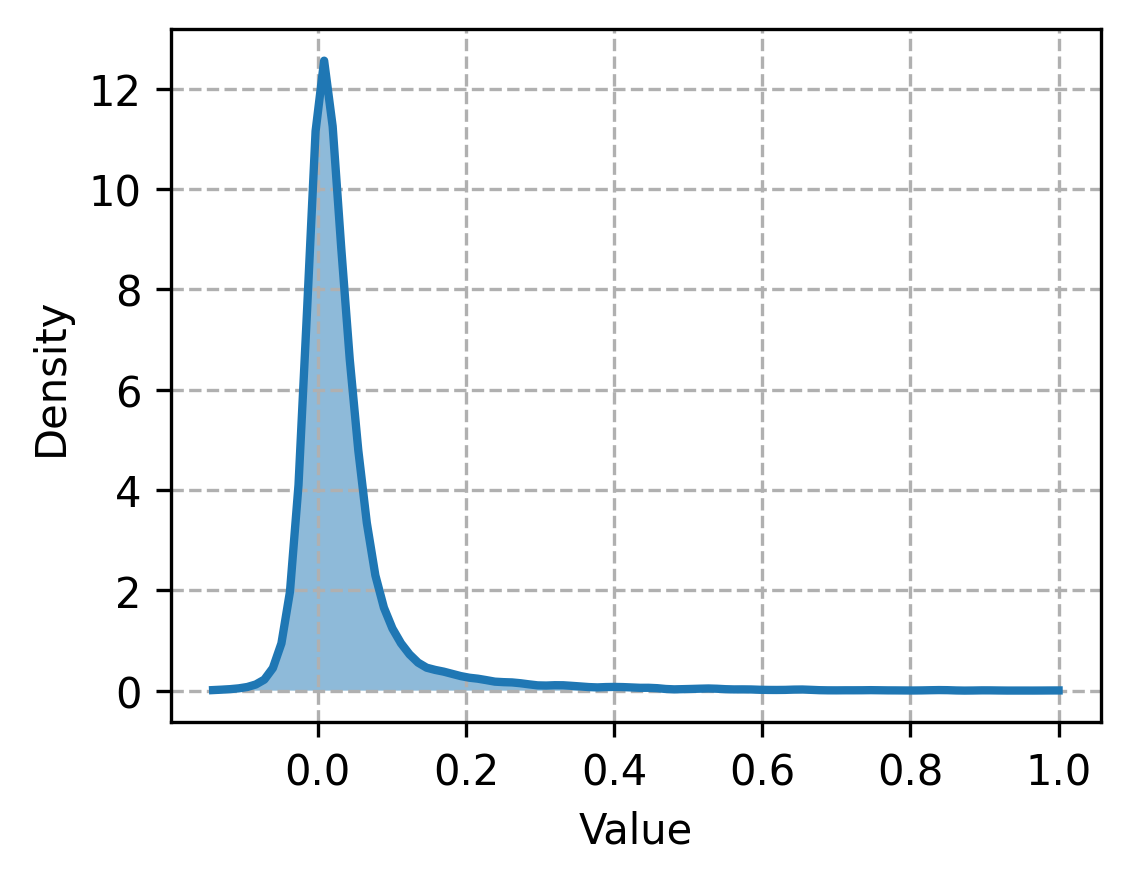}
    }
    \hfill
    \subfigure[Layer 3]{\includegraphics[width=0.22\textwidth]{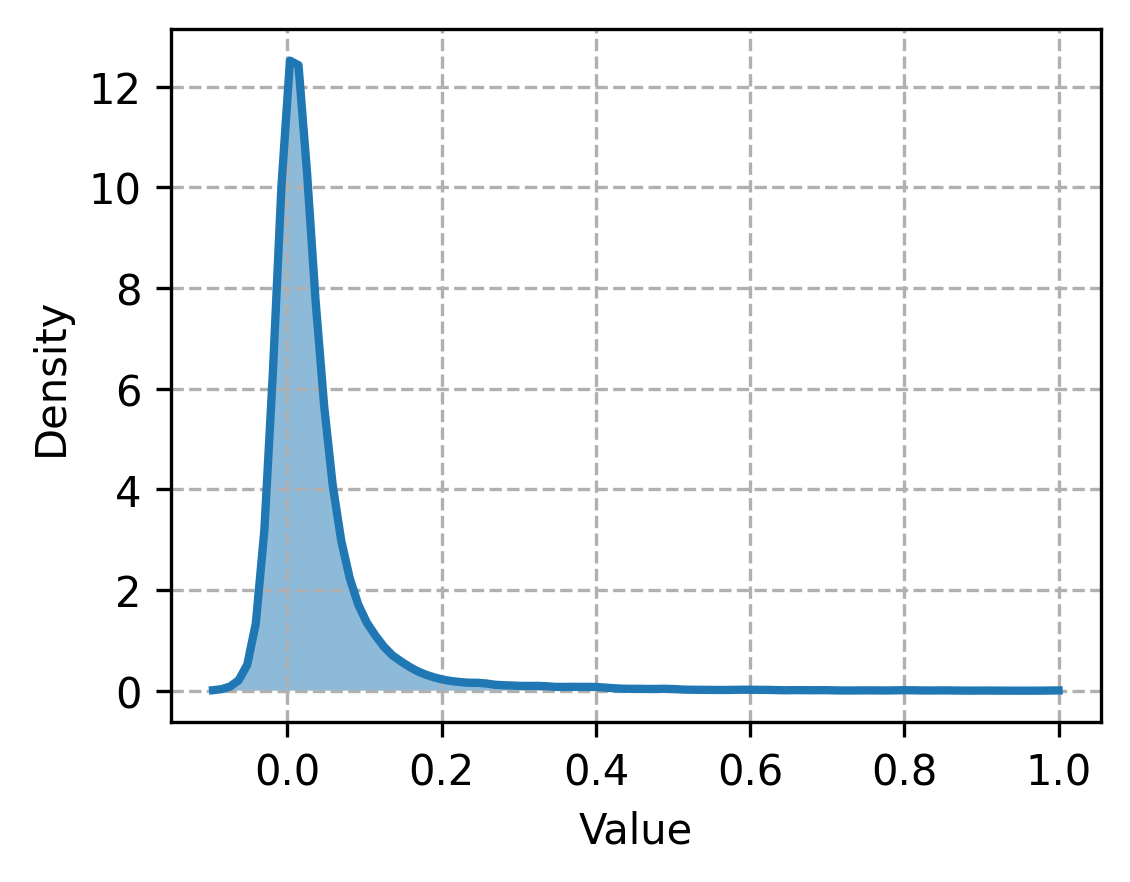}
    }
    \hfill
    \subfigure[Layer 4]{\includegraphics[width=0.22\textwidth]{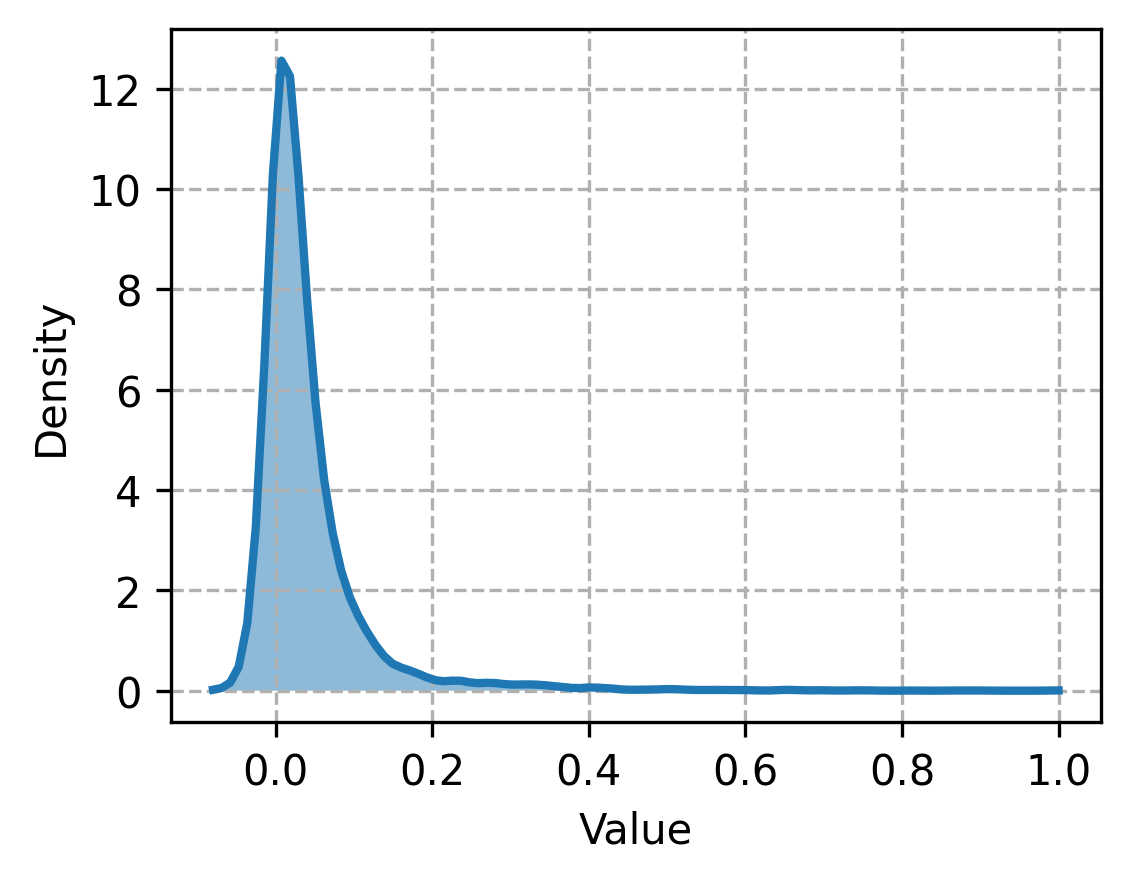}
    }
    \hfill
    \subfigure[Layer 5]{\includegraphics[width=0.22\textwidth]{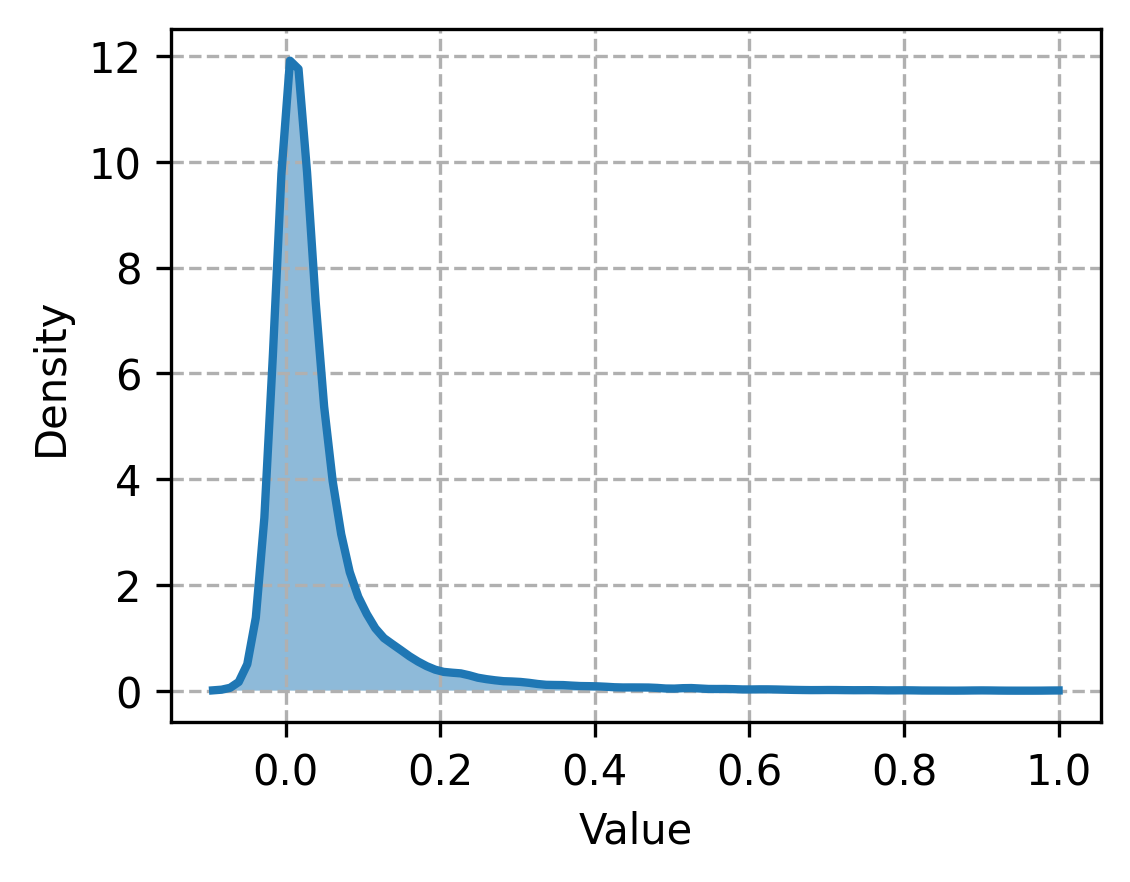}
    }
    \hfill
    \subfigure[Layer 6]{\includegraphics[width=0.22\textwidth]{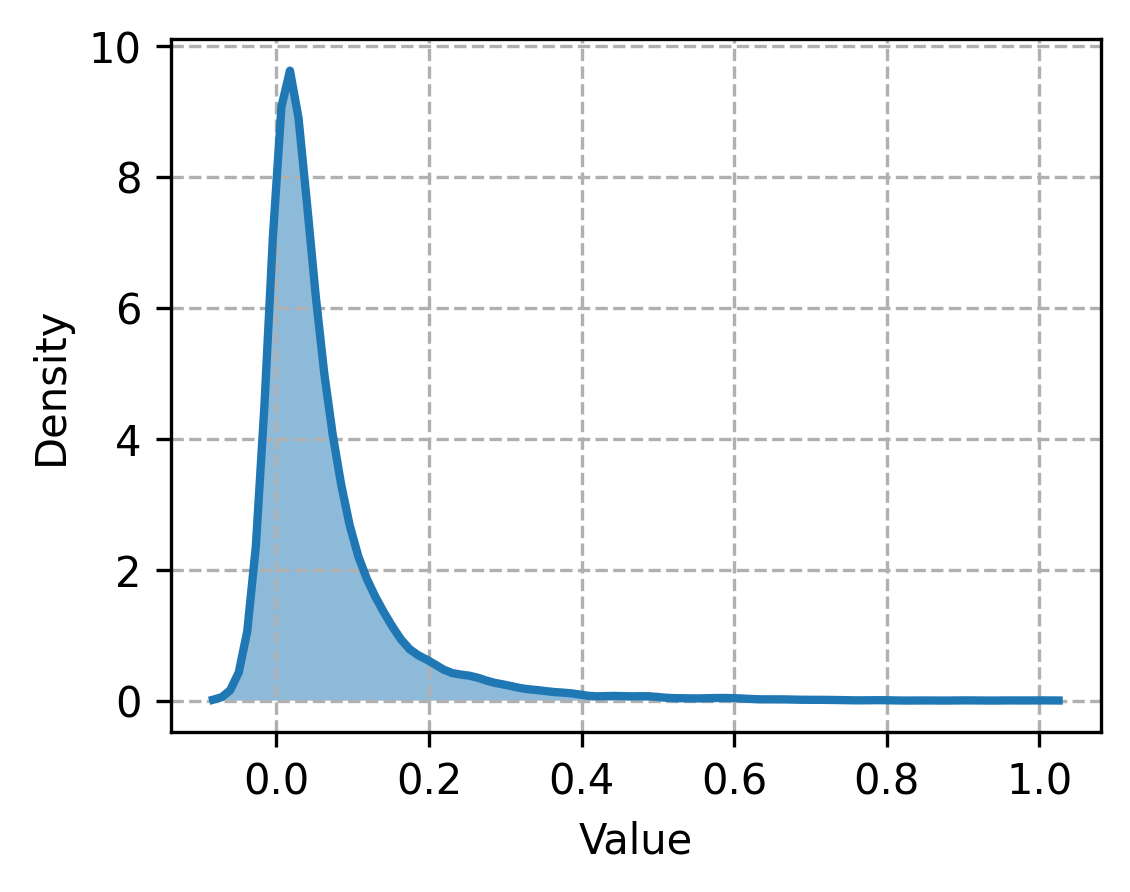}
    }
    \hfill
    \subfigure[Layer 7]{\includegraphics[width=0.22\textwidth]{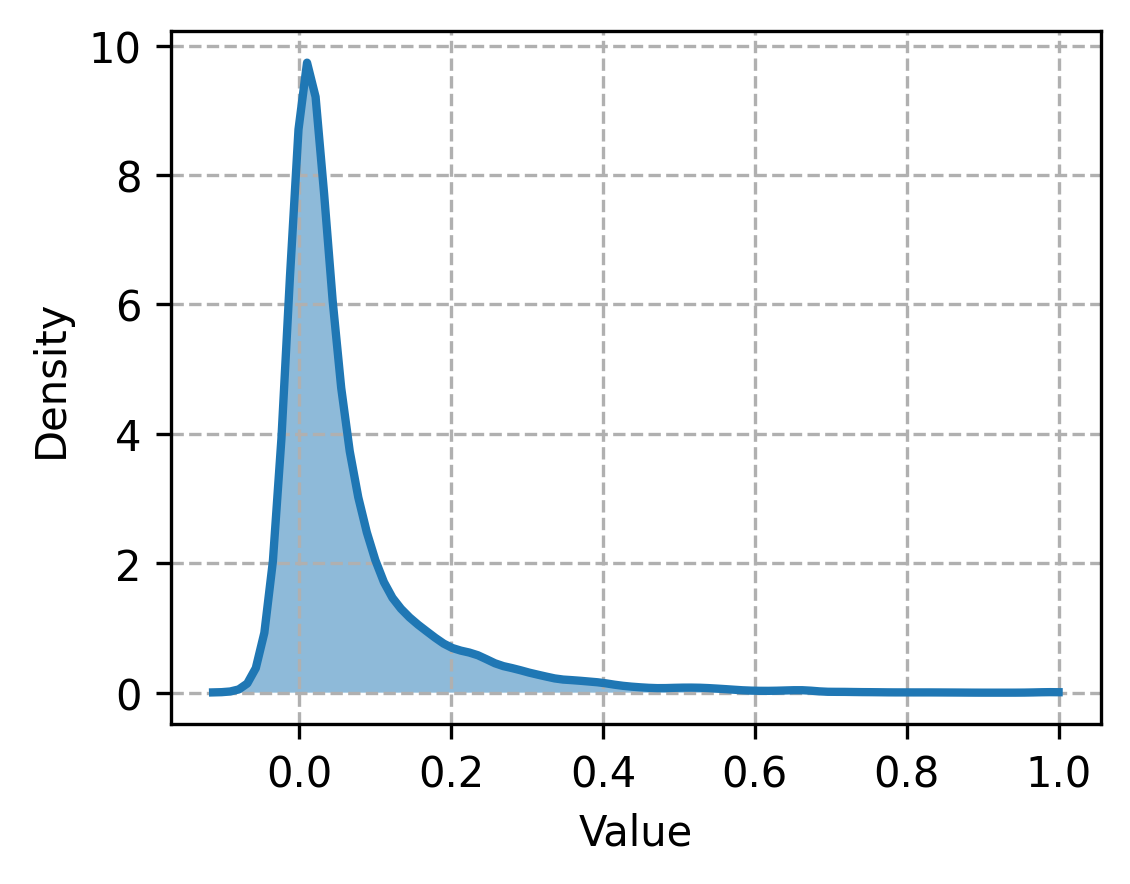}
    }
    \hfill
    \subfigure[Layer 8]{\includegraphics[width=0.22\textwidth]{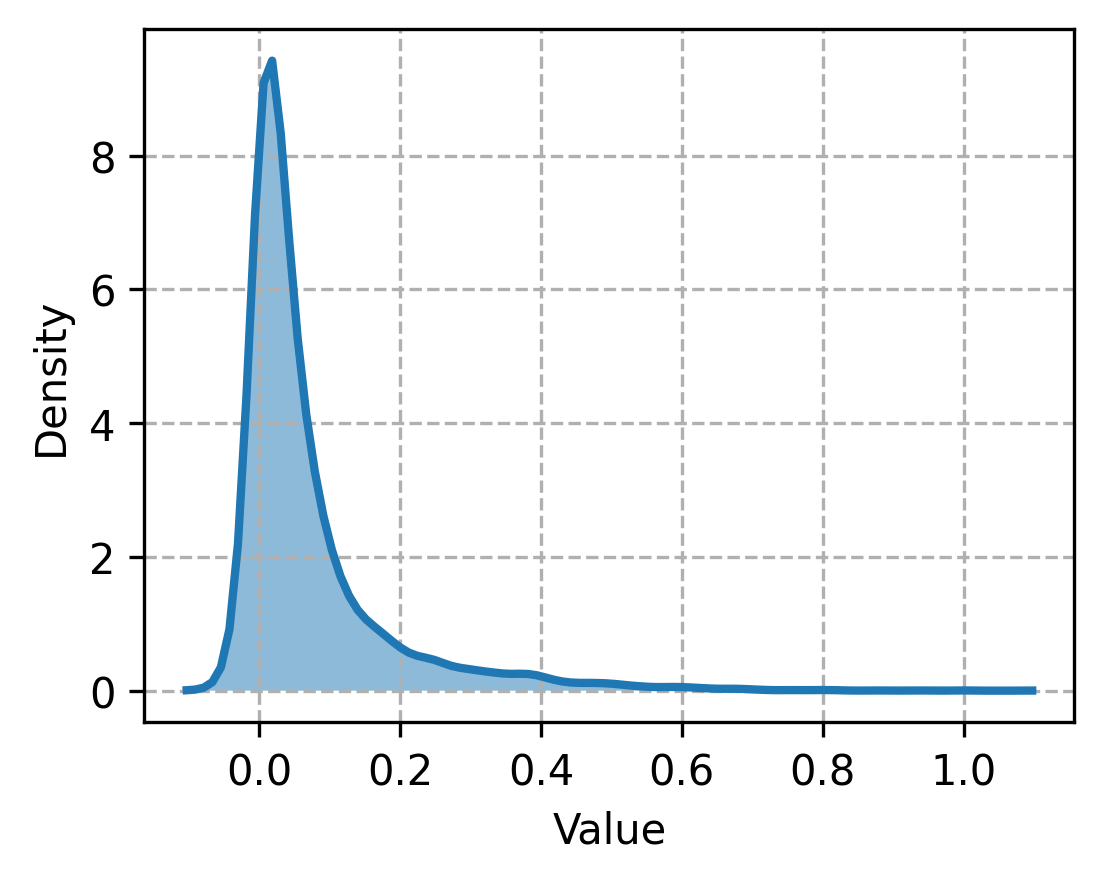}
    }
    \hfill
    \subfigure[Layer 9]{\includegraphics[width=0.22\textwidth]{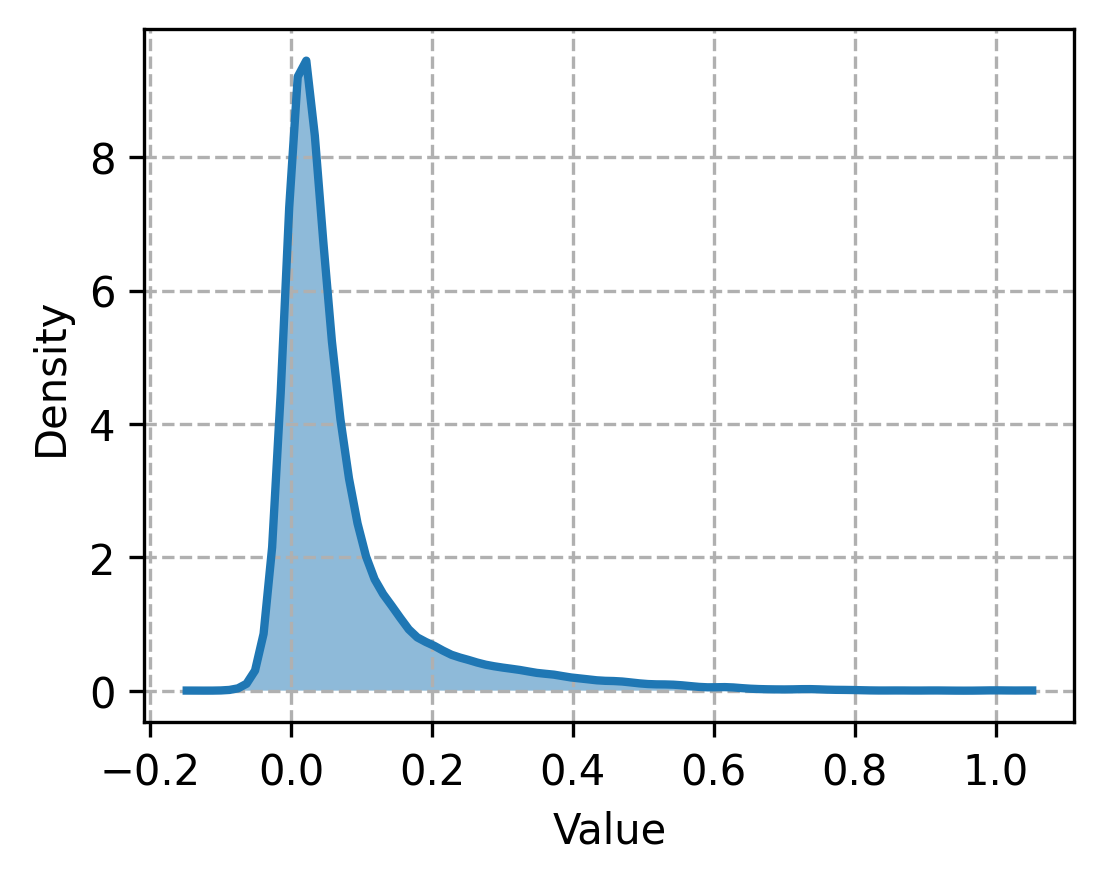}
    }
    \hfill
    \subfigure[Layer 10]{\includegraphics[width=0.22\textwidth]{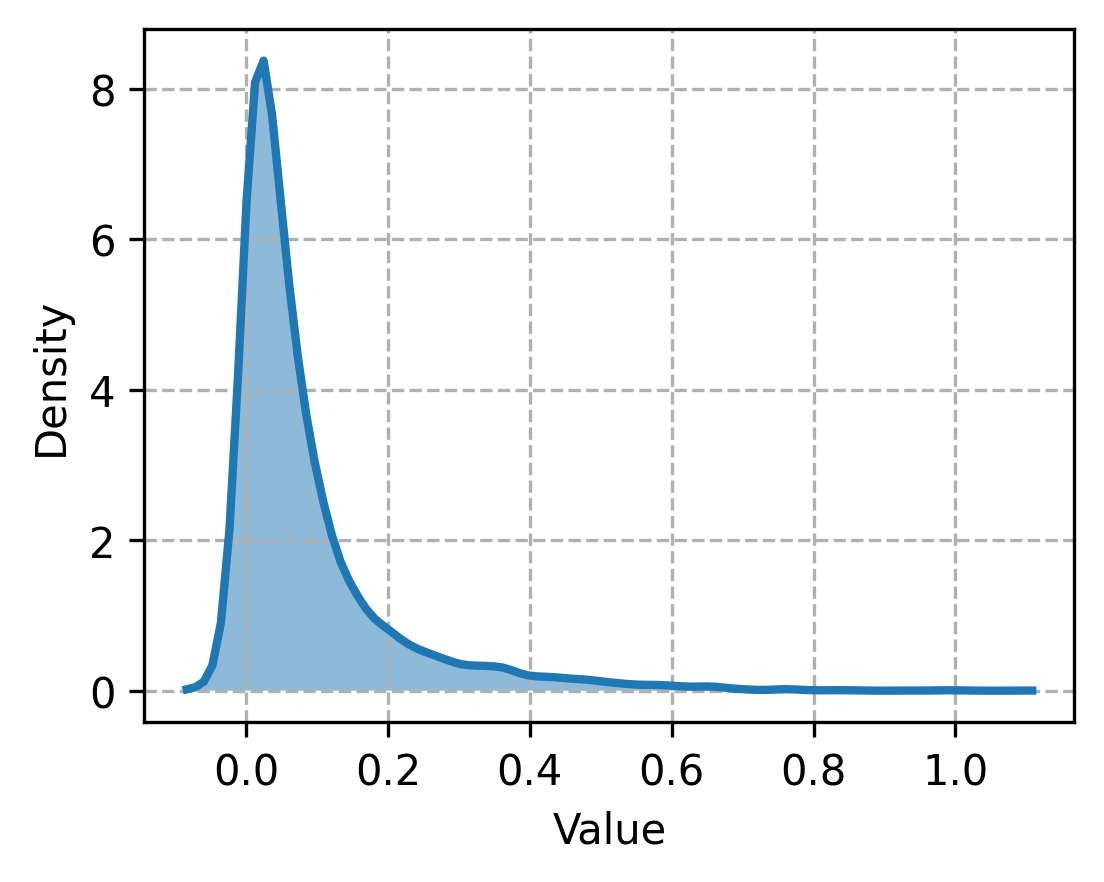}
    }
    \hfill
    \subfigure[Layer 11]{\includegraphics[width=0.22\textwidth]{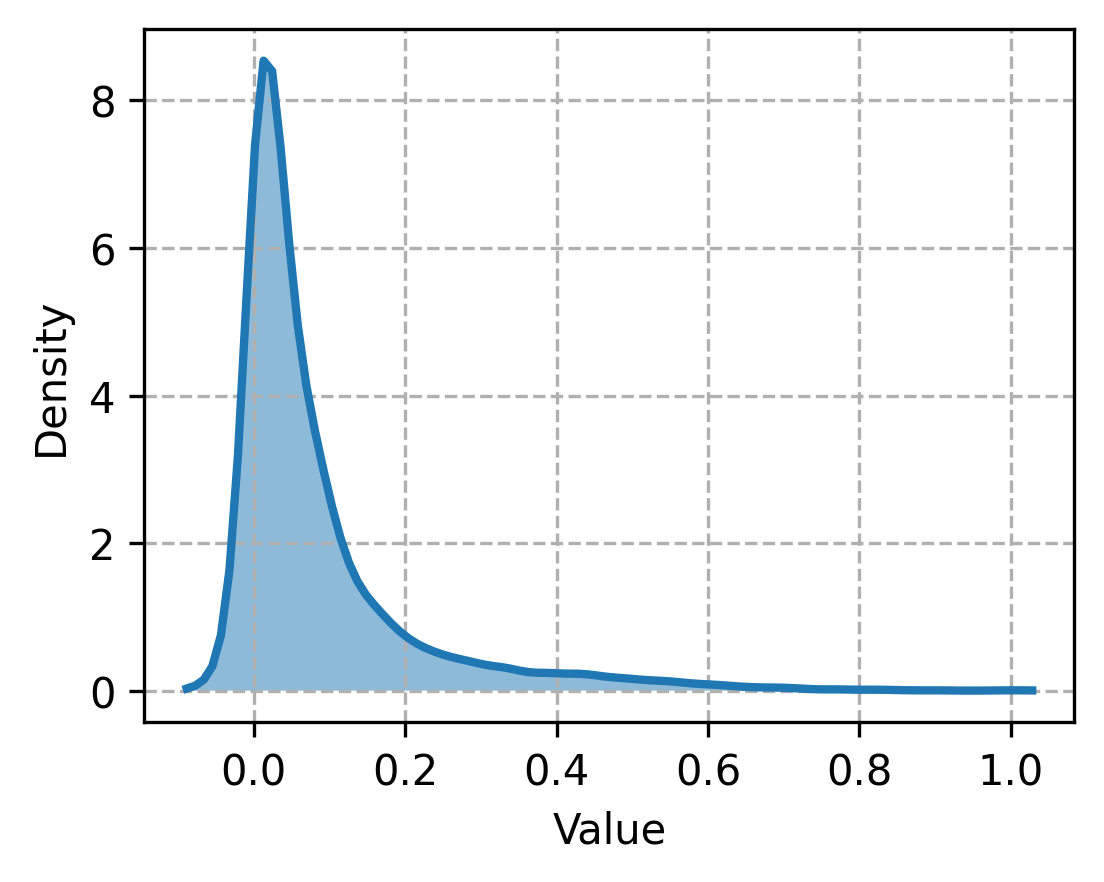}
    }
  \caption{In GPT2-Small, the KDE of elements in \(P\) matrices across layers.}
  \label{fig:superposition_KDE_gpt2-small}
\end{figure*}

\begin{figure*}
    \centering
    \subfigure[Layer 0]{\includegraphics[width=0.22\textwidth]{fig/gpt2-medium/p_matrix/known/KDE/superposition_for_layer_0.png}
    }
    \hfill
    \subfigure[Layer 1]{\includegraphics[width=0.22\textwidth]{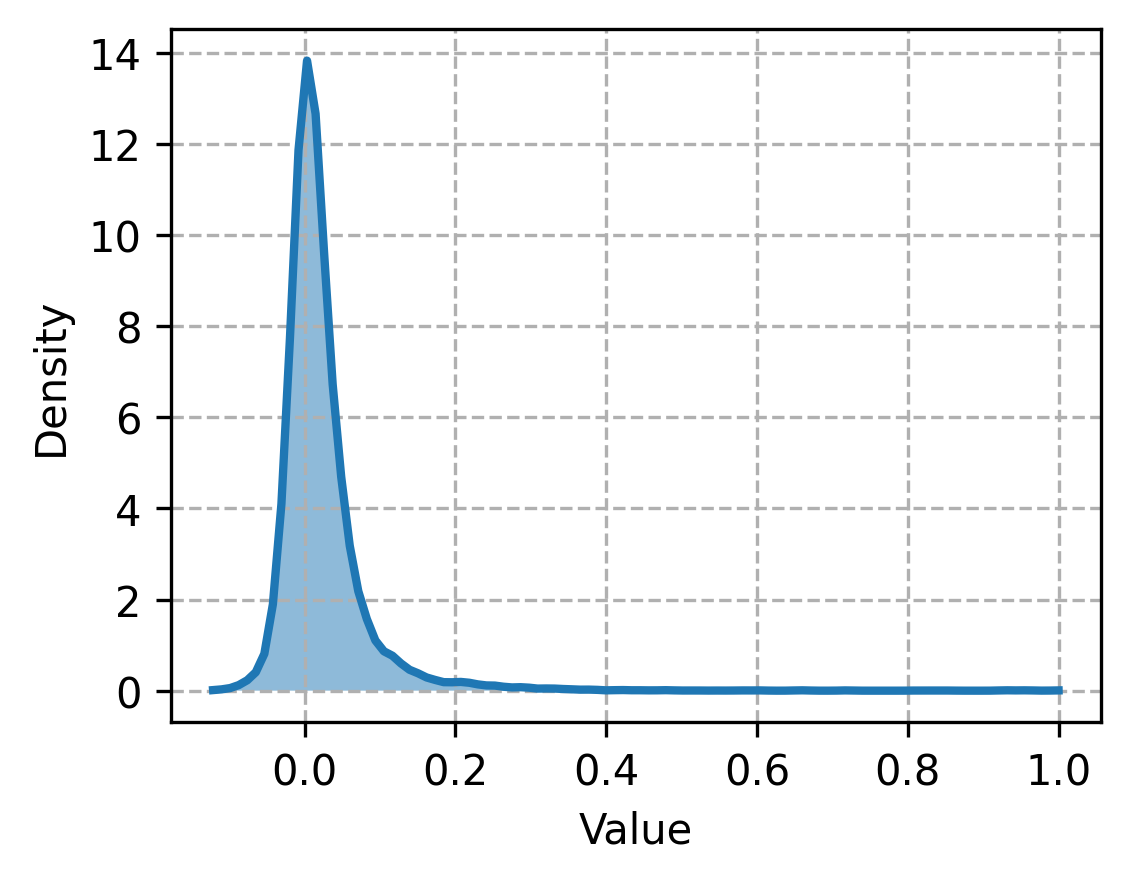}
    }
    \hfill
    \subfigure[Layer 2]{\includegraphics[width=0.22\textwidth]{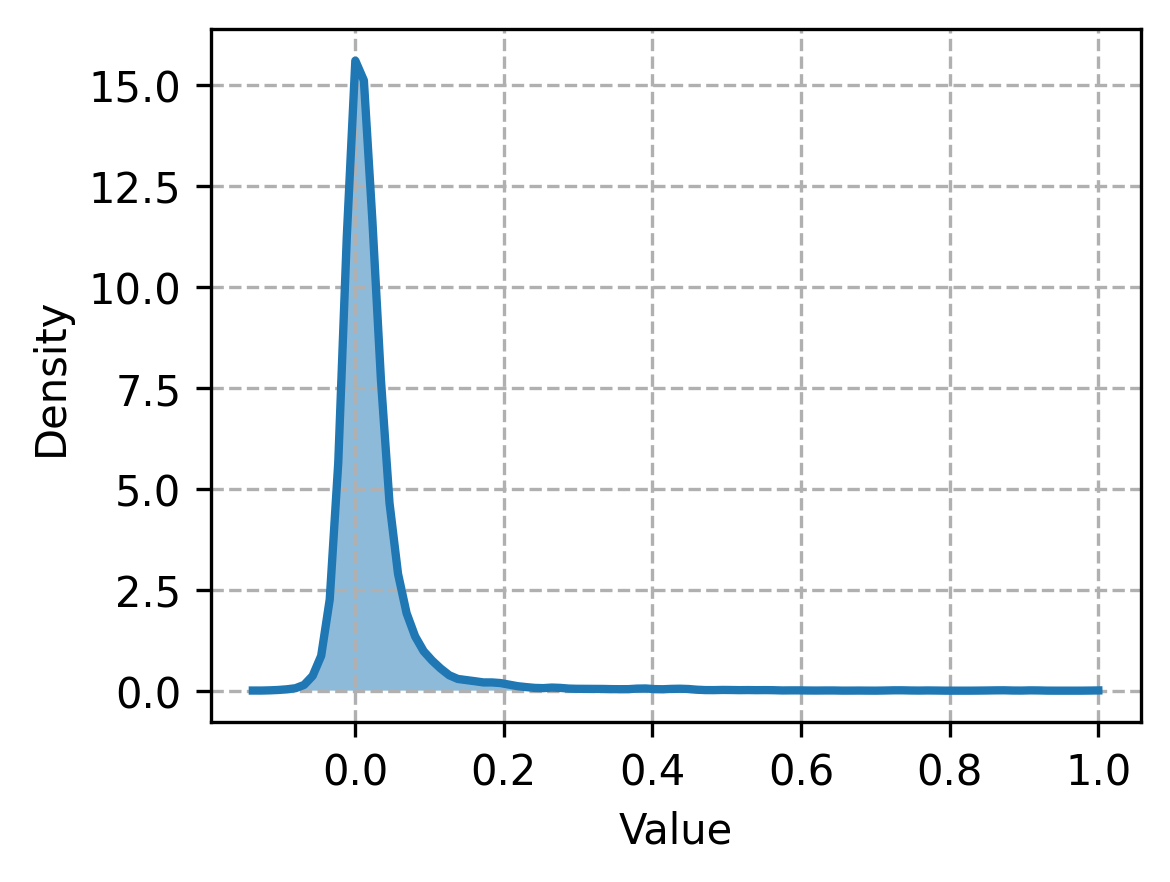}
    }
    \hfill
    \subfigure[Layer 3]{\includegraphics[width=0.22\textwidth]{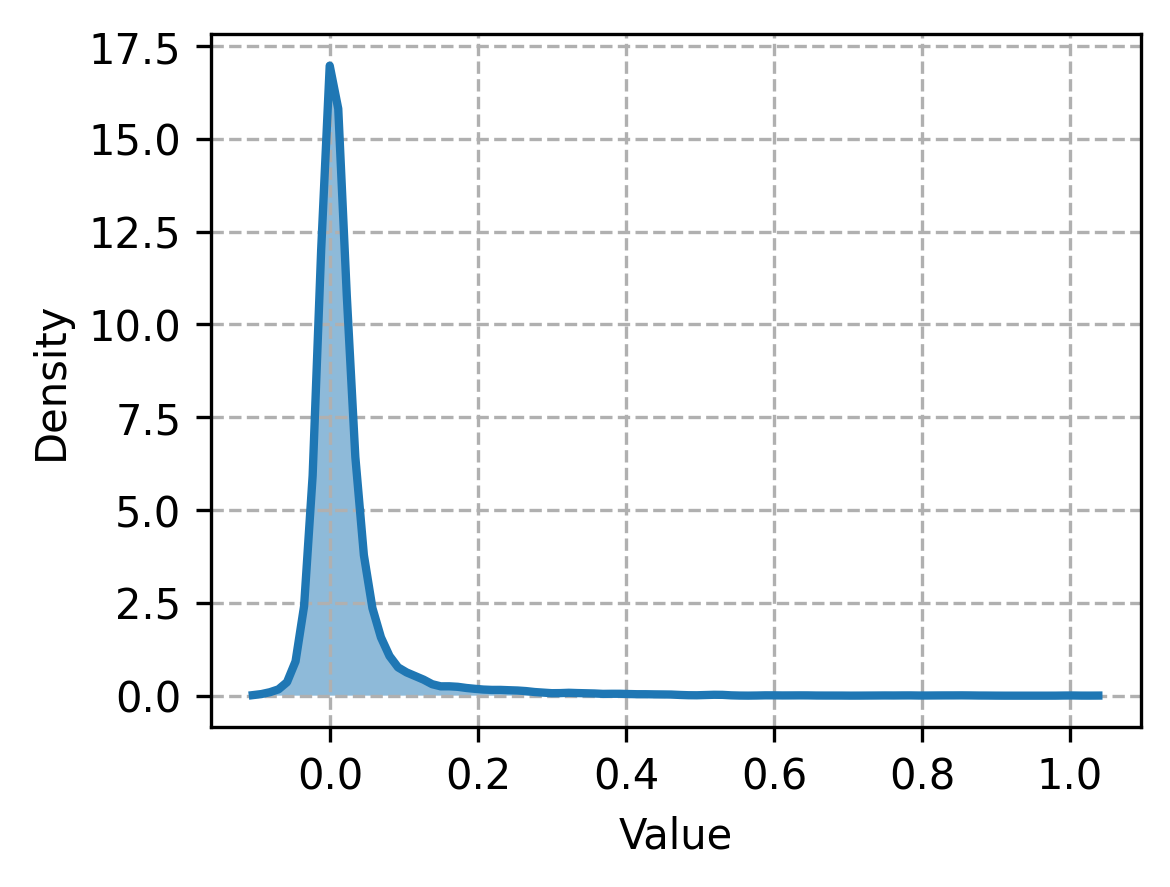}
    }
    \hfill
    \subfigure[Layer 4]{\includegraphics[width=0.22\textwidth]{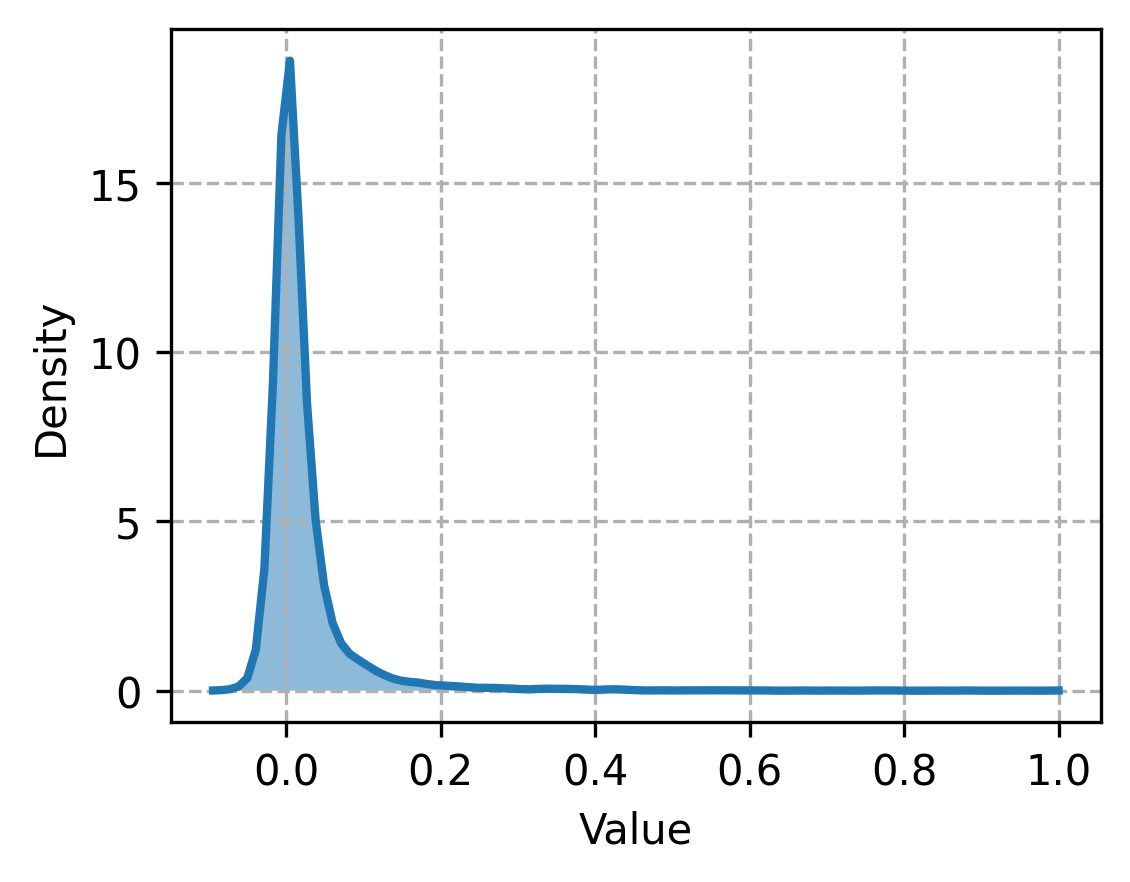}
    }
    \hfill
    \subfigure[Layer 5]{\includegraphics[width=0.22\textwidth]{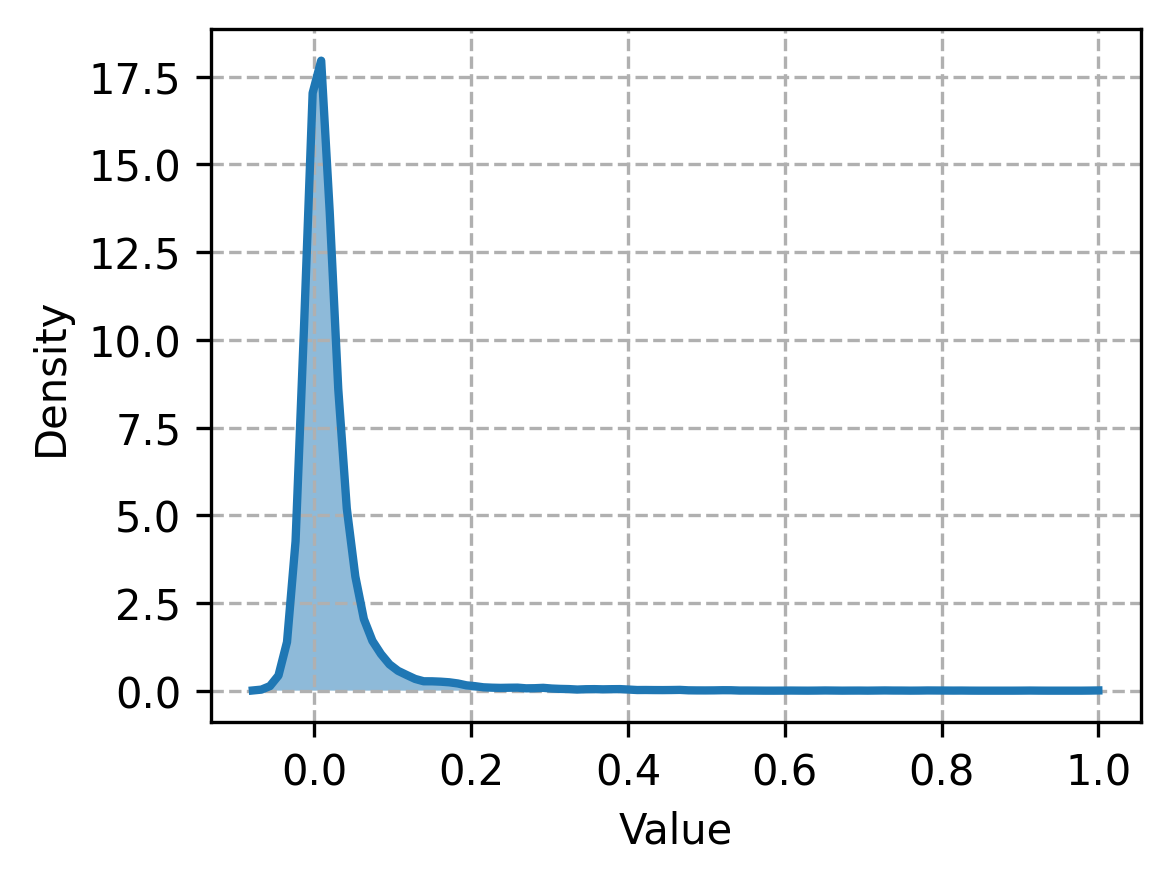}
    }
    \hfill
    \subfigure[Layer 6]{\includegraphics[width=0.22\textwidth]{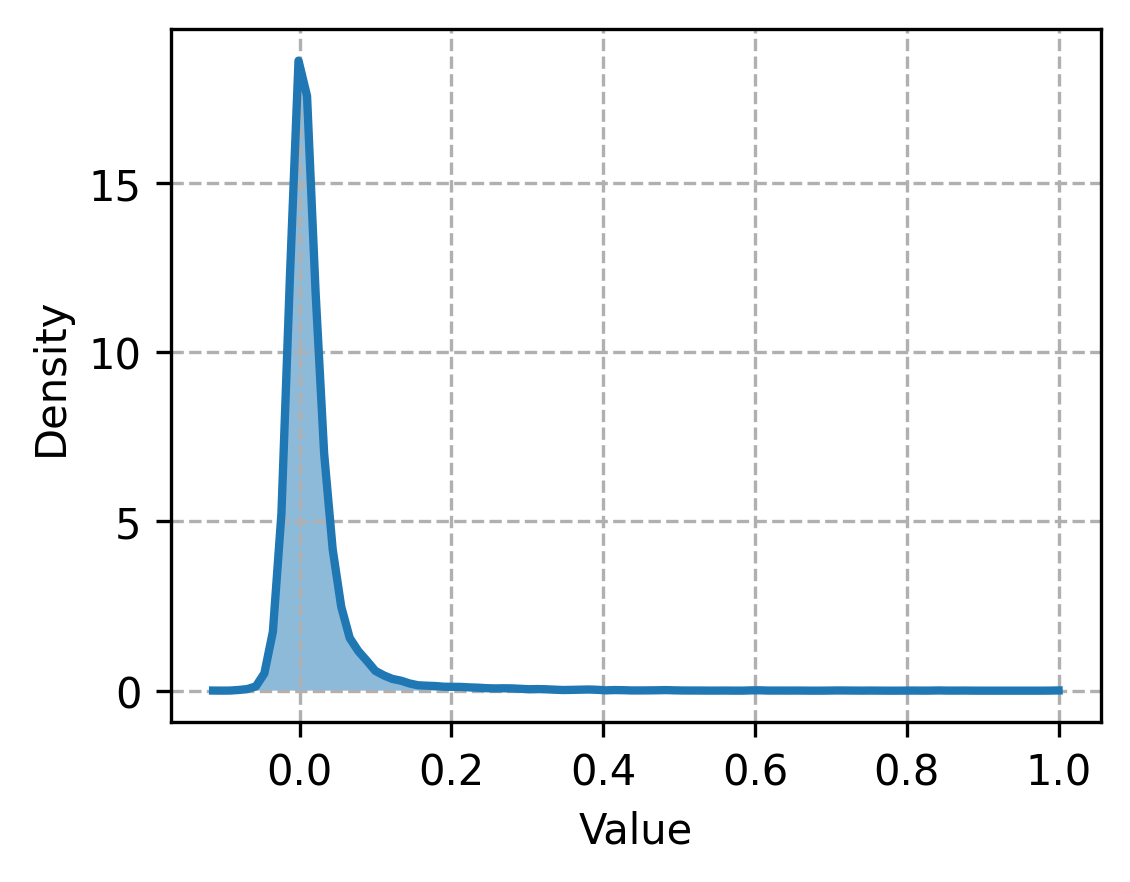}
    }
    \hfill
    \subfigure[Layer 7]{\includegraphics[width=0.22\textwidth]{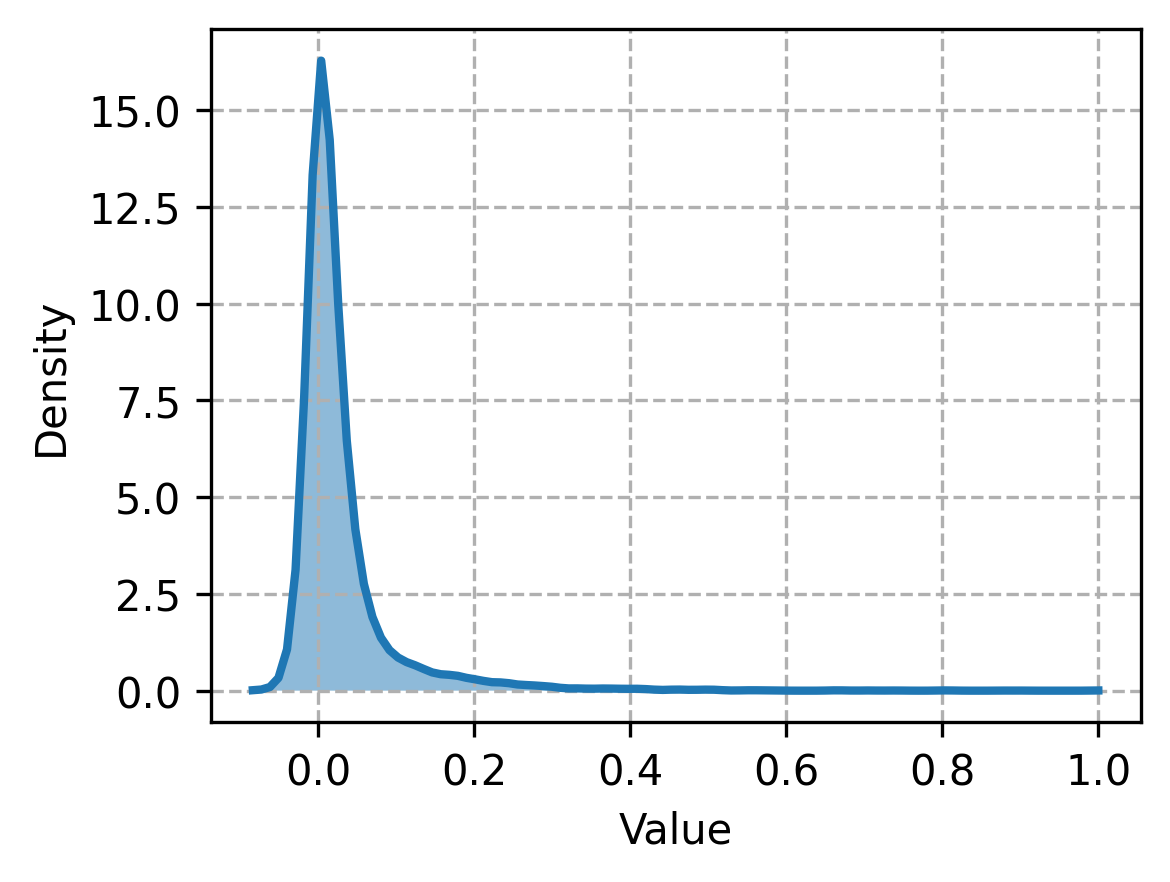}
    }
    \hfill
    \subfigure[Layer 8]{\includegraphics[width=0.22\textwidth]{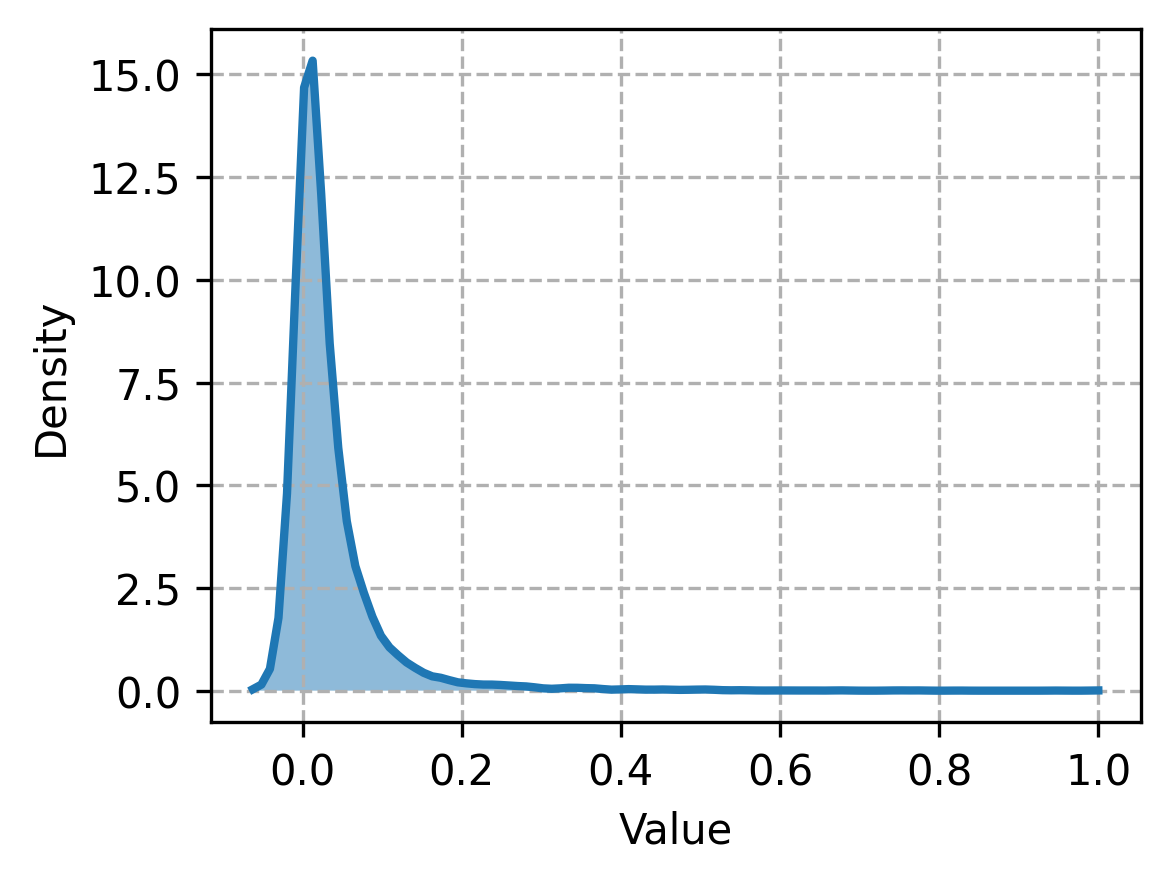}
    }
    \hfill
    \subfigure[Layer 9]{\includegraphics[width=0.22\textwidth]{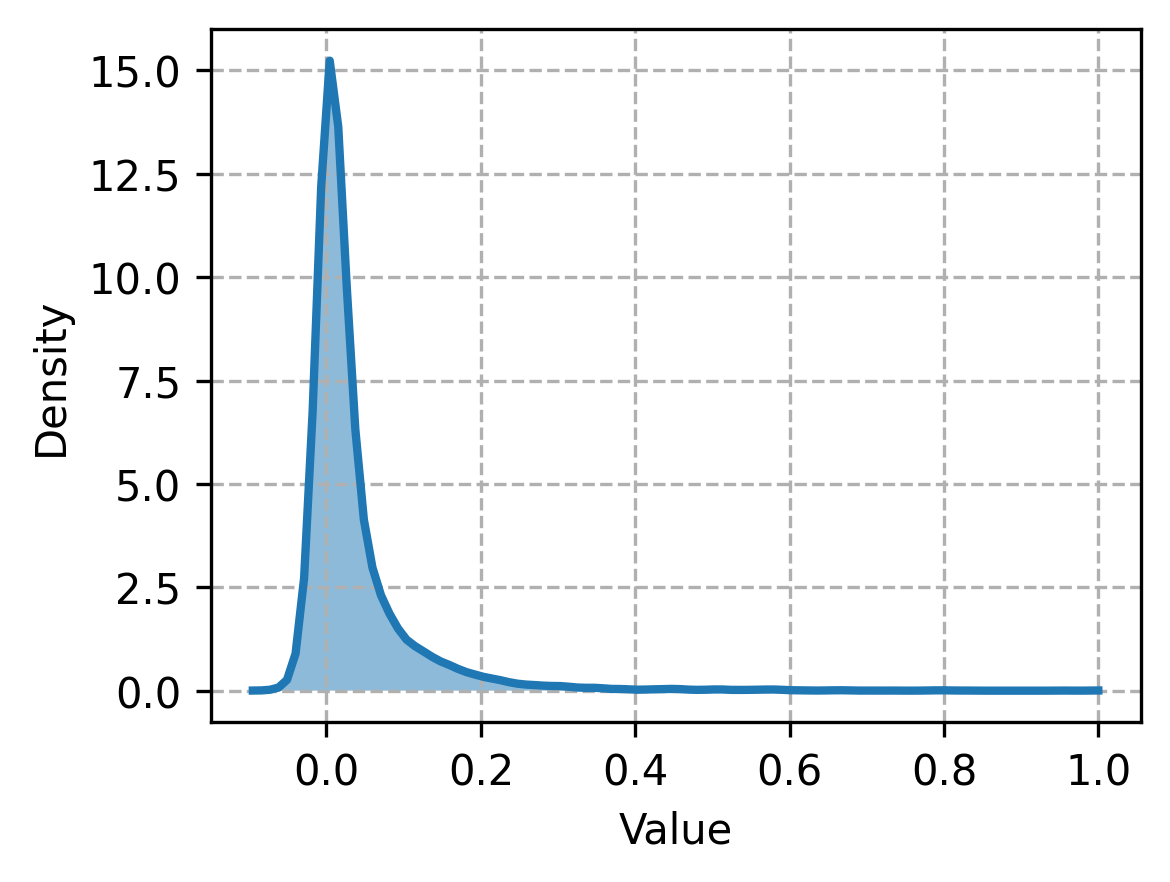}
    }
    \hfill
    \subfigure[Layer 10]{\includegraphics[width=0.22\textwidth]{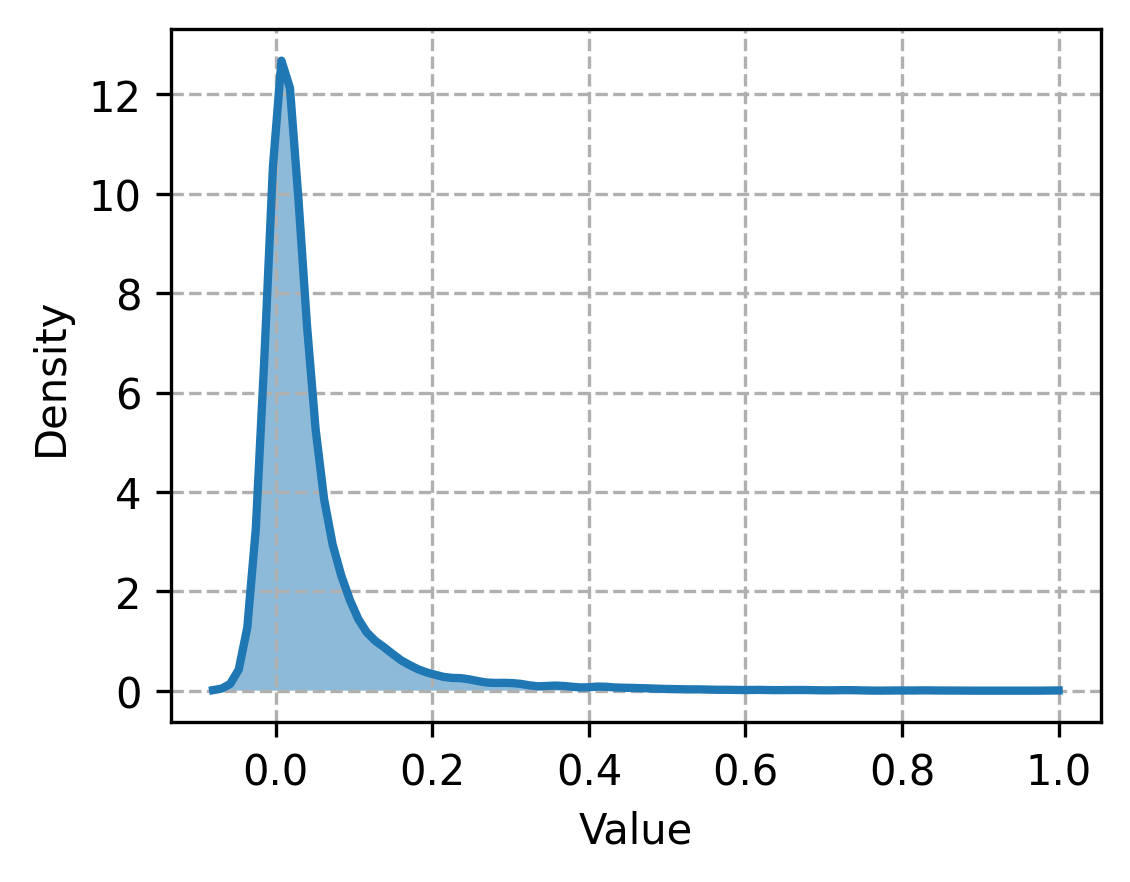}
    }
    \hfill
    \subfigure[Layer 11]{\includegraphics[width=0.22\textwidth]{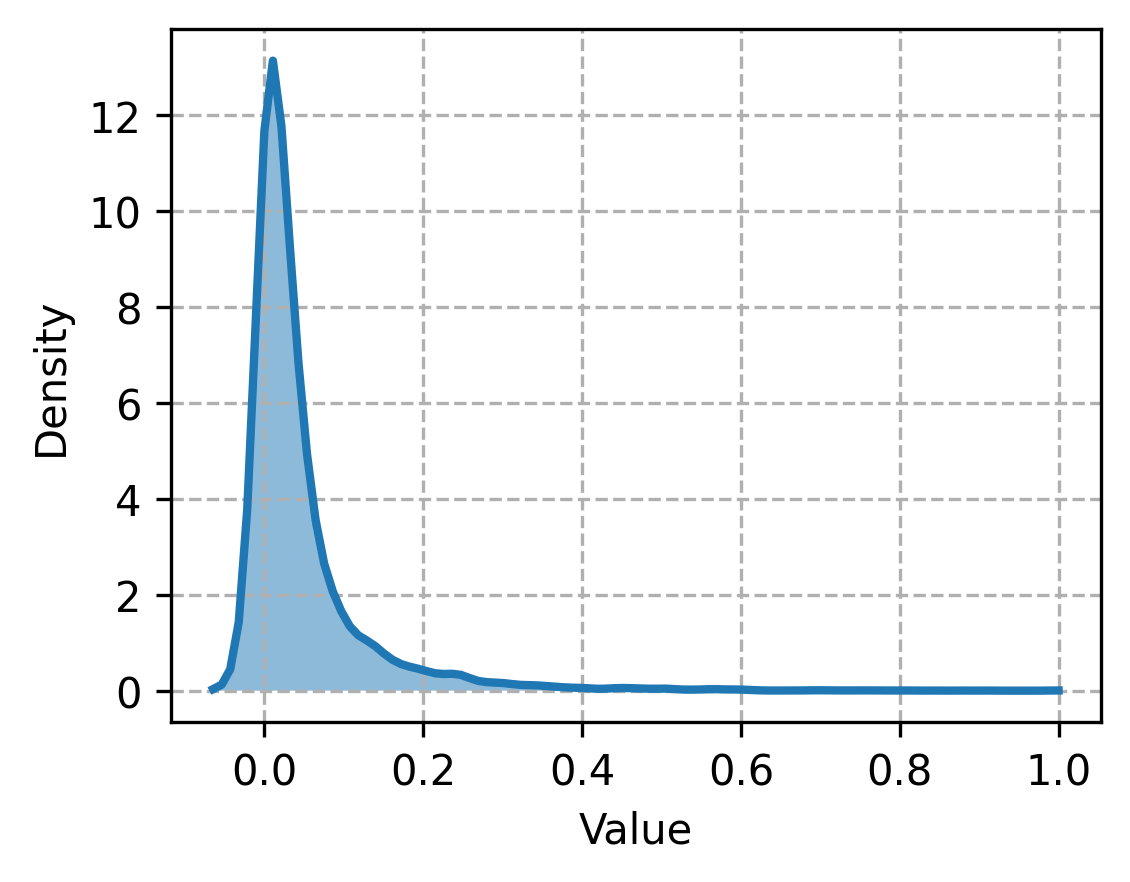}
    }
    \hfill
    \subfigure[Layer 12]{\includegraphics[width=0.22\textwidth]{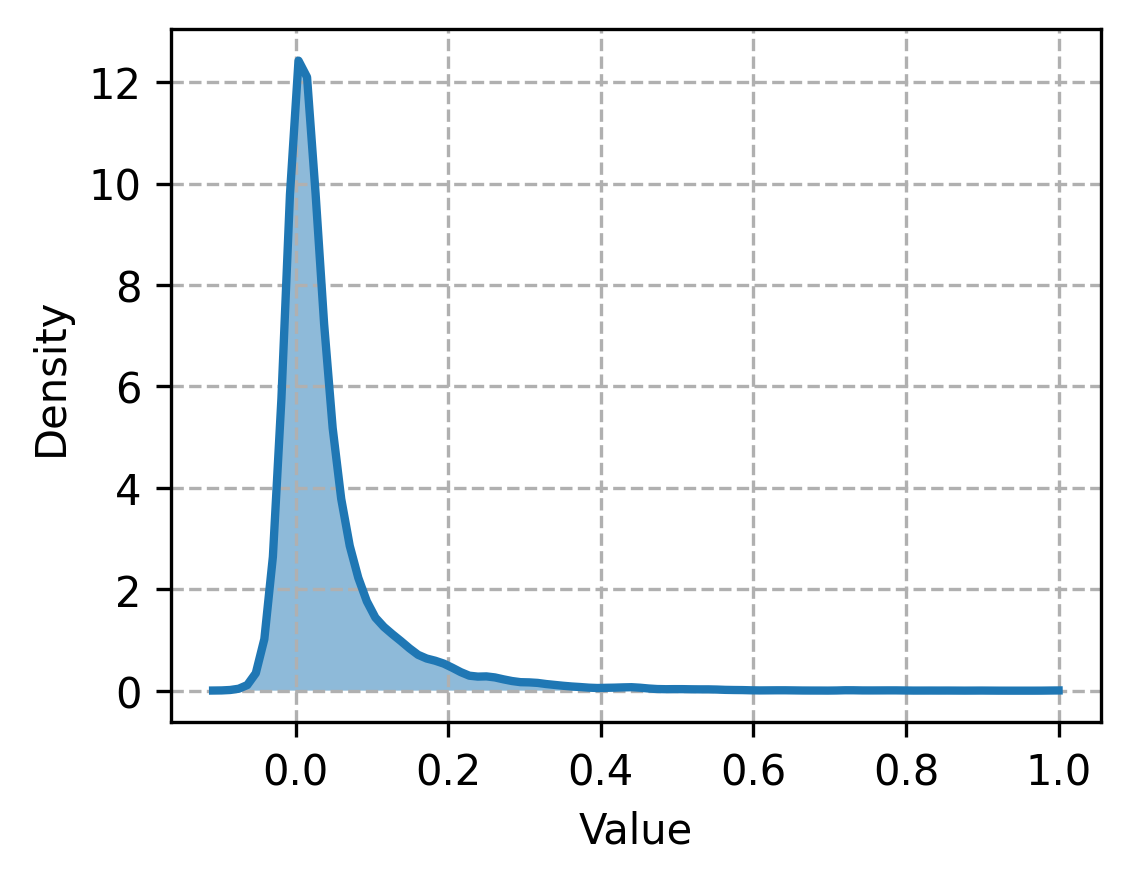}
    }
    \hfill
    \subfigure[Layer 13]{\includegraphics[width=0.22\textwidth]{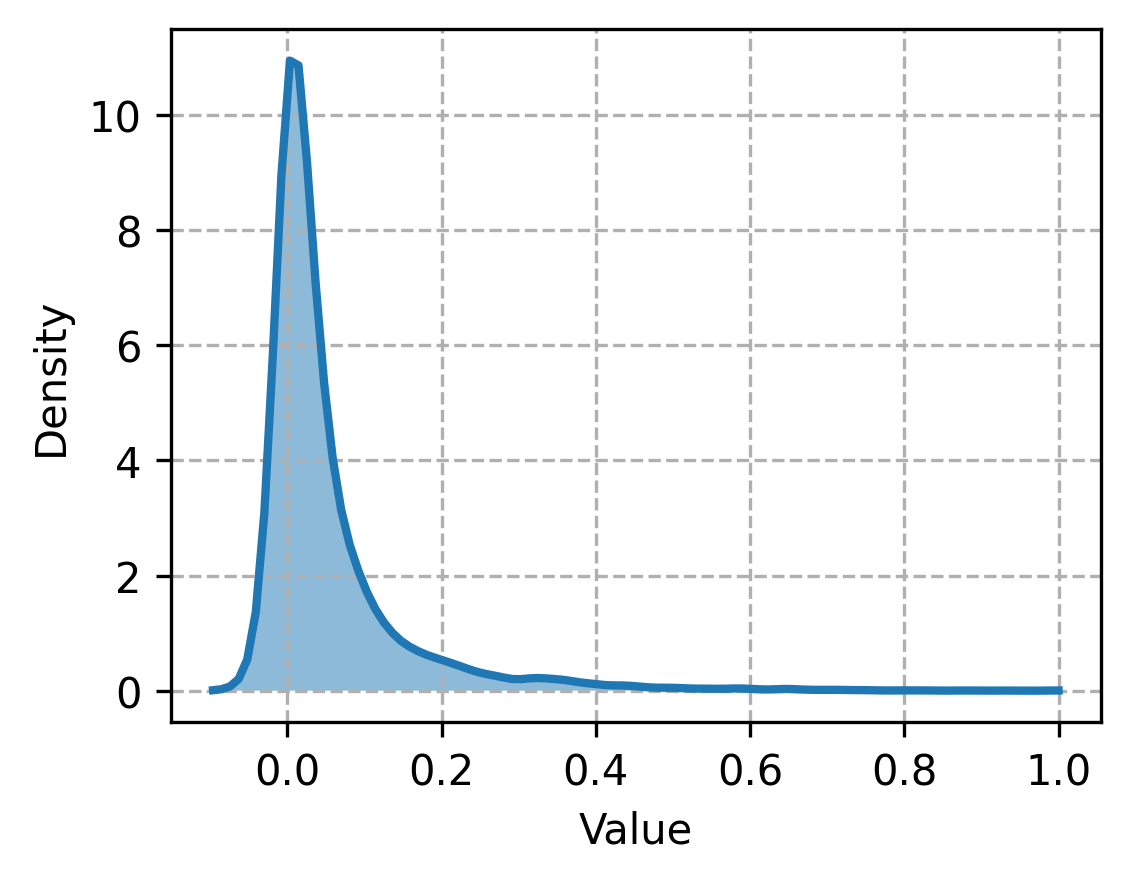}
    }
    \hfill
    \subfigure[Layer 14]{\includegraphics[width=0.22\textwidth]{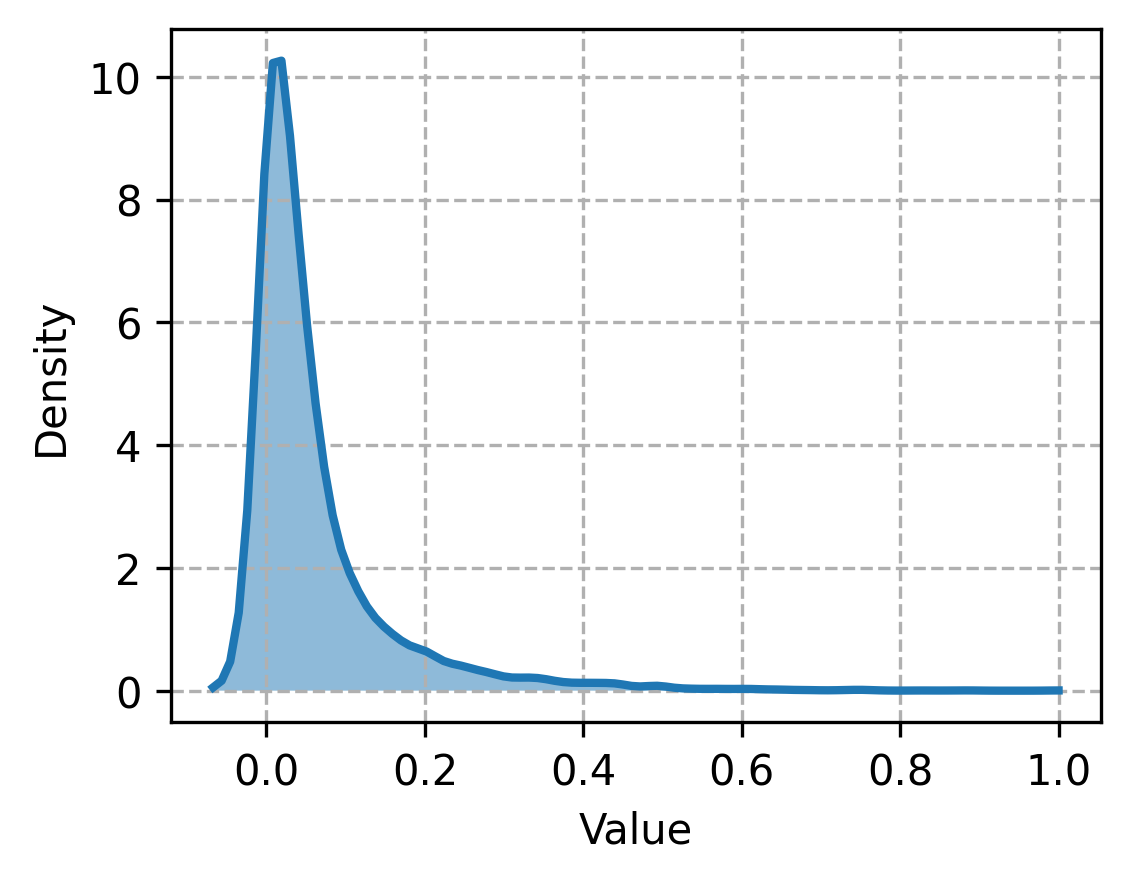}
    }
    \hfill
    \subfigure[Layer 15]{\includegraphics[width=0.22\textwidth]{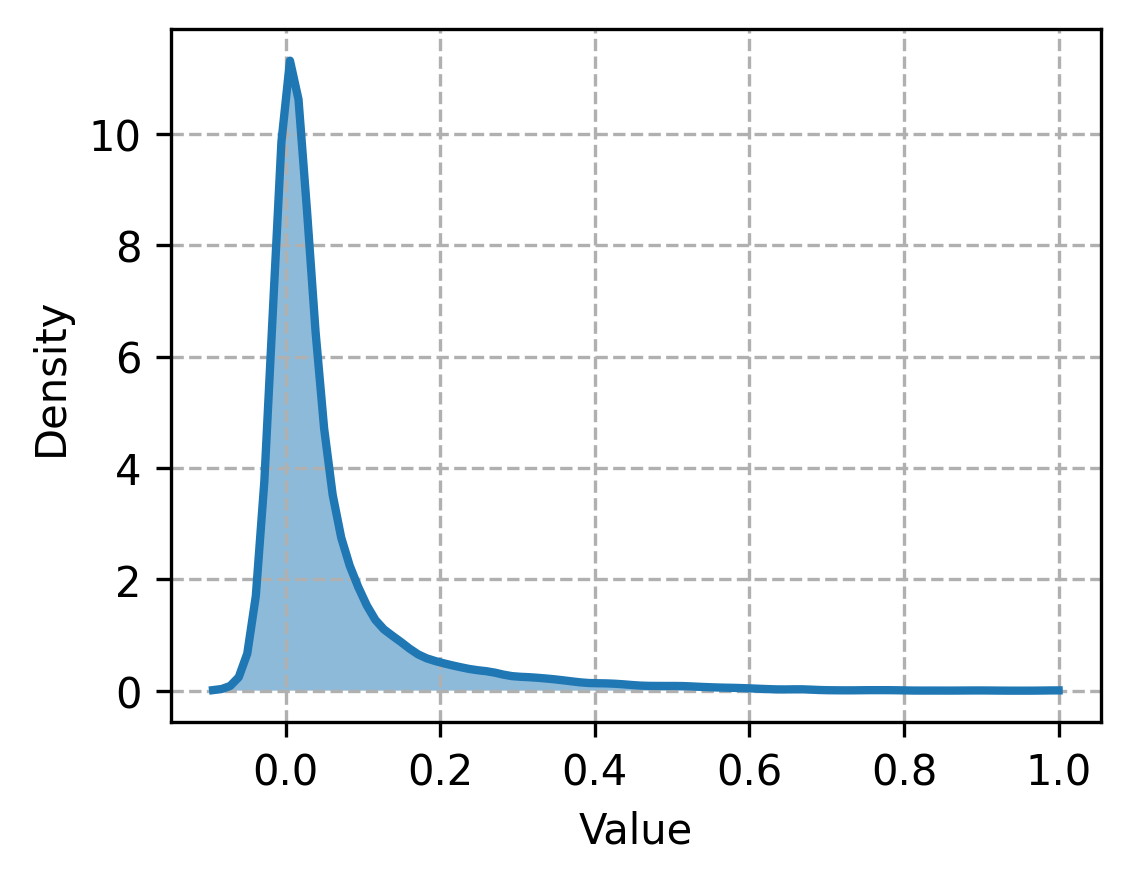}
    }
    \hfill
    \subfigure[Layer 16]{\includegraphics[width=0.22\textwidth]{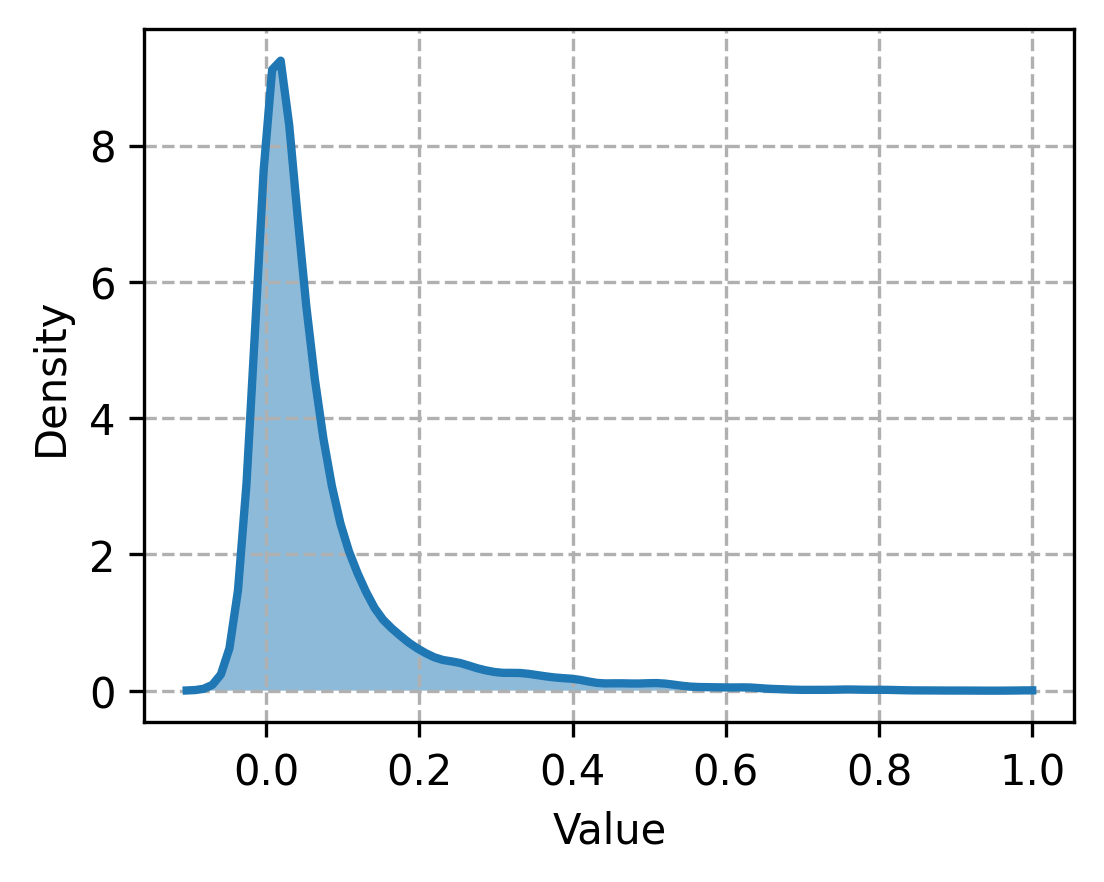}
    }
    \hfill
    \subfigure[Layer 17]{\includegraphics[width=0.22\textwidth]{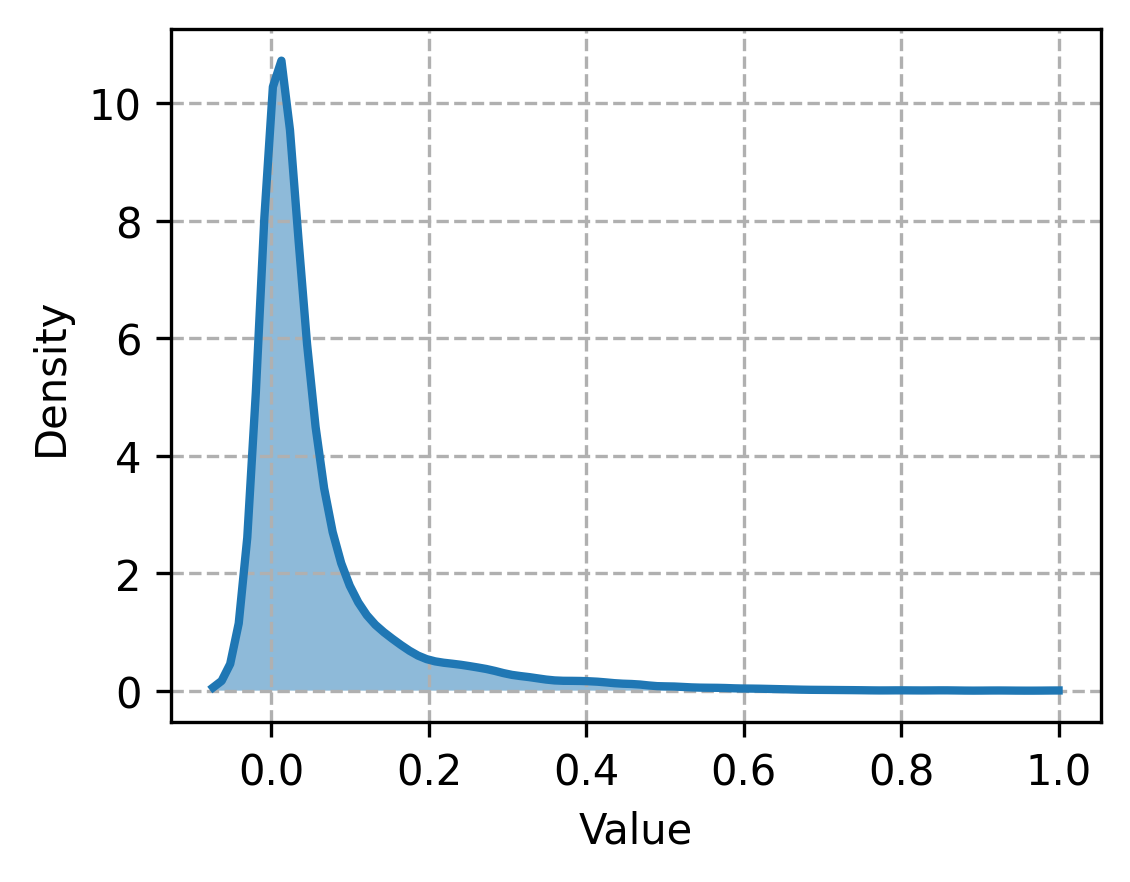}
    }
    \hfill
    \subfigure[Layer 18]{\includegraphics[width=0.22\textwidth]{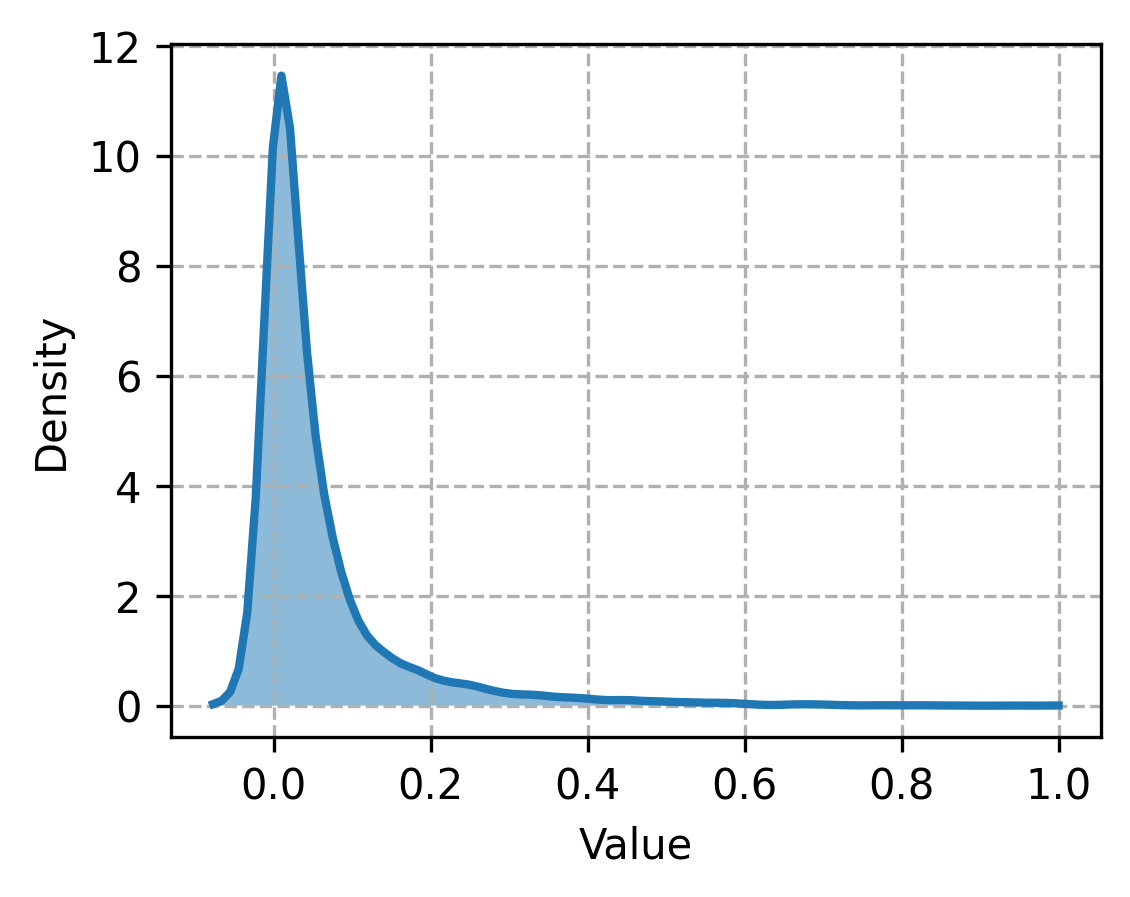}
    }
    \hfill
    \subfigure[Layer 19]{\includegraphics[width=0.22\textwidth]{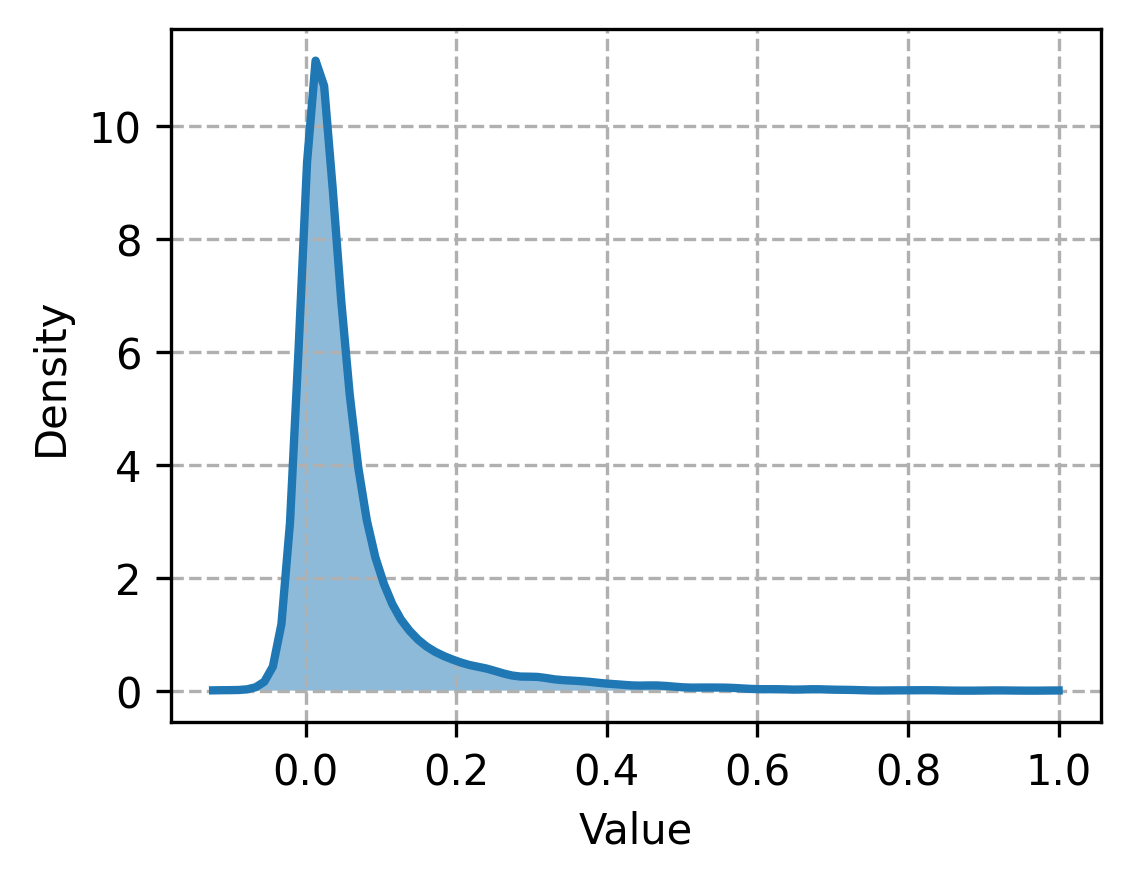}
    }
  \caption{In GPT2-Medium, the KDE of elements in \(P\) matrices across layers (0-19).}
  \label{fig:superposition_KDE_gpt2-medium-part1}
\end{figure*}

\begin{figure*}
    \centering
    \subfigure[Layer 20]{\includegraphics[width=0.22\textwidth]{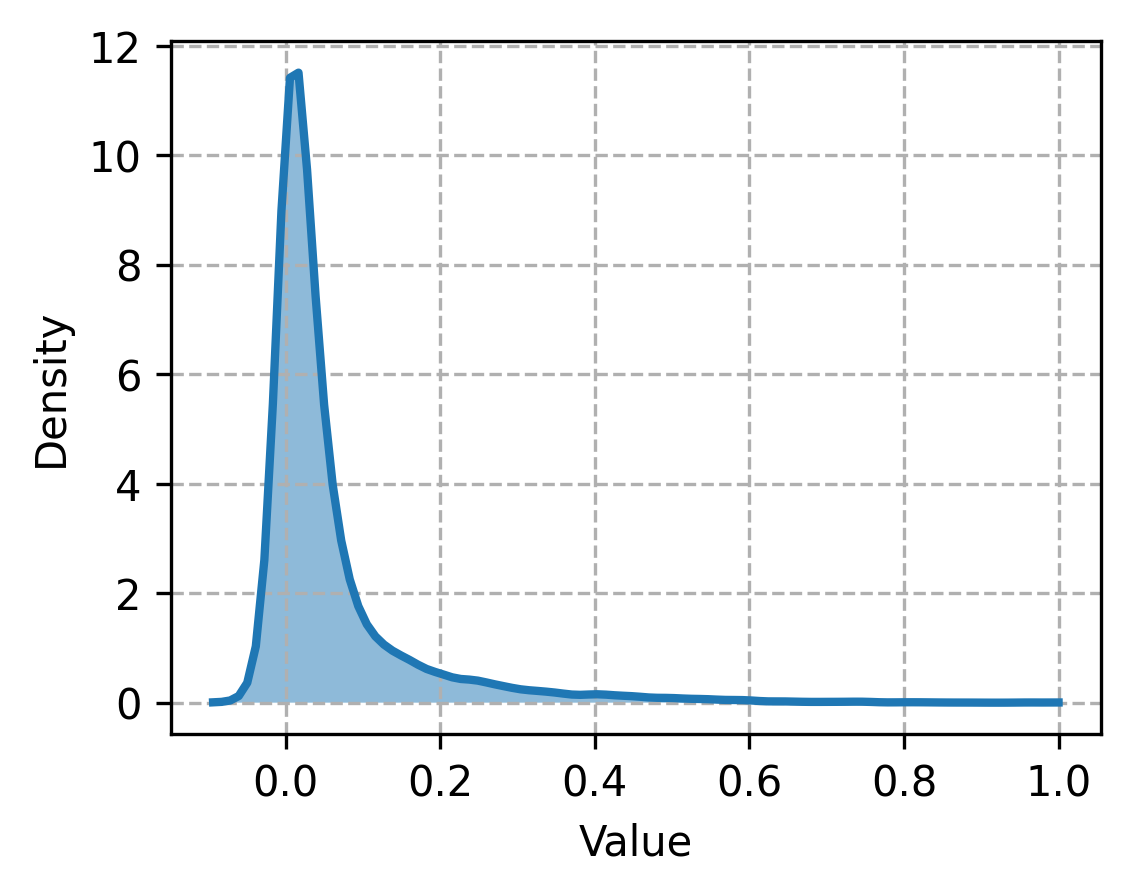}
    }
    \hfill
    \subfigure[Layer 21]{\includegraphics[width=0.22\textwidth]{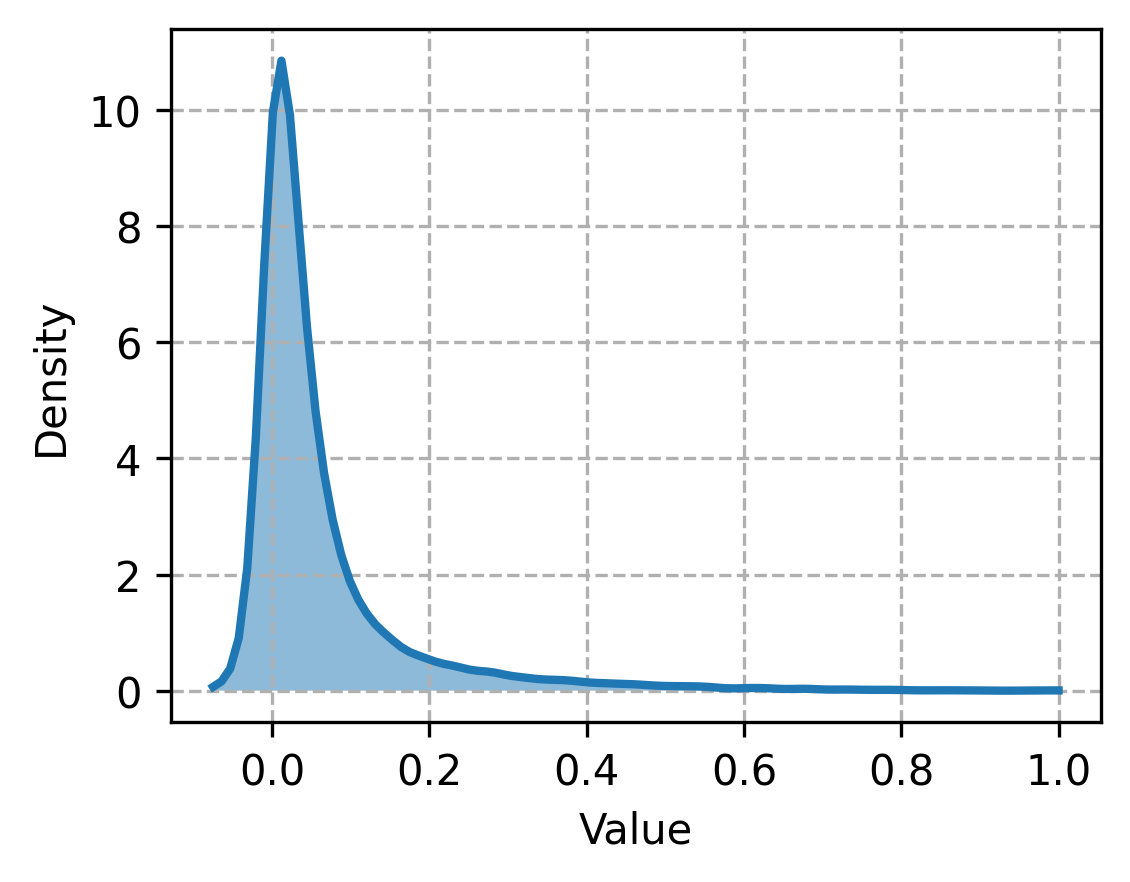}
    }
    \hfill
    \subfigure[Layer 22]{\includegraphics[width=0.22\textwidth]{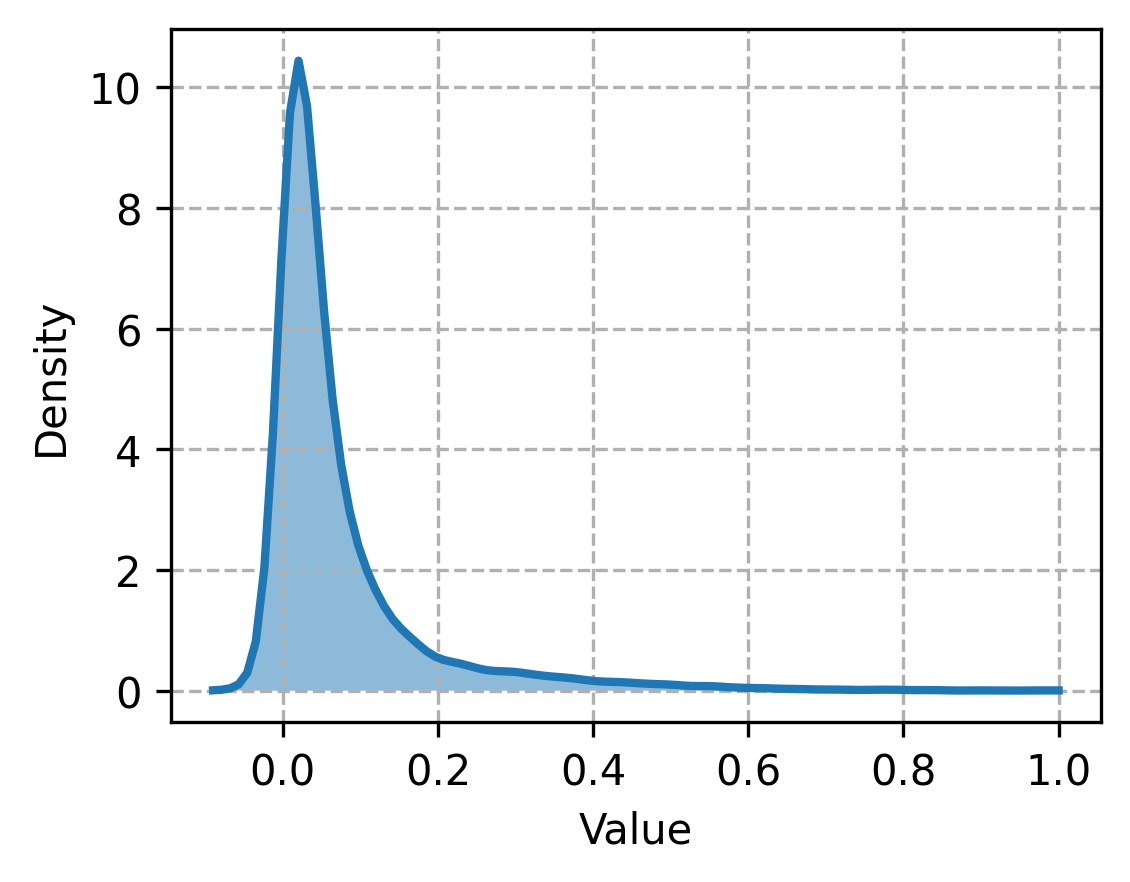}
    }
    \hfill
    \subfigure[Layer 23]{\includegraphics[width=0.22\textwidth]{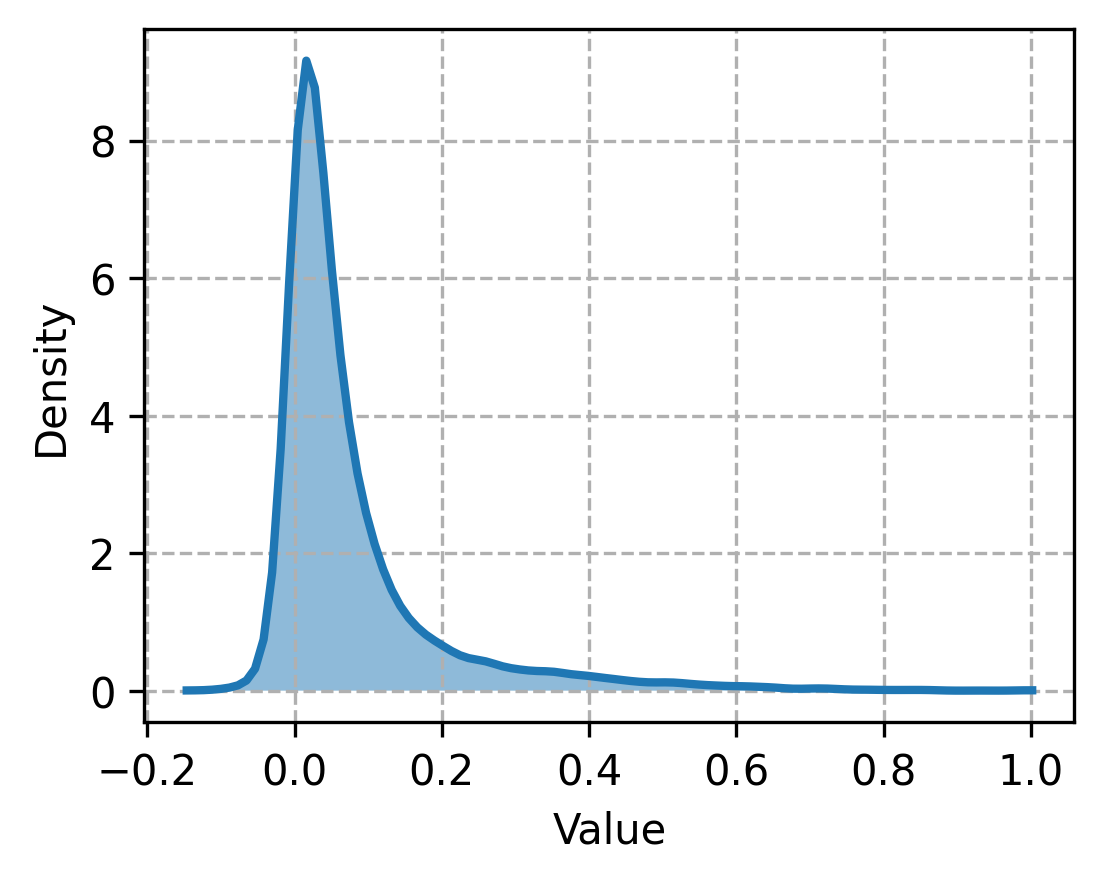}
    }
  \caption{In GPT2-Medium, the KDE of elements in \(P\) matrices across layers (20-23).}
  \label{fig:superposition_KDE_gpt2-medium-part2}
\end{figure*}

\begin{figure*}
    \centering
    \subfigure[Layer 0]{\includegraphics[width=0.22\textwidth]{fig/gpt2-large/p_matrix/known/KDE/superposition_for_layer_0.png}
    }
    \hfill
    \subfigure[Layer 1]{\includegraphics[width=0.22\textwidth]{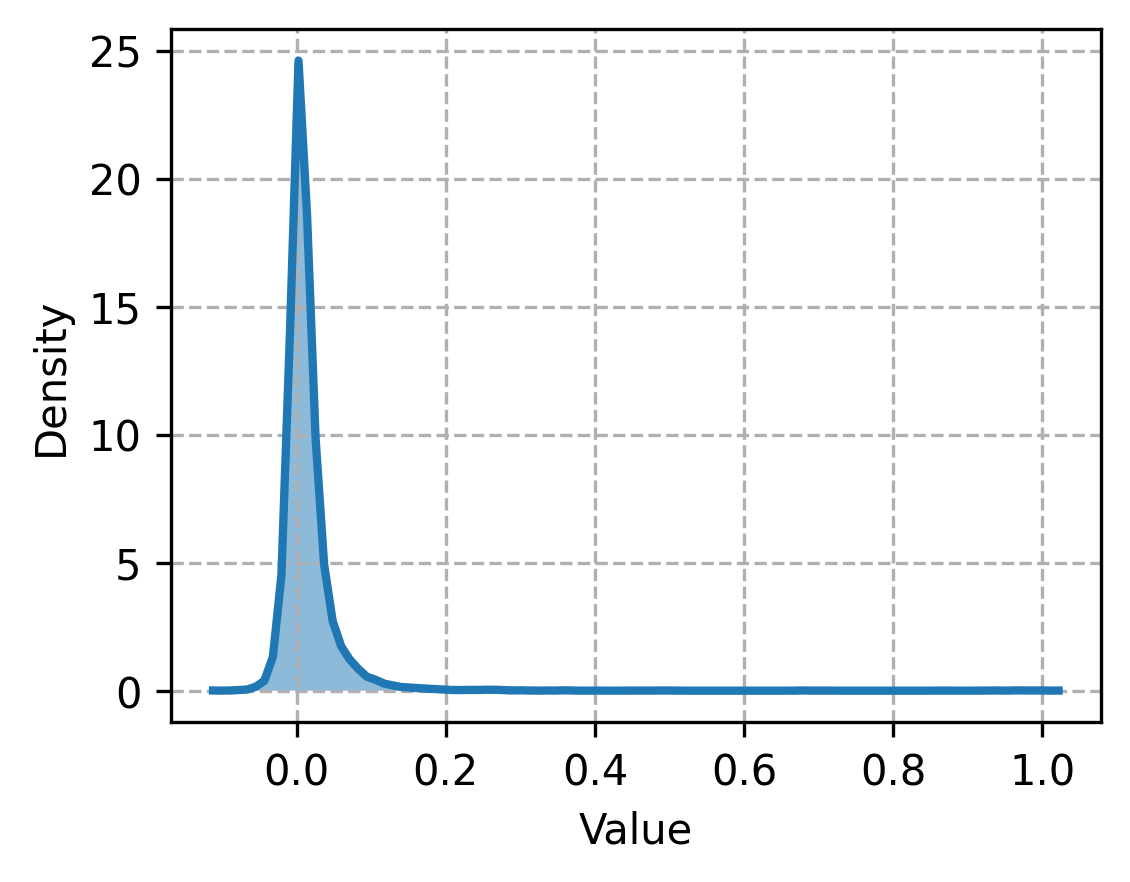}
    }
    \hfill
    \subfigure[Layer 2]{\includegraphics[width=0.22\textwidth]{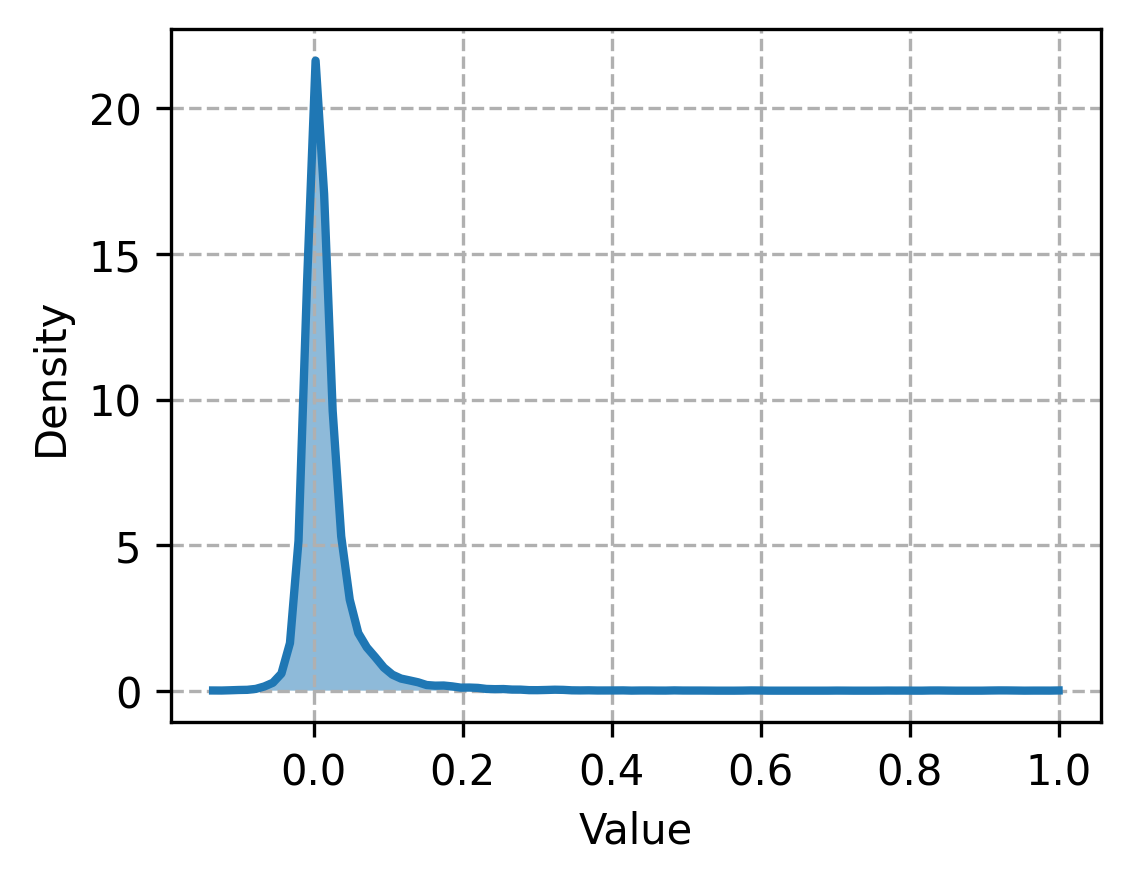}
    }
    \hfill
    \subfigure[Layer 3]{\includegraphics[width=0.22\textwidth]{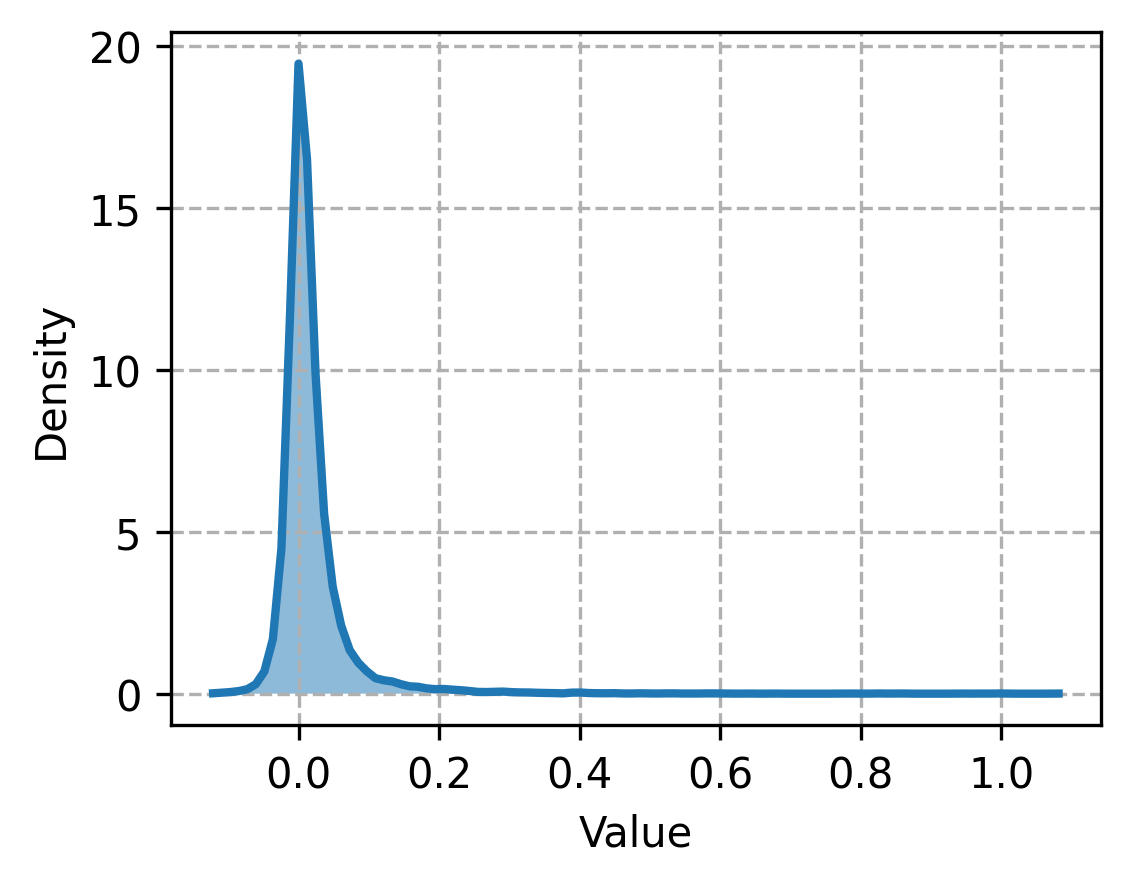}
    }
    \hfill
    \subfigure[Layer 4]{\includegraphics[width=0.22\textwidth]{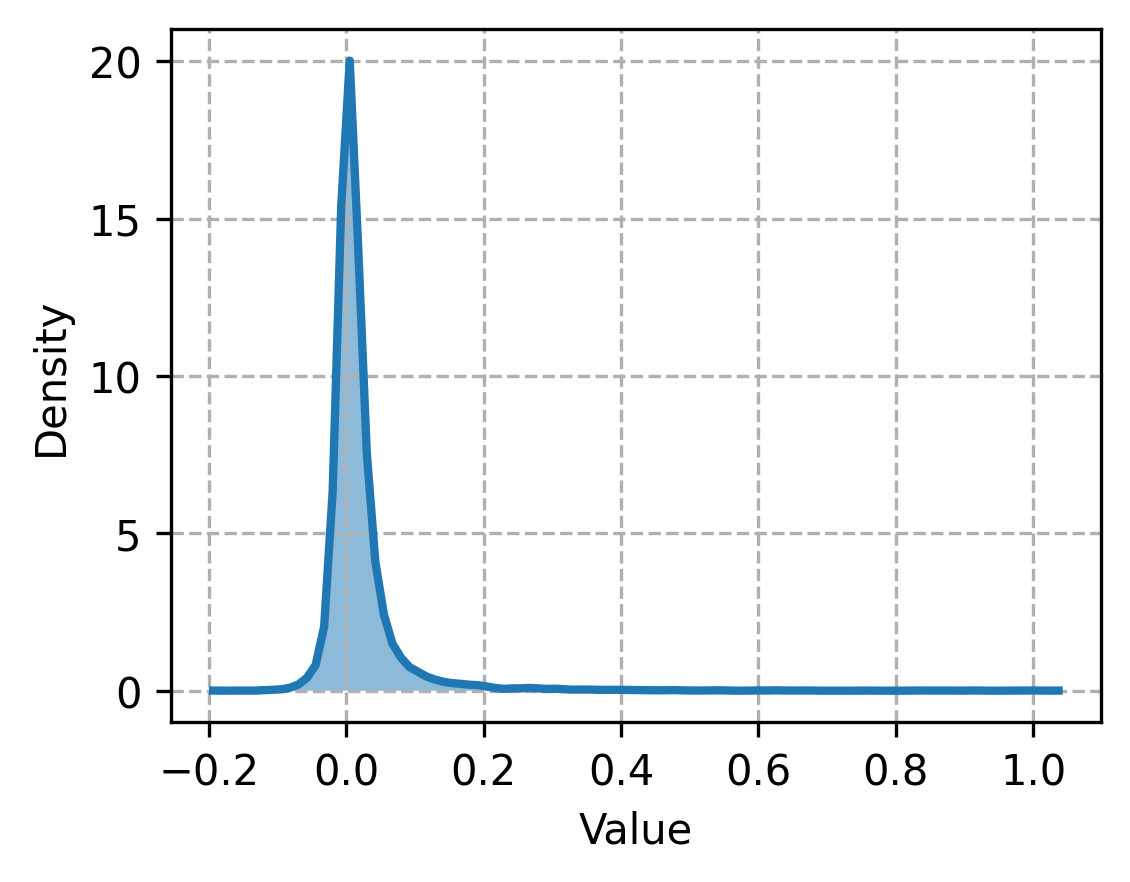}
    }
    \hfill
    \subfigure[Layer 5]{\includegraphics[width=0.22\textwidth]{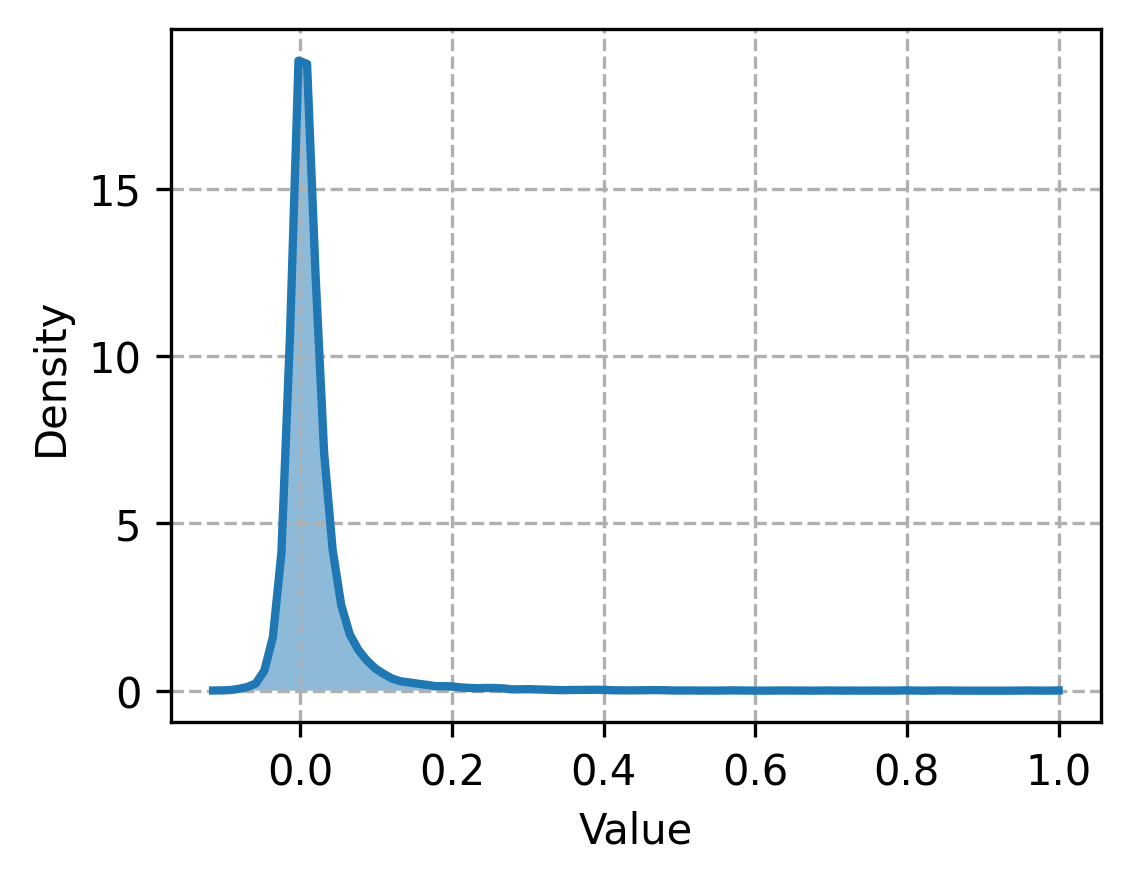}
    }
    \hfill
    \subfigure[Layer 6]{\includegraphics[width=0.22\textwidth]{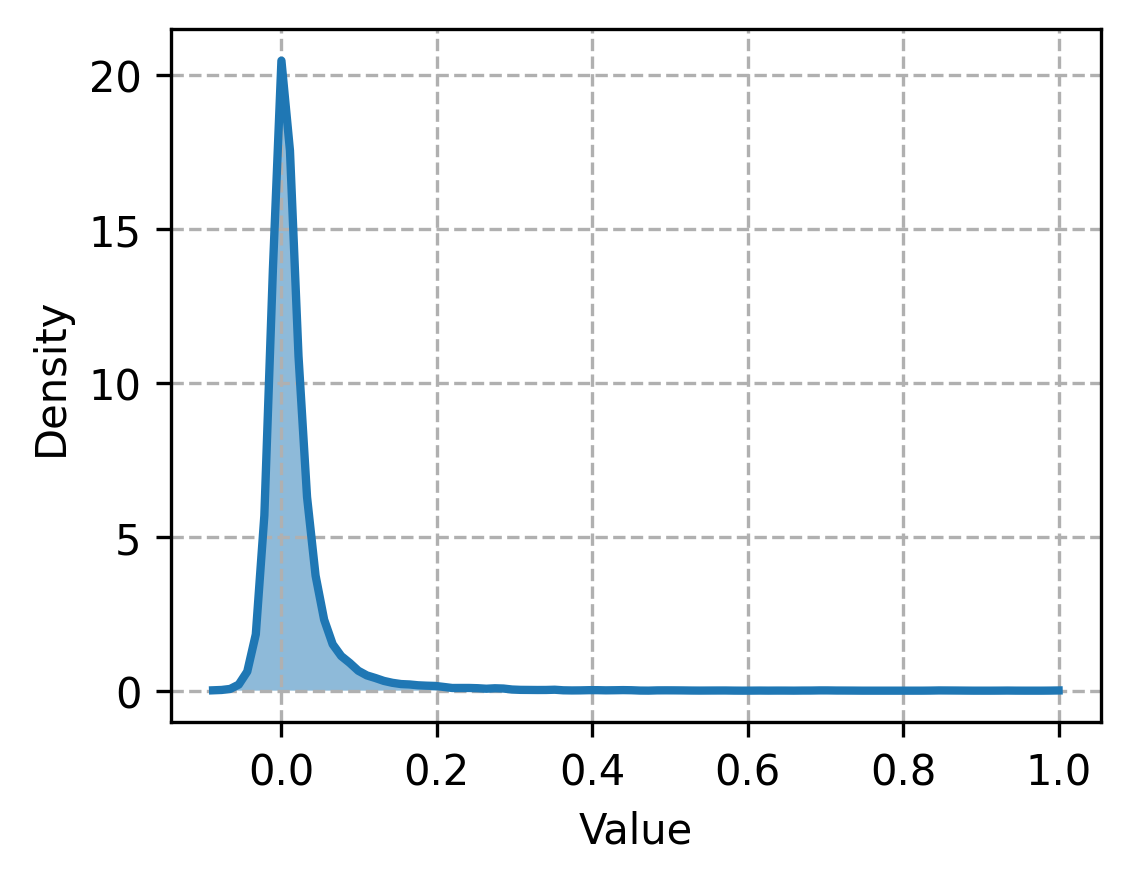}
    }
    \hfill
    \subfigure[Layer 7]{\includegraphics[width=0.22\textwidth]{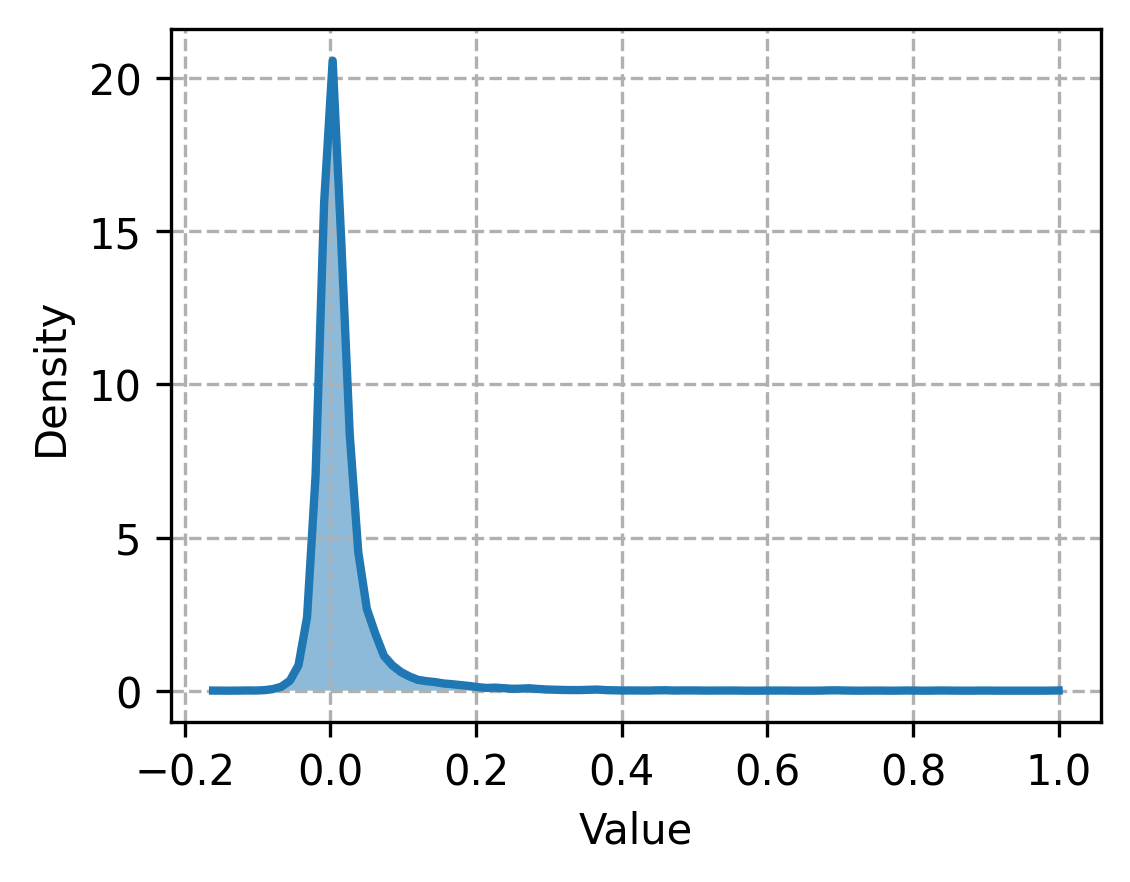}
    }
    \hfill
    \subfigure[Layer 8]{\includegraphics[width=0.22\textwidth]{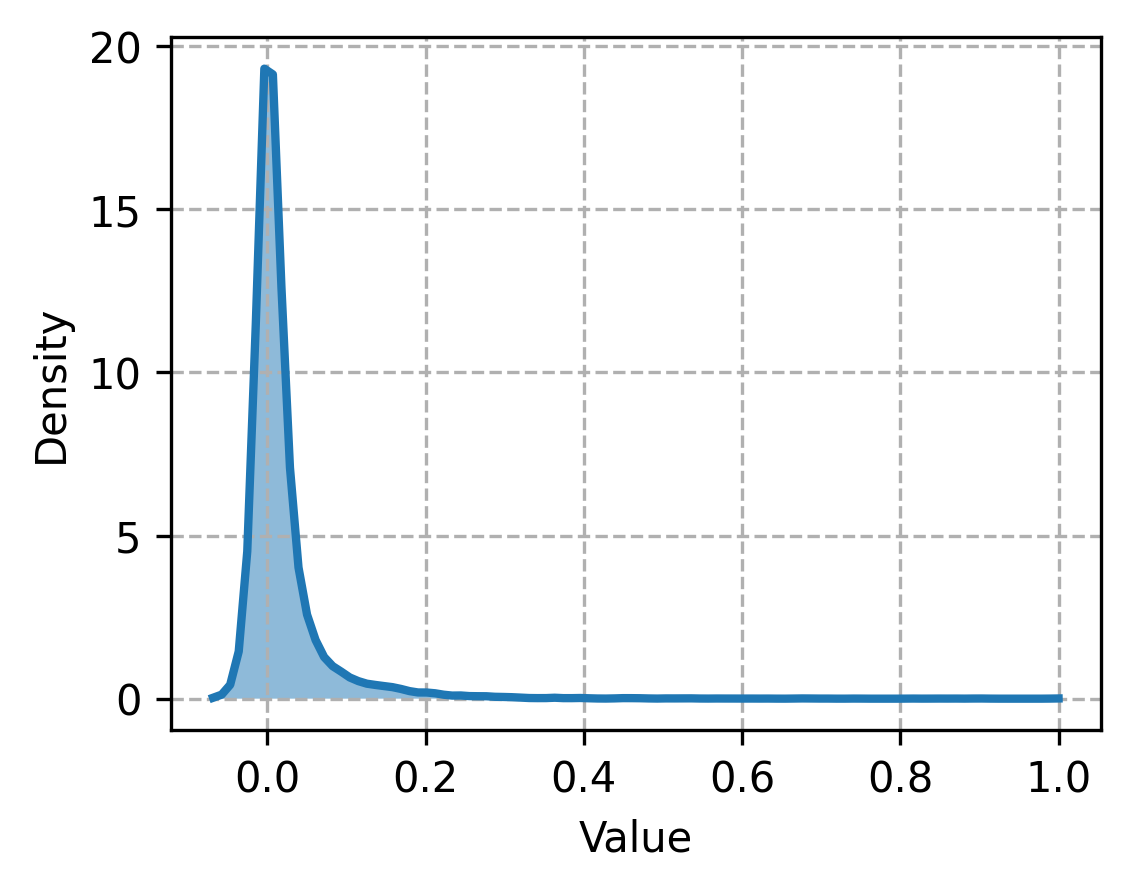}
    }
    \hfill
    \subfigure[Layer 9]{\includegraphics[width=0.22\textwidth]{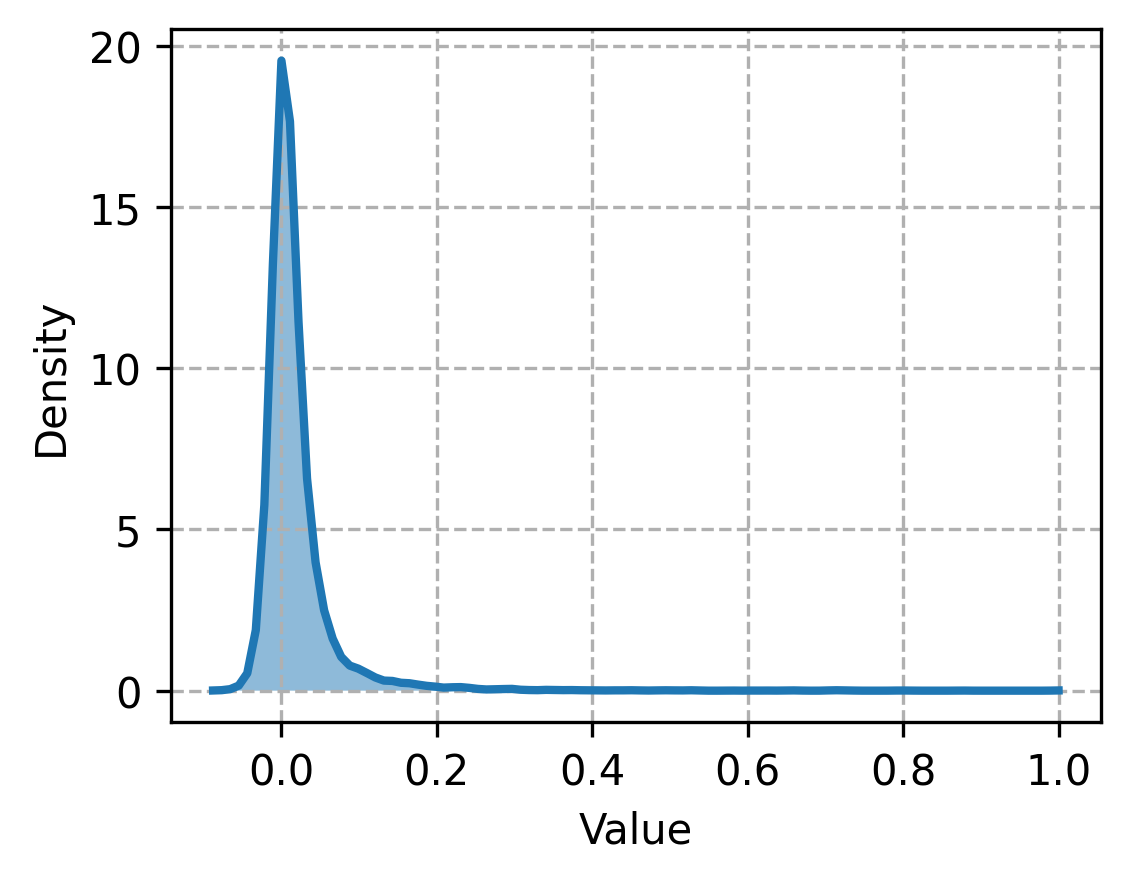}
    }
    \hfill
    \subfigure[Layer 10]{\includegraphics[width=0.22\textwidth]{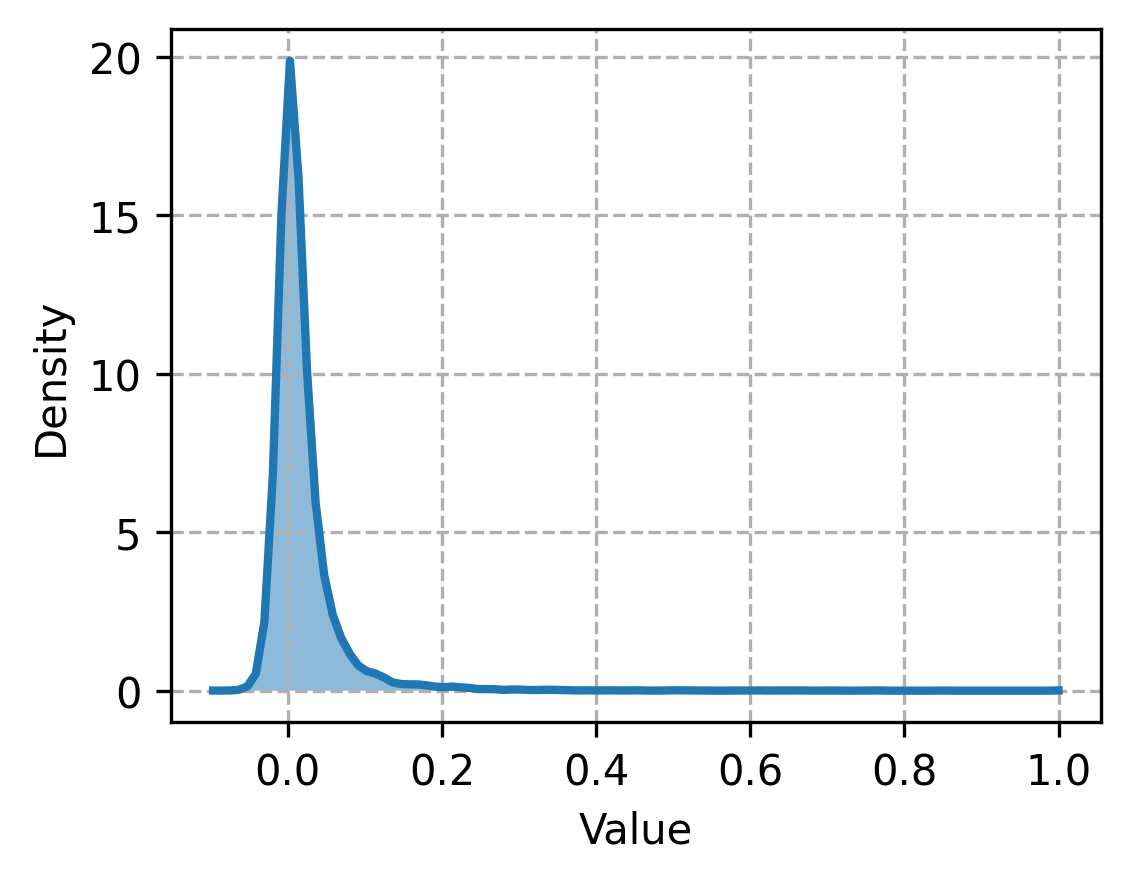}
    }
    \hfill
    \subfigure[Layer 11]{\includegraphics[width=0.22\textwidth]{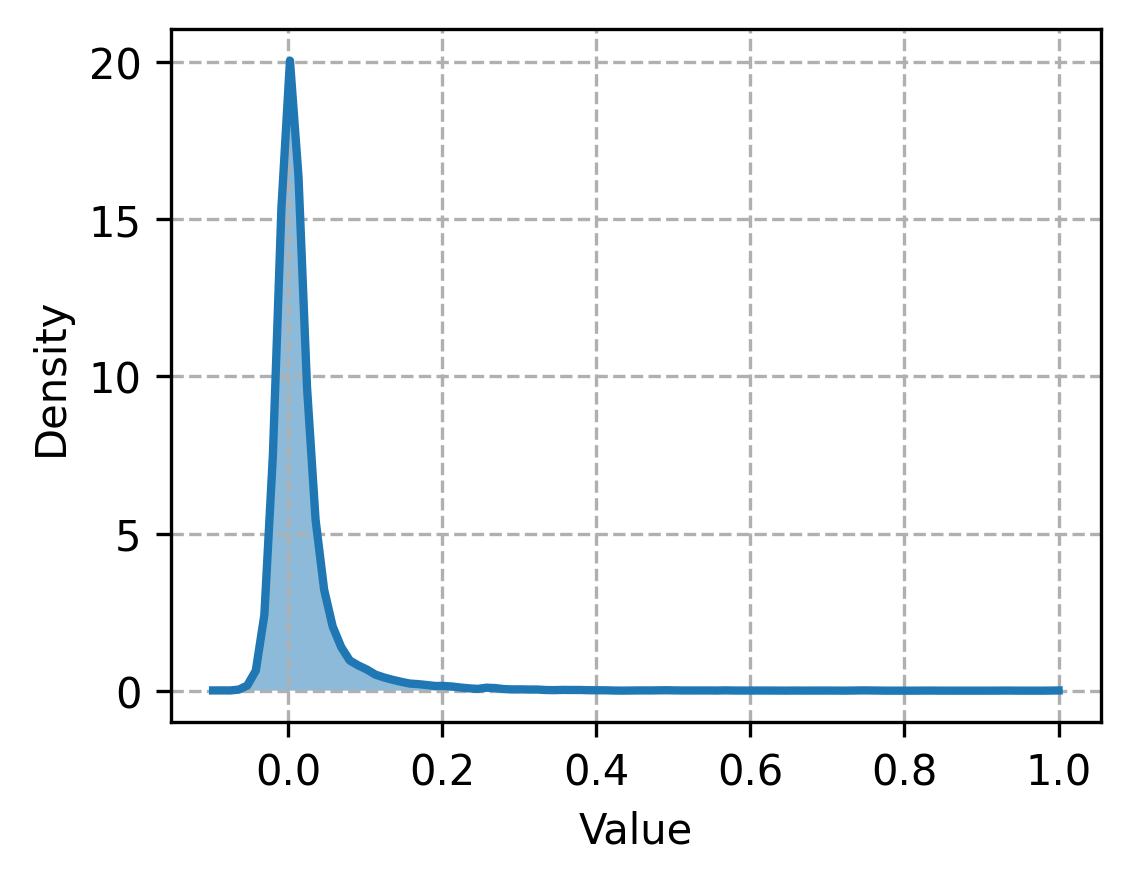}
    }
    \hfill
    \subfigure[Layer 12]{\includegraphics[width=0.22\textwidth]{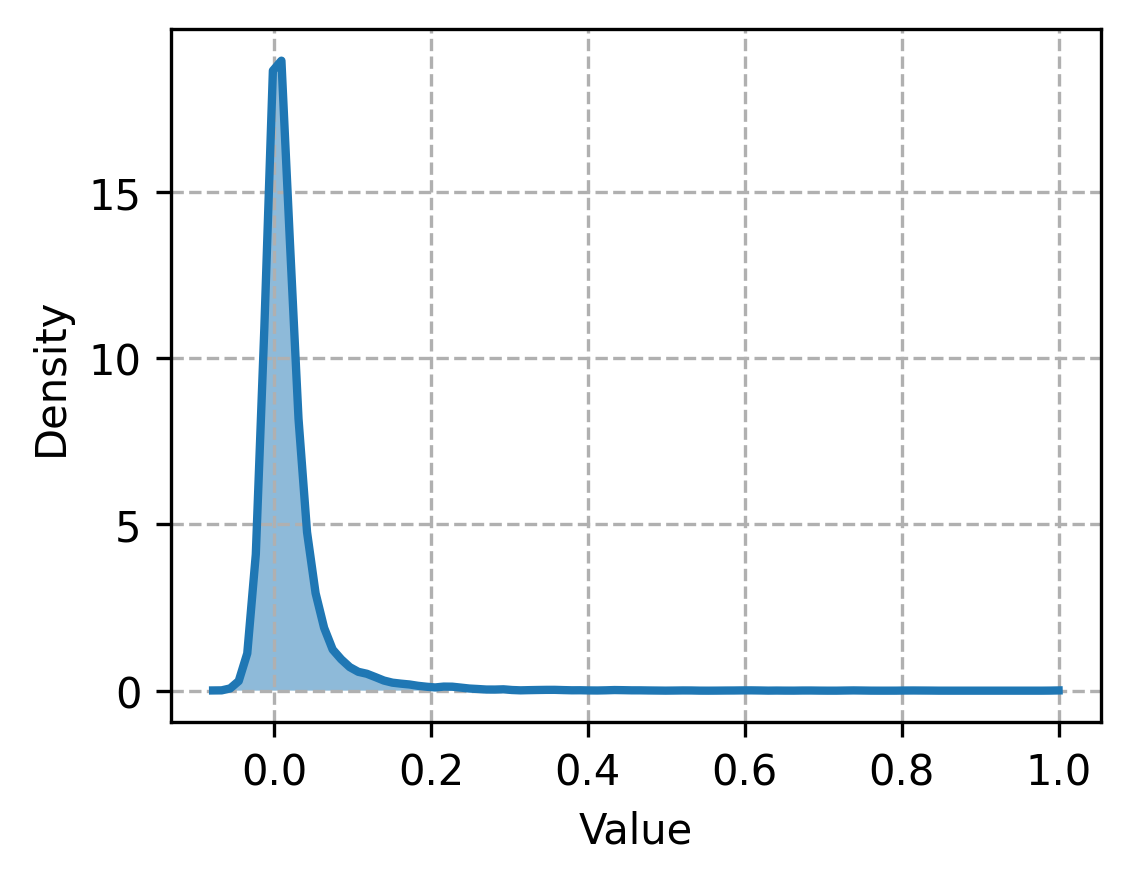}
    }
    \hfill
    \subfigure[Layer 13]{\includegraphics[width=0.22\textwidth]{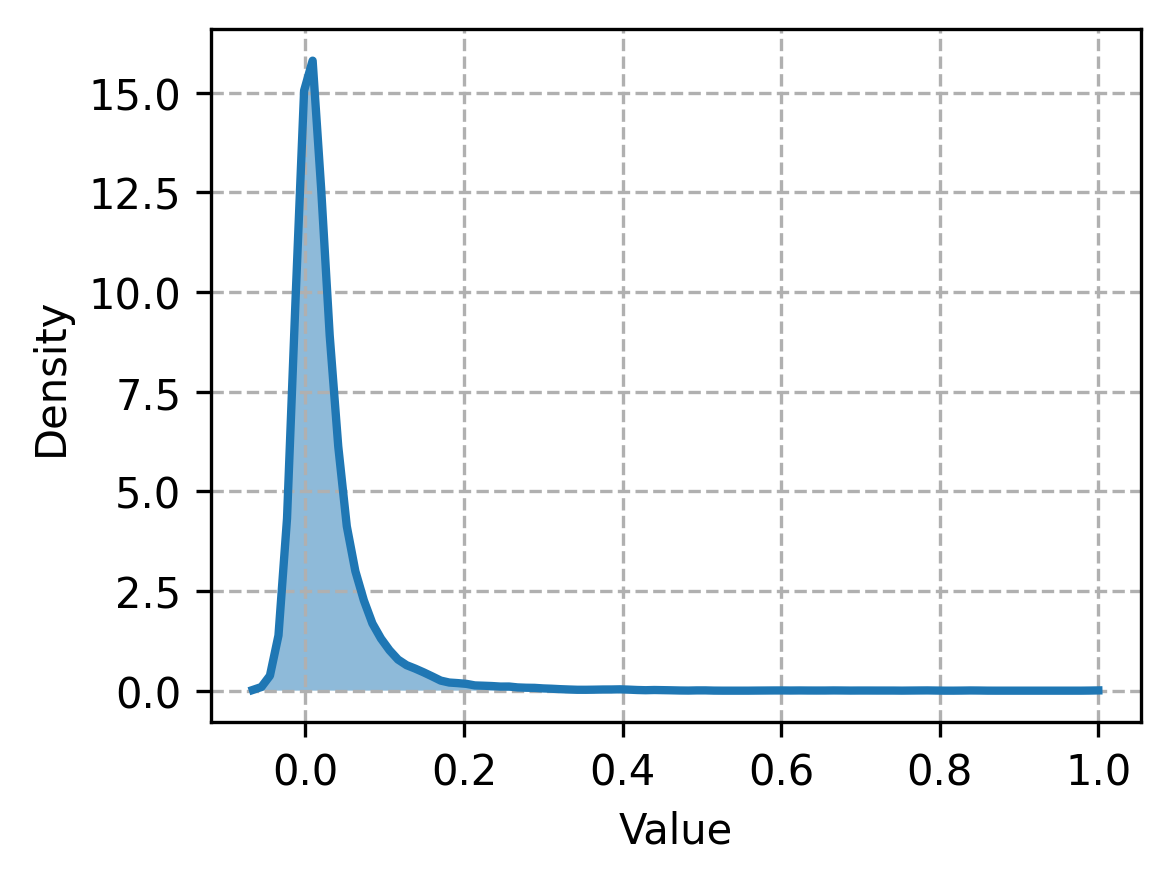}
    }
    \hfill
    \subfigure[Layer 14]{\includegraphics[width=0.22\textwidth]{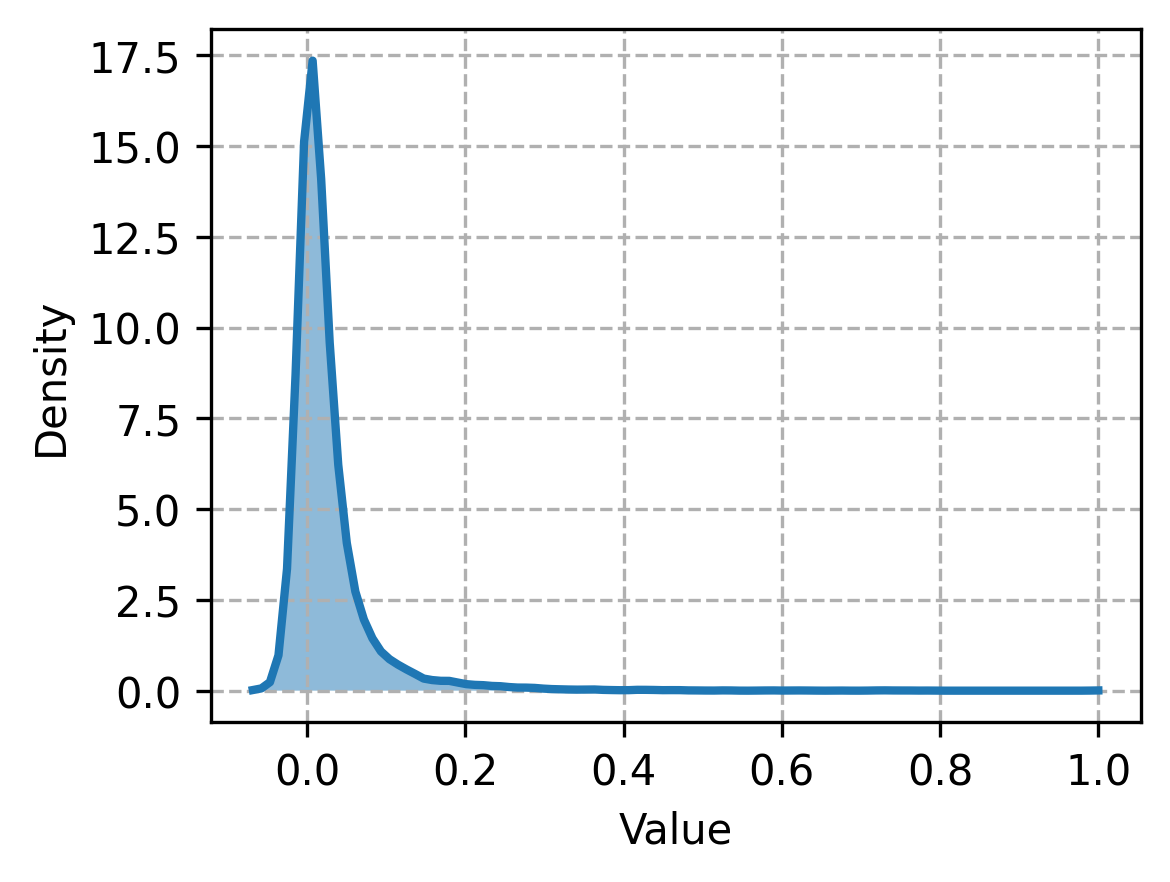}
    }
    \hfill
    \subfigure[Layer 15]{\includegraphics[width=0.22\textwidth]{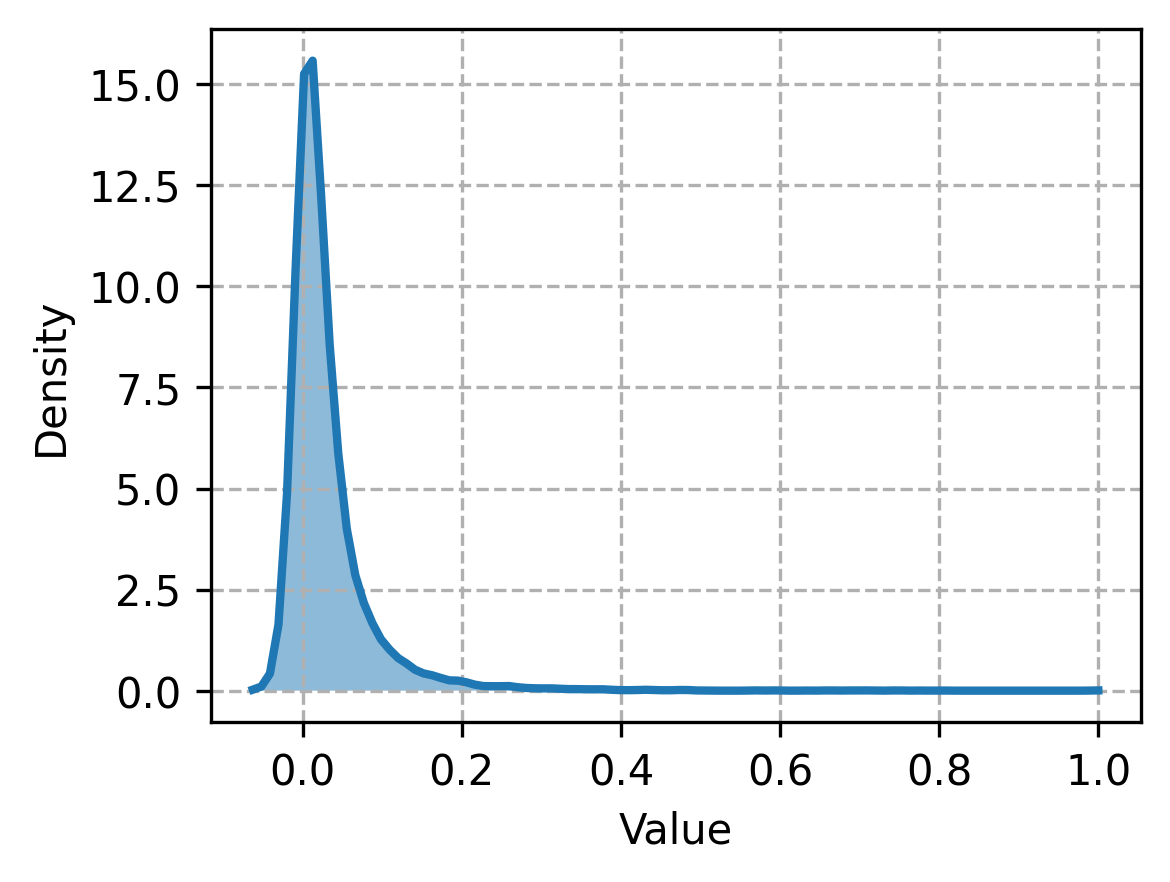}
    }
    \hfill
    \subfigure[Layer 16]{\includegraphics[width=0.22\textwidth]{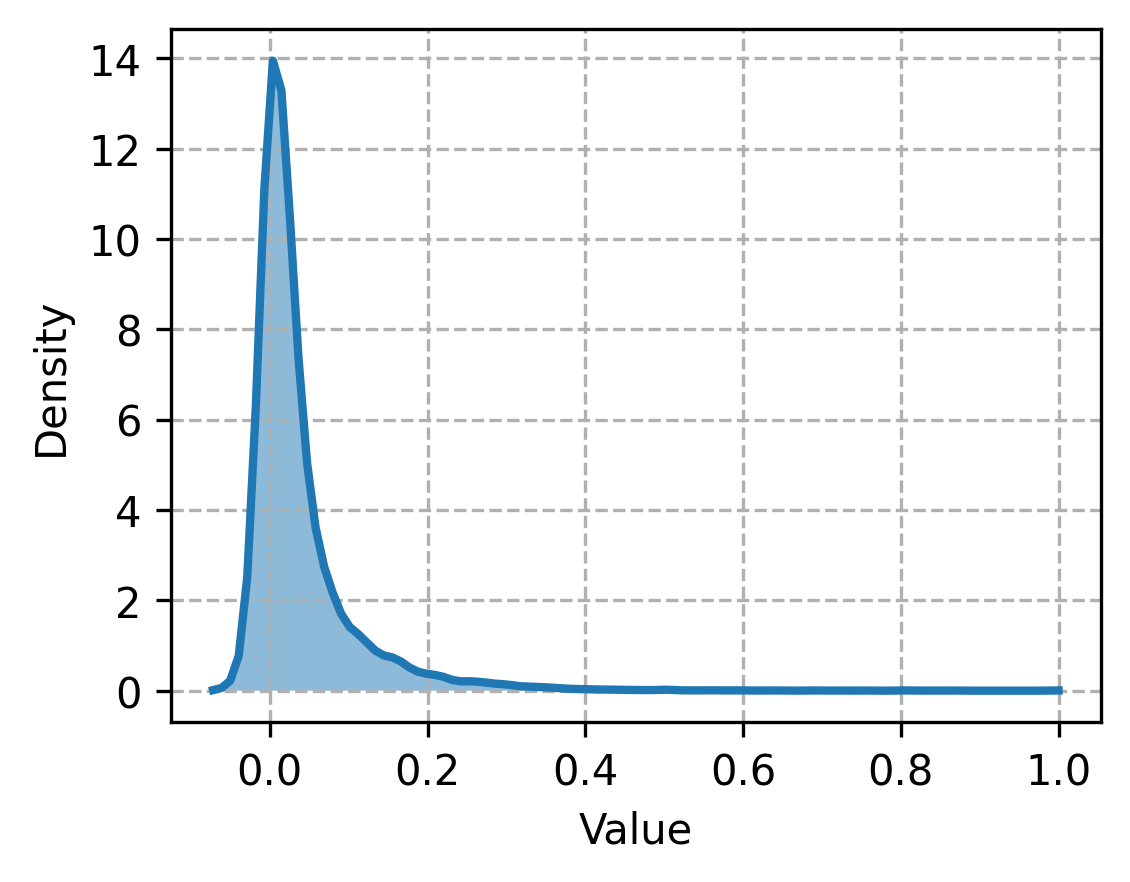}
    }
    \hfill
    \subfigure[Layer 17]{\includegraphics[width=0.22\textwidth]{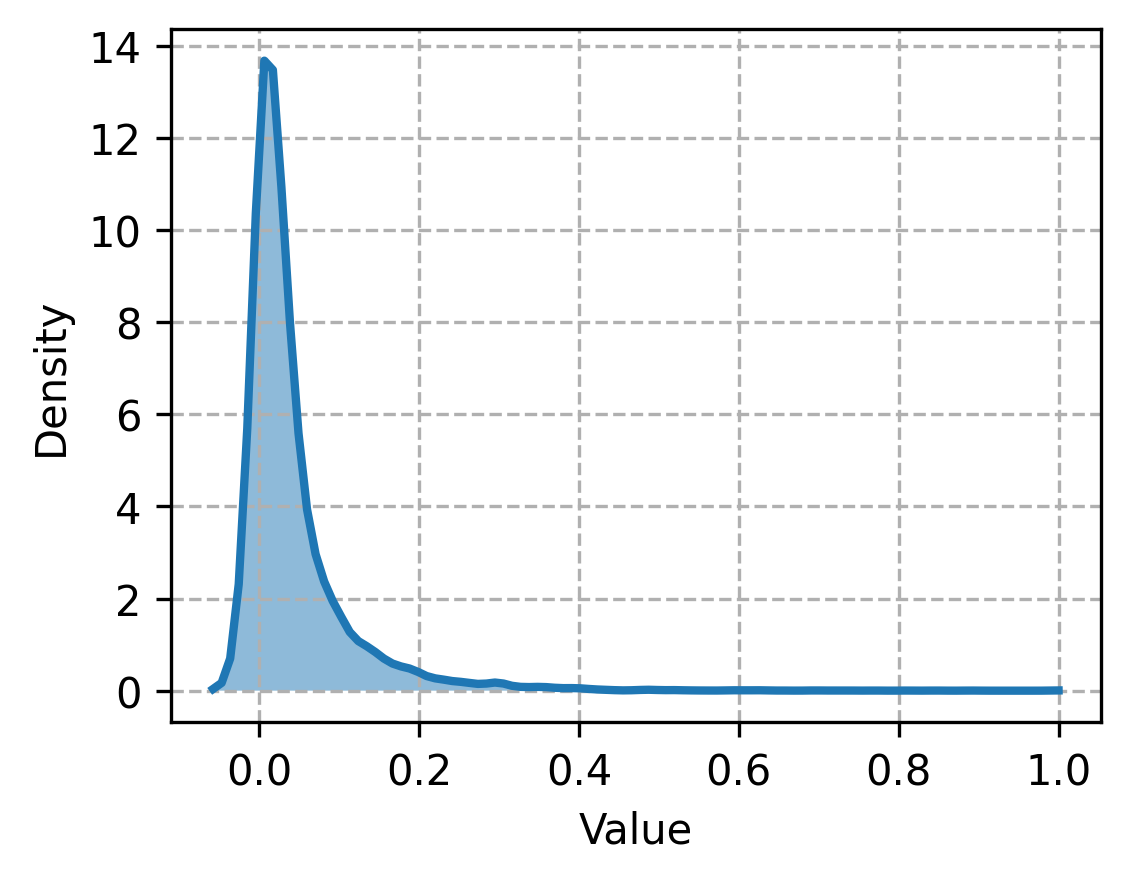}
    }
    \hfill
    \subfigure[Layer 18]{\includegraphics[width=0.22\textwidth]{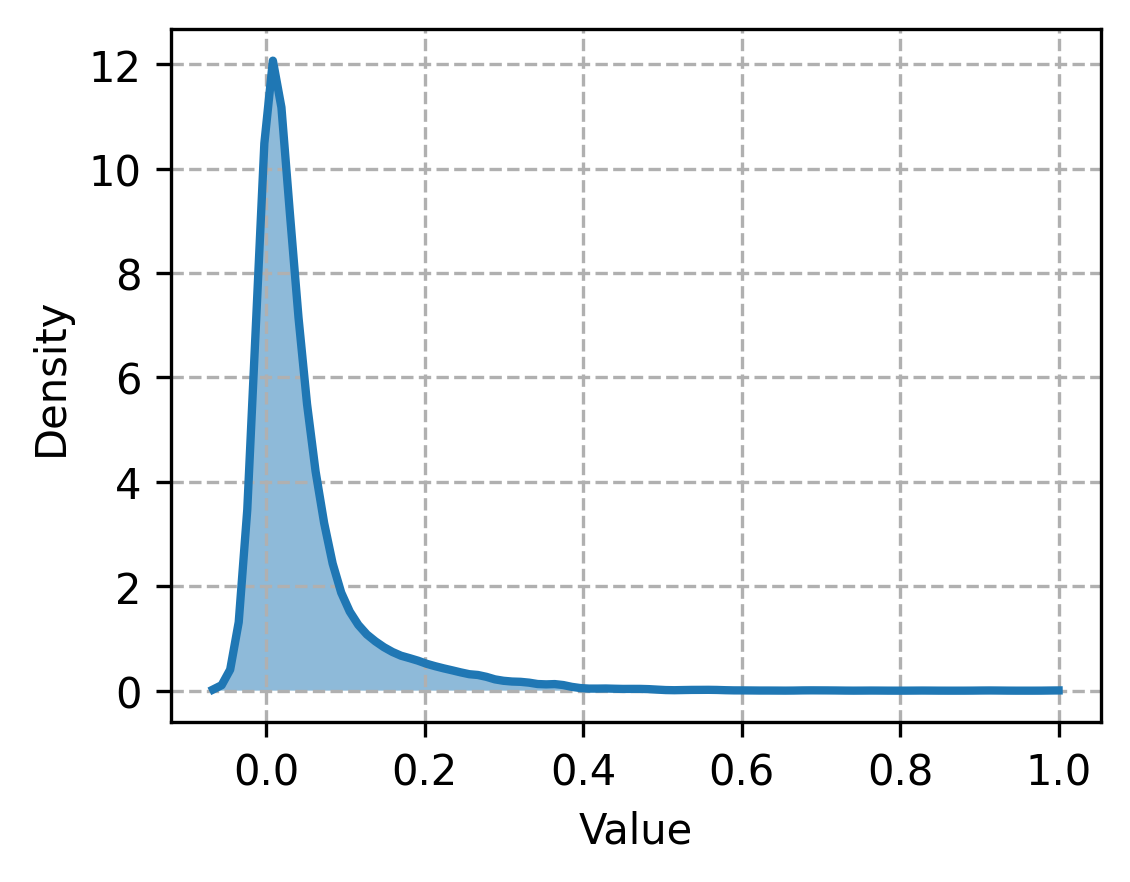}
    }
    \hfill
    \subfigure[Layer 19]{\includegraphics[width=0.22\textwidth]{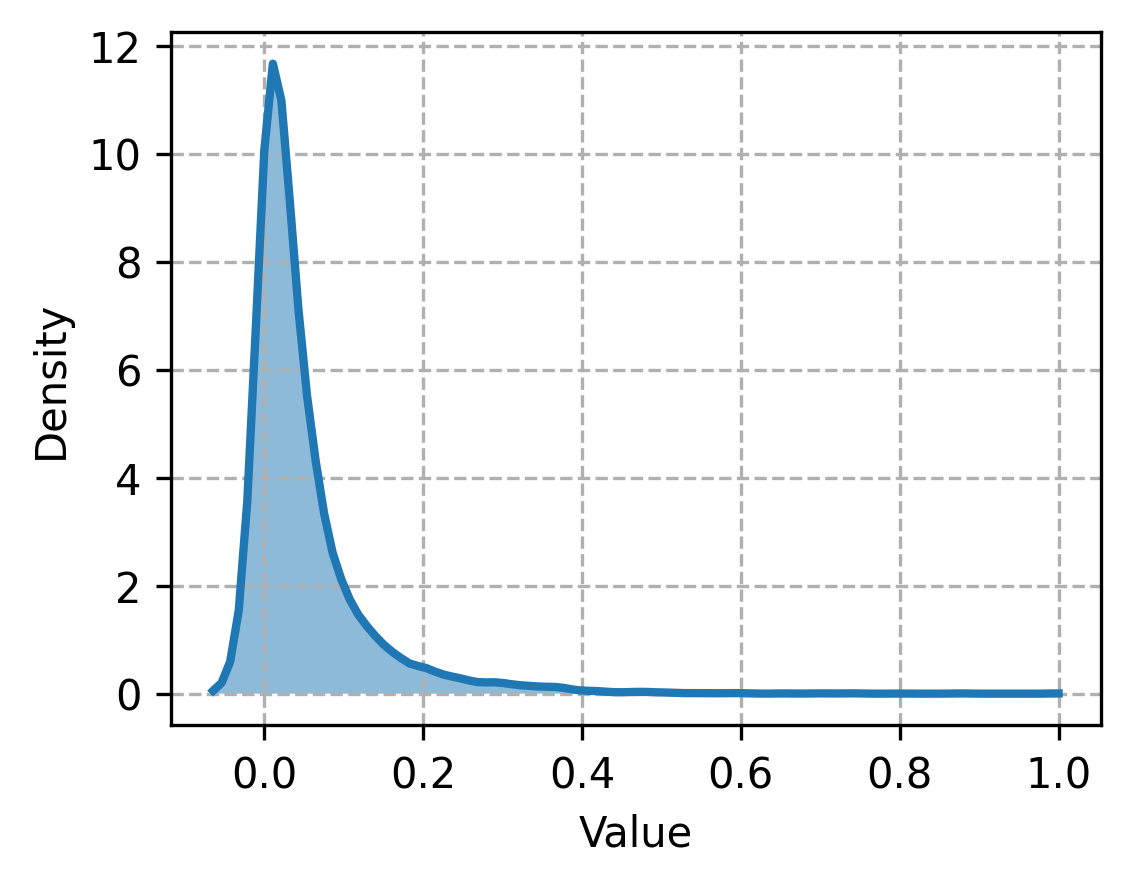}
    }
  \caption{In GPT2-Large, the KDE of elements in \(P\) matrices across layers (0-19).}
  \label{fig:superposition_KDE_gpt2-large-part1}
\end{figure*}

\begin{figure*}
    \centering
    \subfigure[Layer 20]{\includegraphics[width=0.22\textwidth]{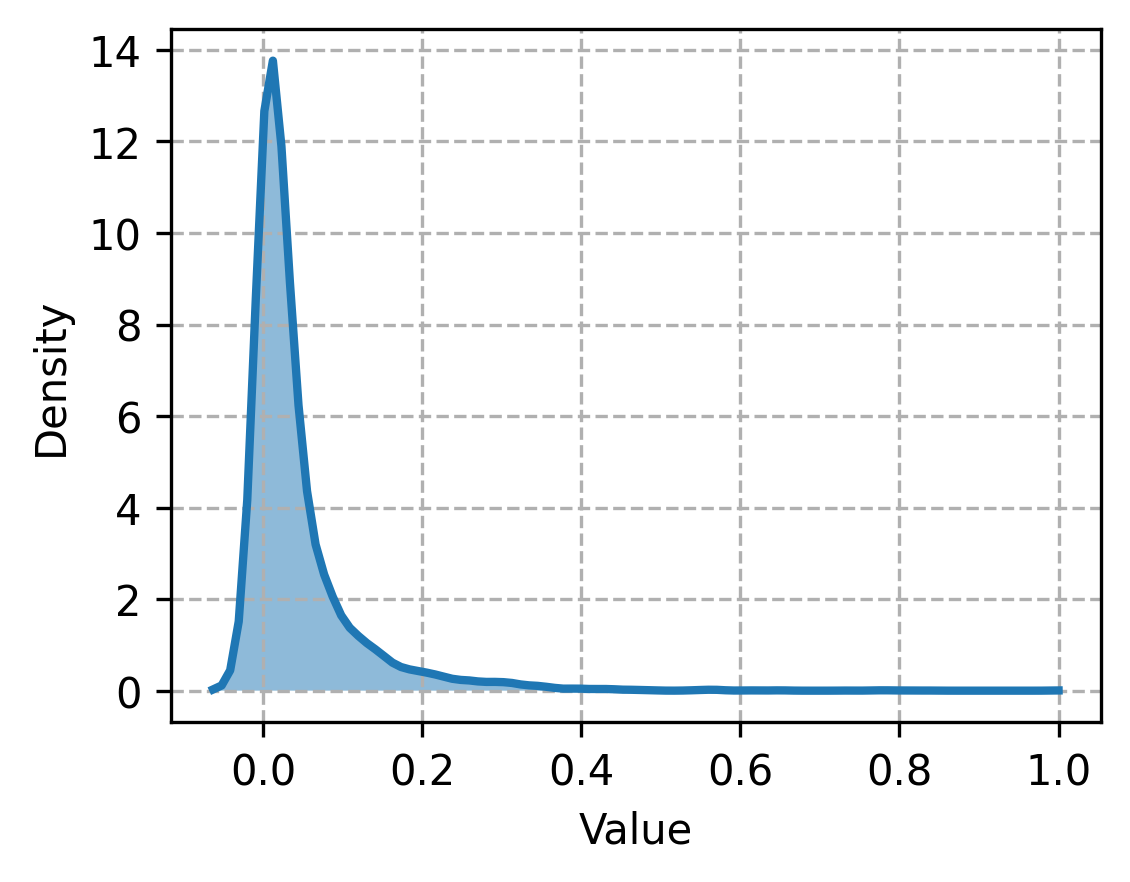}
    }
    \hfill
    \subfigure[Layer 21]{\includegraphics[width=0.22\textwidth]{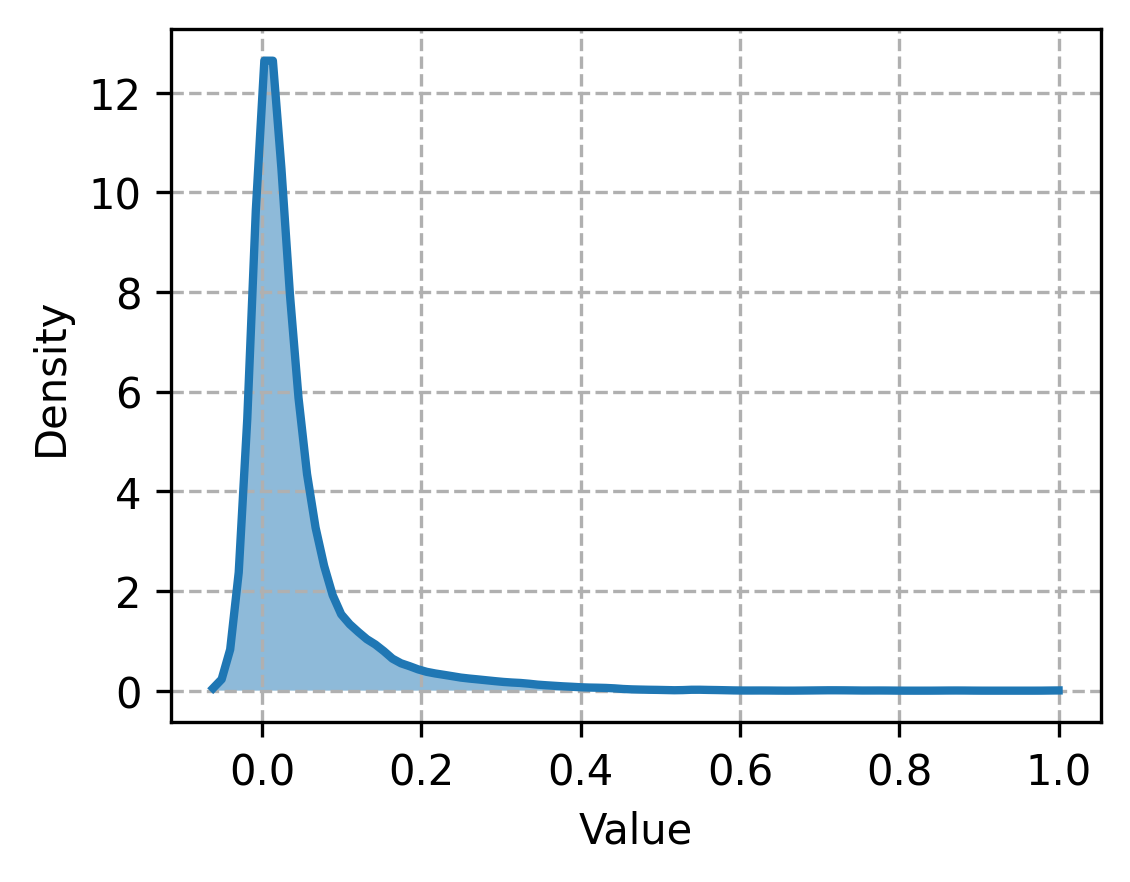}
    }
    \hfill
    \subfigure[Layer 22]{\includegraphics[width=0.22\textwidth]{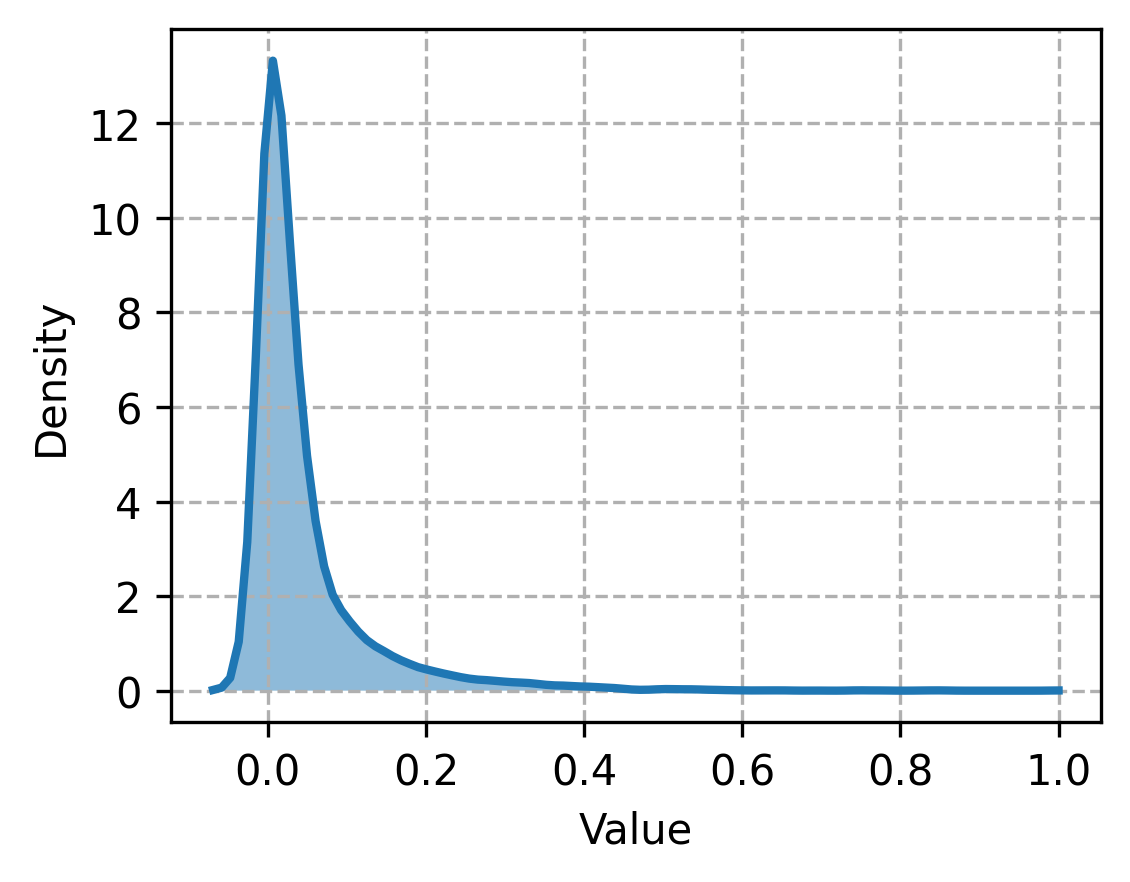}
    }
    \hfill
    \subfigure[Layer 23]{\includegraphics[width=0.22\textwidth]{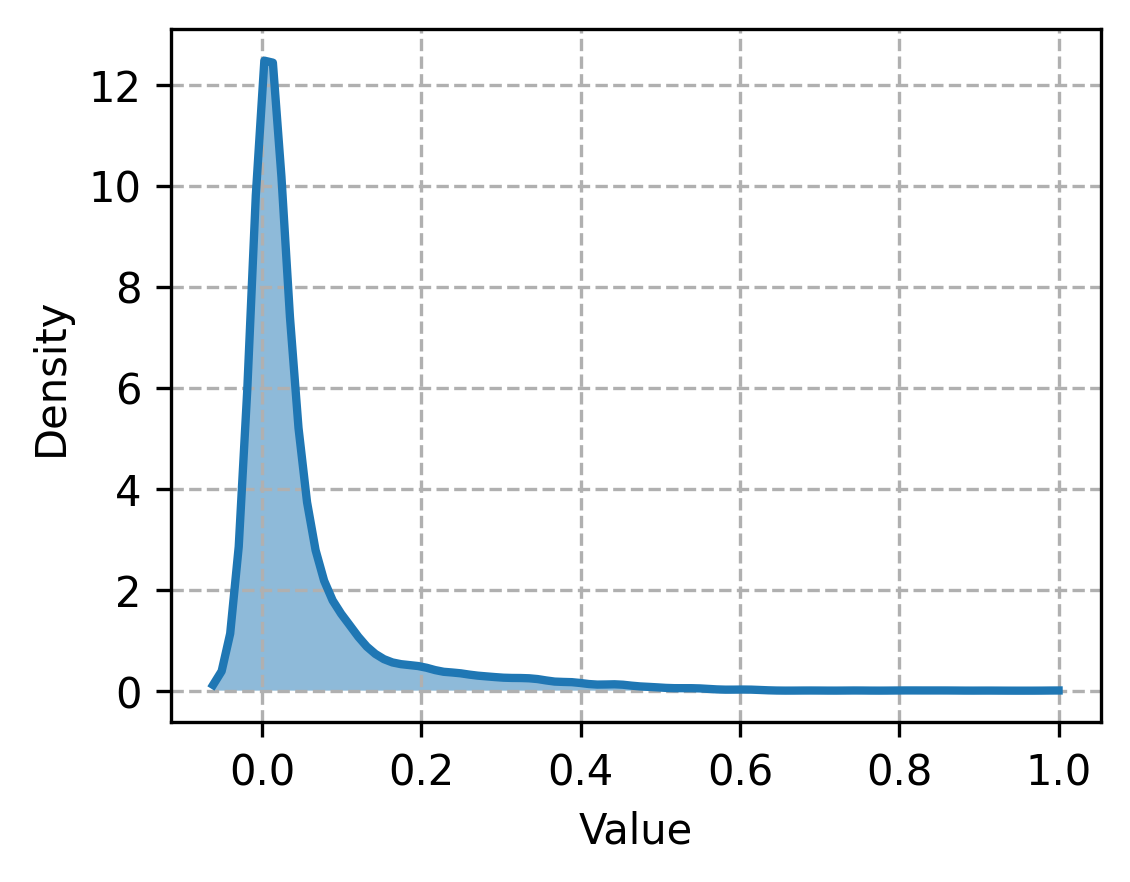}
    }
    \hfill
    \subfigure[Layer 24]{\includegraphics[width=0.22\textwidth]{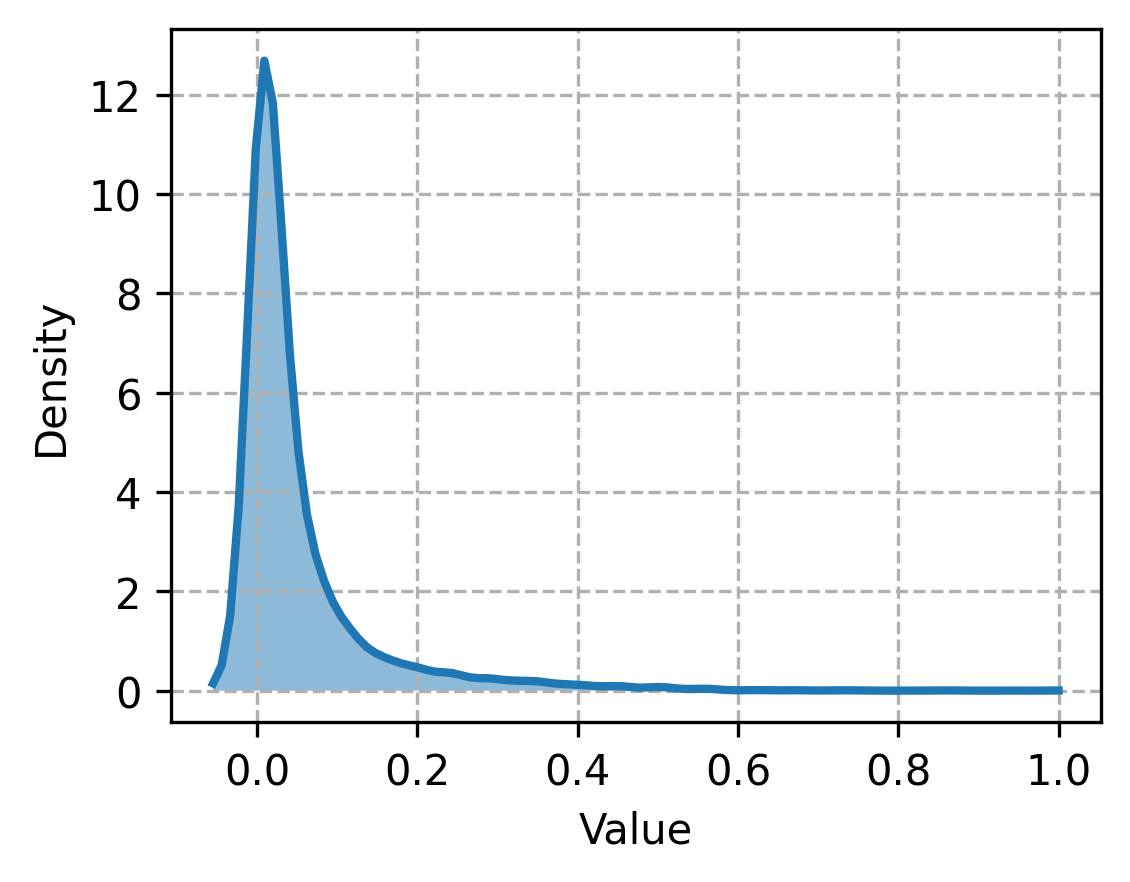}
    }
    \hfill
    \subfigure[Layer 25]{\includegraphics[width=0.22\textwidth]{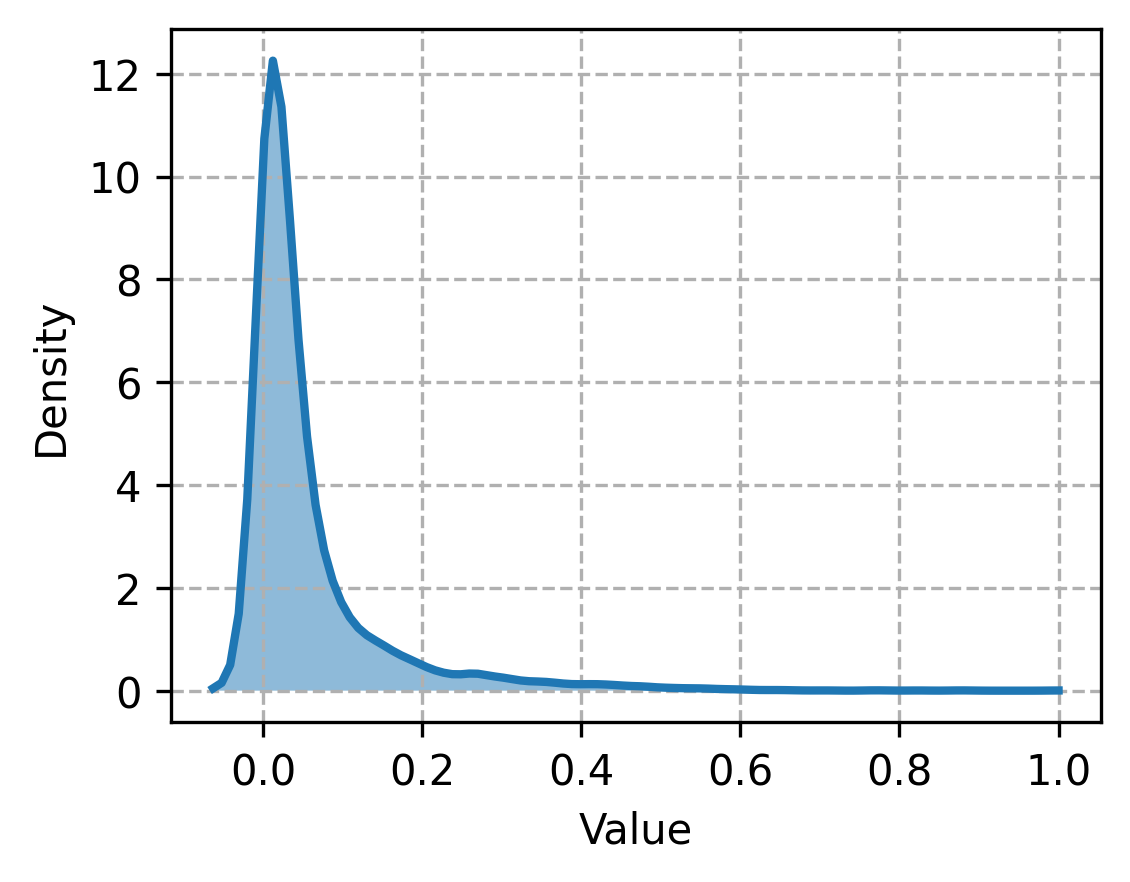}
    }
    \hfill
    \subfigure[Layer 26]{\includegraphics[width=0.22\textwidth]{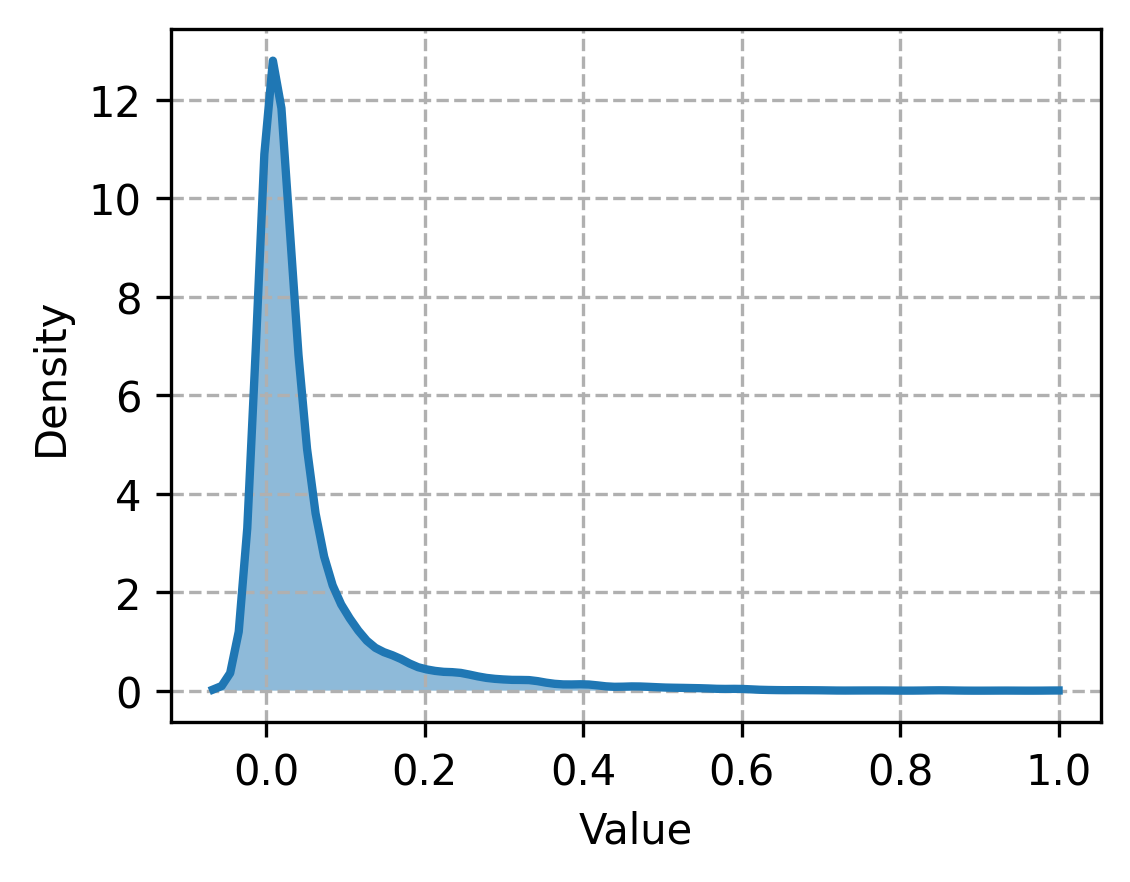}
    }
    \hfill
    \subfigure[Layer 27]{\includegraphics[width=0.22\textwidth]{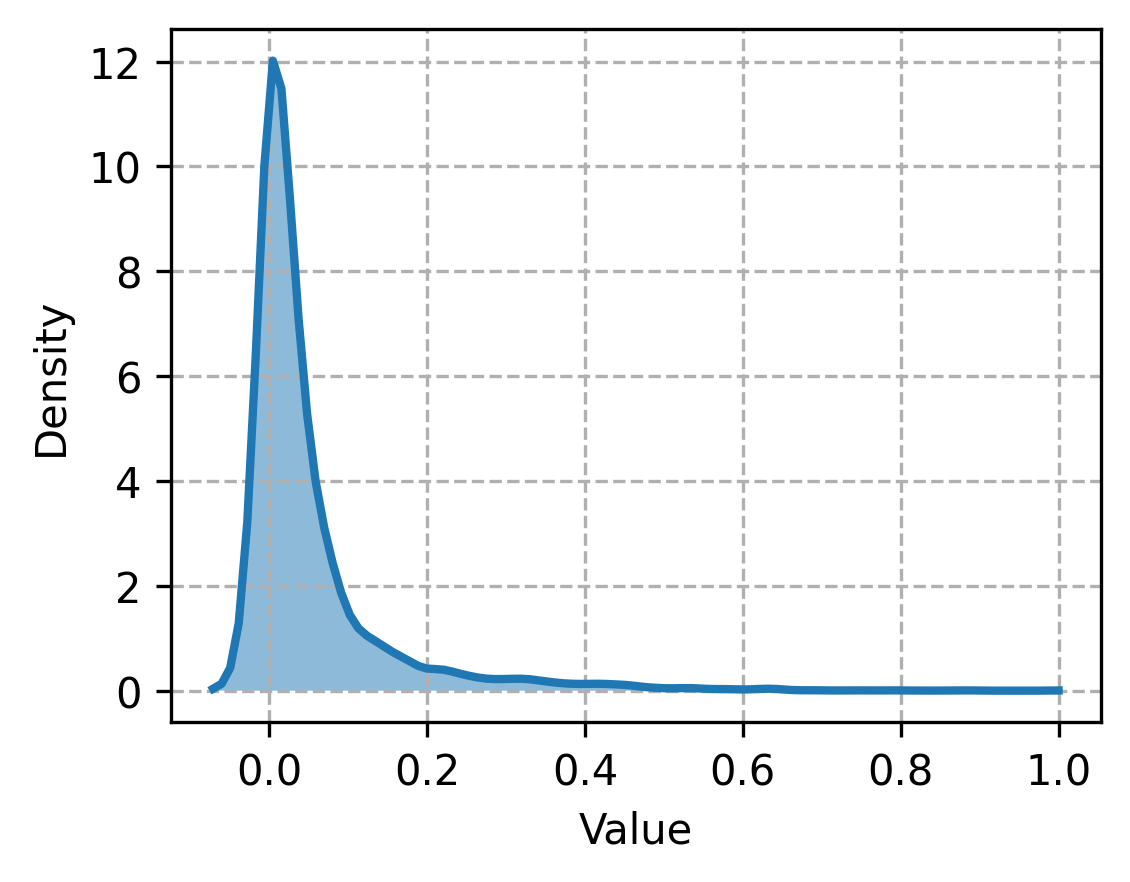}
    }
    \hfill
    \subfigure[Layer 28]{\includegraphics[width=0.22\textwidth]{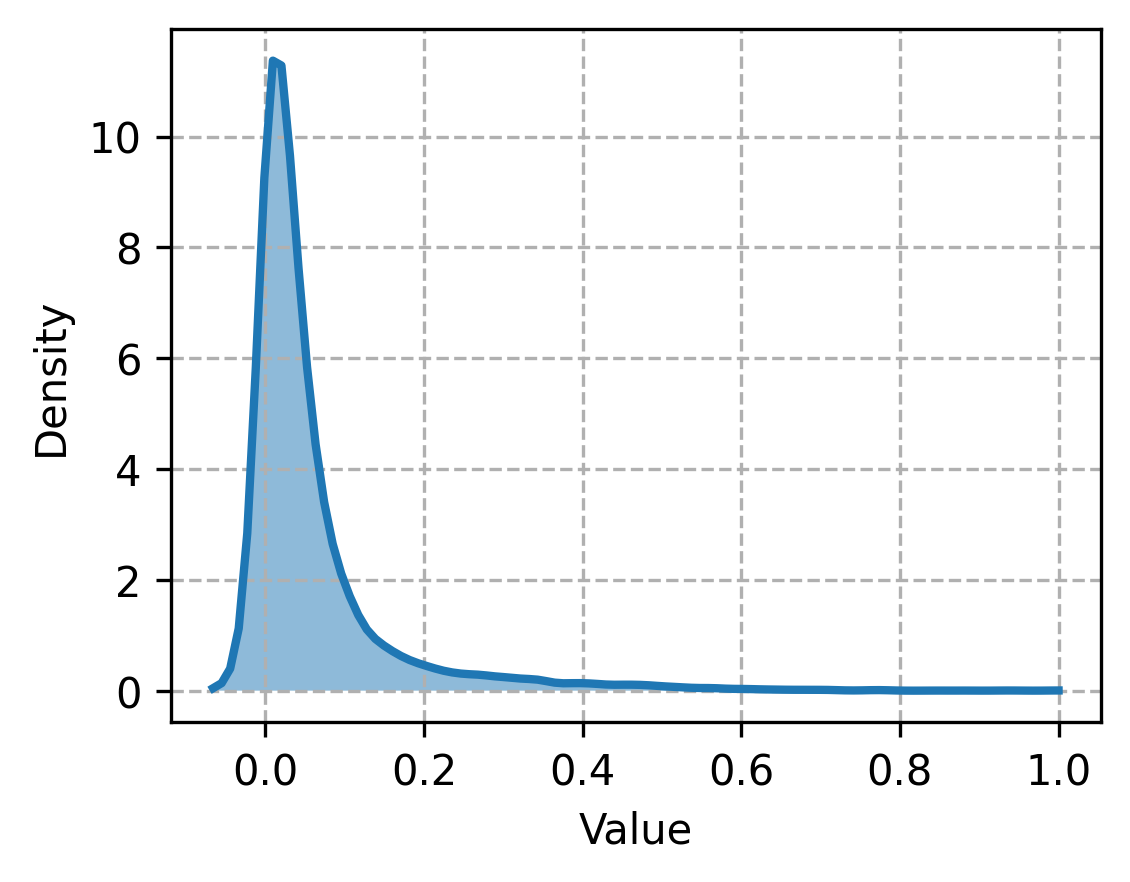}
    }
    \hfill
    \subfigure[Layer 29]{\includegraphics[width=0.22\textwidth]{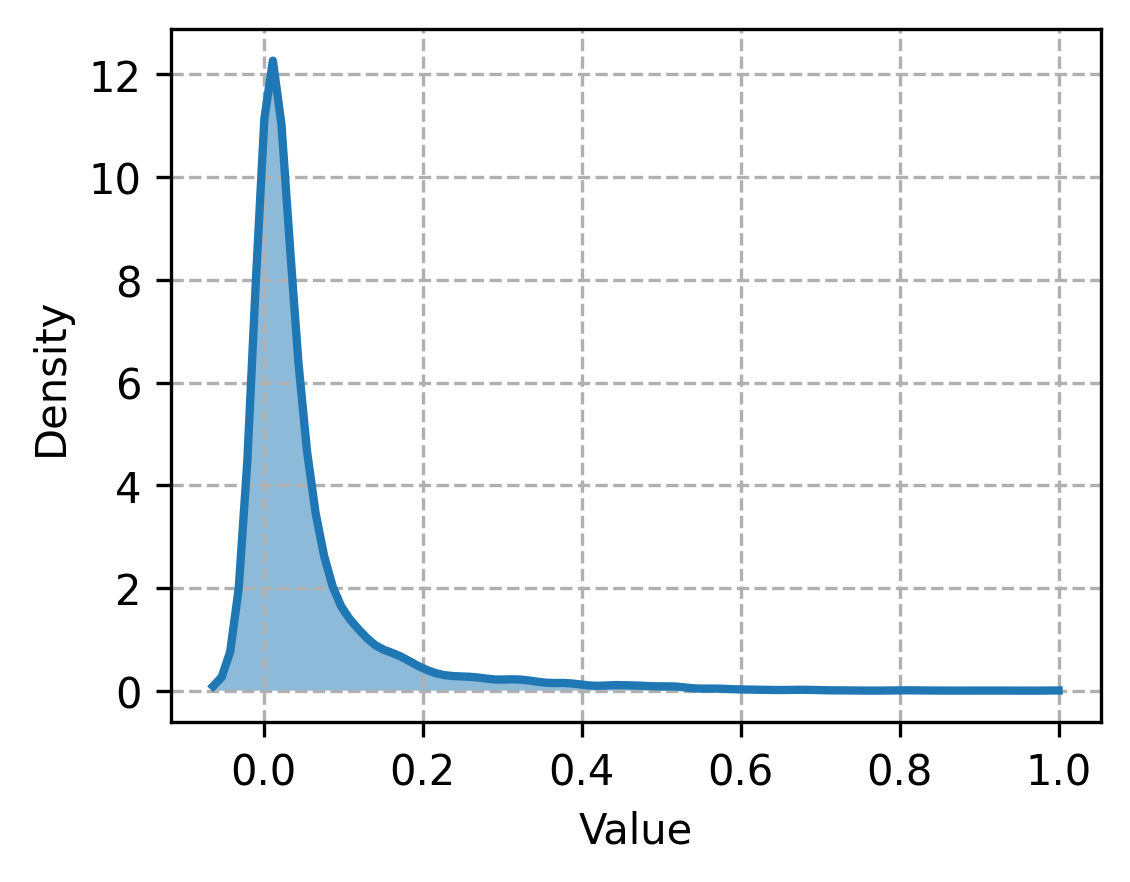}
    }
    \hfill
    \subfigure[Layer 30]{\includegraphics[width=0.22\textwidth]{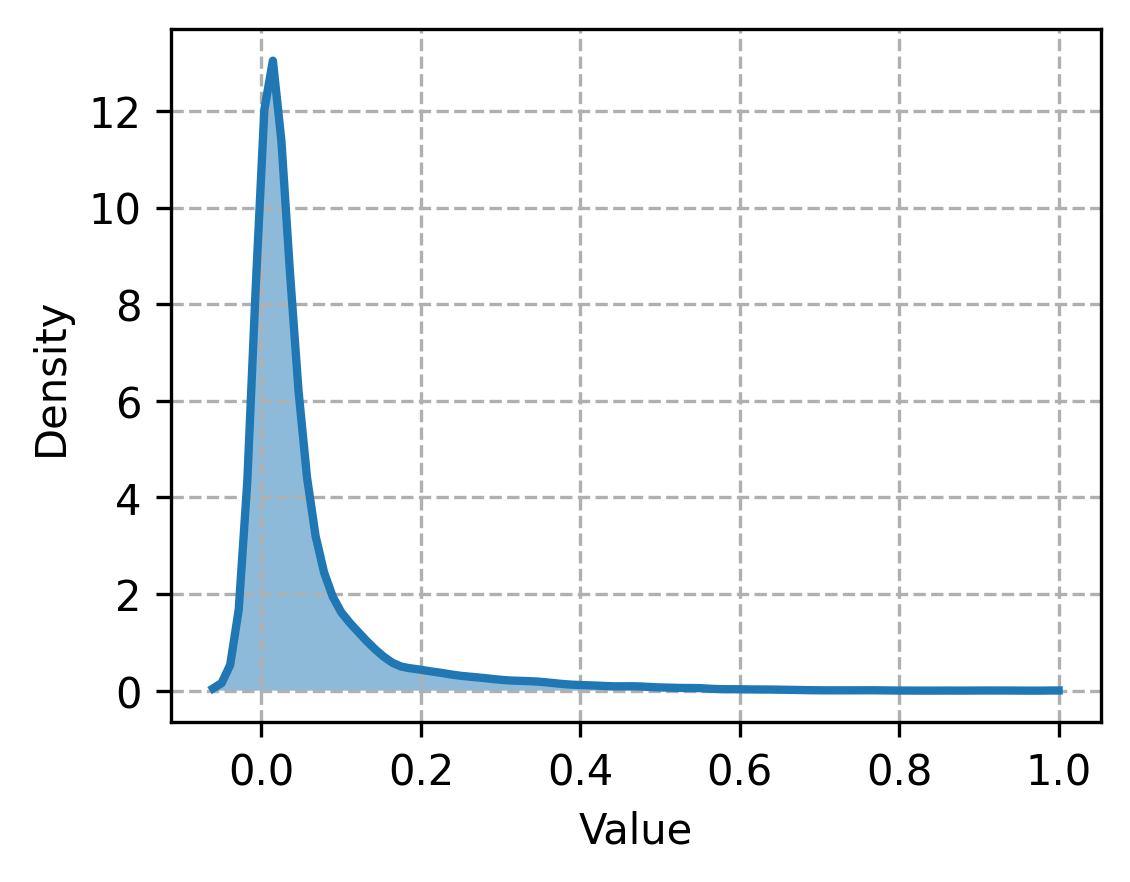}
    }
    \hfill
    \subfigure[Layer 31]{\includegraphics[width=0.22\textwidth]{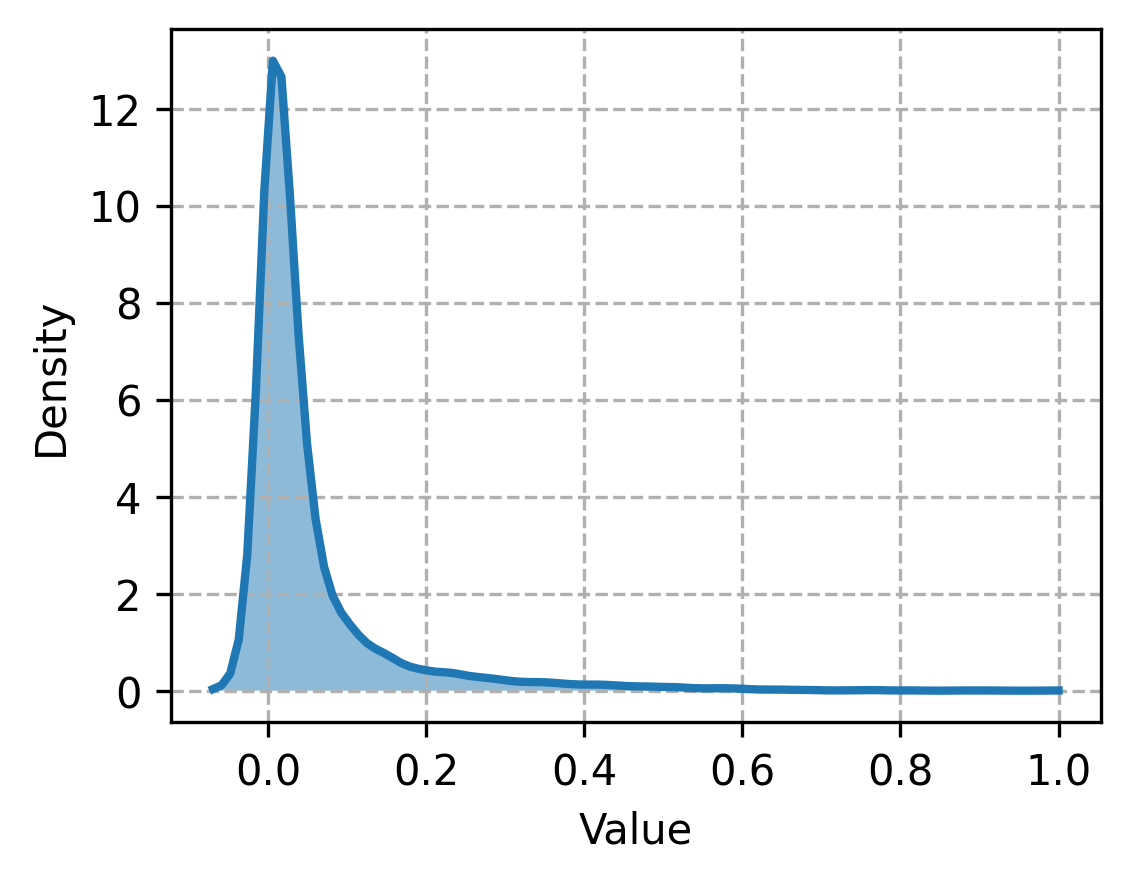}
    }
    \hfill
    \subfigure[Layer 32]{\includegraphics[width=0.22\textwidth]{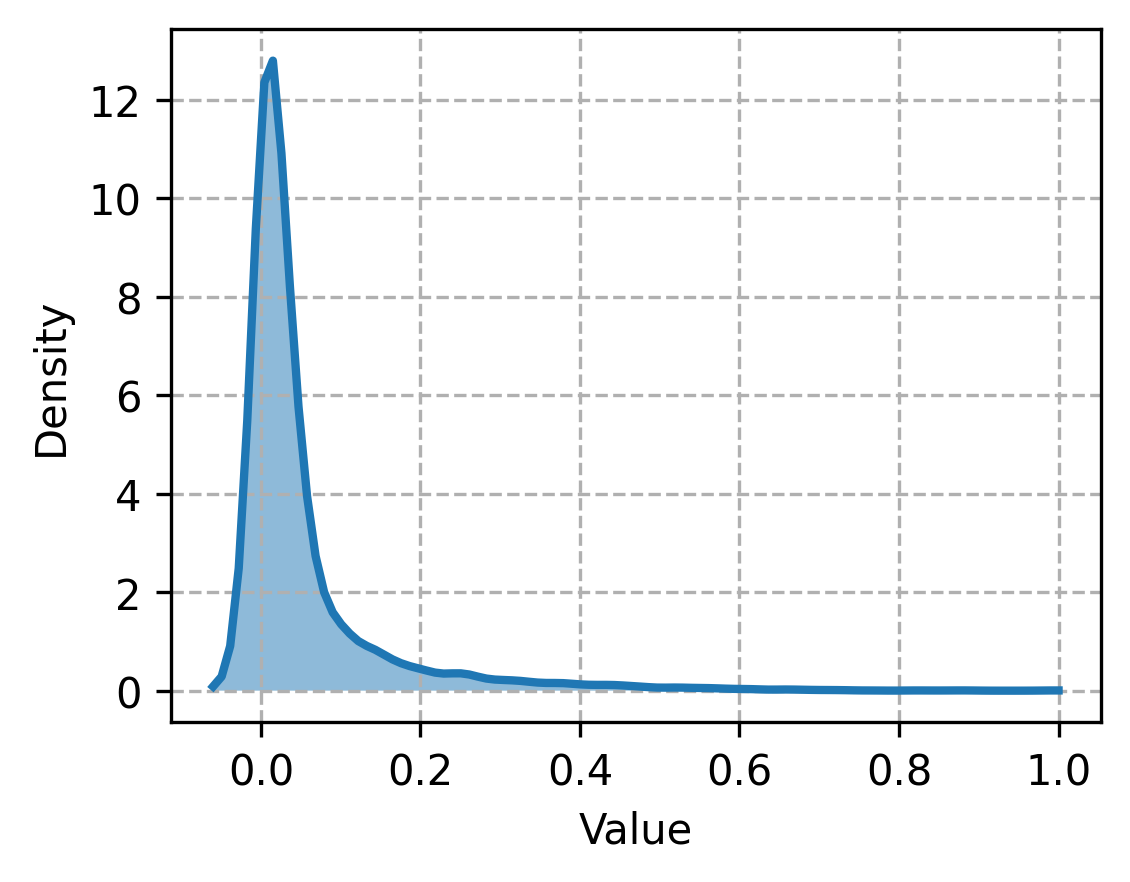}
    }
    \hfill
    \subfigure[Layer 33]{\includegraphics[width=0.22\textwidth]{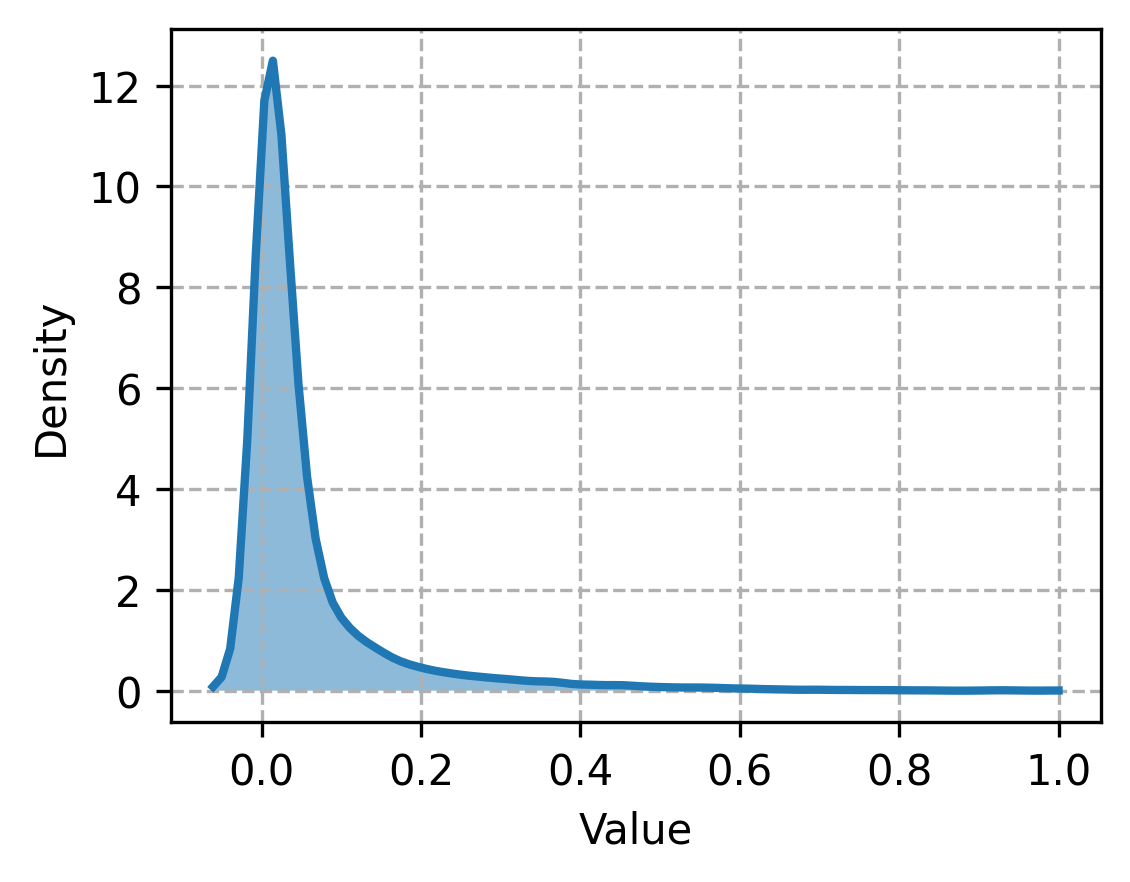}
    }
    \hfill
    \subfigure[Layer 34]{\includegraphics[width=0.22\textwidth]{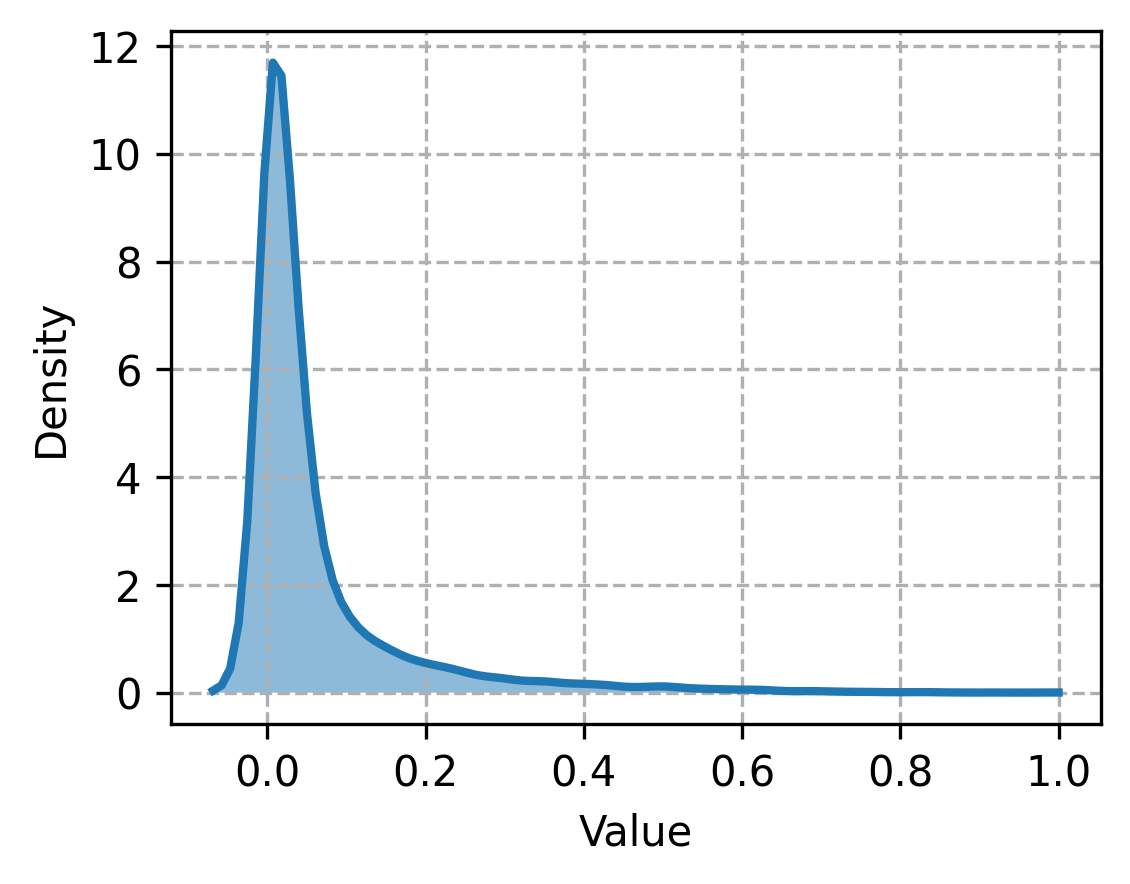}
    }
    \hfill
    \subfigure[Layer 35]{\includegraphics[width=0.22\textwidth]{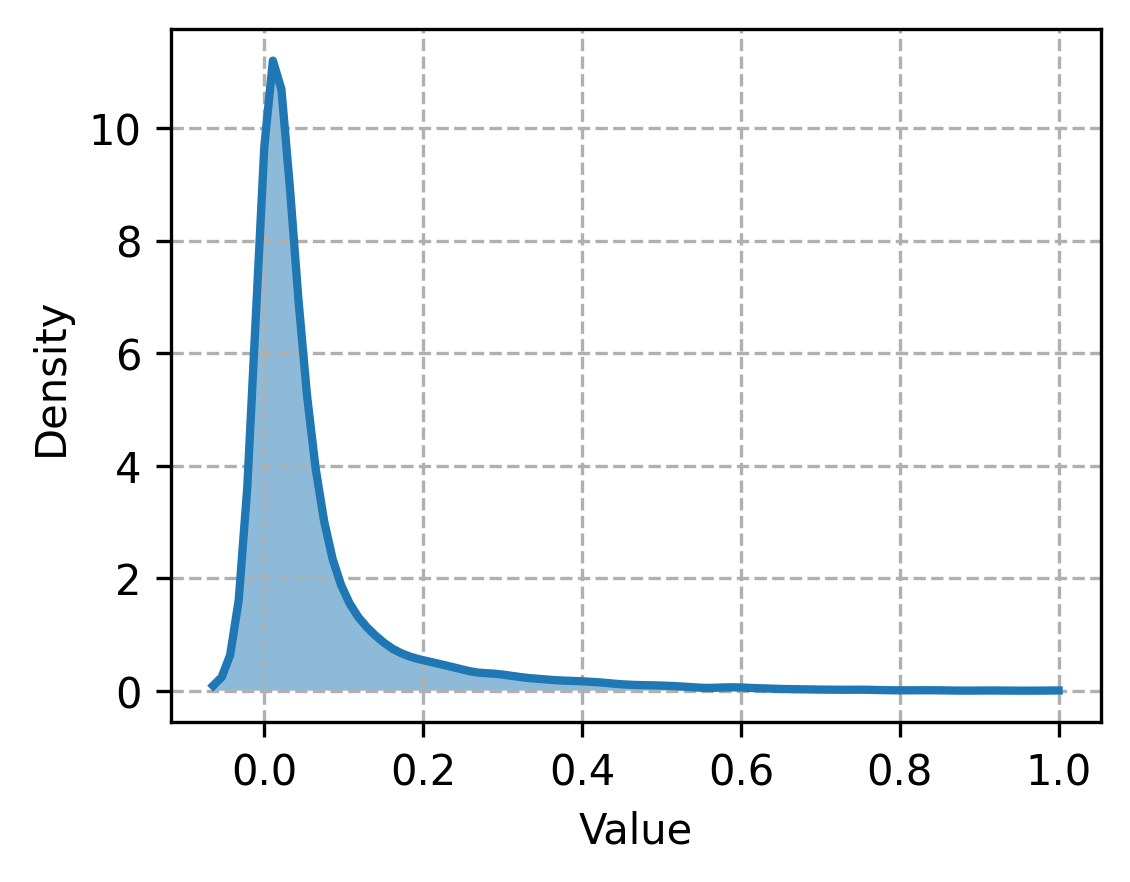}
    }
  \caption{In GPT2-Large, the KDE of elements in \(P\) matrices across layers (20-35).}
  \label{fig:superposition_KDE_gpt2-large-part2}
\end{figure*}

\begin{figure*}
    \centering
    \subfigure[Layer 0]{\includegraphics[width=0.22\textwidth]{fig/EleutherAI/gpt-j-6B/p_matrix/known/KDE/superposition_for_layer_0.png}
    }
    \hfill
    \subfigure[Layer 1]{\includegraphics[width=0.22\textwidth]{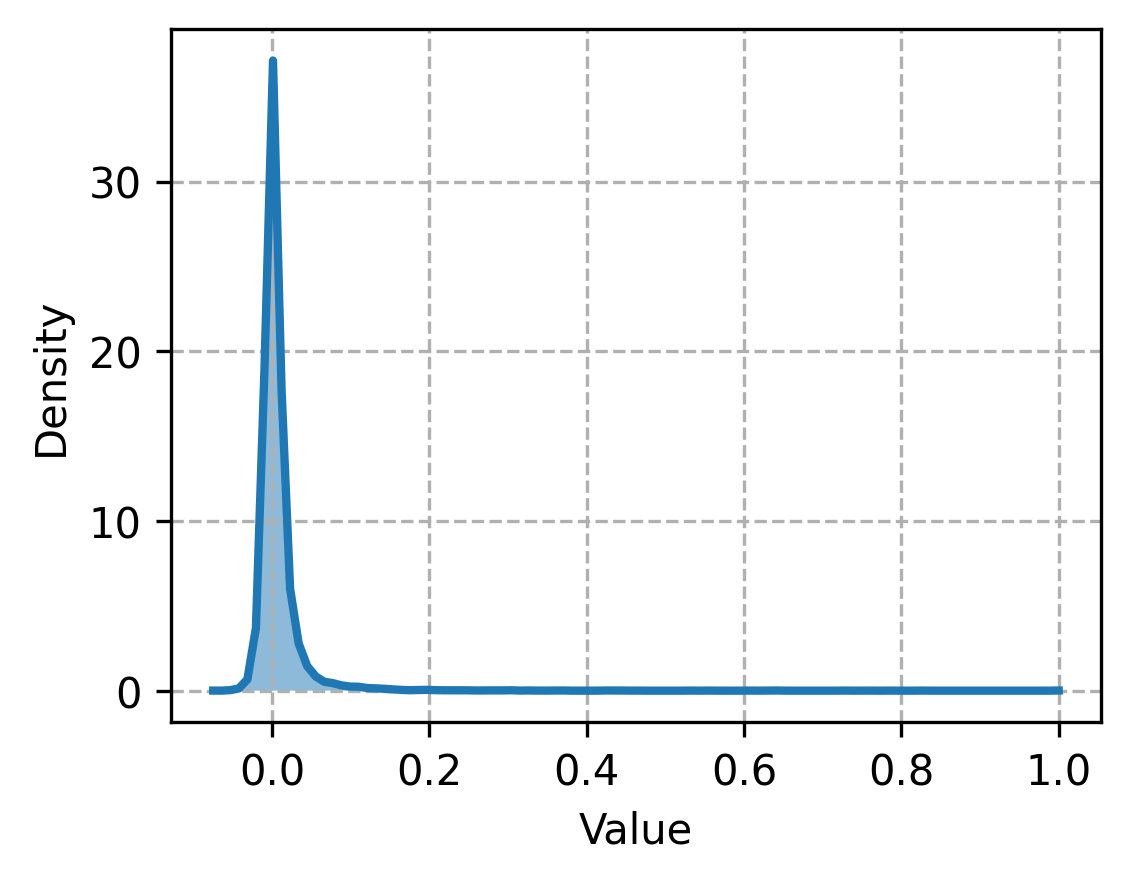}
    }
    \hfill
    \subfigure[Layer 2]{\includegraphics[width=0.22\textwidth]{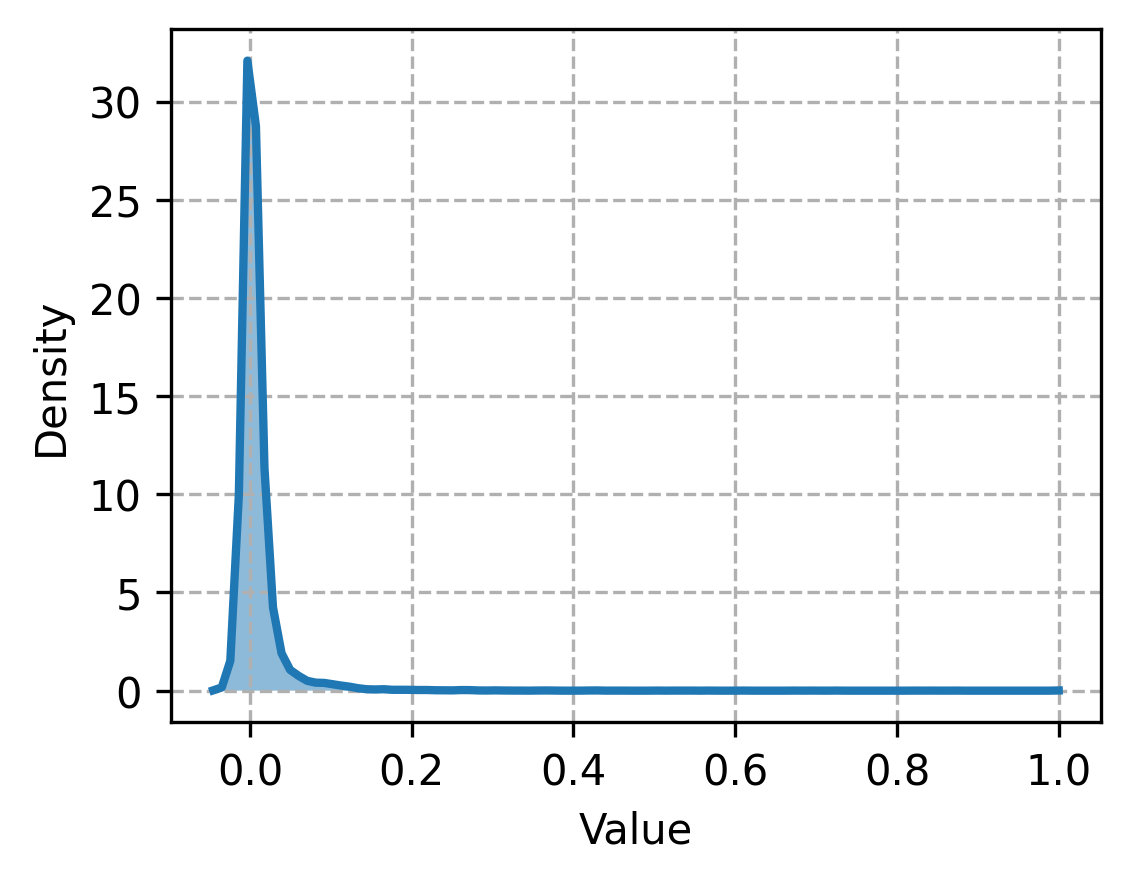}
    }
    \hfill
    \subfigure[Layer 3]{\includegraphics[width=0.22\textwidth]{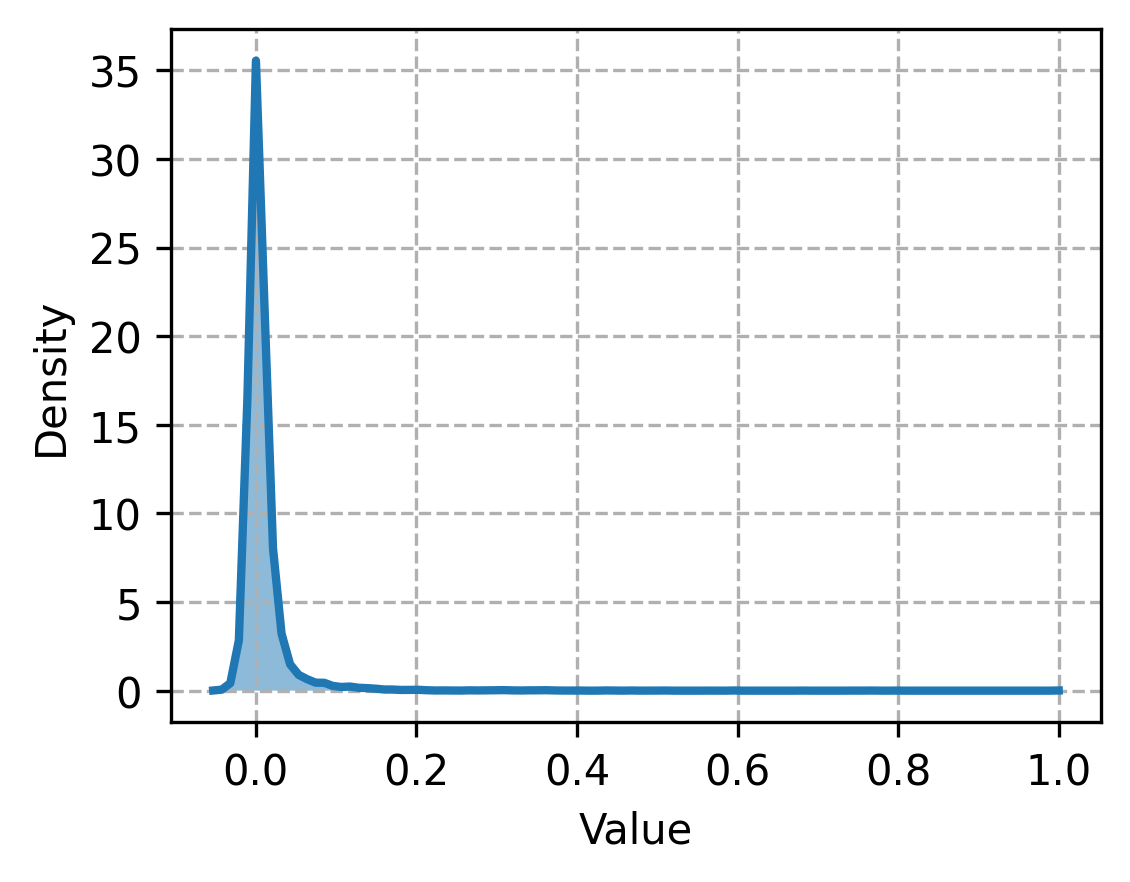}
    }
    \hfill
    \subfigure[Layer 4]{\includegraphics[width=0.22\textwidth]{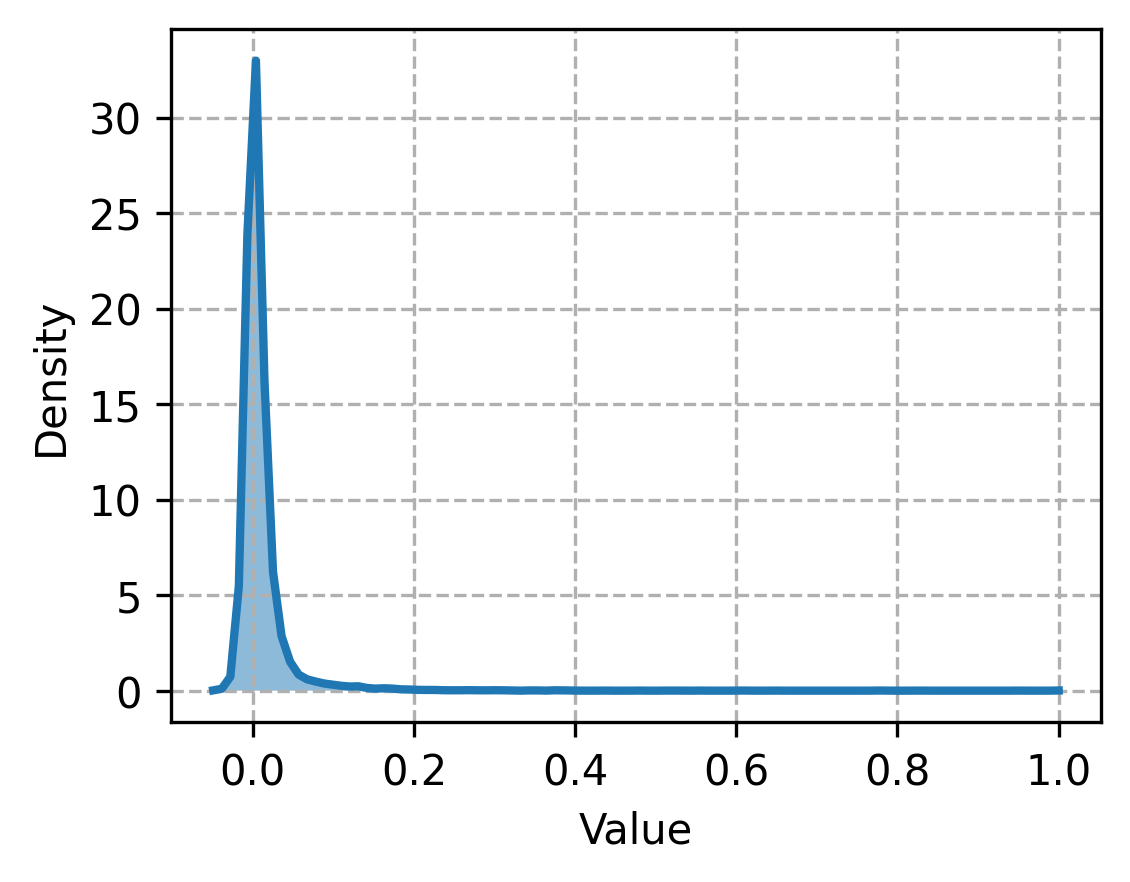}
    }
    \hfill
    \subfigure[Layer 5]{\includegraphics[width=0.22\textwidth]{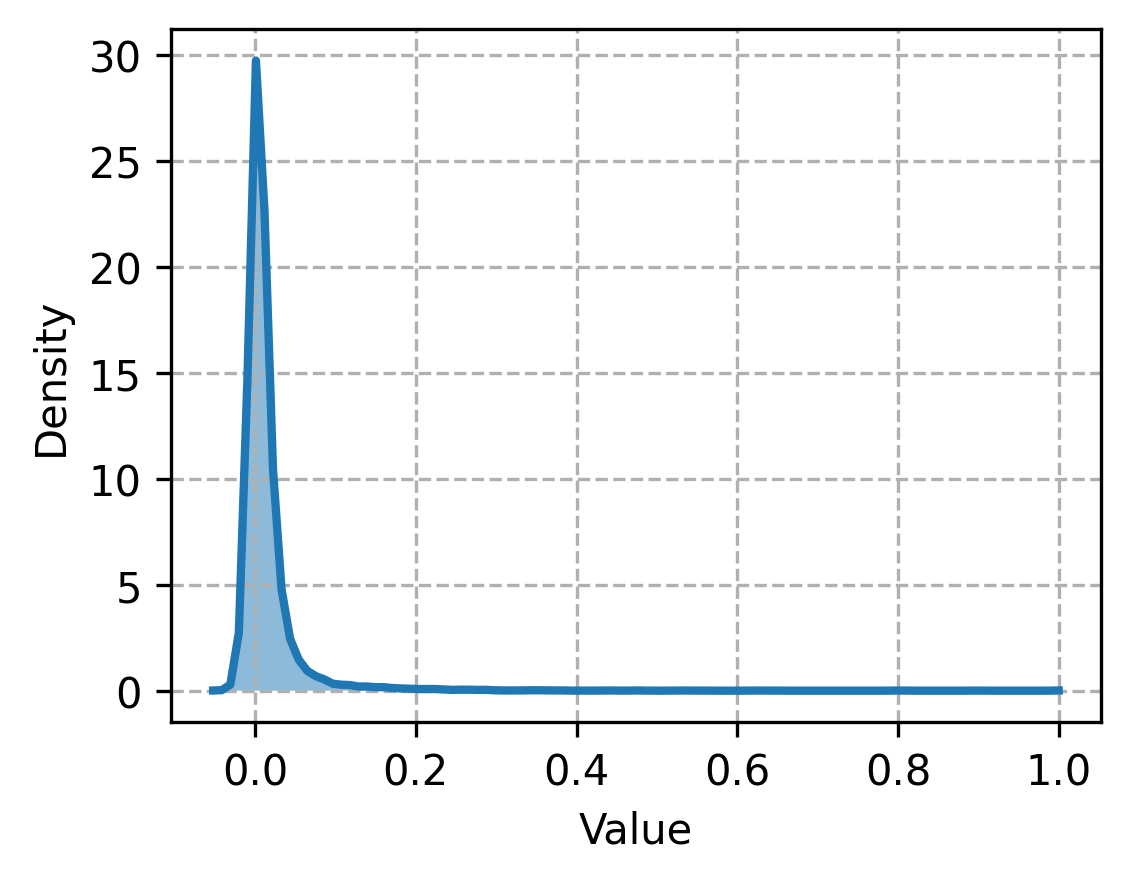}
    }
    \hfill
    \subfigure[Layer 6]{\includegraphics[width=0.22\textwidth]{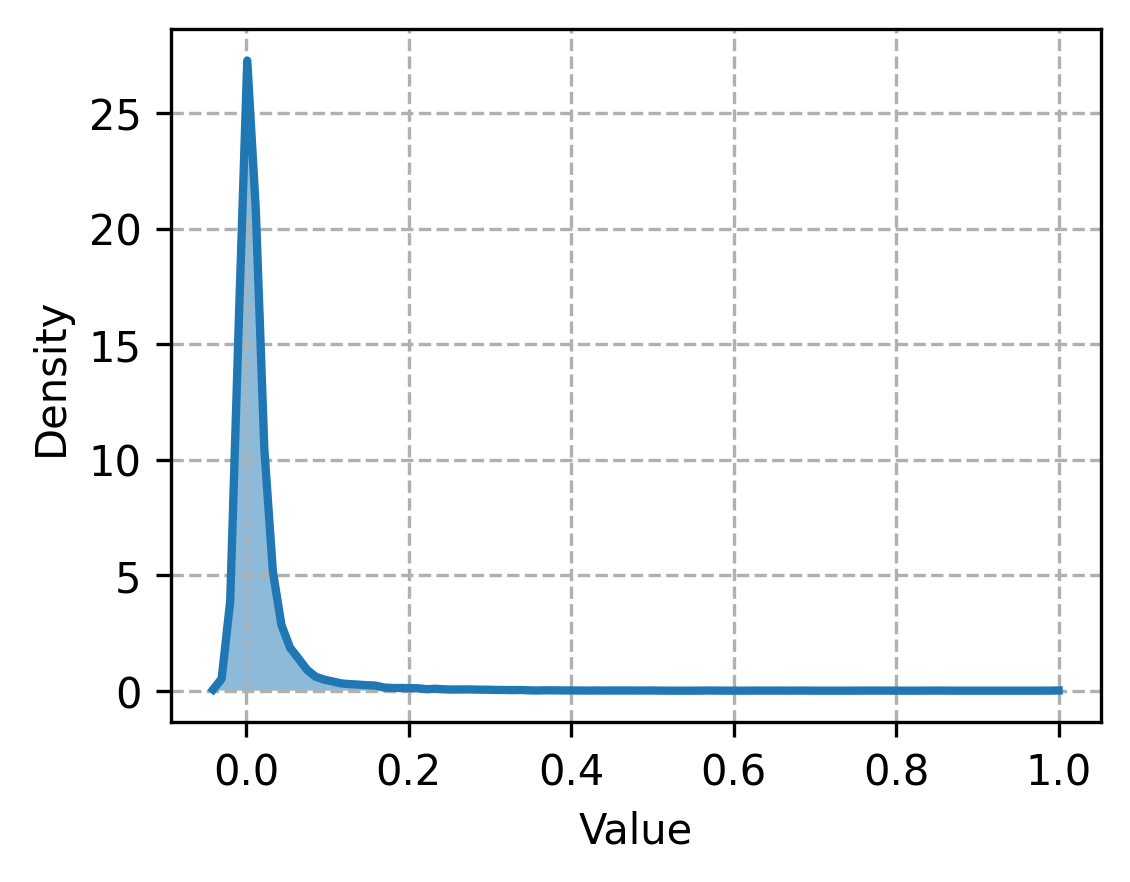}
    }
    \hfill
    \subfigure[Layer 7]{\includegraphics[width=0.22\textwidth]{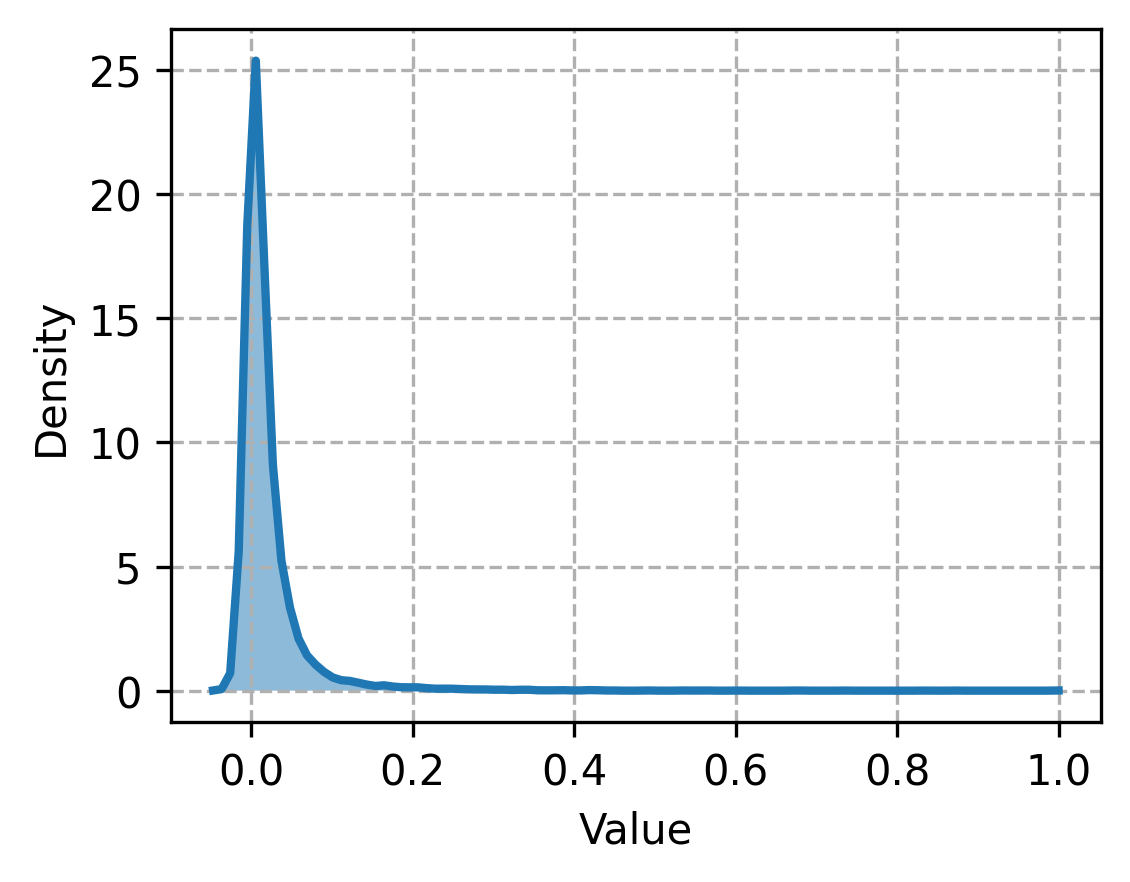}
    }
    \hfill
    \subfigure[Layer 8]{\includegraphics[width=0.22\textwidth]{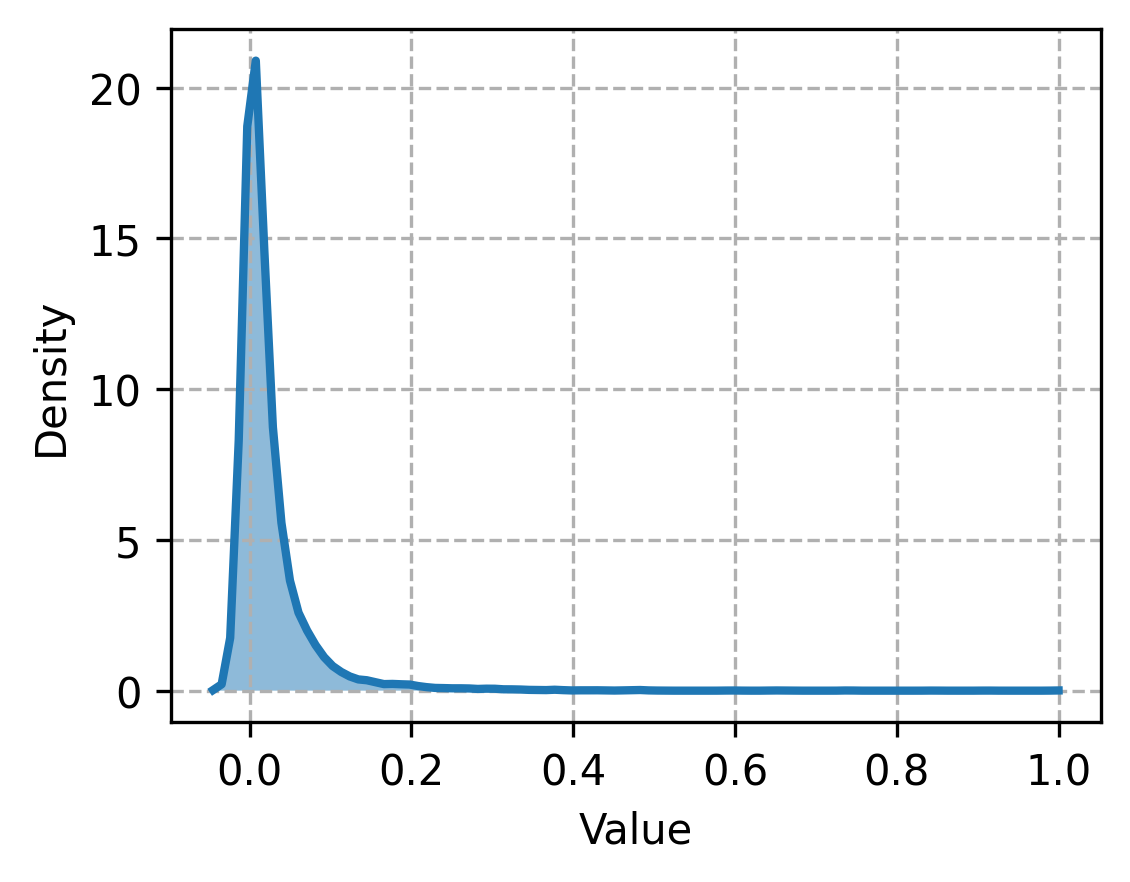}
    }
    \hfill
    \subfigure[Layer 9]{\includegraphics[width=0.22\textwidth]{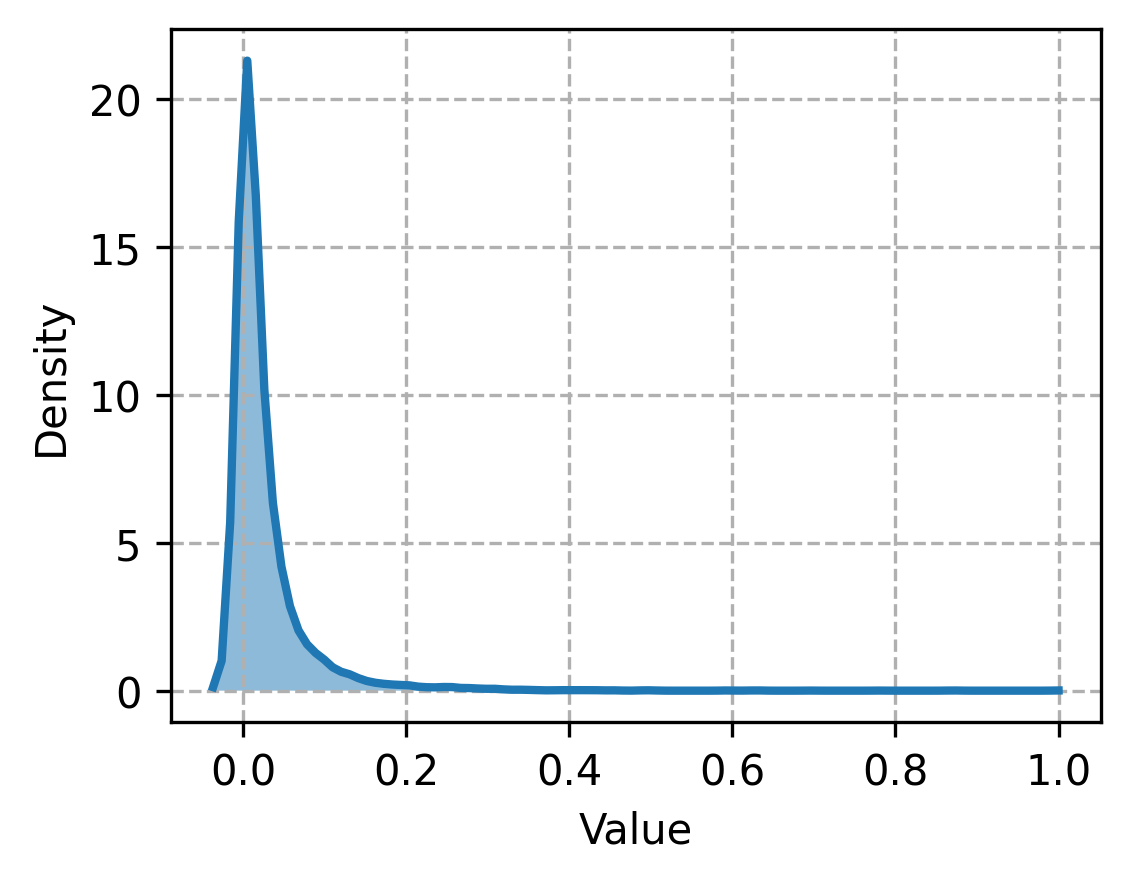}
    }
    \hfill
    \subfigure[Layer 10]{\includegraphics[width=0.22\textwidth]{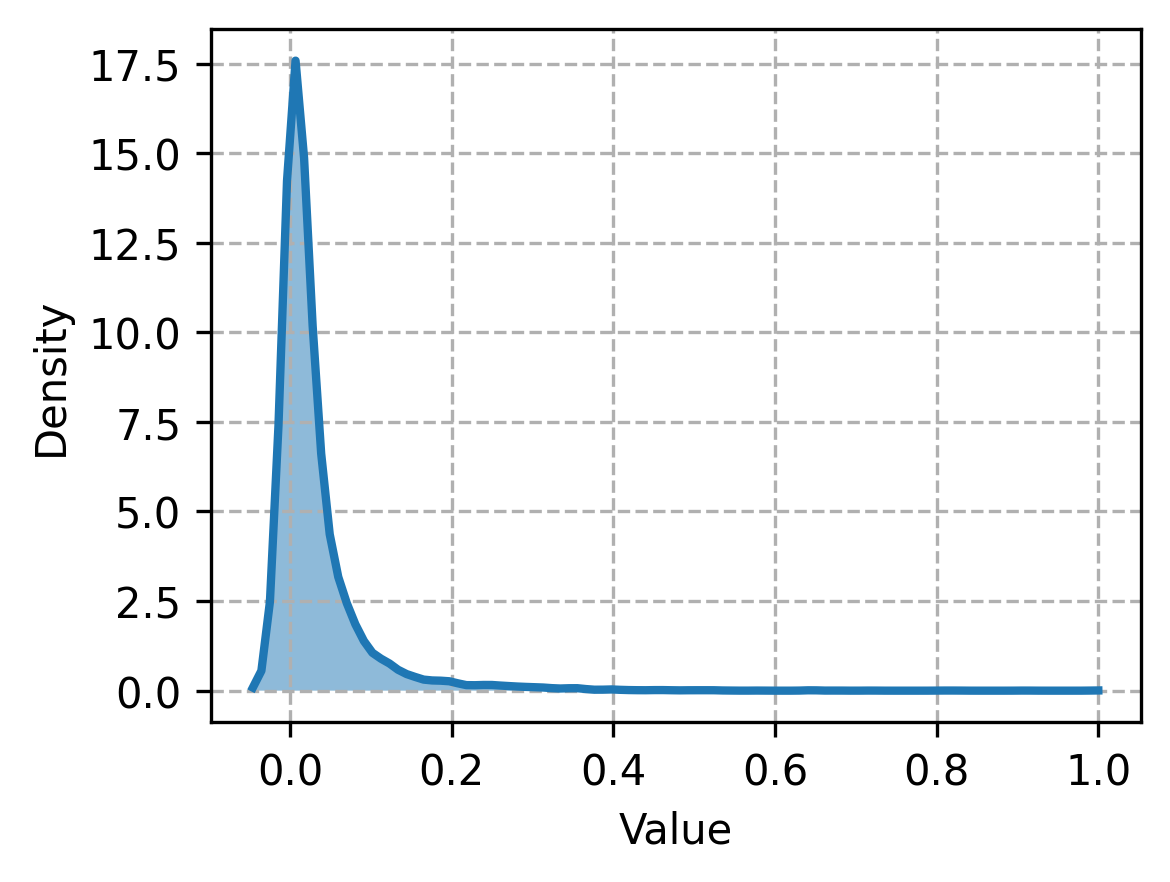}
    }
    \hfill
    \subfigure[Layer 11]{\includegraphics[width=0.22\textwidth]{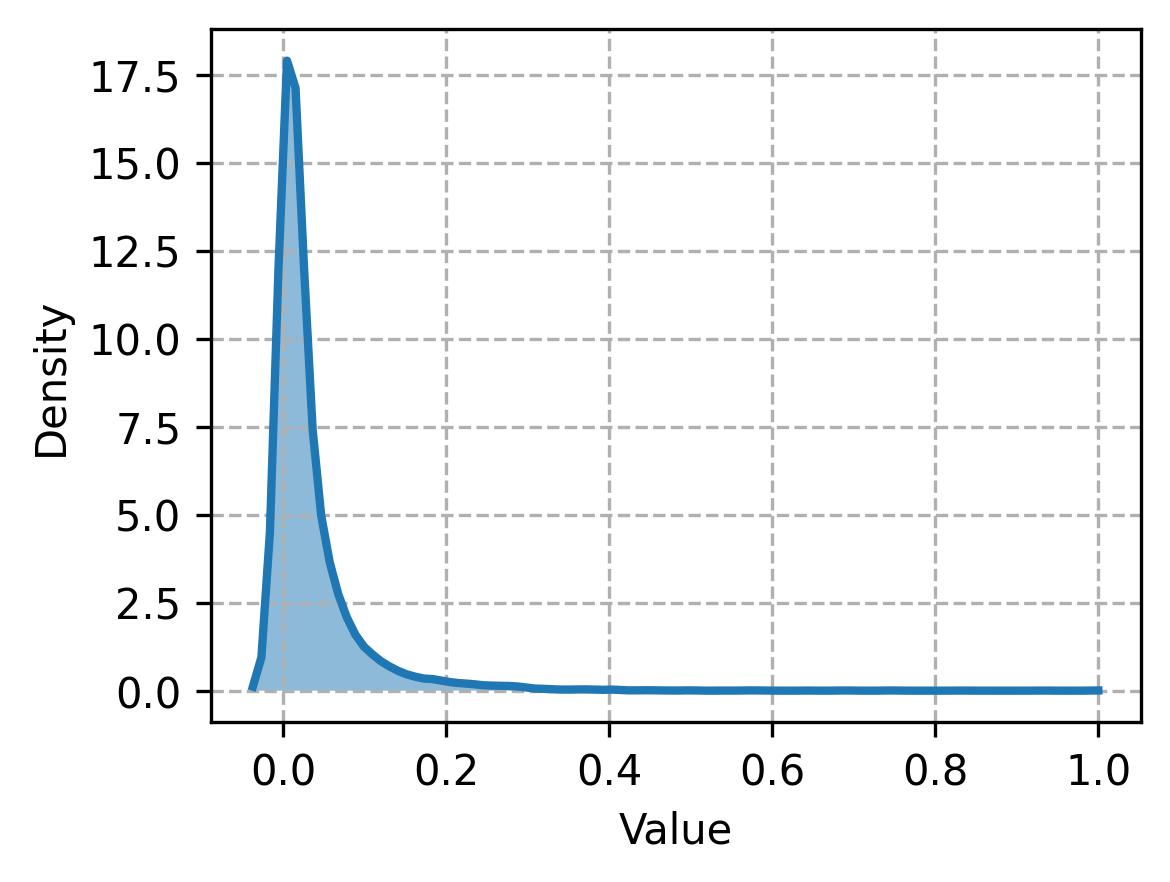}
    }
    \hfill
    \subfigure[Layer 12]{\includegraphics[width=0.22\textwidth]{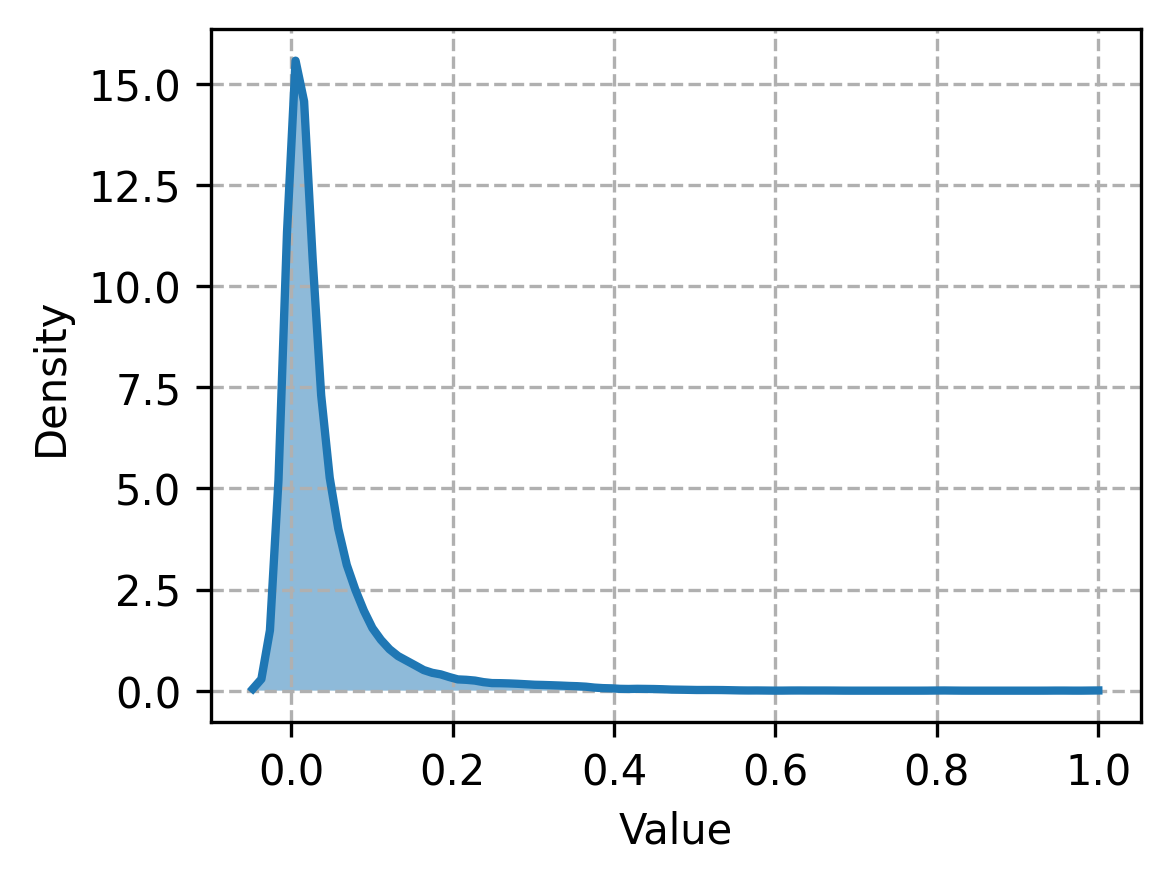}
    }
    \hfill
    \subfigure[Layer 13]{\includegraphics[width=0.22\textwidth]{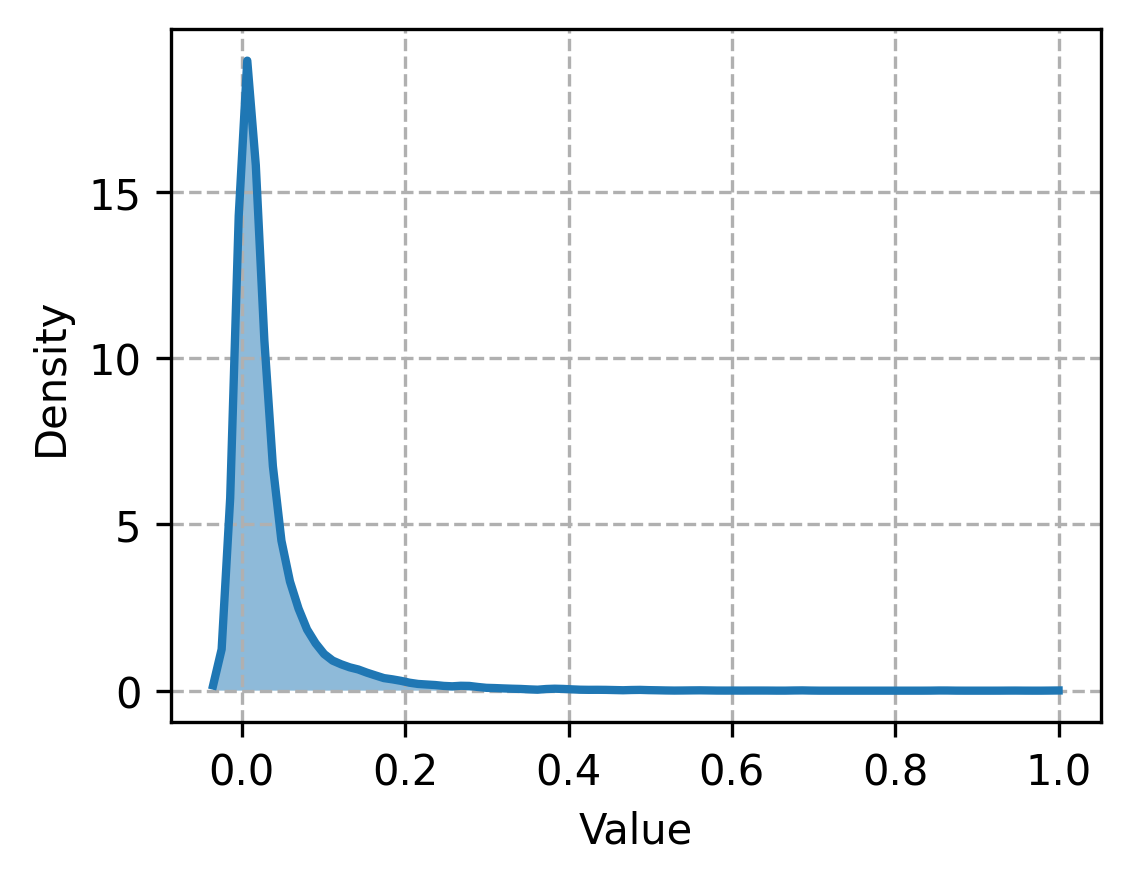}
    }
    \hfill
    \subfigure[Layer 14]{\includegraphics[width=0.22\textwidth]{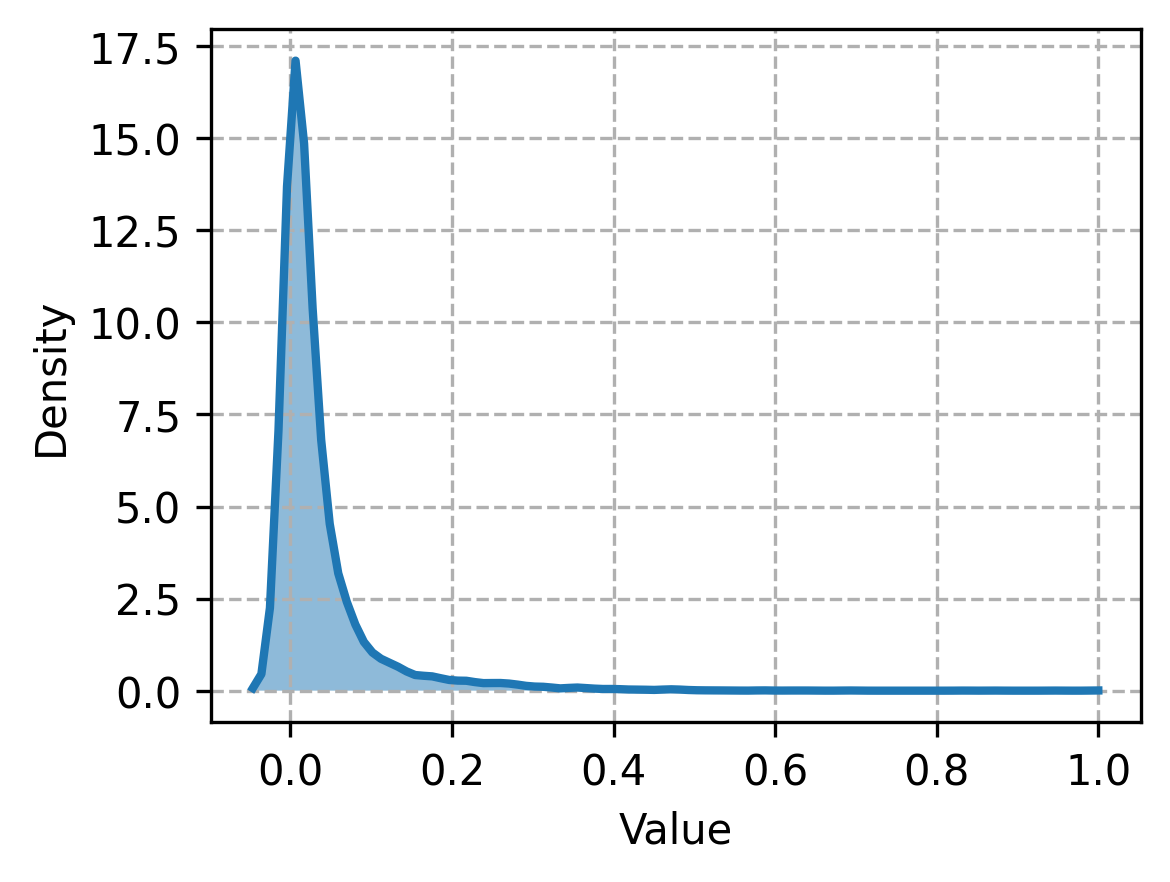}
    }
    \hfill
    \subfigure[Layer 15]{\includegraphics[width=0.22\textwidth]{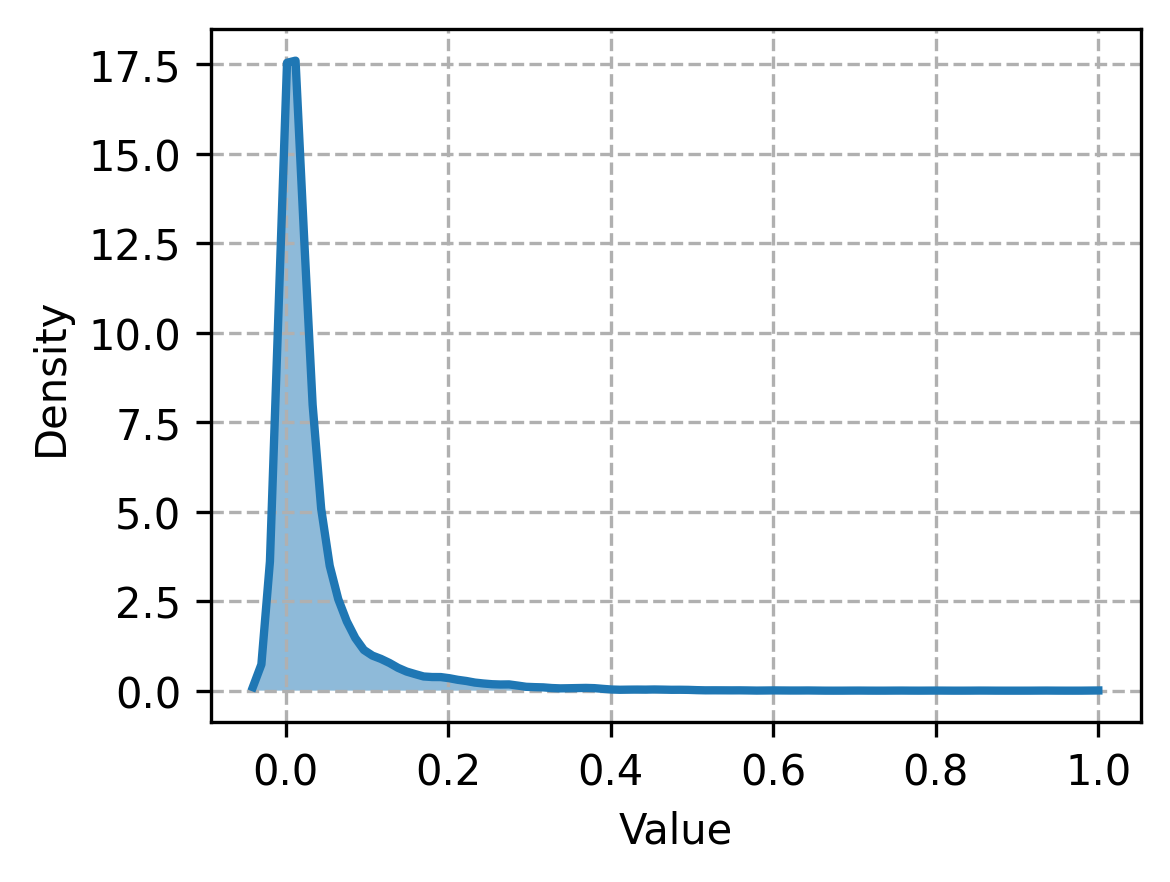}
    }
    \hfill
    \subfigure[Layer 16]{\includegraphics[width=0.22\textwidth]{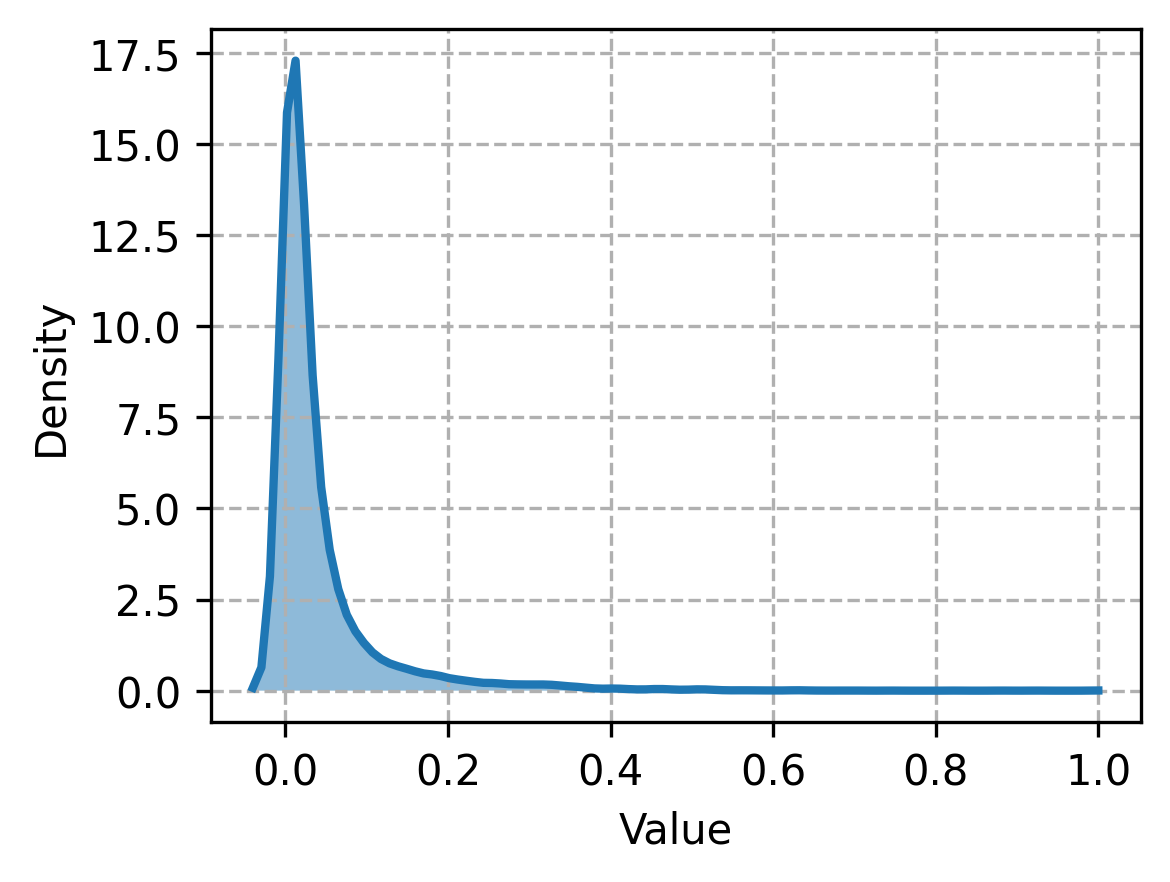}
    }
    \hfill
    \subfigure[Layer 17]{\includegraphics[width=0.22\textwidth]{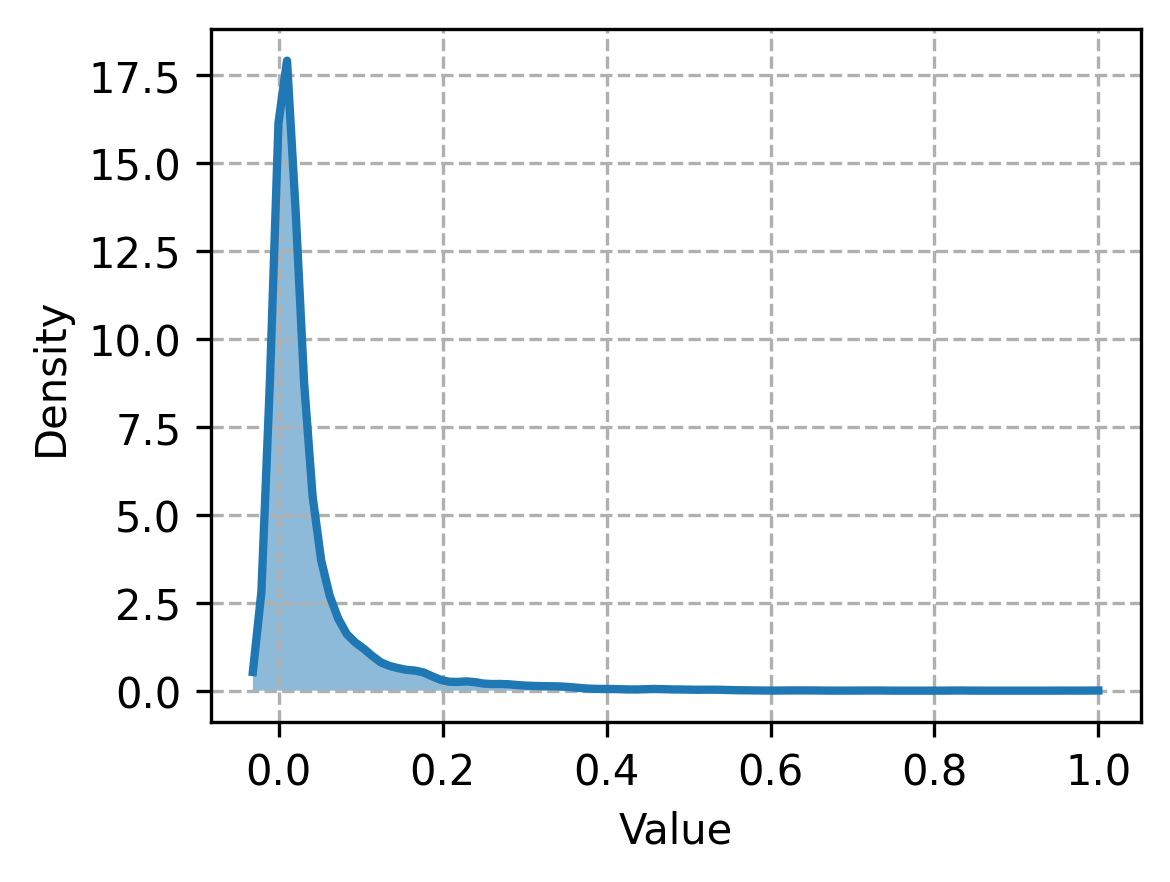}
    }
    \hfill
    \subfigure[Layer 18]{\includegraphics[width=0.22\textwidth]{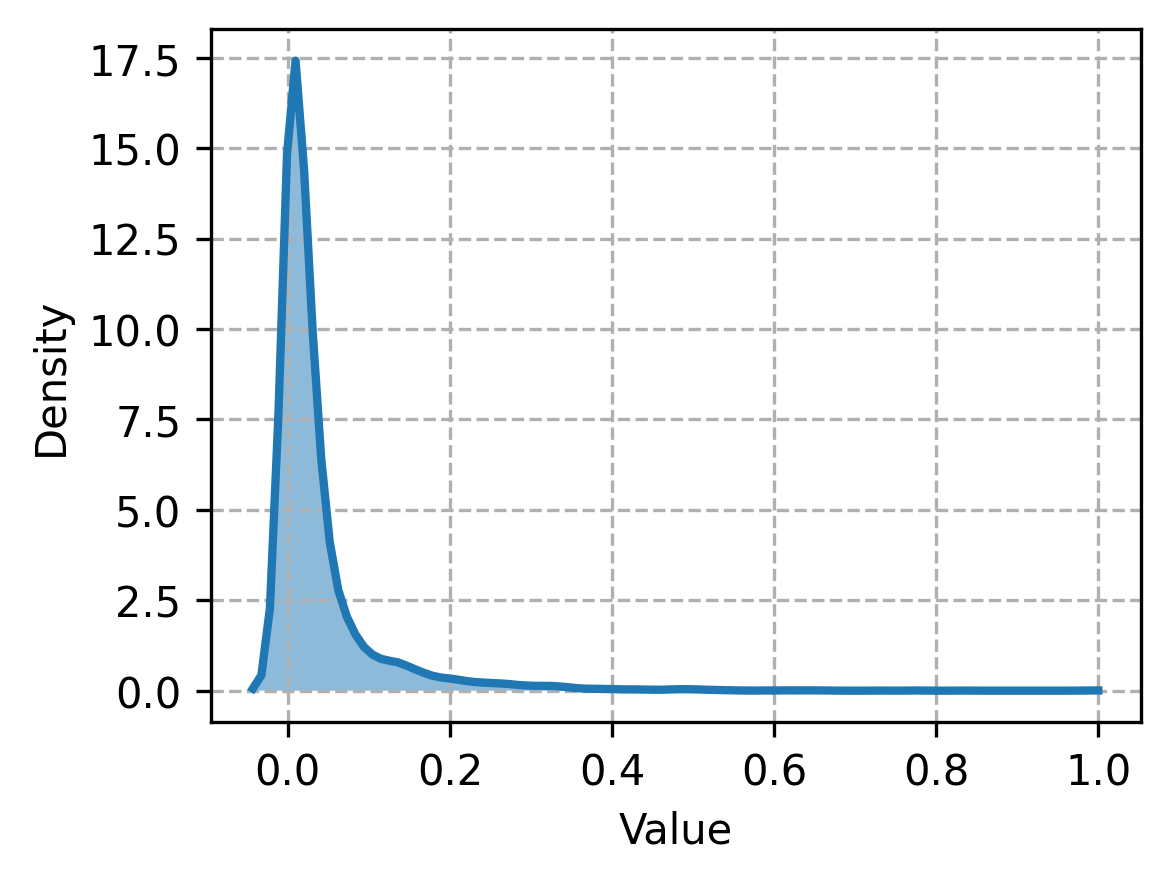}
    }
    \hfill
    \subfigure[Layer 19]{\includegraphics[width=0.22\textwidth]{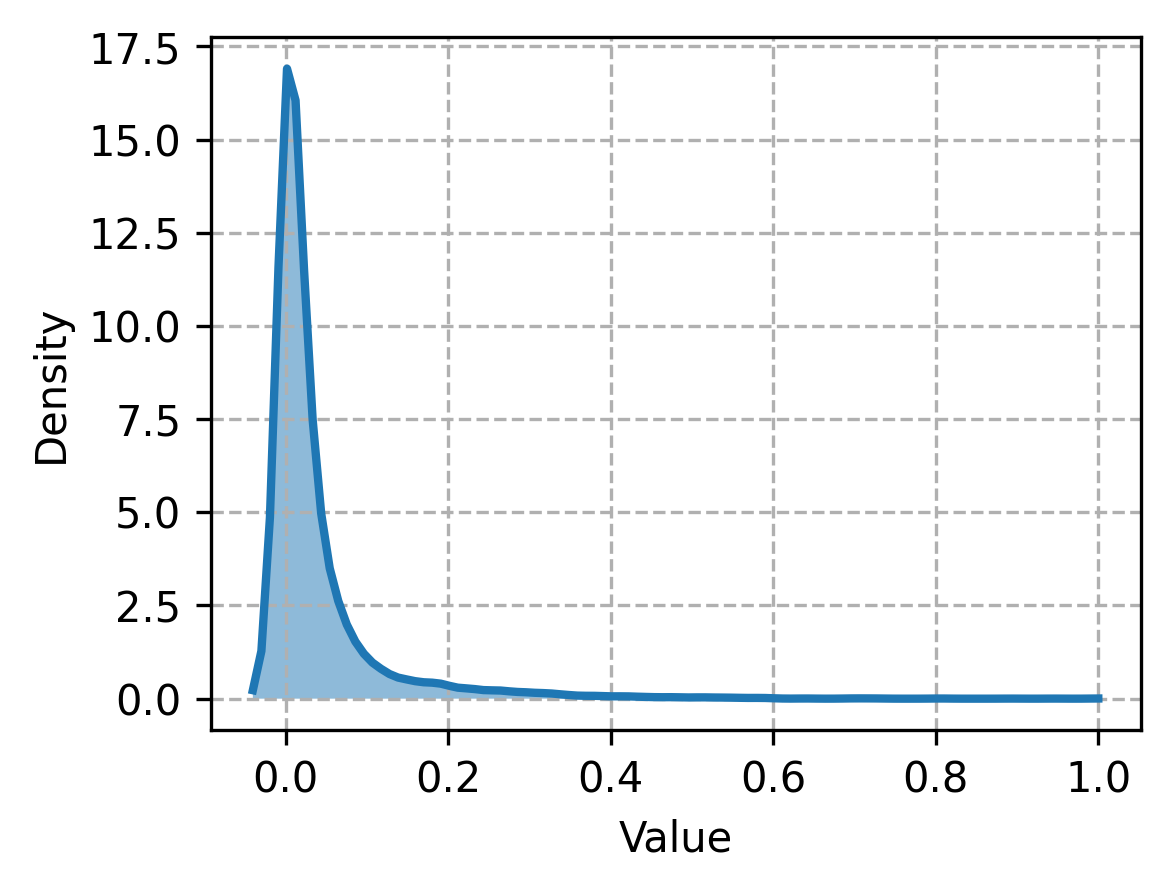}
    }
  \caption{In GPT-J-6B, the KDE of elements in \(P\) matrices across layers (0-19).}
  \label{fig:superposition_KDE_gpt-j-6b-part1}
\end{figure*}

\begin{figure*}
    \centering
    \subfigure[Layer 20]{\includegraphics[width=0.22\textwidth]{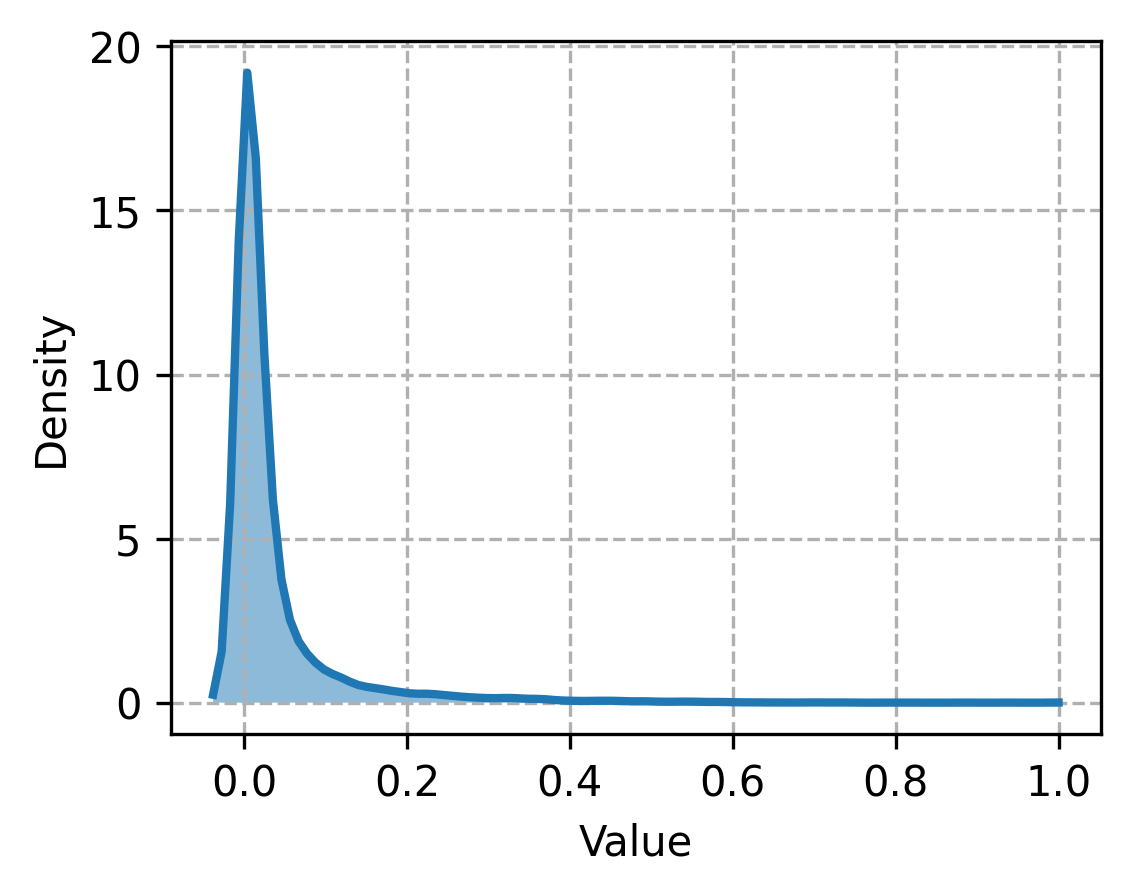}
    }
    \hfill
    \subfigure[Layer 21]{\includegraphics[width=0.22\textwidth]{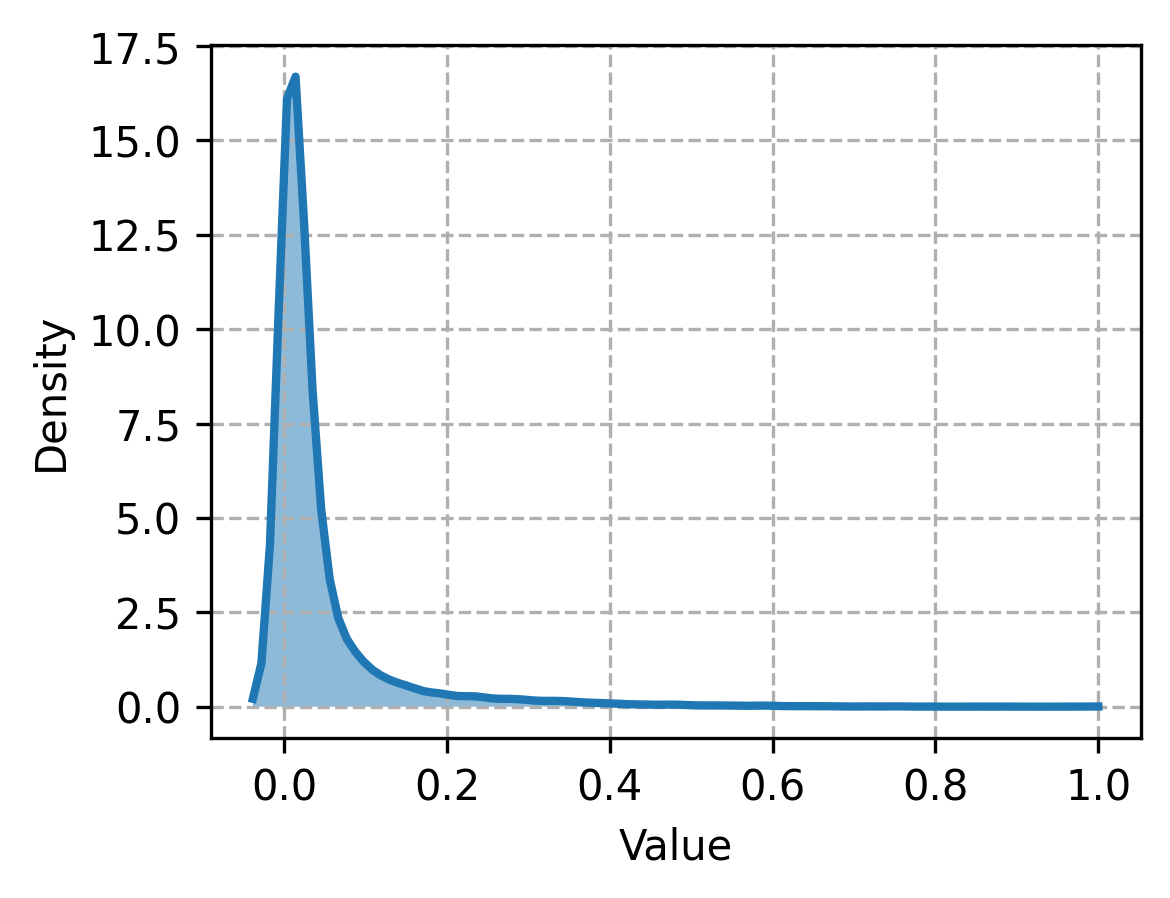}
    }
    \hfill
    \subfigure[Layer 22]{\includegraphics[width=0.22\textwidth]{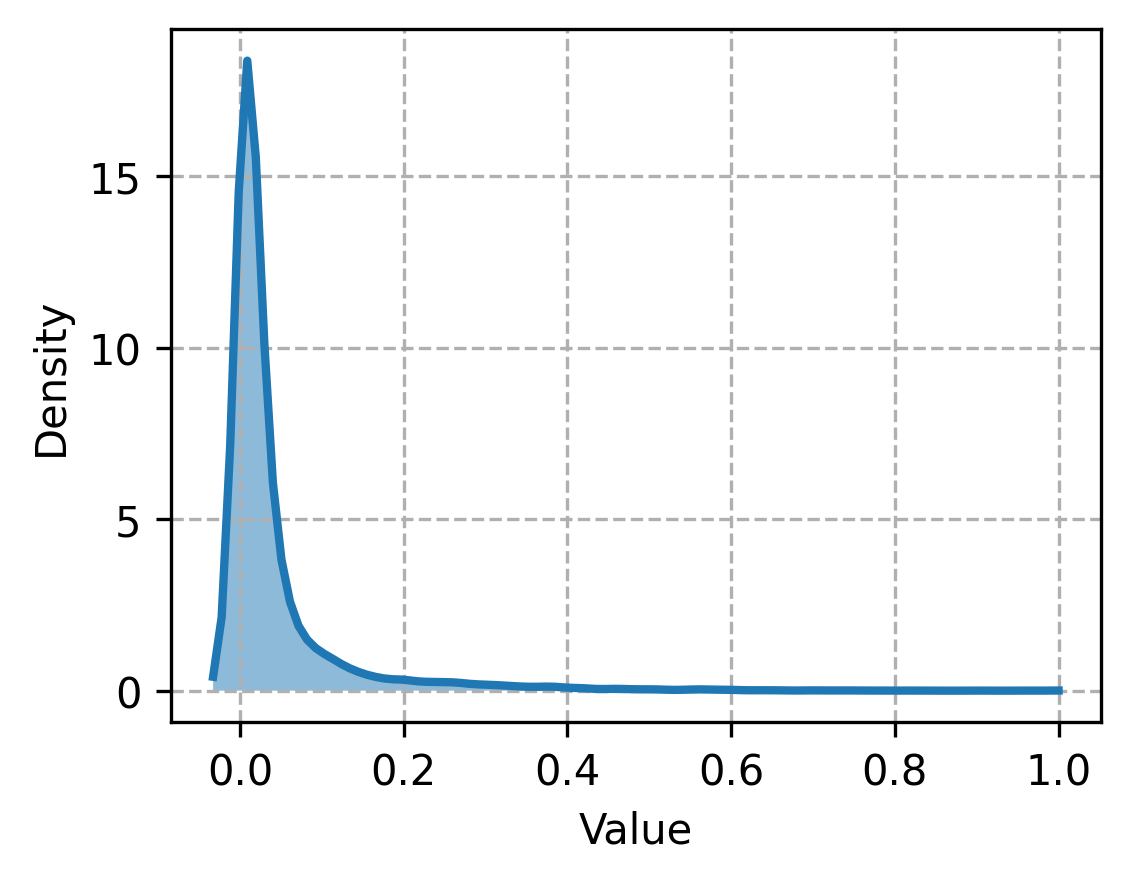}
    }
    \hfill
    \subfigure[Layer 23]{\includegraphics[width=0.22\textwidth]{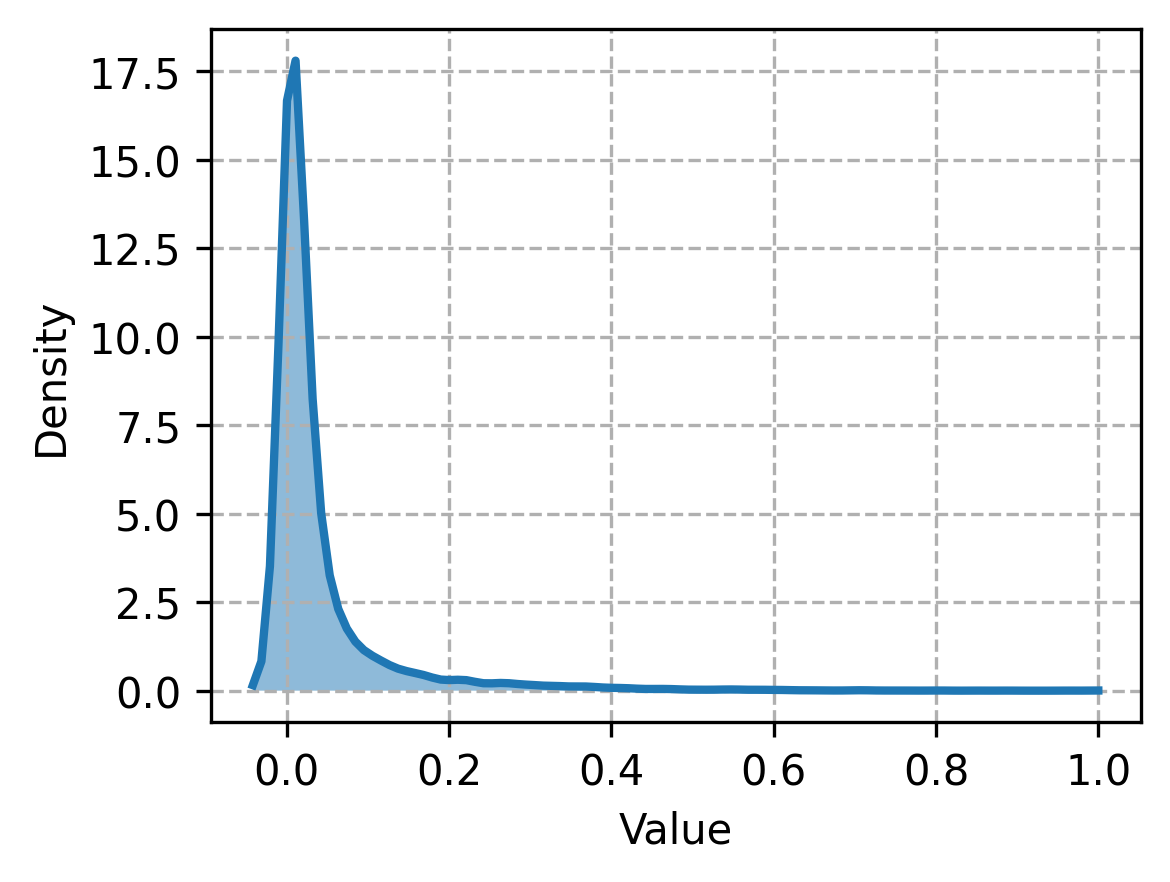}
    }
    \hfill
    \subfigure[Layer 24]{\includegraphics[width=0.22\textwidth]{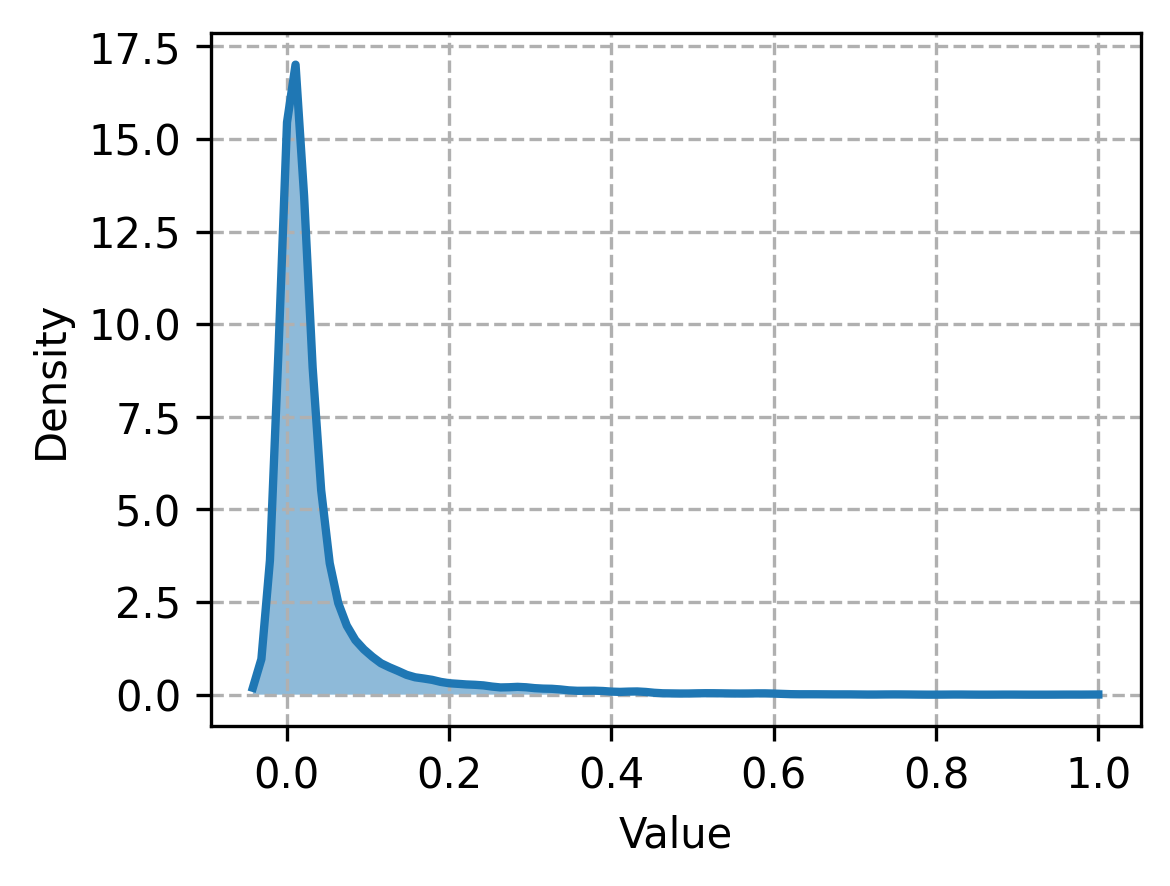}
    }
    \hfill
    \subfigure[Layer 25]{\includegraphics[width=0.22\textwidth]{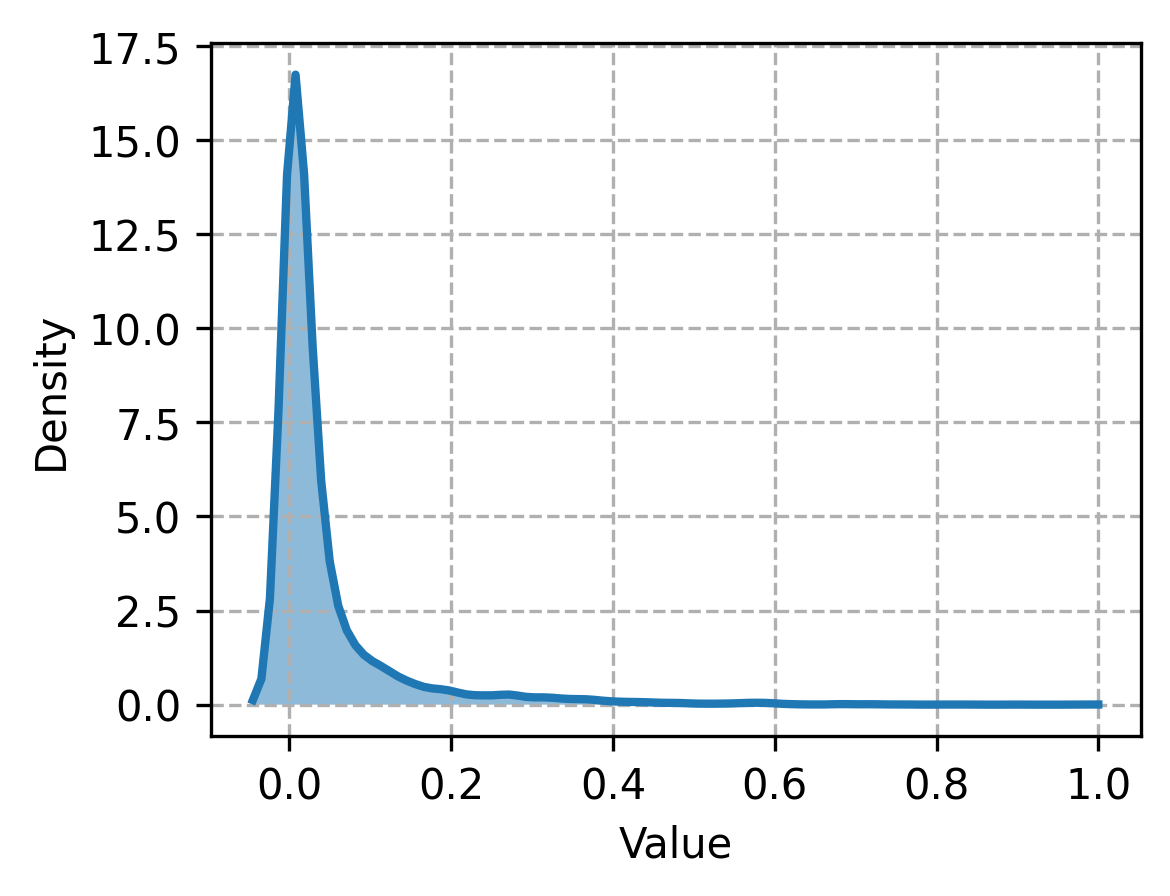}
    }
    \hfill
    \subfigure[Layer 26]{\includegraphics[width=0.22\textwidth]{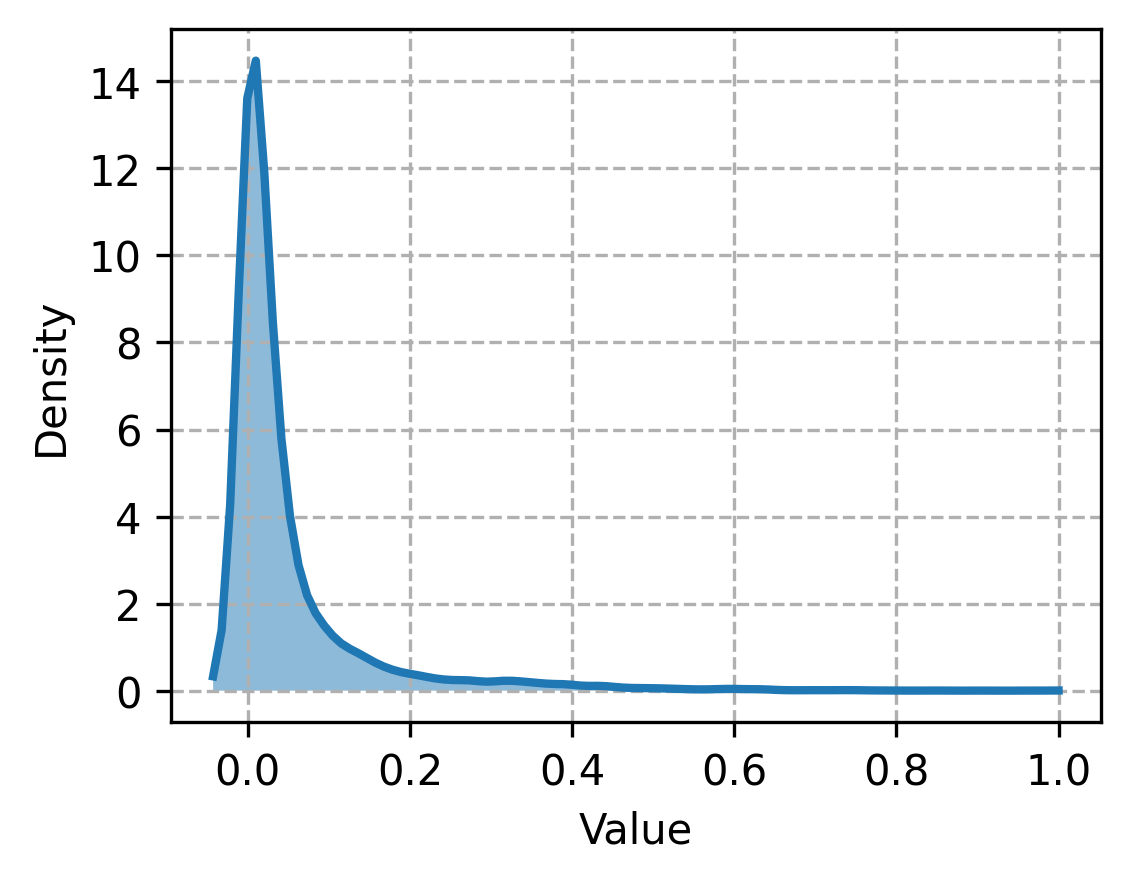}
    }
    \hfill
    \subfigure[Layer 27]{\includegraphics[width=0.22\textwidth]{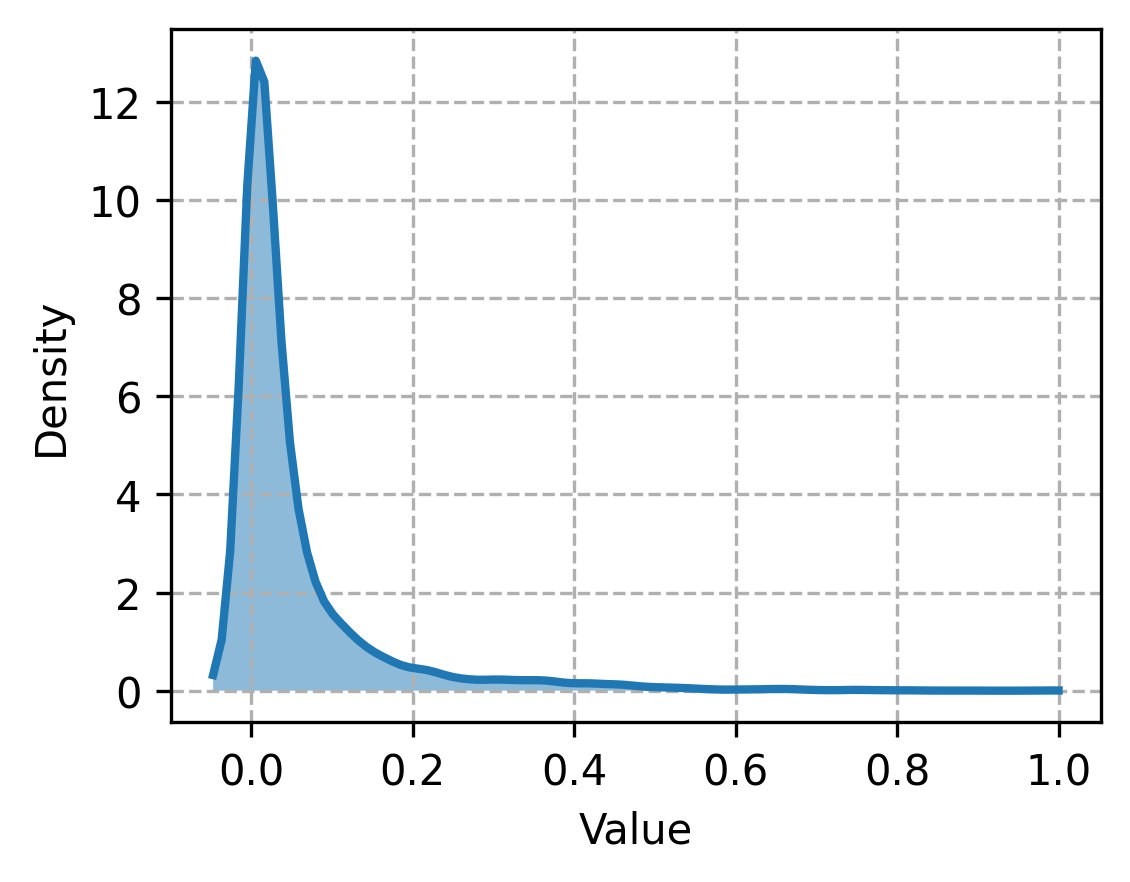}
    }
  \caption{In GPT-J-6B, the KDE of elements in \(P\) matrices across layers (20-27).}
  \label{fig:superposition_KDE_gpt-j-6b-part2}
\end{figure*}

In this section, we will present KDE of the elements of \( P \) matrices across layers for GPT2-Small, GPT2-Medium, GPT2-Large, GPT-J-6B, Pythia-1B, Pythia-2.8B, Pythia-6.9B, Llama2-7B, Llama2-13B, Llama3-8B and Llama3.1-8B. These will be displayed to provide readers with a deepper understanding for the distribution of knowledge superposition, as illustrated in Figures~\ref{fig:superposition_KDE_gpt2-small}-~\ref{fig:superposition_KDE_gpt-j-6b-part2}. Some experiments have been deleted due to Arxiv upload restrictions. If you really need these experimental results, please contact us.

\section{F. Case Study}\label{appendix:F. Case Study}

In this section, we focus on selecting the top 20 knowledge pairs with their corresponding \( p(\cdot,\cdot) \) values from the \( P \) matrix at layer 0 in GPT2-Small, GPT2-Medium, GPT2-Large, and GPT-J-6B. These represent the 20 pairs with the highest degree of superposition among the 128\*128 knowledge pairs.

From Tables~\ref{tab:case_study_gpt2-small}-~\ref{tab:case_study_gpt-j}, we observe the following: First, knowledge pairs with high \( p(\cdot,\cdot) \) values are closely related and often involve content with very similar backgrounds. Second, as the model size increases, the tail of the superposition distribution, where \( p(\cdot,\cdot) \) values are relatively high, becomes progressively clearer. This is particularly evident in GPT-J, where the \( p(\cdot,\cdot) \) value jumps directly from 1 to 0.13 on top 20 pairs, indicating that larger models can discern more fine-grained differences in knowledge. Lastly, we notice that models seem to perform similar operations on seemingly different but actually closely related knowledge in certain layers. For example, in GPT-J, at layer 0, the model assigns a \( p(\cdot,\cdot) \) value of 1 to pairs like "Ilya Nikolaevich Ulyanov" and "Nikolay Nekrasov," "Pierre Trabaud" and "Patrick Rambaud," "Vladimir Mayakovsky" and "Vladimir Bukovsky," and even pairs like "Mac OS X 10.1" and "Windows 8.1," indicating consistent behavior by the model's MLP in layer 0 towards these knowledge pairs.

\section{G. Experimental Detail}\label{appendix:G. Experimental Detail}

Specifically, we intersect the data items from CounterFact that all the language models can correctly complete, and we randomly select 128 items from this intersection. 

The activation extraction is performed at the last token of the subject, which has been shown by multiple studies to be the location where knowledge is extracted in language models (\citealp{meng2022locating}; \citealp{geva2023dissecting}). For instance, in the sentence “The mother tongue of Danielle Darrieux is French,” we extract activation values from the last token of subject, namely "Darrieux", at each layer's MLP in language models. 

For the computation of $C$, we pre-cache the non-centered covariance of activations \(k\) estimated from Wikipedia text \cite{lhoest2021datasets}, specifically from the Wikipedia "20220301.en" dataset.

All our experiments can be conducted on a single 80G A100 GPU.

\section{H. Continuous Editing Causes Models' Failure}\label{appendix:H. Continuous Editing Causes Model's Failure}

In this section, we conduct experiments to illustrate the effects of lifelong editing based on the theoretical framework established in Section Expanding to Lifelong Editing. Specifically, we show that the continual introduction of interference to knowledge unrelated to the edited knowledge (the mathematical formulation of such interference is detailed in Section Expanding to Lifelong Editing) leads to the progressive forgetting of unrelated knowledge by the model.

\begin{figure*}
    \centering
    \subfigure[Case 1]{\includegraphics[width=0.22\textwidth]{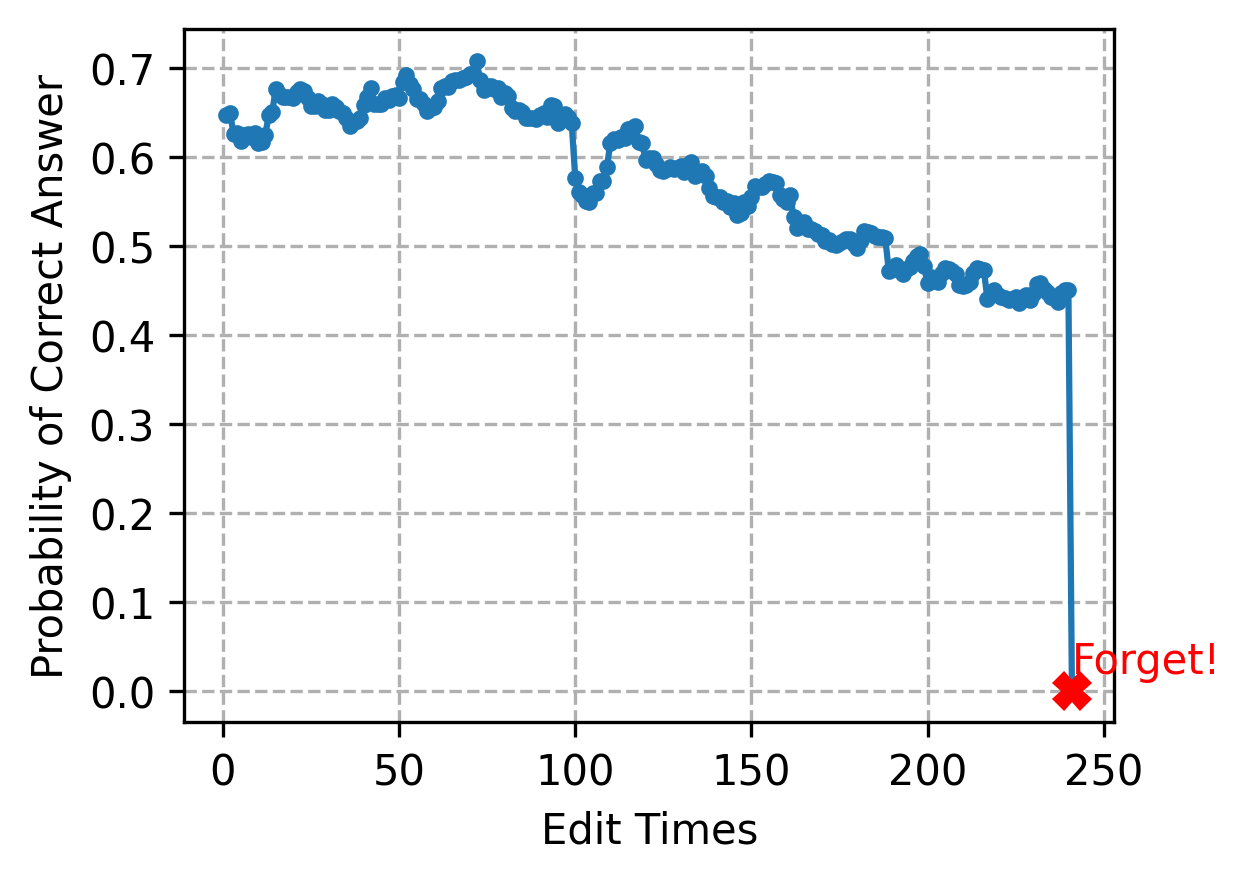}
    }
    \hfill
    \subfigure[Case 2]{\includegraphics[width=0.22\textwidth]{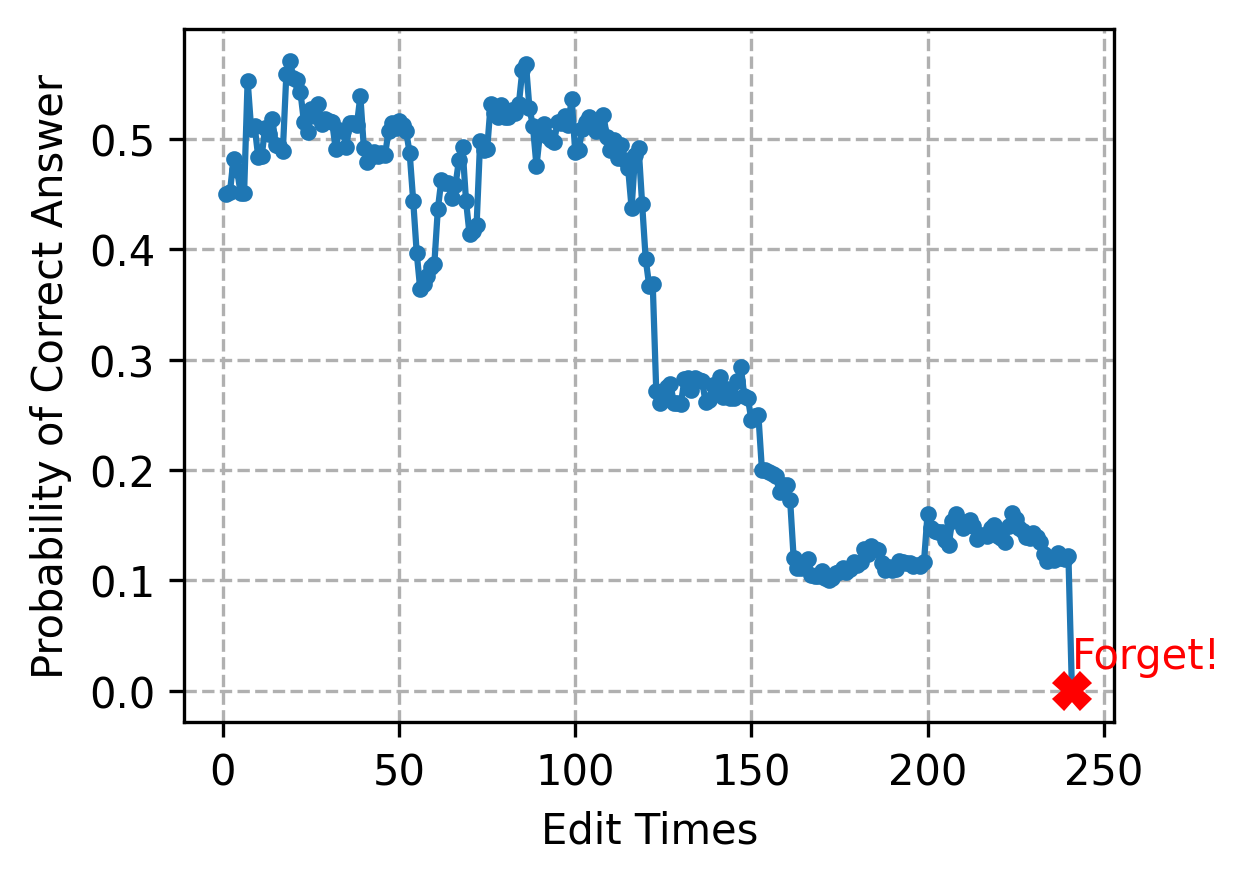}
    }
    \hfill
    \subfigure[Case 3]{\includegraphics[width=0.22\textwidth]{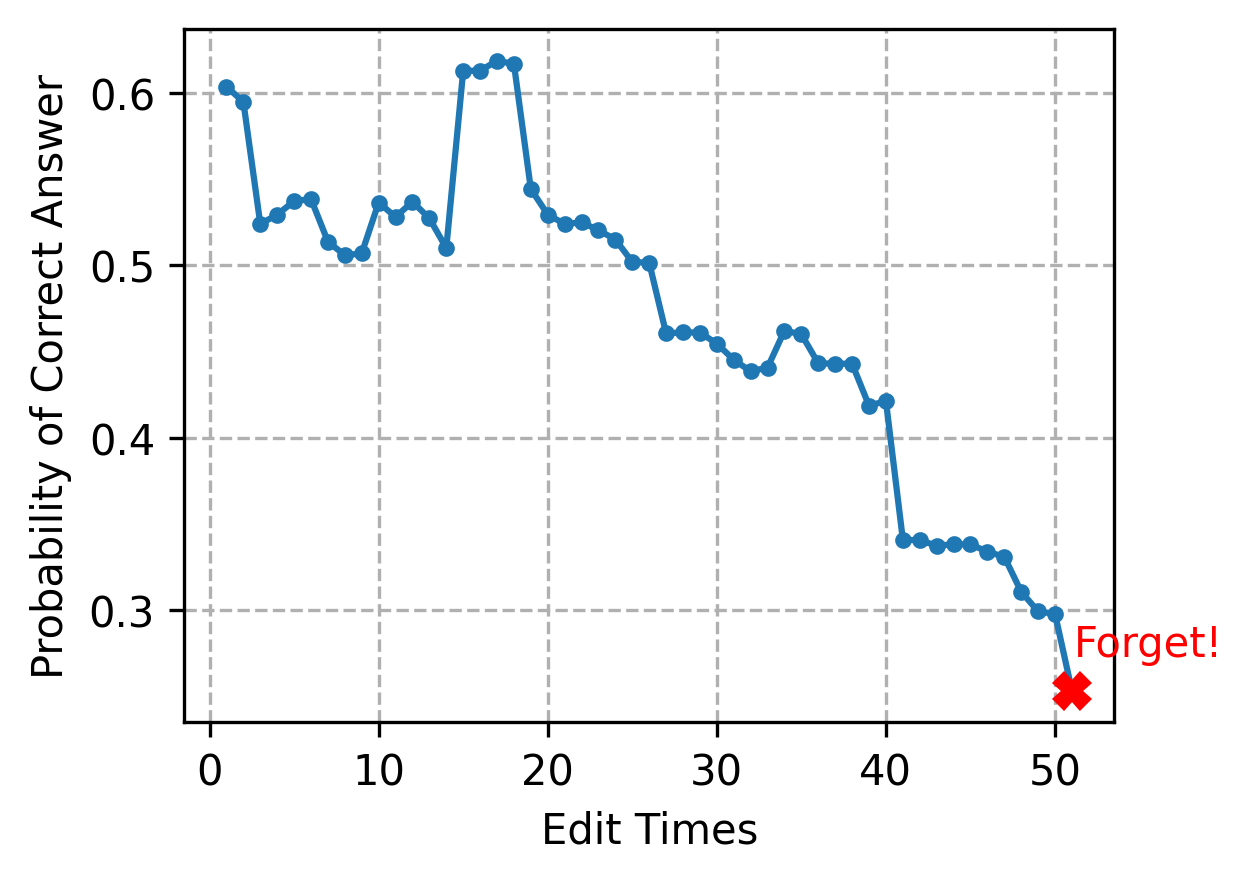}
    }
    \hfill
    \subfigure[Case 4]{\includegraphics[width=0.22\textwidth]{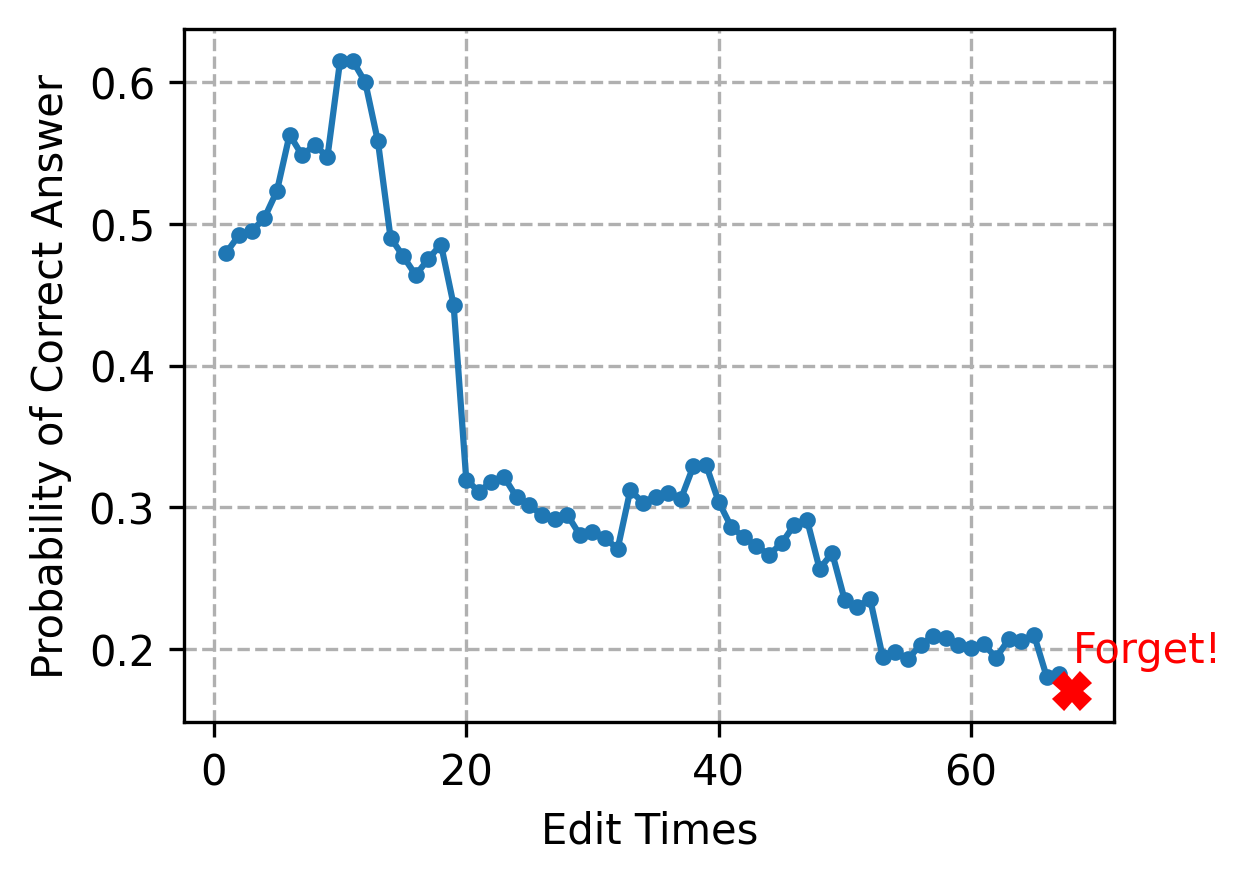}
    }
    \hfill
    \subfigure[Case 5]{\includegraphics[width=0.22\textwidth]{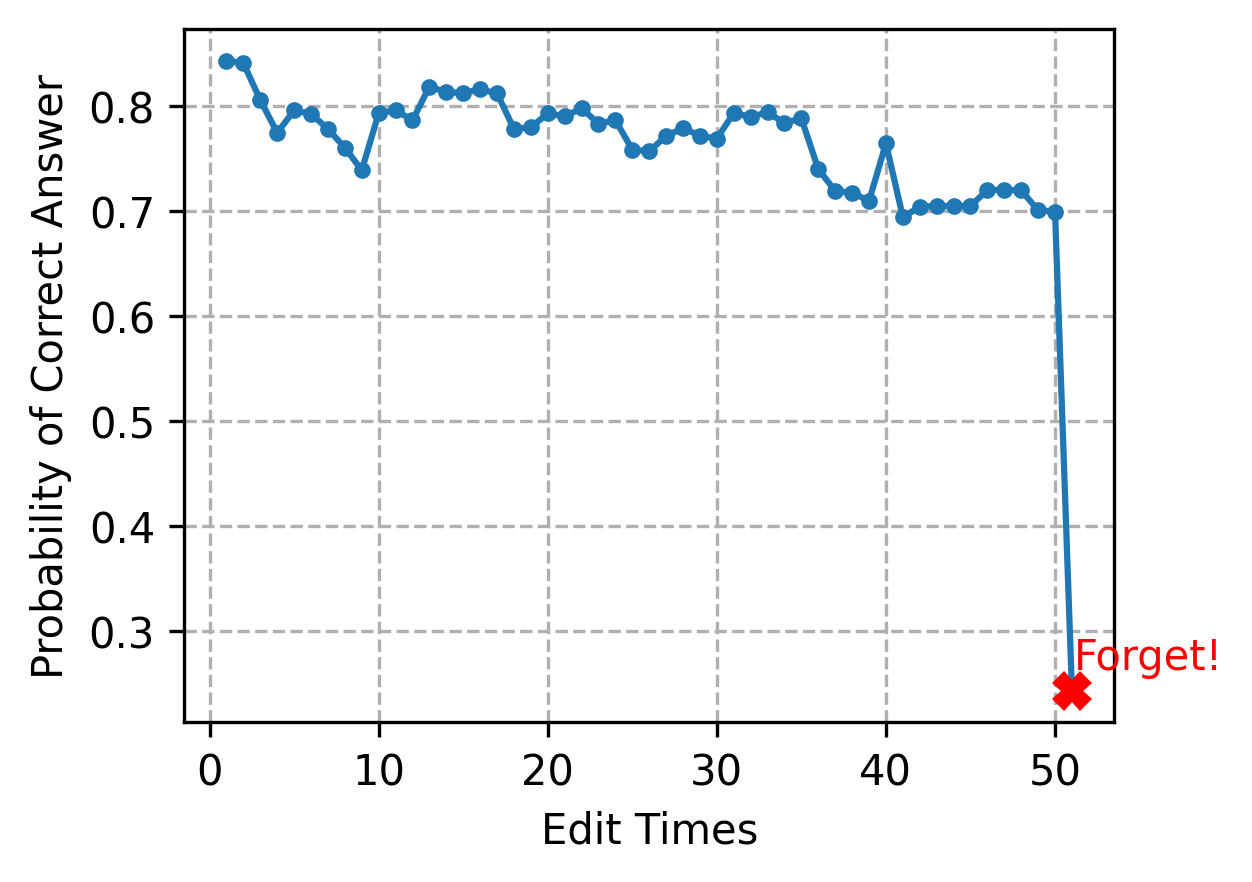}
    }
    \hfill
    \subfigure[Case 6]{\includegraphics[width=0.22\textwidth]{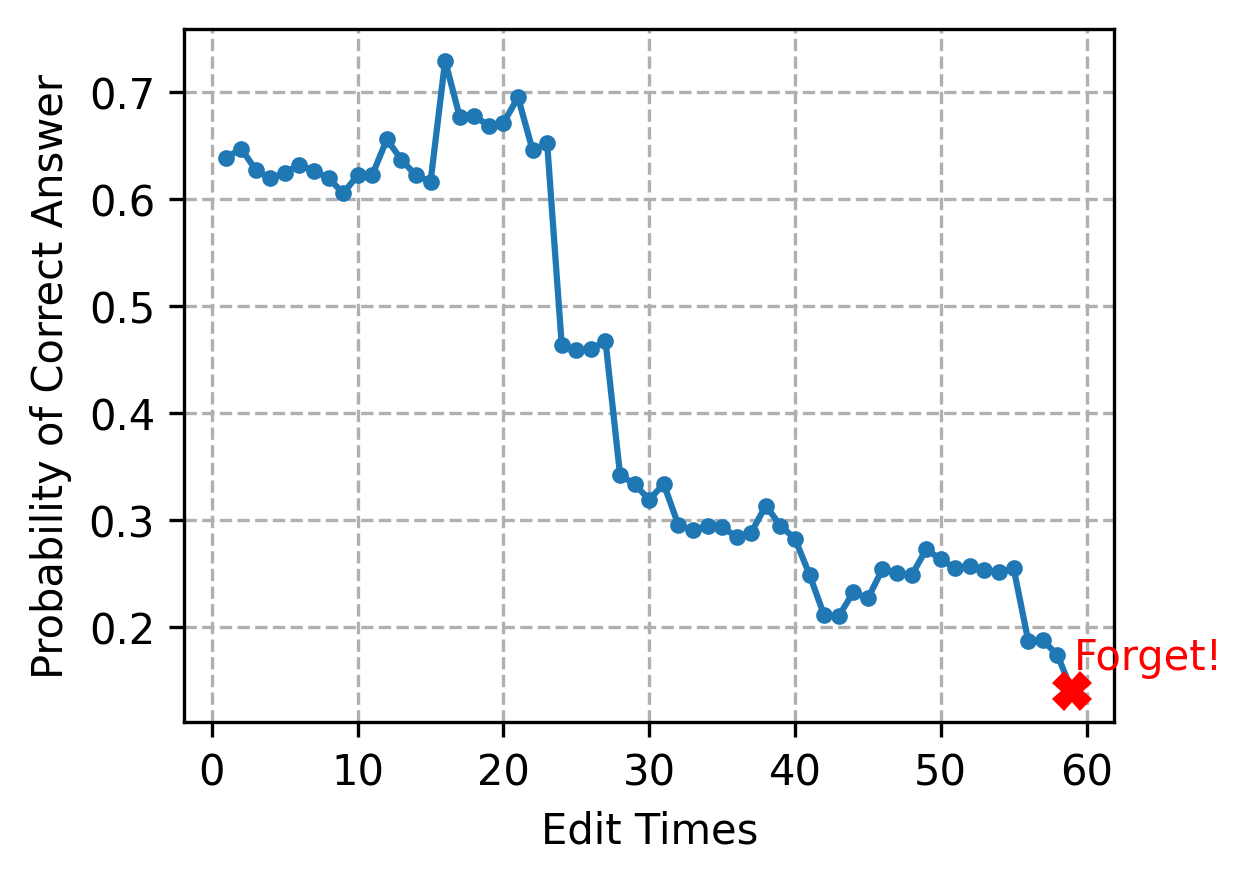}
    }
    \hfill
    \subfigure[Case 7]{\includegraphics[width=0.22\textwidth]{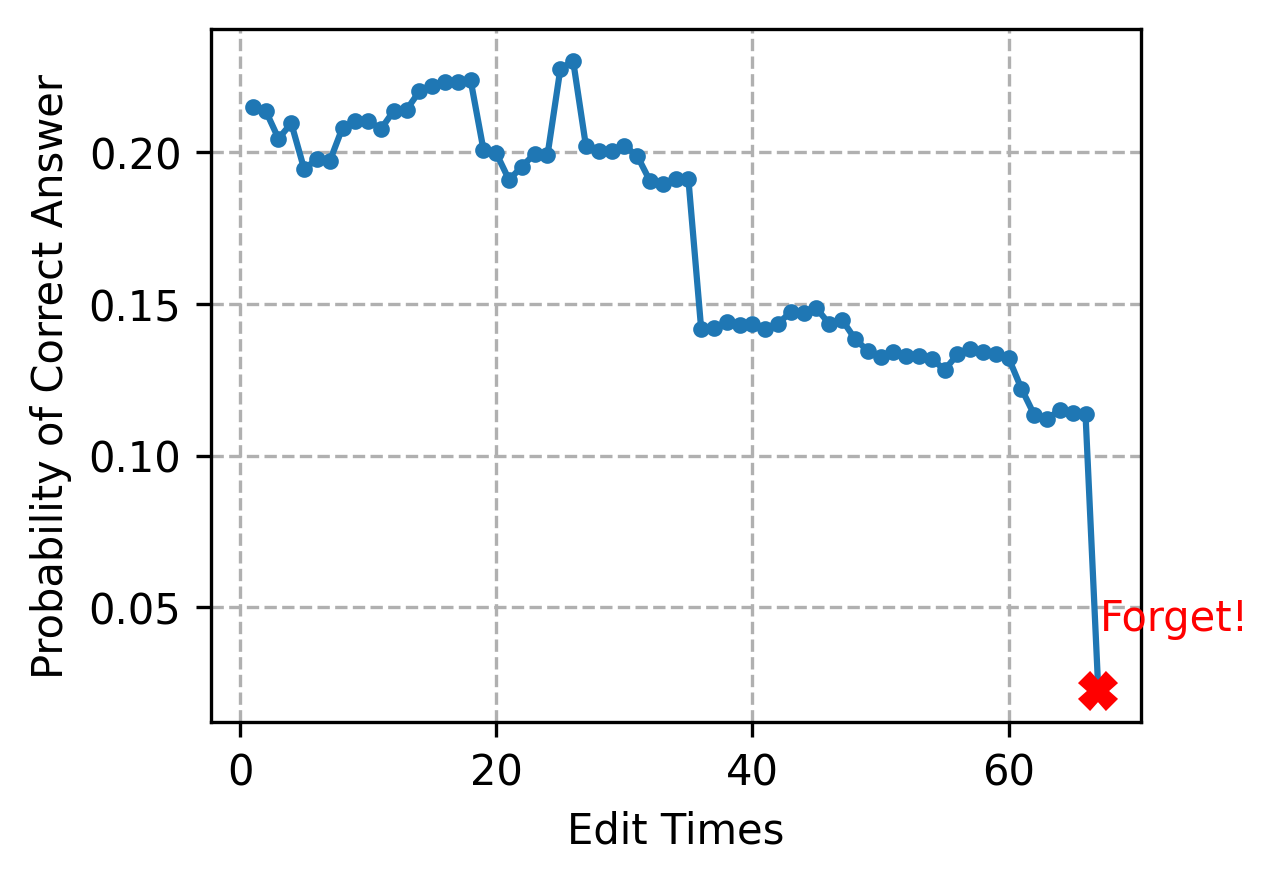}
    }
    \hfill
    \subfigure[Case 8]{\includegraphics[width=0.22\textwidth]{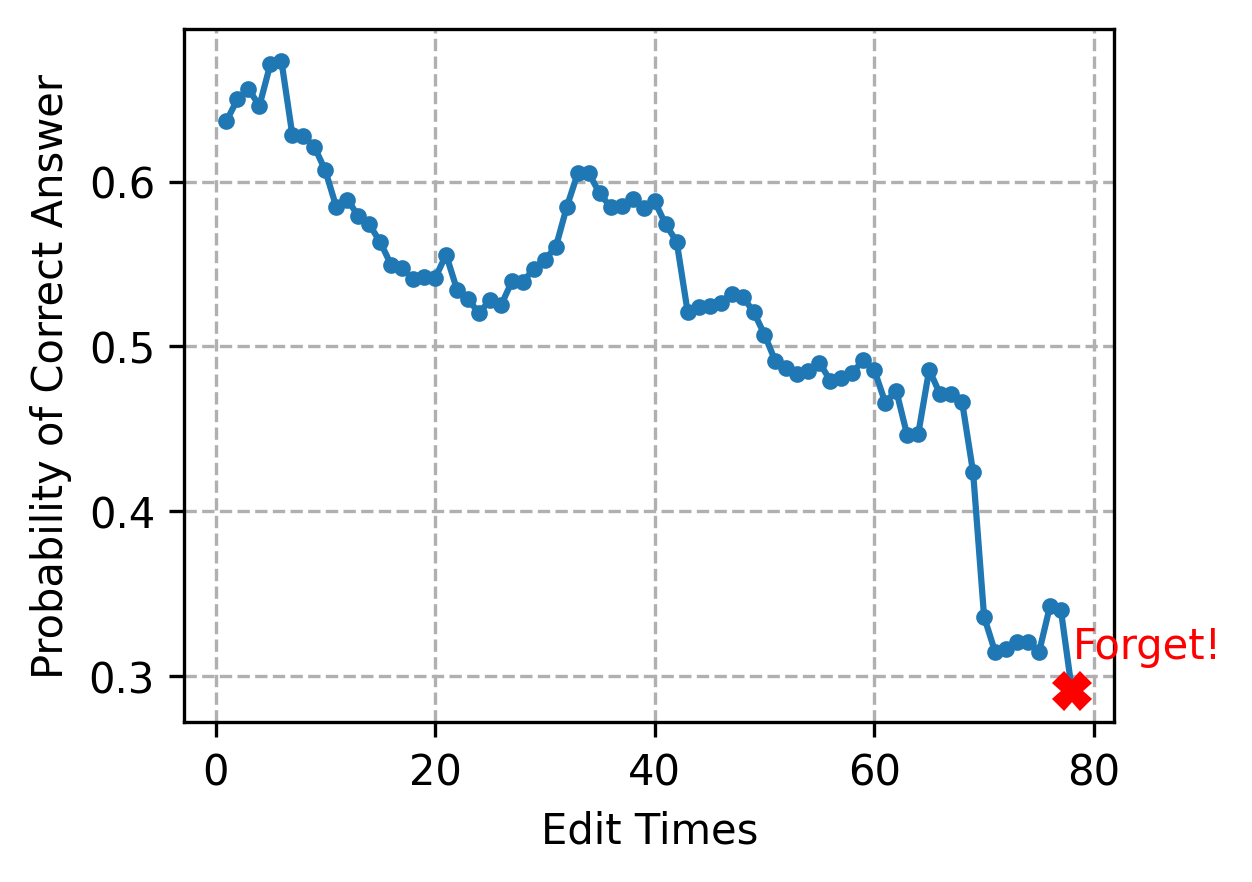}
    }
    \hfill
    \subfigure[Case 9]{\includegraphics[width=0.22\textwidth]{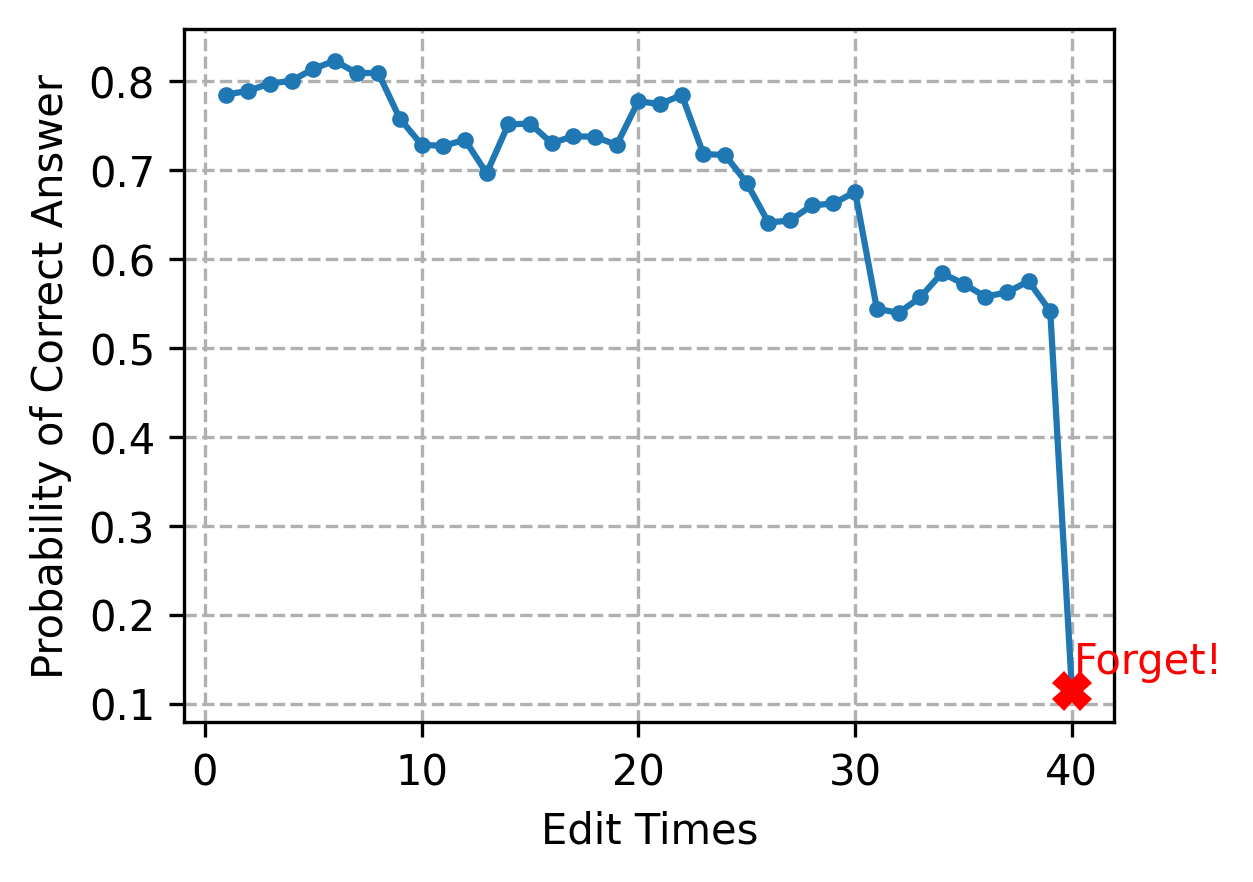}
    }
    \hfill
    \subfigure[Case 10]{\includegraphics[width=0.22\textwidth]{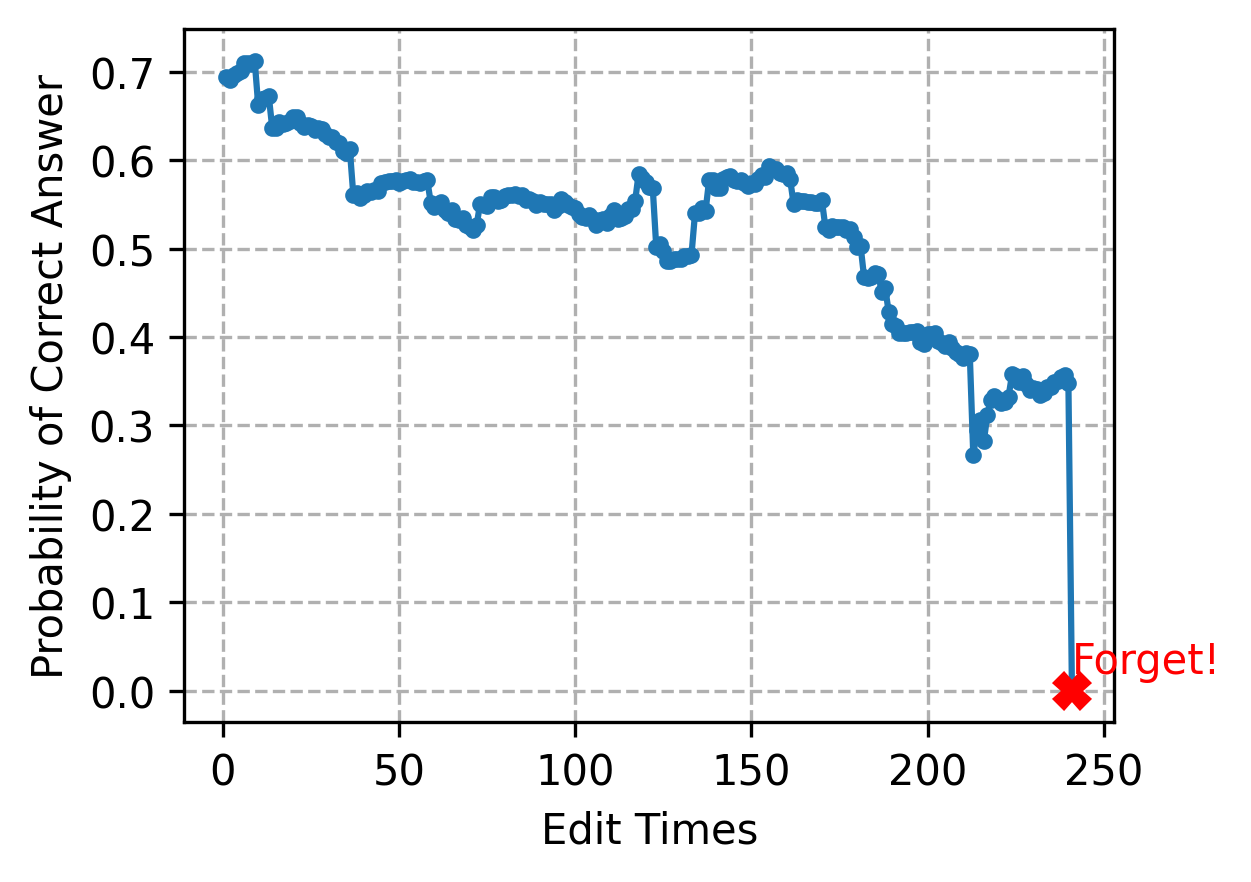}
    }
    \hfill
    \subfigure[Case 11]{\includegraphics[width=0.22\textwidth]{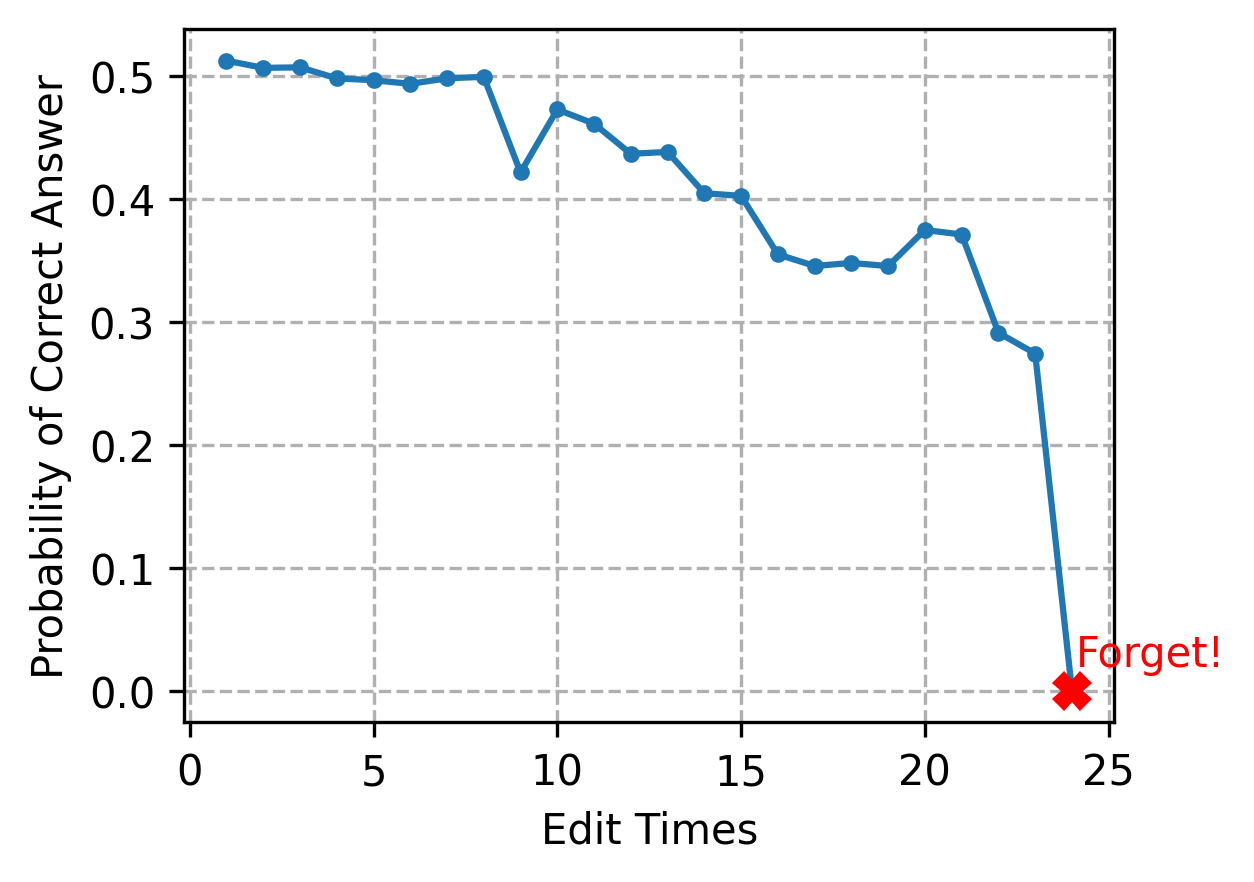}
    }
    \hfill
    \subfigure[Case 12]{\includegraphics[width=0.22\textwidth]{fig/supplementary_results/knowledge_forget_12.png}
    }
    \hfill
    \subfigure[Case 13]{\includegraphics[width=0.22\textwidth]{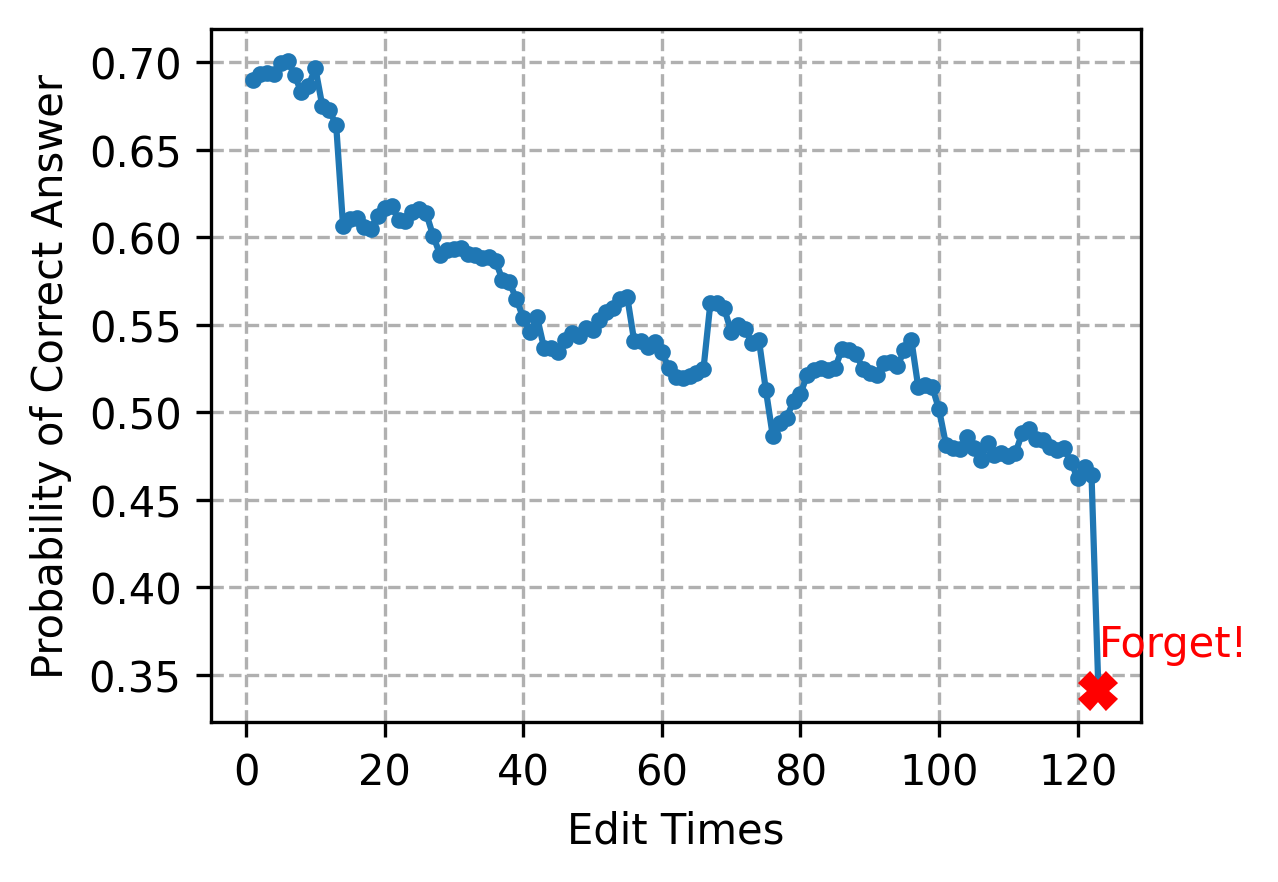}
    }
    \hfill
    \subfigure[Case 14]{\includegraphics[width=0.22\textwidth]{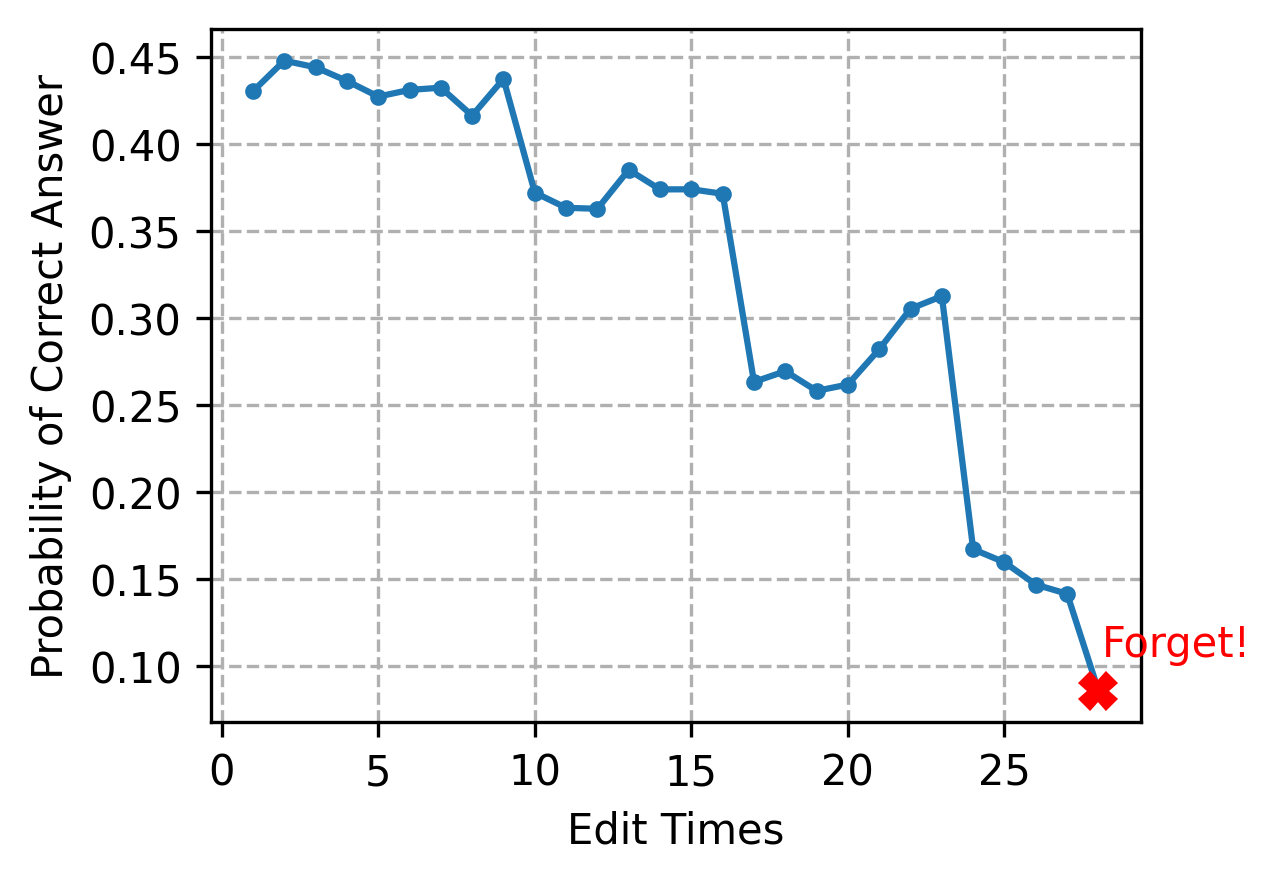}
    }
    \hfill
    \subfigure[Case 15]{\includegraphics[width=0.22\textwidth]{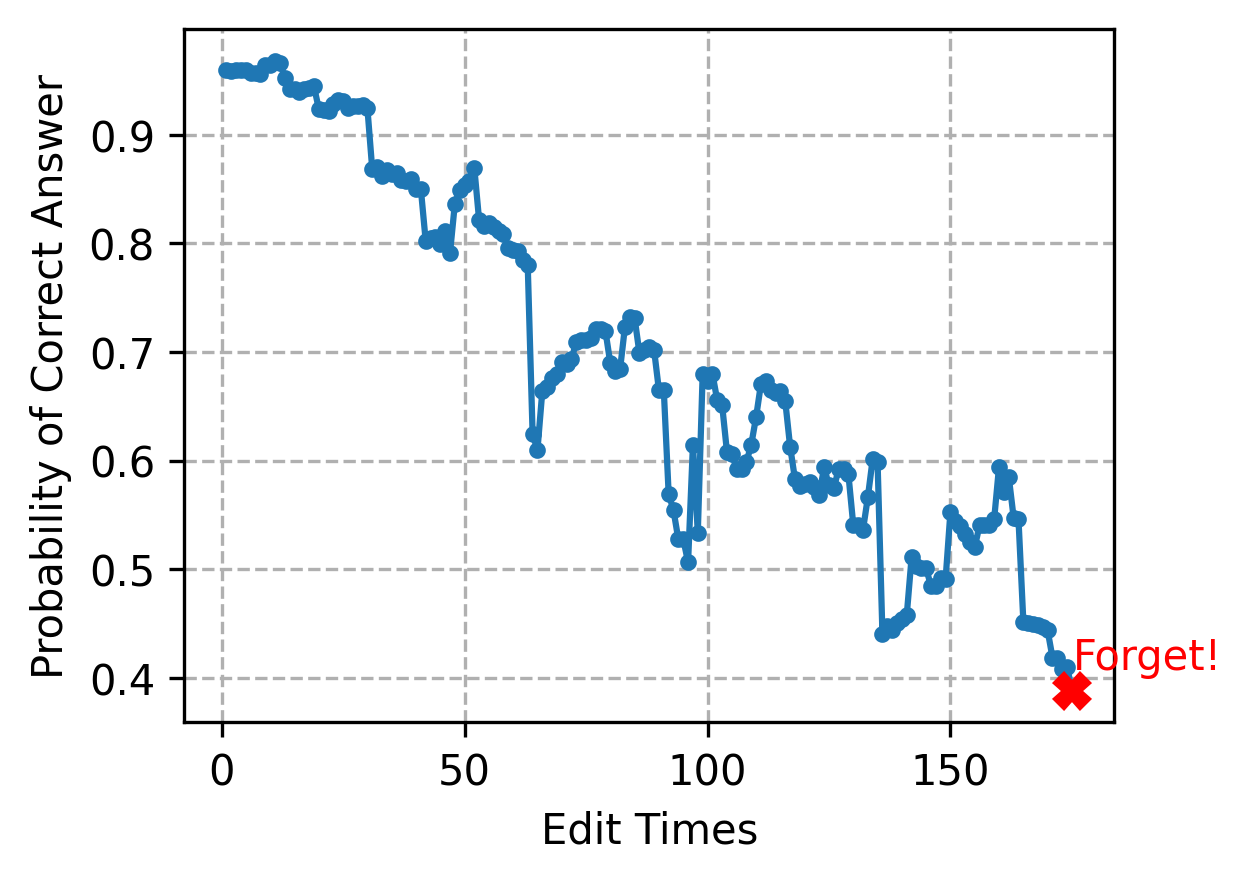}
    }
    \hfill
    \subfigure[Case 16]{\includegraphics[width=0.22\textwidth]{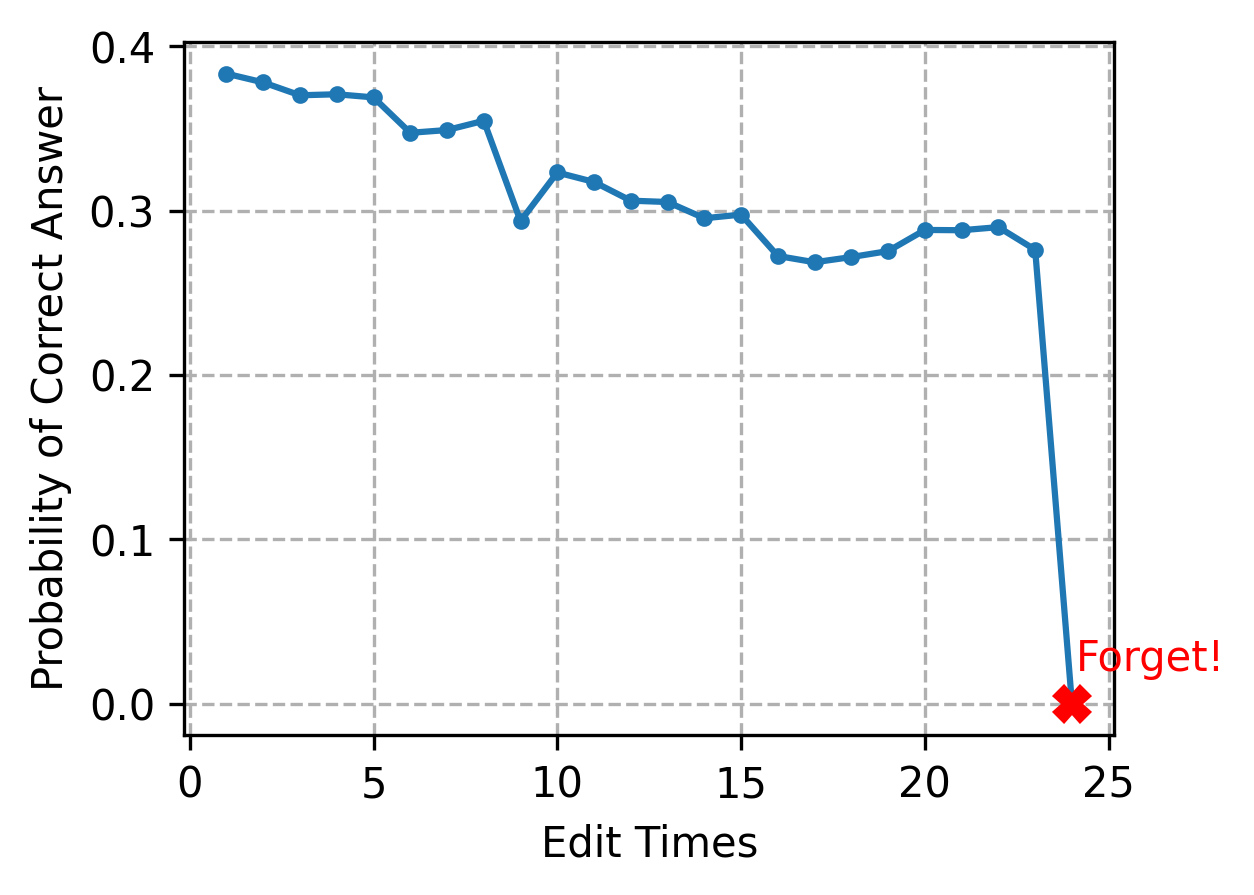}
    }
  \caption{In GPT-J, continuous editing causes irrelevant knowledge to be gradually forgotten.}
  \label{fig:knowledge_forgetting_gpt-j}
\end{figure*}

In our experiments, we use ROME to edit the GPT-J. The dataset is divided into editing data and test data, where the editing data is used to continually modify the model, and the test data is used to record the confidence of unrelated knowledge throughout this process. The editing process continues until the model forgets unrelated knowledge associated with the test data. The experimental results are shown in Figure~\ref{fig:knowledge_forgetting_gpt-j}.

From the results, it is evident that continual editing introduces persistent interference to unrelated knowledge, which progressively reduces the confidence of correct knowledge. As the editing process continues, this interference eventually results in the forgetting of unrelated knowledge.

\end{document}